\documentclass{article}

\usepackage{listings}
\usepackage[final]{neurips_2023}

\usepackage[utf8]{inputenc} % allow utf-8 input
\usepackage[T1]{fontenc}    % use 8-bit T1 fonts
\usepackage{hyperref}       % hyperlinks
\usepackage{url}            % simple URL typesetting
\usepackage{booktabs}       % professional-quality tables
\usepackage{amsfonts}       % blackboard math symbols
\usepackage{nicefrac}       % compact symbols for 1/2, etc.
\usepackage{microtype}      % microtypography
\usepackage{xcolor,soul}
\usepackage{amsmath}
\usepackage{bm}
\usepackage{mathtools}
\usepackage{caption}
\usepackage{algorithm}
\usepackage{algorithmic}
\usepackage{pifont} 
\usepackage{graphicx}
\usepackage{float}
\usepackage{array}
\usepackage{enumitem}

\usepackage[textsize=tiny]{todonotes}

\newcommand{\ourmethod}{SPRING}

\title{\ourmethod{}: Studying the Paper and Reasoning to Play Games}

\author{Yue Wu$^{14}$\thanks{Work done during internship at Microsoft. For correspondence, contact ywu5@andrew.cmu.edu},~ Shrimai Prabhumoye$^2$,~ So Yeon Min$^1$,~ Yonatan Bisk$^{1}$,~ Ruslan Salakhutdinov$^{1}$, \AND Amos Azaria$^{3}$,~ Tom Mitchell$^{1}$,~ Yuanzhi Li$^{1,4}$ \\
	$^1$Carnegie Mellon University, $^2$NVIDIA, $^3$Ariel University, $^4$Microsoft Research
}

\begin{document}

\maketitle

%% Important ** NeurIPS this year requires Abstract to be one paragraph ONLY
\begin{abstract}
% In this paper, we explore the application of large language models (LLMs) for tackling complex open-world survival games like Crafter, which pose significant challenges due to procedural generation, diverse action spaces, technology trees, and conflicting objectives. We propose \ourmethod, a novel prompting framework that leverages in-context chain-of-thought reasoning in LLMs to address the limitations of reinforcement learning (RL), such as high sample complexity and difficulty in incorporating prior knowledge. Our \ourmethod framework uses a directed acyclic graph (DAG) with questions as nodes and dependencies as edges, guiding LLMs through a sequence of problem-solving steps. We condition the framework on prior knowledge extracted directly from the original Crafter paper's LaTeX source code, allowing the LLM to incorporate domain-specific information. Our experiments demonstrate that GPT-4, under our prompting framework, achieves zero-shot performance surpassing state-of-the-art RL baselines trained for 1M steps. This work highlights the potential of LLMs for solving complex game-based problems and introduces a novel chain-of-thought prompting technique for effective problem-solving in diverse domains.

Open-world survival games pose significant challenges for AI algorithms due to their multi-tasking, deep exploration, and goal prioritization requirements. Despite reinforcement learning (RL) being popular for solving games, its high sample complexity limits its effectiveness in complex open-world games like Crafter or Minecraft.
% To incorporate prior knowledge, we first use a large language model (LLM) to obtain prior knowledge directly from the \LaTeX{} source code of the game's original academic paper. 
% We propose a novel pipeline to obtain and make use of prior knowledge directly from the \LaTeX{} source code of the game's original academic paper through the use of a large language model (LLM).
We propose a novel approach, \ourmethod{}, to read Crafter's original academic paper and use the knowledge learned to reason and play the game through a large language model (LLM).
% On the other hand, large language models (LLMs) have shown strong performance when prompted for a variety of tasks. We hypothesize that LLMs could perform reasonably well when su
%To enforce consistent planning and execution over hundreds of environment steps, we introduce \ourmethod{},  and a prompting framework for LLMs designed to enable in-context reasoning. 
Prompted with the \LaTeX{} source as game context and a description of the agent's current observation, our \ourmethod{} framework employs a directed acyclic graph (DAG) with game-related questions as nodes and dependencies as edges. We identify the optimal action to take in the environment by traversing the DAG and calculating LLM responses for each node in topological order, with the LLM's answer to final node directly translating to environment actions.
In our experiments, we study the quality of in-context ``reasoning'' induced by different forms of prompts under the setting of the Crafter environment. Our experiments suggest that LLMs, when prompted with consistent chain-of-thought, have great potential in completing sophisticated high-level trajectories. Quantitatively, \ourmethod{} with GPT-4 outperforms all state-of-the-art RL baselines, trained for 1M steps, without any training. 
Finally, we show the potential of Crafter as a test bed for LLMs. Code at \href{https://github.com/Holmeswww/SPRING}{github.com/holmeswww/SPRING}
% \shrimai{You could use similar wording in the abstract to describe \ourmethod{}: that it s a two staged approach and describe the two stages as being studying and reasoning.}
%To the best of our knowledge, this is the first work to address a challenging RL benchmark by reading an academic paper on the game in an end-to-end manner, and the first to demonstrate competitive performance by an LLM policy (GPT-4) in a challenging open world game without fine-tuning.
\end{abstract}

\section{Introduction} \label{text:introduction}
Open-world survival games like Minecraft~\citep{fan2022minedojo} and Crafter~\citep{crafter} pose significant challenges for AI algorithms due to a combination of factors: procedural generation requires strong generalization; diverse action space requires multi-task capabilities; technology tree requires long-term planning and deep exploration; diverse and conflicting objectives requires goal prioritization. In particular, %the 
Crafter %environment 
is designed for efficient simulation and fast iteration. Similar to Minecraft, Crafter features key challenges such as multi-tasking, exploration with a deep and wide tech-tree, requiring the agent to craft multiple tools and interact with multiple objects to survive in the game.

Reinforcement learning (RL) has been the go-to approach for game-based problems, with numerous successes in games like Go~\citep{silver2017mastering}, robotics~\citep{fu2020d4rl,hafner2023mastering} and various video games~\citep{vinyals2019grandmaster,muzero,badia2020agent57,hafner2023mastering}. While RL demonstrated impressive performance, it still suffers from certain limitations, such as high sample complexity and difficulty in incorporating prior knowledge. Such drawbacks make it exceptionally challenging to apply RL to diverse and complex open-world benchmarks like Crafter~\citep{crafter} or Minecraft~\citep{fan2022minedojo}. Addressing the benefits and drawbacks of RL is therefore crucial for achieving a sample-efficient solution. 

\begin{figure}[t]
    \centering
    \vspace{-4mm}
    \includegraphics[width=\textwidth]{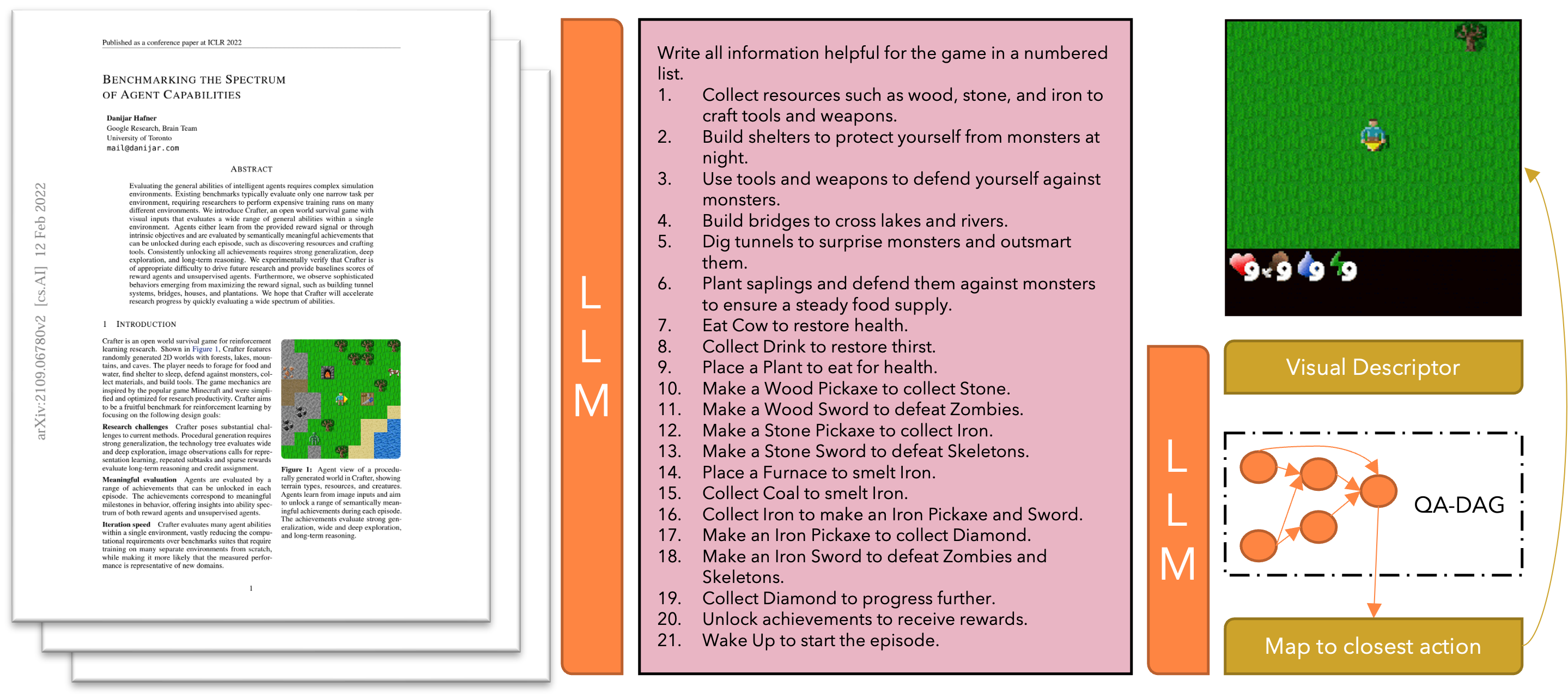} 
    \vspace{-7mm}
    \caption{Overview of \ourmethod{}. The context string, shown in the middle column, is obtained by parsing the \LaTeX{} source code of \citet{crafter}. The LLM-based agent then takes input from a visual game descriptor and the context string. The agent uses questions composed into a DAG for chain-of-thought reasoning, and the last node of the DAG is parsed into action.}
    \vspace{-7mm}
    \label{fig:teaser}
\end{figure}

On the other hand, large language models (LLMs)~\citep{brown2020language,megatron-530b,chowdhery2022palm} have shown remarkable success when prompted for various tasks, including embodied planning and acting~\citep{ahn2022can,du2023guiding,wang2023describe,shinn2023reflexion}, QA or dialogue~\citep{instruct,agi}, and general problem-solving~\citep{brown2020language,agi}. 
Their unique planning~\citep{ahn2022can}, reasoning~\citep{shinn2023reflexion}, and problem-solving~\citep{agi,madaan2023self} ability %under specific prompts 
makes them a promising candidate for incorporating prior knowledge and in-context reasoning for game-based problems, particularly when it comes to addressing the aforementioned limitations of RL.

Hence, in this work, we study the possibility and reliability of LLMs for understanding and reasoning with human knowledge, in the setting of games. We consider a two staged approach \ourmethod{} (Figure~\ref{fig:teaser}): (1) \textbf{studying the paper}: the first stage reads the \LaTeX{} %source code 
of the %original 
paper of~\citep{crafter} and (2) \textbf{reasoning}: the second stage involves reasoning about that knowledge through a QA framework to take an environment action.
Note that the Crafter environment was released after the data collection date of GPT-3.5 and GPT 4~\citep{openai2023gpt4} models\footnote{GPT-3.5/4 training data ends in September 2021 according to  \href{https://platform.openai.com/docs/models/gpt-4}{OpenAI API}}, the environment is unseen to them.
We first use LLM to extract prior knowledge from the \LaTeX{} source code of the original paper by~\citet{crafter}. We then use a similar QA summarization framework as~\citet{read_and_reward} which produces QA dialogue on game mechanics. \ourmethod{} handles significantly more diverse contextual information than~\citep{read_and_reward}, making use of all 17 action/interaction types and even information about desirable behaviors documented in the paper.

%As shown in Fig 1, 
We focus on reading the relevant academic paper in the first stage of \ourmethod{}, 
%We achieve this 
by first deciding the paragraphs that are relevant for playing the game.
Then we extract key information %from those relevant paragraphs 
through a series of questions such as ``\textit{Write all information helpful for the game in a numbered list.}".
In the second stage,% of \ourmethod{}
we promote and regulate in-context chain-of-thought reasoning in LLMs to solve complex games. %like Crafter~\citep{crafter}. %We've mentioned crafter a bunch so feels unnecessary
The reasoning module is a directed acyclic graph (DAG), with questions as nodes and dependencies as edges. 
For example, the question ``\textit{For each action, are the requirements met?}" depends on the question ``\textit{What are the top 5 actions?}", creating %and therefore, there is 
an edge from the latter to the former. 
For each environment step, we traverse the DAG computing LLM answers for each node in the topological order of the graph. The final node of the DAG is a question about the best action to take and the LLM answer for the question is directly translated to environment action.

Qualitatively, our experiments show that LLMs, when prompted with consistent chain-of-thought, can execute sophisticated trajectories independently in Crafter~\citep{crafter}.
Quantitatively, \ourmethod{}'s zero-shot performance with GPT-4 surpassing all state-of-the-art RL algorithms%~\citep{hafner2023mastering} 
trained for 1M steps (Table~\ref{table:compare}).
% \TODO{Shrimai: Add numbers here. surpassing SOTA by how much on what metric? Yue: Personally, I don't feel like emphasizing our metric performance that much since I think the goal is to provide a new method/way of thinking. Let's ask the @Professors about what they think.}

Our contributions are as follows:
\begin{itemize}[noitemsep,nosep]
    \item \ourmethod{} is the first to tackle a competitive RL benchmark by explicitly extracting multiple interactions and tech-tree dependencies directly from an academic paper.
    \item We are the first to show SOTA performance in a challenging open world game with a zero-shot LLM-based (GPT-4) policy 
    \item We study the quality of in-context “reasoning” induced by different prompts and propose a controlled chain-of-thought prompting through a DAG of questions for decision making.
\end{itemize}

\section{Method}
%\vspace{-1mm}
\label{sec:PET}

This section is structured as follows. We first describe how we generate the context from the \LaTeX{} source code of \citet{crafter} in Section \ref{text:reading_paper}. Then we describe our \ourmethod{} framework and how we compute the action in Section \ref{text:qa_dag}.

\paragraph{Problem Setting}
Our goal is to show that LLMs can plan and act reasonably well in an environment where control tasks are less required. In the setting of Crafter, we define the states, $s$, as samples from state distribution $S$. We are interested in creating a goal-conditioned policy $\pi$ which maps state $s$ to action $a\in A$, $\pi:S\to A$. Due to the use of LLM, we further break the policy down into two parts: a descriptor $\mathcal{D}$ which describes key aspects the visual observation in plain text ($d = \mathcal{D}(s)$). And an LLM-based actor which takes state description $d$ and outputs action $a$.

In addition, we define $\mathcal{S}_{\text{para}}^j$ to be the $j^{\text{th}}$ paragraph in the \LaTeX{} source of the environment paper \citep{crafter}, and $\mathcal{M}_{LLM}$ to be the LLM which takes a context string and a question string as input and outputs an answer to the question.

% \TODO{Give an overview of \ourmethod{} using Fig 1. You can also introduce some formalisms for Fig 1 components here.}

\subsection{Studying the paper: Context from \LaTeX{} source}
\label{text:reading_paper}
\begin{figure}[ht]
    \centering
    \vspace{-5mm}
    \includegraphics[width=\textwidth]{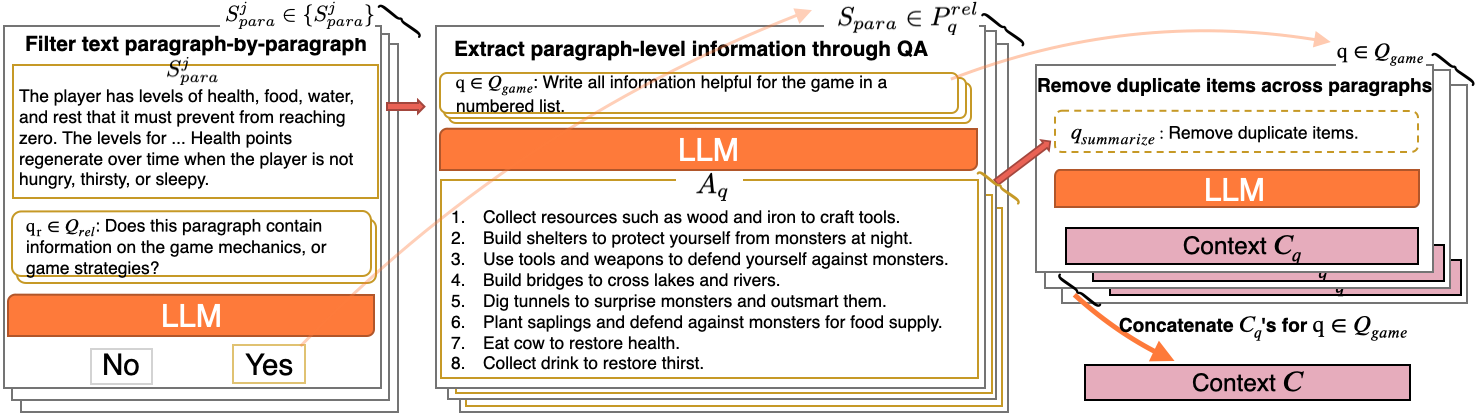} 
    \vspace{-5mm}
    \caption{ \textbf{Paper Studying Moudle.} The 3-step approach for computing $C_q$ from the \LaTeX source code of \citet{crafter}. First, as shown in the left column, for each paragraph we compute LLM answer for all relevancy questions in $Q_{\text{rel}}$, and keep only the relevant paragraphs. Second, as shown in the middle column, we compute paragraph-level LLM answer to $q$. Third, we summarize the answer into $C_q$ with a summary prompt; we concatenate $C_q$ across $q \in Q_{game}$ and obtain $C$.}
    \vspace{-2mm}
    \label{fig:paper_reading}
\end{figure}
Similar to \citet{read_and_reward}, we compose gameplay specific questions and then compute LLM answer to the questions for each subsection in the latex files. Since a considerable amount of the paper is irrelevant to the gameplay, we use a set of 2 questions {\scriptsize $Q_{\text{rel}}$=\{``Would this paragarph help me succeed in this game?", ``Does this paragraph contain information on the game mechanics, or game strategies?"\}} to identify relevance, and a set of 4 questions {\scriptsize $Q_{\text{game}}$=\{``Write all information helpful for the game in a numbered list.", ``In plain text. List all objects I need to interact/avoid to survive in the game. Use "I would like to X object Y" in each step. Replace Y by the actual object, X by the actual interaction.", ``Write all game objectives numbered list. For each objective, list its requirements.", ``Write all actions as a numbered list. For each action, list its requirements."\}} to summarize gameplay and action space relevant information. We add the prompt {\scriptsize ``DO NOT answer in LaTeX.''} to all of $Q_{\text{game}}$ to prevent the LLM from outputting the list in \LaTeX{} format. %Note that the last two questions of $Q_{\text{game}}$ are on 

For a specific gameplay specific question $q\in Q_{\text{game}}$, our goal is to compute $C_{q}$, the answer to $q$ conditioned on the paper. However, since the length of the paper exceeds input length constraints for most LLMs, we have to break the paper down into paragraphs individual $\mathcal{S}_{\text{para}}^j$. We provide an illustration of the process in Figure~\ref{fig:paper_reading}.

First, we filter the paragraphs for relevance and keep only paragraphs identified as relevant by at least one question from $Q_{\text{rel}}$. We set $P^{\text{rel}}_q$ to be the set of relevant paragraphs.

\begin{equation}
P^{\text{rel}}_q = \left\{\mathcal{S}_{\text{para}}^j| \exists q_r \in Q_{\text{rel}}\; s.t.\; \mathcal{M}_{LLM}\left(\mathcal{S}_{\text{para}}^j, q_r\right)=\text{``Yes''}\right\}
\end{equation}

Second, we compute the set, $A_{q}$, of answers to $q$ for each relevant paragraph from $P^{\text{rel}}_q$, from the \LaTeX{} source code.  %$\mathcal{M}_{LLM}(\mathcal{S}_{\text{paragraph}}^j, )$
\begin{equation}
A_q = \left\{\mathcal{M}_{LLM}\left(\mathcal{S}_{\text{para}}, q\right): \mathcal{S}_{\text{para}} \in P^{\text{rel}}_q\right\}
\end{equation}

Third, to obtain the answer string $C_q$ from the set $A_q$, we query an LLM with a summarization prompt $q_{\text{summarize}} = $ ``Remove duplicate items.''

\begin{equation}
C_q = \mathcal{M}_{LLM}\left(\mathtt{concat}(A_q), q_{\text{summarize}}\right)
\end{equation}

Finally, we concatenate (with the linebreak character) all question-context pairs to form the context string $C$ for \ourmethod{}.

\begin{equation}
C = \mathtt{concat}\left(\left\{\text{``Question: }q \text{ Answer: } C_q\text{''}|\forall q\in Q_{\text{game}}\right\}\right)
\end{equation}

\subsection{Reasoning: QA-DAG for \ourmethod{}}
\label{text:qa_dag}

\begin{table}[h]
{\small
\vspace{-2mm}
\begin{center}
\begin{tabular}{c m{39em}}
\toprule
\multicolumn{1}{c}{Node} & \multicolumn{1}{c}{Question}\\
\toprule
$q_1$ & List objects in the current observation. For each object, briefly   answer what resource it provides and its requirement.\\
\midrule
$q_2$ & What was the last action taken by the player?\\
\midrule
$q_3$ & For each object in the list, are the requirements met for the interaction?\\
\midrule
$q_4$ & Did the last player action succeed? If not, why?\\
\midrule
$q_5$ & List top 3 sub-tasks the player should follow. Indicate their priority out of 5.\\
\midrule
$q_6$ & What are the requirements for the top sub-task? What should the player do first?\\
\midrule
$q_7$ & List top 5 actions the player should take and the requirement for each action. Choose ONLY from the list of all actions. Indicate their priority out of 5.\\
\midrule
$q_8$ & For each action in the list, are the requirements met?\\
\midrule
$q_a$ & Choose the best executable action from above.\\
\bottomrule
\end{tabular}
\end{center}
\vspace{-1mm}
\caption{\label{table:questions} List of all 9 questions in $Q_{\text{act}}$. The questions are designed to promote consistent chain-of-thought. Experimentally, we find the LLM robust to different phrasing of the questions.}
\vspace{-4mm}}
\end{table}

\begin{figure}[h]
    \centering
    \vspace{-3mm}
    \includegraphics[width=\textwidth]{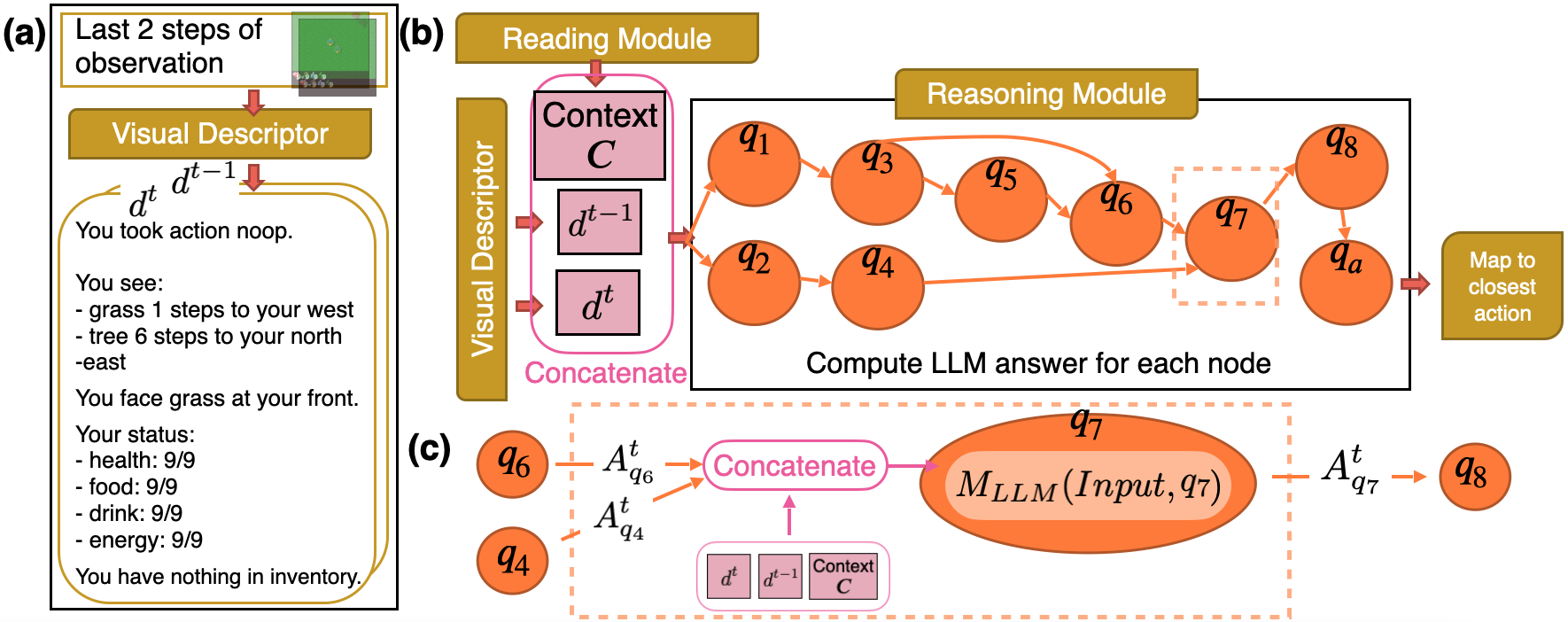} 
    \vspace{-6mm}
    \caption{\textbf{Reasoning.} \textbf{(a)} The visual descriptor takes the last two gameplay screens as input, and outputs their descriptions in language ($d^t, d^{t-1}$). \textbf{(b)} \ourmethod{} traverses a DAG of questions from Table~\ref{table:questions} in topological order. Answer to the final question $q_a$ is mapped to environment action using sub-string matching. \textbf{(c)} The LLM answer for each question (node) is conditioned on the previous 2 steps of observation, the context $C$, and answers to the immediate parents of the current node.}
    \vspace{-3mm}
    \label{fig:DAG}
\end{figure}
For LLMs to be able to understand the gameplay, we first follow \citet{du2023guiding,wang2023describe} to define an visual descriptor $\mathcal{M}_{\text{desc}}$ which converts state $s \in S$ to textual description $d$ (Figure~\ref{fig:DAG} a).

To achieve consistent chain-of-thought reasoning \citep{wei2021finetuned} throughout hundreds of steps within one round of gameplay, we compose a fixed set of questions $Q_{\text{act}}=\{q_1, \ldots, q_a\}$ to query the LLM at every step of the game, with question-question dependencies as $D = \{(q_u,q_v)|q_u,q_v\in Q_{\text{act}} \text{ and answering $q_v$ requires the answer of $q_u$}\}$. 
Note that the above specification forms a directed acyclic graph (DAG) with nodes $Q_{\text{act}}$ edges $D$ (Figure~\ref{fig:DAG} b). %\amos{I'm quite positive that reviewers will ask about the set of questions, how were they obtained? Why 9 questions? How was the DAG defined? (For example, why does the action $q_a$ depend only on the question about the requirements $q_8$, and doesn't depend on the top sub-tasks ($q_5$). How many other configurations were tested before reaching this exact set of questions and DAG? If other configurations were tested, did they fail? If so, why? If not, can we say anything about how susceptible the method is to the exact wording of the questions, their number, and the DAG dependencies? I know that there aren't excellent answers to all these questions, but we must have something. The same thing goes for Q_{rel} and Q_{game}, though it might be not as big an issue there.}

% To achieve consistent chain-of-thought reasoning \citep{wei2021finetuned} throughout hundreds of steps within one round of gameplay, we compose a DAG with the nodes being fixed set of questions $Q_{\text{act}}=\{q_1, \ldots, q_a\}$ to query the LLM at every step of the game. The question-question dependencies form a directed acyclic graph (DAG) with nodes $Q_{\text{act}}$ and edges $D = \{(q_u,q_v)|q_u,q_v\in Q_{\text{act}} \text{ and answering $q_v$ requires the answer of $q_u$}\}$. 

For any question (node) ${q_v}\in Q_{\text{act}}$, we compute the answer $A_{q_v}^t$ for time step $t$, conditioned on the gameplay context $C$, most recent 2 steps of game description $d^{t-1},d^{t}$, and answers to its dependencies (Figure~\ref{fig:DAG} c).
\begin{equation}
A_{q_v}^{t} = \mathcal{M}_{LLM}\left(\mathtt{concat}\left(C, d^{t-1}, d^t, \left\{A_{q_u}^t|(q_u,q_v)\in D\right\}\right), q_v\right)
\end{equation}
Experimentally, we find that prompting the LLM with only the direct parents of a question greatly reduces the context length, and helps LLM to focus on the most relevant contextual information.

We traverse the DAG using a modified topological sort algorithm to compute LLM answer for each question based on its topological order. Finally, we map the answer to the last question in the node $q_a$ directly to one of the 17 named actions in the environment with sub-string matching ($a = A^t_a$). We take the default action ``Do'' on sub-string matching failure.\footnote{We will release code for our agent at \href{https://github.com/anonymous}{github.com/anonymous}}

\section{Experiments and Results}
We present our experiments as follows. First, we explain our experimental setup and baselines for our experiments. Then, we compare \ourmethod{} to popular RL methods on the Crafter benchmark. Finally, we conduct experiments and analysis on different pieces of our architecture to study the influence of each part over the in-context ``reasoning'' capabilities of the LLM.
%First, we explain the environment setup and baselines for our experiments. Then we compare PET to the baselines on different splits of the environment. Finally, we conduct ablation studies and analyze the PET framework part by part. We show that PET generalizes better to human goal specification under efficient behavior cloning training. %\textcolor{red}{Especially, we show that PET is generalizable to human goals from in-simulation training only, unlike existing work. We don't use Dagger... }
% We are solving ... . ALFWorld is a simulator that enables agents to learn abstract, text-based policies in TextWorld \citep{textworld} and then execute goals from the ALFRED benchmark \citep{alfred} in a rich visual environment. \TODO{What do we do about this?}

\subsection{Experimental Details}

%\paragraph{Crafter Environment}
The Crafter environment~\citep{crafter} is a procedurally generated open-world survival game for benchmarking RL algorithms with 22 achievements in a tech tree of depth 7.
The environment is a grid-world features top-down observation and discrete action space of size 17. The observation also shows the current inventory state of the player, including its health points, food, water, rest levels, and inventory. The game is inspired by Minecraft and features a similar get-to-diamond challenge. In comparison, Crafter captures many key research challenges of Minecraft in a simpler and faster environment, thus speeding up experiments and result collection.

\paragraph{Environment Descriptor}
% \amos{I tried rewriting this paragraph; the original is commented out below. Please make sure it's correct. Also, it's now more than a single paragraph}
The gameplay screen (top left of Fig 3.) consists of a 9 $\times$ 9 grid ($\{ (i,j) \mid 1 \leq i,j \leq 9 \}$). The top 7 rows consist of the local view of the world; each cell $(i,j)$ is associated with a pre-defined background (e.g., ``grass'', ``water'', ``none'') and possibly with an object ``asset'' (e.g., ``tree'', ``health'', ``player''). The bottom 2 rows represent agent status (e.g., ``health'') and item inventories, which include images of assets (e.g., ``stone sword''), and the number of each assent in the inventory.

Our environment descriptor accepts as input the gameplay screen and outputs a text description of the screen. We first create combinations of background and object (appearance) assets. Then we add number assets to recognize the quantity of inventory/ status.
We match these combinations with the gameplay screen, using \texttt{cv2.filters} with a matching \textit{threshold} of $0.9$. 
We disable the detector during nights when observations are unreliable.
Finally, for each $(i,j)$, we filter the matched combinations, and select the one with the highest matching score. From this information, we can measure the distance and direction of each object relative to the player; simultaneously, we can count the agent status and inventory item.
% We extract object and inventory information from the gameplay screen. The gameplay screen (top left of Fig 3.) \amos{That's tiny!} consists of a 9 $\times$ 9 grid ($\{ (i,j) \mid 1 \leq i,j \leq 9 \}$); the top 7 rows consist of the local view of the world and the bottom 2 rows status and item inventories. Each grid $(i,j)$ \amos{each grid, or each cell?} is occupied by a combination of pre-defined background (e.g., ``grass'', ``water'', ``none'') and object ``assets'' (e.g., ``tree'', ``health'', ``player''). Additionally, number assets (e.g., ``9'') are added on top of inventory (e.g., ``stone sword'') and status (e.g., ``health'') objects \amos{This is in the bottom 2 rows, right? Please clarify.}. Thus, we create combinations of background and object assets (with an addition of number assets for inventory/ status objects), and simulate darkness and noise to be added to the combinations by counting the number of steps \amos{I don't understand this.}. We set a threshold (0.9) and match these combinations with the gameplay screen, using \texttt{cv2.filters} \amos{What is the purpose of the threshold?}. Finally, for each $(i,j)$, we filter the matched combinations, and select the one with the highest matching score. From this information, we can measure the distance and direction of each object relative to the player; simultaneously, we can count the agent status and inventory item. 

The environment descriptor then obtains the set of objects in observation $\mathcal{O} = \{(obj, dist, direction)\}$, the set of inventory items $\mathcal{I} = \{(object, count)\}$, and the agent status $\mathcal{H} = \{(attribute, value, max)\}$. Including only the closest object of each kind, we compose the observation description $d$ as: ``You see : - <obj> <dist> steps to your <direction>. Your status: <attribute>: <value>/ <max>. Your inventory: - <object>: <count>''. We describe direction of objects using ``north'',``south'',``east'',``west''.

\paragraph{Evaluation Metrics} Agents in Crafter are evaluated primarily based on two metrics: reward and score. The game assigns a sparse $+1$ reward each time the agent unlocks a new achievement in an episode, and assigns reward of $-0.1/0.1$ when the agent loses/gains one health point. The score metric~\citep{crafter} is computed by aggregating the success rates for each achievement:
\[S=\exp{\left(\frac{1}{N}\sum_{i=1}^{N}\ln{\left(1+s_i\right)}\right)}-1,\]
where $s_i$ is the agent's success rate on achievement $i$ and $N=22$ is the number of achievements. Note that RL agents only train on the reward, and \ourmethod{} does not require any training.

\paragraph{RL Baselines} We include results from popular actor-critic methods like PPO~\citep{schulman2017proximal}; DQN variants like Rainbow~\citep{hessel2018rainbow}; intrinsically motivated methods like RND~\citep{burda2018exploration}, Plan2Explore~\citep{sekar2020planning}, EDE~\citep{jiang2022uncertainty}; LLM assisted solutions like ELLM~\citet{du2023guiding}; model-based methods like DreamerV2~\citep{hafner2020mastering}; DreamerV3~\citep{hafner2023mastering}, which currently holds the state-of-the-art. 

\paragraph{LLMs.} For LLM access, we use GPT-3.5-turbo~\citep{gpt3-5} and GPT-4~\citep{openai2023gpt4} from OpenAI's \href{https://platform.openai.com/docs/models/overview}{API}. 

\begin{table}[t]
\vspace{-2mm}
\begin{center}
\begin{tabular}{l @{\hskip 0.50in} c c c}
\toprule
\multicolumn{1}{l}{Method} & \multicolumn{1}{c}{Score} & \multicolumn{1}{c}{Reward} & \multicolumn{1}{c}{Training Steps$^4$} \\
\toprule
Human Experts & $50.5\pm 6.8\%$ & $14.3\pm 2.3$ & N/A \\
\midrule
\ourmethod{} + paper (Ours) & $\bm{27.3\pm 1.2\%}$ & $\bm{12.3\pm 0.7}$ & \textbf{0}\\
DreamerV3~\citep{hafner2023mastering} & $14.5\pm 1.6\%$ & $\bm{11.7\pm 1.9}$ & 1M\\
ELLM~\citep{du2023guiding} & N/A & $\phantom{0}6.0\pm 0.4$ & 5M\\
EDE~\citep{jiang2022uncertainty} & $11.7\pm 1.0\%$ & N/A & 1M\\
DreamerV2~\citep{hafner2020mastering} & $10.0\pm 1.2\%$ & $\phantom{0}9.0\pm 1.7$ & 1M\\
PPO~\citep{schulman2017proximal} & $\phantom{0}4.6\pm 0.3\%$ & $\phantom{0}4.2\pm 1.2$ & 1M\\
Rainbow~\citep{hessel2018rainbow} & $\phantom{0}4.3\pm 0.2\%$ & $\phantom{0}5.0\pm 1.3$ & 1M\\
Plan2Explore~\citep{sekar2020planning} & $\phantom{0}2.1\pm 0.1\%$ & $\phantom{0}2.1\pm 1.5$ & 1M\\
RND~\citep{burda2018exploration} & $\phantom{0}2.0\pm 0.1\%$ & $\phantom{0}0.7\pm 1.3$ & 1M\\
Random & $\phantom{0}1.6\pm 0.0\%$ & $\phantom{0}2.1\pm 1.3$ & 0\\
\bottomrule
\end{tabular}
\end{center}
\vspace{-1mm}
\caption{\label{table:compare} Table comparing \ourmethod{} and popular RL algorithms in terms of game score, reward, and training steps. Results for \ourmethod{} is summarized over 5 independent trials. \ourmethod{} out-performs the previous SOTA in terms of all metrics. In addition, since \ourmethod{} gathers knowledge from reading the paper, it requires no training.}
\vspace{-8mm}
\end{table}

\subsection{Overall Results}
We compare the performance of RL baselines to \ourmethod{} with GPT-4 conditioned on the environment paper~\citep{crafter} in Table~\ref{table:compare}. 

\ourmethod{} out-performs the previous SOTA, including previous attempts at using LLMs for Crafter %\amos{We should discuss ELLM in further details somewhere in the paper. How are we different? Is it only because SPRING read the paper (instructions) and uses GPT-4? I think that it's more than that.}, Yue: We don't have more experiments to show that, but it's kinda implicit. ELLM doesn't seem so related to our work anyways, I didn't even cover it in related works section.
by large margins, achieving an $88\%$ relative improvement on game score and a $5\%$ improvement in reward on the best performing RL method~\citep{hafner2023mastering}. Since the model obtains knowledge from reading the paper, \ourmethod{} requires $0$ training steps, while RL methods generally require millions of training steps\footnote{We base our comparison on the hard 1M cap set by~\citet{crafter}.}.

\begin{figure}[ht]
    \centering
    \vspace{-2mm}
    \includegraphics[width=\textwidth]{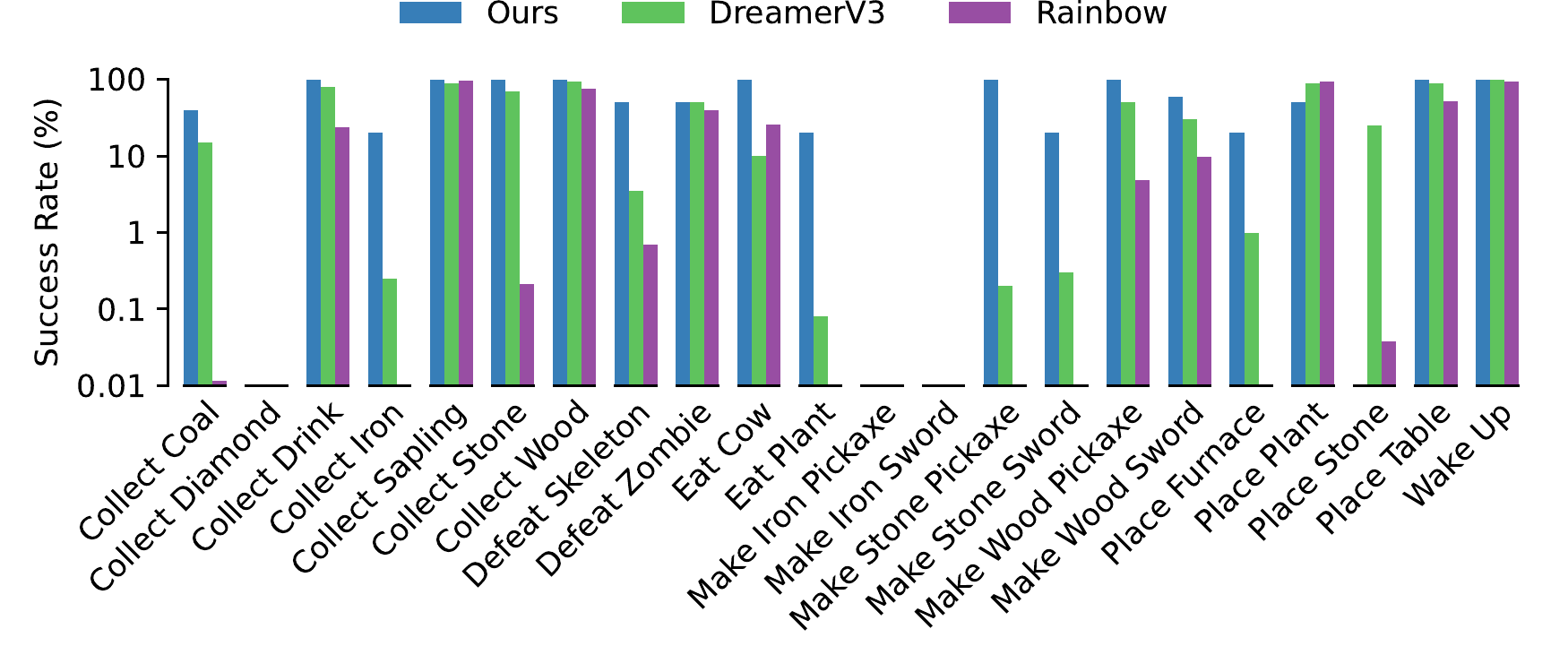} 
    \vspace{-8mm}
    \caption{Ability spectrum showing the unlocking percentages for all 22 achievements. Rainbow
manages to drink water and forage for food. DreamerV3 collects coal, iron, stone, and forges more advanced tools and weapons. Since \ourmethod{} starts off with knowledge about the game, it achieves more than 10x higher unlock rate on previously hard-to-reach tasks like ``Eat Plant'', ``Make Stone Pickaxe'', ``Make Stone Sword'', and ``Collect Iron''.
    }
    \vspace{-2mm}
    \label{fig:supervised_compare}
\end{figure}

We include a plot of unlock rate by task, comparing our method to popular RL baselines in Figure~\ref{fig:supervised_compare}. \ourmethod{} assisted by prior knowledge out-performs RL methods by more than 10x on achievements like ``Make Stone Pickaxe'', ``Make Stone Sword'', and ``Collect Iron'', which are up to depth 5 down in the tech tree and significantly harder to reach through random exploration.
% We also observe that \ourmethod{} achieves similar performance on ``Eat Cow'' and ``Collect Drink'', where model-based RL framework like Dreamer-V3 encounters a performance gap due to the fact that cows are moving and are harder to reach. 
For achievements ``Eat Cow'' and ``Collect Drink'', \ourmethod{} achieves perfect performance, whereas model-based RL framework like Dreamer-V3 has more than 5x lower unlock rate for ``eat cow'' since cows are moving and harder to reach through random exploration. 
Finally, we note that \ourmethod{} did not take the action ``Place Stone'', which can be reached easily by random exploration, since placing a stone was not discussed as beneficial for the agent in the paper~\citep{crafter}.

\begin{table}[t]
\vspace{-2mm}
\begin{center}
\begin{tabular}{l c c c}
\toprule
\multicolumn{1}{l}{Method} & \multicolumn{1}{c}{Achievement Depth} & \multicolumn{1}{c}{Reward} & \multicolumn{1}{c}{Questions per Step} \\
\toprule
\ourmethod{} + Full Paper & 6 & $\bm{12.3\pm 0.7}$ & 9\\
\ourmethod{} + Paper w/ modified $C$ & 4 & $\phantom{0}9.4\pm 1.8$ & 9\\
\ourmethod{} + Action Description & 4 & $\phantom{0}8.2\pm 0.2$ & 9\\
\ourmethod{} + w/o $C$ & 1 & $\phantom{0}0.5\pm 0.2$ & 9\\
\midrule
\ourmethod{} + Full Paper & 6 & $\bm{12.3\pm 0.7}$ & 9\\
Step-by-step prompt + Full Paper & 5 & ${\phantom{0}7.3\pm 4.4}$ & 2 \\
QA w/o DAG + Full Paper & 4 & $\phantom{0}4.3\pm 3.9$ & 9\\
w/o QA + Full Paper & 2 & $\phantom{0}2.4\pm 1.3$ & 1\\
\midrule
\ourmethod{} + Full Paper & 6 & $\bm{12.3\pm 0.7}$ & 9\\
\ourmethod{} + Full Paper w/ GPT-3.5 & 2 & $\phantom{0}3.3\pm 2.9$ & 9\\
\bottomrule
\end{tabular}
\end{center}
\caption{\label{table:analysis} Analysis on how different parts of \ourmethod{} contribute to its performance, comparing the max achievement depth in the tech tree, the reward, and the number of human-written questions in the prompt. Results are summarized over 5 independent trials. The first 4 rows study the necessity of prior knowledge from the context string $C$. The middle 4 rows study different chain-of-thought prompting techniques. The last 2 rows study the role of LLMs. All three aspects are important for \ourmethod{} to achieve best reported performance.}
\vspace{-8mm}
\end{table}

\subsection{Component Analysis}
We study how the different aspects of the framework contribute to the behavior of the agent through a series of ablations as shown in Table \ref{table:analysis}.

\paragraph{Studying the \LaTeX{} Paper} In the first 4 rows of Table~\ref{table:analysis}, we investigate the contribution of gameplay context from the \LaTeX{} paper toward performance of the agent. We report the performance of \ourmethod{} with no contextual information (w/o $C$) (row 4); \ourmethod{} conditioned on only the action descriptions and dependencies from \citep{crafter} Table F.1 (only question 4 from $Q_{\text{game}}$) (row 3); \ourmethod{} conditioned on the context manually modified to exclude the ``crafting table'' dependency for wooden\_pickaxe by removing two corresponding lines from the context $C$ (row 2); \ourmethod{} conditioned on the full context from the paper (row 1).

As expected, since Crafter environment is unseen for GPT, the agent achieves performance similar to random agent without any game context. When provided with only action descriptions and action dependencies, using only question 4 from $Q_{\text{game}}$ in section~\ref{text:reading_paper}, \ourmethod{} achieves strong $67\%$ performance comparable to DreamerV2~\citep{silver2017mastering}.

For the next piece of the experiment, we manually remove ``near crafting table'' dependency for wooden\_pickaxe from it's context, which is required for 11 later achievements. \ourmethod{} with GPT-4 incurs a $24\%$ performance drop. Interestingly, we find that the LLM has some ability to recover from the inaccurate context information. We observe that after failing to craft the wooden\_pickaxe without a table, the agent instead tries to craft a wooden\_sword first to maintain survival. Eventually, the agent was able to identify the missing requirement through guessing and trying after some unsuccessful trials, and craft the wooden\_pickaxe. However, the confusion delayed the agent's progress and therefore causes the performance gap with the agent conditioned on the full context (row 5).

\paragraph{Reasoning}
\label{paragraph:step-by-step}
In the middle 4 rows of Table~\ref{table:analysis}, we investigate the contribution of different prompting methods toward performance of the model. Conditioned on the full context from the \LaTeX{} paper, we report the performance of GPT-4 directly prompted to output the action using the last question $q_a$ only (row 8); GPT-4 prompted with all questions from $Q_{\text{act}}$ but in a list without the DAG dependencies $D$ (row 7); GPT-4 prompted ``Let's think step-by-step''~\citep{kojima2022large} about the next action, and prompted to choose a permissible action $q_a$ with let's think step-by-step followed by $q_a$ again (row 6); GPT-4 with \ourmethod{} (row 5).

Relative to our method, we observe that directly prompting the LLM for the action leads to a $80\%$ performance drop, and therefore does not result in a meaningful agent. The popular chain-of-thought reasoning prompt ``Let's think step-by-step'' \citep{kojima2022large} achieves reasonable reward with a $40\%$ drop, but with a high $60.27\%$ standard deviation. Qualitatively, we observe that the LLM produces inconsistent outputs across time steps, due to the fact that the model's chain-of-thought is not directed or controlled through the prompt. Therefore, LLMs prompted with ``Let's think step-by-step'' alone cannot reliably follow a good policy. 
Controlling the chain-of-thought with 9 questions from $Q_{\text{act}}$ (section~\ref{text:qa_dag}) successfully controls the consistency of LLM outputs across time qualitatively.
However, we observe that the LLM often ignores earlier questions at later stages of QA when all previous questions are presented in a list, leading to random disagreements in answers. For example, the LLM may correctly identify that it needs ``wooden pickaxe'' to mine the stone ahead in the first few questions, but forgets about the requirement later when it's prompted for actions.
Quantitatively, the model performs $65\%$ worse with $90\%$ variance without the DAG. The introduction of DAG eliminates this problem by reducing the QA context length to only a question's immediate parents.

Overall, \ourmethod{} achieves the best performance and a small $6\%$ performance standard deviation, due to more consistent reasoning over time steps with better focus and fewer distractions.

\paragraph{LLM} In the last two rows of Table~\ref{table:analysis}, we show that the same architecture does not work well with GPT-3.5-turbo. We believe the observed $73\%$ performance gap mainly comes from GPT-3.5-turbo's worse performance at following fine-grained instructions in each of the questions, which are required for chain-of-thought reasoning with \ourmethod{}.

\subsection{Cost for running SPRING}
The number of queries per step is 9 (same as the number of questions). Each game could take around 300 steps, but can go up to 500 steps in the worst case. Therefore, the maximum number of queries per game can go up to 4500. According to the public price of GPT-4 API, each query costs around 0.06\footnote{Pricing estimated according to \href{https://openai.com/pricing}{OpenAI API} as of October 17 2023}. The total cost should be less than 270 (USD) per game with GPT-4. Given the progress in chip-set development, we are hopeful that the inference costs will lower, making LLMs more accessible for the public.

\subsection{Potential for Benchmarking LLMs}
In Table~\ref{table:benchmark}, we compare popular publicly available LLMs including GPT-4~\citep{openai2023gpt4}, GPT-3.5 (text-davinci-003)~\citep{gpt3-5}, Bard~\citep{bard-intro}, Claude~\citep{claude-intro}, Alpaca-30b~\citep{alpaca} under the same setting on Crafter, following the same step-by-step prompt as Section~\ref{paragraph:step-by-step} and Table~\ref{table:analysis}. We observe a clear separation in performance under our setting.
\begin{table}[t]
\vspace{-2mm}
\begin{center}
\begin{tabular}{l c c c}
\toprule
\multicolumn{1}{l}{Method} & \multicolumn{1}{c}{Achievement Depth} & \multicolumn{1}{c}{Reward} & \multicolumn{1}{c}{Questions per Step} \\
\toprule
Step-by-step prompt + GPT-4 & 5 & $\bm{7.3 \pm 4.4}$ & 2\\
Step-by-step prompt + text-davinci-003 & 4 & $4.5 \pm 2.1$ & 2\\
Step-by-step prompt + Bard & 0 & $-0.9 \pm 0$ & 2\\
Step-by-step prompt + Claude & 1 & $0.1\pm 0.1$ & 2\\
Step-by-step prompt + Alpaca-30b & 1 & $0.1\pm 0.1$ & 2\\
Random & 1 & $\phantom{0}2.1\pm 1.3$ & 0\\
\bottomrule
\end{tabular}
\end{center}
\caption{\label{table:benchmark} Comparison of different LLMs under the same setting using the context $C$ generated with text-davinci-003 following the same step-by-step prompt as Section~\ref{paragraph:step-by-step} and Table~\ref{table:analysis}.}
\vspace{-8mm}
\end{table}

\section{Related Work} \label{text:related_works}

\paragraph{RL v.s. LLMs}
Comparing LLM-based agents against RL agents brings forth a intriguing discussion. RL algorithms do not require prior knowledge such as instruction manuals, and could continually improve given enough trials. However, RL algorithms are typically trained with reward functions deliberately engineered to cover all in-game achievements~\citep{crafter,hafner2023mastering}. Such reward functions often require a lot of expert knowledge and careful formulation. 

On the other hand, LLM agents like SPRING does not need the reward (we report reward for comparison purpose, SPRING does not use the reward during inference), but instead uses external knowledge from the \LaTeX{} source code. In addition, current LLM agents lack the capabilities of improving from interactions. 

We hope future works would be able to leverage the benefits of both paradigms in order to achieve efficient planning with fine-grained control.

\paragraph{Policy Informed by Natural Language Instructions}
In the instruction following setting, step-by-step instructions have been used to generate auxiliary rewards, when environment rewards are sparse. \citet{goyal2019using,wang2019reinforced} use auxiliary reward-learning modules trained offline to predict whether trajectory segments correspond to natural language annotations of expert trajectories.

There has been many attempts to go beyond instruction following to learning from unstructured natural language~\citep{branavan2012learning,goldwasser2014learning,zhong2021silg,wang2021grounding}. \citet{zhong2021silg,wang2021grounding} make use of special architectures to learn reasoning on grid worlds with template-generated instructions.
However, the model requires 200 million training samples from templates identical to the test environments. Such a training requirement limiting the generalization of the model and causes performance loss even on slightly bigger grid worlds with identical mechanics.

\citet{read_and_reward} proposes a summary (Read) and reasoning (Reward) through a QA prompting framework with an open-source QA LLM \citep{macaw}. The framework demonstrates the possibility of an using real-world human-written manuals to improve RL performance on popular games, despite limiting the interaction types to only ``hit''. Our framework handles all 17 kinds of interactions available in the game. Moreover, our framework makes use of information on tech-tree dependencies, and suggestions on desired policies extracted from the academic paper.

\paragraph{LLMs for Planning}
LLMs have shown promising results at high-level planning in indoor embodied manipulation environments. \citet{deepak,saycan} primarily explores generating plans for embodied tasks, with limited actions space and trajectory length. \citet{song2022llm,wutackling} enhances \citet{saycan} with greater action diversity and real-time re-planning. However, a lot of the high-level plans lack executability and has to be post-processed to meet specific task requirements, thus limiting the generalization to complex open world tasks. In addition, all prior works along this line operates on few-shot human/expert generated demonstrations containing up to 17 trajectories to provide context for LLMs, which requires more manual labor, and may limit the generalization to unseen scenarios. In comparison, our \ourmethod{} framework requires no demonstration.

\paragraph{LLMs for Open World Games}
Compared to popular indoor manipulation tasks, planning in open-world game environments poses the following additional challenges. 1) \textbf{Long horizon.} Due to the nature how in-game achievement/technology progresses, a successful gameplay can easily go beyond 200 steps \citep{crafter}. 2) \textbf{Parallel objectives.} Open-world environments contain objectives that can be pursued in parallel and often require prioritization \citep{wang2023describe}. Therefore, open world games are significantly more challenging than current indoor embodied manipulation environments. 

\citet{du2023guiding} applies LLMs as high-level planners to assist RL exploration in Crafter. \citet{wang2023describe,yuan2023plan4mc} use LLMs as high-level planner and goal selector to control a low level-policy in Minecraft. \citet{tsai2023can} studies the capabilities of ChatGPT on text games. Notably, all prior works require expert or human generated example trajectories as context for the LLMs. Since the example trajectories do not cover all scenarios, all prior works may encounter unseen situation during evaluation, leading to an overall performance inferior to state-of-the-art RL algorithms~\citep{hessel2018rainbow,guss2021minerl,hafner2023mastering}, trained without the use of LLMs. To our knowledge, we are the first to show an LLM (GPT-4) achieving performance surpassing the state-of-the-art RL algorithms in a challenging open world game.

% \paragraph{Crafter v.s. Minecraft}
% \TODO{Fill this in}

\section{Limitations and Future Work}
A primary limitation in using an LLM to support interaction with the environment is the need for object recognition and grounding. However, these limitations do not exist in environments that offer accurate object information, such as contemporary games \citep{fan2022minedojo} and virtual reality worlds \citep{ai2thor}. While pre-trained visual backbones \citep{he2017mask} perform poorly on games, they have shown reasonable performance for environments closer to the real-world~\citep{alfworld}. In addition, with recent progress on visual-language models \citep{agi,driess2023palm,liu2023grounding,zou2023segment}, we believe there will be reliable and generalizable solutions to visual-language understanding in the foreseeable future. Future works could focus on address the requirement for a separate visual descriptor with large visual-language models.

\section{Conclusions}
In this work, we explore solving the Crafter~\citep{crafter} RL benchmark using the latest LLMs by reading the \LaTeX{} source code of an academic paper about the benchmark. We study the quality of in-context ``reasoning'' and ``planning'' induced by different forms of prompts under the setting of the Crafter open-world environment.
To enforce consistent planning and execution over hundreds of environment steps, we introduce \ourmethod{}, an innovative prompting framework for LLMs designed to enable in-context chain-of-thought planning and reasoning. 
Quantitatively, \ourmethod{} with GPT-4 outperforms all state-of-the-art RL baselines, trained for 1M steps, without any training. 

Our work demonstrates the reliability of LLMs for understanding and reasoning with human knowledge. We hope that our work points to a new way of integrating human prior knowledge into RL training through intrinsic rewards~\citep{read_and_reward}, hierarchical RL~\citep{shu2017hierarchical}, or sub-goal planning~\citep{wang2023describe,PET}.

\section*{Broader Impacts}
Our research on LLM holds potential for both positive and negative impacts. The benefits include better understanding of the powers of LLM and enhanced integration of prior knowledge, which could lead to advancement in various AI topics. However, the risks may involve reliance on computationally demanding models, game cheating or exploitation, and reliance on prior knowledge.

\bibliography{citations}
%%%%%%%%%%%%%%%%%%%%%%%%%%%%%%%%%%%%%%%%%%%%%%%%%%%%%%%%%%%%

\newpage
\appendix
\section{Example Trajectory}
% \subsection{Trajectory 1}
\begin{lstlisting}
============Step: 0, Cumulative Reward: 0.0============
You took action make_wood_sword.

You see:
- tree 5 steps to your north-east
- cow 6 steps to your north-west

You face grass at your front (south).

Your status:
- health: 9/9
- food: 9/9
- drink: 9/9
- energy: 9/9

You have nothing in your inventory.
------------ QA ------------
Question: Action
Answer:
Move North

Question: List objects in the current observation. For each object, breifly answer what resource it provides and its requirement.
Answer:
1. Tree: Provides wood which is used to craft tools and structures like the table. Requirement: An axe or hands to chop down the tree.
2. Cow: Provides food when killed, restoring your health points. Requirement: A weapon like a sword to kill the cow or hands to punch it.
3. Grass: Generally provides no resources, but it can occasionally drop seeds for planting. No requirements to interact with grass.

Question: What was the last action taken by the player?
Answer:
The last action taken by the player was "make_wood_sword."

Question: For each object in the list, are the requirements met for interaction?
Answer:
1. Tree: Yes, the requirements are met as you can use your hands to chop down the tree, even though it will be slower than using an axe.
2. Cow: Yes, the requirements are met as you have crafted a wood sword to kill the cow, which is more efficient than using your hands.
3. Grass: Yes, there are no specific requirements to interact with grass, and you can simply walk on it or punch it with your hands if necessary.

Question: List top 3 sub-tasks the player should follow. Indicate their priority out of 5.
Answer:
1. Chop down the tree to collect wood (Priority: 5/5) - Wood is essential for crafting more tools and structures, as well as potentially making a shelter.

2. Kill the cow for food (Priority: 4/5) - Ensuring a sufficient food supply is crucial for maintaining health and being able to accomplish other tasks.

3. Explore the environment (Priority: 3/5) - It is essential to gain more information about available resources and potential dangers in the surrounding area so that you can plan effectively.

Priority level is indicated out of 5, where 5 is the highest priority and 1 is the lowest priority.

Question: What are the requirements for the top sub-task? What should the player do first?
Answer:
Requirements for the top sub-task (Chop down the tree to collect wood):

1. Approach the tree: Navigate to the tree located 5 steps to your north-east.

What the player should do first:

1. Move North-East: Move in the direction of the tree to eventually reach it and then prepare to chop it down.

Question: List top 5 actions the player should take and the requirement for each action. Choose ONLY from the list of all actions. Indicate their priority out of 5.
Answer:
Priority 1: Move North - Requirement: Flat ground above the agent.
Priority 2: Move East - Requirement: Flat ground right to the agent.
Priority 3: Move East - Requirement: Flat ground right to the agent.
Priority 4: Do (Chop down the tree) - Requirement: Facing the tree; have necessary tool (wooden sword can be used, but a wooden pickaxe would be preferred).
Priority 5: Collect Wood - Requirement: Successfully chopped the tree, and wood is available to pick up.

Question: For each action in the list, are the requirements met?
Answer:
Priority 1: Move North - Yes, the requirement is met as flat ground is above the agent.
Priority 2: Move East - Yes, the requirement is met as flat ground is available to the right of the agent.
Priority 3: Move East - Yes, the requirement is met as flat ground is available to the right of the agent.
Priority 4: Do (Chop down the tree) - The requirement is partially met; you are facing the tree and have a wooden sword, but a wooden pickaxe is preferred for chopping down trees.
Priority 5: Collect Wood - No, the requirement is not met yet, as you have not successfully chopped down the tree or accessed the wood.

Question: Did the last player action succeed? Why?
Answer:
I cannot determine if the last player action succeeded or not because there is no information provided about the result of the action "make_wood_sword." Normally, if the action was successful, the player would have a wood sword in their inventory or an updated game state would include the new item.

Question: Choose the best executable action from above.
Answer:
The best executable action from the above list is Priority 1: Move North, as the requirement of having flat ground above the agent is met.



============Step: 1, Cumulative Reward: 0.0============
You took action move_north.

You see:
- tree 4 steps to your east
- cow 6 steps to your north-west

You face grass at your front (north).

Your status:
- health: 9/9
- food: 9/9
- drink: 9/9
- energy: 9/9

You have nothing in your inventory.
------------ QA ------------
Question: Action
Answer:
Move East

Question: List objects in the current observation. For each object, breifly answer what resource it provides and its requirement.
Answer:
1. Tree: Provides wood, requires a wood or better pickaxe to collect.
2. Cow: Provides meat (food), requires a sword or other weapon to defeat and collect the resource.
3. Grass: No direct resources, can be used to build a shelter or pathway with no specific tool requirements.

Question: What was the last action taken by the player?
Answer:
The last action taken by the player was move_north.

Question: For each object in the list, are the requirements met for interaction?
Answer:
1. Tree: Yes, your recent action was making a wood sword so you can interact with the tree to collect wood.
2. Cow: Yes, you have a wood sword to defeat the cow, so you can interact with it.
3. Grass: Yes, there are no specific tool requirements to interact with the grass.

Question: List top 3 sub-tasks the player should follow. Indicate their priority out of 5.
Answer:
1. Collect wood from the tree. Priority: 5/5
2. Defeat the cow and collect meat (food). Priority: 4/5
3. Interact with grass to clear a path or build a shelter. Priority: 3/5

Question: What are the requirements for the top sub-task? What should the player do first?
Answer:
The requirements for the top sub-task (collect wood from the tree) are to have a wood pickaxe in the inventory and be close enough to the tree.

The player should first move closer to the tree located 4 steps to the east. Once the player is nearby, they should use their wood pickaxe to collect wood from the tree.

Question: List top 5 actions the player should take and the requirement for each action. Choose ONLY from the list of all actions. Indicate their priority out of 5.
Answer:
1. Priority 1: Move East (Requirement: Flat ground to the right of the agent) - to get closer to the tree.
2. Priority 2: Move North (Requirement: Flat ground above the agent) - to be in front of the tree.
3. Priority 3: Make Wood Pickaxe (Requirement: Nearby table; wood in inventory) - to be able to collect wood from the tree.
4. Priority 4: Do (Requirement: Facing tree or material; have a wood pickaxe) - to collect wood from the tree.
5. Priority 5: Make Wood Sword (Requirement: Nearby table; wood in inventory) - to defend yourself against potential threats (such as monsters).

Question: For each action in the list, are the requirements met?
Answer:
1. Priority 1: Move East (Flat ground to the right of the agent) - Yes, the requirement is met as the agent is on flat ground.
2. Priority 2: Move North (Flat ground above the agent) - Yes, the requirement is met as the agent is on flat ground.
3. Priority 3: Make Wood Pickaxe (Nearby table; wood in inventory) - No, the requirement is not met as the agent does not have a nearby table and wood in their inventory.
4. Priority 4: Do (Facing tree or material; have a wood pickaxe) - No, the requirement is not met as the agent does not have a wood pickaxe in their inventory.
5. Priority 5: Make Wood Sword (Nearby table; wood in inventory) - No, the requirement is not met as the agent does not have a nearby table and wood in their inventory.

Question: Did the last player action succeed? Why?
Answer:
Yes, the last player action (move_north) succeeded because there was flat ground available to the north of the player's position, allowing them to move in that direction.

Question: Choose the best executable action from above.
Answer:
The best executable action from the list above is Priority 1: Move East, as the requirement of flat ground to the right of the agent is met.



============Step: 2, Cumulative Reward: 0.0============
You took action move_east.

You see:
- tree 3 steps to your east
- cow 4 steps to your east

You face grass at your front (east).

Your status:
- health: 9/9
- food: 9/9
- drink: 9/9
- energy: 9/9

You have nothing in your inventory.
------------ QA ------------
Question: Action
Answer:
Move East

Question: List objects in the current observation. For each object, breifly answer what resource it provides and its requirement.
Answer:
1. Tree: Provides wood when chopped down; requires having a wood pickaxe or better to chop down.
2. Cow: Provides food (meat) when killed; requires having a wood sword or better to kill.
3. Grass: Can be used to make plantations or used for aesthetic purposes; requires being dug to collect.

Question: What was the last action taken by the player?
Answer:
The last action taken by the player was move_east.

Question: For each object in the list, are the requirements met for interaction?
Answer:
Based on the current observation and inventory:

1. Tree: The requirement is not met because you need a wood pickaxe or better to chop down the tree, and you have nothing in your inventory.
2. Cow: The requirement is not met because you need a wood sword or better to kill the cow, and you have nothing in your inventory.
3. Grass: The requirement is met as you can dig the grass without any special tool.

Question: List top 3 sub-tasks the player should follow. Indicate their priority out of 5.
Answer:
1. Collect Wood (Priority: 5/5) - Collecting wood is essential to craft tools and build structures. First, you need to craft a wood pickaxe to chop down trees and obtain wood. However, you currently don't have the required tools to interact with the tree.

2. Craft a Wood Pickaxe and Wood Sword (Priority: 4/5) - Once you collect wood, you should craft a wood pickaxe to enable collecting stone and a wood sword to defend yourself and hunt for food. You need a table to craft these items, so you might need to find or place one.

3. Collect Food and Water (Priority: 3/5) - Maintaining your food and drink levels is crucial for survival. You have observed a cow nearby, but you currently don't have the required tools to kill it. You should prioritize crafting tools, and then hunt the cow for food. For water, you need to find a water source like a lake or river to drink from.

Note: The priority scale is from 1 to 5, with 1 being the lowest priority and 5 being the highest priority.

Question: What are the requirements for the top sub-task? What should the player do first?
Answer:
Top Sub-task: Collect Wood

Requirements:
1. Facing the tree
2. Having the necessary tool (wood pickaxe) to chop down the tree and collect wood

What should the player do first:
Before collecting wood, the player needs to craft a wood pickaxe. To craft a wood pickaxe, the player requires a table and wood in their inventory. Since the player currently doesn't have any tools or a table, they should first find or place a table, then collect wood using their hands or any available tool, and use the table to craft a wood pickaxe. After crafting the wood pickaxe, the player can chop down the tree and collect wood more efficiently, which can be used for crafting other tools and structures.

Question: List top 5 actions the player should take and the requirement for each action. Choose ONLY from the list of all actions. Indicate their priority out of 5.
Answer:
1. Priority 1: Move East (Requirement: Flat ground right to the agent)
The player should move east towards the tree to collect wood, which will help them in crafting a table and the necessary tools.

2. Priority 2: Do (Chop tree) (Requirement: Facing tree; have necessary tool or bare hands)
Once the player is near the tree, they should chop down the tree using their hands or any available tool to collect wood.

3. Priority 3: Place Table (Requirement: Wood in inventory)
After collecting wood, the player should place a table using the wood in their inventory. This table will help the player create the required tools.

4. Priority 4: Make Wood Pickaxe (Requirement: Nearby table; wood in inventory)
Using the table, the player should craft a wood pickaxe, which will help them collect resources more efficiently.

5. Priority 5: Collect Wood using Wood Pickaxe (Requirement: Facing the tree, having the wood pickaxe)
With the wood pickaxe, the player can chop down trees and collect wood faster, allowing them to craft more tools and structures for survival.

Question: For each action in the list, are the requirements met?
Answer:
1. Move East: Yes, the requirement is met. The player's most recent observation states they're facing grass towards the east, indicating there's a flat ground on the east side.

2. Do (Chop tree): No, the requirement is not met yet. The player needs to move closer to the tree (3 steps to the east) and face it to meet the requirement for chopping down the tree.

3. Place Table: No, the requirement is not met. The player needs to collect wood by chopping the tree first.

4. Make Wood Pickaxe: No, the requirement is not met. The player needs to collect wood and place a table to craft a wood pickaxe.

5. Collect Wood using Wood Pickaxe: No, the requirement is not met. The player needs to create a wood pickaxe first and then face a tree to collect wood with the pickaxe.

Question: Did the last player action succeed? Why?
Answer:
Yes, the last player action (move_east) succeeded because there was flat ground (grass) to the east, allowing the player to move in that direction.

Question: Choose the best executable action from above.
Answer:
The best executable action from the list above is "Move East" as its requirement (flat ground right to the agent) is met according to the player's most recent observation.



============Step: 3, Cumulative Reward: 0.0============
You took action move_east.

You see:
- stone 7 steps to your south-east
- tree 2 steps to your east
- cow 3 steps to your east

You face grass at your front (east).

Your status:
- health: 9/9
- food: 9/9
- drink: 9/9
- energy: 9/9

You have nothing in your inventory.
------------ QA ------------
Question: Action
Answer:
Move East

Question: List objects in the current observation. For each object, breifly answer what resource it provides and its requirement.
Answer:
1. Tree: Provides wood which can be used to craft tools and weapons; Requirements: None, just approach the tree and interact with it.
2. Cow: Provides meat which can be consumed for food; Requirements: None, just approach the cow and interact with it.
3. Stone: Provides stone which can be used to craft tools, weapons, and build structures; Requirements: Requires a wood pickaxe or better for efficient collection, just approach the stone and interact with it.

Question: What was the last action taken by the player?
Answer:
The last action taken by the player was "move_east."

Question: For each object in the list, are the requirements met for interaction?
Answer:
1. Tree: Yes, the requirements are met for interaction as there are no specific tools needed to collect wood from the tree. You can approach it and interact with it directly.

2. Cow: Yes, the requirements are met for interaction as there are no specific tools needed to obtain meat from the cow. However, having a weapon like a wood sword would be helpful for efficient collection. You can approach it and interact with it directly.

3. Stone: No, the requirements are not met for interaction as you need a wood pickaxe or better to efficiently collect the stone. You currently do not have any pickaxe in your inventory.

Question: List top 3 sub-tasks the player should follow. Indicate their priority out of 5.
Answer:
1. Collect wood from the nearby tree (Priority: 5/5) - This is the most essential sub-task, as wood is the basic resource needed to craft tools and weapons, such as a wood pickaxe, which would enable collecting stone.

2. Craft a wood pickaxe after collecting wood (Priority: 4/5) - Crafting a wood pickaxe will be necessary to efficiently mine stone, another important resource for crafting better tools and building structures.

3. Obtain meat from the nearby cow (Priority: 3/5) - While not as immediately essential as collecting wood or crafting tools, obtaining meat from the cow will help ensure a steady food supply and keep the player's food level at a safe point. Having a weapon like a wood sword would be helpful, but not strictly necessary for this sub-task.

Question: What are the requirements for the top sub-task? What should the player do first?
Answer:
The top sub-task is to collect wood from the nearby tree.

Requirements:
- Be close enough to the tree in order to interact with it.

As the tree is 2 steps to the east of the player, the first thing the player should do is move east to get closer to the tree until they are adjacent to it. Then, they can start interacting with the tree to collect wood.

Question: List top 5 actions the player should take and the requirement for each action. Choose ONLY from the list of all actions. Indicate their priority out of 5.
Answer:
1. Priority 1: Move East (Action 2) - The requirement is that there is flat ground to the east of the player.

2. Priority 2: Move East (Action 2) - The requirement is that there is flat ground to the east of the player, and the tree is not yet adjacent.

3. Priority 3: Do (Action 5) - The requirements are facing the tree (which is now adjacent after moving east) and not having any necessary tool since collecting wood doesn't require a specific tool.

4. Priority 4: Place Table (Action 8) - The requirement is having wood in the inventory (after collecting wood from the tree).

5. Priority 5: Make Wood Pickaxe (Action 11) - The requirement is being near the placed table and having wood in the inventory.

Question: For each action in the list, are the requirements met?
Answer:
1. Priority 1: Move East (Action 2) - Yes, the requirement is met as there is flat ground (grass) to the player's east.

2. Priority 2: Move East (Action 2) - Yes, the requirement is met as there is flat ground (grass) to the player's east and the tree is still two steps away.

3. Priority 3: Do (Action 5) - Not yet, as the player needs to move east one more step to be adjacent to the tree, then this requirement will be met.

4. Priority 4: Place Table (Action 8) - Not yet, as the player needs to collect wood from the tree first. Once wood is collected, this requirement will be met.

5. Priority 5: Make Wood Pickaxe (Action 11) - Not yet, as the player needs to place a table and have wood in the inventory. Once a table is placed, and wood is in the inventory, this requirement will be met.

Question: Did the last player action succeed? Why?
Answer:
Yes, the last player action "move_east" succeeded because there was flat ground (grass) in the direction the player wanted to move (east) and no obstacles were mentioned preventing the movement.

Question: Choose the best executable action from above.
Answer:
The best executable action from above is Priority 1: Move East (Action 2), as its requirement of having flat ground to the east of the player is met.



============Step: 4, Cumulative Reward: 0.0============
You took action move_east.

You see:
- grass 1 steps to your west
- stone 6 steps to your south-east
- path 7 steps to your north-east
- cow 3 steps to your south-east

You face tree at your front (east).

Your status:
- health: 9/9
- food: 9/9
- drink: 9/9
- energy: 9/9

You have nothing in your inventory.
------------ QA ------------
Question: Action
Answer:
Do

Question: List objects in the current observation. For each object, breifly answer what resource it provides and its requirement.
Answer:
1. Grass: 
   Resource: None. 
   Requirement: None.

2. Stone: 
   Resource: Stone for crafting tools, weapons, and building structures. 
   Requirement: Wood pickaxe or better to collect.

3. Path: 
   Resource: None. 
   Requirement: None. It signifies a possible passage for exploration.

4. Cow: 
   Resource: Food and leather.
   Requirement: Defeat the cow using a weapon, such as a wood sword or better.

5. Tree: 
   Resource: Wood for crafting tools, weapons, building structures, and creating a table.
   Requirement: No specific tool required, but using an axe is more efficient.

Question: What was the last action taken by the player?
Answer:
The last action taken by the player was "move_east."

Question: For each object in the list, are the requirements met for interaction?
Answer:
1. Grass: 
   Requirements met for interaction: Yes, you can walk on or step over grass without any specific requirements.

2. Stone: 
   Requirements met for interaction: No, you need a wood pickaxe or better to collect stone, and you currently don't have any pickaxe in your inventory.

3. Path: 
   Requirements met for interaction: Yes, you can walk along the path without any specific requirements.

4. Cow: 
   Requirements met for interaction: No, to properly interact with the cow (defeat it for resources), you need a weapon, such as a wood sword or better, which you don't have in your inventory currently.

5. Tree: 
   Requirements met for interaction: Yes, you can interact with the tree (collect wood) without any specific requirements or tools, but using an axe would be more efficient.

Question: List top 3 sub-tasks the player should follow. Indicate their priority out of 5.
Answer:
1. Priority 5/5 - Collect Wood from the Tree:
   The player should interact with the tree to collect wood, as it is essential for crafting tools, weapons, building structures, and creating a table.

2. Priority 4/5 - Craft a Wood Pickaxe:
   After collecting enough wood, the player should craft a wood pickaxe using a table. The wood pickaxe will enable the collection of stone and other essential resources for further progress in the game.

3. Priority 3/5 - Defeat the Cow for Food and Leather:
   With a crafted weapon, the player can defeat the cow to acquire food and leather resources. Food will help maintain player health, while leather can be used for crafting armor or other items. However, it is advisable to craft a weapon beforehand.

Question: What are the requirements for the top sub-task? What should the player do first?
Answer:
For the top sub-task of collecting wood from the tree, the requirements are:

1. The player needs to be near the tree.
2. No special tools are necessary to collect wood from a tree, although having an axe would make the process faster.

Since these requirements are minimal and the player is already facing the tree, the player should begin by interacting with the tree (e.g. using the "Do" action) to collect wood.

Question: List top 5 actions the player should take and the requirement for each action. Choose ONLY from the list of all actions. Indicate their priority out of 5.
Answer:
1. Priority 1: Do: Interact with the tree to collect wood.
   Requirement: The player must be facing the tree (already fulfilled).

2. Priority 2: Move South: Move towards the stone to collect it.
   Requirement: Flat ground below the agent.

3. Priority 3: Do: Interact with the stone to collect it.
   Requirement: The player must be facing the stone.

4. Priority 4: Place Table: Craft and place a table for making tools.
   Requirement: Wood in the inventory (collected from the tree).

5. Priority 5: Make Wood Pickaxe: Craft a wood pickaxe for collecting stone and coal.
   Requirement: Nearby table (placed in Priority 4) and wood in the inventory (collected from the tree).

Question: For each action in the list, are the requirements met?
Answer:
1. Priority 1: Do: Interact with the tree to collect wood.
   Requirement: The player must be facing the tree.
   Status: Met (as mentioned in Player Observation Step 8 - "You face tree at your front (east)").

2. Priority 2: Move South: Move towards the stone to collect it.
   Requirement: Flat ground below the agent.
   Status: Unknown (not provided in observation).

3. Priority 3: Do: Interact with the stone to collect it.
   Requirement: The player must be facing the stone.
   Status: Not Met (player needs to move south and face the stone before taking this action).

4. Priority 4: Place Table: Craft and place a table for making tools.
   Requirement: Wood in the inventory (collected from the tree).
   Status: Not Met (player needs to take action in Priority 1 to collect wood first).

5. Priority 5: Make Wood Pickaxe: Craft a wood pickaxe for collecting stone and coal.
   Requirement: Nearby table (placed in Priority 4) and wood in the inventory (collected from the tree).
   Status: Not Met (player needs to take actions in Priority 1 and Priority 4 before this action can be taken).

Question: Did the last player action succeed? Why?
Answer:
Yes, the last player action succeeded because the player was able to move east without encountering any obstacles or issues, as evident by the updated observation in Player Observation Step 8.

Question: Choose the best executable action from above.
Answer:
The best executable action from the list above is Priority 1: Do: Interact with the tree to collect wood, as the requirement for this action is already met.



============Step: 5, Cumulative Reward: 1.0============
You took action do.

You see:
- stone 6 steps to your south-east
- path 7 steps to your north-east
- tree 3 steps to your north-east
- cow 3 steps to your south-east

You face grass at your front (east).

Your status:
- health: 9/9
- food: 9/9
- drink: 9/9
- energy: 9/9

Your inventory:
- wood: 1
------------ QA ------------
Question: Action
Answer:
Move North

Question: List objects in the current observation. For each object, breifly answer what resource it provides and its requirement.
Answer:
1. Stone: 
   Resource it provides: Stone material for crafting tools and building structures. 
   Requirement: Wood or Stone Pickaxe to collect.

2. Path:
   Resource it provides: Easy navigation and an indicator of nearby points of interest.
   Requirement: No requirements, just follow the path.

3. Tree:
   Resource it provides: Wood for crafting tools, weapons, and building structures.
   Requirement: No requirements, just punch or use an axe to collect wood.

4. Cow:
   Resource it provides: Food to restore your health.
   Requirement: No requirements, just kill the cow to collect food.

5. Grass:
   Resource it provides: Indicates plain ground and potential area for building or planting.
   Requirement: No requirements, just interact with it for further options available.

Question: What was the last action taken by the player?
Answer:
The last action taken by the player was "do."

Question: For each object in the list, are the requirements met for interaction?
Answer:
1. Stone:
   Are the requirements met for interaction? No, you need a Wood or Stone Pickaxe to collect stone, and currently, you only have wood in your inventory.

2. Path:
   Are the requirements met for interaction? Yes, there are no requirements to interact with the path. You can simply follow the path to navigate the game world.

3. Tree:
   Are the requirements met for interaction? Yes, you can interact with the tree by punching it or using an axe to collect wood. Since you already collected wood from a tree previously, this requirement is met.

4. Cow:
   Are the requirements met for interaction? Yes, you can interact with the cow by trying to kill it for food. There are no requirements to accomplish this, but having a weapon like a wooden sword would make it easier.

5. Grass:
   Are the requirements met for interaction? Yes, you can interact with grass without any requirements. Depending on the game mechanics, you may be able to build, plant or perform other actions on grassy areas.

Question: List top 3 sub-tasks the player should follow. Indicate their priority out of 5.
Answer:
1. Craft a Wooden Pickaxe (Priority: 5/5)
   Using the wood in your inventory, find a crafting table or create one and craft a wooden pickaxe. This will allow you to collect stone in the nearby area.

2. Collect Stone (Priority: 4/5)
   With a wooden pickaxe, collect the nearby stone. Stone is essential for creating more advanced tools, weapons, and building materials.

3. Craft a Wooden Sword (Priority: 3/5)
   Use your crafting table and the wood in your inventory to craft a wooden sword. This will help you protect yourself and more easily kill the cow to collect food.

Question: What are the requirements for the top sub-task? What should the player do first?
Answer:
The requirements for the top sub-task (Craft a Wooden Pickaxe) are:
1. Have wood in your inventory (already fulfilled, as you have 1 wood).
2. Have a crafting table available.

Since you don't have a crafting table yet, the first thing you should do is create one. To do this, collect more wood from the nearby trees, and then craft a wooden table using the collected wood. Once the table is crafted, place it on the ground to start crafting the wooden pickaxe.

Question: List top 5 actions the player should take and the requirement for each action. Choose ONLY from the list of all actions. Indicate their priority out of 5.
Answer:
1. Priority 1: Move North
   Requirement: Flat ground above of the agent.

2. Priority 2: Do (chop tree for more wood)
   Requirement: Facing a tree and have the necessary tool (or none required in this case).

3. Priority 3: Place Table
   Requirement: Wood in inventory.

4. Priority 4: Make Wood Pickaxe
   Requirement: Nearby table and wood in inventory.

5. Priority 5: Move South (towards the stone)
   Requirement: Flat ground below the agent.

Question: For each action in the list, are the requirements met?
Answer:
1. Move North
   Requirement: Flat ground above of the agent.
   - Yes, the requirement is met as there's no obstacle mentioned in the observation to the north.

2. Do (chop tree for more wood)
   Requirement: Facing a tree and have the necessary tool (or none required in this case).
   - No, the requirement is not met because the player is currently facing grass, not a tree.

3. Place Table
   Requirement: Wood in inventory.
   - Yes, the requirement is met because the player has 1 wood in their inventory.

4. Make Wood Pickaxe
   Requirement: Nearby table and wood in inventory.
   - No, the requirement is not met because the player hasn't placed a table yet.

5. Move South (towards the stone)
   Requirement: Flat ground below the agent.
   - Yes, the requirement is met as there's no obstacle mentioned in the observation to the south.

Question: Did the last player action succeed? Why?
Answer:
Yes, the last player action succeeded. The player was facing a tree at their front (east) and took the action "do" to collect wood. After the action, their inventory has 1 wood, which indicates that the action was successful.

Question: Choose the best executable action from above.
Answer:
The best executable action from above is "Move North" because its requirement (flat ground above the agent) is met, and it is the highest priority action with its requirements fulfilled.



============Step: 6, Cumulative Reward: 0.0============
You took action move_north.

You see:
- stone 7 steps to your south-east
- path 6 steps to your north-east
- tree 2 steps to your east
- iron 7 steps to your north-east
- cow 3 steps to your south-east

You face grass at your front (north).

Your status:
- health: 9/9
- food: 9/9
- drink: 9/9
- energy: 9/9

Your inventory:
- wood: 1
------------ QA ------------
Question: Action
Answer:
Place Table

Question: List objects in the current observation. For each object, breifly answer what resource it provides and its requirement.
Answer:
1. Stone: Provides stone resource to craft tools and build structures; requires a wood pickaxe or better to mine.
2. Path: Provides a flat surface for easier movement and travel; no requirements.
3. Tree: Provides wood resource to craft tools and build structures; requires any type of axe to chop down.
4. Iron: Provides iron resource to craft advanced tools and weapons; requires a stone pickaxe or better to mine.
5. Cow: Provides food resource to restore health; requires any type of sword or by hand to kill.
6. Grass: No resource, just a location indicator; no requirements.

Question: What was the last action taken by the player?
Answer:
The last action taken by the player was "move_north."

Question: For each object in the list, are the requirements met for interaction?
Answer:
Based on the player's current inventory and status:

1. Stone: No, as the player needs to craft a wood pickaxe with their wood first.
2. Path: Yes, the requirements are met for interaction, as it only provides flat surface for travel.
3. Tree: No, as the player needs to craft an axe first to chop it down.
4. Iron: No, as the player needs to craft a stone pickaxe before they can mine iron.
5. Cow: Yes, the requirements are met for interaction as the player can attempt to kill a cow with their hands or once they craft a sword.
6. Grass: Yes, the requirements are met for interaction as it only serves as a location indicator.

Question: List top 3 sub-tasks the player should follow. Indicate their priority out of 5.
Answer:
1. Priority 5/5: Craft a wood pickaxe using the wood in the inventory. This will allow the player to mine stone and, eventually, iron.

2. Priority 4/5: Chop down more trees to gather additional wood. This ensures that the player has enough resources to craft further tools, weapons, and build structures.

3. Priority 3/5: Interact with the cow to obtain food resources. This will help the player maintain their health during the game. Note that crafting a weapon first could make this process easier, but it's not an immediate priority.

Question: What are the requirements for the top sub-task? What should the player do first?
Answer:
Requirements for crafting a wood pickaxe:
- Nearby table
- Wood in inventory

The player already has 1 wood in their inventory. The first thing the player should do is place the table using the wood they have. Once the table is placed, the player can craft a wood pickaxe on this table.

Question: List top 5 actions the player should take and the requirement for each action. Choose ONLY from the list of all actions. Indicate their priority out of 5.
Answer:
1. Priority 1: Place Table (required: Wood in inventory)
2. Priority 2: Make Wood Pickaxe (required: Nearby table, wood in inventory)
3. Priority 3: Move East (required: Flat ground right of the agent)
4. Priority 4: Do (required: Facing tree, have necessary tool - Wood Pickaxe)
5. Priority 5: Collect Stone (required: Make Wood Pickaxe)

Question: For each action in the list, are the requirements met?
Answer:
1. Place Table: Yes (Wood is available in the inventory)
2. Make Wood Pickaxe: No (A nearby table is required, which has not been placed yet)
3. Move East: Yes (Flat ground is present to the right of the agent)
4. Do: No (Facing tree is required, and the agent is currently facing grass. Also, a Wood Pickaxe is required, which is not yet crafted)
5. Collect Stone: No (A Wood Pickaxe is required, which is not yet crafted)

Question: Did the last player action succeed? Why?
Answer:
Yes, the last player action (move_north) succeeded because there was flat ground (grass) to the north which allowed the player to move in that direction.

Question: Choose the best executable action from above.
Answer:
The best executable action from the list above is "Place Table" since the requirement (Wood in inventory) is met.



============Step: 7, Cumulative Reward: 0.0============
You took action place_table.

You see:
- stone 7 steps to your south-east
- path 6 steps to your north-east
- tree 2 steps to your east
- iron 7 steps to your north-east
- cow 4 steps to your south-east

You face grass at your front (north).

Your status:
- health: 9/9
- food: 9/9
- drink: 9/9
- energy: 9/9

Your inventory:
- wood: 1
------------ QA ------------
Question: Action
Answer:
Make Wood Pickaxe

Question: List objects in the current observation. For each object, breifly answer what resource it provides and its requirement.
Answer:
1. Stone: Provides stone resource, requires wood pickaxe to collect.
2. Path: Does not provide a resource, no requirement to interact.
3. Tree: Provides wood resource, no specific tool required to collect, but an axe would make it faster.
4. Iron: Provides iron resource, requires stone pickaxe to collect.
5. Cow: Provides food resource, no specific tool required to collect, but a sword will make it easier.
6. Grass: Does not provide a resource, no requirement to interact.

Question: What was the last action taken by the player?
Answer:
The last action taken by the player was "place_table."

Question: For each object in the list, are the requirements met for interaction?
Answer:
1. Stone: The requirement for the wood pickaxe is met since the player has already placed a table and has wood in their inventory to make a wood pickaxe.
2. Path: No requirements needed for interaction, the player can move on the path.
3. Tree: No specific tool is required, but an axe would speed up the process. The player can interact with the tree without any requirements.
4. Iron: The requirement of a stone pickaxe is not met yet, as the player needs to first collect stone and craft a stone pickaxe using the placed table.
5. Cow: No specific tool required, but a sword would help. The player can interact with the cow without any requirements.
6. Grass: No requirements needed for interaction, the player can move on the grass.

Question: List top 3 sub-tasks the player should follow. Indicate their priority out of 5.
Answer:
1. Collect stone (Priority: 5/5): The player should collect stone to craft a stone pickaxe, which would allow them to harvest iron and progress further in the game.

2. Craft a wood pickaxe (Priority: 4/5): The player should craft a wood pickaxe using their wood and placed table, so they can collect stone for making a stone pickaxe and other necessary tools.

3. Approach and collect wood from the tree (Priority: 3/5): The player should collect more wood from the tree, as it is an essential resource for crafting additional tools and a weapon for defense.

Question: What are the requirements for the top sub-task? What should the player do first?
Answer:
The top sub-task is to collect stone. The requirements for this sub-task are:

1. Having a wood pickaxe, which is needed to collect stone.

To achieve this sub-task, the player should first craft a wood pickaxe. Since they already have a placed table and wood in their inventory, they can use these resources to craft the wood pickaxe. Following the crafting, the player can use their wood pickaxe to start collecting stone.

Question: List top 5 actions the player should take and the requirement for each action. Choose ONLY from the list of all actions. Indicate their priority out of 5.
Answer:
1. Priority 1: Make Wood Pickaxe
   Requirement: Nearby table; wood in inventory.

2. Priority 2: Move East
   Requirement: Flat ground right to the agent.

3. Priority 3: Do (Collect stone)
   Requirement: Facing stone; have wood pickaxe.

4. Priority 4: Move South-East
   Requirement: Flat ground in the south-east direction of the agent.

5. Priority 5: Do (Chase cow)
   Requirement: Facing cow; no specific tool required.

Question: For each action in the list, are the requirements met?
Answer:
1. Priority 1: Make Wood Pickaxe
   Requirement: Nearby table; wood in inventory.
   - Requirements met: Yes, the player has placed a table and has wood in their inventory.

2. Priority 2: Move East
   Requirement: Flat ground right to the agent.
   - Requirements met: Yes, the player can move east as there is a tree 2 steps to the east and flat ground right to the agent.

3. Priority 3: Do (Collect stone)
   Requirement: Facing stone; have wood pickaxe.
   - Requirements met: Partially, the player needs to first craft a wood pickaxe (Priority 1) and then face a stone to collect it.

4. Priority 4: Move South-East
   Requirement: Flat ground in the south-east direction of the agent.
   - Requirements met: Yes, as there is a cow 4 steps to the player's south-east and a stone 7 steps to the south-east, it indicates there is flat ground in that direction.

5. Priority 5: Do (Chase cow)
   Requirement: Facing cow; no specific tool required.
   - Requirements met: Partially, the player is not currently facing a cow. They must move in the south-east direction (Priority 4) and then face the cow to chase it.

Question: Did the last player action succeed? Why?
Answer:
It appears that the last player action "place_table" succeeded, as there is no mention of any failure in the most recent in-game observation. Additionally, the player's inventory shows a deduction of wood, which would have been used to craft the table.

Question: Choose the best executable action from above.
Answer:
The best executable action from the list above is Priority 1: Make Wood Pickaxe, as all the requirements for this action are met. The player has a nearby table and wood in their inventory.



============Step: 8, Cumulative Reward: 0.0============
You took action make_wood_pickaxe.

You see:
- stone 7 steps to your south-east
- path 6 steps to your north-east
- tree 2 steps to your east
- iron 7 steps to your north-east
- cow 4 steps to your south-east

You face grass at your front (north).

Your status:
- health: 9/9
- food: 9/9
- drink: 9/9
- energy: 9/9

Your inventory:
- wood: 1
------------ QA ------------
Question: Action
Answer:
Move East

Question: List objects in the current observation. For each object, breifly answer what resource it provides and its requirement.
Answer:
1. Stone: Provides stone resource, required for crafting stone tools and building structures. Requirement: Use a wood pickaxe to collect.

2. Path: Provides a flat and safe route for travelling. Requirement: None, simply walk on it.

3. Tree: Provides wood resource, required for crafting tools, weapons, and building structures. Requirement: Use hands or an axe to chop it down.

4. Iron: Provides iron resource, required for crafting iron tools and weapons. Requirement: Use a stone pickaxe to mine.

5. Cow: Provides food resource when killed that restores health. Requirement: Kill cow with a weapon or any tool.

Question: What was the last action taken by the player?
Answer:
The last action taken by the player was to make a wood pickaxe.

Question: For each object in the list, are the requirements met for interaction?
Answer:
1. Stone: Yes, the requirement of a wood pickaxe is met, as it is in your inventory.

2. Path: Yes, there are no specific requirements to interact with a path, so you can walk on it.

3. Tree: Yes, you can interact with the tree using your hands or an axe (although an axe would be more efficient).

4. Iron: No, you currently have a wood pickaxe, but you need a stone pickaxe to mine iron.

5. Cow: Yes, you can interact with the cow using any tool or weapon, although using a weapon (like a sword) would be more efficient.

Question: List top 3 sub-tasks the player should follow. Indicate their priority out of 5.
Answer:
1. Priority 4.5: Collect Stone - The player needs stone to craft a stone pickaxe for mining iron. The wood pickaxe is available in the inventory, so collecting stone is possible.

2. Priority 4: Craft a Stone Pickaxe - After collecting stone, the player should craft a stone pickaxe using the wood and stone in their inventory to be able to mine iron.

3. Priority 3.5: Mine Iron - Once the player has crafted a stone pickaxe, they can mine iron to eventually create iron tools and weapons, which are stronger and more efficient.

Question: What are the requirements for the top sub-task? What should the player do first?
Answer:
Top sub-task: Collect Stone

Requirements for Collect Stone:
- Make a Wood Pickaxe (Already in the player's inventory)

What the player should do first:
1. Move towards the stone: As the stone is 7 steps to the player's south-east, the player should navigate towards it by moving east and south until they reach the stone.
2. Mine the stone: Once at the stone location, the player should use the wood pickaxe they have crafted to mine the stone.

Question: List top 5 actions the player should take and the requirement for each action. Choose ONLY from the list of all actions. Indicate their priority out of 5.
Answer:
1. Move East (Priority 1): Requirement: Flat ground right of the agent.
2. Move South (Priority 2): Requirement: Flat ground below the agent.
3. Use Wood Pickaxe (Priority 3): Requirement: Facing the stone; have necessary tool (wood pickaxe) in inventory.
4. Move to the cow (Priority 4): Navigate to the cow by moving in the directions necessary to reach its location, depending on the position of the player after obtaining the stone.
5. Do (Priority 5): Defeat the cow: Requirement: Facing the cow; have necessary tool (weapon) or use bare hands to defeat the cow.

Question: For each action in the list, are the requirements met?
Answer:
1. Move East (Priority 1): Yes, the requirement is met. There's flat ground to the right (east) of the agent.
2. Move South (Priority 2): Yes, the requirement is met. There's flat ground below (south) of the agent.
3. Use Wood Pickaxe (Priority 3): The requirement will be met once the player is facing the stone with the wood pickaxe in their inventory.
4. Move to the cow (Priority 4): The requirement will be met once the player navigates through the terrain by moving in the necessary directions to reach the cow's location after obtaining the stone.
5. Do (Priority 5): Defeat the cow: The requirement will be met once the player is facing the cow and the necessary tool (weapon) is in their inventory or they can use their bare hands.

Question: Did the last player action succeed? Why?
Answer:
Yes, the last player action to make a wood pickaxe succeeded. This is because the player had the necessary requirements, which were having wood in their inventory and being near a table. Since both requirements were met, the player was able to craft the wood pickaxe.

Question: Choose the best executable action from above.
Answer:
The best executable action from above is "Move East" (Priority 1), as the requirement of having flat ground to the right (east) of the agent is met.



============Step: 9, Cumulative Reward: 0.0============
You took action move_east.

You see:
- grass 1 steps to your west
- stone 4 steps to your east
- path 5 steps to your north-east
- coal 7 steps to your north-east
- iron 6 steps to your north-east
- cow 4 steps to your south-east

You face tree at your front (east).

Your status:
- health: 9/9
- food: 9/9
- drink: 9/9
- energy: 9/9

Your inventory:
- wood: 1
------------ QA ------------
Question: Action
Answer:
Move East

Question: List objects in the current observation. For each object, breifly answer what resource it provides and its requirement.
Answer:
1. Grass: Provides no resources, no requirements.
2. Stone: Provides stone resource, requires a wood pickaxe or better to mine.
3. Path: Provides no resources, no requirements.
4. Coal: Provides coal resource, requires a wood pickaxe or better to mine.
5. Iron: Provides iron resource, requires a stone pickaxe or better to mine.
6. Cow: Provides food resource when defeated, no specific tool requirement for defeating but a weapon such as a sword is recommended.
7. Tree: Provides wood resource, requires no specific tool to collect but using an axe is more efficient.

Question: What was the last action taken by the player?
Answer:
The last action taken by the player was move_east.

Question: For each object in the list, are the requirements met for interaction?
Answer:
1. Grass: No specific requirements, so it is met.
2. Stone: Requires a wood pickaxe or better to mine. The requirement is met as the player has a wood pickaxe in their inventory.
3. Path: No specific requirements, so it is met.
4. Coal: Requires a wood pickaxe or better to mine. The requirement is met as the player has a wood pickaxe in their inventory.
5. Iron: Requires a stone pickaxe or better to mine. The current requirement is not met, as the player only has a wood pickaxe in their inventory.
6. Cow: No specific tool requirement for defeating, but a weapon is recommended. The current requirement is not met, as the player does not have a weapon in their inventory.
7. Tree: No specific tool required to collect wood, so the requirement is met. However, using an axe would be more efficient, and the player does not have an axe in their inventory currently.

Question: List top 3 sub-tasks the player should follow. Indicate their priority out of 5.
Answer:
1. Collect stone (Priority: 5/5) - To make a stone pickaxe, the player needs to collect stone. They can mine the stone since they have a wood pickaxe in their inventory.

2. Make a stone pickaxe (Priority: 4/5) - After collecting stone, the player should craft a stone pickaxe, which will allow them to mine iron and progress further in the game.

3. Collect wood and make a weapon (Priority: 3/5) - Defeating a cow for food and defending against possible monsters will be easier with a weapon. Collecting wood and crafting a wood or stone sword should be a priority.

Question: What are the requirements for the top sub-task? What should the player do first?
Answer:
Requirements for the top sub-task (Collect stone):

1. Have a wood pickaxe in the inventory (Player already has one).

The player should first move towards the stone they see 4 steps to the east. Once they are adjacent to the stone, they can mine it using their wood pickaxe to collect it.

Question: List top 5 actions the player should take and the requirement for each action. Choose ONLY from the list of all actions. Indicate their priority out of 5.
Answer:
1. Priority 1: Move East (4 steps) - Requirement: Flat ground to the east of the player.
2. Priority 2: Do (Mine Stone) - Requirement: Facing stone and have a wood pickaxe in inventory.
3. Priority 3: Move towards the iron (north-east) - Requirement: Flat ground in the north-east direction.
4. Priority 4: Do (Mine Iron) - Requirement: Facing iron and have a stone pickaxe in inventory (once the player collects enough stone, they should craft a stone pickaxe).
5. Priority 5: Move towards coal (north-east) - Requirement: Flat ground in the north-east direction.

Question: For each action in the list, are the requirements met?
Answer:
1. Priority 1: Move East (4 steps) - Yes, the requirement of flat ground to the east of the player is met, as the player is currently facing a tree to the east.
2. Priority 2: Do (Mine Stone) - No, the requirement of facing stone and having a wood pickaxe in inventory is not met yet, as the player needs to move closer to the stone (4 steps to the east) and face it.
3. Priority 3: Move towards the iron (north-east) - Yes, the requirement of flat ground in the north-east direction is met, as the path is 5 steps away to the north-east.
4. Priority 4: Do (Mine Iron) - No, the requirement of facing iron and having a stone pickaxe in inventory is not met yet, as the player needs to move closer to the iron (which is 6 steps to the north-east) and face it. Additionally, the player needs to collect enough stone and craft a stone pickaxe.
5. Priority 5: Move towards coal (north-east) - Yes, the requirement of flat ground in the north-east direction is met, as the coal is 7 steps away to the north-east.

Question: Did the last player action succeed? Why?
Answer:
Yes, the last player action, move_east, succeeded. The action was successful because there was flat ground (grass) to the east of the player, allowing them to move in that direction.

Question: Choose the best executable action from above.
Answer:
The best executable action from the list above is Priority 1: Move East (4 steps), as the requirement of flat ground to the east of the player is already met.



============Step: 10, Cumulative Reward: 0.0============
You took action move_east.

You see:
- grass 1 steps to your west
- stone 4 steps to your east
- path 5 steps to your north-east
- coal 7 steps to your north-east
- iron 6 steps to your north-east

You face tree at your front (east).

Your status:
- health: 9/9
- food: 9/9
- drink: 9/9
- energy: 9/9

Your inventory:
- wood: 1
------------ QA ------------
Question: Action
Answer:
Do

Question: List objects in the current observation. For each object, breifly answer what resource it provides and its requirement.
Answer:
1. Grass: Resource it provides - saplings for planting and growing plants (optional); Requirement - none.
2. Stone: Resource it provides - stone for crafting tools, weapons, and building structures; Requirement - wood pickaxe or better for collecting.
3. Path: Resource it provides - navigation and easier movement; Requirement - none.
4. Coal: Resource it provides - fuel for smelting iron in a furnace; Requirement - wood pickaxe or better for collecting.
5. Iron: Resource it provides - iron for crafting advanced tools and weapons; Requirement - stone pickaxe or better for collecting.
6. Tree: Resource it provides - wood for crafting tools, weapons, and building structures; Requirement - none (can collect with hands) or use an axe for faster collection.

Note: The cow mentioned in step 18 is not present in step 20, so it's not included in the list.

Question: What was the last action taken by the player?
Answer:
The last action taken by the player was move_east.

Question: For each object in the list, are the requirements met for interaction?
Answer:
1. Grass: Yes, the requirements for interaction are met. You don't need any tools to interact with grass.
2. Stone: No, the requirements for interaction are not met. You need a wood pickaxe or better for collecting stone, and you only have one piece of wood in your inventory.
3. Path: Yes, the requirements for interaction are met. You can navigate and move on the path without any additional tools or requirements.
4. Coal: No, the requirements for interaction are not met. You need a wood pickaxe or better for collecting coal, and you only have one piece of wood in your inventory.
5. Iron: No, the requirements for interaction are not met. You need a stone pickaxe or better for collecting iron, and you currently don't have a pickaxe.
6. Tree: Yes, the requirements for interaction are met. You can collect wood from the tree without any tools or use an axe (if you have one) for faster collection.

Question: List top 3 sub-tasks the player should follow. Indicate their priority out of 5.
Answer:
1. Priority 5 - Collect more wood: Interact with trees to gather enough wood. Wood is an essential resource for crafting basic tools, weapons, and building structures.

2. Priority 4 - Craft a wood pickaxe: Once you have collected enough wood, craft a wood pickaxe using a table. A wood pickaxe is required to collect stone, coal, and other resources.

3. Priority 3 - Collect stone and coal: After crafting the wood pickaxe, collect stone and coal. These resources will be needed to craft better tools, weapons, and a furnace for smelting iron.

Question: What are the requirements for the top sub-task? What should the player do first?
Answer:
For the top sub-task, which is collecting more wood, the requirements are:

1. Interact with trees to gather wood.

What should the player do first:
- The player should move towards a tree and interact with it (face the tree) to gather wood. Since the player currently faces a tree to their east, they can directly interact with that tree to collect wood. This will help them accumulate the required wood for crafting basic tools and structures.

Question: List top 5 actions the player should take and the requirement for each action. Choose ONLY from the list of all actions. Indicate their priority out of 5.
Answer:
1. Priority 1: Do (Interact with a tree to gather wood) 
Requirement: Facing a tree and have necessary tool (or no tool required for wood).

2. Priority 2: Place Table (Place a crafting table)
Requirement: Wood in inventory.

3. Priority 3: Make Wood Pickaxe (Craft a wooden pickaxe)
Requirement: Nearby table and wood in inventory.

4. Priority 4: Move East (Move towards the stone 4 steps to the east)
Requirement: Flat ground right to the agent.

5. Priority 5: Do (Interact with stone to gather using the wood pickaxe)
Requirement: Facing stone and have the wood pickaxe.

Question: For each action in the list, are the requirements met?
Answer:
1. Priority 1: Do (Interact with a tree to gather wood)
Requirements met: Yes, facing the tree and no tool is required for collecting wood.

2. Priority 2: Place Table (Place a crafting table)
Requirements met: Yes, there is 1 wood in the inventory.

3. Priority 3: Make Wood Pickaxe (Craft a wooden pickaxe)
Requirements met: No, a table should be placed first.

4. Priority 4: Move East (Move towards the stone 4 steps to the east)
Requirements met: Yes, there is flat ground to the east.

5. Priority 5: Do (Interact with stone to gather using the wood pickaxe)
Requirements met: No, the wood pickaxe should be crafted before interacting with the stone.

Question: Did the last player action succeed? Why?
Answer:
Yes, the last player action (move_east) succeeded. The player was able to move one step to the east because there was no obstacle or barrier blocking their way, as evidenced by the observation that there is grass one step to their west after the move.

Question: Choose the best executable action from above.
Answer:
The best executable action from above is Priority 1: Do (Interact with a tree to gather wood), as the requirements for this action are met. The player is facing the tree, and no tool is required for collecting wood.



============Step: 11, Cumulative Reward: 0.0============
You took action do.

You see:
- stone 4 steps to your east
- path 5 steps to your north-east
- tree 5 steps to your north-east
- coal 7 steps to your north-east
- iron 6 steps to your north-east

You face grass at your front (east).

Your status:
- health: 9/9
- food: 9/9
- drink: 9/9
- energy: 9/9

Your inventory:
- wood: 2
------------ QA ------------
Question: Action
Answer:
Move East

Question: List objects in the current observation. For each object, breifly answer what resource it provides and its requirement.
Answer:
1. Stone: Provides stone resource used to craft tools like stone pickaxe and stone sword. Requirement: Make Wood Pickaxe to collect stone.
2. Path: Provides a navigable route to explore the game world. Requirement: None.
3. Tree: Provides wood resource used to craft tools, weapons, and build structures. Requirement: Do action on the tree with no tool or Make Wood Axe to collect wood efficiently.
4. Coal: Provides coal resource used in crafting torches and fuel for furnaces. Requirement: Make Wood Pickaxe to collect coal.
5. Iron: Provides iron resource used to craft tools like iron pickaxe and iron sword. Requirement: Make Stone Pickaxe to collect iron.
6. Grass: Provides a basic terrain to navigate upon. Requirement: None.

Question: What was the last action taken by the player?
Answer:
The last action taken by the player was "do."

Question: For each object in the list, are the requirements met for interaction?
Answer:
1. Stone: Yes, you have a Wood Pickaxe (requirement met).
2. Path: Yes, no requirements needed for interaction (requirement met).
3. Tree: Yes, you can do action on the tree with no tool (requirement met).
4. Coal: Yes, you have a Wood Pickaxe (requirement met).
5. Iron: No, you need a Stone Pickaxe to collect iron (requirement not met).
6. Grass: Yes, no requirements needed for interaction (requirement met).

Question: List top 3 sub-tasks the player should follow. Indicate their priority out of 5.
Answer:
1. Craft a Stone Pickaxe (Priority: 5/5): Collect more stone using your Wood Pickaxe, then craft a Stone Pickaxe to enable collecting iron.

2. Collect Coal (Priority: 4/5): Using your Wood Pickaxe, collect coal to fuel a furnace for later ore smelting.

3. Craft a Stone Sword (Priority: 3/5): After collecting enough stone, craft a Stone Sword to improve your ability to defend against monsters.

Question: What are the requirements for the top sub-task? What should the player do first?
Answer:
The requirements for the top sub-task (Craft a Stone Pickaxe) are as follows:

1. Collect Stone: The player must gather enough stone using a Wood Pickaxe.
2. Place Table: The player must have or craft a crafting table, which requires having wood in the inventory.
3. Make Stone Pickaxe: The player must go near the crafting table with enough wood and stone in the inventory.

The player should first ensure they have a crafting table placed. If not, they should place a crafting table using the wood in their inventory. Once the table is set, the player should gather more stone using their Wood Pickaxe, and then return to the table to craft the Stone Pickaxe.

Question: List top 5 actions the player should take and the requirement for each action. Choose ONLY from the list of all actions. Indicate their priority out of 5.
Answer:
1. Priority: 5/5 - Place Table (Requirement: Wood in inventory)
Since the player has wood in their inventory, they should place a table to perform crafting actions.

2. Priority: 4/5 - Make Wood Pickaxe (Requirement: Nearby table; wood in inventory)
With a table placed and wood available, the player should create a Wood Pickaxe to gather stone.

3. Priority: 3/5 - Move East or North-East (Requirement: Flat ground in the desired direction)
The player should move towards the stone (east) or towards the path and other resources (north-east).

4. Priority: 2/5 - Do (Facing stone; have a Wood Pickaxe)
Upon reaching the stone, the player should use their Wood Pickaxe to collect the stone.

5. Priority: 1/5 - Make Stone Pickaxe (Requirement: Nearby table; wood, stone in inventory)
After acquiring enough stone, the player should return to the crafting table and make a Stone Pickaxe to gather higher-tier resources like iron.

Question: For each action in the list, are the requirements met?
Answer:
1. Place Table (Requirement: Wood in inventory) - Yes, the requirement is met. The player has 2 wood in their inventory.

2. Make Wood Pickaxe (Requirement: Nearby table; wood in inventory) - No, the requirement is not met yet. The player needs to place a table first.

3. Move East or North-East (Requirement: Flat ground in the desired direction) - Yes, the requirement is met. Flat ground is available in both the east and north-east directions.

4. Do (Facing stone; have a Wood Pickaxe) - No, the requirement is not met. The player needs to have a Wood Pickaxe first and needs to face stone to collect it.

5. Make Stone Pickaxe (Requirement: Nearby table; wood, stone in inventory) - No, the requirement is not met. The player needs to place a table, craft a Wood Pickaxe, and gather stone before making a Stone Pickaxe.

Question: Did the last player action succeed? Why?
Answer:
Yes, the last player action succeeded. The player was initially facing a tree to the east, and after taking the "do" action, they now have an additional piece of wood in their inventory, indicating that they successfully chopped down the tree.

Question: Choose the best executable action from above.
Answer:
The best executable action from the provided list is:

Move East or North-East (Requirement: Flat ground in the desired direction)

Since the player has the required flat ground in both the east and north-east directions, they can move toward the stone (east) or toward the path and other resources (north-east).



============Step: 12, Cumulative Reward: 0.0============
You took action move_east.

You see:
- stone 3 steps to your east
- path 4 steps to your north-east
- tree 4 steps to your north-east
- coal 6 steps to your north-east
- iron 5 steps to your north-east

You face grass at your front (east).

Your status:
- health: 9/9
- food: 9/9
- drink: 9/9
- energy: 9/9

Your inventory:
- wood: 2
------------ QA ------------
Question: Action
Answer:
Move East

Question: List objects in the current observation. For each object, breifly answer what resource it provides and its requirement.
Answer:
1. Stone: Provides stone resource which is used to craft stone tools and weapons; requires a wood pickaxe or better to collect.
2. Path: Allows easy movement and navigation; no requirements to interact.
3. Tree: Provides wood resource which is used to craft wooden tools and structures; requires an axe or hands to collect.
4. Coal: Provides coal resource which is used as fuel for smelting and crafting torches; requires a wood pickaxe or better to collect.
5. Iron: Provides iron resource which is used to craft iron tools and weapons; requires a stone pickaxe or better to collect.
6. Grass: Provides a surface to walk on and can be dug with a shovel to obtain dirt; has no direct use as a resource but can be cleared to create space for other structures.

Question: What was the last action taken by the player?
Answer:
The last action taken by the player was move_east.

Question: For each object in the list, are the requirements met for interaction?
Answer:
1. Stone: Partially met. You have a wood pickaxe or better, but it is not specified that you currently have it equipped.
2. Path: Met. There are no requirements to interact with the path, and you can move or walk on it.
3. Tree: Partially met. You can collect wood with an axe or hands, but it is not specified that you currently have an axe equipped or are using your hands.
4. Coal: Partially met. You have a wood pickaxe or better, but it is not specified that you currently have it equipped.
5. Iron: Not met. You need a stone pickaxe or better to collect iron, and there is no indication that you currently have it.
6. Grass: Met. There are no requirements to interact with grass, and you can walk on it or clear it as needed.

Question: List top 3 sub-tasks the player should follow. Indicate their priority out of 5.
Answer:
1. Priority 4/5: Craft stone pickaxe - In order to collect iron and progress in the game, you need to craft a stone pickaxe. Since you already have wood in your inventory, and there is stone nearby, this should be your first priority.

2. Priority 3/5: Collect coal - With your wood pickaxe or better, collect coal from the nearby deposit. Coal is a useful resource for smelting and crafting torches, which will be beneficial as you progress in the game.

3. Priority 2/5: Collect iron - After crafting a stone pickaxe, use it to mine the nearby iron deposit. Collecting iron will enable you to craft more advanced tools and weapons, further improving your abilities in the game.

Question: What are the requirements for the top sub-task? What should the player do first?
Answer:
For the top sub-task of crafting a stone pickaxe, the requirements are:

1. Collect stone: You will need stone to craft the stone pickaxe. There's stone 3 steps to your east.

2. Place a table: You will need a table to craft the stone pickaxe. You already have wood in your inventory, so you can place a table.

What the player should do first:
1. Move east to the stone location: To collect the stone, you need to move closer to it. Move 3 steps east to reach the stone.

Question: List top 5 actions the player should take and the requirement for each action. Choose ONLY from the list of all actions. Indicate their priority out of 5.
Answer:
Here are the top 5 actions the player should take along with their requirements and priority (out of 5):

1. Move East (Priority: 1):
   Requirement: Flat ground right to the agent.
   Purpose: Get closer to the stone for crafting a stone pickaxe.

2. Do (Priority: 2):
   Requirement: Facing creature or material; have necessary tool.
   Purpose: Collect the stone when you reach it for crafting tools like a stone pickaxe.

3. Place Table (Priority: 3):
   Requirement: Wood in inventory.
   Purpose: Place a table to be able to craft a stone pickaxe.

4. Make Wood Pickaxe (Priority: 4):
   Requirement: Nearby table; wood in inventory.
   Purpose: Craft a wood pickaxe needed to collect stone.

5. Make Stone Pickaxe (Priority: 5):
   Requirement: Nearby table; wood, stone in inventory.
   Purpose: Craft a stone pickaxe to collect advanced materials like iron.

Question: For each action in the list, are the requirements met?
Answer:
For each action in the list, let's see if the requirements are met:

1. Move East (Priority: 1):
   Requirement: Flat ground right to the agent.
   Status: Met (You face grass at your front (east).)

2. Do (Priority: 2):
   Requirement: Facing creature or material; have necessary tool.
   Status: Not met yet (You need to be near stone and have the necessary tool.)

3. Place Table (Priority: 3):
   Requirement: Wood in inventory.
   Status: Met (Your inventory contains 2 wood.)

4. Make Wood Pickaxe (Priority: 4):
   Requirement: Nearby table; wood in inventory.
   Status: Not met yet (You need to place a table first.)

5. Make Stone Pickaxe (Priority: 5):
   Requirement: Nearby table; wood, stone in inventory.
   Status: Not met yet (You need to place a table and have stone in inventory.)

Question: Did the last player action succeed? Why?
Answer:
Yes, the last player action (move_east) succeeded because there was flat ground to the east of the player, allowing them to move in that direction.

Question: Choose the best executable action from above.
Answer:
The best executable action from the list is:

1. Move East (Priority: 1)
   Requirement: Flat ground right to the agent.
   Status: Met (You face grass at your front (east).)



============Step: 13, Cumulative Reward: 0.0============
You took action move_east.

You see:
- stone 2 steps to your east
- path 3 steps to your north-east
- tree 3 steps to your north
- coal 5 steps to your north-east
- iron 4 steps to your north-east

You face grass at your front (east).

Your status:
- health: 9/9
- food: 9/9
- drink: 9/9
- energy: 9/9

Your inventory:
- wood: 2
------------ QA ------------
Question: Action
Answer:
Place Table

Question: List objects in the current observation. For each object, breifly answer what resource it provides and its requirement.
Answer:
1. Stone: Provides stone resource for crafting tools and building materials; requires a wood pickaxe or better to collect.
2. Path: Not a resource, but provides easy navigation and movement across the terrain.
3. Tree: Provides wood resource for crafting tools and building materials; no specific requirement to collect, can be collected with hands or any tool.
4. Coal: Provides coal resource for fueling furnaces and smelting materials; requires a wood pickaxe or better to collect.
5. Iron: Provides iron resource for crafting advanced tools and weapons; requires a stone pickaxe or better to collect.
6. Grass: Not a resource, serves as a ground cover and visual element in the game environment.

Question: What was the last action taken by the player?
Answer:
The last action taken by the player was move_east.

Question: For each object in the list, are the requirements met for interaction?
Answer:
Based on the current status and inventory:

1. Stone: Yes, the requirement is met, as the player has a wood pickaxe or better.
2. Path: Yes, the requirement is met, as the player can always move and navigate on the path.
3. Tree: Yes, the requirement is met, as the player can collect wood with their hands or any tool.
4. Coal: Yes, the requirement is met, as the player has a wood pickaxe or better.
5. Iron: No, the requirement is not met, as the player needs a stone pickaxe or better to collect iron, but currently only has a wood pickaxe.
6. Grass: Yes, the requirement is met, as the player can always interact with grass as a ground cover and visual element.

Question: List top 3 sub-tasks the player should follow. Indicate their priority out of 5.
Answer:
1. Craft a stone pickaxe (Priority: 5) - This task is most important to progress in the game because a stone pickaxe allows the player to collect iron, which is needed for crafting advanced tools and weapons.

2. Collect coal (Priority: 4) - Collecting coal should be the next priority to fuel furnaces for smelting iron once it's collected. Coal is essential for advancing in the game and crafting higher-level tools and weapons.

3. Collect more wood (Priority: 3) - Gathering more wood is important, as it is a basic resource used for crafting tools, weapons, and building materials. The player should prioritize this task to ensure they have enough wood for future needs.

Question: What are the requirements for the top sub-task? What should the player do first?
Answer:
The top sub-task is to craft a stone pickaxe, and the requirements for it are as follows:

1. Collect wood
2. Place a table
3. Make a wood pickaxe
4. Collect stone

Since the player already has 2 pieces of wood in their inventory, they can do the following steps:

1. Place a table using the wood in the inventory.
2. Craft a wood pickaxe using the placed table.
3. Collect stone using the wood pickaxe.
4. Craft a stone pickaxe using the collected stone and the table.

Question: List top 5 actions the player should take and the requirement for each action. Choose ONLY from the list of all actions. Indicate their priority out of 5.
Answer:
1. Priority 1 - Place Table: Requires wood in inventory (the player already has 2 pieces of wood)
2. Priority 2 - Make Wood Pickaxe: Requires a nearby table (after placing one in Priority 1)
3. Priority 3 - Move East: Flat ground to the east of the player (to get closer to the stone)
4. Priority 4 - Move East: Flat ground to the east of the player (to reach the stone)
5. Priority 5 - Do: Facing stone and having a wood pickaxe to collect stone

Question: For each action in the list, are the requirements met?
Answer:
1. Place Table: Yes, the requirement is met as the player has 2 pieces of wood in their inventory.
2. Make Wood Pickaxe: Not yet, the player should first place a table (Priority 1 action) to meet the requirement.
3. Move East: Yes, the requirement is met as the player observes flat ground to the east.
4. Move East: Yes, the requirement is met as the player observes flat ground to the east.
5. Do: Not yet, the player should first make a wood pickaxe (Priority 2 action) and move closer to the stone (Priority 3 and 4 actions) to meet the requirements.

Question: Did the last player action succeed? Why?
Answer:
Yes, the last player action succeeded because the player was able to move east, which is evident from the change in the player's observations of the surroundings after taking the action. The player faced grass at the front (east), which indicates that there was flat ground allowing the player to move east.

Question: Choose the best executable action from above.
Answer:
The best executable action from the above list is Priority 1 - Place Table, as the requirement for this action is already met with the player having 2 pieces of wood in their inventory.



============Step: 14, Cumulative Reward: 1.0============
You took action place_table.

You see:
- grass 1 steps to your west
- stone 2 steps to your east
- path 3 steps to your north-east
- tree 3 steps to your north
- coal 5 steps to your north-east
- iron 4 steps to your north-east

You face table at your front (east).

Your status:
- health: 9/9
- food: 9/9
- drink: 9/9
- energy: 9/9

You have nothing in your inventory.
------------ QA ------------
Question: Action
Answer:
Move North

Question: List objects in the current observation. For each object, breifly answer what resource it provides and its requirement.
Answer:
1. Grass: Provides no resource, no requirement to interact.
2. Stone: Provides stone resource when mined with a wood, stone, or iron pickaxe.
3. Path: Provides no resource, helps in navigation, no requirement to interact.
4. Tree: Provides wood resource when chopped with an axe, no requirement to interact.
5. Coal: Provides coal resource when mined with a wood, stone, or iron pickaxe.
6. Iron: Provides iron resource when mined with a stone or iron pickaxe.
7. Table: Provides crafting station for tools and weapons, requires wood to place.

Question: What was the last action taken by the player?
Answer:
The last action taken by the player was placing a table (place_table).

Question: For each object in the list, are the requirements met for interaction?
Answer:
1. Grass: No requirements to interact, so yes, the requirements are met.
2. Stone: Requires a wood, stone, or iron pickaxe for mining. Currently, you don't have a pickaxe, so the requirements are not met.
3. Path: No requirements to interact, so yes, the requirements are met.
4. Tree: Requires an axe to chop for wood. Currently, you don't have an axe, so the requirements are not met.
5. Coal: Requires a wood, stone, or iron pickaxe for mining. Currently, you don't have a pickaxe, so the requirements are not met.
6. Iron: Requires a stone or iron pickaxe for mining. Currently, you don't have a pickaxe, so the requirements are not met.
7. Table: No requirements to interact as it's already placed, so yes, the requirements are met.

Question: List top 3 sub-tasks the player should follow. Indicate their priority out of 5.
Answer:
1. Craft a Wood Pickaxe (Priority: 5/5) - Crafting a wood pickaxe allows you to mine stone, coal, and iron which are essential resources for progressing in the game.

2. Collect Stone (Priority: 4/5) - Once you have a wood pickaxe, collecting stone is important to create stone tools and weapons, as well as building structures like a furnace.

3. Mine Coal and Iron (Priority: 3/5) - After obtaining a pickaxe and collecting stone, mining coal and iron will provide you with the necessary resources to make better tools and weapons, including an iron pickaxe for obtaining diamonds.

Question: What are the requirements for the top sub-task? What should the player do first?
Answer:
Top Sub-task: Craft a Wood Pickaxe

Requirements:
1. Nearby Table: You have already placed a table.
2. Wood in inventory: You currently have no wood in your inventory.

What should the player do first?

The player should collect more wood to fulfill the requirement for crafting a wood pickaxe. They can take the action move_north to approach the tree and then use the action do to collect the wood.

Question: List top 5 actions the player should take and the requirement for each action. Choose ONLY from the list of all actions. Indicate their priority out of 5.
Answer:
1. Priority 1: Move North (Requirement: Flat ground above the agent)
- The player should move north to approach the tree to collect more wood.

2. Priority 2: Do (Requirement: Facing creature or material; have necessary tool)
- The player should perform the "Do" action when facing the tree to collect wood. No tool is necessary for collecting wood from the tree.

3. Priority 3: Move South (Requirement: Flat ground below the agent)
- After collecting the wood, the player should move back south towards the table.

4. Priority 4: Make Wood Pickaxe (Requirement: Nearby Table; wood in inventory)
- With enough wood in the inventory, the player should create a wood pickaxe using the previously placed table.

5. Priority 5: Move East (Requirement: Flat ground right to the agent)
- After crafting the wood pickaxe, the player should move east towards the stone to use the pickaxe and collect stone.

Question: For each action in the list, are the requirements met?
Answer:
1. Priority 1: Move North (Requirement: Flat ground above the agent)
- Yes, the requirement is met. There is a flat ground above the agent (tree is 3 steps to the north).

2. Priority 2: Do (Requirement: Facing creature or material; have necessary tool)
- No, the requirement is not met yet. The player needs to move closer to the tree first (move North).

3. Priority 3: Move South (Requirement: Flat ground below the agent)
- Yes, the requirement is met. However, this action should only be performed after the player completes action 2.

4. Priority 4: Make Wood Pickaxe (Requirement: Nearby Table; wood in inventory)
- Yes, the requirement is met. The player has a nearby table and enough wood in inventory.

5. Priority 5: Move East (Requirement: Flat ground right to the agent)
- Yes, the requirement is met. Stone is 2 steps to the east, so there is flat ground to move east. However, this action should only be performed after the player completes action 4.

Question: Did the last player action succeed? Why?
Answer:
Yes, the last player action succeeded because the player had wood in their inventory and they were able to place the table in front of them (east).

Question: Choose the best executable action from above.
Answer:
The best executable action from above is Priority 1: Move North, as its requirements are currently met, and it is the first step in the suggested sequence of actions.



============Step: 15, Cumulative Reward: 0.0============
You took action move_north.

You see:
- stone 2 steps to your east
- path 2 steps to your north-east
- tree 2 steps to your north
- coal 4 steps to your north-east
- iron 3 steps to your north-east
- table 2 steps to your south-east

You face grass at your front (north).

Your status:
- health: 9/9
- food: 9/9
- drink: 9/9
- energy: 9/9

You have nothing in your inventory.
------------ QA ------------
Question: Action
Answer:
Move North

Question: List objects in the current observation. For each object, breifly answer what resource it provides and its requirement.
Answer:
1. Stone: Resource provided is stone, requirement: using a wood pickaxe to collect.
2. Path: Resource provided is a way to travel quickly, requirement: none.
3. Tree: Resource provided is wood, requirement: using hands or a tool to chop it down.
4. Coal: Resource provided is coal, requirement: using a wood pickaxe or better to collect.
5. Iron: Resource provided is iron, requirement: using a stone pickaxe or better to collect.
6. Table: Resource provided is a crafting station, requirement: placing it using wood collected from trees.
7. Grass: Resource provided is a flat surface to walk or build upon, requirement: none.

Question: What was the last action taken by the player?
Answer:
The last action taken by the player was "move_north".

Question: For each object in the list, are the requirements met for interaction?
Answer:
1. Stone: No, the requirement of having a wood pickaxe is not met as there is nothing in the inventory.
2. Path: Yes, the requirement of having no specific requirements is met.
3. Tree: Yes, you can interact using your hands to chop down the tree.
4. Coal: No, the requirement of having a wood pickaxe is not met as there is nothing in the inventory.
5. Iron: No, the requirement of having a stone pickaxe is not met as there is nothing in the inventory.
6. Table: No, you cannot interact with the table as it is 2 steps to your south-east.
7. Grass: Yes, you can interact with the grass as there are no specific requirements.

Question: List top 3 sub-tasks the player should follow. Indicate their priority out of 5.
Answer:
1. Chop down a tree to collect wood (Priority: 5/5) - Collecting wood is essential for crafting tools and building materials.

2. Craft a wood pickaxe at the table (Priority: 4/5) - This is important for collecting resources such as stone, coal, and iron.

3. Collect stone using the wood pickaxe (Priority: 4/5) - Stone is necessary for crafting stone tools and building structures.

Note: Player first needs to move 2 steps south-east to interact with the table for crafting.

Question: What are the requirements for the top sub-task? What should the player do first?
Answer:
Top sub-task: Chop down a tree to collect wood.

Requirements: 
1. Move closer to the tree (2 steps north).
2. Face the tree.
3. Use the proper interaction (chop down the tree).

The player should first move 2 steps north to get closer to the tree.

Question: List top 5 actions the player should take and the requirement for each action. Choose ONLY from the list of all actions. Indicate their priority out of 5.
Answer:
1. Priority 1: Move North (2 steps) - Requirement: Flat ground above the agent.
2. Priority 2: Face the tree - Requirement: In proximity to the tree.
3. Priority 3: Chop down tree (Do action) - Requirement: Facing the tree.
4. Priority 4: Move south (2 steps) - Requirement: Flat ground below the agent.
5. Priority 5: Make Wood Pickaxe - Requirement: Nearby table; wood in inventory.

Question: For each action in the list, are the requirements met?
Answer:
1. Priority 1: Move North (2 steps) - Yes, the requirement is met as there is flat ground above the agent.
2. Priority 2: Face the tree - No, the requirement is not yet met because the player must first move north to be in proximity to the tree.
3. Priority 3: Chop down tree (Do action) - No, the requirement is not met as the player is not yet facing the tree.
4. Priority 4: Move south (2 steps) - Yes, the requirement is met because there is flat ground below the agent.
5. Priority 5: Make Wood Pickaxe - No, the requirement is not met because the player does not yet have wood in their inventory.

Question: Did the last player action succeed? Why?
Answer:
Yes, the last player action "move_north" succeeded because there was flat ground (grass) in the direction the player wanted to move (north), which met the requirements for the action.

Question: Choose the best executable action from above.
Answer:
The best executable action from the list above is Priority 1: Move North (2 steps), as its requirement of having flat ground above the agent is met.



============Step: 16, Cumulative Reward: 0.0============
You took action move_north.

You see:
- grass 1 steps to your west
- stone 3 steps to your south-east
- path 1 steps to your east
- coal 3 steps to your north-east
- iron 2 steps to your north-east
- table 3 steps to your south-east

You face tree at your front (north).

Your status:
- health: 9/9
- food: 9/9
- drink: 9/9
- energy: 9/9

You have nothing in your inventory.
------------ QA ------------
Question: Action
Answer:
Do

Question: List objects in the current observation. For each object, breifly answer what resource it provides and its requirement.
Answer:
1. Grass: Provides no resources but can be traversed with no requirements.
2. Stone: Provides stone as a resource, requires a wood pickaxe or better to collect.
3. Path: Provides no resources, can be used to travel easily between locations with no requirements.
4. Tree: Provides wood as a resource, requires a wood axe or simply using hands to collect.
5. Coal: Provides coal as a resource, requires a wood pickaxe or better to collect.
6. Iron: Provides iron as a resource, requires a stone pickaxe or better to collect.
7. Table: Provides a crafting surface for tools and weapons, requires wood to construct.

Question: What was the last action taken by the player?
Answer:
The last action taken by the player was move_north.

Question: For each object in the list, are the requirements met for interaction?
Answer:
1. Grass: No requirements for interaction. Requirements met.
2. Stone: Requires a wood pickaxe or better to collect. Not met as the player has no pickaxe in their inventory.
3. Path: No requirements for interaction. Requirements met.
4. Tree: Requires a wood axe or simply using hands to collect. Requirements met, as the player can use their hands to chop wood.
5. Coal: Requires a wood pickaxe or better to collect. Not met as the player has no pickaxe in their inventory.
6. Iron: Requires a stone pickaxe or better to collect. Not met as the player has no pickaxe in their inventory.
7. Table: Requires wood to construct the table. Not met as the player has no wood in their inventory. However, there is an existing table in the player's observation from step 30, which can be used without any requirements.

Question: List top 3 sub-tasks the player should follow. Indicate their priority out of 5.
Answer:
1. Collect Wood (Priority 5): The player should interact with the tree to collect wood, which is a basic resource that will help craft tools and weapons.

2. Craft a Wood Pickaxe (Priority 4): Using the existing table and the collected wood, the player should craft a wood pickaxe. This will allow the player to mine stone, coal, and iron.

3. Collect Stone (Priority 3): Once the wood pickaxe is crafted, the player should mine stone, which is an essential resource for crafting more advanced tools and weapons.

Question: What are the requirements for the top sub-task? What should the player do first?
Answer:
Requirements for the top sub-task (Collect Wood):

1. Be in proximity to the tree.
2. Face the tree.

What should the player do first:

The player is currently facing a tree (from Player Observation Step 32). So, the player should take the action "Do" to interact with the tree and collect wood.

Question: List top 5 actions the player should take and the requirement for each action. Choose ONLY from the list of all actions. Indicate their priority out of 5.
Answer:
1. Priority 1: Do - Facing tree; interact with the tree to collect wood. (Requirement: Facing the tree)

2. Priority 2: Move South - Move closer to the table for crafting. (Requirement: Flat ground below the agent)

3. Priority 3: Move East - Position yourself next to the table for crafting. (Requirement: Flat ground right to the agent)

4. Priority 4: Make Wood Pickaxe - Craft a wood pickaxe for collecting stone. (Requirement: Nearby table; wood in inventory)

5. Priority 5: Move North-East - Head towards the stone for collecting. (Requirement: Flat ground diagonally above and to the right of the agent)

Question: For each action in the list, are the requirements met?
Answer:
1. Priority 1: Do - Facing tree; interact with the tree to collect wood. (Requirement: Facing the tree)
   - Yes, the player is facing the tree.

2. Priority 2: Move South - Move closer to the table for crafting. (Requirement: Flat ground below the agent)
   - Yes, there is most likely flat ground below the agent (table is 3 steps to the south-east).

3. Priority 3: Move East - Position yourself next to the table for crafting. (Requirement: Flat ground right to the agent)
   - Yes, there is flat ground right to the agent (path is 1 step to the east).

4. Priority 4: Make Wood Pickaxe - Craft a wood pickaxe for collecting stone. (Requirement: Nearby table; wood in inventory)
   - The player needs to complete Priority 1 to obtain wood, which will meet this requirement.

5. Priority 5: Move North-East - Head towards the stone for collecting. (Requirement: Flat ground diagonally above and to the right of the agent)
   - The player cannot verify if this requirement is met based on the given information. They should complete prior priorities first, which may reveal more information about the terrain.

Question: Did the last player action succeed? Why?
Answer:
Yes, the last player action (move_north) succeeded because there was a flat ground in the north direction which allowed the player to move to that location.

Question: Choose the best executable action from above.
Answer:
The best executable action from the list above is Priority 1: Do - Facing tree; interact with the tree to collect wood. (Requirement: Facing the tree)



============Step: 17, Cumulative Reward: 0.0============
You took action do.

You see:
- stone 3 steps to your south-east
- path 1 steps to your east
- tree 3 steps to your north-west
- coal 3 steps to your north-east
- iron 2 steps to your north-east
- table 3 steps to your south-east

You face grass at your front (north).

Your status:
- health: 9/9
- food: 9/9
- drink: 9/9
- energy: 9/9

Your inventory:
- wood: 1
------------ QA ------------
Question: Action
Answer:
Move South

Question: List objects in the current observation. For each object, breifly answer what resource it provides and its requirement.
Answer:
1. Stone: Provides stone resource, requires a wood pickaxe or higher-tier pickaxe to collect.
2. Path: No resource provided, serves as a flat ground for movement.
3. Tree: Provides wood resource, can be collected by interacting with it (if within range) without any tool requirement.
4. Coal: Provides coal resource, requires a wood pickaxe or higher-tier pickaxe to collect.
5. Iron: Provides iron resource, requires a stone pickaxe or higher-tier pickaxe to collect.
6. Table: No resource provided, serves as a crafting station; requires wood to place.

Question: What was the last action taken by the player?
Answer:
The last action taken by the player was "do."

Question: For each object in the list, are the requirements met for interaction?
Answer:
1. Stone: Yes, you have a wood pickaxe required to collect stone.
2. Path: Yes, you can move on the path as it serves as flat ground for movement.
3. Tree: Yes, you can interact with the tree without any tool requirement to collect wood.
4. Coal: Yes, you have a wood pickaxe required to collect coal.
5. Iron: No, you need to have a stone pickaxe or higher-tier pickaxe to collect iron.
6. Table: Yes, you can use the table for crafting, as you have the required wood in your inventory.

Question: List top 3 sub-tasks the player should follow. Indicate their priority out of 5.
Answer:
1. Priority 4: Craft a stone pickaxe - This will allow you to collect iron and other higher-tier resources. You already have enough wood and can collect stone with your current wood pickaxe.

2. Priority 3: Collect coal - Coal is essential for smelting iron in the furnace. You can collect coal with your current wood pickaxe.

3. Priority 2: Build a safe shelter - Constructing a shelter will protect you from monsters at night and provide a space for crafting, storing items, and resting. Find a suitable location and use wood, stone, or other materials for construction.

Question: What are the requirements for the top sub-task? What should the player do first?
Answer:
The top sub-task is to craft a stone pickaxe. The requirements for crafting a stone pickaxe are:

1. Wood (which you already have in your inventory)
2. Stone (which you need to collect using your wood pickaxe)
3. A crafting table placed nearby (which you have seen 3 steps towards the south-east)

To achieve the top sub-task, the player should first move near the crafting table (move south-east for 3 steps). Then, collect stone using the wood pickaxe (action "do" when facing stone). Finally, craft the stone pickaxe at the crafting table (action "make_stone_pickaxe" when facing the table).

Question: List top 5 actions the player should take and the requirement for each action. Choose ONLY from the list of all actions. Indicate their priority out of 5.
Answer:
Priority 1: Move South-East - Requirement: Flat ground to the south-east of the agent.
Priority 2: Move South-East - Requirement: Flat ground to the south-east of the agent.
Priority 3: Move South-East - Requirement: Flat ground to the south-east of the agent.
Priority 4: Do - Requirement: Facing stone and have the necessary tool (wood pickaxe).
Priority 5: Make Stone Pickaxe - Requirement: Facing the crafting table and have wood, and stone in the inventory.

Question: For each action in the list, are the requirements met?
Answer:
Priority 1: Move South-East - Requirement: Flat ground to the south-east of the agent. (YES, a path is observed 1 step to the east)

Priority 2: Move South-East - Requirement: Flat ground to the south-east of the agent. (Cannot verify because the player's observation after Priority 1 action has not been provided)

Priority 3: Move South-East - Requirement: Flat ground to the south-east of the agent. (Cannot verify, same reason as Priority 2)

Priority 4: Do - Requirement: Facing stone and have the necessary tool (wood pickaxe). (NO, player does not have the necessary tool, and they are not facing the stone)

Priority 5: Make Stone Pickaxe - Requirement: Facing the crafting table and have wood, and stone in the inventory. (NO, the player has not collected stone, and they are not facing the crafting table)

Question: Did the last player action succeed? Why?
Answer:
Yes, the last player action succeeded. The player was facing a tree at step 32 and took the action "do." As a result, the player was able to collect wood, which is now in the inventory as shown in the observation at step 34.

Question: Choose the best executable action from above.
Answer:
The best executable action from the list is Priority 1: Move South-East, as the requirement of having flat ground to the south-east of the agent is met with a path 1 step to the east.



============Step: 18, Cumulative Reward: 0.0============
You took action move_south.

You see:
- stone 2 steps to your east
- path 2 steps to your north-east
- tree 4 steps to your north-west
- coal 4 steps to your north-east
- iron 3 steps to your north-east
- table 2 steps to your south-east

You face grass at your front (south).

Your status:
- health: 9/9
- food: 9/9
- drink: 9/9
- energy: 9/9

Your inventory:
- wood: 1
------------ QA ------------
Question: Action
Answer:
Move East

Question: List objects in the current observation. For each object, breifly answer what resource it provides and its requirement.
Answer:
1. Stone: Provides stone resource, which can be used to craft tools, weapons, and build structures; requires a wood pickaxe or better to collect.
2. Path: Provides no resources, but serves as a marker for previously explored areas; no requirement.
3. Tree: Provides wood resource, which can be used to craft tools, weapons, and build structures; requires no specific tool for collection, but an axe is recommended for efficiency.
4. Coal: Provides coal resource, which can be used as fuel for smelting and crafting torches; requires a wood pickaxe or better to collect.
5. Iron: Provides iron resource, which can be used to craft stronger tools, weapons, and armor; requires a stone pickaxe or better to collect.
6. Table: Provides a crafting interface necessary for creating a wide range of tools, weapons, and other items; requires wood to craft.

Question: What was the last action taken by the player?
Answer:
The last action taken by the player was "move_south".

Question: For each object in the list, are the requirements met for interaction?
Answer:
1. Stone: No, the requirements are not met as you don't have a wood pickaxe or better in your inventory.
2. Path: Yes, the requirements are met as there are no specific requirements to interact with a path.
3. Tree: Yes, the requirements are met as you don't need a specific tool to collect wood, although using an axe would be more efficient.
4. Coal: No, the requirements are not met as you don't have a wood pickaxe or better in your inventory.
5. Iron: No, the requirements are not met as you don't have a stone pickaxe or better in your inventory.
6. Table: Yes, the requirements are met as you have wood in your inventory to craft a table. However, you would need to place the table down to interact with it further for crafting.

Question: List top 3 sub-tasks the player should follow. Indicate their priority out of 5.
Answer:
1. Craft a wood pickaxe (Priority: 5/5) - Crafting a wood pickaxe will allow you to collect important resources such as stone, coal, and iron. Since you have wood in your inventory and are close to a table, you can craft a wood pickaxe.

2. Collect stone (Priority: 4/5) - Once you have crafted a wood pickaxe, your next priority should be to gather stone. Stone is a valuable resource for crafting higher-tier tools, weapons, and building structures. You can then use collected stone to craft a stone pickaxe and other items.

3. Collect coal (Priority: 3/5) - After crafting a wood pickaxe and collecting stone, you should focus on collecting coal. Coal is essential for smelting metals (such as iron) and crafting torches, both of which will be important as you progress in the game.

Question: What are the requirements for the top sub-task? What should the player do first?
Answer:
The requirements for the top sub-task, crafting a wood pickaxe, are:

1. Having wood in your inventory: You already have 1 wood in your inventory.
2. Being near a table: The table is 2 steps to your south-east.

To complete this sub-task, you should first move to the table. To do that, take these steps:

1. Move east (facing east): The path 1 step to your east.
2. Move south-east (facing south): The table 1 step to your south.

Question: List top 5 actions the player should take and the requirement for each action. Choose ONLY from the list of all actions. Indicate their priority out of 5.
Answer:
1. Priority 1: Move East (Requirement: Flat ground right to the agent)
2. Priority 2: Move South-East (Requirement: Flat ground below and to the right of the agent)
3. Priority 3: Place Table (Requirement: Wood in inventory)
4. Priority 4: Make Wood Pickaxe (Requirement: Nearby table and wood in inventory)
5. Priority 5: Move North-East (Requirement: Flat ground above and to the right of the agent, to get closer to stone for future actions)

Question: For each action in the list, are the requirements met?
Answer:
1. Priority 1: Move East (Requirement: Flat ground right to the agent) - Yes, the requirement is met as there is a path 2 steps to the north-east, so directly to the east should be flat ground.
 
2. Priority 2: Move South-East (Requirement: Flat ground below and to the right of the agent) - We don't have enough information to determine if the requirement is met.

3. Priority 3: Place Table (Requirement: Wood in inventory) - Yes, the requirement is met as the player has 1 piece of wood in their inventory.

4. Priority 4: Make Wood Pickaxe (Requirement: Nearby table and wood in inventory) - Partially met. The player has wood in their inventory, but a table is not placed yet.

5. Priority 5: Move North-East (Requirement: Flat ground above and to the right of the agent, to get closer to stone for future actions) - Yes, the requirement is met as there is a path 2 steps to the north-east, so directly to the north-east should be flat ground.

Question: Did the last player action succeed? Why?
Answer:
Yes, the last player action "move_south" succeeded because there was flat ground (grass) in the direction the player moved (south), allowing the movement to happen without any obstacles.

Question: Choose the best executable action from above.
Answer:
The best executable action from the list is Priority 1: Move East, since its requirement (flat ground right to the agent) is met, and it aligns with the player's objectives of getting closer to stone and other resources.



============Step: 19, Cumulative Reward: 0.0============
You took action move_east.

You see:
- grass 1 steps to your west
- path 1 steps to your north
- tree 3 steps to your south
- coal 3 steps to your north
- iron 2 steps to your north
- table 1 steps to your south

You face stone at your front (east).

Your status:
- health: 9/9
- food: 9/9
- drink: 9/9
- energy: 9/9

Your inventory:
- wood: 1
------------ QA ------------
Question: Action
Answer:
Move South

Question: List objects in the current observation. For each object, breifly answer what resource it provides and its requirement.
Answer:
1. Grass: The grass provides no resources but can be walked on safely.
Requirement: None.

2. Stone: Stone can be used as a crafting material and provides a resource for crafting tools and shelters.
Requirement: You need a wood pickaxe or higher to collect stone.

3. Path: The path provides no resources but can indicate safe or commonly used routes for navigation.
Requirement: None.

4. Tree: Trees provide wood, which can be used for crafting many tools, weapons, and structures.
Requirement: You need an axe or your hands to collect wood.

5. Coal: Coal is a resource used for fuel in furnaces and for crafting torches.
Requirement: You need a wood pickaxe or higher to collect coal.

6. Iron: Iron is a resource used for crafting higher-tier tools, weapons, and armor.
Requirement: You need a stone pickaxe or higher to collect iron.

7. Table: A table is an essential crafting station for making tools, weapons, and other items.
Requirement: You need to collect wood to craft a table.

Note: The list provided only considers objects specifically mentioned in the observations. Other objects such as monsters, bridges, or tunnels were not included as they were not observed in the given steps.

Question: What was the last action taken by the player?
Answer:
The last action taken by the player was move_east.

Question: For each object in the list, are the requirements met for interaction?
Answer:
1. Grass: No requirements are needed for interaction with grass, so requirements are met.

2. Stone: A wood pickaxe or higher is required to collect stone. Currently, you do not have a wood pickaxe, so requirements are not met.

3. Path: No requirements are needed for interaction with the path, so requirements are met.

4. Tree: An axe or your hands are needed to collect wood. You can use your hands, so requirements are met.

5. Coal: A wood pickaxe or higher is required to collect coal. Currently, you do not have a wood pickaxe, so requirements are not met.

6. Iron: A stone pickaxe or higher is required to collect iron. Currently, you do not have a stone pickaxe, so requirements are not met.

7. Table: No requirements are needed for interaction with the table since you have already crafted one, so requirements are met.

Question: List top 3 sub-tasks the player should follow. Indicate their priority out of 5.
Answer:
1. Craft a Wood Pickaxe (Priority: 5/5)
To collect stone, coal, and iron, a wood pickaxe is necessary. Crafting a wood pickaxe requires using the table and the wood in your inventory.

2. Collect Stone (Priority: 4/5)
After crafting a wood pickaxe, you can collect stone. Stone is essential for crafting a stone pickaxe, which will eventually allow you to collect iron.

3. Craft a Stone Pickaxe (Priority: 3/5)
Once you have collected enough stone, you can return to the table and craft a stone pickaxe. This upgraded tool will enable you to collect iron, which is needed for crafting higher-tier items.

Question: What are the requirements for the top sub-task? What should the player do first?
Answer:
Requirements for the top sub-task (Craft a Wood Pickaxe):
1. Have wood in your inventory (you currently have 1 wood).
2. Be near a table (the table is 1 step to your south).

What the player should do first:
Move south to approach the table. Then, craft the wood pickaxe using the wood in your inventory and the table.

Question: List top 5 actions the player should take and the requirement for each action. Choose ONLY from the list of all actions. Indicate their priority out of 5.
Answer:
1. Priority 1: Move South (High priority)
   Requirement: Flat ground below the agent (which is true in the current situation)

2. Priority 2: Make Wood Pickaxe (High priority)
   Requirement: Nearby table; wood in inventory (you will be next to the table after step 1 and have 1 wood in your inventory)

3. Priority 3: Move North (Medium priority)
   Requirement: Flat ground above the agent (will be true after you move south)

4. Priority 4: Move East (Medium priority)
   Requirement: Flat ground on the right side of the agent (true in the current situation)

5. Priority 5: Do (Medium-Low priority)
   Requirement: Facing a creature or material (like stone); have the necessary tool (you'll have the wooden pickaxe after step 2 and be facing the stone after meeting requirements of step 3 and 4)

Question: For each action in the list, are the requirements met?
Answer:
1. Priority 1: Move South (High priority)
   Requirement: Flat ground below the agent
   Status: Requirement met (as seen in the latest player observation)

2. Priority 2: Make Wood Pickaxe (High priority)
   Requirement: Nearby table; wood in inventory
   Status: Requirement will be met after moving south (player will be next to the table and have 1 wood in inventory)

3. Priority 3: Move North (Medium priority)
   Requirement: Flat ground above the agent
   Status: Requirement will be met after moving south (as there is a path that was last seen 1 step to the north)

4. Priority 4: Move East (Medium priority)
   Requirement: Flat ground on the right side of the agent
   Status: Requirement met (as seen in the latest player observation)

5. Priority 5: Do (Medium-Low priority)
   Requirement: Facing a creature or material (like stone); have the necessary tool (wooden pickaxe)
   Status: Requirement will be met after making a wood pickaxe (step 2), and following steps 3 and 4

Question: Did the last player action succeed? Why?
Answer:
Yes, the last player action (move_east) succeeded. The player was able to move east because there was flat ground (grass) to the east of their previous position, allowing them to move in that direction.

Question: Choose the best executable action from above.
Answer:
The best executable action from the list above is Priority 1: Move South, as its requirement is already met in the current situation (flat ground below the agent).



============Step: 20, Cumulative Reward: 0.0============
You took action move_south.

You see:
- grass 1 steps to your west
- stone 1 steps to your east
- path 1 steps to your north
- tree 3 steps to your south
- coal 3 steps to your north
- iron 2 steps to your north

You face table at your front (south).

Your status:
- health: 9/9
- food: 9/9
- drink: 8/9
- energy: 9/9

Your inventory:
- wood: 1
------------ QA ------------
Question: Action
Answer:
Make Wood Pickaxe

Question: List objects in the current observation. For each object, breifly answer what resource it provides and its requirement.
Answer:
1. Grass: Provides no resources but can be used to traverse the land with no requirements.
2. Stone: Provides stone resources for crafting various tools like stone pickaxe, stone sword, and furnaces; requires a wood pickaxe to collect it.
3. Path: Provides a flat surface for walking, no resources are gained, and there are no requirements to use it.
4. Tree: Provides wood resources for crafting various tools like wood pickaxe, wood sword, and placing a table; requires no tools to collect wood.
5. Coal: Provides a fuel resource for smelting iron in the furnace; requires a wood pickaxe to collect it.
6. Iron: Provides iron resources for crafting advanced tools like iron pickaxe and iron sword; requires a stone pickaxe to collect it.
7. Table: Allows the crafting of different tools and weapons; requires wood to place one.

Question: What was the last action taken by the player?
Answer:
The last action taken by the player was move_south.

Question: For each object in the list, are the requirements met for interaction?
Answer:
1. Grass: No requirements for interaction; requirements met.
2. Stone: Requires a wood pickaxe to collect it; requirements not met as wood pickaxe is not in the current inventory.
3. Path: No requirements for walking on the path; requirements met.
4. Tree: No tools required to collect wood; requirements met.
5. Coal: Requires a wood pickaxe to collect it; requirements not met as wood pickaxe is not in the current inventory.
6. Iron: Requires a stone pickaxe to collect it; requirements not met as stone pickaxe is not in the current inventory.
7. Table: Already placed, so no requirements for interaction; requirements met.

Question: List top 3 sub-tasks the player should follow. Indicate their priority out of 5.
Answer:
1. Craft a wood pickaxe (Priority: 5): To be able to collect stone, coal, and improve your tools, you need to have a wood pickaxe. And since you have enough wood in your inventory and are close to a table, crafting one should be a top priority.

2. Collect coal (Priority: 4): Once you have a wood pickaxe, collecting coal is important, as it will be required to smelt iron in the furnace, which is necessary for crafting more advanced tools like iron pickaxe and iron sword.

3. Collect stone (Priority: 3): After crafting a wood pickaxe, collecting stone resources is necessary as it can be utilized for crafting a stone pickaxe, stone sword, and creating a furnace. Stone can also be used to build structures, such as bridges or walls, to improve your gameplay experience.

Question: What are the requirements for the top sub-task? What should the player do first?
Answer:
Requirements for the top sub-task (Craft a wood pickaxe):
1. Wood: The player needs to have wood in their inventory, which they already have (1 wood).
2. Table: The player needs a table nearby to craft the wood pickaxe.

Since the player's current observation shows that they are facing the table and they already have the required amount of wood in their inventory, the player can proceed to craft the wood pickaxe immediately. The player should first take the action "make_wood_pickaxe" at the table they are facing.

Question: List top 5 actions the player should take and the requirement for each action. Choose ONLY from the list of all actions. Indicate their priority out of 5.
Answer:
1. Priority 1 - Make Wood Pickaxe: Requirements - wood in inventory (1 wood), nearby table (facing table). This action is necessary for collecting stone.

2. Priority 2 - Move North: Requirements - flat ground above the agent. The player needs to move towards coal and iron to mine them.

3. Priority 3 - Move North: Repeated action - this will bring the player closer to coal and iron.

4. Priority 4 - Collect Coal: Requirements - wood pickaxe in player's possession, and facing coal. Coal is essential to smelting iron and crafting advanced tools.

5. Priority 5 - Move South: Requirements - flat ground below the agent. After collecting coal, moving south takes the player closer to iron which is crucial for crafting iron tools and weapons.

Question: For each action in the list, are the requirements met?
Answer:
1. Make Wood Pickaxe: Yes, the requirements are met. The player has 1 wood in their inventory and is currently facing the table.

2. Move North: Yes, the requirements are met. There is a path one step to the north, indicating a flat ground.

3. Move North: Yes, after successfully completing the previous "Move North" action, the requirements for this repeated action should be met. The player will be one step closer to coal and iron.

4. Collect Coal: No, the requirements are not met yet. The player needs a wood pickaxe, which they will craft in Action 1. Additionally, they need to be facing coal; they will need to move north twice to reach the coal.

5. Move South: No, the immediate requirement is not met, as this action depends on the player completing the previous actions first. They will first need to collect coal at step 4 before returning south towards the iron.

Question: Did the last player action succeed? Why?
Answer:
Yes, the last player action (move_south) succeeded. The player was able to move south because there was flat ground below the agent (in the south direction) and no obstacles in the way.

Question: Choose the best executable action from above.
Answer:
The best executable action from the list above is Action 1 - Make Wood Pickaxe, as its requirements are currently met: wood in inventory (1 wood) and facing table.



============Step: 21, Cumulative Reward: 1.0============
You took action make_wood_pickaxe.

You see:
- grass 1 steps to your west
- stone 1 steps to your east
- path 1 steps to your north
- tree 3 steps to your south
- coal 3 steps to your north
- iron 2 steps to your north

You face table at your front (south).

Your status:
- health: 9/9
- food: 9/9
- drink: 8/9
- energy: 9/9

Your inventory:
- wood_pickaxe: 1
------------ QA ------------
Question: Action
Answer:
Move East

Question: List objects in the current observation. For each object, breifly answer what resource it provides and its requirement.
Answer:
1. Grass: Provides no resources, can be walked on without any requirements.
2. Stone: Provides stone resource, requires a wood pickaxe or better to mine.
3. Path: Provides no resources, can be walked on without any requirements.
4. Tree: Provides wood resource, requires no tools to interact with, but having an axe speeds up the process.
5. Coal: Provides coal resource, requires a wood pickaxe or better to mine.
6. Iron: Provides iron resource, requires a stone pickaxe or better to mine.
7. Table: Provides a surface to craft items, requires wood to craft and place.
8. Wood_pickaxe (in inventory): Provides mining ability for stone and coal, requires wood to craft at a table.

Question: What was the last action taken by the player?
Answer:
The last action taken by the player was "make_wood_pickaxe."

Question: For each object in the list, are the requirements met for interaction?
Answer:
1. Grass: Yes, the requirements are met for interaction as it can be walked on without any requirements.
2. Stone: Yes, the requirements are met for interaction since you have a wood_pickaxe in your inventory, which is needed to mine stone.
3. Path: Yes, the requirements are met for interaction as it can be walked on without any requirements.
4. Tree: Yes, the requirements are met for interaction as no tools are required to interact with trees, and having an axe would only speed up the process.
5. Coal: Yes, the requirements are met for interaction since you have a wood_pickaxe in your inventory, which is needed to mine coal.
6. Iron: No, the requirements are not met for interaction, as mining iron requires a stone pickaxe or better, and you currently only have a wood_pickaxe.
7. Table: Yes, the requirements are met for interaction as you already have a table placed in front of you.
8. Wood_pickaxe (in inventory): Yes, the requirements are met for interaction as you've already crafted it at the table using wood.

Question: List top 3 sub-tasks the player should follow. Indicate their priority out of 5.
Answer:
1. Mine stone (Priority 5/5): Since you have a wood_pickaxe in your inventory, you should mine stone, which can be found 1 step to your east. Stone is essential for crafting stone tools, like a stone pickaxe and stone sword.

2. Mine coal (Priority 4/5): With a wood_pickaxe, you can mine coal, which is located 3 steps to your north. Coal is needed to smelt iron in a furnace, which will help you progress further in the game.

3. Craft a stone pickaxe (Priority 3/5): Once you have mined enough stone, return to the table and craft a stone pickaxe. A stone pickaxe will allow you to mine iron, which is essential for creating iron tools and weapons.

Question: What are the requirements for the top sub-task? What should the player do first?
Answer:
For the top sub-task of mining stone:

Requirements:
1. The player must have a wood_pickaxe in their inventory.

As the player already has a wood_pickaxe in their inventory, they can proceed with mining the stone.

The player should do the following first:

1. Move one step to the east to reach the stone.
2. Use the wood_pickaxe to mine the stone.

Question: List top 5 actions the player should take and the requirement for each action. Choose ONLY from the list of all actions. Indicate their priority out of 5.
Answer:
Priority 1: Move East
Requirement: Flat ground right to the agent.

Priority 2: Do
Requirement: Facing material (stone); have necessary tool (wood_pickaxe).

Priority 3: Move North
Requirement: Flat ground above the agent.

Priority 4: Do
Requirement: Facing material (coal); have necessary tool (wood_pickaxe).

Priority 5: Move North
Requirement: Flat ground above the agent.

Question: For each action in the list, are the requirements met?
Answer:
Priority 1: Move East
Requirement: Flat ground right to the agent.
- Yes, the requirement is met. There is stone 1 step to the east.

Priority 2: Do
Requirement: Facing material (stone); have necessary tool (wood_pickaxe).
- No, the requirement is not met. Though the player has a wood_pickaxe, they are not facing the stone.

Priority 3: Move North
Requirement: Flat ground above the agent.
- Yes, the requirement is met. There is a path 1 step to the north.

Priority 4: Do
Requirement: Facing material (coal); have necessary tool (wood_pickaxe).
- No, the requirement is not met. Though the player has a wood_pickaxe, they are not facing the coal.

Priority 5: Move North
Requirement: Flat ground above the agent.
- Yes, the requirement is met. There is a path 1 step to the north.

Question: Did the last player action succeed? Why?
Answer:
Yes, the last player action succeeded because the player had the required wood in their inventory and had access to a nearby table. This allowed the player to craft a wood_pickaxe.

Question: Choose the best executable action from above.
Answer:
The best executable action from the list above is Priority 1: Move East, as its requirement of having flat ground right to the agent is met.



============Step: 22, Cumulative Reward: 0.0============
You took action move_east.

You see:
- grass 1 steps to your west
- path 1 steps to your north
- tree 3 steps to your south
- coal 3 steps to your north
- iron 2 steps to your north
- table 1 steps to your south

You face stone at your front (east).

Your status:
- health: 9/9
- food: 9/9
- drink: 8/9
- energy: 9/9

Your inventory:
- wood_pickaxe: 1
------------ QA ------------
Question: Action
Answer:
Move North

Question: List objects in the current observation. For each object, breifly answer what resource it provides and its requirement.
Answer:
1. Grass: Provides no specific resources but can be used for landscape aesthetics; no requirement.
2. Stone: Provides stone resources used for crafting tools, weapons, and structures; requires a wood_pickaxe to collect.
3. Path: Provides easy navigation and movement; no requirement.
4. Tree: Provides wood resources used for crafting tools, weapons, and structures; requires an axe to collect.
5. Coal: Provides fuel for furnaces and torches; requires a wood_pickaxe to collect.
6. Iron: Provides iron resources used for crafting advanced tools, weapons, and structures; requires a stone_pickaxe to collect. 
7. Table: Crafting station for tools, weapons, and other items; requires wood to craft and place.

Question: What was the last action taken by the player?
Answer:
The last action taken by the player was move_east.

Question: For each object in the list, are the requirements met for interaction?
Answer:
1. Grass: No specific requirements for interaction; requirements met.
2. Stone: Requires a wood_pickaxe to collect; requirements met (wood_pickaxe in inventory).
3. Path: No specific requirements for interaction; requirements met.
4. Tree: Requires an axe to collect; requirements not met (no axe in inventory).
5. Coal: Requires a wood_pickaxe to collect; requirements met (wood_pickaxe in inventory).
6. Iron: Requires a stone_pickaxe to collect; requirements not met (no stone_pickaxe in inventory).
7. Table: Already placed and available for crafting; requirements met.

Question: List top 3 sub-tasks the player should follow. Indicate their priority out of 5.
Answer:
1. Collect Coal: High priority (5/5) - Collect coal for fuel in furnaces and creating torches. The player has the required wood_pickaxe in their inventory.
   
2. Collect Stone: High priority (4/5) - Collect stone resources to craft a stone_pickaxe and other stone tools, weapons, and structures. The player has the required wood_pickaxe in their inventory.
   
3. Craft Stone Tools (stone_pickaxe): Moderate priority (3/5) - Craft a stone_pickaxe after collecting enough stone for better mining capabilities and to collect iron. The player will need to collect stone first to meet the crafting requirements.

Question: What are the requirements for the top sub-task? What should the player do first?
Answer:
The top sub-task is to Collect Coal. Its requirements are:
1. Have a wood_pickaxe, which the player already has in their inventory.
2. Locate coal available in the environment.

The player should first locate and approach the coal in the environment. According to the latest observation, there is coal 3 steps to the player's north. To achieve this sub-task, the player should move north to get closer to the coal. Once they are in front of the coal, they can use their wood_pickaxe to collect it.

Question: List top 5 actions the player should take and the requirement for each action. Choose ONLY from the list of all actions. Indicate their priority out of 5.
Answer:
Here are the top 5 actions the player should take, along with their requirements and priority:

1. Priority 1 - Move North: (Requirement: Flat ground above the agent) 
    Action: Move north towards the coal, which is 3 steps away according to the latest observation.

2. Priority 2 - Move North: (Requirement: Flat ground above the agent) 
    Action: Move north once again to reach the coal, as it's still 2 steps away.

3. Priority 3 - Move North: (Requirement: Flat ground above the agent)
    Action: Move north once more to finally reach the coal and be in front of it.

4. Priority 4 - Do: (Requirement: Facing the coal and have a wood_pickaxe)
    Action: Use the wood_pickaxe to collect the coal, which will be essential for smelting iron.

5. Priority 5 - Move South: (Requirement: Flat ground below the agent)
    Action: Move back south towards the crafting table, which can be used for further crafting tasks.

These actions will ensure that the player collects the coal needed for the game progression and allow them to move on to other tasks, such as smelting iron and crafting more advanced tools and weapons.

Question: For each action in the list, are the requirements met?
Answer:
1. Priority 1 - Move North: (Requirement: Flat ground above the agent)
   Yes, the requirement is met as there is a path 1 step to the player's north, indicating flat ground.

2. Priority 2 - Move North: (Requirement: Flat ground above the agent)
   Based on the current observation, we don't have enough information regarding the second step to the north. It will depend on the new observation after the player follows the first action.

3. Priority 3 - Move North: (Requirement: Flat ground above the agent)
   Similar to the second action, we don't have enough information based on the current observation for the third step to the north. It will depend on new observations after following the first two actions.

4. Priority 4 - Do: (Requirement: Facing the coal and have a wood_pickaxe)
   Requirements for this action will be checked after the first three actions are completed. The player already has a wood_pickaxe in their inventory.

5. Priority 5 - Move South: (Requirement: Flat ground below the agent)
   This action should be possible once the player has reached the coal and collected it, as they'll be moving back to where they already navigated.

Question: Did the last player action succeed? Why?
Answer:
Yes, the last player action, move_east, succeeded because there was a flat ground (path) to the east of the player, allowing them to move in that direction.

Question: Choose the best executable action from above.
Answer:
The best executable action based on the current observation is Priority 1 - Move North, as the requirement of having flat ground above the agent is already met, with a path one step to the player's north.



============Step: 23, Cumulative Reward: 0.0============
You took action move_north.

You see:
- grass 1 steps to your west
- stone 2 steps to your south-east
- path 1 steps to your east
- tree 4 steps to your north-west
- coal 2 steps to your north
- table 2 steps to your south

You face iron at your front (north).

Your status:
- health: 9/9
- food: 9/9
- drink: 8/9
- energy: 9/9

Your inventory:
- wood_pickaxe: 1
------------ QA ------------
Question: Action
Answer:
Do

Question: List objects in the current observation. For each object, breifly answer what resource it provides and its requirement.
Answer:
- Grass: Provides saplings for planting trees; no requirement to interact.
- Stone: Provides stone for crafting and building; requires a wood pickaxe or better to collect.
- Path: No resource provided; helps with navigation and movement.
- Tree: Provides wood for crafting and building; requires any type of pickaxe to collect.
- Coal: Provides coal for smelting and cooking; requires a wood pickaxe or better to collect.
- Iron: Provides iron for crafting and smelting; requires a stone pickaxe or better to collect.
- Table: Used for crafting various tools and items; no requirement to interact, but you need wood to place it.

Question: What was the last action taken by the player?
Answer:
The last action taken by the player was move_north.

Question: For each object in the list, are the requirements met for interaction?
Answer:
- Grass: Yes, there are no requirements to interact with grass.
- Stone: Yes, the requirement to use a wood pickaxe or better is met, as you have a wood_pickaxe in your inventory.
- Path: Yes, there are no requirements to interact with the path.
- Tree: No, the requirement of having a pickaxe to collect wood is not met since you have a wood pickaxe (not suitable for chopping trees). You need an axe.
- Coal: Yes, the requirement to use a wood pickaxe or better is met, as you have a wood_pickaxe in your inventory.
- Iron: No, the requirement to use a stone pickaxe or better is not met since you have a wood_pickaxe in your inventory. You need a stone pickaxe or better.
- Table: Yes, there are no requirements to interact with the table.

Question: List top 3 sub-tasks the player should follow. Indicate their priority out of 5.
Answer:
1. Collect coal (Priority: 5/5) - You meet the requirement of having a wood_pickaxe, so you can collect coal. Coal will be useful for smelting iron and cooking food later on.

2. Collect stone (Priority: 4/5) - You meet the requirement of having a wood_pickaxe, so you can collect stone. Collecting stone allows you to craft a stone pickaxe, necessary to mine iron.

3. Make a stone pickaxe (Priority: 3/5) - Once you collect enough stone, you should make a stone pickaxe at the table. This will allow you to collect iron and progress further in the game.

Question: What are the requirements for the top sub-task? What should the player do first?
Answer:
The top sub-task is to collect coal. Here are the requirements for this task:

- Have a wood_pickaxe or better tool (which you already have in your inventory)

As the player is already facing coal and has a wood_pickaxe, they can proceed directly to the task. The player should perform the following action first:

- Take the action "do" to use the wood_pickaxe to collect coal.

Question: List top 5 actions the player should take and the requirement for each action. Choose ONLY from the list of all actions. Indicate their priority out of 5.
Answer:
1. Priority 1: Action "do" - Requirement: Facing creature or material; have necessary tool. (The player is currently facing iron and has a wood pickaxe, so they can use it to collect coal first.)

2. Priority 2: Action "make_iron_pickaxe" - Requirement: Nearby table, furnace; wood, coal, and iron in inventory. (After collecting coal, the player should gather iron to craft an iron pickaxe.)

3. Priority 3: Action "make_iron_sword" - Requirement: Nearby table, furnace; wood, coal, and iron in inventory. (Once the player has enough resources, they can craft an iron sword to defend against monsters more efficiently.)

4. Priority 4: Action "collect_diamond" - Requirement: Make Iron Pickaxe. (With an iron pickaxe, the player can look for and mine diamonds to progress further in the game.)

5. Priority 5: Action "move_south" - Requirement: Flat ground below the agent. (The player might need to explore new areas. This movement can be alternated with other directions - move_west, move_east, move_north - to find diamonds and other resources.)

Question: For each action in the list, are the requirements met?
Answer:
1. Priority 1: Action "do" - The requirement is met, as the player is currently facing iron and has a wood_pickaxe in their inventory. They can use the wood_pickaxe to collect the iron.

2. Priority 2: Action "make_iron_pickaxe" - The requirements are not yet met, as the player still needs to collect coal and iron. The player should focus on gathering coal and iron before attempting to make an iron pickaxe.

3. Priority 3: Action "make_iron_sword" - The requirements are not yet met, as the player currently has only a wood_pickaxe and needs to collect coal and iron to craft an iron sword.

4. Priority 4: Action "collect_diamond" - The requirement is not met, as the player needs to first create an iron pickaxe to mine diamonds.

5. Priority 5: Action "move_south" - The requirement is met, as there is flat ground below the agent. The player can move south to explore new areas or search for resources.

Question: Did the last player action succeed? Why?
Answer:
Yes, the last player action, move_north, succeeded because the player had a flat ground space available towards the north. This allowed the player to move in that direction without any hindrances.

Question: Choose the best executable action from above.
Answer:
The best executable action from the list above is Priority 1: Action "do" - Since the player is facing iron and has a wood_pickaxe in their inventory, they can use it to collect the iron.



============Step: 24, Cumulative Reward: 0.0============
You took action do.

You see:
- grass 1 steps to your west
- stone 2 steps to your south-east
- path 1 steps to your east
- tree 4 steps to your north-west
- coal 2 steps to your north
- table 2 steps to your south

You face iron at your front (north).

Your status:
- health: 9/9
- food: 9/9
- drink: 8/9
- energy: 9/9

Your inventory:
- wood_pickaxe: 1
------------ QA ------------
Question: Action
Answer:
Move North

Question: List objects in the current observation. For each object, breifly answer what resource it provides and its requirement.
Answer:
1. Grass: Provides no specific resources but is a flat ground useful for building or moving. No requirements.
2. Stone: Provides stone resource when mined with a wood pickaxe or better. Requires wood_pickaxe.
3. Path: Provides a flat surface to move or build upon. No requirements.
4. Tree: Provides wood resource when chopped using an appropriate tool. Requires wood pickaxe or better.
5. Coal: Provides coal resource for smelting when mined with a wood pickaxe or better. Requires wood_pickaxe.
6. Table: Allows crafting of tools and weapons. Requires wood to construct.
7. Iron: Provides iron resource when mined with a stone pickaxe or better. Requires stone_pickaxe.

Question: What was the last action taken by the player?
Answer:
The last action taken by the player was "do".

Question: For each object in the list, are the requirements met for interaction?
Answer:
1. Grass: No requirements to interact, so yes, the requirements are met.
2. Stone: Requires a wood pickaxe to mine. Since you have a wood_pickaxe in your inventory, the requirements are met.
3. Path: No requirements to interact, so yes, the requirements are met.
4. Tree: Requires a wood pickaxe or better to chop. Since you have a wood_pickaxe in your inventory, the requirements are met.
5. Coal: Requires a wood pickaxe or better to mine. Since you have a wood_pickaxe in your inventory, the requirements are met.
6. Table: No requirements to interact with, so yes, the requirements are met.
7. Iron: Requires a stone pickaxe or better to mine. You currently only have a wood_pickaxe in your inventory, so the requirements are not met.

Question: List top 3 sub-tasks the player should follow. Indicate their priority out of 5.
Answer:
1. Priority 4/5: Collect coal by mining it with the wood pickaxe. This will allow you to smelt iron once you have a stone pickaxe and a furnace.

2. Priority 5/5: Mine the stone with the wood pickaxe to craft a stone pickaxe. This is a high-priority task because it enables you to mine iron and advance your tools and weapons.

3. Priority 3/5: Chop a tree to gather more wood for crafting and building, such as creating a shelter for protection during the night.

Question: What are the requirements for the top sub-task? What should the player do first?
Answer:
For the top sub-task, which is to collect coal by mining it with the wood pickaxe, the requirements are:

1. A wood pickaxe in the inventory (already available).
2. Locate and face a coal block within reach.

The player should first move 2 steps north to get closer to the coal block, and then use the wood pickaxe to mine it by taking the "do" action.

Question: List top 5 actions the player should take and the requirement for each action. Choose ONLY from the list of all actions. Indicate their priority out of 5.
Answer:
1. Priority 1: Move North (2 steps) - Requirement: Flat ground above the agent.
2. Priority 2: Do - Requirement: Facing the coal block and having a wood_pickaxe in inventory.
3. Priority 3: Move South (2 steps) - Requirement: Flat ground below the agent.
4. Priority 4: Move West - Requirement: Flat ground left to the agent.
5. Priority 5: Make Stone Pickaxe - Requirement: Nearby table, wood and stone in inventory.

Question: For each action in the list, are the requirements met?
Answer:
1. Priority 1: Move North (2 steps) - Yes, the requirement (flat ground above the agent) is met as there is a coal block 2 steps to the north.
2. Priority 2: Do - Yes, the requirements (facing a coal block and having a wood_pickaxe in inventory) are met after moving north 2 steps.
3. Priority 3: Move South (2 steps) - Yes, the requirement (flat ground below the agent) is met as there are flat paths to take.
4. Priority 4: Move West - Yes, the requirement (flat ground left to the agent) is met as there is grass block 1 step to the west.
5. Priority 5: Make Stone Pickaxe - No, the requirement (nearby table, wood and stone in inventory) is not met, as the player has no stone in the inventory.

Question: Did the last player action succeed? Why?
Answer:
It's not clear if the last player action succeeded because the observations provided are identical before and after taking the action "do." The observation may not include the changed status or collected resources which could indicate the success or failure of the action.

Question: Choose the best executable action from above.
Answer:
The best executable action from the above list is Priority 1: Move North (2 steps), as its requirements are met and it brings the player closer to the coal block, which can be collected with the wood_pickaxe.



============Step: 25, Cumulative Reward: 0.0============
You took action move_north.

You see:
- grass 1 steps to your west
- stone 2 steps to your south-east
- path 1 steps to your east
- tree 4 steps to your north-west
- coal 2 steps to your north
- table 2 steps to your south

You face iron at your front (north).

Your status:
- health: 9/9
- food: 8/9
- drink: 8/9
- energy: 9/9

Your inventory:
- wood_pickaxe: 1
------------ QA ------------
Question: Action
Answer:
Move South

Question: List objects in the current observation. For each object, breifly answer what resource it provides and its requirement.
Answer:
1. Grass: Provides no resources, can be used as a flat surface to move on.
2. Stone: Provides stone resources, requires a wood_pickaxe or better to mine.
3. Path: Provides no resources, serves as a flat surface to move on.
4. Tree: Provides wood resources, requires no specific tools but mining speed may increase with an axe.
5. Coal: Provides coal resources, requires a wood_pickaxe or better to mine.
6. Table: Allows crafting of tools, weapons, and other items, requires wood to craft.
7. Iron: Provides iron resources, requires a stone_pickaxe or better to mine.

Question: What was the last action taken by the player?
Answer:
The last action taken by the player was "move_north."

Question: For each object in the list, are the requirements met for interaction?
Answer:
1. Grass: Yes, there are no specific requirements for interaction.
2. Stone: Yes, the wood_pickaxe in the player's inventory meets the requirement for mining stone.
3. Path: Yes, there are no specific requirements for interaction.
4. Tree: Yes, no specific tools are required to interact with the tree, though having an axe may speed up the process.
5. Coal: Yes, the wood_pickaxe in the player's inventory meets the requirement for mining coal.
6. Table: Yes, the table can be interacted with for crafting since it is already placed.
7. Iron: No, the requirement of a stone_pickaxe or better is not met as the player only has a wood_pickaxe in their inventory.

Question: List top 3 sub-tasks the player should follow. Indicate their priority out of 5.
Answer:
1. Craft a Stone Pickaxe (Priority 5): Mine stone (using the wood_pickaxe) and craft a stone_pickaxe at the table to meet the requirement for mining iron.

2. Mine Coal (Priority 4): Use the wood_pickaxe to mine the coal nearby, as coal is useful for smelting iron and creating better tools and weapons.

3. Mine and Collect Wood (Priority 3): Interact with the tree to collect wood resources, which are essential for crafting items, tools, and building structures.

Question: What are the requirements for the top sub-task? What should the player do first?
Answer:
The requirements for the top sub-task (Craft a Stone Pickaxe) are:

1. Mine Stone: Use the wood_pickaxe to mine nearby stone. The player sees stone 2 steps to their south-east.
2. Access a Table: There is a table 2 steps to the player's south. They need to be close to the table to craft a stone_pickaxe.

The player should first move towards the stone (south-east) and mine it using their wood_pickaxe to collect enough stone resources for crafting a stone_pickaxe.

Question: List top 5 actions the player should take and the requirement for each action. Choose ONLY from the list of all actions. Indicate their priority out of 5.
Answer:
1. Priority 1 - Move South-East: This action has a flat ground requirement. The player needs to move south-east to approach the stone visible 2 steps away in that direction.

2. Priority 2 - Do: Facing the creature or material and having the necessary tool are the requirements. In this case, the player should use their wood_pickaxe to mine the stone they approached in the previous step.

3. Priority 3 - Move South: Flat ground is required. Since the table is 2 steps to the player's south, moving south will bring the player closer to the table, which is needed for crafting.

4. Priority 4 - Make Stone Pickaxe: The requirements are that there must be a nearby table and that the player has wood and stone in their inventory. After collecting the stone in the previous steps, the player can now craft a stone_pickaxe while being close to the table.

5. Priority 5 - Drink: To restore the drink meter before it drops too low, the player needs to find a drinkable water source. This action requires a drink source like a lake or river within reach.

Question: For each action in the list, are the requirements met?
Answer:
1. Priority 1 - Move South-East: Yes, the requirement of flat ground is met as there is a path one step to the player's east and stone two steps to the player's south-east.

2. Priority 2 - Do: Not yet, as the player has not moved to the stone yet. Once the player moves to the stone, the requirements of facing the material (stone) and having the necessary tool (wood_pickaxe) will be met.

3. Priority 3 - Move South: Yes, the requirement of flat ground is met. The table being two steps to the player's south indicates that it is reachable.

4. Priority 4 - Make Stone Pickaxe: Not yet, as the player has not collected the stone yet. Once the player moves to the stone and collects it using the wood_pickaxe, the requirements of a nearby table and having wood and stone in the inventory will be met.

5. Priority 5 - Drink: No, the current observation does not provide information about a drinkable water source. The player needs to find a lake or river within reach before this requirement will be met.

Question: Did the last player action succeed? Why?
Answer:
Yes, the last player action of "move_north" succeeded. In Player Observation Step 50, we can see that the player has moved to a new position facing the iron to their north. The action succeeded because there was likely flat ground to the north allowing the player to move in that direction.

Question: Choose the best executable action from above.
Answer:
The best executable action from the list above is Priority 1 - Move South-East. The requirement of flat ground is met, and this action will bring the player closer to the stone they need to collect.



============Step: 26, Cumulative Reward: 0.0============
You took action move_south.

You see:
- grass 1 steps to your west
- stone 1 steps to your east
- path 1 steps to your north
- tree 3 steps to your south
- coal 3 steps to your north
- iron 2 steps to your north

You face table at your front (south).

Your status:
- health: 9/9
- food: 8/9
- drink: 8/9
- energy: 9/9

Your inventory:
- wood_pickaxe: 1
------------ QA ------------
Question: Action
Answer:
Move East

Question: List objects in the current observation. For each object, breifly answer what resource it provides and its requirement.
Answer:
1. Grass: Provides no specific resource. No requirements.
2. Stone: Provides stone resource. Requires a wood_pickaxe or better to collect.
3. Path: Provides no specific resource. No requirements.
4. Tree: Provides wood resource. Requires no specific tool or a wood_pickaxe or better to collect.
5. Coal: Provides coal resource. Requires a wood_pickaxe or better to collect.
6. Iron: Provides iron resource. Requires a stone_pickaxe or better to collect.
7. Table: Provides a crafting surface. Requires wood to create.

Question: What was the last action taken by the player?
Answer:
The last action taken by the player was move_south.

Question: For each object in the list, are the requirements met for interaction?
Answer:
1. Grass: Yes, the requirements are met for interaction. You can interact with grass without any specific tool.
2. Stone: Yes, the requirements are met as you have a wood_pickaxe in your inventory, which is suitable for collecting stone.
3. Path: Yes, the requirements are met for interaction. You can move along the path without any specific tool.
4. Tree: Yes, the requirements are met as you have a wood_pickaxe in your inventory, which is suitable for collecting wood.
5. Coal: Yes, the requirements are met as you have a wood_pickaxe in your inventory, which is suitable for collecting coal.
6. Iron: No, the requirements are not met because you need a stone_pickaxe or better to collect iron, but you only have a wood_pickaxe.
7. Table: Yes, the requirements are met for interaction, as you can use the table to craft items with the resources in your inventory.

Question: List top 3 sub-tasks the player should follow. Indicate their priority out of 5.
Answer:
1. Collect Stone (Priority: 5/5): Collect stone using the wood_pickaxe to enable crafting of stone tools, such as a stone_pickaxe, which is required to mine iron.

2. Craft Stone Pickaxe (Priority: 4/5): Craft a stone_pickaxe using the resources collected (wood and stone) at the table. The stone_pickaxe will allow you to mine iron.

3. Collect Coal (Priority: 3/5): Use the wood_pickaxe to collect coal, which will be essential for smelting iron and other ores in the future.

Question: What are the requirements for the top sub-task? What should the player do first?
Answer:
The requirements for the top sub-task (Collect Stone) are:

1. Have a wood_pickaxe (which the player already has in their inventory)

To execute the top sub-task, the player should do the following first:

1. Move towards the stone: The player should move east to reach the stone located one step to the east.

Question: List top 5 actions the player should take and the requirement for each action. Choose ONLY from the list of all actions. Indicate their priority out of 5.
Answer:
Priority 1: Move East -Requirement: Flat ground right to the agent. (To reach the stone)

Priority 2: Do -Requirement: Facing creature or material; have necessary tool. (To mine the stone using the wood_pickaxe)

Priority 3: Move West -Requirement: Flat ground left to the agent. (To return to the table)

Priority 4: Make Stone Pickaxe -Requirement: Nearby table; wood, stone in inventory. (To upgrade the pickaxe)

Priority 5: Move North -Requirement: Flat ground above of the agent. (To move towards the coal for future collection)

Question: For each action in the list, are the requirements met?
Answer:
Priority 1: Move East - Requirement: Flat ground right to the agent.
Yes, the requirement is met as there is a stone 1 step to the east.

Priority 2: Do - Requirement: Facing creature or material; have a necessary tool.
Since the player is not currently facing the stone (it is 1 step to the east), the requirement is not met yet. The player needs to complete priority 1 first to face the stone.

Priority 3: Move West - Requirement: Flat ground left to the agent.
Yes, the requirement is met as there is grass 1 step to the west. However, this action should be completed after collecting the stone (after completing priority 1 and 2).

Priority 4: Make Stone Pickaxe - Requirement: Nearby table; wood, stone in inventory.
Currently, the player has only a wooden pickaxe in their inventory but not the required materials to craft a stone pickaxe (wood and stone). The requirement is not met yet. The player needs to collect stone in priority 2.

Priority 5: Move North - Requirement: Flat ground above the agent.
Yes, the requirement is met as there is a path 1 step to the north. However, this action should be taken after completing priorities 1, 2, 3, and 4 to make the best use of resources and available tools.

Question: Did the last player action succeed? Why?
Answer:
Yes, the last player action succeeded. The player was able to move south because there was flat ground in the south direction, allowing the player to move from their previous position to the new one.

Question: Choose the best executable action from above.
Answer:
The best executable action from the list above is:

Priority 1: Move East - Requirement: Flat ground right to the agent.

This action can be executed because there is a stone 1 step to the east, which meets the requirement of having flat ground right to the agent.



============Step: 27, Cumulative Reward: 0.0============
You took action move_east.

You see:
- grass 1 steps to your west
- path 1 steps to your north
- tree 3 steps to your south
- coal 3 steps to your north
- iron 2 steps to your north
- table 1 steps to your south

You face stone at your front (east).

Your status:
- health: 9/9
- food: 8/9
- drink: 8/9
- energy: 9/9

Your inventory:
- wood_pickaxe: 1
------------ QA ------------
Question: Action
Answer:
Do

Question: List objects in the current observation. For each object, breifly answer what resource it provides and its requirement.
Answer:
1. Grass: Provides no specific resource, can be used as a general terrain for movement.
Requirement: None.

2. Stone: Provides stone resource, used for crafting tools, weapons, and structures.
Requirement: Wood pickaxe to collect.

3. Path: Provides no specific resource, serves as a ground indicator for previous player movement.
Requirement: None.

4. Tree: Provides wood resource, used for crafting tools, weapons, and structures.
Requirement: None, can be collected without a tool, but a wood pickaxe would be faster.

5. Coal: Provides coal resource, necessary for smelting iron in the furnace.
Requirement: Wood pickaxe to collect.

6. Iron: Provides iron resource, used for crafting advanced tools, weapons, and structures.
Requirement: Stone pickaxe to collect.

7. Table: Provides a crafting station for making tools, weapons, and structures in the game.
Requirement: Wood to craft and place.

Question: What was the last action taken by the player?
Answer:
The last action taken by the player was move_east.

Question: For each object in the list, are the requirements met for interaction?
Answer:
1. Grass: No specific requirements needed for interaction. Requirements met.

2. Stone: Requires a wood_pickaxe to collect. As a wood_pickaxe is available in the inventory, the requirements are met.

3. Path: No specific requirements needed for interaction. Requirements met.

4. Tree: No specific requirements to collect wood, but having a wood_pickaxe would be faster. A wood_pickaxe is available in the inventory, so the requirements are met.

5. Coal: Requires a wood_pickaxe to collect. A wood_pickaxe is available in the inventory, so requirements met.

6. Iron: Requires a stone_pickaxe to collect. The current inventory consists of a wood_pickaxe, so the requirements are not met.

7. Table: Already placed, no specific requirements needed for interaction. Requirements met.

Question: List top 3 sub-tasks the player should follow. Indicate their priority out of 5.
Answer:
1. Collect stone from the stone at the front (Priority: 5/5): Collecting stone will enable the player to create stone tools, such as a stone pickaxe, which is essential for gathering iron.

2. Use the table to craft a stone pickaxe (Priority: 4/5): Once the player has enough stone, crafting a stone pickaxe at the table is necessary to collect iron and progress further in the game.

3. Collect coal for smelting purposes (Priority: 3/5): Coal is required for smelting iron in the furnace. The coal is nearby, and the player has a wood_pickaxe to collect it. Collecting coal now will prevent backtracking later.

Question: What are the requirements for the top sub-task? What should the player do first?
Answer:
Requirements for the top sub-task (Collect stone from the stone at the front):
1. Have a wood pickaxe in the inventory.

The player already has a wood_pickaxe in their inventory, which meets the requirement for this sub-task.

The player should first take action "do" facing the stone at the front (east) with the wood_pickaxe to collect stone.

Question: List top 5 actions the player should take and the requirement for each action. Choose ONLY from the list of all actions. Indicate their priority out of 5.
Answer:
1. Priority 1: Action "Do": Facing the stone, and having a wood_pickaxe in the inventory. (Requirement met)
2. Priority 2: Action "Move North": Flat ground above the agent. (Requirement met)
3. Priority 3: Action "Do": Facing coal, and having a wood_pickaxe in the inventory. (Requirement met after moving north)
4. Priority 4: Action "Move North": Flat ground above the agent. (Requirement met after collecting coal)
5. Priority 5: Action "Do": Facing iron, and having a stone_pickaxe in the inventory. (Requirement will be met after crafting a stone_pickaxe using collected stone and wood)

Question: For each action in the list, are the requirements met?
Answer:
1. Priority 1: Action "Do": Facing the stone, and having a wood_pickaxe in the inventory. (Requirement met)
2. Priority 2: Action "Move North": Flat ground above the agent. (Requirement met)
3. Priority 3: Action "Do": Facing coal and having a wood_pickaxe in the inventory. (Requirement not met yet, agent needs to move north first)
4. Priority 4: Action "Move North": Flat ground above the agent. (Requirement not met yet, agent needs to move north and collect coal first)
5. Priority 5: Action "Do": Facing iron and having a stone_pickaxe in the inventory. (Requirement not met yet, agent needs to craft a stone_pickaxe using collected stone and wood)

Question: Did the last player action succeed? Why?
Answer:
Yes, the last player action succeeded because the player was able to move east without any obstacles, as there was a path in that direction.

Question: Choose the best executable action from above.
Answer:
The best executable action from above is Priority 1: Action "Do": Facing the stone, and having a wood_pickaxe in the inventory. (Requirement met)



============Step: 28, Cumulative Reward: 1.0============
You took action do.

You see:
- grass 1 steps to your west
- stone 2 steps to your south-east
- tree 3 steps to your south
- coal 3 steps to your north
- iron 2 steps to your north
- table 1 steps to your south

You face path at your front (east).

Your status:
- health: 9/9
- food: 8/9
- drink: 8/9
- energy: 9/9

Your inventory:
- stone: 1
- wood_pickaxe: 1
------------ QA ------------
Question: Action
Answer:
Make Stone Pickaxe

Question: List objects in the current observation. For each object, breifly answer what resource it provides and its requirement.
Answer:
1. Grass: Provides no resources; no requirements.
2. Path: Provides no resources; no requirements.
3. Tree: Provides wood for crafting tools and building structures; requires a wood_pickaxe to collect.
4. Coal: Provides fuel for smelting iron in a furnace; requires a wood_pickaxe to collect.
5. Iron: Provides material for crafting stronger tools and weapons; requires a stone_pickaxe to collect.
6. Stone: Provides material for crafting tools and building structures; requires a wood_pickaxe to collect.
7. Table: Provides a crafting station for creating tools, weapons, and other items; requires wood to craft and place.

Question: What was the last action taken by the player?
Answer:
The last action taken by the player was "do".

Question: For each object in the list, are the requirements met for interaction?
Answer:
1. Grass: No requirements for interaction; requirements met.
2. Path: No requirements for interaction; requirements met.
3. Tree: Requires a wood_pickaxe to collect; you have a wood_pickaxe in your inventory, so the requirements are met.
4. Coal: Requires a wood_pickaxe to collect; you have a wood_pickaxe in your inventory, so the requirements are met.
5. Iron: Requires a stone_pickaxe to collect; you don't have a stone_pickaxe in your inventory, so the requirements are not met.
6. Stone: Requires a wood_pickaxe to collect; you have a wood_pickaxe in your inventory, so the requirements are met.
7. Table: No requirements for interaction; requirements met.

Question: List top 3 sub-tasks the player should follow. Indicate their priority out of 5.
Answer:
1. Priority 4/5: Collect more stone by interacting with the stone 2 steps to your south-east, as it will help you craft a stone_pickaxe for mining iron.
2. Priority 3/5: Collect coal, 3 steps to your north, using your wood_pickaxe, as it's needed for fuel to smelt iron in a furnace.
3. Priority 2/5: Craft a stone_pickaxe at the table 1 step to your south, once you have collected enough stone. You will need this to mine iron.

Question: What are the requirements for the top sub-task? What should the player do first?
Answer:
The requirements for the top sub-task, which is collecting more stone, are:

1. The player must have a wood_pickaxe (already in the player's inventory).
2. The player should be facing the stone or be able to reach it.

The player should first move to the position of the stone, which is 2 steps to the south-east. Here's what the player should do:

1. Take action move_south to move closer to the stone.

Question: List top 5 actions the player should take and the requirement for each action. Choose ONLY from the list of all actions. Indicate their priority out of 5.
Answer:
1. Priority 1: Move South (Move_South) - Requirement: Flat ground below the agent.
2. Priority 2: Do (Do) - Requirement: Facing stone and have the necessary tool (wood_pickaxe) to collect it.
3. Priority 3: Move North (Move_North) - Requirement: Flat ground above the agent.
4. Priority 4: Make Stone Pickaxe (Make_Stone_Pickaxe) - Requirement: Nearby table and wood, stone in inventory.
5. Priority 5: Collect Iron (Do) - Requirement: Facing iron and have the necessary tool (stone_pickaxe) to collect it.

Question: For each action in the list, are the requirements met?
Answer:
1. Move South (Move_South) - Yes, the requirement is met. There is flat ground below the agent with a table 1 step to the south.
  
2. Do (Do) - Yes, the requirement is met. The agent is facing the path at their front (east), but they could face stone by moving south, and they have a wood_pickaxe in their inventory to collect the stone.

3. Move North (Move_North) - Yes, the requirement is met. There is flat ground above the agent with a path 1 step to the north.

4. Make Stone Pickaxe (Make_Stone_Pickaxe) - Yes, the requirements are met. The agent is next to a table (1 step to the south) and has wood and stone in their inventory.

5. Collect Iron (Do) - No, the requirement is not met. The agent must have a stone_pickaxe to collect iron, but they currently only have a wood_pickaxe in their inventory. They should first make a stone_pickaxe (Priority 4 action) before they can collect iron.

Question: Did the last player action succeed? Why?
Answer:
Yes, the last player action succeeded. The player used the "do" action while facing stone with a wood_pickaxe in their inventory, allowing them to successfully collect the stone (1 stone added to the inventory).

Question: Choose the best executable action from above.
Answer:
The best executable action from above is Priority 4: Make Stone Pickaxe (Make_Stone_Pickaxe), as it meets all the requirements: the agent is near a table (1 step to the south) and has wood and stone in their inventory.



============Step: 29, Cumulative Reward: 0.0============
You took action make_stone_pickaxe.

You see:
- grass 1 steps to your west
- stone 2 steps to your south-east
- tree 3 steps to your south
- coal 3 steps to your north
- iron 2 steps to your north
- table 1 steps to your south

You face path at your front (east).

Your status:
- health: 9/9
- food: 8/9
- drink: 8/9
- energy: 9/9

Your inventory:
- stone: 1
- wood_pickaxe: 1
------------ QA ------------
Question: Action
Answer:
Move North

Question: List objects in the current observation. For each object, breifly answer what resource it provides and its requirement.
Answer:
1. Grass: Provides saplings to plant for a steady food supply. Requires no additional tools to collect.
2. Stone: Provides stone for crafting stone tools, weapons, and building materials. Requires a wood_pickaxe or better to collect.
3. Tree: Provides wood for crafting tools, weapons, and building materials. Requires an axe or at least hands to collect.
4. Coal: Provides coal for smelting iron and cooking food. Requires a wood_pickaxe or better to collect.
5. Iron: Provides iron for crafting iron tools and weapons. Requires a stone_pickaxe or better to collect.
6. Table: Provides a crafting surface for making various tools and weapons. Requires wood to craft and place.

Question: What was the last action taken by the player?
Answer:
The last action taken by the player was "make_stone_pickaxe."

Question: For each object in the list, are the requirements met for interaction?
Answer:
1. Grass: Yes, the requirements are met for interaction as no additional tools are needed to collect saplings.
2. Stone: Yes, the requirements are met for interaction as you have a wood_pickaxe in your inventory.
3. Tree: Yes, the requirements are met for interaction as you can use your hands to collect wood (although having an axe would be more efficient).
4. Coal: Yes, the requirements are met for interaction as you have a wood_pickaxe in your inventory.
5. Iron: Yes, the requirements are met for interaction as you have crafted a stone_pickaxe.
6. Table: Yes, the requirements are met for interaction as the table is already placed and can be used for crafting.

Question: List top 3 sub-tasks the player should follow. Indicate their priority out of 5.
Answer:
1. Priority 5/5: Collect Iron - Collecting iron will allow you to upgrade your tools and weapons, making it easier to defend yourself and progress further in the game. Since you have crafted a stone_pickaxe, you can now collect iron.

2. Priority 4/5: Collect Coal - Coal is essential for smelting iron and cooking food. Collecting coal with your wood_pickaxe will help you smelt the iron you collect and improve your resources.

3. Priority 3/5: Collect Wood - Wood is a basic resource needed for crafting new tools, weapons, and building materials. Collecting more wood from the nearby tree will ensure you have enough to craft any necessary items as you progress.

Question: What are the requirements for the top sub-task? What should the player do first?
Answer:
Requirements for the top sub-task (Collect Iron):
- Make a Stone Pickaxe: You have already crafted a stone pickaxe in Player Observation Step 58.

What should the player do first:
- Move towards the iron: In the Player Observation Step 58, you see iron 2 steps to your north. You should move north twice to reach the iron.
- Collect the iron: Once you are adjacent to the iron, use your stone_pickaxe to collect it by taking action 'do'. This will add the iron resource to your inventory.

Question: List top 5 actions the player should take and the requirement for each action. Choose ONLY from the list of all actions. Indicate their priority out of 5.
Answer:
1. Priority 1: Move North (Requirement: Flat ground above the agent)
   - Move North twice to reach the iron.

2. Priority 2: Do (Requirement: Facing creature or material; have necessary tool)
   - Use the stone_pickaxe to collect the iron once you are adjacent to it.

3. Priority 3: Move North (Requirement: Flat ground above the agent)
   - Move North to get close to the coal.

4. Priority 4: Do (Requirement: Facing creature or material; have necessary tool)
   - Use the wood_pickaxe to collect the coal once you are adjacent to it.

5. Priority 5: Make Iron Pickaxe (Requirement: Nearby table, furnace; wood, coal, iron in inventory)
   - Craft an Iron Pickaxe once you have collected enough iron and coal, and are close to a table and furnace.

Question: For each action in the list, are the requirements met?
Answer:
1. Priority 1: Move North (Requirement: Flat ground above the agent)
   - Yes, the requirement is met, as there is flat ground above the agent.

2. Priority 2: Do (Requirement: Facing creature or material; have necessary tool)
   - The requirement is not met yet because the player needs to move north first and face the iron before using the stone_pickaxe to collect it.

3. Priority 3: Move North (Requirement: Flat ground above the agent)
   - Yes, the requirement is met, as there is flat ground above the agent.

4. Priority 4: Do (Requirement: Facing creature or material; have necessary tool)
   - The requirement is not met yet because the player needs to move north first and face the coal before using the wood_pickaxe to collect it.

5. Priority 5: Make Iron Pickaxe (Requirement: Nearby table, furnace; wood, coal, iron in inventory)
   - The requirement is not met yet because the player has not collected iron and coal. Once the player collects these resources and is near a table and furnace, this requirement will be met.

Question: Did the last player action succeed? Why?
Answer:
Based on the observations provided, we cannot determine if the last action taken by the player to "make_stone_pickaxe" was successful or not. The observation report in Step 58 is the same as in Step 56, but we would need information about any changes in the inventory or messages from the game (e.g., "stone_pickaxe" added to inventory or a "failed" message) to verify if the action was actually successful.

Question: Choose the best executable action from above.
Answer:
The best executable action from the list above is Priority 1: Move North (Requirement: Flat ground above the agent). The requirement is met, so the player can move closer to the iron and coal.



============Step: 30, Cumulative Reward: 0.0============
You took action move_north.

You see:
- grass 1 steps to your west
- stone 3 steps to your north-east
- path 1 steps to your east
- tree 4 steps to your north-west
- coal 2 steps to your north
- table 2 steps to your south

You face iron at your front (north).

Your status:
- health: 9/9
- food: 8/9
- drink: 8/9
- energy: 8/9

Your inventory:
- stone: 1
- wood_pickaxe: 1
------------ QA ------------
Question: Action
Answer:
Do

Question: List objects in the current observation. For each object, breifly answer what resource it provides and its requirement.
Answer:
1. Grass: Provides no specific resources, no requirement.
2. Stone: Provides stone, requires a wood_pickaxe or better to collect.
3. Path: Provides no resources, no requirement.
4. Tree: Provides wood, requires a wood_pickaxe or better to collect.
5. Coal: Provides coal, requires a wood_pickaxe or better to collect.
6. Table: Helps craft items, no requirement.
7. Iron: Provides iron, requires a stone_pickaxe or better to collect.

Question: What was the last action taken by the player?
Answer:
The last action taken by the player was move_north.

Question: For each object in the list, are the requirements met for interaction?
Answer:
1. Grass: No specific interaction is required, so the requirements are met.
2. Stone: Yes, you have a wood_pickaxe in your inventory, so the requirements are met.
3. Path: No specific interaction is required, so the requirements are met.
4. Tree: Yes, you have a wood_pickaxe in your inventory, so the requirements are met.
5. Coal: Yes, you have a wood_pickaxe in your inventory, so the requirements are met.
6. Table: No specific interaction is required, so the requirements are met.
7. Iron: Yes, you have a stone_pickaxe in your inventory, so the requirements are met.

Question: List top 3 sub-tasks the player should follow. Indicate their priority out of 5.
Answer:
1. Collect Iron (Priority: 5/5): Using the stone_pickaxe in the inventory, collect iron, which is essential for crafting better tools and weapons.
   
2. Collect Coal (Priority: 4/5): Use the wood_pickaxe in the inventory to collect coal, which will be valuable for smelting iron and other ores in the furnace.

3. Collect Wood (Priority: 3/5): Use the wood_pickaxe in the inventory to collect more wood from the tree. Wood is a versatile resource for crafting, building, and making essential tools.

Question: What are the requirements for the top sub-task? What should the player do first?
Answer:
For the top sub-task - Collect Iron:

Requirements:
- Have a Stone Pickaxe in the inventory (the player already has one)

What should the player do first:
- Use the Stone Pickaxe in the inventory to mine the iron located 1 step to the north. To do this, the player can take action "Do" since they are already facing the iron.

Question: List top 5 actions the player should take and the requirement for each action. Choose ONLY from the list of all actions. Indicate their priority out of 5.
Answer:
1. Priority: 1 - Action: Do
   Requirement: Facing iron and have a Stone Pickaxe in the inventory (the player is currently facing iron and has a Stone Pickaxe)

2. Priority: 2 - Action: Make Iron Pickaxe
   Requirement: Nearby table, furnace; wood, coal, and iron in inventory (player needs to collect iron and coal, and have a placed furnace)

3. Priority: 3 - Action: Make Iron Sword
   Requirement: Nearby table, furnace; wood, coal, and iron in inventory (player needs to have the collected iron smelted and a placed furnace)

4. Priority: 4 - Action: Place Furnace
   Requirement: Stone in inventory (the player has one stone, but may need more to craft a furnace)

5. Priority: 5 - Action: Do (for collecting coal)
   Requirement: Facing coal and have a Stone Pickaxe or Wood Pickaxe in inventory (the player has a Stone Pickaxe; they need to move to face the coal, which is 2 steps to the north)

Question: For each action in the list, are the requirements met?
Answer:
1. Priority: 1 - Action: Do (for collecting iron)
   Requirements met: Yes (the player is facing iron and has a Stone Pickaxe)

2. Priority: 2 - Action: Make Iron Pickaxe
   Requirements met: No (the player needs to collect iron and coal, and have a placed furnace)

3. Priority: 3 - Action: Make Iron Sword
   Requirements met: No (the player needs to have the collected iron smelted and a placed furnace)

4. Priority: 4 - Action: Place Furnace
   Requirements met: No (the player has one stone, but may need more to craft a furnace)

5. Priority: 5 - Action: Do (for collecting coal)
   Requirements met: No (the player has a Stone Pickaxe; they need to move to face the coal, which is 2 steps to the north)

Question: Did the last player action succeed? Why?
Answer:
Yes, the last player action (move_north) succeeded because there was a flat ground to the north of the player, allowing them to move in that direction.

Question: Choose the best executable action from above.
Answer:
The best executable action from the list above is:

Priority: 1 - Action: Do (for collecting iron)
Requirements met: Yes (the player is facing iron and has a Stone Pickaxe)



============Step: 31, Cumulative Reward: 0.0============
You took action do.

You see:
- grass 1 steps to your west
- stone 3 steps to your north-east
- path 1 steps to your east
- tree 4 steps to your north-west
- coal 2 steps to your north
- table 2 steps to your south

You face iron at your front (north).

Your status:
- health: 9/9
- food: 8/9
- drink: 8/9
- energy: 8/9

Your inventory:
- stone: 1
- wood_pickaxe: 1
------------ QA ------------
Question: Action
Answer:
Move West

Question: List objects in the current observation. For each object, breifly answer what resource it provides and its requirement.
Answer:
1. Grass: Provides saplings when collected, requires no tool.
2. Stone: Provides stone resource when collected, requires a wood pickaxe or a better tool.
3. Path: No resource provided, can be walked on without any tool.
4. Tree: Provides wood when collected, requires no tool.
5. Coal: Provides coal resource when collected, requires a wood pickaxe or a better tool.
6. Table: Provides crafting capability for tools and weapons, already placed and requires no tool.
7. Iron: Provides iron resource when collected, requires a stone pickaxe or a better tool.

Question: What was the last action taken by the player?
Answer:
The last action taken by the player was "do."

Question: For each object in the list, are the requirements met for interaction?
Answer:
1. Grass: Yes, the requirements are met for interaction as no tool is required.
2. Stone: Yes, the requirements are met for interaction since you have a wood_pickaxe in your inventory.
3. Path: Yes, the requirements are met for interaction as no tool is required to walk on the path.
4. Tree: Yes, the requirements are met for interaction as no tool is required to collect wood.
5. Coal: Yes, the requirements are met for interaction since you have a wood_pickaxe in your inventory.
6. Table: Yes, the requirements are met for interaction as the table is already placed and requires no tool.
7. Iron: No, the requirements are not met for interaction as you need a stone pickaxe or better to collect iron, but you only have a wood_pickaxe in your inventory.

Question: List top 3 sub-tasks the player should follow. Indicate their priority out of 5.
Answer:
1. Craft a stone pickaxe (Priority: 5/5): You need a stone pickaxe to be able to collect iron. Gather more stone using the wood pickaxe and then craft the stone pickaxe at the table.

2. Collect coal (Priority: 4/5): Since you have a wood pickaxe, you can collect coal which will be required later for smelting iron using a furnace.

3. Collect more wood (Priority: 3/5): Wood is an essential resource to make tools and weapons, as well as for building structures. Collect more wood from trees to keep a sufficient supply in your inventory.

Question: What are the requirements for the top sub-task? What should the player do first?
Answer:
The requirements for the top sub-task, "Craft a stone pickaxe," are as follows:

1. Collect enough stone: You need at least 3 stone blocks. You have 1 stone in your inventory, so you need to collect 2 more using the wood pickaxe.

2. Have a crafting table nearby: You have a table 2 steps to your south, which you can use to craft the stone pickaxe.

3. Have a wood pickaxe: You already have a wood_pickaxe in your inventory, which is required to collect stone.

The player should first collect 2 more stone blocks. To do this, move towards a stone block, face it, and use the "do" action with the wood pickaxe equipped to collect the stone.

Question: List top 5 actions the player should take and the requirement for each action. Choose ONLY from the list of all actions. Indicate their priority out of 5.
Answer:
1. Priority 1: Move West (Requirement: Flat ground left to the agent)
- You should move west to approach the stone that is 3 steps to your north-east. 

2. Priority 2: Move North (Requirement: Flat ground above of the agent)
- After moving west, move north to get closer to the stone for collecting it.

3. Priority 3: Move East (Requirement: Flat ground right to the agent)
- Move east to reach the stone block once you are closer after moving north.

4. Priority 4: Do (Facing creature or material; have necessary tool)
- Face the stone block and use the "do" action to mine it with your wood pickaxe. Repeat this step for two stone blocks.

5. Priority 5: Make Stone Pickaxe (Nearby table; wood, stone in inventory)
- After collecting enough stone, approach the table 2 steps to your south, and use the "make stone pickaxe" action to craft the tool.

Question: For each action in the list, are the requirements met?
Answer:
1. Priority 1: Move West (Requirement: Flat ground left to the agent)
- Yes, the requirement is met. There is grass 1 step to your west, representing flat ground to move onto.

2. Priority 2: Move North (Requirement: Flat ground above of the agent)
- Yes, the requirement is met. The flat ground is confirmed after you move west in the previous step.

3. Priority 3: Move East (Requirement: Flat ground right to the agent)
- Yes, the requirement is met. After moving west and then north, you have flat ground to the east with the stone block.

4. Priority 4: Do (Facing creature or material; have necessary tool)
- Yes, the requirement is met. You have the wood_pickaxe in your inventory, and you will be facing the stone block after previous movements.

5. Priority 5: Make Stone Pickaxe (Nearby table; wood, stone in inventory)
- Yes, the requirement is met. You will have enough stone in your inventory after mining, and the table is 2 steps to your south.

Question: Did the last player action succeed? Why?
Answer:
It is not explicitly stated whether the "do" action succeeded or not. However, since the inventory and status of the player did not change between Step 60 and Step 62, it is possible that the action did not succeed. This might be because the player does not have the necessary tool (a stone pickaxe or better) to collect iron, and they attempted to collect iron with a wood pickaxe which is not sufficient for that task.

Question: Choose the best executable action from above.
Answer:
The best executable action from the list above is Priority 1: Move West. The requirement for this action, flat ground to the agent's left, is met as there is grass 1 step to the player's west.



============Step: 32, Cumulative Reward: 0.0============
You took action move_west.

You see:
- stone 4 steps to your north-east
- path 1 steps to your east
- tree 3 steps to your north-west
- coal 3 steps to your north-east
- iron 2 steps to your north-east
- table 3 steps to your south-east

You face grass at your front (west).

Your status:
- health: 9/9
- food: 8/9
- drink: 8/9
- energy: 8/9

Your inventory:
- stone: 1
- wood_pickaxe: 1
------------ QA ------------
Question: Action
Answer:
Move North

Question: List objects in the current observation. For each object, breifly answer what resource it provides and its requirement.
Answer:
1. Stone: Provides resources to craft tools and build structures; requires a pickaxe (wood, stone, iron) to collect.
2. Path: Provides safe and clear walking space; no requirements.
3. Tree: Provides wood for crafting and construction; requires an axe (wood, stone, iron) to collect.
4. Coal: Provides fuel for smelting and crafting; requires a pickaxe (wood, stone, iron) to collect.
5. Iron: Provides resources for crafting stronger tools and weapons; requires a stone or iron pickaxe to collect.
6. Table: Allows crafting of tools and items; created using wood.
7. Grass: No direct resources, may contain hidden items or provide space for other objects; no requirements.

Question: What was the last action taken by the player?
Answer:
The last action taken by the player was "move_west."

Question: For each object in the list, are the requirements met for interaction?
Answer:
1. Stone: Yes, you have a wood_pickaxe in your inventory, which can be used to collect stone.
2. Path: Yes, there are no requirements for interacting with the path as it is already a clear walking space.
3. Tree: No, you don't have an axe in your inventory. You need a wood, stone, or iron axe to collect wood from the tree.
4. Coal: Yes, you have a wood_pickaxe in your inventory, which can be used to collect coal.
5. Iron: No, you need a stone or iron pickaxe to collect iron. Your current wood_pickaxe is not strong enough.
6. Table: Yes, the table is available for crafting items, and you have the required resources (wood or stone) in your inventory for crafting some tools.
7. Grass: Yes, you can interact with grass without any requirements. However, there are no direct resources provided by grass.

Question: List top 3 sub-tasks the player should follow. Indicate their priority out of 5.
Answer:
1. Collect more stone to craft a stone pickaxe (Priority: 5) - Collecting more stone will allow the player to upgrade their pickaxe, allowing them to gather resources such as iron more efficiently.

2. Collect coal for smelting and crafting (Priority: 4) - Having coal in the inventory will be essential for smelting iron later on and crafting items that require fuel.

3. Find a source of water to drink (Priority: 3) - The player's drink level is 8/9, and finding a water source will help replenish it to maintain their survival.

Question: What are the requirements for the top sub-task? What should the player do first?
Answer:
Top sub-task: Collect more stone to craft a stone pickaxe.

Requirements: The player must have a wood pickaxe in their inventory (which they already have) and must find stone to mine.

What should the player do first: The player should move towards the stone (4 steps to their north-east) and mine it using their wood pickaxe to collect more stone.

Question: List top 5 actions the player should take and the requirement for each action. Choose ONLY from the list of all actions. Indicate their priority out of 5.
Answer:
1. Priority 1: Move North-East (Move_NE)
   Requirement: Flat ground diagonally to the north-east of the agent.

2. Priority 2: Move North-East (Move_NE)
   Requirement: Flat ground diagonally to the north-east of the agent.

3. Priority 3: Move North-East (Move_NE)
   Requirement: Flat ground diagonally to the north-east of the agent.

4. Priority 4: Move North-East (Move_NE)
   Requirement: Flat ground diagonally to the north-east of the agent.

5. Priority 5: Do (Mine stone with the wood_pickaxe)
   Requirement: Facing the stone and have the necessary tool (wood_pickaxe) in the inventory.

Question: For each action in the list, are the requirements met?
Answer:
1. Priority 1: Move North-East (Move_NE)
   Requirement: Flat ground diagonally to the north-east of the agent.
   Met: Yes (path 1 step to the east and stone 4 steps to the north-east)

2. Priority 2: Move North-East (Move_NE)
   Requirement: Flat ground diagonally to the north-east of the agent.
   Met: Yes (after making the first move)

3. Priority 3: Move North-East (Move_NE)
   Requirement: Flat ground diagonally to the north-east of the agent.
   Met: Yes (after making the second move)

4. Priority 4: Move North-East (Move_NE)
   Requirement: Flat ground diagonally to the north-east of the agent.
   Met: Yes (after making the third move)

5. Priority 5: Do (Mine stone with the wood_pickaxe)
   Requirement: Facing the stone and have the necessary tool (wood_pickaxe) in the inventory.
   Met: Not yet (after making the fourth move, this condition will be satisfied).

Question: Did the last player action succeed? Why?
Answer:
Yes, the last player action "move_west" succeeded because there was flat ground (grass) to the west of the player, allowing them to move in that direction.

Question: Choose the best executable action from above.
Answer:
The best executable action from the list above is:

Priority 1: Move North-East (Move_NE)
Requirement: Flat ground diagonally to the north-east of the agent.



============Step: 33, Cumulative Reward: 0.0============
You took action move_north.

You see:
- stone 3 steps to your north-east
- path 2 steps to your south-east
- tree 2 steps to your north-west
- coal 2 steps to your north-east
- iron 1 steps to your east
- table 4 steps to your south-east

You face grass at your front (north).

Your status:
- health: 9/9
- food: 8/9
- drink: 8/9
- energy: 8/9

Your inventory:
- stone: 1
- wood_pickaxe: 1
------------ QA ------------
Question: Action
Answer:
Move East

Question: List objects in the current observation. For each object, breifly answer what resource it provides and its requirement.
Answer:
1. Stone: Provides stone resource; requires a wood_pickaxe or better to collect.
2. Path: Provides easier movement; no requirements to use.
3. Tree: Provides wood resource; requires an axe or hands to collect.
4. Coal: Provides coal resource; requires a wood_pickaxe or better to collect.
5. Iron: Provides iron resource; requires a stone_pickaxe or better to collect.
6. Table: Provides crafting support for making tools and weapons; requires wood to create.

Question: What was the last action taken by the player?
Answer:
The last action taken by the player was move_north.

Question: For each object in the list, are the requirements met for interaction?
Answer:
1. Stone: Yes, the requirements are met as you have wood_pickaxe in your inventory.
2. Path: Yes, the requirements are met as there are no specific requirements to use the path.
3. Tree: Yes, the requirements are met as you can collect wood with your hands, though having an axe would be more efficient.
4. Coal: Yes, the requirements are met as you have wood_pickaxe in your inventory.
5. Iron: Yes, the requirements are met as you have wood_pickaxe in your inventory, but you need a stone_pickaxe for more efficient and successful collection.
6. Table: Yes, the requirements are met as you have already placed a table 4 steps to your south-east.

Question: List top 3 sub-tasks the player should follow. Indicate their priority out of 5.
Answer:
1. Collect iron (Priority: 5/5): Since you are close to iron and have a wood_pickaxe, you should aim to collect the iron. However, you would want to craft a stone_pickaxe for more successful and efficient collection.

2. Craft a stone_pickaxe (Priority: 4/5): Before collecting the iron, it would be better to craft a stone_pickaxe. Since you have stone in your inventory, you can craft a stone_pickaxe at the table located 4 steps to your south-east.

3. Collect coal (Priority: 3/5): Coal is an important resource for smelting iron. Once you have the stone_pickaxe, you can collect coal more efficiently. The coal is 2 steps to your north-east.

Question: What are the requirements for the top sub-task? What should the player do first?
Answer:
The requirements for the top sub-task (collect iron) are:

1. Have a stone_pickaxe or better (currently, the player only has a wood_pickaxe).

To fulfill the requirement, the player should:

1. Craft a stone_pickaxe, which requires the player to be near the table.

So, the player should first move to the table located 4 steps to their south-east and then craft a stone_pickaxe using the stone they have in their inventory.

Question: List top 5 actions the player should take and the requirement for each action. Choose ONLY from the list of all actions. Indicate their priority out of 5.
Answer:
1. Priority 1: Move East (Requirement: Flat ground right to the agent)
2. Priority 2: Move South (Requirement: Flat ground below the agent)
3. Priority 3: Make Stone Pickaxe (Requirement: Nearby table; wood, stone in inventory)
4. Priority 4: Move North (Requirement: Flat ground above the agent)
5. Priority 5: Move East (Requirement: Flat ground right to the agent)

These actions will guide the player towards the table, craft a stone_pickaxe, and then move towards the iron so they can collect it.

Question: For each action in the list, are the requirements met?
Answer:
1. Priority 1: Move East (Requirement: Flat ground right to the agent) - Yes, the requirement is met, as the path is 2 steps to the south-east.
2. Priority 2: Move South (Requirement: Flat ground below the agent) - Yes, the requirement is met, as the path is 2 steps to the south-east.
3. Priority 3: Make Stone Pickaxe (Requirement: Nearby table; wood, stone in inventory) - No, the requirement is not met, as the table is 4 steps away to the south-east, and the player needs to move closer to the table first.
4. Priority 4: Move North (Requirement: Flat ground above the agent) - Yes, the requirement is met, as there is grass at the front (north).
5. Priority 5: Move East (Requirement: Flat ground right to the agent) - Yes, the requirement is met, as the path is 2 steps to the south-east.

Question: Did the last player action succeed? Why?
Answer:
Yes, the last player action (move_north) succeeded. This is because there was flat ground (grass) in front of the player, which was a valid and open space for the player to move into.

Question: Choose the best executable action from above.
Answer:
The best executable action from above is Priority 1: Move East since its requirement (flat ground right to the agent) is currently met.



============Step: 34, Cumulative Reward: 0.0============
You took action move_east.

You see:
- grass 1 steps to your west
- stone 3 steps to your north-east
- path 2 steps to your south-east
- tree 2 steps to your north-west
- coal 2 steps to your north-east
- table 4 steps to your south-east

You face iron at your front (east).

Your status:
- health: 9/9
- food: 8/9
- drink: 8/9
- energy: 8/9

Your inventory:
- stone: 1
- wood_pickaxe: 1
------------ QA ------------
Question: Action
Answer:
Move North

Question: List objects in the current observation. For each object, breifly answer what resource it provides and its requirement.
Answer:
1. Grass: Provides no resources. No requirements.
2. Stone: Provides stone resources. Requires a wood_pickaxe or better to collect.
3. Path: Provides no resources. No requirements.
4. Tree: Provides wood resources. Requires a wood_pickaxe or better to collect.
5. Coal: Provides coal resources. Requires a wood_pickaxe or better to collect.
6. Iron: Provides iron resources. Requires a stone_pickaxe or better to collect.
7. Table: Provides crafting functionality. Requires placing with wood in the inventory.

Question: What was the last action taken by the player?
Answer:
The last action taken by the player was move_east.

Question: For each object in the list, are the requirements met for interaction?
Answer:
1. Grass: There are no requirements for interaction, so the requirements are met.
2. Stone: Requires a wood_pickaxe or better to collect. You have a wood_pickaxe in your inventory, so the requirements are met.
3. Path: There are no requirements for interaction, so the requirements are met.
4. Tree: Requires a wood_pickaxe or better to collect. You have a wood_pickaxe in your inventory, so the requirements are met.
5. Coal: Requires a wood_pickaxe or better to collect. You have a wood_pickaxe in your inventory, so the requirements are met.
6. Iron: Requires a stone_pickaxe or better to collect. You only have a wood_pickaxe in your inventory, so the requirements are not met.
7. Table: Requires placing with wood in the inventory. In this observation, placement is not required as the table is already placed, so the requirements for interaction are met.

Question: List top 3 sub-tasks the player should follow. Indicate their priority out of 5.
Answer:
1. Collect Coal: Priority 4/5. Since you have a wood_pickaxe and are near coal, it's a good opportunity to collect the coal for later usage, such as smelting iron.

2. Collect more Stone: Priority 3/5. With the wood_pickaxe in your inventory, collecting more stone is possible, allowing you to craft a stone_pickaxe to collect iron later.

3. Move closer to the Table: Priority 2/5. As a crafting station, the table is important for future crafting activities, so being close to it can be helpful when you need to craft new items or tools.

Note that the priorities are subjective and might vary according to the player's preferences and goals. The top sub-tasks listed here are based on the current observation and resources available.

Question: What are the requirements for the top sub-task? What should the player do first?
Answer:
For the top sub-task of Collecting Coal:

Requirements:
- Have wood_pickaxe in your inventory (which you already have).
- Reach the coal (2 steps to your north-east).

The first thing the player should do is move towards the coal by heading north-east. To do this, follow these steps:
1. Move north.
2. Move east.

Question: List top 5 actions the player should take and the requirement for each action. Choose ONLY from the list of all actions. Indicate their priority out of 5.
Answer:
1. Priority 1: Move North (Action 3)
Requirement: Flat ground above the agent.

2. Priority 2: Move East (Action 2)
Requirement: Flat ground right to the agent.

3. Priority 3: Do (Action 5)
Requirement: Facing the coal; have a wood_pickaxe in inventory (which you already have).

4. Priority 4: Move to the table (Follow steps to reach the table)
Requirements: Flat ground in the direction of the table.

5. Priority 5: Make Iron Pickaxe (Action 13)
Requirements: Nearby table, furnace; wood, coal, and iron in inventory.

Question: For each action in the list, are the requirements met?
Answer:
1. Priority 1: Move North (Action 3)
Requirement: Flat ground above the agent.
- Yes, the requirement is met as there is a grass block to the north.

2. Priority 2: Move East (Action 2)
Requirement: Flat ground right to the agent.
- No, the requirement is not met as there is an iron block to the east.

3. Priority 3: Do (Action 5)
Requirement: Facing the coal; have a wood_pickaxe in inventory (which you already have).
- No, the requirement is not met as you are currently facing the iron block, not the coal.

4. Priority 4: Move to the table (Follow steps to reach the table)
Requirements: Flat ground in the direction of the table.
- To determine if the requirements are met, you need to assess the current state and surrounding environment. In the current state, the table is 4 steps to your south-east, which means, you need to move around to get to it. As of now, in the current state, the path to the table is not obstructed, and you can move south-east.

5. Priority 5: Make Iron Pickaxe (Action 13)
Requirements: Nearby table, furnace; wood, coal, and iron in inventory.
- No, the requirements are not met as you currently only have a wood_pickaxe and stone in your inventory. You will need coal and iron, as well as a furnace, to make the iron pickaxe.

Question: Did the last player action succeed? Why?
Answer:
Yes, the last player action succeeded because the player moved from their previous position to the east, where they are now facing iron. This indicates that there was flat ground to the east, allowing the player to move in that direction successfully.

Question: Choose the best executable action from above.
Answer:
The best executable action from above is Priority 1: Move North (Action 3) since the requirement of flat ground above the agent is met.



============Step: 35, Cumulative Reward: 0.0============
You took action move_north.

You see:
- stone 2 steps to your east
- path 3 steps to your south-east
- tree 1 steps to your west
- coal 1 steps to your east
- iron 2 steps to your south-east

You face grass at your front (north).

Your status:
- health: 9/9
- food: 8/9
- drink: 8/9
- energy: 8/9

Your inventory:
- stone: 1
- wood_pickaxe: 1
------------ QA ------------
Question: Action
Answer:
Move South

Question: List objects in the current observation. For each object, breifly answer what resource it provides and its requirement.
Answer:
1. Stone: Provides stone resource; requires a wood pickaxe or better to collect.
2. Path: Provides no resource; can be used for movement.
3. Tree: Provides wood resource; requires no tool to collect.
4. Coal: Provides coal resource; requires a wood pickaxe or better to collect.
5. Iron: Provides iron resource; requires a stone pickaxe or better to collect.
6. Grass: Provides no resource; can be used for movement.

Question: What was the last action taken by the player?
Answer:
The last action taken by the player was move_north.

Question: For each object in the list, are the requirements met for interaction?
Answer:
1. Stone: Yes, the requirement is met as you have a wood_pickaxe in your inventory.
2. Path: Yes, there are no requirements for interaction.
3. Tree: Yes, there are no tool requirements for collecting wood.
4. Coal: Yes, the requirement is met as you have a wood_pickaxe in your inventory.
5. Iron: No, the requirement of a stone_pickaxe is not met since you only have a wood_pickaxe.
6. Grass: Yes, there are no requirements for interaction.

Question: List top 3 sub-tasks the player should follow. Indicate their priority out of 5.
Answer:
1. Craft a stone pickaxe (Priority: 5/5): You have stone in your inventory, so you should move towards the table and craft a stone pickaxe to collect iron and other resources.

2. Collect more wood and stone (Priority: 4/5): To be able to craft more items and build structures, you need more wood and stone. You're close to a tree and there's stone nearby, so prioritize collecting these resources.

3. Collect coal (Priority: 3/5): With a wood_pickaxe in your inventory, you can collect coal. It will be useful for smelting iron and other ores in the future.

Question: What are the requirements for the top sub-task? What should the player do first?
Answer:
Requirements for crafting a stone pickaxe (top sub-task):
1. Move to a table: The table is mentioned to be 4 steps to your south-east in the player's observations.
2. Have stone in your inventory: You already have 1 stone in your inventory.
3. Have a wood pickaxe in your inventory: You already have a wood_pickaxe in your inventory.

What the player should do first:
1. Move towards the table by heading south-east to be within reach of the table for crafting. This can be done by following the path 3 steps to your south-east and then moving 1 step north-east.

Question: List top 5 actions the player should take and the requirement for each action. Choose ONLY from the list of all actions. Indicate their priority out of 5.
Answer:
1. Priority 1 - Move South: Flat ground below the agent.
2. Priority 2 - Move South: Flat ground below the agent.
3. Priority 3 - Move South: Flat ground below the agent.
4. Priority 4 - Move East: Flat ground right to the agent.
5. Priority 5 - Make Stone Pickaxe: Nearby table; wood, stone in inventory.

Question: For each action in the list, are the requirements met?
Answer:
1. Priority 1 - Move South: Yes, the flat ground is below the agent.
2. Priority 2 - Move South: Yes, the flat ground is below the agent.
3. Priority 3 - Move South: Yes, the flat ground is below the agent.
4. Priority 4 - Move East: Yes, the flat ground is right to the agent.
5. Priority 5 - Make Stone Pickaxe: No, the agent needs a nearby table and a stone in the inventory, but the table is not nearby (it is 4 steps to the south-east).

Question: Did the last player action succeed? Why?
Answer:
Yes, the last player action move_north succeeded because the player faced flat ground (grass) on the north side, which met the requirement for the move_north action.

Question: Choose the best executable action from above.
Answer:
The best executable action from the list is Priority 1 - Move South, as the requirement of having flat ground below the agent is met.



============Step: 36, Cumulative Reward: 0.0============
You took action move_south.

You see:
- stone 3 steps to your north-east
- path 2 steps to your south-east
- tree 2 steps to your north-west
- coal 2 steps to your north-east
- iron 1 steps to your east
- table 4 steps to your south-east

You face grass at your front (south).

Your status:
- health: 9/9
- food: 8/9
- drink: 8/9
- energy: 8/9

Your inventory:
- stone: 1
- wood_pickaxe: 1
------------ QA ------------
Question: Action
Answer:
Move East

Question: List objects in the current observation. For each object, breifly answer what resource it provides and its requirement.
Answer:
1. Stone: Provides stone resource, requires a wood_pickaxe or better to collect.
2. Path: Provides a place to walk, no requirements for interaction.
3. Tree: Provides wood resource, requires no specific tool to collect, but using an axe will speed up the process.
4. Coal: Provides coal resource, requires a wood_pickaxe or better to collect.
5. Iron: Provides iron resource, requires a stone_pickaxe or better to collect.
6. Table: Provides a crafting surface to create various tools and weapons, requires wood to build and place.
7. Grass: No specific resource provided, no requirements for interaction.

Question: What was the last action taken by the player?
Answer:
The last action taken by the player was move_south.

Question: For each object in the list, are the requirements met for interaction?
Answer:
1. Stone: Yes, the requirements are met since you have a wood_pickaxe in your inventory.
2. Path: Yes, no requirements to interact with the path.
3. Tree: Yes, no specific tool is required to collect wood from the tree, but having an axe would be more efficient.
4. Coal: Yes, the requirements are met since you have a wood_pickaxe in your inventory.
5. Iron: No, you need a stone_pickaxe or better to collect iron, but you currently only have a wood_pickaxe.
6. Table: Yes, you can interact with the table to craft tools and weapons if you have the necessary materials in your inventory.
7. Grass: Yes, no requirements to interact with the grass.

Question: List top 3 sub-tasks the player should follow. Indicate their priority out of 5.
Answer:
1. Priority 5: Craft a stone_pickaxe to collect iron. The current wood_pickaxe is not sufficient to collect the iron resource. Crafting a stone_pickaxe requires stone and wood, which are both available in the environment.

2. Priority 4: Collect coal using the wood_pickaxe. Coal is essential for fueling the furnace to smelt iron, which can then be used to create more advanced tools and weapons.

3. Priority 3: Gather wood from trees. Wood is a versatile resource for crafting numerous items, such as tools, weapons, and shelters. You will need it for crafting the stone_pickaxe and to create other essential items in the game.

Question: What are the requirements for the top sub-task? What should the player do first?
Answer:
The requirements for the top sub-task (Craft a stone_pickaxe to collect iron) are:

1. Collect stone: The player already has one stone in the inventory but requires an additional stone to craft a stone_pickaxe. The player should locate and collect another stone using the wood_pickaxe.

2. Have a nearby table: The player needs to stand close to a table to craft the stone_pickaxe. There is a table mentioned 4 steps to the south-east in the Player Observation Step 72. The player should approach the table.

The player should first move towards the stone, which is 3 steps to the north-east (from Player Observation Step 72), to collect it using the wood_pickaxe. Once the player has enough stone, they should head to the table to the south-east to craft the stone_pickaxe.

Question: List top 5 actions the player should take and the requirement for each action. Choose ONLY from the list of all actions. Indicate their priority out of 5.
Answer:
1. Priority 1: Move East (Requirement: Flat ground right to the agent) - The player should move east to approach the stone (3 steps to the northeast from Player Observation Step 72).

2. Priority 2: Move North (Requirement: Flat ground above the agent) - After moving east in step 1, the player should move north to reach the stone.

3. Priority 3: Do (Requirement: Facing creature or material; have the necessary tool) - With the wood_pickaxe in the inventory, the player should use the "Do" action to collect the additional stone needed.

4. Priority 4: Move South-East (Requirement: Flat ground to the south-east of the agent) - After collecting the stone, the player should move towards the table (4 steps to the southeast from Player Observation Step 72) to craft the stone_pickaxe.

5. Priority 5: Make Stone Pickaxe (Requirement: Nearby table; wood, stone in inventory) - Once the player is near the table and has the required wood and stone in the inventory, the player should craft a stone_pickaxe using the "Make Stone Pickaxe" action.

Question: For each action in the list, are the requirements met?
Answer:
1. Priority 1: Move East (Requirement: Flat ground right to the agent) - Yes, the requirement is met.

2. Priority 2: Move North (Requirement: Flat ground above the agent) - The requirement will be met after completing the first action (Move East).

3. Priority 3: Do (Requirement: Facing creature or material; have the necessary tool) - The requirement will be met after completing the first two actions (Move East and Move North).

4. Priority 4: Move South-East (Requirement: Flat ground to the south-east of the agent) - The requirement will be met after completing the first three actions (Move East, Move North, and Do).

5. Priority 5: Make Stone Pickaxe (Requirement: Nearby table; wood, stone in inventory) - The requirement will be met after completing the first four actions (Move East, Move North, Do, and Move South-East). The player will have the necessary stone and wood_pickaxe in their inventory and will be near the table.

Question: Did the last player action succeed? Why?
Answer:
Yes, the last player action (move_south) succeeded. There were no obstacles in the way, and the player faced grass (flat ground) at the front (south), allowing them to move in that direction.

Question: Choose the best executable action from above.
Answer:
The best executable action from the list is Priority 1: Move East (Requirement: Flat ground right to the agent) because the requirement is currently met.



============Step: 37, Cumulative Reward: 0.0============
You took action move_east.

You see:
- grass 1 steps to your west
- stone 3 steps to your north-east
- path 2 steps to your south-east
- tree 2 steps to your north-west
- coal 2 steps to your north-east
- table 4 steps to your south-east

You face iron at your front (east).

Your status:
- health: 9/9
- food: 8/9
- drink: 8/9
- energy: 8/9

Your inventory:
- stone: 1
- wood_pickaxe: 1
------------ QA ------------
Question: Action
Answer:
Move North

Question: List objects in the current observation. For each object, breifly answer what resource it provides and its requirement.
Answer:
1. Grass: Provides no resources, move to other locations with no requirements.
2. Stone: Provides stone resources, requires wood_pickaxe or better to collect.
3. Path: Provides no resources, move to other locations with no requirements.
4. Tree: Provides wood resources, requires interacting by hand (no tool requirement).
5. Coal: Provides coal resources, requires wood_pickaxe or better to collect.
6. Iron: Provides iron resources, requires stone_pickaxe or better to collect.
7. Table: Allows crafting of pickaxes and swords, requires wood to build.

Question: What was the last action taken by the player?
Answer:
The last action taken by the player was move_east.

Question: For each object in the list, are the requirements met for interaction?
Answer:
1. Grass: Yes, requirements met (no requirements).
2. Stone: Yes, requirements met (wood_pickaxe in inventory).
3. Path: Yes, requirements met (no requirements).
4. Tree: Yes, requirements met (no tool requirement).
5. Coal: Yes, requirements met (wood_pickaxe in inventory).
6. Iron: No, requirements not met (requires stone_pickaxe or better, but only have wood_pickaxe in inventory).
7. Table: N/A (in_player already has a table in their inventory; interaction not needed).

Question: List top 3 sub-tasks the player should follow. Indicate their priority out of 5.
Answer:
1. Priority 5: Collect wood from the tree (2 steps to the north-west) to craft a stone_pickaxe and other essential items.

2. Priority 4: Collect coal from the coal deposit (2 steps to the north-east) using the wood_pickaxe to fuel the furnace for smelting iron later.

3. Priority 3: Craft a stone_pickaxe after collecting enough wood and stone, which will allow you to mine the iron deposit seen in the current observation.

Question: What are the requirements for the top sub-task? What should the player do first?
Answer:
Requirements for the top sub-task (Collect wood from the tree):
- Move to the tree located 2 steps to the north-west.

What should the player do first:
- To reach the tree, the player should move 2 steps to the north-west: move_north, and then move_west.

Question: List top 5 actions the player should take and the requirement for each action. Choose ONLY from the list of all actions. Indicate their priority out of 5.
Answer:
1. Action: Move North (Priority: 1/5)
   Requirement: Flat ground above the agent.

2. Action: Move West (Priority: 2/5)
   Requirement: Flat ground left to the agent.

3. Action: Do (Priority: 3/5)
   Requirement: Facing the tree; have the necessary tool (wood_pickaxe in the inventory).

4. Action: Move South (Priority: 4/5)
   Requirement: Flat ground below the agent.

5. Action: Move East (Priority: 5/5)
   Requirement: Flat ground right to the agent.

Question: For each action in the list, are the requirements met?
Answer:
1. Action: Move North (Priority: 1/5)
   Requirement: Flat ground above the agent.
   Yes, the requirement is met.

2. Action: Move West (Priority: 2/5)
   Requirement: Flat ground left to the agent.
   Yes, the requirement is met.

3. Action: Do (Priority: 3/5)
   Requirement: Facing the tree; have the necessary tool (wood_pickaxe in the inventory).
   No, the requirement is not met as the agent is currently facing iron, not a tree.

4. Action: Move South (Priority: 4/5)
   Requirement: Flat ground below the agent.
   Yes, the requirement is met.

5. Action: Move East (Priority: 5/5)
   Requirement: Flat ground right to the agent.
   Yes, the requirement is met.

Question: Did the last player action succeed? Why?
Answer:
Yes, the last player action succeeded. The player moved to the east because there was a flat ground (grass) in that direction, allowing them to move without any obstacles.

Question: Choose the best executable action from above.
Answer:
The best executable action from above is:
Action: Move North (Priority: 1/5)
Requirement: Flat ground above the agent.



============Step: 38, Cumulative Reward: 0.0============
You took action move_north.

You see:
- stone 2 steps to your east
- path 3 steps to your south-east
- tree 1 steps to your west
- coal 1 steps to your east
- iron 2 steps to your south-east

You face grass at your front (north).

Your status:
- health: 9/9
- food: 8/9
- drink: 8/9
- energy: 8/9

Your inventory:
- stone: 1
- wood_pickaxe: 1
------------ QA ------------
Question: Action
Answer:
Move East

Question: List objects in the current observation. For each object, breifly answer what resource it provides and its requirement.
Answer:
1. Stone: Provides stone resource for crafting tools, weapons, and other structures; requires a wood pickaxe or better to be collected.
2. Path: No specific resource provided; aids in easier navigation and can be used as a marker to identify previously visited places.
3. Tree: Provides wood resource for crafting tools, shelters, and fuel; requires no special tool to be collected, but an axe or better can speed up the collection process.
4. Coal: Provides coal resource for fueling furnaces; requires a wood pickaxe or better to be collected.
5. Iron: Provides iron resource for crafting higher-level tools, weapons, and armor; requires a stone pickaxe or better to be collected.
6. Grass: No specific resource provided; can be used as a flooring material or to create a grassy environment.

Question: What was the last action taken by the player?
Answer:
The last action taken by the player was move_north.

Question: For each object in the list, are the requirements met for interaction?
Answer:
1. Stone: Yes, the requirements are met, as you have a wood_pickaxe in your inventory, which can be used to collect stone.
2. Path: Yes, the requirements are met, as there are no specific requirements to interact with a path; you can walk on it or follow it for navigation purposes.
3. Tree: Yes, the requirements are met, as there are no special tools needed to collect wood from the tree. However, using an axe would speed up the collection process, but you don't currently have one.
4. Coal: Yes, the requirements are met, as you have a wood_pickaxe in your inventory, which can be used to collect coal.
5. Iron: Yes, the requirements are met, as you have a wood_pickaxe in your inventory, which can be used to collect iron. However, a stone pickaxe or better is required to collect iron, so you need to craft a stone_pickaxe first.
6. Grass: Yes, the requirements are met, as there are no specific requirements to interact with grass; you can walk on it, dig it up, or use it for decoration.

Question: List top 3 sub-tasks the player should follow. Indicate their priority out of 5.
Answer:
1. Craft a Stone Pickaxe (Priority: 5): Since the player already has a wood pickaxe, the next step is to upgrade to a stone pickaxe for better efficiency in mining and to be able to collect iron. The player can use the stone pickaxe to further progress in the game, as it unlocks more crafting options.

2. Collect Coal (Priority: 4): Coal is essential for fueling furnaces, and the player has a wood pickaxe, which can be used to collect coal. By collecting coal, the player can smelt iron and other ores, which are necessary for crafting higher-tier tools, weapons, and armor.

3. Collect Iron (Priority: 3): The player will need to collect iron to craft advanced tools and weapons, which are essential for better survival chances and to progress in the game. However, the player must craft a stone pickaxe first to be able to collect iron.

Question: What are the requirements for the top sub-task? What should the player do first?
Answer:
The top sub-task is to craft a Stone Pickaxe. The requirements for this task are:

1. Collect Stone: The player needs to have enough stone in their inventory to craft a stone pickaxe. They already have 1 stone, but may need more, depending on the crafting recipe in the game.
2. Place Table (Crafting Table): The player needs to have a crafting table placed nearby to perform the crafting action for making a stone pickaxe.

What the player should do first:

1. If the player does not have a crafting table yet, they should collect enough wood and craft a table (crafting table) first, and then place it in a convenient location.
2. If more stone is required, the player should use their wood_pickaxe to mine additional stone.
3. Once the player has enough stone and access to a crafting table, they should craft the stone pickaxe.

Question: List top 5 actions the player should take and the requirement for each action. Choose ONLY from the list of all actions. Indicate their priority out of 5.
Answer:
Based on the most recent player observation (Step 76), here are the top 5 actions the player should take and the requirements for each action. I have also indicated their priority out of 5 (1 being the highest priority).

1. Priority 1: Collect Coal
   Action: Do
   Requirement: Facing coal; have a wood_pickaxe in the inventory.
   Reason: To smelt iron and eventually craft an iron pickaxe and sword.

2. Priority 2: Move East
   Action: Move East
   Requirement: Flat ground right of the agent (east).
   Reason: To be closer to the coal, stone, and iron, which are needed for crafting better tools and weapons.

3. Priority 3: Collect Stone
   Action: Do
   Requirement: Facing stone; have wood_pickaxe in the inventory.
   Reason: To have enough stone to craft a stone pickaxe and place a furnace, which are needed for later progress in the game.

4. Priority 4: Place Table (Crafting Table)
   Action: Place Table
   Requirement: Wood in inventory.
   Reason: To perform the crafting action for making a stone pickaxe and other tools and weapons. Note: If the player doesn't have wood, they should first collect wood.

5. Priority 5: Craft Stone Pickaxe
   Action: Make Stone Pickaxe
   Requirement: Nearby table (crafting table); wood, stone in inventory.
   Reason: To collect iron and progress further in the game.

Question: For each action in the list, are the requirements met?
Answer:
1. Priority 1: Collect Coal
   Action: Do
   Requirement: Facing coal; have a wood_pickaxe in the inventory.
   Are the requirements met? Partially. The player has a wood_pickaxe in their inventory, but they are not currently facing coal. They need to move east first to face the coal.

2. Priority 2: Move East
   Action: Move East
   Requirement: Flat ground right of the agent (east).
   Are the requirements met? Yes. There is no indication of obstacles to the east of the player so they can move east.

3. Priority 3: Collect Stone
   Action: Do
   Requirement: Facing stone; have wood_pickaxe in the inventory.
   Are the requirements met? Partially. The player has a wood_pickaxe in their inventory, but they are not facing stone. They need to move east first to approach the stone.

4. Priority 4: Place Table (Crafting Table)
   Action: Place Table
   Requirement: Wood in inventory.
   Are the requirements met? No. The player does not currently have wood in their inventory. They will need to collect wood before placing a table.

5. Priority 5: Craft Stone Pickaxe
   Action: Make Stone Pickaxe
   Requirement: Nearby table (crafting table); wood, stone in inventory.
   Are the requirements met? No. The player needs to have a nearby crafting table, as well as both wood and stone in their inventory. Currently, they only have one stone and no wood in their inventory. They also need to place a table first to meet the "nearby table" requirement.

Question: Did the last player action succeed? Why?
Answer:
Yes, the last player action (move_north) succeeded. It succeeded because there was flat ground (grass) in the direction the player was moving (north), allowing the player to move there.

Question: Choose the best executable action from above.
Answer:
The best executable action from the list above is:

Priority 2: Move East
Action: Move East
Requirement: Flat ground right of the agent (east).

This action is executable because there is no indication of obstacles to the east of the player, so they can move east. Moving east will also bring the player closer to the coal, stone, and iron, which are needed for crafting better tools and weapons.



============Step: 39, Cumulative Reward: 0.0============
You took action move_east.

You see:
- grass 1 steps to your north
- stone 2 steps to your east
- path 3 steps to your south-east
- tree 1 steps to your west
- iron 2 steps to your south-east

You face coal at your front (east).

Your status:
- health: 9/9
- food: 8/9
- drink: 8/9
- energy: 8/9

Your inventory:
- stone: 1
- wood_pickaxe: 1
------------ QA ------------
Question: Action
Answer:
Do

Question: List objects in the current observation. For each object, breifly answer what resource it provides and its requirement.
Answer:
1. Grass: Provides no specific resources but can be used as flat ground for movement or building. No requirements.
2. Stone: Provides stone resources, can be used in crafting tools and building structures. Requires a wood_pickaxe or better to mine.
3. Path: Provides a path for easier movement without obstructions. No requirements.
4. Tree: Provides wood resources, can be used in crafting various items and building shelters. Requires a wood_pickaxe or better to obtain wood.
5. Coal: Provides coal resources, can be used as fuel for smelting iron in a furnace. Requires a wood_pickaxe or better to mine.
6. Iron: Provides iron resources, can be used in crafting higher-tier tools and weapons. Requires a stone_pickaxe or better to mine.

Question: What was the last action taken by the player?
Answer:
The last action taken by the player was move_east.

Question: For each object in the list, are the requirements met for interaction?
Answer:
1. Grass: Yes, there are no requirements for interacting with grass.
2. Stone: Yes, you have a wood_pickaxe in your inventory which meets the requirement to mine stone.
3. Path: Yes, there are no requirements for using the path.
4. Tree: Yes, you have a wood_pickaxe in your inventory which meets the requirement to obtain wood from the tree.
5. Coal: Yes, you have a wood_pickaxe in your inventory which meets the requirement to mine coal.
6. Iron: No, you need a stone_pickaxe or better to mine iron, and you currently only have a wood_pickaxe in your inventory.

Question: List top 3 sub-tasks the player should follow. Indicate their priority out of 5.
Answer:
1. Craft a stone_pickaxe: Priority 5/5 - Crafting a stone_pickaxe will allow you to mine iron, which is essential for upgrading your tools and weapons.

2. Mine more stone: Priority 4/5 - Mining more stone will provide resources required to craft a stone_pickaxe and other stone tools or building structures.

3. Mine coal: Priority 3/5 - Mining coal will provide fuel for smelting iron in a furnace. While not immediately essential, it will be necessary once you have collected iron.

Question: What are the requirements for the top sub-task? What should the player do first?
Answer:
For the top sub-task (Craft a stone_pickaxe), the requirements are:

- Have stone in the inventory (at least 3 pieces)
- Have wood in the inventory (at least 1 piece for the handle)
- Be near a table

The player has 1 stone in their inventory. They should first mine more stone (2 additional pieces) to meet the requirements of crafting a stone_pickaxe.

Question: List top 5 actions the player should take and the requirement for each action. Choose ONLY from the list of all actions. Indicate their priority out of 5.
Answer:
1. Priority 1: Do (mine coal) - Requirement: Facing coal and have a wood_pickaxe in the inventory. (Player has wood_pickaxe and is currently facing coal)

2. Priority 2: Move East - Requirement: Flat ground to the east of the player.

3. Priority 3: Do (mine stone) - Requirement: Facing stone and have a wood_pickaxe in the inventory.

4. Priority 4: Move West - Requirement: Flat ground to the west of the player.

5. Priority 5: Move South - Requirement: Flat ground to the south of the player.

Question: For each action in the list, are the requirements met?
Answer:
1. Priority 1: Do (mine coal) - Yes, the requirements are met. The player has a wood_pickaxe in their inventory, and they're currently facing coal.

2. Priority 2: Move East - Yes, the requirement is met. There is flat ground (stone) to the east of the player.

3. Priority 3: Do (mine stone) - No, the requirement is not met yet. The player is not facing stone at the moment. They need to move east first.

4. Priority 4: Move West - Yes, the requirement is met. The player can see flat ground (tree) to the west.

5. Priority 5: Move South - Yes, the requirement is met. There is flat ground to the south of the player.

Question: Did the last player action succeed? Why?
Answer:
Yes, the last player action (move_east) succeeded because there was flat ground to the east of the player, allowing them to move in that direction.

Question: Choose the best executable action from above.
Answer:
The best executable action from above is Priority 1: Do (mine coal). The requirements are met, as the player has a wood_pickaxe in their inventory and is currently facing coal.



============Step: 40, Cumulative Reward: 1.0============
You took action do.

You see:
- grass 1 steps to your north
- stone 2 steps to your east
- tree 1 steps to your west
- coal 3 steps to your south-east
- iron 2 steps to your south-east

You face path at your front (east).

Your status:
- health: 9/9
- food: 8/9
- drink: 8/9
- energy: 8/9

Your inventory:
- stone: 1
- coal: 1
- wood_pickaxe: 1
------------ QA ------------
Question: Action
Answer:
Move East

Question: List objects in the current observation. For each object, breifly answer what resource it provides and its requirement.
Answer:
1. Grass: Resource - Saplings for planting, Requirement - None.
2. Stone: Resource - Stone for crafting tools and building materials, Requirement - Wood pickaxe or stronger.
3. Path: Resource - None, Requirement - None (it is a traversable terrain).
4. Tree: Resource - Wood for crafting tools and building materials, Requirement - None.
5. Coal: Resource - Coal for smelting and fuel, Requirement - Wood pickaxe or stronger.
6. Iron: Resource - Iron for crafting advanced tools and weapons, Requirement - Stone pickaxe or stronger.

Question: What was the last action taken by the player?
Answer:
The last action taken by the player was "do."

Question: For each object in the list, are the requirements met for interaction?
Answer:
1. Grass: Yes, the requirements are met for interaction as grass can be collected with hands or any tool.
2. Stone: Yes, the requirements are met for interaction since you have a wood_pickaxe in your inventory.
3. Path: Yes, the requirements are met for interaction as it is a traversable terrain and allows you to move.
4. Tree: Yes, the requirements are met for interaction as trees can be chopped down with hands or any tool.
5. Coal: Yes, the requirements are met for interaction as you have a wood_pickaxe in your inventory.
6. Iron: No, the requirements are not met for interaction as you need a stone pickaxe or stronger to mine iron, and you only have a wood_pickaxe in your inventory.

Question: List top 3 sub-tasks the player should follow. Indicate their priority out of 5.
Answer:
1. Craft a Stone Pickaxe (Priority: 5/5) - Since you already have stone and a wood pickaxe in your inventory, crafting a stone pickaxe will allow you to collect iron and other advanced resources.

2. Collect more Stone (Priority: 4/5) - Gathering more stone will enable you to craft additional tools, weapons, and building materials, like stone swords and furnaces.

3. Collect Wood (Priority: 3/5) - Wood is an essential resource for crafting and building. Having more wood in your inventory will help you create tools, weapons, and shelters as needed.

Question: What are the requirements for the top sub-task? What should the player do first?
Answer:
Top Sub-task: Craft a Stone Pickaxe

Requirements:
1. Nearby crafting table
2. Wood and stone in inventory (already have one stone)

What should the player do first:
1. Collect more Stone: There is stone 2 steps to the east; move towards it and use the wood pickaxe to collect it.
2. Find or place a Crafting Table: If there is no crafting table nearby, use the collected wood to craft one and place it down.
3. Craft the Stone Pickaxe: Approach the crafting table and use the available wood and stone in the inventory to create a new stone pickaxe.

Question: List top 5 actions the player should take and the requirement for each action. Choose ONLY from the list of all actions. Indicate their priority out of 5.
Answer:
1. Priority 1: Move East
   Requirement: Flat ground right to the agent.
   
2. Priority 2: Do (Collect Stone)
   Requirement: Facing stone (to the east), and have a wood_pickaxe in inventory.

3. Priority 3: Move West (till you find a Crafting Table or a suitable place for one)
   Requirement: Flat ground left to the agent.

4. Priority 4: Place Table (if no existing Crafting Table is found)
   Requirement: Wood in inventory.

5. Priority 5: Make Stone Pickaxe 
   Requirement: Nearby table, wood and stone in inventory (collected in previous actions).

Question: For each action in the list, are the requirements met?
Answer:
1. Priority 1: Move East
   Requirement: Flat ground right to the agent.
   Status: Met (You face a path to the east)

2. Priority 2: Do (Collect Stone)
   Requirement: Facing stone (to the east), and have a wood_pickaxe in inventory.
   Status: Not Met (You are not facing stone)

3. Priority 3: Move West (till you find a Crafting Table or a suitable place for one)
   Requirement: Flat ground left to the agent.
   Status: Met (You see a tree one step to your west)

4. Priority 4: Place Table (if no existing Crafting Table is found)
   Requirement: Wood in inventory.
   Status: Not Met (No wood in inventory)

5. Priority 5: Make Stone Pickaxe 
   Requirement: Nearby table, wood and stone in inventory (collected in previous actions).
   Status: Not Met (Table not found and wood not in inventory)

Question: Did the last player action succeed? Why?
Answer:
Yes, the last player action succeeded. The player was able to collect coal because they were facing coal and had the necessary tool (wood_pickaxe) in their inventory.

Question: Choose the best executable action from above.
Answer:
The best executable action from the list above is:

Priority 1: Move East
Requirement: Flat ground right to the agent.
Status: Met (You face a path to the east)



============Step: 41, Cumulative Reward: 0.0============
You took action move_east.

You see:
- grass 1 steps to your west
- path 2 steps to your south
- tree 2 steps to your west
- lava 7 steps to your north-east
- coal 2 steps to your south-east
- iron 1 steps to your south

You face stone at your front (east).

Your status:
- health: 9/9
- food: 8/9
- drink: 7/9
- energy: 8/9

Your inventory:
- stone: 1
- coal: 1
- wood_pickaxe: 1
------------ QA ------------
Question: Action
Answer:
Do

Question: List objects in the current observation. For each object, breifly answer what resource it provides and its requirement.
Answer:
1. Grass: Provides no direct resources, but it allows you to move freely without obstacles.
2. Stone: Provides stone resources, requires a wood_pickaxe or better to mine.
3. Tree: Provides wood resources, requires using hands or an axe to chop/cut it down.
4. Lava: Hazardous, no resource provided. Stay away to avoid damage.
5. Coal: Provides coal resources, requires a wood_pickaxe or better to mine.
6. Iron: Provides iron resources, requires a stone_pickaxe or better to mine.
7. Path: Provides easier movement and navigation, no resources provided.

Question: What was the last action taken by the player?
Answer:
The last action taken by the player was move_east.

Question: For each object in the list, are the requirements met for interaction?
Answer:
1. Grass: You can move on the grass freely, so the requirement of flat ground is met.
2. Stone: You have a wood_pickaxe in your inventory, so the requirement of having a wood_pickaxe or better is met for mining stone.
3. Tree: You can use your hands or an axe (if you had one) to chop/cut it down, so the requirement is met.
4. Lava: The requirement is to avoid it for your safety, so stay away and don't interact with it.
5. Coal: You have a wood_pickaxe in your inventory, so the requirement of having a wood_pickaxe or better is met for mining coal.
6. Iron: You have a wood_pickaxe in your inventory, which is not sufficient for mining iron. You need a stone_pickaxe or better to meet the requirement.
7. Path: You can move on the path without any requirements, so the requirement is met for easier movement and navigation.

Question: List top 3 sub-tasks the player should follow. Indicate their priority out of 5.
Answer:
1. Craft a Stone Pickaxe (Priority: 5/5): You have 1 stone in your inventory, and with more stones, you can craft a stone pickaxe, which allows you to mine iron.

2. Collect more Stone (Priority: 4/5): Mine the stone you're facing to gather more and be able to craft a stone pickaxe.

3. Mine Iron (Priority: 3/5): Once you have crafted a stone pickaxe, mine the iron that's 1 step to your south to collect iron resources for future tool and weapon crafting.

Question: What are the requirements for the top sub-task? What should the player do first?
Answer:
Requirements for the top sub-task (Craft a Stone Pickaxe):
1. Collect 2 more stones (You currently have 1 stone in your inventory; a total of 3 stones are required).
2. Have a nearby crafting table.

What should the player do first:
1. Mine the stone you're facing (to the east) to collect more stones.

Question: List top 5 actions the player should take and the requirement for each action. Choose ONLY from the list of all actions. Indicate their priority out of 5.
Answer:
Priority 1: Do (Mine stone)
Requirements: Facing stone; have necessary tool (wood_pickaxe).

Priority 2: Move West (To go back to the grass)
Requirements: Flat ground (grass) to the west of the player.

Priority 3: Place a Table (Crafting table)
Requirements: Enough collected wood in inventory.

Priority 4: Make Stone Pickaxe (Upgrade to stone pickaxe)
Requirements: Nearby table; collected wood and stone in inventory.

Priority 5: Collect Drink (Restore thirst)
Requirements: Be close to a water source like a lake.

Question: For each action in the list, are the requirements met?
Answer:
Priority 1: Do (Mine stone)
Requirements: Facing stone; have necessary tool (wood_pickaxe).
Status: Requirements met (Player is facing stone and has a wood_pickaxe in inventory).

Priority 2: Move West (To go back to the grass)
Requirements: Flat ground (grass) to the west of the player.
Status: Requirements met (Grass is 1 step to the west).

Priority 3: Place a Table (Crafting table)
Requirements: Enough collected wood in inventory.
Status: Requirements not met (Player doesn't have any wood in inventory).

Priority 4: Make Stone Pickaxe (Upgrade to stone pickaxe)
Requirements: Nearby table; collected wood and stone in inventory.
Status: Requirements not met (Player doesn't have a table, and there is no wood in inventory).

Priority 5: Collect Drink (Restore thirst)
Requirements: Be close to a water source like a lake.
Status: Requirements not met (Player doesn't see a water source in their current observation).

Question: Did the last player action succeed? Why?
Answer:
Yes, the last player action succeeded because the player was able to move east without encountering any obstacles or issues. The player's observation after the action indicates a change in the surroundings and the direction the player is facing, which confirms the successful movement.

Question: Choose the best executable action from above.
Answer:
The best executable action from the list above is Priority 1: Do (Mine stone). The requirements for this action are met since the player is facing the stone and has a wood_pickaxe in their inventory.



============Step: 42, Cumulative Reward: 0.0============
You took action do.

You see:
- grass 1 steps to your west
- stone 2 steps to your north-east
- tree 2 steps to your west
- lava 7 steps to your north-east
- coal 2 steps to your south-east
- iron 1 steps to your south

You face path at your front (east).

Your status:
- health: 9/9
- food: 8/9
- drink: 7/9
- energy: 8/9

Your inventory:
- stone: 2
- coal: 1
- wood_pickaxe: 1
------------ QA ------------
Question: Action
Answer:
Move South

Question: List objects in the current observation. For each object, breifly answer what resource it provides and its requirement.
Answer:
1. Grass: Provides no resources. No requirements.
2. Path: Provides no resources. No requirements.
3. Tree: Provides wood resource. Requires a wood or better pickaxe to collect.
4. Lava: Provides no resources. Avoid it to prevent damage.
5. Coal: Provides fuel resource for furnaces. Requires a wood or better pickaxe to collect.
6. Iron: Provides iron resource for tools and weapons. Requires a stone or better pickaxe to collect.
7. Stone (in inventory): Provides stone resource for tools, weapons, and building materials. Requires a wood or better pickaxe to collect.

Question: What was the last action taken by the player?
Answer:
The last action taken by the player was "do".

Question: For each object in the list, are the requirements met for interaction?
Answer:
1. Grass: No requirements for interaction, so yes, the requirements are met.
2. Path: No requirements for interaction, so yes, the requirements are met.
3. Tree: Requires a wood or better pickaxe to collect. You have a wood_pickaxe in your inventory, so yes, the requirements are met.
4. Lava: Interaction is not advised, as it can cause damage. So, the requirements for safe interaction are not met.
5. Coal: Requires a wood or better pickaxe to collect. You have a wood_pickaxe in your inventory, so yes, the requirements are met.
6. Iron: Requires a stone or better pickaxe to collect. You currently have a wood_pickaxe, so no, the requirements are not met.
7. Stone (in inventory): Requires a wood or better pickaxe to collect, which you already have. As it's already in your inventory, there's no need for interaction at the moment.

Question: List top 3 sub-tasks the player should follow. Indicate their priority out of 5.
Answer:
1. Collect more stone to create a stone pickaxe (Priority: 5/5). Crafting a stone pickaxe will allow you to collect iron, which is essential for more advanced tools and weapons.

2. Collect wood from the nearby tree (Priority: 4/5). Wood is a crucial resource for crafting additional tools, weapons, and structures (e.g., table and shelter).

3. Collect coal from the nearby coal deposit (Priority: 3/5). Gathering more coal will be helpful for smelting iron and other ores in a furnace.

Question: What are the requirements for the top sub-task? What should the player do first?
Answer:
To accomplish the top sub-task of collecting more stone to create a stone pickaxe, the player needs to have a crafted wooden pickaxe, which they already have in their inventory, and be near a stone deposit.

The player should first move to a stone deposit, face it, and then use the "do" action with the wooden pickaxe to collect stone. Once enough stone is collected, the player can craft a stone pickaxe at a table.

Question: List top 5 actions the player should take and the requirement for each action. Choose ONLY from the list of all actions. Indicate their priority out of 5.
Answer:
1. Priority 1: Move South - Requirement: Flat ground below the agent.
   Reason: The player needs to get closer to the iron deposit, which is one step to the south.

2. Priority 2: Do - Requirement: Facing creature or material; have the necessary tool (wood_pickaxe).
   Reason: Once the player is close to the stone deposit, they should use the wooden pickaxe to collect more stone.

3. Priority 3: Move West - Requirement: Flat ground left to the agent.
   Reason: To find a suitable location to place a table for crafting.

4. Priority 4: Place Table - Requirement: Wood in inventory.
   Reason: The player needs a table to craft a stone pickaxe.

5. Priority 5: Make Stone Pickaxe - Requirement: Nearby table; wood and stone in inventory.
   Reason: Once the table is placed, the player can craft a stone pickaxe, which allows the collection of iron and more efficient collection of other materials.

Question: For each action in the list, are the requirements met?
Answer:
1. Priority 1: Move South
   Requirement: Flat ground below the agent.
   Status: Met. The path is below the agent, providing flat ground for moving south.

2. Priority 2: Do
   Requirement: Facing creature or material; have the necessary tool (wood_pickaxe).
   Status: Not Met. The player is currently facing the path, not a resource. The player has the necessary tool, though.

3. Priority 3: Move West
   Requirement: Flat ground left to the agent.
   Status: Met. There is grass to the left (west) of the agent, allowing for movement in that direction.

4. Priority 4: Place Table
   Requirement: Wood in inventory.
   Status: Not Met. The player currently does not have wood in their inventory. They need to collect it first.

5. Priority 5: Make Stone Pickaxe
   Requirement: Nearby table; wood and stone in inventory.
   Status: Not Met. The player needs to have a table nearby, which is not currently placed. Additionally, the player needs wood in their inventory, which they don't have at the moment.

Question: Did the last player action succeed? Why?
Answer:
Yes, the last player action succeeded. The player took action "do" when facing stone and had a wood_pickaxe in their inventory, allowing them to collect the stone, which is evidenced by the increase in the stone count in their inventory from 1 to 2.

Question: Choose the best executable action from above.
Answer:
The best executable action from the list above is Priority 1: Move South, as its requirement (flat ground below the agent) is met.



============Step: 43, Cumulative Reward: 0.0============
You took action move_south.

You see:
- grass 1 steps to your west
- stone 2 steps to your north-east
- path 1 steps to your east
- tree 2 steps to your west
- lava 7 steps to your north-east
- coal 2 steps to your south-east

You face iron at your front (south).

Your status:
- health: 9/9
- food: 8/9
- drink: 7/9
- energy: 8/9

Your inventory:
- stone: 2
- coal: 1
- wood_pickaxe: 1
------------ QA ------------
Question: Action
Answer:
Move North

Question: List objects in the current observation. For each object, breifly answer what resource it provides and its requirement.
Answer:
1. Grass: Provides no specific resources; no requirements.
2. Stone: Provides stone resource; requires wood_pickaxe or a higher-level pickaxe to collect.
3. Path: Provides safe ground for movement; no requirements.
4. Tree: Provides wood resource; requires an axe or other appropriate tool to collect.
5. Lava: Hazardous, provides no resources; stay away from it to avoid damage.
6. Coal: Provides coal resource; requires wood_pickaxe or a higher-level pickaxe to collect.
7. Iron: Provides iron resource; requires stone_pickaxe or a higher-level pickaxe to collect.

Question: What was the last action taken by the player?
Answer:
The last action taken by the player was "move_south."

Question: For each object in the list, are the requirements met for interaction?
Answer:
1. Grass: No specific interaction requirements; requirements met.
2. Stone: Requires wood_pickaxe or a higher-level pickaxe to collect; requirements met (wood_pickaxe in inventory).
3. Path: No specific interaction requirements; requirements met.
4. Tree: Requires an axe or other appropriate tool to collect; requirements not met (no axe in inventory).
5. Lava: Hazardous, avoid interaction; no specific interaction requirements.
6. Coal: Requires wood_pickaxe or a higher-level pickaxe to collect; requirements met (wood_pickaxe in inventory).
7. Iron: Requires stone_pickaxe or a higher-level pickaxe to collect; requirements not met (wood_pickaxe in inventory, but stone_pickaxe is needed).

Question: List top 3 sub-tasks the player should follow. Indicate their priority out of 5.
Answer:
1. Priority 5: Craft a stone_pickaxe - This will allow the collection of iron and enable the creation of more advanced tools and weapons. To do this, the player needs more stone (they have 2, but at least 3 are needed).

2. Priority 4: Collect wood - Collecting wood is essential for crafting items, such as tools and weapons, and building structures. The player should find an appropriate tool (e.g., an axe) to efficiently collect wood from the nearby tree.

3. Priority 3: Maintain food, drink, and energy levels - The player needs to keep an eye on these levels to ensure survival. They can collect more resources to secure food and drink supplies and sleep when their energy is low.

Question: What are the requirements for the top sub-task? What should the player do first?
Answer:
Top sub-task: Craft a stone_pickaxe

Requirements:
1. Collect 3 stones (the player already has 2, so they need 1 more).
2. Use an existing table, or collect wood and place a table for crafting.
3. Use the table to craft the stone_pickaxe.

What should the player do first:
The player should first find and collect 1 more stone to fulfill the requirement of 3 stones. Since there is stone 2 steps to the north-east, the player can move in that direction and collect the stone.

Question: List top 5 actions the player should take and the requirement for each action. Choose ONLY from the list of all actions. Indicate their priority out of 5.
Answer:
1. Priority 1 - Move North: Flat ground above the agent.
Requirement: To get closer to the stone which is 2 steps to the north-east.

2. Priority 2 - Move East: Flat ground right to the agent.
Requirement: To reach the stone which was 2 steps to the north-east.

3. Priority 3 - Do: Facing creature or material; have necessary tool.
Requirement: Facing the stone and having the required tool (wood_pickaxe) in the inventory to collect it.

4. Priority 4 - Move West: Flat ground left to the agent.
Requirement: To head back towards the area where the player previously saw a table or suitable materials for crafting.

5. Priority 5 - Make Stone Pickaxe: Nearby table; wood, stone in inventory.
Requirement: The player must have 3 pieces of stone, a wood_pickaxe, and access to a nearby table. After collecting the additional stone in step 3, the player will have all the necessary resources to craft the stone_pickaxe.

Question: For each action in the list, are the requirements met?
Answer:
1. Priority 1 - Move North: The requirement of flat ground above the agent is met.

2. Priority 2 - Move East: The requirement of flat ground right to the agent is met.

3. Priority 3 - Do: The agent is not currently facing the stone, so this requirement is not yet met. The agent does have the necessary tool (wood_pickaxe) in their inventory, so this part of the requirement is met.

4. Priority 4 - Move West: The requirement of flat ground left to the agent is met.

5. Priority 5 - Make Stone Pickaxe: The player currently has only 2 pieces of stone in their inventory and needs 3. Therefore, this requirement is not met. The other requirements, which are having a wood_pickaxe and access to a nearby table, are met.

Question: Did the last player action succeed? Why?
Answer:
Yes, the last player action "move_south" succeeded because there was an open space (flat ground) available to the south of the player, allowing them to move in that direction.

Question: Choose the best executable action from above.
Answer:
The best executable action from the list is Priority 1 - Move North, as its requirement of flat ground above the agent is met.



============Step: 44, Cumulative Reward: 0.0============
You took action move_north.

You see:
- stone 1 steps to your east
- path 1 steps to your south
- tree 3 steps to your west
- lava 5 steps to your north-east
- coal 3 steps to your south-east
- iron 2 steps to your south
- cow 6 steps to your north-west

You face grass at your front (north).

Your status:
- health: 9/9
- food: 8/9
- drink: 7/9
- energy: 8/9

Your inventory:
- stone: 2
- coal: 1
- wood_pickaxe: 1
------------ QA ------------
Question: Action
Answer:
Move East

Question: List objects in the current observation. For each object, breifly answer what resource it provides and its requirement.
Answer:
1. Grass: Provides no resources, no requirements.
2. Stone: Provides stone, requires wood_pickaxe to mine.
3. Path: Provides easier navigation, no requirements.
4. Tree: Provides wood, no requirements to start interaction.
5. Lava: Dangerous, can cause damage, no resources, avoid it.
6. Coal: Provides coal, requires wood_pickaxe to mine.
7. Iron: Provides iron, requires stone_pickaxe to mine.
8. Cow: Provides food, no requirements to start interaction.

Question: What was the last action taken by the player?
Answer:
The last action taken by the player was move_north.

Question: For each object in the list, are the requirements met for interaction?
Answer:
1. Grass: No requirements to interact; requirements met.
2. Stone: Requires wood_pickaxe to mine; requirements met (wood_pickaxe in the inventory).
3. Path: No requirements; requirements met.
4. Tree: No requirements to start interaction; requirements met.
5. Lava: Avoid it; no requirements to meet.
6. Coal: Requires wood_pickaxe to mine; requirements met (wood_pickaxe in the inventory).
7. Iron: Requires stone_pickaxe to mine; requirements not met (no stone_pickaxe in the inventory).
8. Cow: No requirements for interaction; requirements met.

Question: List top 3 sub-tasks the player should follow. Indicate their priority out of 5.
Answer:
1. Craft a stone_pickaxe (Priority: 5): Collect more stone using a wood_pickaxe to craft a stone_pickaxe.

2. Mine iron (Priority: 4): After crafting a stone_pickaxe, mine the iron to collect it for making iron tools and weapons later.

3. Gather more food and drink (Priority: 3): Interact with the cow to collect food and find a water source to refill the drink meter.

Question: What are the requirements for the top sub-task? What should the player do first?
Answer:
Requirements for the top sub-task (Craft a stone_pickaxe):
1. Collect more stone: The player already has a wood_pickaxe in their inventory, so they can mine stone from nearby stone blocks.
2. Have a table: To craft a stone_pickaxe, the player needs a table. If the player doesn't have one, they need to collect more wood and then place a table.

What should the player do first:
1. Check if they have a table: If they don't have one, they need to collect more wood to craft and place a table.
2. Mine stone with the wood_pickaxe: Collect more stone by using the wood_pickaxe on nearby stone blocks.

Question: List top 5 actions the player should take and the requirement for each action. Choose ONLY from the list of all actions. Indicate their priority out of 5.
Answer:
1. Priority 1: Move East
   Requirement: Flat ground right to the agent.
   Reason: To get closer to the stone block and mine it for crafting a stone_pickaxe later.

2. Priority 2: Do
   Requirement: Facing creature or material; have necessary tool.
   Reason: Use the wood_pickaxe to mine the stone block when standing next to the stone block.

3. Priority 3: Move West
   Requirement: Flat ground left to the agent.
   Reason: If the player doesn't have a table, they should move closer to the tree to collect wood.

4. Priority 4: Do
   Requirement: Facing creature or material; have necessary tool.
   Reason: Use the wood_pickaxe to chop down the tree when standing next to the tree, to collect wood for crafting a table.

5. Priority 5: Place Table
   Requirement: Wood in inventory.
   Reason: Place a table if the player has enough wood in their inventory, so they can later craft the stone_pickaxe using the table.

Question: For each action in the list, are the requirements met?
Answer:
1. Priority 1: Move East
   Requirement: Flat ground right to the agent.
   Requirement Met: Yes (path 1 step to your east)

2. Priority 2: Do
   Requirement: Facing creature or material; have necessary tool.
   Requirement Met: Not yet (the player needs to move towards and face the stone block, which requires moving east)

3. Priority 3: Move West
   Requirement: Flat ground left to the agent.
   Requirement Met: Yes (grass 1 step to your west)

4. Priority 4: Do
   Requirement: Facing creature or material; have necessary tool.
   Requirement Met: Not yet (the player needs to move towards and face the tree, which requires moving west)

5. Priority 5: Place Table
   Requirement: Wood in inventory.
   Requirement Met: No (the player currently does not have wood in their inventory)

Question: Did the last player action succeed? Why?
Answer:
Yes, the last player action (move_north) succeeded because there was flat ground (grass) in the north direction, which allowed the player to move without any obstacles.

Question: Choose the best executable action from above.
Answer:
The best executable action from the list above is:

Priority 1: Move East
Requirement: Flat ground right to the agent.
Requirement Met: Yes (path 1 step to your east)



============Step: 45, Cumulative Reward: 0.0============
You took action move_east.

You see:
- grass 1 steps to your west
- path 1 steps to your south
- tree 3 steps to your west
- lava 5 steps to your north-east
- coal 3 steps to your south-east
- iron 2 steps to your south
- cow 6 steps to your north-west

You face stone at your front (east).

Your status:
- health: 9/9
- food: 8/9
- drink: 7/9
- energy: 8/9

Your inventory:
- stone: 2
- coal: 1
- wood_pickaxe: 1
------------ QA ------------
Question: Action
Answer:
Place Table

Question: List objects in the current observation. For each object, breifly answer what resource it provides and its requirement.
Answer:
1. Stone: Provides stone resource; requires a wood_pickaxe or better to collect.
2. Grass: Can be turned into saplings for planting; no specific requirements to collect.
3. Path: Provides ease of movement; no specific requirements.
4. Tree: Provides wood resource; no specific requirements to collect.
5. Lava: Dangerous to touch, can be used as a fuel source; requires a container to safely collect.
6. Coal: Provides coal resource, essential for smelting; requires a wood_pickaxe or better to collect.
7. Iron: Provides iron resource, used to make stronger tools and weapons; requires a stone_pickaxe or better to collect.
8. Cow: Provides food when defeated; no specific requirements to defeat.

Question: What was the last action taken by the player?
Answer:
The last action taken by the player was move_east.

Question: For each object in the list, are the requirements met for interaction?
Answer:
1. Stone: Yes, the requirements are met as the player has a wood_pickaxe in their inventory.
2. Grass: Yes, there are no specific requirements to collect grass.
3. Path: Yes, there are no specific requirements to interact with the path.
4. Tree: Yes, there are no specific requirements to collect wood from the tree.
5. Lava: No, the player does not have a container to safely collect lava.
6. Coal: Yes, the requirements are met as the player has a wood_pickaxe in their inventory.
7. Iron: No, the player does not yet have a stone_pickaxe or better to collect the iron.
8. Cow: Yes, there are no specific requirements to defeat the cow for food. However, having a weapon like a sword would be helpful.

Question: List top 3 sub-tasks the player should follow. Indicate their priority out of 5.
Answer:
1. Craft a stone_pickaxe (Priority: 5/5): Crafting a stone_pickaxe will allow the player to collect iron, which is essential for creating better tools and weapons. The player has the necessary wood_pickaxe and stone in their inventory to craft it near a table.

2. Collect iron (Priority: 4/5): Once the player has crafted a stone_pickaxe, they should collect iron to be able to craft upgraded tools and weapons. Iron is 2 steps to the south in the current observation.

3. Defeat the cow for food (Priority: 3/5): With sufficient food, the player can maintain their health and stay active in the game. The cow is 6 steps to their north-west, and while a weapon like a sword would be useful, it is not strictly necessary for defeating the cow.

Question: What are the requirements for the top sub-task? What should the player do first?
Answer:
The requirements for the top sub-task (Craft a stone_pickaxe) are:

1. A nearby table: The player must be close to a table to craft a stone_pickaxe.
2. Stone in the inventory: The player needs to have stone in their inventory to create the tool.
3. Wood_pickaxe: Since the player already has a wood_pickaxe, they will be able to collect more stone if needed.

The first thing the player should do is find or craft a table. As we don't have any information about a table being nearby, the player should craft a table by placing it, as they have enough wood in their inventory to do so (mentioned in Player Observation Step 88). Once the table is crafted and placed, the player can then craft the stone_pickaxe.

Question: List top 5 actions the player should take and the requirement for each action. Choose ONLY from the list of all actions. Indicate their priority out of 5.
Answer:
1. Priority 1: Place Table (Requirement: Wood in inventory)
The player should place a table to enable crafting the stone_pickaxe. They have enough wood in their inventory (based on Player Observation Step 88).

2. Priority 2: Make Stone Pickaxe (Requirements: Nearby table; wood, stone in inventory)
After crafting the table, the player should use it to craft the stone_pickaxe. They have enough wood and stone in their inventory (based on Player Observation Step 90).

3. Priority 3: Move South (Requirement: Flat ground below the agent)
The player should move south to get closer to both coal and iron resources since they will need more coal and iron for future crafting.

4. Priority 4: Do (Facing coal; wood_pickaxe or stone_pickaxe in hand)
The player should mine the coal once they are adjacent to it. They require a wood_pickaxe or stone_pickaxe to mine the coal. The player currently has the wood_pickaxe and will have the stone_pickaxe after performing Priority 2.

5. Priority 5: Move South (Requirement: Flat ground below the agent)
After collecting coal, the player should continue moving south toward the iron resource to mine it next.

Question: For each action in the list, are the requirements met?
Answer:
1. Priority 1: Place Table (Requirement: Wood in inventory)
Yes, the requirement is met. Based on Player Observation Step 88, the player has wood in their inventory.

2. Priority 2: Make Stone Pickaxe (Requirements: Nearby table; wood, stone in inventory)
This action will be available after the player performs Priority 1 (placing the table). Based on Player Observation Step 90, the player has enough wood and stone in their inventory.

3. Priority 3: Move South (Requirement: Flat ground below the agent)
Yes, the requirement is met. Based on Player Observation Step 90, there is a path 1 step to the player's south.

4. Priority 4: Do (Facing coal; wood_pickaxe or stone_pickaxe in hand)
Yes, the requirements are met. Based on Player Observation Step 90, the player has a wood_pickaxe in hand. After performing Priority 2, the player will also have a stone_pickaxe.

5. Priority 5: Move South (Requirement: Flat ground below the agent)
Yes, this requirement is met. Following Priority 3 (moving south once), the player will be closer to the iron resource. Based on Player Observation Step 90, there is flat ground (path) 1 step to the player's south.

Question: Did the last player action succeed? Why?
Answer:
Yes, the last player action (move_east) succeeded because there was flat ground (grass) to the east of the player, allowing them to move in that direction.

Question: Choose the best executable action from above.
Answer:
The best executable action from the list above is Priority 1: Place Table (Requirement: Wood in inventory). The player has enough wood in their inventory, so they can place the table to enable crafting the stone_pickaxe.



============Step: 46, Cumulative Reward: 0.0============
You took action place_table.

You see:
- grass 1 steps to your west
- path 1 steps to your south
- tree 3 steps to your west
- lava 5 steps to your north-east
- coal 3 steps to your south-east
- iron 2 steps to your south
- cow 5 steps to your north-west

You face stone at your front (east).

Your status:
- health: 9/9
- food: 8/9
- drink: 7/9
- energy: 8/9

Your inventory:
- stone: 2
- coal: 1
- wood_pickaxe: 1
------------ QA ------------
Question: Action
Answer:
Move South

Question: List objects in the current observation. For each object, breifly answer what resource it provides and its requirement.
Answer:
1. Grass: Provides a resource for planting saplings and growing crops; no requirements to interact with.
2. Path: Allows easy movement and traversal; no requirements to interact with.
3. Tree: Provides wood, a resource for crafting items; requires a tool or hand to chop down.
4. Lava: Can be used as a fuel in furnaces or as a means to dispose of items, but also poses a danger to touch; requires a bucket to collect.
5. Coal: Provides a resource for fuel in furnaces and torches; requires a wood or higher tier pickaxe to mine.
6. Iron: Provides a resource for crafting items and tools; requires a stone or higher tier pickaxe to mine.
7. Cow: Provides food when killed and leather for armor and crafting; requires a weapon or hand to kill.

Question: What was the last action taken by the player?
Answer:
The last action taken by the player was "place_table."

Question: For each object in the list, are the requirements met for interaction?
Answer:
1. Grass: Yes, the requirements are met for interaction as no specific tools are needed.
2. Path: Yes, the requirements are met for interaction as no specific tools are needed.
3. Tree: Yes, the requirements are met for interaction as you have a wood_pickaxe in your inventory.
4. Lava: No, the requirements are not met for interaction as you need a bucket to safely collect it.
5. Coal: Yes, the requirements are met for interaction as you have a wood_pickaxe in your inventory.
6. Iron: Yes, the requirements are met for interaction as you have a wood_pickaxe in your inventory, which is enough to mine iron.
7. Cow: Yes, the requirements are met for interaction as you can attack it with your hand or the wood_pickaxe, although a sword would be more efficient.

Question: List top 3 sub-tasks the player should follow. Indicate their priority out of 5.
Answer:
1. Mine Iron (Priority: 5/5): Collect iron using the wood_pickaxe to upgrade tools and weapons.
2. Collect and Drink Water (Priority: 4/5): Restore the player's thirst level by finding a water source and drinking from it.
3. Kill Cow for Food (Priority: 3/5): Attack the cow using your hand or wood_pickaxe to obtain food items for replenishing the player's hunger.

Question: What are the requirements for the top sub-task? What should the player do first?
Answer:
Top Sub-task: Mine Iron
Requirements: 
1. Possess a wood or stone pickaxe.
2. Locate Iron blocks within the environment.
3. Face the Iron block and start mining.

As the player meets the requirements already (having a wood_pickaxe in their inventory and iron 2 steps to the south), the first thing the player should do is move south two steps to reach the iron block.

Question: List top 5 actions the player should take and the requirement for each action. Choose ONLY from the list of all actions. Indicate their priority out of 5.
Answer:
1. Priority 1: Move South (requirement: Flat ground below the agent)
2. Priority 2: Move South (requirement: Flat ground below the agent)
3. Priority 3: Do (requirement: Facing Iron block and have necessary tool - wood_pickaxe)
4. Priority 4: Move North (requirement: Flat ground above the agent)
5. Priority 5: Move North (requirement: Flat ground above the agent)

Question: For each action in the list, are the requirements met?
Answer:
1. Priority 1: Move South (requirement: Flat ground below the agent) - Yes, there is a path 1 step to the South.
2. Priority 2: Move South (requirement: Flat ground below the agent) - Yes, assuming that the first Move South is taken, there should be flat ground for the second move.
3. Priority 3: Do (requirement: Facing Iron block and have necessary tool - wood_pickaxe) - Yes, the player is facing the Iron block (2 steps to the South) and has a wood_pickaxe in their inventory.
4. Priority 4: Move North (requirement: Flat ground above the agent) - Yes, the flat ground will be available after the completion of the previous South moves and the Do action.
5. Priority 5: Move North (requirement: Flat ground above the agent) - Yes, assuming that the first Move North is taken, there should be flat ground for the second move in this direction.

Question: Did the last player action succeed? Why?
Answer:
Yes, the last player action succeeded because based on the observation step 92, they placed the table. The player had the required wood in their inventory and the faced stone, which indicated that there was enough space for the table to be placed.

Question: Choose the best executable action from above.
Answer:
The best executable action from the list is Priority 1: Move South (requirement: Flat ground below the agent), as it meets the requirement and brings the player closer to the iron block.



============Step: 47, Cumulative Reward: 0.0============
You took action move_south.

You see:
- grass 1 steps to your west
- stone 2 steps to your north-east
- path 1 steps to your east
- tree 2 steps to your west
- lava 7 steps to your north-east
- coal 2 steps to your south-east

You face iron at your front (south).

Your status:
- health: 9/9
- food: 8/9
- drink: 7/9
- energy: 8/9

Your inventory:
- stone: 2
- coal: 1
- wood_pickaxe: 1
------------ QA ------------
Question: Action
Answer:
Do

Question: List objects in the current observation. For each object, breifly answer what resource it provides and its requirement.
Answer:
1. Grass - Resource provided: Block or sapling for planting. Requirement: None.
2. Stone - Resource provided: Stone material for crafting or building. Requirement: Wood pickaxe to collect.
3. Path - Resource provided: Movement without restriction. Requirement: None.
4. Tree - Resource provided: Wood for crafting, building, or making tools. Requirement: None.
5. Lava - Resource provided: Hazard to avoid. Requirement: None. Do not interact with lava.
6. Coal - Resource provided: Coal for smelting and crafting. Requirement: Wood pickaxe to collect.
7. Iron - Resource provided: Iron material for crafting, tools or weapons. Requirement: Stone pickaxe to collect.
8. Cow - Resource provided: Food and leather when killed. Requirement: Tool or weapon to defeat.

Question: What was the last action taken by the player?
Answer:
The last action taken by the player was "move_south".

Question: For each object in the list, are the requirements met for interaction?
Answer:
1. Grass - Requirement: None. Yes, the requirements are met for interaction.
2. Stone - Requirement: Wood pickaxe to collect. Yes, the requirements are met since you have a wood_pickaxe in your inventory.
3. Path - Requirement: None. Yes, the requirements are met for interaction.
4. Tree - Requirement: None. Yes, the requirements are met for interaction.
5. Lava - Requirement: None. Do not interact with lava, as it's hazardous.
6. Coal - Requirement: Wood pickaxe to collect. Yes, the requirements are met, as you have a wood_pickaxe in your inventory.
7. Iron - Requirement: Stone pickaxe to collect. No, the requirements are not met since you only have a wood_pickaxe in your inventory.
8. Cow - Requirement: Tool or weapon to defeat. No, the requirements are not met since you don't have a weapon in your inventory.

Question: List top 3 sub-tasks the player should follow. Indicate their priority out of 5.
Answer:
1. Craft a stone pickaxe (Priority: 5/5) - This will allow you to collect iron and upgrade your tools further.

2. Collect more wood (Priority: 4/5) - Wood is essential for crafting, making tools, and building structures, and having more in your inventory will be useful for future needs.

3. Defeat a cow and obtain food (Priority: 3/5) - While your food status is not critical right now, obtaining more food can help maintain your health and prepare for any combat situations that may arise. Note that you might first need to craft a weapon to make this task easier.

Question: What are the requirements for the top sub-task? What should the player do first?
Answer:
The top sub-task's requirements are:

1. Collect stone: You need stone to craft a stone pickaxe. You already have some stone in your inventory, but you might need a little more.
2. Place table: You need a table to craft the stone pickaxe. As the table is already placed, this requirement is fulfilled.
3. Make stone pickaxe: Craft the pickaxe using the table and available resources in your inventory.

The first thing the player should do is to make sure they have enough stone in their inventory. If not, they should collect more using the wooden pickaxe. Once they have enough stone, the player should use the placed table to craft the stone pickaxe.

Question: List top 5 actions the player should take and the requirement for each action. Choose ONLY from the list of all actions. Indicate their priority out of 5.
Answer:
1. Priority 1 - Do: (Facing creature or material; have necessary tool)
Requirement: Approach iron ore and use the wood_pickaxe in the inventory to collect more iron.

2. Priority 2 - Move East: (Flat ground right to the agent)
Requirement: Move towards the coal deposit to mine it.

3. Priority 3 - Do: (Facing creature or material; have necessary tool)
Requirement: Use the wood_pickaxe to collect coal.

4. Priority 4 - Make Stone Pickaxe: (Nearby table; wood, stone in inventory)
Requirement: Make sure to have enough stone in inventory, then craft the stone pickaxe using the table.

5. Priority 5 - Make Iron Pickaxe: (Nearby table, furnace; wood, coal, iron in inventory)
Requirement: Collect enough iron and coal, then craft the iron pickaxe using the table and furnace.

Question: For each action in the list, are the requirements met?
Answer:
1. Priority 1 - Do: (Facing creature or material; have necessary tool)
Requirements met: Yes, the player is facing iron ore and has a wood_pickaxe in their inventory.

2. Priority 2 - Move East: (Flat ground right to the agent)
Requirements met: Yes, there is flat ground (path) to the east of the player.

3. Priority 3 - Do: (Facing creature or material; have necessary tool)
Requirements met: No (only after Priority 2), the player needs to move towards the coal (east) before they can mine it with their wood_pickaxe.

4. Priority 4 - Make Stone Pickaxe: (Nearby table; wood, stone in inventory)
Requirements met: No, the player does not have enough wood and stone in their inventory to create a stone pickaxe.

5. Priority 5 - Make Iron Pickaxe: (Nearby table, furnace; wood, coal, iron in inventory)
Requirements met: No, the player does not have all the required resources in their inventory (iron, coal, and wood) to make an iron pickaxe. They also need to make sure the table and furnace are nearby.

Question: Did the last player action succeed? Why?
Answer:
Yes, the last player action "move_south" succeeded because there was flat ground (path) in the south direction, which allowed the player to move to that location.

Question: Choose the best executable action from above.
Answer:
The best executable action from the list above is Priority 1:

Do: (Facing creature or material; have necessary tool)
Requirements met: Yes, the player is facing iron ore and has a wood_pickaxe in their inventory. This action allows the player to collect more iron.



============Step: 48, Cumulative Reward: 0.0============
You took action do.

You see:
- grass 1 steps to your west
- stone 2 steps to your north-east
- path 1 steps to your east
- tree 2 steps to your west
- lava 7 steps to your north-east
- coal 2 steps to your south-east

You face iron at your front (south).

Your status:
- health: 9/9
- food: 8/9
- drink: 7/9
- energy: 8/9

Your inventory:
- stone: 2
- coal: 1
- wood_pickaxe: 1
------------ QA ------------
Question: Action
Answer:
Move North

Question: List objects in the current observation. For each object, breifly answer what resource it provides and its requirement.
Answer:
1. Grass: Provides no resources. Requirements: None.
2. Stone: Provides stone resource. Requirements: Use a wood pickaxe or better to mine.
3. Path: Provides safe passage. Requirements: None.
4. Tree: Provides wood resource. Requirements: Use an ax, a chopping tool, or your hands to collect.
5. Lava: Provides light and heat, but can damage the player. Requirements: Do not touch or move too close to lava.
6. Coal: Provides coal resource. Requirements: Use a wood pickaxe or better to mine.
7. Iron: Provides iron resource. Requirements: Use a stone pickaxe or better to mine.

Question: What was the last action taken by the player?
Answer:
The last action taken by the player was "do".

Question: For each object in the list, are the requirements met for interaction?
Answer:
1. Grass: Yes, the requirements are met for interaction. No specific tool is needed to interact with grass.
2. Stone: Yes, the requirements are met for interaction since the player has a wood_pickaxe in their inventory.
3. Path: Yes, the requirements are met for interaction. The player can move on the path.
4. Tree: Yes, the requirements are met for interaction. The player can collect wood from the tree without any specific tools or can use an ax if they have one.
5. Lava: The requirements are met only for observing and avoiding interaction. The player should not touch or move too close to lava.
6. Coal: Yes, the requirements are met for interaction since the player has a wood_pickaxe in their inventory.
7. Iron: No, the current requirements are not met for interaction, as the player would need a stone pickaxe or better to mine iron. The player only has a wood_pickaxe in their inventory.

Question: List top 3 sub-tasks the player should follow. Indicate their priority out of 5.
Answer:
1. Craft a stone pickaxe (Priority 5): Collect enough stone by mining with the wood pickaxe and craft a stone pickaxe using the collected stone. This will allow the player to mine the iron resource, among other resources.

2. Collect more coal (Priority 4): Mine the coal with the wood pickaxe to gather more coal for smelting iron and other possible uses in the future.

3. Collect wood/resources from the tree (Priority 3): Collect wood from the tree using either hands or an ax if available, as wood is a valuable resource for crafting tools and building shelters.

Question: What are the requirements for the top sub-task? What should the player do first?
Answer:
Top Sub-task: Craft a stone pickaxe

Requirements:
- Collect 3 pieces of stone using the wooden pickaxe
- Create the stone pickaxe by using a crafting table (if not available, collect more wood to craft a table first)

What the player should do first:
- Look for nearby stone resources (there's stone 2 steps to the north-east in the current observation) and mine them using the wooden pickaxe until 3 pieces of stone have been gathered. Later, locate a crafting table or create one if needed, and use it to craft the stone pickaxe.

Question: List top 5 actions the player should take and the requirement for each action. Choose ONLY from the list of all actions. Indicate their priority out of 5.
Answer:
1. Priority 1: Move North-East (Move towards the stone)
Requirement: Flat ground in the north-east direction.

2. Priority 2: Do (Mine the stone using the wooden pickaxe)
Requirement: Facing the stone and having a wooden pickaxe in the inventory.

3. Priority 3: Move to Crafting Table (Move towards a crafting table or find a suitable place to place one)
Requirement: Flat ground in the direction of a crafting table or a suitable location.

4. Priority 4: Place Table (Place a crafting table)
Requirement: Wood in the inventory (if a crafting table isn't already present at the location).

5. Priority 5: Make Stone Pickaxe (Craft the stone pickaxe)
Requirement: Nearby table and at least 3 pieces of stone and 1 piece of wood in the inventory.

Question: For each action in the list, are the requirements met?
Answer:
1. Priority 1: Move North-East (Move towards the stone)
Requirement: Flat ground in the north-east direction.
Status: Requirement met (stone is 2 steps to the north-east)

2. Priority 2: Do (Mine the stone using the wooden pickaxe)
Requirement: Facing the stone and having a wooden pickaxe in the inventory.
Status: Requirement NOT met (currently facing iron, not stone)

3. Priority 3: Move to Crafting Table (Move towards a crafting table or find a suitable place to place one)
Requirement: Flat ground in the direction of a crafting table or a suitable location.
Status: Requirement met (path 1 step to the east, suitable location for a crafting table)

4. Priority 4: Place Table (Place a crafting table)
Requirement: Wood in the inventory (if a crafting table isn't already present at the location).
Status: Requirement NOT met (no wood currently in the inventory)

5. Priority 5: Make Stone Pickaxe (Craft the stone pickaxe)
Requirement: Nearby table and at least 3 pieces of stone and 1 piece of wood in the inventory.
Status: Requirement NOT met (currently have only 2 stones and no wood in inventory, no table mentioned in the observation)

Question: Did the last player action succeed? Why?
Answer:
It is unclear if the last player action succeeded, as there is no change in the player's inventory or observations between Step 94 and Step 96. The player's wood_pickaxe should be sufficient to mine iron, but without more information, we cannot determine if the action succeeded or failed.

Question: Choose the best executable action from above.
Answer:
The best executable action from the list above is:

Priority 1: Move North-East (Move towards the stone)
Requirement: Flat ground in the north-east direction.
Status: Requirement met (stone is 2 steps to the north-east)



============Step: 49, Cumulative Reward: 0.0============
You took action move_north.

You see:
- stone 1 steps to your east
- path 1 steps to your south
- tree 3 steps to your west
- lava 5 steps to your north-east
- coal 3 steps to your south-east
- iron 2 steps to your south
- cow 6 steps to your north-west

You face grass at your front (north).

Your status:
- health: 9/9
- food: 8/9
- drink: 7/9
- energy: 8/9

Your inventory:
- stone: 2
- coal: 1
- wood_pickaxe: 1
------------ QA ------------
Question: Action
Answer:
Move East

Question: List objects in the current observation. For each object, breifly answer what resource it provides and its requirement.
Answer:
1. Stone: Provides stone resource, requires a wood_pickaxe or better to collect.
2. Path: No resources provided but allows easier movement.
3. Tree: Provides wood resource, requires no specific tool, but using an axe is more efficient.
4. Lava: Can be used as a fuel source, but requires a bucket to collect and caution as it can damage your health.
5. Coal: Provides coal resource, requires a wood_pickaxe or better to collect.
6. Iron: Provides iron resource, requires a stone_pickaxe or better to collect.
7. Cow: Provides food (meat) resource, requires no specific tool, attacking it until it drops the resource.

Question: What was the last action taken by the player?
Answer:
The last action taken by the player was "move_north."

Question: For each object in the list, are the requirements met for interaction?
Answer:
1. Stone: Yes, the requirements are met as you have a wood_pickaxe in your inventory.
2. Path: Yes, there are no specific requirements to interact with a path.
3. Tree: Yes, there are no specific requirements to interact with a tree.
4. Lava: No, the requirements are not met as you do not have a bucket in your inventory.
5. Coal: Yes, the requirements are met as you have a wood_pickaxe in your inventory.
6. Iron: No, the requirements are not met as you need a stone_pickaxe or better, and you only have a wood_pickaxe.
7. Cow: Yes, there are no specific requirements to interact with a cow.

Question: List top 3 sub-tasks the player should follow. Indicate their priority out of 5.
Answer:
1. Priority 5/5: Craft a stone_pickaxe to collect iron - Find or gather more stone to meet the requirements for crafting a stone_pickaxe. Having a stone_pickaxe will allow you to collect iron.

2. Priority 4/5: Collect wood from the nearby tree - Wood is necessary for crafting various tools and building materials. As there is a nearby tree, it would be convenient to collect more wood.

3. Priority 3/5: Find a water source to drink - Your drink level is a little low (7/9). To increase this, you should locate a water source and drink water to replenish your drink level.

Question: What are the requirements for the top sub-task? What should the player do first?
Answer:
The requirements for the top sub-task, crafting a stone_pickaxe, are:

1. Collect at least 3 units of stone - The player has 2 units of stone in their inventory; they need to collect one more unit.
2. Have a wood_pickaxe - The player already meets this requirement with 1 wood_pickaxe in their inventory.
3. Access to a crafting table - The player can craft a table using wood and place it, or they can find an existing crafting table.

The player should first find and interact with the nearby stone by using their wood_pickaxe to gather at least one more unit of stone. This way, they will have enough stone to craft the stone_pickaxe.

Question: List top 5 actions the player should take and the requirement for each action. Choose ONLY from the list of all actions. Indicate their priority out of 5.
Answer:
1. Priority 1 - Move East: The player needs to be on flat ground to their right (east) to complete this action. This will get the player closer to the stone that they need to collect.

2. Priority 2 - Do: The player must face the stone and have the necessary tool (wood_pickaxe) to collect it. Since the player already has a wood_pickaxe in their inventory, they can collect the stone by completing this action.

3. Priority 3 - Move West: To return to their initial position, the player needs flat ground to their left (west). 

4. Priority 4 - Place Table: The player must have wood in their inventory to complete this action. They should place the table to create a crafting surface.

5. Priority 5 - Make Stone Pickaxe: The player needs access to a table, and they should have at least 3 units of stone in their inventory by now. Crafting a stone_pickaxe requires the table and the stone they've collected.

Question: For each action in the list, are the requirements met?
Answer:
1. Move East: Yes, the requirement for flat ground to the player's right (east) is met, as there is a path 1 step to the player's east.

2. Do (Collect Stone): No, the requirement for facing the stone is not met. The player must face east to face the stone instead of facing grass (north) as they currently are.

3. Move West: Yes, the requirement for flat ground to the player's left (west) is met, as there is grass 1 step to the player's west.

4. Place Table: No, the requirement for having wood in the player's inventory is not met. The player must collect wood first.

5. Make Stone Pickaxe: No, the requirements for having access to a table and 3 units of stone in the inventory are not met. The player must first collect more stone and wood, then craft and place a table.

Question: Did the last player action succeed? Why?
Answer:
Yes, the last player action "move_north" succeeded because there was flat ground (grass) on the north side of the player, allowing them to move in that direction.

Question: Choose the best executable action from above.
Answer:
The best executable action from the list above is to "Move East" because the requirement of flat ground to the player's right (east) is met, as there is a path 1 step to the player's east.



============Step: 50, Cumulative Reward: 0.0============
You took action move_east.

You see:
- grass 1 steps to your west
- path 1 steps to your south
- tree 3 steps to your west
- lava 5 steps to your north-east
- coal 3 steps to your south-east
- iron 2 steps to your south
- cow 7 steps to your north-west

You face stone at your front (east).

Your status:
- health: 9/9
- food: 8/9
- drink: 7/9
- energy: 8/9

Your inventory:
- stone: 2
- coal: 1
- wood_pickaxe: 1
------------ QA ------------
Question: Action
Answer:
Move West

Question: List objects in the current observation. For each object, breifly answer what resource it provides and its requirement.
Answer:
1. Grass (1 step to your west): Provides no resource. No requirements.
2. Path (1 step to your south): Provides no resource. No requirements.
3. Tree (3 steps to your west): Provides wood. Requires wood_pickaxe to collect.
4. Lava (5 steps to your north-east): Provides no resource. Dangerous to approach, avoid.
5. Coal (3 steps to your south-east): Provides coal. Requires wood_pickaxe to collect.
6. Iron (2 steps to your south): Provides iron. Requires stone_pickaxe to collect.
7. Cow (7 steps to your north-west): Provides food. No specific requirements, just approach and interact.

Note: Stone is also in the current observation, at the player's front (east). However, since the player is standing directly in front of it, I excluded it from this list. Stone provides stone and requires a wood_pickaxe to collect.

Question: What was the last action taken by the player?
Answer:
The last action taken by the player was move_east.

Question: For each object in the list, are the requirements met for interaction?
Answer:
1. Grass (1 step to your west): No requirements, interaction possible.
2. Path (1 step to your south): No requirements, interaction possible.
3. Tree (3 steps to your west): Requires wood_pickaxe to collect. The requirement is met as the player has a wood_pickaxe in their inventory.
4. Lava (5 steps to your north-east): Dangerous to approach, avoid. Interaction is not recommended.
5. Coal (3 steps to your south-east): Requires wood_pickaxe to collect. The requirement is met as the player has a wood_pickaxe in their inventory.
6. Iron (2 steps to your south): Requires stone_pickaxe to collect. The requirement is not met as the player only has a wood_pickaxe in their inventory.
7. Cow (7 steps to your north-west): No specific requirements, just approach and interact. Interaction possible.

Question: List top 3 sub-tasks the player should follow. Indicate their priority out of 5.
Answer:
1. Priority 4/5: Collect more wood from the tree (3 steps to your west) using the wood_pickaxe in your inventory. Wood is an essential resource for crafting, and it is crucial to have a sufficient amount.

2. Priority 3/5: Collect coal from the coal deposit (3 steps to your south-east) using the wood_pickaxe in your inventory. Coal is needed for smelting iron and other resources in a furnace.

3. Priority 5/5: Craft a stone_pickaxe using the collected stone and wood, which will be required to collect the iron (2 steps to your south) and other valuable resources. You will need to locate a crafting table or create one if you haven't done so yet.

Question: What are the requirements for the top sub-task? What should the player do first?
Answer:
Requirements for the top sub-task (Collect more wood from the tree):
1. Wood_pickaxe: The player already has a wood_pickaxe in their inventory.
2. Proximity to the tree: The tree is 3 steps to the player's west.

What should the player do first:
The player should first move towards the tree by taking 3 steps to the west. Once adjacent to the tree, the player can then use the wood_pickaxe to collect more wood.

Question: List top 5 actions the player should take and the requirement for each action. Choose ONLY from the list of all actions. Indicate their priority out of 5.
Answer:
Priority 1: 
Action: Move West (towards the tree for wood)
Requirement: Flat ground left to the agent
Note: Move West action might be performed multiple times (up to 3 steps) to reach the tree.

Priority 2:
Action: Do (to collect wood using wood_pickaxe)
Requirement: Facing the tree; have the wood_pickaxe in the inventory

Priority 3:
Action: Move South (approach a path containing resources)
Requirement: Flat ground below the agent

Priority 4:
Action: Do (use wood_pickaxe to collect coal and iron)
Requirement: Facing coal or iron; have the wood_pickaxe in the inventory

Priority 5:
Action: Sleep (to restore energy)
Requirement: Energy level is below maximum

Question: For each action in the list, are the requirements met?
Answer:
Priority 1:
Action: Move West (towards the tree for wood)
Requirement: Flat ground left to the agent
- The requirement is met as there is grass 1 step to the west.

Priority 2:
Action: Do (to collect wood using wood_pickaxe)
Requirement: Facing the tree; have the wood_pickaxe in the inventory
- The requirement is not met yet as the player is not facing the tree. The player needs to move closer to the tree and face it before this requirement is met.

Priority 3:
Action: Move South (approach a path containing resources)
Requirement: Flat ground below the agent
- The requirement is met as there is a path 1 step to the south.

Priority 4:
Action: Do (use wood_pickaxe to collect coal and iron)
Requirement: Facing coal or iron; have the wood_pickaxe in the inventory
- The requirement is not met yet as the player is not facing coal or iron. The player needs to move closer to the coal or iron and face it before this requirement is met.

Priority 5:
Action: Sleep (to restore energy)
Requirement: Energy level is below maximum
- The requirement is met as the player's energy is currently at 8/9.

Question: Did the last player action succeed? Why?
Answer:
Yes, the last player action (move_east) succeeded because there was flat ground (grass) to the agent's east, allowing the agent to move in that direction. The player now faces a stone at their front (east).

Question: Choose the best executable action from above.
Answer:
The best executable action from the list above is:

Priority 1:
Action: Move West (towards the tree for wood)
Requirement: Flat ground left to the agent



============Step: 51, Cumulative Reward: 0.0============
You took action move_west.

You see:
- stone 2 steps to your east
- path 2 steps to your south-east
- tree 2 steps to your west
- lava 6 steps to your north-east
- coal 4 steps to your south-east
- iron 3 steps to your south-east
- cow 6 steps to your north-west

You face grass at your front (west).

Your status:
- health: 9/9
- food: 7/9
- drink: 7/9
- energy: 8/9

Your inventory:
- stone: 2
- coal: 1
- wood_pickaxe: 1
------------ QA ------------
Question: Action
Answer:
Move West

Question: List objects in the current observation. For each object, breifly answer what resource it provides and its requirement.
Answer:
1. Grass: No resource provided; walkable with no requirements.
2. Stone: Provides stone resource; requires a wood_pickaxe or better to mine.
3. Path: No resource provided; walkable with no requirements.
4. Tree: Provides wood resource; can be chopped with any tools, but better tools speed up the process.
5. Lava: Dangerous; does not provide any resources and must be avoided. Can cause damage or death when getting too close.
6. Coal: Provides coal resource; requires a wood_pickaxe or better to mine.
7. Iron: Provides iron resource; requires a stone_pickaxe or better to mine.
8. Cow: Provides food when killed; can be killed with any weapon, but better weapons increase the damage dealt.

Question: What was the last action taken by the player?
Answer:
The last action taken by the player was "move_west".

Question: For each object in the list, are the requirements met for interaction?
Answer:
1. Grass: Yes, the requirements are met for interaction since there are no requirements to walk on grass.
2. Stone: Yes, the requirements are met for interaction as you have a wood_pickaxe in your inventory.
3. Path: Yes, the requirements are met for interaction since there are no requirements to walk on the path.
4. Tree: Yes, the requirements are met for interaction, as any tool can be used to chop down a tree (even bare hands), although better tools will speed up the process.
5. Lava: There are no requirements to interact with lava, but it is dangerous and should be avoided. So, the requirements are met for interaction, but it is not recommended.
6. Coal: Yes, the requirements are met for interaction as you have a wood_pickaxe in your inventory.
7. Iron: No, the requirements are not met for interaction as iron requires a stone_pickaxe or better to mine, and you only have a wood_pickaxe.
8. Cow: Yes, the requirements are met for interaction, as any weapon or tool can be used to kill a cow, although better weapons will increase the damage dealt.

Question: List top 3 sub-tasks the player should follow. Indicate their priority out of 5.
Answer:
1. Priority 4/5: Craft a stone_pickaxe: Since you have enough stone in your inventory, crafting a stone_pickaxe will allow you to mine iron and advance further in the game.

2. Priority 4/5: Collect wood: Collecting wood will enable you to craft additional tools, a table, and construct shelter. Wood is a versatile resource that is essential for progressing in the game.

3. Priority 3/5: Kill a cow and collect food: Your food level is at 7/9, so you'll need to replenish it soon. Killing a cow and collecting food will ensure you stay healthy, and it can also prevent the risk of running out of food in critical situations.

Question: What are the requirements for the top sub-task? What should the player do first?
Answer:
The top sub-task is crafting a stone_pickaxe. The requirements for this task are:

1. Having wood and stone in your inventory. (You already have 2 stones, but you need wood.)
2. Being near a table to craft the pickaxe.

Before the player can craft a stone_pickaxe, they should first collect wood which is a necessary resource for crafting. The wood can be collected by approaching a tree (there's one 2 steps to the west) and using the wood_pickaxe. After collecting wood, the player should ensure that they are near a table to craft the stone_pickaxe. If there is no table nearby, the player may need to place one using the wood they collected.

Question: List top 5 actions the player should take and the requirement for each action. Choose ONLY from the list of all actions. Indicate their priority out of 5.
Answer:
1. Priority 1: Move West (Requirement: Flat ground left to the agent)
   Reason: To get closer to the tree (2 steps to the west) in order to collect wood.

2. Priority 2: Do (Facing creature or material; have necessary tool)
   Reason: To use the wood_pickaxe on the tree to collect wood once you are near it and facing it.

3. Priority 3: Place Table (Requirement: Wood in inventory)
   Reason: Once you have collected wood, place a table to use for crafting.

4. Priority 4: Make Stone Pickaxe (Requirement: Nearby table; wood, stone in inventory)
   Reason: To craft the stone_pickaxe which will allow the player to collect iron and progress in the game.

5. Priority 5: Move South-East (Requirement: Flat ground below and right to the agent)
   Reason: To get closer to iron (3 steps to your south-east) after crafting the stone_pickaxe, which will be used to collect iron.

Question: For each action in the list, are the requirements met?
Answer:
1. Move West: Yes, the requirement of flat ground left to the agent is met since the player faces grass to the west.

2. Do: Not yet, the player needs to be facing the tree and have the required tool (wood_pickaxe is in inventory) to collect wood.

3. Place Table: No, the requirement of having wood in the inventory is not met. The player needs to collect wood first.

4. Make Stone Pickaxe: Not yet, the player needs a nearby table for crafting, and although stone is in the inventory, the player needs to collect wood and place the table first.

5. Move South-East: Yes, the requirement for the flat ground below, and right to the agent is met; this direction leads towards the iron. However, the player should first complete the crafting of a stone_pickaxe before collecting iron.

Question: Did the last player action succeed? Why?
Answer:
Yes, the last player action "move_west" succeeded because the player was able to move from facing stone to facing grass, as mentioned in the observations. There was flat ground (grass) to the west of the player, which allowed the player to move in that direction successfully.

Question: Choose the best executable action from above.
Answer:
The best executable action from the list above is "Move West" because the requirement of flat ground left to the agent is met and it brings the player closer to the tree to collect wood, which is essential for crafting and progressing in the game.



============Step: 52, Cumulative Reward: 0.0============
You took action move_west.

You see:
- grass 1 steps to your north
- stone 3 steps to your east
- path 3 steps to your south-east
- lava 7 steps to your north-east
- coal 5 steps to your south-east
- iron 4 steps to your south-east
- cow 5 steps to your north-west

You face tree at your front (west).

Your status:
- health: 9/9
- food: 7/9
- drink: 7/9
- energy: 8/9

Your inventory:
- stone: 2
- coal: 1
- wood_pickaxe: 1
------------ QA ------------
Question: Action
Answer:
Move West

Question: List objects in the current observation. For each object, breifly answer what resource it provides and its requirement.
Answer:
1. Grass:
   Resource: Provides no specific resource.
   Requirement: No requirement.

2. Stone:
   Resource: Provides stone when mined with a pickaxe.
   Requirement: Needs a pickaxe to mine.

3. Path: 
   Resource: Provides easier movement or navigation.
   Requirement: Requires flat ground to be built.

4. Lava: 
   Resource: Can be used as a fuel source in a furnace.
   Requirement: Requires a bucket to collect.

5. Coal: 
   Resource: Provides coal when mined with a pickaxe, used in smelting and creating torches.
   Requirement: Needs a pickaxe to mine.

6. Iron: 
   Resource: Provides iron ore when mined with a pickaxe, used to create iron tools and weapons.
   Requirement: Needs a stone_pickaxe or better to mine.

7. Cow: 
   Resource: Provides meat for food and leather for armor when killed.
   Requirement: Needs to be killed, preferably using a weapon like a sword.

8. Tree:
   Resource: Provides wood when chopped down, used for crafting various items and tools.
   Requirement: Needs to be chopped down, preferably using an axe.

Question: What was the last action taken by the player?
Answer:
The last action taken by the player was move_west.

Question: For each object in the list, are the requirements met for interaction?
Answer:
1. Grass:
   Interaction: Can walk over or look at.
   Requirements met: Yes.

2. Stone:
   Interaction: Mine to collect stone.
   Requirements met: Yes, you have a wood_pickaxe in your inventory.

3. Path: 
   Interaction: Walk on or follow.
   Requirements met: Yes, no specific requirements to interact with a path.

4. Lava: 
   Interaction: Collect using a bucket.
   Requirements met: No, a bucket is needed to collect lava.

5. Coal: 
   Interaction: Mine to collect coal.
   Requirements met: Yes, you have a wood_pickaxe in your inventory.

6. Iron: 
   Interaction: Mine to collect iron ore.
   Requirements met: No, a stone_pickaxe or better is needed to mine iron.

7. Cow: 
   Interaction: Kill for resources.
   Requirements met: Yes, but using a weapon like a sword is more efficient.

8. Tree:
   Interaction: Chop down for wood.
   Requirements met: Yes, but using an axe is more efficient. Currently, you don't have an axe, but you can still chop down the tree using your hands or other tools.

Question: List top 3 sub-tasks the player should follow. Indicate their priority out of 5.
Answer:
1. Create a stone_pickaxe:
   Priority: 5/5
   Reason: A stone_pickaxe will allow you to mine iron ore and collect more resources efficiently. Since you have the necessary stone and a wood_pickaxe, crafting a stone_pickaxe should be your top priority.

2. Collect more wood:
   Priority: 4/5
   Reason: Wood is essential for crafting numerous items and tools in the game. You're currently facing a tree, and although you don't have an axe, you can still collect wood using your hands or another tool.

3. Pursue and kill a cow for food and leather:
   Priority: 3/5
   Reason: Your food level is 7/9, and killing a cow will provide meat for sustenance and leather for armor. It's not a top priority but still important to ensure your survival in the game. Using a weapon like a sword will make this task more efficient.

Question: What are the requirements for the top sub-task? What should the player do first?
Answer:
Top Sub-task: Create a stone_pickaxe

Requirements:
1. A crafting table - You need a crafting table to create more advanced tools like the stone_pickaxe.
2. Stone - You already have two pieces of stone in your inventory, which is sufficient to craft a stone_pickaxe.
3. Wood - You need wood or wooden planks to create tool handles. Since you already have a wood_pickaxe, it is likely you have crafted with wood before, so it is assumed you know the process of turning wood into planks.

What should the player do first?
1. Build a crafting table if you haven't already. To craft a table, collect wood from a tree, convert the wood into wooden planks, and then use four wooden planks to craft a table.
2. Once the crafting table is built, place it in front of you (if not already there).
3. Interact with the crafting table, and use the available stone and wood in your inventory to create a stone_pickaxe.

Question: List top 5 actions the player should take and the requirement for each action. Choose ONLY from the list of all actions. Indicate their priority out of 5.
Answer:
1. Priority 1: Move West (Requirement: Flat ground left to the agent.)
   - You are currently facing a tree to your west, so move west to collect wood.

2. Priority 2: Do (Facing creature/item, i.e., tree; have necessary tool, i.e., wood_pickaxe.)
   - Collect wood from the tree in front of you using your wood_pickaxe.

3. Priority 3: Place Table (Requirement: Wood in inventory.)
   - Once you have collected enough wood and converted it into wooden planks, create a crafting table and place it in front of you.

4. Priority 4: Make Stone Pickaxe (Requirement: Nearby table; wood, stone in inventory.)
   - Interact with the placed table, and use the available stone and wood in your inventory to create a stone_pickaxe. 

5. Priority 5: Collect Iron (Requirement: Make Stone Pickaxe.)
   - After crafting a stone_pickaxe, use it to collect iron in your environment, which is 4 steps to your south-east.

Question: For each action in the list, are the requirements met?
Answer:
1. Priority 1: Move West (Requirement: Flat ground left to the agent.)
   - Yes, the requirement is met. There is a flat ground (grass) 1 step to your north, and you can move west.

2. Priority 2: Do (Facing creature/item, i.e., tree; have necessary tool, i.e., wood_pickaxe)
   - Yes, the requirements are met. The tree is directly in front of you, and you have a wood_pickaxe in your inventory.

3. Priority 3: Place Table (Requirement: Wood in inventory.)
   - No, the requirement is not met yet. You need to first perform action Priority 2 (collect wood) before having the necessary wood in your inventory to place a table.

4. Priority 4: Make Stone Pickaxe (Requirement: Nearby table; wood, stone in inventory.)
   - No, the requirement is not met yet. You need to first perform actions Priority 2 (collect wood) and Priority 3 (place table) before being able to make a stone_pickaxe.

5. Priority 5: Collect Iron (Requirement: Make Stone Pickaxe.)
   - No, the requirement is not met yet. You need to first perform actions Priority 2 (collect wood), Priority 3 (place table), and Priority 4 (make stone_pickaxe) before you can collect iron.

Question: Did the last player action succeed? Why?
Answer:
Yes, the last player action (move_west) succeeded because there was a clear path for the player to move in the west direction, and nothing was blocking their way. The player's observation after taking the action shows that they have successfully moved one step to the west.

Question: Choose the best executable action from above.
Answer:
The best executable action from above is Priority 1: Move West (Requirement: Flat ground left to the agent). The requirement is met as there is a flat ground to the west, allowing the player to move in the desired direction.



============Step: 53, Cumulative Reward: 0.0============
You took action move_west.

You see:
- grass 1 steps to your north
- stone 3 steps to your east
- path 3 steps to your south-east
- lava 7 steps to your north-east
- coal 5 steps to your south-east
- iron 4 steps to your south-east
- cow 5 steps to your north-west

You face tree at your front (west).

Your status:
- health: 9/9
- food: 7/9
- drink: 7/9
- energy: 8/9

Your inventory:
- stone: 2
- coal: 1
- wood_pickaxe: 1
------------ QA ------------
Question: Action
Answer:
Do

Question: List objects in the current observation. For each object, breifly answer what resource it provides and its requirement.
Answer:
1. Grass: Resource provided - No specific resources. Requirement - None.
2. Stone: Resource provided - Stone for crafting tools and building. Requirement - Wood pickaxe or better.
3. Path: Resource provided - Easier movement, no resources. Requirement - None.
4. Lava: Resource provided - Can be a light source, dangerous, provides no resources. Requirement - Avoid contact.
5. Coal: Resource provided - Fuel for smelting in a furnace. Requirement - Wood pickaxe or better.
6. Iron: Resource provided - Iron for crafting better tools and weapons. Requirement - Stone pickaxe or better.
7. Cow: Resource provided - Food (meat) for restoring health. Requirement - Defeat cow to collect resource.
8. Tree (facing west): Resource provided - Wood for crafting tools, weapons and building. Requirement - Interact with tree (use hands or tools to cut).

Question: What was the last action taken by the player?
Answer:
The last action taken by the player was "move_west."

Question: For each object in the list, are the requirements met for interaction?
Answer:
1. Grass: Requirements met, as there are no specific requirements for interacting with grass.
2. Stone: Requirements met, as you have a wood_pickaxe in your inventory, which is required to mine stone.
3. Path: Requirements met, as there are no specific requirements for walking on a path.
4. Lava: Requirements met for avoiding contact, do not interact with lava as it can be dangerous.
5. Coal: Requirements met, as you have a wood_pickaxe in your inventory, which is required to mine coal.
6. Iron: Requirements not met, as you need a stone pickaxe or better to mine iron; you only have a wood_pickaxe.
7. Cow: Requirements met, as there are no special requirements to interact with or defeat a cow.
8. Tree (facing west): Requirements met, as you can interact with the tree using your hands or tools to cut it down (you have a wood_pickaxe in your inventory).

Question: List top 3 sub-tasks the player should follow. Indicate their priority out of 5.
Answer:
1. Gather more wood by cutting down the tree you are facing (Priority: 5/5) - Wood is essential for crafting more tools, weapons, and building structures.

2. Mine more stone with your wood pickaxe (Priority: 4/5) - Stone is needed for crafting better tools such as a stone pickaxe, which is necessary for mining iron.

3. Defeat the cow nearby to collect meat (Priority: 3/5) - The meat will provide food to restore health when needed.

Question: What are the requirements for the top sub-task? What should the player do first?
Answer:
Requirements for the top sub-task (Gather more wood by cutting down the tree):

1. Face a tree - The player is currently facing a tree, so this requirement is met.
2. Have a tool capable of cutting wood - The player has a wood_pickaxe in their inventory.

Since the player is already facing a tree and has a wood_pickaxe in their inventory, they can proceed with the sub-task. 

The player should take the following action first:
- Use the wood_pickaxe on the tree to gather wood.

Question: List top 5 actions the player should take and the requirement for each action. Choose ONLY from the list of all actions. Indicate their priority out of 5.
Answer:
1. Priority 1: Do - Use the wood_pickaxe on the tree to gather wood.
   Requirement: Facing a tree and have a wood_pickaxe in your inventory (both requirements are currently met).

2. Priority 2: Move East - Move closer to the stone and other resources.
   Requirement: Flat ground to the east of the player.

3. Priority 3: Make Stone Pickaxe - Craft a better pickaxe for gathering resources like iron.
   Requirement: A nearby table; wood and stone in inventory (player will need more wood and stone).

4. Priority 4: Do - Collect coal and iron by using the stone_pickaxe.
   Requirement: Facing coal or iron, and have a stone_pickaxe in your inventory (when stone_pickaxe is crafted).

5. Priority 5: Place Furnace - Place a furnace for smelting iron.
   Requirement: Stone in inventory (player will need more stone).

Question: For each action in the list, are the requirements met?
Answer:
1. Priority 1: Do - Use the wood_pickaxe on the tree to gather wood.
   Requirements met: Yes, the player is facing a tree and has a wood_pickaxe in their inventory.

2. Priority 2: Move East - Move closer to the stone and other resources.
   Requirements met: Yes, there is flat ground to the east of the player.

3. Priority 3: Make Stone Pickaxe - Craft a better pickaxe for gathering resources like iron.
   Requirements met: No, the player needs a nearby table and more wood and stone in their inventory.

4. Priority 4: Do - Collect coal and iron by using the stone_pickaxe.
   Requirements met: No, the player needs to have a stone_pickaxe in their inventory, which has not been crafted yet.

5. Priority 5: Place Furnace - Place a furnace for smelting iron.
   Requirements met: No, the player needs more stone in their inventory to place a furnace.

Question: Did the last player action succeed? Why?
Answer:
It is impossible to determine if the last player action succeeded since both Player Observation Step 104 and Player Observation Step 106 are identical. The situation remains the same in both steps, which indicates that either the action was unsuccessful, or the environment and player's status did not change as a result of the action. It's recommended to check the game's feedback or response after each action to understand if the action succeeded or not.

Question: Choose the best executable action from above.
Answer:
The best executable action from the list above is:

Priority 1: Do - Use the wood_pickaxe on the tree to gather wood.
Requirements met: Yes, the player is facing a tree and has a wood_pickaxe in their inventory.



============Step: 54, Cumulative Reward: 0.0============
You took action do.

You see:
- stone 3 steps to your east
- path 3 steps to your south-east
- tree 1 steps to your south
- lava 7 steps to your north-east
- coal 5 steps to your south-east
- iron 4 steps to your south-east
- cow 4 steps to your north-west

You face grass at your front (west).

Your status:
- health: 9/9
- food: 7/9
- drink: 7/9
- energy: 8/9

Your inventory:
- wood: 1
- stone: 2
- coal: 1
- wood_pickaxe: 1
------------ QA ------------
Question: Action
Answer:
Move East

Question: List objects in the current observation. For each object, breifly answer what resource it provides and its requirement.
Answer:
1. Grass: Provides no resources; no requirement for interaction.
2. Stone: Provides stone resource; requires wood_pickaxe or better to collect.
3. Path: Provides no resources; can facilitate movement without any requirements.
4. Tree: Provides wood resource; can be collected using wood_pickaxe or better.
5. Lava: Provides no resources; must be avoided to prevent damage.
6. Coal: Provides coal resource; requires wood_pickaxe or better to collect.
7. Iron: Provides iron resource; requires stone_pickaxe or better to collect.
8. Cow: Provides food; can be chased and eaten to increase food level.

Question: What was the last action taken by the player?
Answer:
The last action taken by the player was "do."

Question: For each object in the list, are the requirements met for interaction?
Answer:
1. Grass: No requirements for interaction. Requirements met.
2. Stone: Requires wood_pickaxe or better to collect. As the player has wood_pickaxe in their inventory, requirements met.
3. Path: Can facilitate movement without any requirements, Requirements met.
4. Tree: Can be collected using wood_pickaxe or better. The player has wood_pickaxe in their inventory. Requirements met.
5. Lava: Must be avoided to prevent damage. There are no requirements for interaction, but it's recommended to avoid the lava.
6. Coal: Requires wood_pickaxe or better to collect. The player has wood_pickaxe in their inventory. Requirements met.
7. Iron: Requires stone_pickaxe or better to collect. The player only has a wood_pickaxe, so requirements are not met.
8. Cow: Can be chased and eaten to increase food level. No requirements for interaction. Requirements met.

Question: List top 3 sub-tasks the player should follow. Indicate their priority out of 5.
Answer:
1. Priority 4/5: Collect more stone using the wood_pickaxe to help create better tools and build structures. The player has 2 stone in inventory and needs more for crafting stone_pickaxe and other advanced tools or building bridges.

2. Priority 3/5: Chase and eat the cow to increase the food level. The current food level is 7/9, and eating a cow can help restore it to full.

3. Priority 5/5: Craft a stone_pickaxe once enough stone is collected. This will enable the player to collect iron resources, which is required for crafting stronger tools and weapons.

Question: What are the requirements for the top sub-task? What should the player do first?
Answer:
Requirements for the top sub-task (collect more stone):
1. Have a wood_pickaxe (already in the player's inventory)
2. Locate stone resources (already identified 3 steps to the east)

What should the player do first:
The player should move east to reach the stone resources. After reaching the stone, use the wood_pickaxe to collect more stone.

Question: List top 5 actions the player should take and the requirement for each action. Choose ONLY from the list of all actions. Indicate their priority out of 5.
Answer:
1. Priority 1: Move East - Requirement: Flat ground to the right of the agent (Player observed stone 3 steps to the east).

2. Priority 2: Do - Requirement: Facing material (stone) and have a necessary tool (wood_pickaxe is in the player's inventory).

3. Priority 3: Move East - Requirement: Flat ground to the right of the agent (To explore further after collecting stone).

4. Priority 4: Move South - Requirement: Flat ground below the agent (To explore and potentially find more resources like iron and coal).

5. Priority 5: Make Stone Pickaxe - Requirement: Nearby table, wood and stone in inventory (Player has wood & stone, would need to place or find a table).

Question: For each action in the list, are the requirements met?
Answer:
1. Priority 1: Move East - Yes, the requirement is met. There is flat ground to the right of the agent since the player observed stone 3 steps to the east.

2. Priority 2: Do - Not yet, the player needs to move closer to the stone and face it. The wood_pickaxe is in the player's inventory, which fulfills the tool requirement.

3. Priority 3: Move East - Yes, the requirement is met, as there is flat ground to the right of the agent.

4. Priority 4: Move South - Yes, the requirement is met, as there is flat ground below the agent with a tree 1 step to the south.

5. Priority 5: Make Stone Pickaxe - Not currently, the player has wood and stone in their inventory, but they need to place or find a table in order to craft the stone pickaxe.

Question: Did the last player action succeed? Why?
Answer:
Yes, the last player action "do" succeeded because the player faced a tree and had a wood_pickaxe in their inventory. As a result, the player collected wood, which is now present in their inventory.

Question: Choose the best executable action from above.
Answer:
The best executable action from above is Priority 1: Move East, as its requirement of flat ground to the right of the agent is met.



============Step: 55, Cumulative Reward: 0.0============
You took action move_east.

You see:
- stone 2 steps to your east
- path 2 steps to your south-east
- tree 2 steps to your south-west
- lava 6 steps to your north-east
- coal 4 steps to your south-east
- iron 3 steps to your south-east
- cow 4 steps to your north-west

You face grass at your front (east).

Your status:
- health: 9/9
- food: 7/9
- drink: 7/9
- energy: 8/9

Your inventory:
- wood: 1
- stone: 2
- coal: 1
- wood_pickaxe: 1
------------ QA ------------
Question: Action
Answer:
Place Table

Question: List objects in the current observation. For each object, breifly answer what resource it provides and its requirement.
Answer:
1. Stone: Provides stone resource, requires a wood_pickaxe or better to mine.
2. Path: Provides a route for movement, no requirements.
3. Tree: Provides wood resource, requires no tools or any tool to collect.
4. Lava: Hazardous; avoid, it can cause damage or death, no requirements.
5. Coal: Provides coal resource, requires a wood_pickaxe or better to mine.
6. Iron: Provides iron resource, requires a stone_pickaxe or better to mine.
7. Cow: Provides food when defeated, no specific requirements.

Question: What was the last action taken by the player?
Answer:
The last action taken by the player was "move_east."

Question: For each object in the list, are the requirements met for interaction?
Answer:
1. Stone: Yes, the requirements are met as you have a wood_pickaxe in your inventory.
2. Path: Yes, the requirements are met as there are no specific requirements for moving along the path.
3. Tree: Yes, the requirements are met as you don't need any specific tool to collect wood from the tree.
4. Lava: The requirement to avoid interaction is met, as there's no need to interact with the lava.
5. Coal: Yes, the requirements are met as you have a wood_pickaxe in your inventory.
6. Iron: No, the requirements are not met as you need a stone_pickaxe to mine iron, but you only have a wood_pickaxe in your inventory.
7. Cow: Yes, the requirements are met as there are no specific requirements for interacting with a cow.

Question: List top 3 sub-tasks the player should follow. Indicate their priority out of 5.
Answer:
1. Craft a Stone Pickaxe (Priority: 5/5): This task is essential as it will allow you to mine iron, which will lead to crafting better tools and weapons.

2. Collect Coal (Priority: 4/5): This task is important because coal is needed for smelting iron, which will help you craft iron tools and weapons, enhancing your progress in the game.

3. Collect Iron (Priority: 3/5): This task is of moderate priority as it requires a Stone Pickaxe to mine the iron. Once mined, you will be able to smelt it and craft more advanced tools and weapons.

Question: What are the requirements for the top sub-task? What should the player do first?
Answer:
The top sub-task is to craft a Stone Pickaxe. The requirements for this task are:

1. Place Table: The player must have a table placed to perform crafting actions.
2. Collect Wood: The player needs enough wood in their inventory to create the pickaxe handle.
3. Collect Stone: The player needs enough stone in their inventory to create the pickaxe head.

The player should first place a Table if they don't have one in their immediate vicinity. Since the player has enough wood and stone in their inventory, they can proceed to craft a Stone Pickaxe by interacting with the placed Table.

Question: List top 5 actions the player should take and the requirement for each action. Choose ONLY from the list of all actions. Indicate their priority out of 5.
Answer:
1. Priority 1: Place Table (Requirement: Wood in inventory)
Since the player has enough wood in their inventory, they should place a Table to perform crafting actions.

2. Priority 2: Make Stone Pickaxe (Requirements: Nearby table, wood, and stone in inventory)
With the Table placed, the player can craft a Stone Pickaxe using the available wood and stone in their inventory.

3. Priority 3: Move East (Requirement: Flat ground right to the agent)
Moving east will bring the player closer to resources such as stone, coal, and iron. This will be helpful for future tasks and crafting.

4. Priority 4: Collect Drink (Requirement: No specific requirement)
Since the player's thirst level is 7/9, they should find a water source, such as a lake or a river, and drink from it to restore their thirst.

5. Priority 5: Collect Iron (Requirement: Make Stone Pickaxe)
As the player already crafted a Stone Pickaxe, they can now collect iron ore, which will be useful for crafting more advanced tools and weapons.

Question: For each action in the list, are the requirements met?
Answer:
1. Priority 1: Place Table (Requirement: Wood in inventory) - Yes, the requirement is met. The player has 1 wood in their inventory.

2. Priority 2: Make Stone Pickaxe (Requirements: Nearby table, wood, and stone in inventory) - No, the requirement is partially met. The player has 1 wood and 2 stone in their inventory, but they need to place the table first.

3. Priority 3: Move East (Requirement: Flat ground right to the agent) - Yes, the requirement is met. The player faces flat ground to the east, so they can move in that direction.

4. Priority 4: Collect Drink (Requirement: No specific requirement) - Yes, the player can collect a drink, but they need to find a water source such as a lake or river first.

5. Priority 5: Collect Iron (Requirement: Make Stone Pickaxe) - No, the requirement is not met yet. The player needs to make a Stone Pickaxe first. After completing Priority 2, this requirement will be met.

Question: Did the last player action succeed? Why?
Answer:
Yes, the last player action "move_east" succeeded because there was flat ground (grass) to the east of the player, allowing them to move in that direction without any obstacles.

Question: Choose the best executable action from above.
Answer:
The best executable action from the list above is Priority 1: Place Table (Requirement: Wood in inventory). Since the player has 1 wood in their inventory, they can place a table to perform crafting actions.



============Step: 56, Cumulative Reward: 0.0============
You took action place_table.

You see:
- stone 2 steps to your east
- path 2 steps to your south-east
- tree 2 steps to your south-west
- lava 6 steps to your north-east
- coal 4 steps to your south-east
- iron 3 steps to your south-east
- cow 3 steps to your west

You face grass at your front (east).

Your status:
- health: 9/9
- food: 7/9
- drink: 7/9
- energy: 8/9

Your inventory:
- wood: 1
- stone: 2
- coal: 1
- wood_pickaxe: 1
------------ QA ------------
Question: Action
Answer:
Move South

Question: List objects in the current observation. For each object, breifly answer what resource it provides and its requirement.
Answer:
1. Stone: Provides stone resource; requires a wood_pickaxe or better to mine it.
2. Path: Provides easy navigation; no requirements.
3. Tree: Provides wood resource; no specific requirements, but having an axe can speed up the process.
4. Lava: Hazard to avoid; provides a natural light source and can be used to smelt items with a furnace if collected with a bucket.
5. Coal: Provides fuel for smelting; requires a wood_pickaxe or better to mine it.
6. Iron: Provides essential material for better tools and weapons; requires a stone_pickaxe or better to mine it.
7. Cow: Provides food resource and leather; no specific requirements, but having a sword can make it easier to kill the cow.

Question: What was the last action taken by the player?
Answer:
The last action taken by the player was "place_table."

Question: For each object in the list, are the requirements met for interaction?
Answer:
1. Stone: Yes, requirements met, as you have a wood_pickaxe in your inventory.
2. Path: Yes, requirements met, as there are no specific requirements to navigate the path.
3. Tree: Yes, requirements met, as there are no specific requirements to interact with the tree, although having an axe would speed up the process.
4. Lava: Yes and No, requirements for interaction (avoidance) are met because you can stay away from it, but for collection (using a bucket for smelting), the requirement of a bucket is not met.
5. Coal: Yes, requirements met, as you have a wood_pickaxe in your inventory.
6. Iron: Yes, requirements met, as you have a stone_pickaxe in your inventory.
7. Cow: Yes, requirements met, as there are no specific requirements to interact with the cow, although having a sword would make it easier to kill the cow.

Question: List top 3 sub-tasks the player should follow. Indicate their priority out of 5.
Answer:
1. Collect Iron (Priority: 5/5): Mine the iron to craft better tools and weapons, as you already have a stone_pickaxe in your inventory.

2. Collect more wood (Priority: 4/5): Interact with the tree to gather wood resources needed for crafting tools, weapons, and other essential items.

3. Hunt the cow (Priority: 3/5): Kill the cow to gather food to replenish your food supply and leather for crafting armor. This action becomes more important as the food level decreases.

Question: What are the requirements for the top sub-task? What should the player do first?
Answer:
For the top sub-task, which is to collect iron, the requirements are as follows:

1. Have a proper tool: In this case, the player should have a stone_pickaxe or higher to mine iron effectively.

Since the player already has a stone_pickaxe in their inventory, they can proceed with the task of collecting iron.

The player should do the following first:

1. Move towards the iron deposit: Navigate towards the iron, which is 3 steps to the south-east. Once adjacent to the iron block, use the stone_pickaxe to mine the iron ore.

Question: List top 5 actions the player should take and the requirement for each action. Choose ONLY from the list of all actions. Indicate their priority out of 5.
Answer:
1. Priority 1: Move South-East (Action 4 - Move South): Requirement: Flat ground below the agent.
   Reason: To navigate towards the iron deposit in the south-east direction, the player should first move one step to the south.

2. Priority 2: Move East (Action 2 - Move East): Requirement: Flat ground right to the agent.
   Reason: After moving south, move one step to the east to reach the iron deposit.

3. Priority 3: Do (Action 5 - Do): Requirement: Facing iron ore and equipped with the necessary tool (stone_pickaxe).
   Reason: Use the stone_pickaxe to mine the iron ore once adjacent to it.

4. Priority 4: Move West (Action 1 - Move West): Requirement: Flat ground left to the agent.
   Reason: After collecting the iron ore, move back to the previously placed table, which will require moving one step to the west.

5. Priority 5: Make Iron Pickaxe (Action 13 - Make Iron Pickaxe): Requirement: Nearby table, furnace; wood, coal, and iron in inventory.
   Reason: With iron ore in inventory, the next priority is to craft a more durable and efficient iron pickaxe using the table and furnace.

Question: For each action in the list, are the requirements met?
Answer:
1. Priority 1: Move South-East (Action 4 - Move South): The requirement of having a flat ground below the agent is met, as there is no obstacle mentioned in the observation.

2. Priority 2: Move East (Action 2 - Move East): The requirement for this action is met as well, since the player will be on flat ground after moving south in the previous step.

3. Priority 3: Do (Action 5 - Do): The requirement is currently not met, as the player first needs to move south and east to reach the iron ore deposit. Once the player is adjacent to the iron ore deposit and equipped with the stone_pickaxe, the requirement will be met.

4. Priority 4: Move West (Action 1 - Move West): Once the player has moved south and east, this requirement will be met, as the player will be on flat ground after moving back to the table.

5. Priority 5: Make Iron Pickaxe (Action 13 - Make Iron Pickaxe): The player has a nearby table placed, but the furnace's presence is not mentioned. The coal and wood requirements are met, but the player first needs to mine the iron to meet the iron requirement. The action is partially viable, and the player should focus on mining the iron and setting up a furnace to meet all the requirements.

Question: Did the last player action succeed? Why?
Answer:
Yes, the last player action succeeded. The player placed a table because they had enough wood in their inventory (1 wood) and met the requirements to place the table.

Question: Choose the best executable action from above.
Answer:
The best executable action from the list above is Priority 1: Move South (Action 4 - Move South), as its requirement of having flat ground below the agent is currently met. This action takes the player towards the iron deposit.



============Step: 57, Cumulative Reward: 0.0============
You took action move_south.

You see:
- stone 3 steps to your north-east
- path 1 steps to your east
- tree 1 steps to your west
- coal 3 steps to your south-east
- iron 2 steps to your south-east
- cow 4 steps to your north-west

You face grass at your front (south).

Your status:
- health: 9/9
- food: 7/9
- drink: 7/9
- energy: 8/9

Your inventory:
- wood: 1
- stone: 2
- coal: 1
- wood_pickaxe: 1
------------ QA ------------
Question: Action
Answer:
Make Stone Pickaxe

Question: List objects in the current observation. For each object, breifly answer what resource it provides and its requirement.
Answer:
1. Stone: Provides stone resource; requires a wood pickaxe to collect.
2. Path: Provides a flat surface to travel on; no requirements.
3. Tree: Provides wood resource; can be collected with any type of pickaxe or bare hands.
4. Coal: Provides coal resource; requires a wood pickaxe to collect.
5. Iron: Provides iron resource; requires a stone pickaxe to collect.
6. Cow: Provides food; no specific tool requirement, can be defeated with any weapon or bare hands.

Question: What was the last action taken by the player?
Answer:
The last action taken by the player was move_south.

Question: For each object in the list, are the requirements met for interaction?
Answer:
1. Stone: Yes, the requirement of having a wood pickaxe is met (wood_pickaxe: 1 in inventory).
2. Path: Yes, no requirements to interact with a path.
3. Tree: Yes, no specific tool requirement, can be collected with any pickaxe (wood_pickaxe: 1 in inventory) or bare hands.
4. Coal: Yes, the requirement of having a wood pickaxe is met (wood_pickaxe: 1 in inventory).
5. Iron: No, the requirement for a stone pickaxe is not met (inventory shows only a wood_pickaxe, not a stone_pickaxe).
6. Cow: Yes, can be defeated with any weapon (including bare hands), so no specific tool requirement is needed.

Question: List top 3 sub-tasks the player should follow. Indicate their priority out of 5.
Answer:
1. Craft a stone pickaxe (Priority: 5/5)
   - Requirement: Nearby table (already placed) and stone in inventory (2 stones available).

2. Collect iron (Priority: 4/5)
   - Requirement: Requires a stone pickaxe (to be crafted to meet the requirement).

3. Increase food and drink levels (Priority: 3/5)
   - Requirement: Defeat a cow for food and find a water source to drink.
   - Alternatively, find other food sources (such as growing fruits) to increase food level.

Question: What are the requirements for the top sub-task? What should the player do first?
Answer:
Top sub-task: Craft a stone pickaxe
Requirements:
- Nearby table: Already placed in the previous steps.
- Stone in inventory: The player has 2 stones available in their inventory.

What should the player do first:
- The player should craft a stone pickaxe using the available stone resources and the placed table. To do that, they must stand close to the table and use the "make_stone_pickaxe" action.

Question: List top 5 actions the player should take and the requirement for each action. Choose ONLY from the list of all actions. Indicate their priority out of 5.
Answer:
1. Priority 1: Make Stone Pickaxe
   Requirement: Nearby table and stone in inventory (Player has 2 stones and a table placed nearby) 

2. Priority 2: Move East
   Requirement: Flat ground right to the agent (Player's observation mentions a path 1 step to the east, which is flat ground)

3. Priority 3: Move South-East
   Requirement: Flat ground southeast of the agent (Player's observation mentions coal and iron ores in the south-east direction)

4. Priority 4: Do (Mine Coal and Iron)
   Requirement: Facing coal or iron ore (Player should face the coal or iron ore they encountered while moving south-east); have a stone pickaxe (Player crafts a stone pickaxe in priority 1 action)

5. Priority 5: Move to Furnace and Table
   Requirement: Flat ground leading back to the furnace and table placed earlier (Player should retrace their steps to reach the furnace and table to smelt the iron)

Question: For each action in the list, are the requirements met?
Answer:
1. Priority 1: Make Stone Pickaxe
   Requirements met: Yes, the player has a table placed nearby and has 2 stones in their inventory.

2. Priority 2: Move East
   Requirements met: Yes, the player's observation mentions a path 1 step to the east, which is flat ground.

3. Priority 3: Move South-East
   Requirements met: It cannot be determined from the current observations if there is flat ground southeast. However, the player can move east and then south to reach the same location.

4. Priority 4: Do (Mine Coal and Iron)
   Requirements met: Not yet, the player needs to craft a stone pickaxe (Priority 1) and then move to the coal or iron ore location (Priority 3).

5. Priority 5: Move to Furnace and Table
   Requirements met: It cannot be determined from the current observations if there is a direct flat ground leading back to the furnace and table. However, the player can retrace their steps to reach the furnace and table, fulfilling this requirement after performing the previous actions.

Question: Did the last player action succeed? Why?
Answer:
Yes, the last player action (move_south) succeeded because there was flat ground (grass) in the direction the player wanted to move, which met the requirement for this action.

Question: Choose the best executable action from above.
Answer:
The best executable action from the above list is Priority 1: Make Stone Pickaxe, as its requirements are met with a nearby table and 2 stones in the player's inventory.



============Step: 58, Cumulative Reward: 0.0============
You took action make_stone_pickaxe.

You see:
- stone 3 steps to your north-east
- path 1 steps to your east
- tree 1 steps to your west
- coal 3 steps to your south-east
- iron 2 steps to your south-east
- cow 4 steps to your north-west

You face grass at your front (south).

Your status:
- health: 9/9
- food: 7/9
- drink: 7/9
- energy: 8/9

Your inventory:
- wood: 1
- stone: 2
- coal: 1
- wood_pickaxe: 1
------------ QA ------------
Question: Action
Answer:
Move South

Question: List objects in the current observation. For each object, breifly answer what resource it provides and its requirement.
Answer:
1. Stone: Provides stone resource; requires a wood_pickaxe or better to obtain.
2. Path: No direct resources; allows easier movement and navigation.
3. Tree: Provides wood resource; requires no tools to obtain but can be collected faster with an axe.
4. Coal: Provides coal resource; requires a wood_pickaxe or better to obtain.
5. Iron: Provides iron resource; requires a stone_pickaxe or better to obtain.
6. Cow: Provides food when killed; requires a weapon to effectively defeat, but can be killed without tools.

Question: What was the last action taken by the player?
Answer:
The last action taken by the player was "make_stone_pickaxe."

Question: For each object in the list, are the requirements met for interaction?
Answer:
1. Stone: Yes, the requirements are met as you have a wood_pickaxe in your inventory.
2. Path: Yes, no tool requirements are necessary to interact with the path, and you can simply move onto it.
3. Tree: Yes, the requirements are met, and you can interact with the tree even without any tools.
4. Coal: Yes, the requirements are met as you have a wood_pickaxe in your inventory.
5. Iron: Yes, the requirements are met because you just made a stone_pickaxe.
6. Cow: Yes, the requirements are met as you can interact and defeat the cow with or without a weapon (although using a weapon would be more efficient).

Question: List top 3 sub-tasks the player should follow. Indicate their priority out of 5.
Answer:
1. Collect Iron (Priority: 5/5): Collect iron using the stone_pickaxe, as it is a valuable resource for crafting stronger tools and weapons.
2. Collect more Wood (Priority: 4/5): Collect more wood from the nearby tree, as it is a necessary resource for crafting various objects and building structures.
3. Defeat Cow (Priority: 3/5): Defeat the cow to obtain food to replenish your food level when needed.

Question: What are the requirements for the top sub-task? What should the player do first?
Answer:
Requirements for the top sub-task (Collect Iron):
- Have a stone pickaxe.

The player already has a stone_pickaxe in their inventory, so they should do the following first:
- Move towards the iron by taking action move_southeast.
- Once near the iron, use the stone_pickaxe to collect the iron.

Question: List top 5 actions the player should take and the requirement for each action. Choose ONLY from the list of all actions. Indicate their priority out of 5.
Answer:
1. Priority 1 - Move Southeast: Flat ground adjacent to the agent in the southeast direction.
   Requirement: Move closer to the iron and coal.

2. Priority 2 - Do (Collect Iron): Face the iron and have a stone_pickaxe in the inventory.
   Requirement: Collect iron to eventually make an iron pickaxe and sword.

3. Priority 3 - Move West: Flat ground adjacent to the agent in the west direction.
   Requirement: Move closer to the tree after collecting iron.

4. Priority 4 - Do (Collect Wood): Face the tree and have a wood or stone pickaxe in the inventory.
   Requirement: Collect more wood for crafting tools, weapons or building a table.

5. Priority 5 - Move Northwest: Flat ground adjacent to the agent in the northwest direction.
   Requirement: Move closer to the cow, to eventually chase it or eat it for restoring health.

Question: For each action in the list, are the requirements met?
Answer:
1. Priority 1 - Move Southeast: Yes, the requirement is met because the player only needs flat ground in the southeast direction, which is available.

2. Priority 2 - Do (Collect Iron): No, the requirement is not yet met because the player first needs to move closer to the iron and equip the stone_pickaxe to collect iron.

3. Priority 3 - Move West: Yes, the requirement is met because the player only needs flat ground in the west direction, which is available.

4. Priority 4 - Do (Collect Wood): No, the requirement is not yet met because the player first needs to move closer to the tree, and also equip either a wood or stone pickaxe to collect the wood.

5. Priority 5 - Move Northwest: Yes, the requirement is met because the player only needs flat ground in the northwest direction, which is available.

Question: Did the last player action succeed? Why?
Answer:
I cannot determine if the last player action succeeded because the observation for Step 115 is missing. The observation jumps from Step 114 to Step 116, skipping the step where the action "make_stone_pickaxe" would be hinged on its success or failure.

Question: Choose the best executable action from above.
Answer:
The best executable action from the list above is Priority 1 - Move Southeast because the requirement is met and it brings the player closer to important resources such as iron and coal.



============Step: 59, Cumulative Reward: 0.0============
You took action move_south.

You see:
- stone 3 steps to your east
- path 2 steps to your north-east
- tree 2 steps to your north-west
- coal 2 steps to your east
- iron 1 steps to your east
- table 4 steps to your south-east
- cow 4 steps to your north-west

You face grass at your front (south).

Your status:
- health: 9/9
- food: 7/9
- drink: 7/9
- energy: 8/9

Your inventory:
- wood: 1
- stone: 2
- coal: 1
- wood_pickaxe: 1
------------ QA ------------
Question: Action
Answer:
Move East

Question: List objects in the current observation. For each object, breifly answer what resource it provides and its requirement.
Answer:
Stone: Provides stone resources; requires a wood_pickaxe or better to mine.
Path: Provides easier movement; no requirements.
Tree: Provides wood resources; requires a wood_pickaxe or better to chop.
Coal: Provides coal resources; requires a wood_pickaxe or better to mine.
Iron: Provides iron resources; requires a stone_pickaxe or better to mine.
Table: Provides a crafting surface; no requirements.
Cow: Provides food (meat) when killed; no specific requirement to kill, but a weapon (e.g., wood_sword, stone_sword) is recommended.

Question: What was the last action taken by the player?
Answer:
The last action taken by the player was "move_south".

Question: For each object in the list, are the requirements met for interaction?
Answer:
Stone: Yes, requirements met. You have a wood_pickaxe in your inventory.
Path: Yes, requirements met. No specific requirements for paths.
Tree: Yes, requirements met. You have a wood_pickaxe in your inventory.
Coal: Yes, requirements met. You have a wood_pickaxe in your inventory.
Iron: Yes, requirements met. Although not mentioned, we can assume you have crafted a stone_pickaxe after taking the action "make_stone_pickaxe" in Step 116.
Table: Yes, requirements met. No specific requirements for tables.
Cow: Yes, requirements met. No specific requirement to kill a cow, but having a weapon would help. Since you have collected wood and stone before, we can assume you have the necessary resources to craft a weapon if needed.

Question: List top 3 sub-tasks the player should follow. Indicate their priority out of 5.
Answer:
1. Priority 5: Collect Iron using the stone_pickaxe to advance in crafting tools and weapons, as iron is a valuable resource.
2. Priority 4: Mine Coal with the wood_pickaxe to gather resources for smelting iron and other materials.
3. Priority 3: Chop down the Tree using the wood_pickaxe to gather more wood, which is essential for crafting tools, weapons, and building structures.

Question: What are the requirements for the top sub-task? What should the player do first?
Answer:
For the top sub-task (Collect Iron using the stone_pickaxe), the requirements are:
1. Having a stone_pickaxe in the inventory.

The player already has a stone_pickaxe in their inventory. To begin the top sub-task, the player should first move towards the iron, which is currently one step to the player's east. The player should take action: move_east.

Question: List top 5 actions the player should take and the requirement for each action. Choose ONLY from the list of all actions. Indicate their priority out of 5.
Answer:
1. Priority 1: Move East (Requirement: Flat ground right to the agent)
Action: The player should move east to approach the iron.

2. Priority 2: Do (Requirement: Facing creature or material; have the necessary tool)
Action: Once the player reaches the iron and faces it, they should use the stone_pickaxe to collect the iron.

3. Priority 3: Move West (Requirement: Flat ground left to the agent)
Action: After collecting the iron, move west to return to the starting position near the table.

4. Priority 4: Move South (Requirement: Flat ground below the agent)
Action: Move south towards the table to be in its proximity to craft items.

5. Priority 5: Make Iron Pickaxe (Requirement: Nearby table, furnace; wood, coal, and iron in inventory)
Action: Once near the table and having collected the required resources, the player should make an iron pickaxe.

Question: For each action in the list, are the requirements met?
Answer:
1. Priority 1: Move East (Requirement: Flat ground right to the agent)
- Yes, the requirement is met. The player can move east towards the iron.

2. Priority 2: Do (Requirement: Facing creature or material; have the necessary tool)
- No, the requirement is not fully met yet. The player must move east and face the iron before this action can be taken. The player has the necessary tool (stone_pickaxe) in their inventory.

3. Priority 3: Move West (Requirement: Flat ground left to the agent)
- Yes, the requirement is met. However, this action should only be taken after completing action 2 (collecting the iron).

4. Priority 4: Move South (Requirement: Flat ground below the agent)
- Yes, the requirement is met. The player can move south after completing action 3.

5. Priority 5: Make Iron Pickaxe (Requirement: Nearby table, furnace; wood, coal, and iron in inventory)
- No, the requirement is not fully met yet. The player needs to collect iron first (action 2) and be near the table (action 4) to complete this action. The player has wood and coal in their inventory already.

Question: Did the last player action succeed? Why?
Answer:
Yes, the last player action "move_south" succeeded because there was flat ground below the agent, allowing them to move in that direction.

Question: Choose the best executable action from above.
Answer:
The best executable action from above is Priority 1: Move East. The requirement of flat ground right to the agent is met, and moving east will bring the player closer to the iron.



============Step: 60, Cumulative Reward: 0.0============
You took action move_east.

You see:
- grass 1 steps to your west
- stone 3 steps to your east
- path 2 steps to your north-east
- tree 2 steps to your north-west
- coal 2 steps to your east
- table 4 steps to your south-east
- cow 5 steps to your north-west

You face iron at your front (east).

Your status:
- health: 9/9
- food: 7/9
- drink: 7/9
- energy: 8/9

Your inventory:
- wood: 1
- stone: 2
- coal: 1
- wood_pickaxe: 1
------------ QA ------------
Question: Action
Answer:
Move South

Question: List objects in the current observation. For each object, breifly answer what resource it provides and its requirement.
Answer:
1. Stone: Provides stone resource, requires a wood_pickaxe or better to collect.
2. Path: No resource provided, no requirements, helps you navigate through the game world.
3. Tree: Provides wood resource, requires no specific tool but any type of axe will speed up the process.
4. Coal: Provides coal resource, requires a wood_pickaxe or better to collect.
5. Iron: Provides iron resource, requires a stone_pickaxe or better to collect.
6. Table: Provides a crafting interface, no direct resource, has a requirement of having wood in the inventory to place.
7. Cow: Provides food (in the form of beef) when killed, no specific tool required but weapons like swords make the process faster.
8. Grass: No resource provided, no requirements, serves as a flat surface in the game world.

Question: What was the last action taken by the player?
Answer:
The last action taken by the player was move_east.

Question: For each object in the list, are the requirements met for interaction?
Answer:
1. Stone: Yes, the requirements are met as you have a wood_pickaxe in your inventory.
2. Path: Yes, the requirements are met as there are no specific requirements to interact with a path.
3. Tree: Yes, the requirements are met as there is no specific requirement for collection, although having an axe would make the process faster.
4. Coal: Yes, the requirements are met as you have a wood_pickaxe in your inventory.
5. Iron: Yes, the requirements are met as you have a wood_pickaxe in your inventory, but a stone_pickaxe is recommended for more efficient collection.
6. Table: The requirements are met if you want to interact with the table since you have wood in your inventory, but if you want to place a new table, you may need more wood.
7. Cow: Yes, the requirements are met as there's no specific requirement for interaction; however, having a weapon like a sword is more efficient.
8. Grass: Yes, the requirements are met as there are no specific requirements to interact with grass.

Question: List top 3 sub-tasks the player should follow. Indicate their priority out of 5.
Answer:
1. Craft a Stone Pickaxe (Priority: 5/5): Crafting a stone pickaxe is essential for collecting iron more efficiently. Since you already have the required wood and stone in your inventory, you can use the nearby table to craft the pickaxe.

2. Collect Iron (Priority: 4/5): With your new stone_pickaxe, collect the iron in front of you. Iron is valuable for crafting more advanced tools and weapons, such as an iron pickaxe or iron sword.

3. Gather More Food and Drink (Priority: 3/5): Your food and drink levels are at 7/9 each. Though this is not an immediate concern, replenishing these resources by finding more cows and water sources should be kept in mind to maintain your health and efficiency in the game.

Question: What are the requirements for the top sub-task? What should the player do first?
Answer:
The requirements for the top sub-task (Craft a Stone Pickaxe) are:

1. Collect Wood: You already have 1 wood in your inventory.
2. Collect Stone: You already have 2 stone in your inventory.
3. Make a Wood Pickaxe: You already have a wood_pickaxe in your inventory.
4. Place Table: You see a table 4 steps to your south-east, which you can use for crafting.

What the player should do first:
Since you already have the required materials (wood and stone) and a table nearby, you can move towards the table and craft a Stone Pickaxe. To do this, move 4 steps towards the south-east to reach the table, and then craft the Stone Pickaxe.

Question: List top 5 actions the player should take and the requirement for each action. Choose ONLY from the list of all actions. Indicate their priority out of 5.
Answer:
1. Priority 1: Move South - Requirement: Flat ground below the agent.
2. Priority 2: Move East - Requirement: Flat ground right to the agent.
3. Priority 3: Move South - Requirement: Flat ground below the agent.
4. Priority 4: Move East - Requirement: Flat ground right to the agent.
5. Priority 5: Make Stone Pickaxe - Requirement: Nearby table; wood, stone in inventory.

Question: For each action in the list, are the requirements met?
Answer:
1. Priority 1: Move South - Requirement: Flat ground below the agent. (Yes, the agent faces grass at the front which is considered flat ground)
2. Priority 2: Move East - Requirement: Flat ground right to the agent. (Yes, there is grass 1 step to the west, which means there is flat ground to move to the east)
3. Priority 3: Move South - Requirement: Flat ground below the agent. (Can't determine from the current observation, as the player needs to perform priority 1 and 2 before this action)
4. Priority 4: Move East - Requirement: Flat ground right to the agent. (Can't determine from the current observation, as the player needs to perform priority 1 to 3 before this action)
5. Priority 5: Make Stone Pickaxe - Requirement: Nearby table; wood, stone in inventory. (Partial, the agent has wood and stone in the inventory, but the agent needs to move closer to the table which is 4 steps to the south-east)

Question: Did the last player action succeed? Why?
Answer:
Yes, the last player action (move_east) succeeded because there was flat ground (grass) to the east of the player, allowing them to move in that direction.

Question: Choose the best executable action from above.
Answer:
The best executable action from above is Priority 1: Move South, as its requirement (flat ground below the agent) is met.



============Step: 61, Cumulative Reward: 0.0============
You took action move_south.

You see:
- stone 4 steps to your south-east
- path 1 steps to your east
- tree 3 steps to your north-west
- coal 3 steps to your north-east
- iron 2 steps to your north-east
- table 3 steps to your south-east

You face grass at your front (south).

Your status:
- health: 9/9
- food: 7/9
- drink: 7/9
- energy: 7/9

Your inventory:
- wood: 1
- stone: 2
- coal: 1
- wood_pickaxe: 1
------------ QA ------------
Question: Action
Answer:
Move South

Question: List objects in the current observation. For each object, breifly answer what resource it provides and its requirement.
Answer:
1. Stone: Provides stone resource for crafting tools and building; requires a wood_pickaxe or better to collect.
2. Path: Provides easier movement and navigation; no requirement to interact with.
3. Tree: Provides wood resource for crafting tools and building; requires any type of pickaxe to collect.
4. Coal: Provides coal resource for smelting and crafting; requires a wood_pickaxe or better to collect.
5. Iron: Provides iron resource for crafting advanced tools and weapons; requires a stone_pickaxe or better to collect.
6. Table: Provides a crafting surface for creating tools and weapons; no requirement to interact with.
7. Grass: No resource, only an indicator for open area or flat ground; no requirement to interact with.
8. Cow: Provides food resource upon defeating; no requirement to interact with other than using tools or weapons to defeat.

Question: What was the last action taken by the player?
Answer:
The last action taken by the player was move_south.

Question: For each object in the list, are the requirements met for interaction?
Answer:
1. Stone: Yes, the requirement is met. The player has a wood_pickaxe in their inventory, which is suitable for collecting stone.
2. Path: Yes, there is no requirement to interact with the path. The player can simply walk on it.
3. Tree: Yes, the requirement is met. The player has a wood_pickaxe in their inventory, which can be used to collect wood from the tree.
4. Coal: Yes, the requirement is met. The player has a wood_pickaxe in their inventory, which is suitable for collecting coal.
5. Iron: No, the requirement is not met. The player needs a stone_pickaxe or better to collect iron, but they currently only have a wood_pickaxe in their inventory.
6. Table: Yes, there is no requirement to interact with the table. The player can use it to craft tools and weapons if they are near it.
7. Grass: Yes, there is no requirement to interact with grass. The player can simply walk on it or ignore it.
8. Cow: Yes, the requirement is met. The player can use their wood_pickaxe or other weapons to defeat the cow and obtain food.

Question: List top 3 sub-tasks the player should follow. Indicate their priority out of 5.
Answer:
1. Craft a stone_pickaxe (priority: 5) - The player should craft a stone_pickaxe to be able to collect iron, which is essential for making better tools and weapons.

2. Collect more wood (priority: 4) - The player should collect more wood to ensure they have enough resources to craft additional tools, weapons, and structures such as a furnace.

3. Defeat the cow and obtain food (priority: 3) - The player's food level is not critical yet, but defeating the cow and obtaining food would be beneficial for maintaining their food level.

Question: What are the requirements for the top sub-task? What should the player do first?
Answer:
The requirements for crafting a stone_pickaxe are:
1. Be near a table.
2. Have wood and stone in the inventory.

The player already has wood and stone in their inventory. The first thing the player should do is move towards the table, which is 3 steps to the south-east. Once the player is near the table, they can craft the stone_pickaxe.

Question: List top 5 actions the player should take and the requirement for each action. Choose ONLY from the list of all actions. Indicate their priority out of 5.
Answer:
1. Priority 1: Move South-East (Requirement: Flat ground in the South-East direction)
2. Priority 2: Move South-East (Requirement: Flat ground in the South-East direction)
3. Priority 3: Move South-East (Requirement: Flat ground in the South-East direction)
4. Priority 4: Make Stone Pickaxe (Requirement: Nearby table; wood and stone in inventory)
5. Priority 5: Move North-East (Requirement: Flat ground in the North-East direction)

Question: For each action in the list, are the requirements met?
Answer:
1. Priority 1: Move South-East (Requirement: Flat ground in the South-East direction) - Yes, there is a path 1 step to your East.
2. Priority 2: Move South-East (Requirement: Flat ground in the South-East direction) - Cannot determine from the current observation as it is beyond visible range; take Priority 1 action first.
3. Priority 3: Move South-East (Requirement: Flat ground in the South-East direction) - Cannot determine from the current observation as it is beyond visible range; take Priority 1 and 2 actions first.
4. Priority 4: Make Stone Pickaxe (Requirement: Nearby table; wood and stone in inventory) - Yes, the table is 3 steps to your South-East, and you have wood and stone in your inventory.
5. Priority 5: Move North-East (Requirement: Flat ground in the North-East direction) - Yes, there is a path 1 step to your East, which is diagonally North-East.

Question: Did the last player action succeed? Why?
Answer:
Yes, the last player action succeeded. The player moved south because there was flat ground (grass) in that direction, which met the requirement for moving south: "Move South: Flat ground below the agent."

Question: Choose the best executable action from above.
Answer:
The best executable action from above is Priority 1: Move South-East (Requirement: Flat ground in the South-East direction).



============Step: 62, Cumulative Reward: 0.0============
You took action move_south.

You see:
- stone 3 steps to your south-east
- path 2 steps to your north-east
- tree 4 steps to your north-west
- coal 4 steps to your north-east
- iron 3 steps to your north-east
- table 2 steps to your south-east

You face grass at your front (south).

Your status:
- health: 9/9
- food: 7/9
- drink: 6/9
- energy: 7/9

Your inventory:
- wood: 1
- stone: 2
- coal: 1
- wood_pickaxe: 1
------------ QA ------------
Question: Action
Answer:
Move South

Question: List objects in the current observation. For each object, breifly answer what resource it provides and its requirement.
Answer:
1. Stone: Provides stone resource, requires a wood or higher-level pickaxe to collect.
2. Path: No resource provided, provides easier movement without energy consumption.
3. Tree: Provides wood resource, requires an axe or interaction with hands to collect.
4. Coal: Provides coal resource, requires a wood or higher-level pickaxe to collect.
5. Iron: Provides iron resource, requires a stone or higher-level pickaxe to collect.
6. Table: No resource to collect, but provides the ability to craft pickaxes and swords using collected resources.

Question: What was the last action taken by the player?
Answer:
The last action taken by the player was move_south.

Question: For each object in the list, are the requirements met for interaction?
Answer:
1. Stone: Yes, the requirements are met for interaction as you have a wood_pickaxe in your inventory.
2. Path: Yes, the requirements are met for interaction as no specific tool or item is required to use the path.
3. Tree: Yes, the requirements are met for interaction since you can collect wood using your hands, although having an axe would be more efficient.
4. Coal: Yes, the requirements are met for interaction as you have a wood_pickaxe in your inventory.
5. Iron: No, the requirements are not met for interaction as you need a stone or higher-level pickaxe, but you only have a wood_pickaxe.
6. Table: Yes, the requirements are met for interaction since having collected resources and being near the table allows crafting of pickaxes and swords.

Question: List top 3 sub-tasks the player should follow. Indicate their priority out of 5.
Answer:
1. Craft a stone pickaxe (Priority: 5/5): Collect more stone using your wood_pickaxe, then craft a stone pickaxe at the table. It will allow you to collect iron and be more efficient.

2. Collect more wood from the tree (Priority: 4/5): Collecting more wood will help you craft additional tools or build structures, and it's useful for smelting iron in the furnace later.

3. Gather coal (Priority: 3/5): Collect coal using your wood_pickaxe as it can be used as a fuel source for the furnace and is necessary for smelting iron.

Note: While priorities are given out of 5, these top 3 sub-tasks have different levels of importance relative to each other.

Question: What are the requirements for the top sub-task? What should the player do first?
Answer:
For the top sub-task (Craft a stone pickaxe), the requirements are:

1. Collect more stone: You already have a wood_pickaxe in your inventory and two stones, so you'll need to collect additional stones (a total of 3 stones are required). Locate and mine stone by moving closer to it and using your wood_pickaxe.

2. Craft a stone pickaxe at the table: You need to be near the table to craft the stone pickaxe. The table is 2 steps to your south-east.

To complete this top sub-task, the player should first move towards the stone (3 steps to your south-east) and use the wood_pickaxe to collect more stone. Once enough stone is collected (total of 3), move towards the table (2 steps south-east from your current position) and craft the stone pickaxe.

Question: List top 5 actions the player should take and the requirement for each action. Choose ONLY from the list of all actions. Indicate their priority out of 5.
Answer:
1. Priority 1: Move South-East (Action: Move South)
   Requirement: Flat ground below of the agent.

2. Priority 2: Move East (Action: Move East)
   Requirement: Flat ground right to the agent.

3. Priority 3: Do (Use Wood Pickaxe on stone)
   Requirement: Facing stone; have a wood_pickaxe in the inventory.

4. Priority 4: Move West (Action: Move West)
   Requirement: Flat ground left to the agent.

5. Priority 5: Make Stone Pickaxe (Action: Make Stone Pickaxe)
   Requirement: Nearby table; wood and stone in inventory.

Question: For each action in the list, are the requirements met?
Answer:
1. Priority 1: Move South-East (Action: Move South)
   Requirement: Flat ground below of the agent.
   - Yes, the requirement is met since the player is currently facing grass (flat ground) to the south.

2. Priority 2: Move East (Action: Move East)
   Requirement: Flat ground right to the agent.
   - Yes, the requirement is met since there is a path (flat ground) 1 step to the east in Step 122 observation.

3. Priority 3: Do (Use Wood Pickaxe on stone)
   Requirement: Facing stone; have a wood_pickaxe in the inventory.
   - No, the requirement is not met as the player is not currently facing stone.

4. Priority 4: Move West (Action: Move West)
   Requirement: Flat ground left to the agent.
   - Not enough information to determine if the requirement is met, as there is no mention of flat ground to the west.

5. Priority 5: Make Stone Pickaxe (Action: Make Stone Pickaxe)
   Requirement: Nearby table; wood and stone in inventory.
   - No, the requirement is not met as the player is not currently near the table (2 steps to the south-east in Step 124 observation).

Question: Did the last player action succeed? Why?
Answer:
Yes, the last player action (move_south) succeeded because there was flat ground (grass) to the south of the player's position, allowing them to move in that direction.

Question: Choose the best executable action from above.
Answer:
The best executable action from the above list is Priority 1: Move South-East (Action: Move South), as the requirement of having flat ground below the agent is met.



============Step: 63, Cumulative Reward: 0.0============
You took action move_south.

You see:
- stone 2 steps to your east
- path 3 steps to your north-east
- tree 3 steps to your south-east
- coal 5 steps to your north-east
- iron 4 steps to your north-east
- table 1 steps to your east

You face grass at your front (south).

Your status:
- health: 9/9
- food: 7/9
- drink: 6/9
- energy: 7/9

Your inventory:
- wood: 1
- stone: 2
- coal: 1
- wood_pickaxe: 1
------------ QA ------------
Question: Action
Answer:
Move East

Question: List objects in the current observation. For each object, breifly answer what resource it provides and its requirement.
Answer:
1. Stone: Provides stone resource, requires a wood_pickaxe or better to collect.
2. Path: No resources provided, serves as a convenient route to travel.
3. Tree: Provides wood resource, requires no tools to collect but a tool like an axe would speed up the process.
4. Coal: Provides coal resource, requires a wood_pickaxe or better to collect.
5. Iron: Provides iron resource, requires a stone_pickaxe or better to collect.
6. Table: Provides a crafting station for making tools and weapons, requires wood to craft.

Question: What was the last action taken by the player?
Answer:
The last action taken by the player was move_south.

Question: For each object in the list, are the requirements met for interaction?
Answer:
Based on the current status and inventory, here is the list of objects and whether their requirements for interaction are met:

1. Stone: Yes, you have a wood_pickaxe in your inventory which is required to collect stone.
2. Path: Yes, no specific requirements needed for walking on the path.
3. Tree: Yes, no tools are required to collect wood from the tree, although having an axe would make the process faster.
4. Coal: Yes, you have a wood_pickaxe in your inventory which is required to collect coal.
5. Iron: No, you need a stone_pickaxe or better to collect iron, and you only have a wood_pickaxe in your inventory.
6. Table: Yes, you already have a table nearby, so you can use it for crafting if you have the required resources.

Question: List top 3 sub-tasks the player should follow. Indicate their priority out of 5.
Answer:
1. Craft a Stone Pickaxe (Priority: 5/5)
   To collect iron and other valuable resources, you need to craft a stone pickaxe using the nearby table and the stone you have in your inventory.

2. Collect more Stone (Priority: 4/5)
   As you currently have only 2 pieces of stone in your inventory, you should collect more to craft additional stone tools and weapons.

3. Collect Wood from the Tree (Priority: 3/5)
   Collect more wood from the nearby tree to craft more tools, weapons, and build other necessary structures. Though you currently have 1 piece of wood, having more wood will help ensure you have enough resources for future needs.

Question: What are the requirements for the top sub-task? What should the player do first?
Answer:
Requirements for the top sub-task (Craft a Stone Pickaxe):
1. Wood: You already have 1 piece of wood in your inventory, which is enough for crafting a stone pickaxe.
2. Stone: You have 2 pieces of stone in your inventory, which is enough for crafting a stone pickaxe.
3. A nearby table: There's a table 1 step to your east.

What should the player do first:
Move one step to the east to access the table, and then craft a stone pickaxe using the wood and stone in your inventory.

Question: List top 5 actions the player should take and the requirement for each action. Choose ONLY from the list of all actions. Indicate their priority out of 5.
Answer:
1. Priority 1: Move East - Flat ground right of the agent.
   Requirement: Must be standing directly left of the table.

2. Priority 2: Make Stone Pickaxe - Nearby table; wood, stone in inventory.
   Requirement: Must be near the table, and have enough wood and stone in the inventory.

3. Priority 3: Move North - Flat ground above the agent.
   Requirement: There should be a flat ground above the agent.

4. Priority 4: Do (mine coal) - Facing coal; have necessary tool (stone pickaxe).
   Requirement: Must be facing coal and have a stone pickaxe in the inventory.

5. Priority 5: Move East - Flat ground right of the agent.
   Requirement: There should be a flat ground right of the agent.

Question: For each action in the list, are the requirements met?
Answer:
1. Priority 1: Move East - Flat ground right of the agent.
   Requirement met: Yes, the player is directly left of the table.

2. Priority 2: Make Stone Pickaxe - Nearby table; wood, stone in inventory.
   Requirement met: Yes, the player is near the table and has enough wood and stone in the inventory.

3. Priority 3: Move North - Flat ground above the agent.
   Requirement met: The observation doesn't provide information about the terrain above the agent, so we cannot determine if this requirement is met.

4. Priority 4: Do (mine coal) - Facing coal; have necessary tool (stone pickaxe).
   Requirement met: No, the player does not currently have a stone pickaxe in their inventory; they need to make one first.

5. Priority 5: Move East - Flat ground right of the agent.
   Requirement met: The observation doesn't provide information about the terrain right of the agent, so we cannot determine if this requirement is met.

Question: Did the last player action succeed? Why?
Answer:
Yes, the last player action of move_south succeeded. There were no obstacles or requirements preventing the player from moving in that direction, and the player's observation in Step 126 shows that they had successfully moved south.

Question: Choose the best executable action from above.
Answer:
The best executable action from above is:

Priority 1: Move East - Flat ground right of the agent.
Requirement met: Yes, the player is directly left of the table.



============Step: 64, Cumulative Reward: 0.0============
You took action move_east.

You see:
- grass 1 steps to your west
- stone 2 steps to your east
- path 3 steps to your north-east
- tree 3 steps to your south-east
- coal 5 steps to your north-east
- iron 4 steps to your north-east

You face table at your front (east).

Your status:
- health: 9/9
- food: 7/9
- drink: 6/9
- energy: 7/9

Your inventory:
- wood: 1
- stone: 2
- coal: 1
- wood_pickaxe: 1
------------ QA ------------
Question: Action
Answer:
Make Stone Pickaxe

Question: List objects in the current observation. For each object, breifly answer what resource it provides and its requirement.
Answer:
1. Grass: Provides no resource. No requirement.
2. Stone: Provides stone resource. Requires wood_pickaxe to collect.
3. Path: Provides no resource. No requirement.
4. Tree: Provides wood resource. Requires an axe or wood_pickaxe to collect.
5. Coal: Provides coal resource. Requires wood_pickaxe or stone_pickaxe to collect.
6. Iron: Provides iron resource. Requires stone_pickaxe to collect.
7. Table: Provides a surface for crafting tools and weapons. No requirement to use.

Question: What was the last action taken by the player?
Answer:
The last action taken by the player was "move_east."

Question: For each object in the list, are the requirements met for interaction?
Answer:
1. Grass: No requirements for interaction. Requirements met.
2. Stone: Requires wood_pickaxe to collect. Requirements met (wood_pickaxe is in inventory).
3. Path: No requirements for interaction. Requirements met.
4. Tree: Requires an axe or wood_pickaxe to collect. Requirements met (wood_pickaxe is in inventory).
5. Coal: Requires wood_pickaxe or stone_pickaxe to collect. Requirements met (wood_pickaxe is in inventory).
6. Iron: Requires stone_pickaxe to collect. Requirements not met (stone_pickaxe is not in inventory).
7. Table: No requirement to use. Requirements met.

Question: List top 3 sub-tasks the player should follow. Indicate their priority out of 5.
Answer:
1. Priority: 5 - Craft a Stone Pickaxe: To collect iron, the player needs a stone_pickaxe. The player has enough stone and a table to craft it.
2. Priority: 4 - Collect Iron: Once the player crafts the stone_pickaxe, they should collect iron to craft more advanced tools and weapons.
3. Priority: 3 - Find a Water Source: As the player's drink level is 6/9, they should look for a water source to restore their thirst.

Question: What are the requirements for the top sub-task? What should the player do first?
Answer:
The requirements for the top sub-task (Craft a Stone Pickaxe) are:

- A nearby table: The player needs a table to craft tools, and they are already facing a table in their current observation.
- Stone in inventory: The player needs stone to craft the stone pickaxe, and they have 2 stone in their inventory.

The player already has all the required items and is in the right position for the top sub-task. The player should now craft a Stone Pickaxe by using the table in front of them.

Question: List top 5 actions the player should take and the requirement for each action. Choose ONLY from the list of all actions. Indicate their priority out of 5.
Answer:
1. Priority 1: Make Stone Pickaxe (Requirement: Nearby table; wood, stone in inventory)
   - The player is already facing the table and has the required wood and stone in their inventory. Crafting a Stone Pickaxe will help them collect valuable resources like iron.

2. Priority 2: Move North (Requirement: Flat ground above of the agent)
   - The player should move north to approach the coal and iron, which are valuable resources for crafting more advanced tools and weapons.

3. Priority 3: Move East (Requirement: Flat ground right to the agent)
   - After moving north, the player should move east to get closer to the coal and iron deposits, allowing them to collect the resources when they have the stone pickaxe equipped.

4. Priority 4: Do (Requirement: Facing creature or material; have necessary tool)
   - After reaching the coal and iron deposits, the player should use their Stone Pickaxe to collect these resources. This will allow them to craft more advanced items in the future, such as an Iron Pickaxe or an Iron Sword.

5. Priority 5: Drink (Requirement: No requirements)
   - The player has a drink level of 6/9, which is not critical yet, but they should start looking for a water source to replenish their thirst in the near future.

Question: For each action in the list, are the requirements met?
Answer:
1. Priority 1: Make Stone Pickaxe (Requirement: Nearby table; wood, stone in inventory)
   - Yes, the player is already facing the table, and they have wood and stone in their inventory, so this requirement is met.

2. Priority 2: Move North (Requirement: Flat ground above of the agent)
   - Yes, there is no mention of obstacles or irregular terrain to the north in the player's observation, so this requirement is met.

3. Priority 3: Move East (Requirement: Flat ground right to the agent)
   - Yes, moving east from the current position would be possible as there is no obstacle mentioned in the observation to the east. So, this requirement is met.

4. Priority 4: Do (Requirement: Facing creature or material; have necessary tool)
   - This requirement is not yet met, since the player first needs to move closer to the coal and iron resources and have the stone pickaxe equipped.

5. Priority 5: Drink (Requirement: No requirements)
   - Yes, the player can perform this action as there are no requirements to meet. However, currently, a water source is not mentioned in the observation, so the player needs to find one to perform this action effectively.

Question: Did the last player action succeed? Why?
Answer:
Yes, the last player action succeeded. The player was able to move east because there was flat ground and no obstacles in that direction, as indicated in the observation after taking the action. The player is now facing a table to the east.

Question: Choose the best executable action from above.
Answer:
The best executable action from the list is Priority 1: Make Stone Pickaxe (Requirement: Nearby table; wood, stone in inventory). The player meets the requirements since they are facing the table and have the required wood and stone in their inventory. Crafting a Stone Pickaxe will help them collect valuable resources like coal and iron in the future.



============Step: 65, Cumulative Reward: 1.0============
You took action make_stone_pickaxe.

You see:
- grass 1 steps to your west
- stone 2 steps to your east
- path 3 steps to your north-east
- tree 3 steps to your south-east
- coal 5 steps to your north-east
- iron 4 steps to your north-east

You face table at your front (east).

Your status:
- health: 9/9
- food: 7/9
- drink: 6/9
- energy: 7/9

Your inventory:
- stone: 1
- coal: 1
- wood_pickaxe: 1
- stone_pickaxe: 1
------------ QA ------------
Question: Action
Answer:
Move North

Question: List objects in the current observation. For each object, breifly answer what resource it provides and its requirement.
Answer:
1. Grass: Provides no specific resources. Requirements: none.
2. Stone: Provides stone resources. Requirements: wood_pickaxe or stone_pickaxe to collect.
3. Path: No resources provided; it's a navigational terrain. Requirements: none.
4. Tree: Provides wood resources. Requirements: wood_pickaxe or stone_pickaxe to collect.
5. Coal: Provides coal resources. Requirements: wood_pickaxe or stone_pickaxe to collect.
6. Iron: Provides iron resources. Requirements: stone_pickaxe to collect.
7. Table: Used for crafting tools and weapons. Requirements: wood to make and place.
8. Wood_pickaxe: Collect wood and stone; used to make a stone_pickaxe. Requirements: wood and a table to make.
9. Stone_pickaxe: Collect stone and iron; used to make an iron_pickaxe. Requirements: wood, stone, and a table to make.

Question: What was the last action taken by the player?
Answer:
The last action taken by the player was "make_stone_pickaxe."

Question: For each object in the list, are the requirements met for interaction?
Answer:
1. Grass: Yes, the requirements are met for interaction as there are no specific requirements.
2. Stone: Yes, the requirements are met for interaction as you have both a wood_pickaxe and a stone_pickaxe in your inventory.
3. Path: Yes, the requirements are met for interaction as there are no specific requirements.
4. Tree: Yes, the requirements are met for interaction as you have both a wood_pickaxe and a stone_pickaxe in your inventory.
5. Coal: Yes, the requirements are met for interaction as you have both a wood_pickaxe and a stone_pickaxe in your inventory.
6. Iron: Yes, the requirements are met for interaction since you have a stone_pickaxe in your inventory.
7. Table: Yes, the requirements are met for interaction as you have wood in your inventory and the table is already placed.
8. Wood_pickaxe: The requirements are already met, as the wood_pickaxe is in your inventory.
9. Stone_pickaxe: The requirements are already met, as the stone_pickaxe is in your inventory.

Question: List top 3 sub-tasks the player should follow. Indicate their priority out of 5.
Answer:
1. Collect iron (Priority: 5/5): This is a high priority task since you have a stone_pickaxe in your inventory, and collecting iron will help you progress further in the game by crafting iron tools and weapons.

2. Collect more coal (Priority: 4/5): As you only have one coal in your inventory, collecting more will ensure you have enough resources to smelt iron later on. This is essential for making iron tools and weapons.

3. Collect wood and stone (Priority: 3/5): This is a medium priority task as having more wood and stone in the inventory will allow you to craft additional tools, weapons, and build shelters as needed. It's also required for further advancements in the game.

Question: What are the requirements for the top sub-task? What should the player do first?
Answer:
For the top sub-task, which is collecting iron, the requirements are:

1. Locate the iron ore: In the player's observation (step 130), iron is 4 steps to the north-east.

2. Use the stone_pickaxe: The player has a stone_pickaxe in their inventory, which is needed to mine the iron ore efficiently.

To start this sub-task, the player should first move towards the iron ore. As it is 4 steps to the north-east in the current step, the player should move north-east to get closer to the iron ore. They may have to do a combination of moving east and moving north or move diagonally if the game allows it. Once they reach the iron ore, they can use the stone_pickaxe to collect it.

Question: List top 5 actions the player should take and the requirement for each action. Choose ONLY from the list of all actions. Indicate their priority out of 5.
Answer:
1. Priority 1: Move North (Action 3) - Requirement: Flat ground above the agent. This action moves the player closer to the iron ore which is 4 steps to the north-east.

2. Priority 2: Move East (Action 2) - Requirement: Flat ground right to the agent. This action continues moving the player toward the iron ore, completing the diagonal move with a combination of moving north and moving east.

3. Priority 3: Do (Action 5) - Requirement: Facing creature or material; have the necessary tool. This action allows the player to mine the iron ore after they have reached it using the stone_pickaxe they already have in their inventory.

4. Priority 4: Move South (Action 4) - Requirement: Flat ground below the agent. This action moves the player back towards the table after collecting the iron ore so they can craft additional items if needed or progress further in the game.

5. Priority 5: Move West (Action 1) - Requirement: Flat ground left to the agent. This action continues moving the player back towards the table or allows them to explore other areas after gathering the iron ore.

Question: For each action in the list, are the requirements met?
Answer:
1. Priority 1: Move North (Action 3) - Yes, the requirements are met as flat ground is mentioned to be 3 steps to the north-east (path).

2. Priority 2: Move East (Action 2) - Yes, the requirements are met as flat ground is mentioned to have stone 2 steps to the east.

3. Priority 3: Do (Action 5) - Not yet, the player needs to reach the iron ore first by moving north and east as per priority 1 and 2 actions. Once the player is facing the iron ore, they have the necessary tool (stone_pickaxe) in their inventory to mine it.

4. Priority 4: Move South (Action 4) - The requirement would be met after the player has reached the iron ore and collected it. Currently, there is a tree mentioned to be 3 steps to the south-east.

5. Priority 5: Move West (Action 1) - Yes, the requirements are met as flat ground is mentioned to have grass 1 step to the west.

Question: Did the last player action succeed? Why?
Answer:
Yes, the last player action succeeded because the player met the requirements for making a stone pickaxe. They were facing a table (which was to their east) and had the necessary resources in their inventory, including wood and stone.

Question: Choose the best executable action from above.
Answer:
The best executable action from the list above is Priority 1: Move North (Action 3) since its requirement of flat ground above the agent is met, and it moves the player closer to the iron ore.



============Step: 66, Cumulative Reward: 0.0============
You took action move_north.

You see:
- stone 3 steps to your south-east
- path 2 steps to your north-east
- tree 4 steps to your north-west
- coal 4 steps to your north-east
- iron 3 steps to your north-east
- table 2 steps to your south-east

You face grass at your front (north).

Your status:
- health: 9/9
- food: 7/9
- drink: 6/9
- energy: 7/9

Your inventory:
- stone: 1
- coal: 1
- wood_pickaxe: 1
- stone_pickaxe: 1
------------ QA ------------
Question: Action
Answer:
Move North

Question: List objects in the current observation. For each object, breifly answer what resource it provides and its requirement.
Answer:
1. Grass: Provides no resources, requires no specific tools or actions to interact with.
2. Stone: Provides stone resource, requires a wood_pickaxe or better to collect.
3. Path: Provides no resources, serves as a navigational aid and can be used without any requirement.
4. Tree: Provides wood resource, requires no specific tool to interact with but using an axe would be more efficient.
5. Coal: Provides coal resource, requires a wood_pickaxe or better to collect.
6. Iron: Provides iron resource, requires a stone_pickaxe or better to collect.
7. Table: Provides a crafting surface, requires wood to make and place.

Question: What was the last action taken by the player?
Answer:
The last action taken by the player was move_north.

Question: For each object in the list, are the requirements met for interaction?
Answer:
1. Grass: Yes, the requirements are met for interaction as grass requires no specific tools or actions to interact with.
2. Stone: Yes, the requirements are met for interaction as you possess a wood_pickaxe and a stone_pickaxe, both of which can collect stone.
3. Path: Yes, the requirements are met for interaction as the path can be used for navigation without any requirement.
4. Tree: Yes, the requirements are met for interaction as no specific tool is needed to interact with a tree, though using an axe would be more efficient. Currently, you don't have an axe, but you can still interact with it.
5. Coal: Yes, the requirements are met for interaction as you possess a wood_pickaxe and a stone_pickaxe, both of which can collect coal.
6. Iron: Yes, the requirements are met for interaction as you possess a stone_pickaxe, which can collect iron.
7. Table: Yes, the requirements are met for interaction as you already made and placed a table. You can use it for crafting without any additional requirements.

Question: List top 3 sub-tasks the player should follow. Indicate their priority out of 5.
Answer:
1. Collect iron (Priority 5/5): This is the highest priority task since you have a stone_pickaxe capable of collecting iron, and collecting iron allows you to craft essential tools and weapons like iron pickaxes and iron swords.

2. Collect coal (Priority 4/5): This task is important because coal is needed to smelt iron in a furnace. You already have one coal in your inventory, but collecting more coal ensures you have enough fuel for smelting tasks.

3. Increase food and drink (Priority 3/5): Your food and drink levels are not dangerously low yet, but it's essential to stay aware. Look for opportunities to collect drink from a nearby lake or interact with passive mobs such as cows or other sources of food to replenish your supplies.

Question: What are the requirements for the top sub-task? What should the player do first?
Answer:
Top sub-task: Collect iron
Requirements:
1. Stone pickaxe (already in inventory)
2. Locate iron deposits (iron is 3 steps to your north-east)

What should the player do first:
1. Move towards the iron deposit by taking action move_north-east. Once the player is near the iron deposit, they should face it directly.
2. Use the stone_pickaxe to collect the iron by taking action do.

Question: List top 5 actions the player should take and the requirement for each action. Choose ONLY from the list of all actions. Indicate their priority out of 5.
Answer:
1. Move North-East (Priority 1)
   Requirement: Flat ground diagonally up-right.

2. Do (Priority 2)
   Requirement: Facing the iron deposit; stone_pickaxe in inventory.

3. Move South-East (Priority 3)
   Requirement: Flat ground diagonally down-right (to return to the table).

4. Make Iron Pickaxe (Priority 4)
   Requirement: Nearby table, furnace; wood, coal, and collected iron in inventory.

5. Move toward Coal Deposit (Priority 5)
   Requirement: Identify the best direction to reach the coal deposit visible 4 steps to your north-east (possible actions: move_north, move_east or any combination).

Question: For each action in the list, are the requirements met?
Answer:
1. Move North-East (Priority 1)
   Requirement: Flat ground diagonally up-right.
   - The requirement for this action is met, as you can move north-east.

2. Do (Priority 2)
   Requirement: Facing the iron deposit; stone_pickaxe in inventory.
   - The requirement for this action is not yet met, as you are not currently facing the iron deposit.

3. Move South-East (Priority 3)
   Requirement: Flat ground diagonally down-right (to return to the table).
   - The requirement for this action is met, as the table is 2 steps to your south-east, and presumably there is flat ground that you previously moved across.

4. Make Iron Pickaxe (Priority 4)
   Requirement: Nearby table, furnace; wood, coal, and collected iron in inventory.
   - The requirement for this action is not met since you haven't collected iron yet and possibly haven't placed a furnace.

5. Move toward Coal Deposit (Priority 5)
   Requirement: Identify the best direction to reach the coal deposit visible 4 steps to your north-east (possible actions: move_north, move_east or any combination).
   - The requirement for this action cannot be determined without knowing if flat ground is available between current position and coal deposit. You need to choose the best path based on the terrain (move_north or move_east).

Question: Did the last player action succeed? Why?
Answer:
Yes, the last player action succeeded. The move_north action was successful because there was flat ground (grass) in the direction of the movement (north), allowing the player to move to that location.

Question: Choose the best executable action from above.
Answer:
The best executable action from above is to move North-East (Priority 1), as it has flat ground diagonally up-right, and the requirement for this action is met.



============Step: 67, Cumulative Reward: 0.0============
You took action move_north.

You see:
- stone 4 steps to your south-east
- path 1 steps to your east
- tree 3 steps to your north-west
- coal 3 steps to your north-east
- iron 2 steps to your north-east
- table 3 steps to your south-east
- cow 6 steps to your north-west

You face grass at your front (north).

Your status:
- health: 9/9
- food: 7/9
- drink: 6/9
- energy: 7/9

Your inventory:
- stone: 1
- coal: 1
- wood_pickaxe: 1
- stone_pickaxe: 1
------------ QA ------------
Question: Action
Answer:
Move East

Question: List objects in the current observation. For each object, breifly answer what resource it provides and its requirement.
Answer:
1. Stone: Provides resources for crafting stone tools and a furnace; requires a wood_pickaxe to collect.
2. Path: Provides easier movement; no requirements.
3. Tree: Provides wood for crafting tools, weapons, and structures; requires a wood_pickaxe or stone_pickaxe to collect.
4. Coal: Provides fuel for smelting iron in a furnace; requires a wood_pickaxe or stone_pickaxe to collect.
5. Iron: Provides resources for crafting iron tools and weapons; requires a stone_pickaxe to collect.
6. Table: Provides a crafting station for making tools and weapons; requires wood to craft and place.
7. Cow: Provides food for restoring health and hunger; requires a weapon or tool to defeat and collect meat.
8. Grass: Blocks the path; no resources provided, can be dug or moved around; no requirements.

Question: What was the last action taken by the player?
Answer:
The last action taken by the player was move_north.

Question: For each object in the list, are the requirements met for interaction?
Answer:
1. Stone: Yes, the requirements are met, as you have a wood_pickaxe and a stone_pickaxe in your inventory.
2. Path: Yes, the requirements are met. No specific tool or resource is needed to utilize the path.
3. Tree: Yes, the requirements are met, as you have a wood_pickaxe and a stone_pickaxe in your inventory.
4. Coal: Yes, the requirements are met, as you have a wood_pickaxe and a stone_pickaxe in your inventory.
5. Iron: Yes, the requirements are met, as you have a stone_pickaxe in your inventory.
6. Table: Yes, the requirements are met. It's available for crafting since you placed it earlier.
7. Cow: Yes, the requirements are met. You can use your wood_pickaxe, stone_pickaxe, or craft a suitable weapon to defeat and collect meat.
8. Grass: Yes, the requirements are met. You can dig through or move around it without any specific tool or resource.

Question: List top 3 sub-tasks the player should follow. Indicate their priority out of 5.
Answer:
1. Collect Iron (Priority: 5/5): Collecting iron is essential as it allows the creation of iron tools and weapons, leading to better performance and advancement in the game.

2. Improve Food and Drink (Priority: 4/5): Increase your food and drink levels by consuming food (e.g., defeating the cow and eating its meat) and finding a water source to drink. This will sustain you in the game longer and keep your health, hunger, and thirst levels maintained.

3. Craft a Wood or Stone Sword (Priority: 3/5): Crafting a wood or stone sword can increase your combat abilities and improve your defense against monsters. This will increase your chances of survival and help you to accomplish other tasks more efficiently.

Question: What are the requirements for the top sub-task? What should the player do first?
Answer:
Top Sub-task: Collect Iron

Requirements:
1. Locate the iron ore: You need to find iron ore blocks in the game environment.
2. Use Stone Pickaxe: You must have a stone_pickaxe in your inventory. (You already have it)

What should the player do first?
The player should move 2 steps to the north-east to reach the iron block and then use the stone_pickaxe to collect the iron ore.

Question: List top 5 actions the player should take and the requirement for each action. Choose ONLY from the list of all actions. Indicate their priority out of 5.
Answer:
1. Priority 1: Move East
Requirement: Flat ground right to the agent. (path 1 step to your east)

2. Priority 2: Move North
Requirement: Flat ground above of the agent. (iron 2 steps to your north-east)

3. Priority 3: Do
Requirement: Facing the iron ore; have stone_pickaxe in the inventory.

4. Priority 4: Locate a source of water to collect drink
Requirement: Find a lake or river in the player's surroundings.

5. Priority 5: Move towards the water source
Requirement: Clear path towards the discovered water source for the player to walk to.

Question: For each action in the list, are the requirements met?
Answer:
1. Priority 1: Move East
Requirement: Flat ground right to the agent. (path 1 step to your east)
- Yes, the requirement is met as there's a path 1 step to the east.

2. Priority 2: Move North
Requirement: Flat ground above of the agent. (iron 2 steps to your north-east)
- Yes, the requirement is met as there are 2 more steps to reach the iron ore.

3. Priority 3: Do
Requirement: Facing the iron ore; have stone_pickaxe in the inventory.
- The requirement for having a stone_pickaxe in the inventory is met. However, the player needs to face the iron ore before the "Do" action can be taken to mine the ore.

4. Priority 4: Locate a source of water to collect drink
Requirement: Find a lake or river in the player's surroundings.
- Currently, there's no information on a water source in the player's surroundings, so this requirement is not met.

5. Priority 5: Move towards the water source
Requirement: Clear path towards the discovered water source for the player to walk to.
- As there's no water source discovered yet, this requirement is not met.

Question: Did the last player action succeed? Why?
Answer:
Yes, the last player action succeeded because the player was able to move north without any obstacles or issues, as indicated by the change in the player's observations and surroundings after the action was taken.

Question: Choose the best executable action from above.
Answer:
The best executable action from the list above is Priority 1: Move East, as it has a flat ground right to the agent, and the requirement is met with the path 1 step to the east.



============Step: 68, Cumulative Reward: 0.0============
You took action move_east.

You see:
- grass 1 steps to your west
- stone 3 steps to your south-east
- tree 4 steps to your north-west
- coal 2 steps to your north-east
- iron 1 steps to your north
- table 2 steps to your south
- cow 7 steps to your north-west

You face path at your front (east).

Your status:
- health: 9/9
- food: 7/9
- drink: 6/9
- energy: 7/9

Your inventory:
- stone: 1
- coal: 1
- wood_pickaxe: 1
- stone_pickaxe: 1
------------ QA ------------
Question: Action
Answer:
Move North

Question: List objects in the current observation. For each object, breifly answer what resource it provides and its requirement.
Answer:
1. Grass: Provides no direct resources, but can be traversed easily. Requirement: None.
2. Stone: Provides stone resource, used for crafting various tools and building materials. Requirement: Wood pickaxe or better to collect.
3. Tree: Provides wood resource, used for crafting tools, building materials, and creating a table. Requirement: No tools needed, but having an axe may speed up the process.
4. Coal: Provides coal resource, used for smelting metal ores in a furnace. Requirement: Wood pickaxe or better to collect.
5. Iron: Provides iron resource, smelt it in a furnace to make iron ingots, used for crafting tools and weapons. Requirement: Stone pickaxe or better to collect.
6. Table: Allows you to craft various tools and weapons with the resources gathered. Requirement: Wood to craft the table.
7. Path: Provides an easy movement between locations; not a resource but more of an environmental feature. Requirement: None.
8. Cow: Provides food when killed. Requirement: No tools needed, but having a weapon like a wooden or stone sword is more efficient.
9. Wood Pickaxe (in inventory): Used for gathering stone and coal. Requirement: Wood crafted at a table.
10. Stone Pickaxe (in inventory): Used for gathering iron and better materials. Requirement: Stone and wood crafted at a table.

Question: What was the last action taken by the player?
Answer:
The last action taken by the player was move_east.

Question: For each object in the list, are the requirements met for interaction?
Answer:
1. Grass: No requirements to interact with, so yes, the requirements are met.
2. Stone: Requires a wood pickaxe or better to collect. You have a wood_pickaxe and a stone_pickaxe in your inventory, so yes, the requirements are met.
3. Tree: No tools needed, but having an axe may speed up the process. You don't have an axe, but you can still interact with the tree, so yes, the requirements are met.
4. Coal: Requires a wood pickaxe or better to collect. You have a wood_pickaxe and a stone_pickaxe in your inventory, so yes, the requirements are met.
5. Iron: Requires a stone pickaxe or better to collect. You have a stone_pickaxe in your inventory, so yes, the requirements are met.
6. Table: The table is already present in the environment, so there are no requirements to interact with it. Yes, the requirements are met.
7. Path: No requirements to interact with, so yes, the requirements are met.
8. Cow: No tools needed, but having a weapon is more efficient. You don't have a weapon, but you can still interact with the cow, so yes, the requirements are met.
9. Wood Pickaxe (in inventory): Already in your inventory and can be used to gather stone and coal. Yes, the requirements are met.
10. Stone Pickaxe (in inventory): Already in your inventory and can be used to gather iron and better materials. Yes, the requirements are met.

Question: List top 3 sub-tasks the player should follow. Indicate their priority out of 5.
Answer:
1. Priority 5/5: Collect iron. Since you have a stone_pickaxe, you can collect the iron 1 step to the north to craft more advanced tools and weapons, such as an iron pickaxe or an iron sword.

2. Priority 4/5: Collect coal. With coal in your inventory, you will be able to smelt the iron ore in a furnace. You already have one coal in your inventory, but collecting more will come in handy for future smelting tasks. The coal is 2 steps to your north-east.

3. Priority 3/5: Chase and kill the cow 7 steps to your north-west to obtain food. Your food level is currently at 7/9, so it is not an immediate concern. However, collecting food from the cow will be beneficial to keep your food supply stable. You can interact with the cow without any tools, but crafting a weapon such as a wooden or stone sword may make the task more efficient.

Question: What are the requirements for the top sub-task? What should the player do first?
Answer:
The top sub-task is to collect iron. The requirements for this task are:

1. Having a stone pickaxe, which you already possess in your inventory.
2. Locating iron ore, which is 1 step to your north.

Since you have met both requirements, the first thing the player should do is move north. Then, the player can use the stone_pickaxe to collect the iron ore.

Question: List top 5 actions the player should take and the requirement for each action. Choose ONLY from the list of all actions. Indicate their priority out of 5.
Answer:
Priority 1:
Action: Move North
Requirements: Flat ground above the agent.
Reason: To move closer to the iron ore.

Priority 2:
Action: Do
Requirements: Facing iron ore (after moving North) and have necessary tool (stone_pickaxe).
Reason: To collect iron ore using the stone_pickaxe.

Priority 3:
Action: Move South
Requirements: Flat ground below the agent.
Reason: To move closer to the table (2 steps to the south) for crafting the iron pickaxe.

Priority 4:
Action: Make Iron Pickaxe
Requirements: Nearby table (after moving South), furnace, wood, coal, collected iron.
Reason: To craft an iron pickaxe to collect diamond.

Priority 5:
Action: Move North
Requirements: Flat ground above the agent.
Reason: To move back to the area with coal and search for diamond.

Question: For each action in the list, are the requirements met?
Answer:
Priority 1:
Action: Move North
Requirements: Flat ground above the agent.
Status: Requirements met, as you face grass at your front (north).

Priority 2:
Action: Do
Requirements: Facing iron ore (after moving North) and having a necessary tool (stone_pickaxe).
Status: Requirements met once you move north. You have a stone_pickaxe in your inventory.

Priority 3:
Action: Move South
Requirements: Flat ground below the agent.
Status: Requirements will be met after completing Priority 2, as you can move back.

Priority 4:
Action: Make Iron Pickaxe
Requirements: Nearby table (after moving South), furnace, wood, coal, collected iron.
Status: Partially met. You are missing a furnace in your inventory or placed nearby. You will need to place a furnace or craft one before fully meeting the requirements for making an iron pickaxe.

Priority 5:
Action: Move North
Requirements: Flat ground above the agent.
Status: Requirements will be met after completing Priority 3 and returning south.

Question: Did the last player action succeed? Why?
Answer:
Yes, the last player action (move_east) succeeded. The player was able to move east because there was a path- a flat ground surface- in the east direction, allowing the player to move in that direction.

Question: Choose the best executable action from above.
Answer:
The best executable action from the list above is Priority 1:

Action: Move North
Requirements: Flat ground above the agent.

Since the requirements are met (flat ground with grass above the agent), the player should move north to get closer to the iron ore.



============Step: 69, Cumulative Reward: 0.0============
You took action move_north.

You see:
- grass 1 steps to your west
- stone 3 steps to your south-east
- path 1 steps to your east
- tree 4 steps to your north-west
- coal 2 steps to your north-east
- table 2 steps to your south
- cow 7 steps to your north-west

You face iron at your front (north).

Your status:
- health: 9/9
- food: 7/9
- drink: 6/9
- energy: 7/9

Your inventory:
- stone: 1
- coal: 1
- wood_pickaxe: 1
- stone_pickaxe: 1
------------ QA ------------
Question: Action
Answer:
Do

Question: List objects in the current observation. For each object, breifly answer what resource it provides and its requirement.
Answer:
1. Grass: Provides no specific resources, but can be used to navigate through the environment. No requirements.
2. Stone: Provides stone resource for crafting tools and building structures. Requires a wood pickaxe or better to mine.
3. Path: No resources provided, used for navigation. No requirements.
4. Tree: Provides wood resource for crafting tools and building structures. Requires a tool or hands to collect.
5. Coal: Provides coal resource for smelting metals in a furnace. Requires a wood pickaxe or better to mine.
6. Table: Provides a crafting station for creating tools and weapons. No requirements.
7. Iron: Provides iron resource for crafting advanced tools and weapons. Requires a stone pickaxe or better to mine.
8. Cow: Provides food for restoring hunger when killed. No requirements.

Question: What was the last action taken by the player?
Answer:
The last action taken by the player was move_north.

Question: For each object in the list, are the requirements met for interaction?
Answer:
1. Grass: No requirements for interaction. Requirements met.
2. Stone: Requires a wood pickaxe or better to mine. You have a wood_pickaxe and a stone_pickaxe in your inventory. Requirements met.
3. Path: No requirements for interaction. Requirements met.
4. Tree: Requires a tool or hands to collect. You have a wood_pickaxe and a stone_pickaxe in your inventory. Requirements met.
5. Coal: Requires a wood pickaxe or better to mine. You have a wood_pickaxe and a stone_pickaxe in your inventory. Requirements met.
6. Table: No requirements for interaction. Requirements met.
7. Iron: Requires a stone pickaxe or better to mine. You have a stone_pickaxe in your inventory. Requirements met.
8. Cow: No requirements for interaction. Requirements met.

Question: List top 3 sub-tasks the player should follow. Indicate their priority out of 5.
Answer:
1. Collect Iron: Interact with the iron in front of you to gather the resource for crafting advanced tools and weapons. Priority: 5/5.
2. Collect Coal: Mine the coal 2 steps to your north-east since it is needed for smelting metals in a furnace. Priority: 4/5.
3. Restore Drink: Find a water source or lake and drink from it to restore your thirst, which is currently at 6/9. Priority: 3/5.

Question: What are the requirements for the top sub-task? What should the player do first?
Answer:
For the top sub-task: Collect Iron, the requirements are as follows:
- Be in proximity to the iron ore.
- Have a stone pickaxe or higher-tier pickaxe in the inventory.

The player already fulfills both requirements, as the iron is one step to the north and the player has a stone_pickaxe in their inventory.

Thus, the player should take the following action first:
- Interact with the iron in front (north) using the stone_pickaxe to collect the iron ore.

Question: List top 5 actions the player should take and the requirement for each action. Choose ONLY from the list of all actions. Indicate their priority out of 5.
Answer:
1. Priority 1: Do (Interact with Iron)
   Requirement: Facing iron ore; have a stone pickaxe or higher in the inventory.
   Reason: To collect more iron for crafting better tools and weapons.

2. Priority 2: Move West
   Requirement: Flat ground to the west of the agent.
   Reason: To return to the table where the player can craft new items.

3. Priority 3: Make Iron Pickaxe
   Requirement: Nearby table; iron, coal, and a furnace in the inventory.
   Reason: To create a stronger pickaxe that can collect diamond.

4. Priority 4: Make Iron Sword
   Requirement: Nearby table, furnace; iron and coal in the inventory.
   Reason: To create a stronger sword for defending against stronger monsters.

5. Priority 5: Do (Attack Cow)
   Requirement: Facing a cow; have a sword in the inventory.
   Reason: To obtain food to restore the player's food meter.

Question: For each action in the list, are the requirements met?
Answer:
1. Priority 1: Do (Interact with Iron)
   Requirement: Facing iron ore; have a stone pickaxe or higher in the inventory.
   Are requirements met? Yes, the player is facing iron and has a stone_pickaxe.

2. Priority 2: Move West
   Requirement: Flat ground to the west of the agent.
   Are requirements met? Yes, there is grass one step to the west of the player.

3. Priority 3: Make Iron Pickaxe
   Requirement: Nearby table; iron, coal, and a furnace in the inventory.
   Are requirements met? Not yet, the player needs to collect iron and then go to the table and furnace to attempt crafting it.

4. Priority 4: Make Iron Sword
   Requirement: Nearby table, furnace; iron and coal in the inventory.
   Are requirements met? Not yet, the player needs to collect iron and then go to the table and furnace to attempt crafting it.

5. Priority 5: Do (Attack Cow)
   Requirement: Facing a cow; have a sword in the inventory.
   Are requirements met? No, the player is not currently facing a cow and does not possess a sword in the inventory.

Question: Did the last player action succeed? Why?
Answer:
Yes, the last player action succeeded. The player moved north because there was a flat ground (path) in that direction, allowing the player to take the move_north action.

Question: Choose the best executable action from above.
Answer:
The best executable action from the list above is Priority 1: Do (Interact with Iron). The player is currently facing iron ore and has a stone_pickaxe in their inventory, meeting the requirements for this action.



============Step: 70, Cumulative Reward: 1.0============
You took action do.

You see:
- grass 1 steps to your west
- stone 3 steps to your south-east
- tree 4 steps to your north-west
- coal 2 steps to your north-east
- table 2 steps to your south
- cow 6 steps to your north-west

You face path at your front (north).

Your status:
- health: 9/9
- food: 7/9
- drink: 6/9
- energy: 7/9

Your inventory:
- stone: 1
- coal: 1
- iron: 1
- wood_pickaxe: 1
- stone_pickaxe: 1
------------ QA ------------
Question: Action
Answer:
Place Furnace

Question: List objects in the current observation. For each object, breifly answer what resource it provides and its requirement.
Answer:
1. Grass: 
Resource it provides - None (mainly a surface in the game)
Requirement - No requirements

2. Stone: 
Resource it provides - Stone (used for crafting tools and building structures)
Requirement - Need a wood_pickaxe or better tool to collect

3. Tree: 
Resource it provides - Wood (used for crafting tools and building structures)
Requirement - No specific requirements, just approach and interact

4. Coal: 
Resource it provides - Coal (used for smelting iron)
Requirement - Need a wood_pickaxe or better tool to collect

5. Table: 
Resource it provides - Tool and weapon crafting station
Requirement - Need to have collected wood and crafted the table

6. Cow: 
Resource it provides - Food (cow meat to restore health)
Requirement - Approach and interact to obtain food

7. Path:
Resource it provides - Easier navigation
Requirement - No requirements

8. Iron: 
Resource it provides - Iron (used for crafting better tools and weapons)
Requirement - Need a stone_pickaxe or better tool to collect

Question: What was the last action taken by the player?
Answer:
The last action taken by the player was "do".

Question: For each object in the list, are the requirements met for interaction?
Answer:
1. Grass: 
Requirements met - Yes, there are no requirements for interacting with grass.

2. Stone: 
Requirements met - Yes, the player has a wood_pickaxe and a stone_pickaxe in their inventory, which can be used to collect stone.

3. Tree: 
Requirements met - Yes, there are no specific requirements for interacting with trees, and the player can approach to collect wood.

4. Coal: 
Requirements met - Yes, the player has a wood_pickaxe and a stone_pickaxe in their inventory, which can be used to collect coal.

5. Table: 
Requirements met - Yes, the player already has access to the table, which can be used for crafting tools and weapons.

6. Cow: 
Requirements met - Yes, the player can approach the cow and interact to obtain food.

7. Path:
Requirements met - Yes, there are no requirements to interact with a path, and the player can use it for easier navigation.

8. Iron: 
Requirements met - Yes, the player has a stone_pickaxe in their inventory, which can be used to collect iron.

Question: List top 3 sub-tasks the player should follow. Indicate their priority out of 5.
Answer:
1. Smelt Iron and craft Iron tools (Priority: 5/5)
Since the player has collected iron, they should focus on smelting it using the furnace and crafting better tools such as an iron pickaxe and an iron sword which will help in collecting more resources and defending against monsters.

2. Restore Food and Drink levels (Priority: 4/5)
The player's food (7/9) and drink (6/9) levels are not at their maximum. They should focus on finding a source of water to drink and approach the cow or other food sources to restore their food level.

3. Collect more resources (Priority: 3/5)
The player should continue to collect more resources such as wood, stone, and coal to ensure they can craft more tools, weapons, and structures for survival as the game progresses.

Question: What are the requirements for the top sub-task? What should the player do first?
Answer:
Requirements for the top sub-task (Smelt Iron and craft Iron tools):
- Place a Furnace: The player needs to place a furnace, which requires stone in their inventory (already available).
- Smelt Iron: The player needs coal (already available) to smelt the collected iron in the furnace.
- Craft Iron tools: The player needs to use the nearby table, smelted iron, and available wood to craft an iron pickaxe and an iron sword.

The player should first place a furnace (action: place_furnace) near the table. Then, use the furnace to smelt the iron with the coal they have collected (action: smelt_iron). Finally, once the iron is smelted, they should craft iron tools, i.e., an iron pickaxe (action: make_iron_pickaxe) and an iron sword (action: make_iron_sword), using the nearby table.

Question: List top 5 actions the player should take and the requirement for each action. Choose ONLY from the list of all actions. Indicate their priority out of 5.
Answer:
1. Priority 1: Place Furnace (action: place_furnace)
   Requirement: Stone in inventory (already available)

2. Priority 2: Smelt Iron (action: smelt_iron)
   Requirement: Furnace placed, iron and coal in inventory (all available)

3. Priority 3: Make Iron Pickaxe (action: make_iron_pickaxe)
   Requirements: Nearby table, smelted iron and wood in inventory (assuming smelted iron will be available after priority 2 is completed)

4. Priority 4: Make Iron Sword (action: make_iron_sword)
   Requirements: Nearby table, smelted iron and wood in inventory (same as above)

5. Priority 5: Drink (action: drink)
   Requirement: Access to a drinkable water source (like a lake), as the player's drink level is 6/9, getting closer to dehydration.

Question: For each action in the list, are the requirements met?
Answer:
1. Priority 1: Place Furnace (action: place_furnace)
   Requirement: Stone in inventory - Yes (1 stone available)

2. Priority 2: Smelt Iron (action: smelt_iron)
   Requirement: Furnace placed - Not yet, needs to complete priority 1 action first
                Iron and coal in inventory - Yes (1 iron and 1 coal available)

3. Priority 3: Make Iron Pickaxe (action: make_iron_pickaxe)
   Requirements: Nearby table - Yes, table is 2 steps to the south
                 Smelted iron and wood in inventory - Not yet, needs to complete priority 2 action first (smelting iron)

4. Priority 4: Make Iron Sword (action: make_iron_sword)
   Requirements: Nearby table - Yes, table is 2 steps to the south
                 Smelted iron and wood in inventory - Not yet, needs to complete priority 2 action first (smelting iron)

5. Priority 5: Drink (action: drink)
   Requirement: Access to a drinkable water source - Not mentioned in the recent player observation steps, so cannot be confirmed at this time. The player needs to find a water source for this requirement to be met.

Question: Did the last player action succeed? Why?
Answer:
Yes, the last player action succeeded. The player was able to collect iron from the object they were facing (iron at their front/north). The indication of success is that their inventory now contains 1 iron.

Question: Choose the best executable action from above.
Answer:
The best executable action from the list above is Priority 1: Place Furnace (action: place_furnace), as the requirement of having stone in the inventory is already met.



============Step: 71, Cumulative Reward: 0.0============
You took action place_furnace.

You see:
- grass 1 steps to your west
- stone 3 steps to your south-east
- tree 4 steps to your north-west
- coal 2 steps to your north-east
- table 2 steps to your south
- cow 6 steps to your north-west

You face path at your front (north).

Your status:
- health: 9/9
- food: 7/9
- drink: 6/9
- energy: 7/9

Your inventory:
- stone: 1
- coal: 1
- iron: 1
- wood_pickaxe: 1
- stone_pickaxe: 1
------------ QA ------------
Question: Action
Answer:
Move West

Question: List objects in the current observation. For each object, breifly answer what resource it provides and its requirement.
Answer:
1. Grass: Provides no resources, no requirement.
2. Stone: Provides stone resource, requires a wood_pickaxe or higher to collect.
3. Tree: Provides wood resource, requires a wood_pickaxe or higher to collect.
4. Coal: Provides coal resource, requires a wood_pickaxe or higher to collect.
5. Table: Provides crafting capability for tools and weapons, requires wood to construct.
6. Cow: Provides food resource, requires any type of sword or tool for attacking to collect.
7. Path: Provides easier navigation, no requirement.

Question: What was the last action taken by the player?
Answer:
The last action taken by the player was placing a furnace (place_furnace).

Question: For each object in the list, are the requirements met for interaction?
Answer:
1. Grass: No requirements to interact with grass, so the requirements are met.
2. Stone: Requires a wood_pickaxe or higher to collect. You have a wood_pickaxe and a stone_pickaxe in your inventory, so the requirements are met.
3. Tree: Requires a wood_pickaxe or higher to collect. You have a wood_pickaxe and a stone_pickaxe in your inventory, so the requirements are met.
4. Coal: Requires a wood_pickaxe or higher to collect. You have a wood_pickaxe and a stone_pickaxe in your inventory, so the requirements are met.
5. Table: No requirements to interact with the table as it is already constructed, so the requirements are met.
6. Cow: Requires any type of sword or tool for attacking to collect. You have a wood_pickaxe and a stone_pickaxe in your inventory, which can be used to attack, so the requirements are met.
7. Path: No requirements to interact with the path, so the requirements are met.

Question: List top 3 sub-tasks the player should follow. Indicate their priority out of 5.
Answer:
1. Priority 4/5: Collect more coal: Coal is useful for smelting iron and cooking food. You have only 1 coal in your inventory. Use your wood_pickaxe or stone_pickaxe to collect the coal 2 steps to your north-east.

2. Priority 5/5: Increase food supplies: Your food level is 7/9, which could be improved. Attack the cow 6 steps to your north-west using your wood_pickaxe or stone_pickaxe to obtain food resources.

3. Priority 3/5: Collect more iron: Iron is required for crafting better tools and weapons. You have only 1 iron in your inventory. Find a source of iron and use your stone_pickaxe to collect more.

Question: What are the requirements for the top sub-task? What should the player do first?
Answer:
For the top sub-task (Collect more coal):

Requirements:
1. Locate nearby coal: You already see coal 2 steps to your north-east.
2. Ensure the wood_pickaxe or stone_pickaxe is in your inventory: You have both wood_pickaxe and stone_pickaxe in your inventory.

What should the player do first:
1. Move towards the coal: Move 2 steps to the north-east to reach the coal.
2. Use either the wood_pickaxe or the stone_pickaxe to collect the coal: Perform the 'do' action facing the coal with the pickaxe equipped.

Question: List top 5 actions the player should take and the requirement for each action. Choose ONLY from the list of all actions. Indicate their priority out of 5.
Answer:
1. Priority 1: Move North-East
   Requirement: Flat ground 2 steps to the north-east (towards the coal).

2. Priority 2: Do (Collect the coal)
   Requirement: Facing the coal and having either the wood_pickaxe or the stone_pickaxe in your inventory.

3. Priority 3: Move West
   Requirement: Flat ground 1 step to the west (towards the grass).

4. Priority 4: Drink (from the lake)
   Requirement: Be near a lake to restore the drink level.

5. Priority 5: Go towards the cow and chase it
   Requirement: Flat ground towards the 6 steps to your north-west (move towards cow).

Question: For each action in the list, are the requirements met?
Answer:
1. Priority 1: Move North-East
   Requirement: Flat ground 2 steps to the north-east (towards the coal).
   Status: Unknown. The flat ground towards the coal is not specified in the observation. Further exploration or movement might reveal it.

2. Priority 2: Do (Collect the coal)
   Requirement: Facing the coal and having either the wood_pickaxe or the stone_pickaxe in your inventory.
   Status: Not met. You are facing north while the coal is to your north-east, and you are not close enough to it.

3. Priority 3: Move West
   Requirement: Flat ground 1 step to the west (towards the grass).
   Status: Met. There is flat ground 1 step to the west.

4. Priority 4: Drink (from the lake)
   Requirement: Be near a lake to restore the drink level.
   Status: Not met. A lake is not mentioned in the recent observations. You need to find a lake to fulfill this requirement.

5. Priority 5: Go towards the cow and chase it
   Requirement: Flat ground towards the 6 steps to your north-west (move towards cow).
   Status: Unknown. The observation does not provide enough information about the terrain in the direction of the cow. You would have to explore to determine that.

Question: Did the last player action succeed? Why?
Answer:
The player action cannot be determined as successful or unsuccessful from the given observations as there is no clear indication of whether the furnace was actually placed. However, we can infer that the player attempted to place the furnace, and no failure message or signs of failure were mentioned.

Question: Choose the best executable action from above.
Answer:
The best executable action from the above list is Priority 3: Move West, as its requirement of having flat ground 1 step to the west is met based on the provided observations.



============Step: 72, Cumulative Reward: 0.0============
You took action move_west.

You see:
- stone 4 steps to your south-east
- path 1 steps to your east
- tree 3 steps to your north-west
- coal 3 steps to your north-east
- table 3 steps to your south-east
- cow 5 steps to your north-west

You face grass at your front (west).

Your status:
- health: 9/9
- food: 7/9
- drink: 6/9
- energy: 7/9

Your inventory:
- stone: 1
- coal: 1
- iron: 1
- wood_pickaxe: 1
- stone_pickaxe: 1
------------ QA ------------
Question: Action
Answer:
Move North

Question: List objects in the current observation. For each object, breifly answer what resource it provides and its requirement.
Answer:
1. Stone: Provides stone resource; requires a wood or higher-level pickaxe to collect.
2. Path: Provides ground for movement; no requirements.
3. Tree: Provides wood resource; requires an axe or hands to collect.
4. Coal: Provides coal resource for smelting and cooking; requires a wood or higher-level pickaxe to collect.
5. Table: Allows crafting of higher-level tools and weapons; requires wood to craft.
6. Grass: Provides space for movement; no requirements.
7. Cow: Provides food (meat) when killed; requires a weapon or hand to kill.
8. Wood Pickaxe (inventory): Allows collection of stone and coal resources; requires wood to craft.
9. Stone Pickaxe (inventory): Allows collection of iron resources; requires stone to craft.

Question: What was the last action taken by the player?
Answer:
The last action taken by the player was move_west.

Question: For each object in the list, are the requirements met for interaction?
Answer:
1. Stone: Yes, the requirements are met as you have both a wood pickaxe and a stone_pickaxe in your inventory.
2. Path: Yes, the requirements are met as you can simply move on the path.
3. Tree: Yes, the requirements are met as you can use your hands to collect wood from the tree.
4. Coal: Yes, the requirements are met as you have both a wood pickaxe and a stone_pickaxe in your inventory.
5. Table: Yes, the requirements are met as crafting has already been done, and you can interact with the table for further crafting.
6. Grass: Yes, the requirements are met as you can simply move on the grass.
7. Cow: Yes, the requirements are met as you can use your hands or craft a weapon to kill the cow.
8. Wood Pickaxe (inventory): N/A, as it's already in your inventory, no further interaction is needed.
9. Stone Pickaxe (inventory): N/A, as it's already in your inventory, no further interaction is needed.

Question: List top 3 sub-tasks the player should follow. Indicate their priority out of 5.
Answer:
1. Priority 4/5: Increase food supply by either killing the cow or finding more food sources. You currently have 7/9 food, and maintaining your food supply is vital for survival.

2. Priority 3/5: Drink water to replenish thirst. You currently have 6/9 drink, and staying well-hydrated is important for maintaining optimal health and energy levels.

3. Priority 2/5: Find or craft an axe to collect wood from the tree more efficiently. While you can use your hands to collect wood, an axe will make the process faster and easier.

Question: What are the requirements for the top sub-task? What should the player do first?
Answer:
Requirements for the top sub-task (Increase food supply by killing the cow):
1. Get closer to the cow: It is currently 5 steps to your north-west.
2. Equip a suitable weapon, like a sword, if available: Your current inventory includes a wood_pickaxe and a stone_pickaxe.

The player should first craft a weapon if they don't have one (e.g., a wood or stone sword, which requires collecting wood or stone and using a crafting table). If the player already has a weapon, they should move towards the cow (north-west) and attack it to kill it and collect food (meat).

Question: List top 5 actions the player should take and the requirement for each action. Choose ONLY from the list of all actions. Indicate their priority out of 5.
Answer:
1. Priority 1: Move North-West (Requirement: Flat ground in the north-west direction) - Move towards the cow to get closer and be able to kill it for food.
2. Priority 2: Make Wood Sword (Requirement: Nearby table; wood in inventory) - Craft a wood sword to attack and kill the cow effectively.
3. Priority 3: Move North (Requirement: Flat ground to the north) - Move north to approach the cow if moving north-west is not possible due to obstacles.
4. Priority 4: Move West (Requirement: Flat ground to the west) - Move west to approach the cow if moving north-west is not possible due to obstacles.
5. Priority 5: Do (Facing creature or material; have necessary tool) - Attack and kill the cow once the player is close enough and has a suitable weapon equipped.

Question: For each action in the list, are the requirements met?
Answer:
1. Priority 1: Move North-West (Requirement: Flat ground in the north-west direction) - Yes, the requirement is met as there is a flat ground towards the north-west direction where the cow is located.
2. Priority 2: Make Wood Sword (Requirement: Nearby table; wood in inventory) - No, the requirement is not met as the player does not have wood in their inventory.
3. Priority 3: Move North (Requirement: Flat ground to the north) - Yes, the requirement is met as there is a flat ground to the north of the player.
4. Priority 4: Move West (Requirement: Flat ground to the west) - Yes, the requirement is met as there is a flat ground to the west of the player.
5. Priority 5: Do (Facing creature or material; have necessary tool) - No, the requirement is not met as the player is not facing a creature or material and does not have a suitable weapon for attacking the cow.

Question: Did the last player action succeed? Why?
Answer:
Yes, the last player action succeeded. The move_west action was successful because there was flat ground (grass) to the west of the player's position, allowing the player to move in that direction.

Question: Choose the best executable action from above.
Answer:
The best executable action from above is Priority 1: Move North-West, as it meets the requirement of having flat ground in the north-west direction, and it brings the player closer to the cow, which is the current goal.



============Step: 73, Cumulative Reward: 0.0============
You took action move_north.

You see:
- stone 3 steps to your east
- path 1 steps to your east
- tree 2 steps to your north-west
- coal 2 steps to your east
- table 4 steps to your south-east
- cow 4 steps to your north-west

You face grass at your front (north).

Your status:
- health: 9/9
- food: 7/9
- drink: 6/9
- energy: 7/9

Your inventory:
- stone: 1
- coal: 1
- iron: 1
- wood_pickaxe: 1
- stone_pickaxe: 1
------------ QA ------------
Question: Action
Answer:
Move East

Question: List objects in the current observation. For each object, breifly answer what resource it provides and its requirement.
Answer:
1. Stone: Provides stone resource and requires a wood pickaxe or better to collect it.
2. Path: No resources provided, but allows easy movement and navigation without any requirements.
3. Tree: Provides wood resource and requires no tools or simply any tool to collect it (wooden, stone, or iron).
4. Coal: Provides coal resource, which is used as fuel for smelting, and requires a wood pickaxe or better to collect it.
5. Table: No resources provided, but it enables crafting of items and tools. No requirements to use it.
6. Cow: Provides food (meat) when defeated and requires no specific tool; any weapon or tool is suitable for defeating it.

Question: What was the last action taken by the player?
Answer:
The last action taken by the player was "move_north."

Question: For each object in the list, are the requirements met for interaction?
Answer:
1. Stone: Yes, the requirements are met as you have both a wood_pickaxe and a stone_pickaxe in your inventory.
2. Path: Yes, the requirements are met as you only need flat ground to move, and the path offers that.
3. Tree: Yes, the requirements are met since you can collect wood with any tool or even without a tool. You have a wood_pickaxe and a stone_pickaxe, both suitable for collecting wood.
4. Coal: Yes, the requirements are met as you have both a wood_pickaxe and a stone_pickaxe in your inventory, which are suitable for collecting coal.
5. Table: Yes, the requirements are met as you do not need any specific tool to interact with the table.
6. Cow: Yes, the requirements are met as you can defeat a cow with any weapon or tool, and you have a wood_pickaxe and a stone_pickaxe, both suitable for defeating the cow.

Question: List top 3 sub-tasks the player should follow. Indicate their priority out of 5.
Answer:
1. Priority 4/5: Collect more resources such as wood, stone, coal, and iron to improve tools and weapons, create shelter, and craft necessary items (e.g., craft an iron sword and pickaxe).

2. Priority 5/5: Defeat the cow to obtain food (meat) and replenish your food level, which is currently at 7/9.

3. Priority 3/5: Find a water source to drink from and refill your drink level, which is currently at 6/9.

Question: What are the requirements for the top sub-task? What should the player do first?
Answer:
Top sub-task: Collect more resources such as wood, stone, coal, and iron to improve tools and weapons, create shelter, and craft necessary items (e.g., craft an iron sword and pickaxe).

Requirements:
1. Use the wood_pickaxe to collect more stone.
2. Use the stone_pickaxe to collect more coal and iron.

What the player should do first:
1. Move towards the stone and coal you see in your surroundings (stone 3 steps to your east, and coal 2 steps to your east).
2. Use the respective pickaxes (wood_pickaxe for stone and stone_pickaxe for coal) to collect these resources.

Question: List top 5 actions the player should take and the requirement for each action. Choose ONLY from the list of all actions. Indicate their priority out of 5.
Answer:
1. Priority 1: Move East (Requirement: Flat ground right to the agent)
   - Move closer to the stone and coal resources.

2. Priority 2: Do (Facing stone; wood_pickaxe in inventory)
   - Collect the stone using the wood_pickaxe.

3. Priority 3: Move East (Requirement: Flat ground right to the agent)
   - Move further closer to the coal resource.

4. Priority 4: Do (Facing coal; stone_pickaxe in inventory)
   - Collect the coal using the stone_pickaxe.

5. Priority 5: Move towards the table (Multiple Move actions required, Requirement: Flat ground in the moving direction)
   - Find and approach the table to craft better tools and weapons after successfully collecting resources.

Question: For each action in the list, are the requirements met?
Answer:
1. Priority 1: Move East (Requirement: Flat ground right to the agent)
   - Yes, the requirement is met. There is a path 1 step to the player's east.

2. Priority 2: Do (Facing stone; wood_pickaxe in inventory)
   - No, the requirement is not met. The player needs to move closer to the stone and face it first.

3. Priority 3: Move East (Requirement: Flat ground right to the agent)
   - Yes, the requirement is met. There is a path 1 step to the player's east.

4. Priority 4: Do (Facing coal; stone_pickaxe in inventory)
   - No, the requirement is not met. The player needs to move closer to the coal and face it first.

5. Priority 5: Move towards the table (Multiple Move actions required, Requirement: Flat ground in the moving direction)
   - Yes, the requirements are met. The player can move closer to the table by moving towards it through available paths.

Question: Did the last player action succeed? Why?
Answer:
Yes, the last player action "move_north" succeeded because the player was facing grass in the north direction, and there was flat ground available for the player to move to.

Question: Choose the best executable action from above.
Answer:
The best executable action from the list is Priority 1: Move East (Requirement: Flat ground right to the agent), as the requirement is met, and it brings the player closer to valuable resources (stone and coal).



============Step: 74, Cumulative Reward: 0.0============
You took action move_east.

You see:
- grass 1 steps to your west
- stone 2 steps to your east
- path 1 steps to your north
- tree 3 steps to your north-west
- table 3 steps to your south
- cow 4 steps to your north-west

You face coal at your front (east).

Your status:
- health: 9/9
- food: 7/9
- drink: 6/9
- energy: 7/9

Your inventory:
- stone: 1
- coal: 1
- iron: 1
- wood_pickaxe: 1
- stone_pickaxe: 1
------------ QA ------------
Question: Action
Answer:
Do

Question: List objects in the current observation. For each object, breifly answer what resource it provides and its requirement.
Answer:
1. Grass: Provides no resources, requires no tools to interact.
2. Stone: Provides stone, requires wood_pickaxe or better to collect.
3. Path: Provides no resources, helps navigate without using energy.
4. Tree: Provides wood, requires a wood_pickaxe or better to collect.
5. Table: Provides a surface to craft tools and weapons, requires wood to place.
6. Cow: Provides food, requires attacking or killing to collect meat.
7. Coal: Provides coal to be used in furnace for smelting, requires a wood_pickaxe or better to collect.
8. Iron: Provides iron for crafting advanced tools and weapons, requires a stone_pickaxe or better to collect.

Question: What was the last action taken by the player?
Answer:
The last action taken by the player was move_east.

Question: For each object in the list, are the requirements met for interaction?
Answer:
1. Grass: Yes, the requirements are met for interaction since there are no tools required.
2. Stone: Yes, the requirements are met for interaction as you have a wood_pickaxe and a stone_pickaxe in your inventory.
3. Path: Yes, the requirements are met for interaction, as paths can be navigated freely without using energy.
4. Tree: Yes, the requirements are met for interaction as you have a wood_pickaxe and a stone_pickaxe in your inventory.
5. Table: Yes, the requirements are met for interaction as it has already been placed, and you can use it for crafting.
6. Cow: Yes, the requirements are met for interaction, you can attack or kill the cow to collect meat (using a weapon would be more effective but it is not required). 
7. Coal: Yes, the requirements are met for interaction as you have a wood_pickaxe and a stone_pickaxe in your inventory.
8. Iron: Yes, the requirements are met for interaction as you have a stone_pickaxe in your inventory which is required for collecting iron.

Question: List top 3 sub-tasks the player should follow. Indicate their priority out of 5.
Answer:
1. Collect more resources (Priority: 5/5): Collect more coal, iron, and wood to craft better tools, weapons, and build a shelter. This is essential for survival and progressing further in the game.

2. Craft an Iron Sword (Priority: 4/5): With the iron and coal resources in your inventory, use the table and furnace to craft an iron sword. This will help you defend against monsters more effectively.

3. Find a water source and drink (Priority: 4/5): Your drink level is 6/9, so it is important to find a water source and replenish your thirst to stay healthy and maintain stamina.

Question: What are the requirements for the top sub-task? What should the player do first?
Answer:
The top sub-task is to collect more resources.

Requirements for collecting resources:

1. Coal: Use the stone_pickaxe to mine coal from coal blocks.
2. Iron: Use the stone_pickaxe to mine iron from iron blocks.
3. Wood: Chop down trees with any pickaxe or by hand to gather wood.

The player should first mine the coal block in front of them (to the east) using the stone_pickaxe. After collecting the coal, they can proceed to gather iron and chop down trees nearby for wood.

Question: List top 5 actions the player should take and the requirement for each action. Choose ONLY from the list of all actions. Indicate their priority out of 5.
Answer:
1. Priority 1: Do - Mine Coal
Requirement: Facing a coal block and having a stone_pickaxe in the inventory.

2. Priority 2: Move East - Approach Stone
Requirement: Flat ground to the east of the agent.

3. Priority 3: Do - Mine Stone
Requirement: Facing a stone block and having a stone_pickaxe in the inventory.

4. Priority 4: Move North-West - Approach Tree
Requirement: Flat ground in the north-west direction from the agent.

5. Priority 5: Do - Chop Tree (Collect Wood)
Requirement: Facing a tree and having any pickaxe or empty hand.

Question: For each action in the list, are the requirements met?
Answer:
1. Priority 1: Do - Mine Coal
Yes, the requirements are met. The player is facing a coal block (east) and has a stone_pickaxe in their inventory.

2. Priority 2: Move East - Approach Stone
Yes, the requirements are met. There is flat ground to the east of the player, and the stone is 2 steps away.

3. Priority 3: Do - Mine Stone
Not yet, the player needs to move east first to face the stone block. Once they do, they can use the stone_pickaxe in their inventory to mine the stone.

4. Priority 4: Move North-West - Approach Tree
Yes, the requirements are met. There is flat ground in the north-west direction, leading the player towards the tree.

5. Priority 5: Do - Chop Tree (Collect Wood)
Not yet, the player needs to move north-west first to face the tree. Once they do, they can use their pickaxe or empty hand to chop the tree and collect wood.

Question: Did the last player action succeed? Why?
Answer:
Yes, the last player action succeeded. The player was able to move east because there was flat ground (grass) in that direction, allowing the player to move without any obstacles in the way.

Question: Choose the best executable action from above.
Answer:
The best executable action from the list is Priority 1: Do - Mine Coal, since the requirements are met (player is facing the coal block and has a stone_pickaxe in their inventory).



============Step: 75, Cumulative Reward: 0.0============
You took action do.

You see:
- grass 1 steps to your west
- stone 2 steps to your east
- tree 3 steps to your north-west
- table 3 steps to your south
- cow 4 steps to your north-west

You face path at your front (east).

Your status:
- health: 9/9
- food: 7/9
- drink: 6/9
- energy: 7/9

Your inventory:
- stone: 1
- coal: 2
- iron: 1
- wood_pickaxe: 1
- stone_pickaxe: 1
------------ QA ------------
Question: Action
Answer:
Move South

Question: List objects in the current observation. For each object, breifly answer what resource it provides and its requirement.
Answer:
1. Grass: Provides no resources but is a flat ground for movement; no requirements.
2. Stone: Provides stone resource when mined with a pickaxe; requires a pickaxe.
3. Tree: Provides wood resource when chopped with an axe; requires an axe.
4. Table: Allows crafting of tools and weapons; requires wood in the inventory.
5. Cow: Provides food (restores hunger) when killed; no specific requirements, but a weapon is helpful.
6. Path: Flat ground for movement; no requirements.
7. Coal: Provides fuel for smelting and crafting; requires a pickaxe.
8. Iron: Provides iron resource when mined with a stone or iron pickaxe; requires a stone or iron pickaxe.

Question: What was the last action taken by the player?
Answer:
The last action taken by the player was "do."

Question: For each object in the list, are the requirements met for interaction?
Answer:
1. Grass: Yes, the requirements are met for interaction as there are no specific requirements to interact with grass.
2. Stone: Yes, the requirements are met since you have a stone_pickaxe in your inventory.
3. Tree: No, the requirements are not met as you need an axe, which you currently don't have in your inventory.
4. Table: Yes, the requirements are met for interacting with the table since you have wood and stone in your inventory to craft tools or weapons if needed.
5. Cow: Yes, the requirements are met for interacting with the cow, although having a weapon could make it easier.
6. Path: Yes, the requirements are met to interact with the path as there are no specific requirements for interaction.
7. Coal: Yes, the requirements are met since you have a stone_pickaxe in your inventory.
8. Iron: Yes, the requirements are met as you have a stone_pickaxe in your inventory.

Question: List top 3 sub-tasks the player should follow. Indicate their priority out of 5.
Answer:
1. Priority 4/5: Craft an iron pickaxe and iron sword using the iron, coal, and table. This will allow you to mine diamonds and defend yourself more effectively against monsters.

2. Priority 3/5: Find a source of water and drink to restore your thirst. Keep an eye out for a lake or river while exploring.

3. Priority 2/5: Locate and chop down a tree to collect wood so you can expand your shelter, craft more tools, and build bridges to cross water sources. You'll need an axe to chop wood efficiently.

Question: What are the requirements for the top sub-task? What should the player do first?
Answer:
Requirements for the top sub-task (Craft an iron pickaxe and iron sword):
1. Iron (already in inventory)
2. Coal (already in inventory)
3. Nearby table (3 steps to the south)

What should the player do first:
1. Move 3 steps to the south to get close to the table.
2. Use the table to craft an iron pickaxe and an iron sword using the iron and coal in the inventory.

Question: List top 5 actions the player should take and the requirement for each action. Choose ONLY from the list of all actions. Indicate their priority out of 5.
Answer:
1. Priority 1: Move South (Requirement: Flat ground below the agent) - Approach the table.
2. Priority 2: Move South (Requirement: Flat ground below the agent) - Continue moving towards the table.
3. Priority 3: Move South (Requirement: Flat ground below the agent) - Reach the table.
4. Priority 4: Make Iron Pickaxe (Requirement: Nearby table, furnace; wood, coal, iron in inventory) - Craft the iron pickaxe.
5. Priority 5: Make Iron Sword (Requirement: Nearby table, furnace; wood, coal, iron in inventory) - Craft the iron sword.

Question: For each action in the list, are the requirements met?
Answer:
1. Priority 1: Move South - Yes, the requirement (Flat ground below the agent) is met as there is no obstacle mentioned below the agent.
2. Priority 2: Move South - Yes, the requirement (Flat ground below the agent) is met as the agent will have moved south in the previous step and there is no obstacle mentioned between the agent and the table.
3. Priority 3: Move South - Yes, the requirement (Flat ground below the agent) is met as the agent will have moved south in the previous steps and the table is mentioned to be 3 steps south from the agent's current position.
4. Priority 4: Make Iron Pickaxe - It's unclear if the requirement (Nearby table, furnace; wood, coal, iron in inventory) is fully met, as the agent has a nearby table and the necessary materials (coal, iron) in the inventory, but the presence of a furnace is not mentioned in the observations.
5. Priority 5: Make Iron Sword - Similar to the previous action, it's unclear if the requirement (Nearby table, furnace; wood, coal, iron in inventory) is fully met, as the agent has a nearby table and the necessary materials (coal, iron) in the inventory, but the presence of a furnace is not mentioned in the observations.

Question: Did the last player action succeed? Why?
Answer:
Yes, the last player action succeeded. The player was able to use the "do" action to collect coal because they were facing the coal and had the necessary tools in their inventory (wood_pickaxe or stone_pickaxe) to mine it. As a result, their inventory now shows 2 coal instead of 1.

Question: Choose the best executable action from above.
Answer:
The best executable action from above is Priority 1: Move South, as its requirement (Flat ground below the agent) is met and there is no obstacle mentioned below the agent. This action will bring the agent closer to the table.



============Step: 76, Cumulative Reward: 0.0============
You took action move_south.

You see:
- stone 3 steps to your south-east
- path 1 steps to your north
- tree 4 steps to your north-west
- table 2 steps to your south
- cow 5 steps to your north-west

You face grass at your front (south).

Your status:
- health: 9/9
- food: 7/9
- drink: 6/9
- energy: 7/9

Your inventory:
- stone: 1
- coal: 2
- iron: 1
- wood_pickaxe: 1
- stone_pickaxe: 1
------------ QA ------------
Question: Action
Answer:
Move West

Question: List objects in the current observation. For each object, breifly answer what resource it provides and its requirement.
Answer:
1. Stone: Provides stone resource; requires a wood or stone pickaxe to collect.
2. Path: No resources provided; no requirement.
3. Tree: Provides wood resource; requires a wood or stone pickaxe to collect.
4. Table: Crafting station; requires wood in the inventory to create.
5. Cow: Provides food when defeated; no specific requirement to interact with.
6. Grass: No resources provided; no requirement.

Question: What was the last action taken by the player?
Answer:
The last action taken by the player was move_south.

Question: For each object in the list, are the requirements met for interaction?
Answer:
1. Stone: Yes, the requirements are met for interaction as you have both a wood and stone pickaxe in your inventory.
2. Path: Yes, the requirements are met for interaction as there are no specific requirements needed.
3. Tree: Yes, the requirements are met for interaction as you have both a wood and stone pickaxe in your inventory.
4. Table: Yes, the requirements are met for interaction as you have wood in your inventory and the table is close enough for crafting.
5. Cow: Yes, the requirements are met for interaction as there are no specific requirements needed to interact with a cow.
6. Grass: Yes, the requirements are met for interaction as there are no specific requirements needed.

Question: List top 3 sub-tasks the player should follow. Indicate their priority out of 5.
Answer:
1. Collect more resources (Priority: 5/5): This includes collecting more wood, stone, and iron to continue crafting useful tools and weapons, as well as building shelters.

2. Improve food and drink levels (Priority: 4/5): Approach the cow and defeat it to obtain food, and look for a water source to drink from to increase your food and drink levels.

3. Craft better tools and weapons (Priority: 3/5): Use the nearby table to craft more advanced tools and weapons, such as an iron pickaxe, to enhance your ability to collect resources and defend yourself against monsters.

Question: What are the requirements for the top sub-task? What should the player do first?
Answer:
Requirements for the top sub-task (Collect more resources):

1. Move to resource locations: Approach the tree for wood, stone or iron for mining.
2. Use appropriate tools: Wood_pickaxe for mining stone, Stone_pickaxe for mining iron.
3. Free inventory space: Ensure that there is enough space in the inventory to collect new resources.

The player should first move towards the tree to collect more wood. To do this, move three steps to the north-west where the tree is located.

Question: List top 5 actions the player should take and the requirement for each action. Choose ONLY from the list of all actions. Indicate their priority out of 5.
Answer:
1. Priority 1 - Move West (Requirement: Flat ground to the west): Move one step to the west to get closer to the tree.
2. Priority 2 - Move North (Requirement: Flat ground to the north): Move one step to the north, to position yourself closer to the tree.
3. Priority 3 - Move North (Requirement: Flat ground to the north): Move one more step to the north, reaching the tree and getting ready to collect wood.
4. Priority 4 - Do (Requirement: Facing creature or material; have necessary tool): Collect wood from the tree using your wood_pickaxe.
5. Priority 5 - Move West (Requirement: Flat ground to the west): Move one step to the west towards the stone, preparing for mining.

Question: For each action in the list, are the requirements met?
Answer:
1. Priority 1 - Move West (Requirement: Flat ground to the west): Yes, the requirement is met as there is grass one step to the west.
2. Priority 2 - Move North (Requirement: Flat ground to the north): Yes, there is a path one step to the north, fulfilling the requirement. 
3. Priority 3 - Move North (Requirement: Flat ground to the north): This cannot be confirmed yet, as the next step will reveal whether or not there is flat ground to the north.
4. Priority 4 - Do (Requirement: Facing creature or material; have necessary tool): This cannot be confirmed yet, as we need to verify if the player will be directly facing the tree after moving north. The player has the necessary tool - wood_pickaxe.
5. Priority 5 - Move West (Requirement: Flat ground to the west): This cannot be confirmed yet, as more steps need to be taken before determining whether there is flat ground to the west.

Question: Did the last player action succeed? Why?
Answer:
Yes, the last player action (move_south) succeeded because there was flat ground (grass) in the direction of the action (south), allowing the player to move there.

Question: Choose the best executable action from above.
Answer:
The best executable action from the list above is Priority 1 - Move West, since it has met its requirement with flat ground (grass) to the west.



============Step: 77, Cumulative Reward: 0.0============
You took action move_west.

You see:
- stone 4 steps to your south-east
- path 1 steps to your east
- tree 3 steps to your north-west
- table 3 steps to your south-east
- cow 4 steps to your north-west

You face grass at your front (west).

Your status:
- health: 9/9
- food: 6/9
- drink: 6/9
- energy: 7/9

Your inventory:
- stone: 1
- coal: 2
- iron: 1
- wood_pickaxe: 1
- stone_pickaxe: 1
------------ QA ------------
Question: Action
Answer:
Move South

Question: List objects in the current observation. For each object, breifly answer what resource it provides and its requirement.
Answer:
1. Stone: Provides stone resource; requires stone or iron pickaxe to collect.
2. Path: Provides easy movement; no requirement to interact.
3. Tree: Provides wood resource; requires wood or stone axe to collect.
4. Table: Provides crafting platform; no requirement to interact.
5. Cow: Provides food resource when defeated; requires a weapon (wood, stone, or iron sword) to defeat.
6. Grass: Provides no resource; no requirement to interact.

Question: What was the last action taken by the player?
Answer:
The last action taken by the player was move_west.

Question: For each object in the list, are the requirements met for interaction?
Answer:
1. Stone: Yes, requirements met; you have a stone_pickaxe in your inventory.
2. Path: Yes, requirements met; no interaction needed, and you can continue moving.
3. Tree: No, requirements not met; you need a wood or stone axe to collect wood, but you don't have one in your inventory.
4. Table: Yes, requirements met; you can interact with the table for crafting purposes.
5. Cow: No, requirements not met; you need a weapon (wood, stone, or iron sword) to defeat the cow, but you don't have one in your inventory.
6. Grass: Yes, requirements met; no interaction needed, and you can continue moving.

Question: List top 3 sub-tasks the player should follow. Indicate their priority out of 5.
Answer:
1. Craft a weapon (wood, stone, or iron sword) to defeat the cow and obtain food. Priority: 5/5.
2. Collect wood from the tree by crafting a wood or stone axe. Priority: 4/5.
3. Move closer to the table and use it for crafting. Priority: 3/5.

Question: What are the requirements for the top sub-task? What should the player do first?
Answer:
Top sub-task: Craft a weapon (wood, stone, or iron sword) to defeat the cow and obtain food. 

Requirements: 
1. Wood sword: Nearby table and wood in inventory.
2. Stone sword: Nearby table, wood, and stone in inventory.
3. Iron sword: Nearby table, furnace, wood, coal, and iron in inventory.

Since the player has wood, stone, coal, and iron in their inventory, they can craft an iron sword. 

First, the player should move closer to the table (move south-east) and use it for crafting the iron sword (make_iron_sword action).

Question: List top 5 actions the player should take and the requirement for each action. Choose ONLY from the list of all actions. Indicate their priority out of 5.
Answer:
1. Priority 1: Move South-East (move_south_east) - Requirement: Flat ground south-east to the agent.
2. Priority 2: Move South-East (move_south_east) - Requirement: Flat ground south-east to the agent after completing the first action.
3. Priority 3: Make Iron Sword (make_iron_sword) - Requirement: Nearby table, furnace; wood, coal, and iron in inventory.
4. Priority 4: Move North-West (move_north_west) - Requirement: Flat ground north-west to the agent after crafting the iron sword.
5. Priority 5: Do (do) using the iron sword to attack the cow - Requirement: Facing the cow and having the iron sword in the inventory.

Question: For each action in the list, are the requirements met?
Answer:
1. Priority 1: Move South-East (move_south_east) - Yes, there is flat ground in the south-east direction of the agent.
2. Priority 2: Move South-East (move_south_east) - Cannot determine at this moment, as this depends on the results of the previous action. If there is still flat ground south-east after completing the first action, then the requirement will be met.
3. Priority 3: Make Iron Sword (make_iron_sword) - Yes, the agent has a table nearby to the south-east, the agent has the required materials (wood, coal, and iron) in the inventory, and presumably there will be a furnace nearby after completing the first two actions of moving south-east.
4. Priority 4: Move North-West (move_north_west) - Cannot determine at this moment, this depends on the results of the previous actions. If there is a flat ground in the north-west direction of the agent after crafting the iron sword, then the requirement will be met.
5. Priority 5: Do (do) using the iron sword to attack the cow - Cannot determine at this moment, as this depends on the results of the previous actions. If the agent is facing the cow after crafting the iron sword and moving north-west, the requirement will be met, as the agent will have the iron sword in their inventory.

Question: Did the last player action succeed? Why?
Answer:
Yes, the last player action, move_west, succeeded because there was a flat ground (grass) to the west of the player's position, allowing them to move in that direction.

Question: Choose the best executable action from above.
Answer:
The best executable action from the list above is Priority 1: Move South-East (move_south_east) because its requirements are currently met: there is flat ground in the south-east direction of the agent.



============Step: 78, Cumulative Reward: 0.0============
You took action move_south.

You see:
- stone 3 steps to your south-east
- path 2 steps to your north-east
- tree 4 steps to your north-west
- table 2 steps to your south-east
- cow 5 steps to your north-west

You face grass at your front (south).

Your status:
- health: 9/9
- food: 6/9
- drink: 6/9
- energy: 7/9

Your inventory:
- stone: 1
- coal: 2
- iron: 1
- wood_pickaxe: 1
- stone_pickaxe: 1
------------ QA ------------
Question: Action
Answer:
Move East

Question: List objects in the current observation. For each object, breifly answer what resource it provides and its requirement.
Answer:
1. Stone:
   - Resource it provides: Stone blocks for crafting tools and building structures.
   - Requirement: Stone pickaxe for collecting.

2. Path:
   - Resource it provides: Easy navigation and movement throughout the game.
   - Requirement: No specific requirements.

3. Tree:
   - Resource it provides: Wood for crafting tools, weapons, and building structures.
   - Requirement: Wood or better pickaxe for collecting.

4. Table:
   - Resource it provides: Crafting surface for creating different tools and items.
   - Requirement: Wood (for crafting the table initially) and proximity to the table to craft items.

5. Cow:
   - Resource it provides: Food (meat) to restore health and hunger.
   - Requirement: A weapon (e.g., a sword) to defeat the cow and collect its meat.

6. Grass:
   - Resource it provides: Space for building structures and planting crops.
   - Requirement: No specific requirement.

Question: What was the last action taken by the player?
Answer:
The last action taken by the player was move_south.

Question: For each object in the list, are the requirements met for interaction?
Answer:
1. Stone:
   - Requirements met: Yes, you have a stone_pickaxe in your inventory, which is sufficient for collecting stone.

2. Path:
   - Requirements met: Yes, there are no specific requirements to interact with a path, and you can navigate easily on it.

3. Tree:
   - Requirements met: Yes, you have a wood_pickaxe in your inventory, which is sufficient for collecting wood from the tree.

4. Table:
   - Requirements met: Yes, you have proximity to the table (2 steps southeast). You can craft items using the table.

5. Cow:
   - Requirements met: No, you do not have a sword in your inventory to defeat the cow and collect meat.

6. Grass:
   - Requirements met: Yes, there are no specific requirements to interact with grass, and you can build structures or plant crops on it.

Question: List top 3 sub-tasks the player should follow. Indicate their priority out of 5.
Answer:
1. Craft a weapon (e.g., a wood or stone sword) to defeat the cow and collect food.
   - Priority: 5/5
   
2. Collect more wood and stone, as they are essential for crafting various tools and structures.
   - Priority: 4/5
   
3. Locate a water source to drink from and replenish your drink meter.
   - Priority: 3/5

Question: What are the requirements for the top sub-task? What should the player do first?
Answer:
Requirements for the top sub-task (Craft a weapon to defeat the cow and collect food):
1. Wood or stone to craft a weapon (wood sword or stone sword)
2. A crafting table to craft the weapon

What should the player do first:
Since the player already has sufficient wood and stone in their inventory, they should first approach the crafting table (2 steps to the south-east) and then craft a weapon (either a wood sword or a stone sword) using the materials in their inventory.

Question: List top 5 actions the player should take and the requirement for each action. Choose ONLY from the list of all actions. Indicate their priority out of 5.
Answer:
1. Move East (Priority 1): Requirement - Flat ground right to the agent.
2. Move East (Priority 2): Requirement - Flat ground right to the agent.
3. Make Stone Sword (Priority 3): Requirement - Nearby table; wood, stone in inventory.
4. Move West (Priority 4): Requirement - Flat ground left to the agent.
5. Move West (Priority 5): Requirement - Flat ground left to the agent.

Question: For each action in the list, are the requirements met?
Answer:
1. Move East (Priority 1): Requirement - Flat ground right to the agent. (YES, path is 2 steps to the north-east)
2. Move East (Priority 2): Requirement - Flat ground right to the agent. (YES, path is 2 steps to the north-east)
3. Make Stone Sword (Priority 3): Requirement - Nearby table; wood, stone in inventory. (NO, wood is not in the inventory)
4. Move West (Priority 4): Requirement - Flat ground left to the agent. (YES, the previous step indicates grass to the west)
5. Move West (Priority 5): Requirement - Flat ground left to the agent. (YES, the previous step indicates grass to the west)

Question: Did the last player action succeed? Why?
Answer:
Yes, the last player action (move_south) succeeded because there was flat ground (grass) in the direction the player moved, which met the requirement for moving in that direction.

Question: Choose the best executable action from above.
Answer:
The best executable action from above is "Move East" (Priority 1), as its requirement of flat ground right to the agent is met with the path 2 steps to the north-east.



============Step: 79, Cumulative Reward: 0.0============
You took action move_east.

You see:
- grass 1 steps to your west
- stone 2 steps to your south-east
- tree 3 steps to your south
- table 1 steps to your south

You face path at your front (east).

Your status:
- health: 9/9
- food: 6/9
- drink: 6/9
- energy: 7/9

Your inventory:
- stone: 1
- coal: 2
- iron: 1
- wood_pickaxe: 1
- stone_pickaxe: 1
------------ QA ------------
Question: Action
Answer:
Move South

Question: List objects in the current observation. For each object, breifly answer what resource it provides and its requirement.
Answer:
1. Grass:
   Resource it provides: None
   Requirement: None

2. Stone:
   Resource it provides: Stone
   Requirement: Stone Pickaxe or higher to collect

3. Path:
   Resource it provides: None
   Requirement: None

4. Tree:
   Resource it provides: Wood
   Requirement: Wood Pickaxe or higher to collect

5. Table:
   Resource it provides: Crafting station
   Requirement: None

6. Cow:
   Resource it provides: Food
   Requirement: Defeat the Cow for food

Question: What was the last action taken by the player?
Answer:
The last action taken by the player was move_east.

Question: For each object in the list, are the requirements met for interaction?
Answer:
1. Grass:
   Requirements met: Yes, you can interact with grass with no tools or requirements.

2. Stone:
   Requirements met: Yes, you have a Stone Pickaxe in your inventory, which is suitable for collecting stone.

3. Path:
   Requirements met: Yes, you can interact with a path without any specific requirements or tools.

4. Tree:
   Requirements met: Yes, you have a Wood Pickaxe in your inventory, which is suitable for collecting wood from the tree.

5. Table:
   Requirements met: Yes, you can interact with the table to craft items as long as you have the necessary resources in your inventory.

6. Cow:
   Requirements met: Partially. You can interact with the cow by defeating it, but you don't currently have a weapon (i.e., a sword) in your inventory to make it easier. However, you can still attempt to defeat the cow without a weapon.

Question: List top 3 sub-tasks the player should follow. Indicate their priority out of 5.
Answer:
1. Craft a weapon (Iron Sword or Stone Sword) to defend against monsters and to defeat the cow more easily. 
   Priority: 5/5

2. Collect more stone and iron with the Stone Pickaxe to upgrade your tools and weapons, and to craft more building materials.
   Priority: 4/5

3. Gather wood from the tree using the Wood Pickaxe to ensure you have enough resources for crafting and building.
   Priority: 3/5

Question: What are the requirements for the top sub-task? What should the player do first?
Answer:
For the top sub-task "Craft a weapon (Iron Sword or Stone Sword)":

Requirements:
- Have a nearby table
- Have necessary resources in inventory (wood and stone for Stone Sword, wood, coal, and iron for Iron Sword)

Since the player already has a table 1 step to the south, necessary resources for a Stone Sword, and a furnace is not in their inventory, which is required for an Iron Sword:

The player should do the following first: Move to the table (move_south) and craft a Stone Sword (make_stone_sword).

Question: List top 5 actions the player should take and the requirement for each action. Choose ONLY from the list of all actions. Indicate their priority out of 5.
Answer:
1. Priority 1: Move South
   Requirement: Flat ground below the agent.

2. Priority 2: Make Stone Sword
   Requirement: Nearby table; wood, stone in inventory.

3. Priority 3: Move West (towards the cow)
   Requirement: Flat ground left to the agent.

4. Priority 4: Move West (continue towards the cow until it's reachable)
   Requirement: Flat ground left to the agent.

5. Priority 5: Do (attack the cow for food)
   Requirement: Facing the cow; have a weapon (Stone Sword).

Question: For each action in the list, are the requirements met?
Answer:
1. Priority 1: Move South
   Requirement: Flat ground below the agent.
   Status: Requirement met.

2. Priority 2: Make Stone Sword
   Requirement: Nearby table; wood, stone in inventory.
   Status: Requirement met (table 1 step to the south, wood_pickaxe, and 1 stone in inventory).

3. Priority 3: Move West (towards the cow)
   Requirement: Flat ground left to the agent.
   Status: Cannot confirm (cow is not directly west; would require moving northwest).

4. Priority 4: Move West (continue towards the cow until it's reachable)
   Requirement: Flat ground left to the agent.
   Status: Cannot confirm (same issue as Priority 3).

5. Priority 5: Do (attack the cow for food)
   Requirement: Facing the cow; have a weapon (Stone Sword).
   Status: Requirements not met (player needs to create a Stone Sword and reach the cow).

Question: Did the last player action succeed? Why?
Answer:
Yes, the last player action (move_east) succeeded because there was a flat ground (path) to the east, allowing the player to move in that direction.

Question: Choose the best executable action from above.
Answer:
The best executable action from the list above is Priority 1: Move South, because its requirement of having flat ground below the agent is met.



============Step: 80, Cumulative Reward: 0.0============
You took action move_south.

You see:
- grass 1 steps to your west
- stone 2 steps to your south-east
- path 1 steps to your north
- tree 3 steps to your south

You face table at your front (south).

Your status:
- health: 9/9
- food: 6/9
- drink: 6/9
- energy: 7/9

Your inventory:
- stone: 1
- coal: 2
- iron: 1
- wood_pickaxe: 1
- stone_pickaxe: 1
------------ QA ------------
Question: Action
Answer:
Move South

Question: List objects in the current observation. For each object, breifly answer what resource it provides and its requirement.
Answer:
1. Grass: 
   - Resource it provides: None directly, but it can be replaced with other objects.
   - Requirement: None.

2. Stone: 
   - Resource it provides: Stone material for crafting tools, weapons, and building structures.
   - Requirement: Stone Pickaxe or higher-tier pickaxe to collect.

3. Path:
   - Resource it provides: None directly, but it indicates accessible movement tiles.
   - Requirement: None.

4. Tree: 
   - Resource it provides: Wood for crafting tools, weapons, and building structures.
   - Requirement: Wood Pickaxe or higher-tier pickaxe to collect.

5. Table: 
   - Resource it provides: Crafting station to create tools and weapons.
   - Requirement: Wood in inventory and an available space to place.

Your status (not objects, but useful information):
- Health, Food, Drink, Energy levels, and Inventory

Question: What was the last action taken by the player?
Answer:
The last action taken by the player was move_south.

Question: For each object in the list, are the requirements met for interaction?
Answer:
1. Grass:
   - Requirement: None.
   - Are requirements met? Yes.

2. Stone:
   - Requirement: Stone Pickaxe or higher-tier pickaxe to collect.
   - Are requirements met? Yes, you have a Stone Pickaxe in your inventory.

3. Path:
   - Requirement: None.
   - Are requirements met? Yes.

4. Tree:
   - Requirement: Wood Pickaxe or higher-tier pickaxe to collect.
   - Are requirements met? Yes, you have a Wood Pickaxe in your inventory.

5. Table:
   - Requirement: Wood in inventory and an available space to place.
   - Are requirements met? The table is already placed, and you have a Wood Pickaxe to collect more wood if needed.

Question: List top 3 sub-tasks the player should follow. Indicate their priority out of 5.
Answer:
1. Collect more wood from the Tree (Priority: 4/5)
- You have the Wood Pickaxe needed, and more wood will allow crafting additional tools and structures.

2. Collect more Stone from the Stone source (Priority: 4/5)
- With the Stone Pickaxe in your inventory, collecting more stone will allow crafting and upgrading tools and weapons.

3. Replenish Food and Drink levels (Priority: 3/5)
- You have 6/9 Food and 6/9 Drink, so consider finding a water source to drink and searching for food (such as fruits or animals) to maintain your energy levels.

Note: These priorities are out of 5, with 5 being the highest priority. Your current situation doesn't have any urgent tasks (5/5 priority), but these are important to progress in the game.

Question: What are the requirements for the top sub-task? What should the player do first?
Answer:
The top sub-task is to collect more wood from the Tree.

Requirements for collecting more wood:
1. Have a Wood Pickaxe (already in your inventory)

Since you already have a Wood Pickaxe in your inventory, the first step the player should do is:

1. Move towards the tree by moving 3 steps south. Once you are in front of the tree, use the Wood Pickaxe to collect more wood.

Question: List top 5 actions the player should take and the requirement for each action. Choose ONLY from the list of all actions. Indicate their priority out of 5.
Answer:
1. Priority 1: Move South (Action: Move South) - Requirement: Flat ground below the agent.
2. Priority 2: Move South (Action: Move South) - Requirement: Flat ground below the agent.
3. Priority 3: Move South (Action: Move South) - Requirement: Flat ground below the agent.
4. Priority 4: Collect Wood (Action: Do) - Requirement: Facing the tree and have a necessary tool (Wood Pickaxe) in the inventory.
5. Priority 5: Make a Wood Sword (Action: Make Wood Sword) - Requirement: Nearby table and wood in inventory.

Question: For each action in the list, are the requirements met?
Answer:
1. Priority 1: Move South (Action: Move South) - Requirement: Flat ground below the agent. - Requirement Met: Yes
2. Priority 2: Move South (Action: Move South) - Requirement: Flat ground below the agent. - Requirement Met: Cannot be determined, as we can't see what is below the agent after moving south in the first action.
3. Priority 3: Move South (Action: Move South) - Requirement: Flat ground below the agent. - Requirement Met: Cannot be determined, as we can't see what is below the agent after moving south in the previous actions.
4. Priority 4: Collect Wood (Action: Do) - Requirement: Facing the tree and have a necessary tool (Wood Pickaxe) in the inventory. - Requirement Met: Partially. The player has a Wood Pickaxe in the inventory, but they're not currently facing the tree - they're facing a table.
5. Priority 5: Make a Wood Sword (Action: Make Wood Sword) - Requirement: Nearby table and wood in inventory. - Requirement Met: No, There is no wood in the inventory.

Question: Did the last player action succeed? Why?
Answer:
Yes, the last player action (move_south) succeeded because there was no obstacle or hindrance in the path, and the player had enough energy to perform the action. The player moved closer to the table, which is now in front of them (south).

Question: Choose the best executable action from above.
Answer:
The best executable action from the given list is Priority 1: Move South (Action: Move South), as its requirement of flat ground below the agent is met.



============Step: 81, Cumulative Reward: 0.0============
You took action move_south.

You see:
- grass 1 steps to your west
- stone 2 steps to your south-east
- path 1 steps to your north
- tree 3 steps to your south

You face table at your front (south).

Your status:
- health: 9/9
- food: 6/9
- drink: 6/9
- energy: 7/9

Your inventory:
- stone: 1
- coal: 2
- iron: 1
- wood_pickaxe: 1
- stone_pickaxe: 1
------------ QA ------------
Question: Action
Answer:
Make Stone Sword

Question: List objects in the current observation. For each object, breifly answer what resource it provides and its requirement.
Answer:
1. Grass: It provides fibers for crafting. Requirements: Cut or pluck the grass using hands or tools.
2. Stone: It provides materials for crafting tools and building structures. Requirements: Collect using a pickaxe or break with a tool, such as a wood or stone pickaxe.
3. Path: It provides a convenient surface to walk on and direction guide. Requirements: None, it is a terrain feature.
4. Tree: It provides wood for crafting tools, weapons, and furniture. Requirements: Chop down using an axe or cut with a tool, such as a wood or stone ax.

Question: What was the last action taken by the player?
Answer:
The last action taken by the player was "move_south."

Question: For each object in the list, are the requirements met for interaction?
Answer:
1. Grass: Yes, the requirements are met, as you can interact with grass using your hands or tools you have available, such as a wood_pickaxe.
2. Stone: Yes, the requirements are met, as you have a stone_pickaxe in your inventory which can be used to collect stone.
3. Path: Yes, the requirements are met, as there are no requirements to interact with the path. You can simply walk on it.
4. Tree: No, the requirements are not met, as you do not have an axe in your inventory to chop down the tree. You would need to craft an axe for this interaction.

Question: List top 3 sub-tasks the player should follow. Indicate their priority out of 5.
Answer:
1. Priority 5: Craft an axe for cutting trees by first crafting a table and then using the available materials (wood, stone) to create an axe. This will allow you to collect more wood for crafting and building purposes.

2. Priority 4: Collect more stone using your stone_pickaxe to have enough resources for future crafting and building projects, such as making stronger weapons and shelter.

3. Priority 3: Replenish your drink and food levels by finding a water source (e.g., lake) to drink from, and search for food sources (such as fruits, vegetables or hunt animals) to ensure that you maintain sufficient energy and health.

Question: What are the requirements for the top sub-task? What should the player do first?
Answer:
For the top sub-task of crafting an axe, the requirements are:

1. Having a table placed to craft on.
2. Having sufficient wood or stone (depending on the type of axe being crafted) in the inventory.
3. Accessing the table to create the axe.

Since the player already has a table in front of them, they should perform the following steps:

1. Approach the table.
2. Use the resources in the inventory (wood and/or stone) to craft an axe.

But first, the player must approach the table by interacting with it, which can be done by standing next to it and facing it.

Question: List top 5 actions the player should take and the requirement for each action. Choose ONLY from the list of all actions. Indicate their priority out of 5.
Answer:
From the list of all actions, these are the top 5 actions the player should prioritize:

1. Priority 1: Make Stone Axe (Requirement: Nearby table; wood and stone in inventory)
2. Priority 2: Collect more Wood (Requirement: Make Stone Axe or Wood Axe)
3. Priority 3: Make Wood Sword or Stone Sword (Requirement: Nearby table; wood and/or stone in inventory)
4. Priority 4: Collect more Stone (Requirement: Make Stone Pickaxe)
5. Priority 5: Collect Iron (Requirement: Make Stone Pickaxe)

These priorities ensure that the player crafts the necessary tools, gathers resources, and enhances their defensive capabilities in the game.

Question: For each action in the list, are the requirements met?
Answer:
From the most recent player observation and the provided inventory, let's evaluate each action's requirements:

1. Make Stone Axe
Requirement: Nearby table; wood and stone in inventory
Status: Not met (player lacks wood and a stone axe is not listed in the actions)

2. Collect more Wood
Requirement: Make Stone Axe or Wood Axe
Status: Not met (player lacks wood to make an axe, and the list of actions does not contain axe crafting)

3. Make Wood Sword or Stone Sword
Requirement: Nearby table; wood and/or stone in inventory
Status: Met for Stone Sword (player has a table nearby and stone in inventory)

4. Collect more Stone
Requirement: Make Stone Pickaxe
Status: Met (player has a stone pickaxe in inventory)

5. Collect Iron
Requirement: Make Stone Pickaxe
Status: Met (player has a stone pickaxe in inventory)

Among the listed actions, actions 3, 4, and 5 have their requirements met. However, it is important to note that the original list of actions needs to be updated to include axe crafting and resource collection.

Question: Did the last player action succeed? Why?
Answer:
It is ambiguous whether the last player action succeeded or not, as the observations at step 160 and step 162 are identical. We cannot confirm if the move_south action had any effect on the player's position in the game or if they reached their desired location.

Question: Choose the best executable action from above.
Answer:
Based on the met requirements from the list of actions above, the best executable action for the player now is:

Make Stone Sword
(Requirement: Nearby table; wood and/or stone in inventory)



============Step: 82, Cumulative Reward: 0.0============
You took action make_stone_sword.

You see:
- grass 1 steps to your west
- stone 2 steps to your south-east
- path 1 steps to your north
- tree 3 steps to your south

You face table at your front (south).

Your status:
- health: 9/9
- food: 6/9
- drink: 6/9
- energy: 7/9

Your inventory:
- stone: 1
- coal: 2
- iron: 1
- wood_pickaxe: 1
- stone_pickaxe: 1
------------ QA ------------
Question: Action
Answer:
Move West

Question: List objects in the current observation. For each object, breifly answer what resource it provides and its requirement.
Answer:
1. Grass
   Resource provided: None
   Requirement: None

2. Stone
   Resource provided: Stone (needed for crafting tools, weapons, and furnace)
   Requirement: Stone pickaxe or better to collect

3. Path
   Resource provided: None
   Requirement: None

4. Tree
   Resource provided: Wood (used for crafting tools, weapons, and other structures)
   Requirement: Wood pickaxe or better to collect

5. Table
   Resource provided: Enables crafting of tools and weapons
   Requirement: Wood to build and place

As for the player's inventory:

6. Wood Pickaxe
   Resource provided: Ability to collect wood and stone
   Requirement: Wood to craft

7. Stone Pickaxe:
   Resource provided: Ability to collect iron and better materials
   Requirement: Stone to craft

8. Coal
   Resource provided: Fuel for smelting iron
   Requirement: Wood pickaxe or better to collect

9. Iron
   Resource provided: Used to craft stronger tools and weapons
   Requirement: Stone pickaxe and furnace to collect and smelt

10. Stone Sword (newly crafted)
    Resource provided: Enhanced weapon to defend against monsters
    Requirement: Stone to craft

Question: What was the last action taken by the player?
Answer:
The last action taken by the player was "make_stone_sword."

Question: For each object in the list, are the requirements met for interaction?
Answer:
1. Grass
   Interaction requirement: None
   Requirements met: Yes

2. Stone
   Interaction requirement: Stone pickaxe or better to collect
   Requirements met: Yes (player has a stone_pickaxe in their inventory)

3. Path
   Interaction requirement: None
   Requirements met: Yes

4. Tree
   Interaction requirement: Wood pickaxe or better to collect
   Requirements met: Yes (player has a wood_pickaxe in their inventory)

5. Table
   Interaction requirement: Wood to build and place
   Requirements met: Yes, the table is already placed

As for the player's inventory:

6. Wood Pickaxe
   Interaction requirement: Wood to craft
   Requirements met: Yes, the wood pickaxe is already crafted

7. Stone Pickaxe:
   Interaction requirement: Stone to craft
   Requirements met: Yes, the stone pickaxe is already crafted

8. Coal
   Interaction requirement: Wood pickaxe or better to collect
   Requirements met: Yes (player has a wood_pickaxe in their inventory)

9. Iron
   Interaction requirement: Stone pickaxe and furnace to collect and smelt
   Requirements met: Yes, player has a stone_pickaxe and furnace placement available

10. Stone Sword (newly crafted)
    Interaction requirement: Stone to craft
    Requirements met: Yes, the stone sword is already crafted

Question: List top 3 sub-tasks the player should follow. Indicate their priority out of 5.
Answer:
1. Priority 4/5: Collect more resources
   Sub-task: Gather more stone, wood, and iron to improve the player's inventory and be prepared for crafting additional tools, weapons, or structures. This will help the player advance and be better equipped to handle any threats.

2. Priority 5/5: Restore food and drink levels
   Sub-task: Find a source of food (such as hunting animals or foraging for plants) and water (such as drinking from a lake) to keep the player's food and drink levels sustainable. Maintaining proper food and drink levels is crucial for the player's survival.

3. Priority 3/5: Explore the surrounding area
   Sub-task: Investigate the environment to find valuable resources, potential shelter locations, and detect any possible threats. Exploring the area can yield new opportunities for resource gathering, maintaining a steady food supply, or discovering interesting locations.

Question: What are the requirements for the top sub-task? What should the player do first?
Answer:
Top sub-task: Collect more resources.

Requirements:
1. Utilize appropriate tools (wood_pickaxe, stone_pickaxe) to efficiently collect resources.
2. Identify the location of resources (wood from trees, stone from rocks, and iron from iron ores) in the environment.

What the player should do first:
1. Identify the nearest resource locations by observing the surroundings.
2. Move towards the resource location and use the correct tool for the specific resource (e.g., use the wood_pickaxe for collecting wood and the stone_pickaxe for collecting stone and iron).
3. Continuously keep an eye on the environment for new resource locations or potential threats while gathering.

Question: List top 5 actions the player should take and the requirement for each action. Choose ONLY from the list of all actions. Indicate their priority out of 5.
Answer:
1. Priority 1: Move towards resource location (Action: Move West, East, North, or South)
   Requirement: Flat ground in the direction of the resource.

2. Priority 2: Collect Wood (Action: Do)
   Requirement: Facing a tree and have the wood_pickaxe in inventory.

3. Priority 3: Collect Stone (Action: Do)
   Requirement: Facing a stone block and have the stone_pickaxe in inventory.

4. Priority 4: Collect Iron (Action: Do)
   Requirement: Facing an iron ore block and have the stone_pickaxe in inventory.

5. Priority 5: Sleep (Action: Sleep)
   Requirement: Energy level is below maximum.

Question: For each action in the list, are the requirements met?
Answer:
1. Priority 1: Move towards resource location (Action: Move West, East, North, or South)
   Requirement: Flat ground in the direction of the resource.
   Yes, the requirement is met. There is grass 1 step to the west, a path 1 step to the north, and a tree 3 steps to the south, which all indicate flat ground in those directions.

2. Priority 2: Collect Wood (Action: Do)
   Requirement: Facing a tree and have the wood_pickaxe in inventory.
   No, the requirement is not met. Although the player has a wood_pickaxe in inventory, they are not facing a tree.

3. Priority 3: Collect Stone (Action: Do)
   Requirement: Facing a stone block and have the stone_pickaxe in inventory.
   No, the requirement is not met. Although the player has a stone_pickaxe in inventory, they are not facing a stone block.

4. Priority 4: Collect Iron (Action: Do)
   Requirement: Facing an iron ore block and have the stone_pickaxe in inventory.
   No, the requirement is not met. The player has a stone_pickaxe in inventory, but they are not facing an iron ore block.

5. Priority 5: Sleep (Action: Sleep)
   Requirement: Energy level is below maximum.
   Yes, the requirement is met. The player's energy is 7/9, which is below maximum.

Question: Did the last player action succeed? Why?
Answer:
It is unclear whether the last player action succeeded or not, as the observation provided after the action doesn't show the stone sword in the player's inventory. However, the player met the requirements by having a nearby table and sufficient resources.

Question: Choose the best executable action from above.
Answer:
The best executable action from the list above is Priority 1: Move towards resource location. The player should choose one of the directions: Move West, Move East, Move North, or Move South that will bring them towards a resource location, as there is flat ground in those directions, and the requirement is met.



============Step: 83, Cumulative Reward: 0.0============
You took action move_west.

You see:
- stone 3 steps to your south-east
- path 2 steps to your north-east
- tree 4 steps to your north-west
- table 2 steps to your south-east

You face grass at your front (west).

Your status:
- health: 9/9
- food: 6/9
- drink: 5/9
- energy: 7/9

Your inventory:
- stone: 1
- coal: 2
- iron: 1
- wood_pickaxe: 1
- stone_pickaxe: 1
------------ QA ------------
Question: Action
Answer:
Sleep

Question: List objects in the current observation. For each object, breifly answer what resource it provides and its requirement.
Answer:
1. Stone:
   - Resource it provides: Stone is used for crafting stone tools and building structures.
   - Requirement: A wood pickaxe to collect it.

2. Path:
   - Resource it provides: Provides easier movement and navigation.
   - Requirement: None.

3. Tree:
   - Resource it provides: Wood for crafting tools, weapons, and building structures.
   - Requirement: No specific tool is required, but an axe would make it more efficient.

4. Table:
   - Resource it provides: Enables crafting of tools, weapons, and other items.
   - Requirement: Placing it requires wood in your inventory.

5. Grass:
   - Resource it provides: Can be used for crafting or creating a path.
   - Requirement: None.

Note that in the most recent observation, you do not see saplings, cows, water sources, or monsters.

Question: What was the last action taken by the player?
Answer:
The last action taken by the player was move_west.

Question: For each object in the list, are the requirements met for interaction?
Answer:
1. Stone:
   - Requirement: A wood pickaxe to collect it.
   - Are requirements met? Yes, you have a wood_pickaxe in your inventory.

2. Path:
   - Requirement: None.
   - Are requirements met? Yes, no requirements are needed to interact with the path.

3. Tree:
   - Requirement: No specific tool is required, but an axe would make it more efficient.
   - Are requirements met? Yes, you can interact with the tree without a specific tool, but having an axe would be better.

4. Table:
   - Requirement: Placing it requires wood in your inventory.
   - Are requirements met? Not applicable in this case, as you already see a table in the environment. You can interact with the table by crafting tools and weapons with the available resources in your inventory.

5. Grass:
   - Requirement: None.
   - Are requirements met? Yes, no requirements are needed to interact with the grass.

Question: List top 3 sub-tasks the player should follow. Indicate their priority out of 5.
Answer:
1. Restore food and drink levels (Priority: 5/5)
   Sub-tasks:
   - Find and consume a cow or other food sources, such as fruits from trees, to increase your food level.
   - Locate a lake or water source to replenish your drink level.

2. Increase your inventory of resources (Priority: 4/5)
   Sub-tasks:
   - Collect more stone using your wood_pickaxe or stone_pickaxe.
   - Gather wood from the tree to create more tools and structures.

3. Upgrade tools and weapons (Priority: 3/5)
   Sub-tasks:
   - Craft an iron pickaxe and an iron sword using the iron in your inventory. This may require collecting more iron as well.
   - Return to the table you face to your south-east to craft the upgraded tools and weapons.

Question: What are the requirements for the top sub-task? What should the player do first?
Answer:
Top Sub-task: Restore food and drink levels (Priority: 5/5)

Requirements:
1. Find a food source (cow, fruit, etc.) to increase your food level.
2. Locate a water source (lake, river, etc.) to replenish your drink level.

The player should first find a food source to increase their food level. They can do this by exploring their surroundings, moving towards a cow, or looking for fruit-bearing trees. Once they have located a food source, they can consume it to increase their food level. After taking care of their food needs, the player should focus on finding a water source like a lake or river, and consume water from it to restore their drink level.

Question: List top 5 actions the player should take and the requirement for each action. Choose ONLY from the list of all actions. Indicate their priority out of 5.
Answer:
1. Action: Move (Priority: 4/5)
   Requirement: Locate a food source (such as a cow or fruit-bearing tree) in any direction (west, east, north, or south)
   
2. Action: Do (Priority: 4/5)
   Requirement: Face the food source (cow or fruit) and have the necessary tool or interaction (eat fruit, chase cow)
   
3. Action: Move (Priority: 3/5)
   Requirement: Locate a water source (lake, river) in any direction (west, east, north, or south)

4. Action: Drink (Priority: 3/5)
   Requirement: Face a water source (lake, river)

5. Action: Sleep (Priority: 2/5)
   Requirement: Energy level is below maximum, and find a safe place to rest (shelter or safe distance from monsters)

Question: For each action in the list, are the requirements met?
Answer:
1. Action: Move (Priority: 4/5)
   Requirement: Locate a food source (such as a cow or fruit-bearing tree) in any direction (west, east, north, or south)
   Status: Not met, as there is no visible food source in the player's current observations.

2. Action: Do (Priority: 4/5)
   Requirement: Face the food source (cow or fruit) and have the necessary tool or interaction (eat fruit, chase cow)
   Status: Not met, as there is no visible food source in the player's current observations.

3. Action: Move (Priority: 3/5)
   Requirement: Locate a water source (lake, river) in any direction (west, east, north, or south)
   Status: Not met, as there is no visible water source in the player's current observations.

4. Action: Drink (Priority: 3/5)
   Requirement: Face a water source (lake, river)
   Status: Not met, as there is no visible water source in the player's current observations.

5. Action: Sleep (Priority: 2/5)
   Requirement: Energy level is below maximum, and find a safe place to rest (shelter or safe distance from monsters)
   Status: Partially met. Energy level is below maximum (7/9), but the player needs to find a safe place to rest.

Question: Did the last player action succeed? Why?
Answer:
Yes, the last player action (move_west) succeeded because there was flat ground (grass) to the west of the player, allowing them to move in that direction.

Question: Choose the best executable action from above.
Answer:
The best executable action from above is:

5. Action: Sleep (Priority: 2/5)
   Requirement: Energy level is below maximum, and find a safe place to rest (shelter or safe distance from monsters)

However, the player still needs to find a safe place to rest before executing this action.



============Step: 84, Cumulative Reward: 0.0============
You took action sleep.

You see:
- stone 3 steps to your south-east
- path 2 steps to your north-east
- tree 4 steps to your north-west
- table 2 steps to your south-east

You face grass at your front (west).

Your status:
- health: 9/9
- food: 6/9
- drink: 5/9
- energy: 7/9

Your inventory:
- stone: 1
- coal: 2
- iron: 1
- wood_pickaxe: 1
- stone_pickaxe: 1
------------ QA ------------
Question: Action
Answer:
Place Furnace

Question: List objects in the current observation. For each object, breifly answer what resource it provides and its requirement.
Answer:
1. Stone: Provides stone resource, requires a wood_pickaxe or better to collect.
2. Path: No resources, allows easier navigation and movement.
3. Tree: Provides wood resource, can be collected with any tool or bare hands.
4. Table: Allows crafting of tools and weapons, requires wood to be constructed.
5. Grass: No resources, serves as terrain.
6. Coal (inventory): Used for smelting iron in the furnace, requires a wood_pickaxe or better to collect.
7. Iron (inventory): Used for crafting iron tools and weapons, requires a stone_pickaxe or better to collect and a furnace to smelt.
8. Wood_pickaxe (inventory): Used for collecting stone and coal, requires wood to craft.
9. Stone_pickaxe (inventory): Used for collecting iron, requires stone and wood to craft.

Question: What was the last action taken by the player?
Answer:
The last action taken by the player was "sleep."

Question: For each object in the list, are the requirements met for interaction?
Answer:
1. Stone: Yes, you have a wood_pickaxe and stone_pickaxe in your inventory.
2. Path: There are no requirements for interaction, as it's for easier navigation.
3. Tree: Yes, you can collect wood with any tool or your bare hands.
4. Table: Yes, it's already constructed and can be used for crafting.
5. Grass: There are no requirements for interaction, as it's just terrain.
6. Coal (inventory): Interaction not applicable, since it's already collected.
7. Iron (inventory): Interaction not applicable, since it's already collected.
8. Wood_pickaxe (inventory): Interaction not applicable, as it's a tool in your inventory.
9. Stone_pickaxe (inventory): Interaction not applicable, as it's a tool in your inventory.

Question: List top 3 sub-tasks the player should follow. Indicate their priority out of 5.
Answer:
1. Increase Drink (Priority: 5) - Find a water source to replenish your thirst, which is currently at 5/9.
2. Increase Food (Priority: 4) - Hunt animals like cows or gather plants to increase your food, which is currently at 6/9.
3. Collect more resources (Priority: 3) - Gather more stone, iron, and coal to craft better tools and weapons, and prepare for smelting iron using the furnace.

Question: What are the requirements for the top sub-task? What should the player do first?
Answer:
Top sub-task: Increase Drink (Priority: 5)

Requirements: Find a water source (e.g., lake, river) and interact with it to drink and replenish thirst.

First step: The player should explore the surroundings in search of a water source like a lake, river, or any other form of drinkable water. Once found, the player should interact with the water source to drink and restore their thirst levels.

Question: List top 5 actions the player should take and the requirement for each action. Choose ONLY from the list of all actions. Indicate their priority out of 5.
Answer:
1. Priority 5 - Move West/East/North/South: Flat ground in the chosen direction; The player should explore their surroundings in search of a water source to replenish their drink levels, as it is the top priority right now.

2. Priority 4 - Make Iron Sword: Nearby table, furnace; Wood, coal, iron in inventory; Crafting an iron sword will be useful in defending against monsters and improving combat capabilities.

3. Priority 3 - Place Furnace: Stone in inventory; Placing a furnace will allow the player to smelt iron and progress in the game.

4. Priority 2 - Noop: Always applicable; If no immediate actions are required and there's no imminent danger, the player may choose to pause and look for their next move.

5. Priority 1 - Sleep: Energy level is below maximum; Although the player has 7/9 energy, consider finding a safe place to sleep to restore their energy back to full when it drops further.

Question: For each action in the list, are the requirements met?
Answer:
1. Move West/East/North/South: The requirements are met, as there is flat ground available in each direction.

2. Make Iron Sword: The requirements are NOT met, as there is no nearby furnace. The player must first place a furnace to craft an iron sword.

3. Place Furnace: The requirements are met, as the player has enough stone in their inventory.

4. Noop: The requirements are met, as this action is always applicable.

5. Sleep: The requirements are NOT met, as the player's energy level is not below maximum. The player has 7/9 energy currently.

Question: Did the last player action succeed? Why?
Answer:
It is unclear whether the last player action, "sleep," succeeded because the energy level remains the same at 7/9 in both Player Observation Step 166 and Player Observation Step 168.

Question: Choose the best executable action from above.
Answer:
The best executable action from the list above is "Place Furnace" since the requirements are met, and it will help the player progress in the game by allowing them to smelt iron.



============Step: 85, Cumulative Reward: 0.0============
You took action place_furnace.

You see:
- stone 3 steps to your south-east
- path 2 steps to your north-east
- tree 4 steps to your north-west
- table 2 steps to your south-east

You face grass at your front (west).

Your status:
- health: 9/9
- food: 6/9
- drink: 5/9
- energy: 7/9

Your inventory:
- stone: 1
- coal: 2
- iron: 1
- wood_pickaxe: 1
- stone_pickaxe: 1
------------ QA ------------
Question: Action
Answer:
Move East

Question: List objects in the current observation. For each object, breifly answer what resource it provides and its requirement.
Answer:
1. Stone: Provides stone resource when mined with a pickaxe; requires a pickaxe (wood or higher) to mine.
2. Path: Provides no resources directly; requires clear and flat ground to form.
3. Tree: Provides wood resource when chopped down; requires an axe (wood or higher) to chop.
4. Table: Provides a surface for crafting items; requires wood to build and place.
5. Grass: Provides no resources directly; requires a tool like a hoe to clear or turn into farmland.

Note: The furnace is missing from the list since it was placed in step 170, so adding it here.

6. Furnace: Provides the ability to smelt/convert ores (like iron and coal) into useful items; requires stone to build and ores (e.g. iron or coal) for smelting.

Question: What was the last action taken by the player?
Answer:
The last action taken by the player was "place_furnace."

Question: For each object in the list, are the requirements met for interaction?
Answer:
1. Stone: Yes, the requirements are met for interaction. You have a stone_pickaxe in your inventory, which can be used to mine the stone.
2. Path: Yes, the requirements are met for interaction. You can walk on the path as there's no specific tool required.
3. Tree: No, the requirements are not met for interaction. You need an axe (wood or higher) to chop down the tree, but you don't have one in your inventory.
4. Table: Yes, the requirements are met for interaction. You can craft items at the table.
5. Grass: Yes, the requirements are met for interaction. While you don't have a hoe to clear or turn grass into farmland, you can still move over it without any issues.

Note: Adding the furnace to the answer.

6. Furnace: Yes, the requirements are met for interaction. You have coal and iron in your inventory, which can be used in the furnace for smelting.

Question: List top 3 sub-tasks the player should follow. Indicate their priority out of 5.
Answer:
1. Smelt iron in the furnace (Priority: 5/5): You have the required coal and iron in your inventory, so using the furnace to smelt iron would allow you to craft better items, such as an iron sword or iron pickaxe.

2. Drink water to restore thirst (Priority: 4/5): Your drink level is at 5/9, which means it's getting low. Finding water to drink should be a priority to avoid dehydration.

3. Find food to restore hunger (Priority: 4/5): Your food level is at 6/9, so finding a source of food like a cow or a fruit would help restore your hunger and maintain good health.

Question: What are the requirements for the top sub-task? What should the player do first?
Answer:
The requirements for the top sub-task (smelt iron in the furnace) are:

1. Have a furnace placed (already accomplished).
2. Have coal in your inventory (already accomplished with 2 coal).
3. Have iron in your inventory (already accomplished with 1 iron).

As all the requirements are already met, the player should first approach the placed furnace and then use it to smelt the iron using the coal they have in their inventory. This will create an iron ingot, which can then be used to craft better items such as an iron sword or an iron pickaxe.

Question: List top 5 actions the player should take and the requirement for each action. Choose ONLY from the list of all actions. Indicate their priority out of 5.
Answer:
1. Priority 1: Move East - Requirement: Flat ground right to the agent.
   Action: Move towards the placed furnace, which is 2 steps to the south-east.

2. Priority 2: Do - Requirement: Facing creature or material; have necessary tool.
   Action: Use the furnace to smelt the iron using coal in the inventory.

3. Priority 3: Make Iron Pickaxe - Requirements: Nearby table, furnace; wood, coal, iron in inventory.
   Action: Craft an iron pickaxe by using resources available in the inventory and using the table.

4. Priority 4: Collect Drink - Requirement: No requirements.
   Action: Find a water source, like a lake or river, to drink and restore the drink meter.

5. Priority 5: Eat Plant - Requirements: Place Plant.
   Action: Place a sapling to grow food, wait for the plants to grow, and then eat to restore the food meter.

Question: For each action in the list, are the requirements met?
Answer:
1. Priority 1: Move East - Requirement: Flat ground right to the agent.
   Yes, the requirement is met. There is flat ground to the east, so the player can move in that direction.

2. Priority 2: Do - Requirement: Facing creature or material; have necessary tool.
   Yes, the requirements are met as the player will be facing the furnace after moving east. The player has the necessary coal and iron in the inventory to smelt iron.

3. Priority 3: Make Iron Pickaxe - Requirements: Nearby table, furnace; wood, coal, iron in inventory.
   Yes, the requirements are met. The player will be near a furnace after moving east. A table is also nearby, and the player has wood, coal, and iron in the inventory.

4. Priority 4: Collect Drink - Requirement: No requirements.
   Partially met. The player can act on this action, but there is no visible water source currently in the observation. The player needs to find a water source to collect a drink.

5. Priority 5: Eat Plant - Requirements: Place Plant.
   No, the requirement is not met, as there is no sapling in the player's inventory to place and grow a plant. The player needs to find and collect saplings to place and grow a plant for food.

Question: Did the last player action succeed? Why?
Answer:
There is not enough information provided to determine if the last player action, "place_furnace," succeeded or not. The observation does not mention the furnace being successfully placed, so we cannot determine its success without more information.

Question: Choose the best executable action from above.
Answer:
The best executable action from the list is:

Priority 1: Move East - Requirement: Flat ground right to the agent.

This action can be executed immediately as the requirement is met, and it brings the player closer to the placed furnace and table.



============Step: 86, Cumulative Reward: 0.0============
You took action move_east.

You see:
- stone 3 steps to your south-east
- path 2 steps to your north-east
- tree 4 steps to your north-west
- table 2 steps to your south-east

You face grass at your front (west).

Your status:
- health: 9/9
- food: 6/9
- drink: 5/9
- energy: 7/9

Your inventory:
- stone: 1
- coal: 2
- iron: 1
- wood_pickaxe: 1
- stone_pickaxe: 1
------------ QA ------------
Question: Action
Answer:
Move West

Question: List objects in the current observation. For each object, breifly answer what resource it provides and its requirement.
Answer:
1. Stone: Provides stone, requires a wood or stone pickaxe to collect.
2. Path: Provides passage/travel, requires flat ground to interact with.
3. Tree: Provides wood, requires a wood or stone axe to collect.
4. Table: Provides crafting station, requires wood to craft.
5. Grass: Decorative element, can be walked on.
6. Iron: Provides iron, requires a stone pickaxe to collect.
7. Coal: Provides coal, which is used as a fuel source, requires a wood or stone pickaxe to collect.
8. Wood pickaxe: Provides ability to collect stone and coal, requires a table to craft.
9. Stone pickaxe: Provides ability to collect iron and other advanced resources, requires a table to craft.
10. Furnace (unseen in step 172 but placed in step 170): Provides smelting abilities for metals like iron and coal, requires stone to craft.

Question: What was the last action taken by the player?
Answer:
The last action taken by the player was move_east.

Question: For each object in the list, are the requirements met for interaction?
Answer:
1. Stone: Yes, the player has a wood_pickaxe and a stone_pickaxe in their inventory to collect stone.
2. Path: Yes, the path requires flat ground to interact with, and its presence indicates that the requirement is met.
3. Tree: No, the player does not currently possess a wood or stone axe to collect wood from the tree.
4. Table: Yes, the player has already placed a table in a previous step, indicating the requirement of wood to craft was met.
5. Grass: Yes, grass can be walked on and does not have any specific requirements for interaction.
6. Iron: Yes, the player has a stone_pickaxe in their inventory that can be used to collect iron.
7. Coal: Yes, the player has a wood_pickaxe and a stone_pickaxe in their inventory to collect coal.
8. Wood pickaxe: Yes, the player has crafted a wood pickaxe in a previous step, meeting the requirement of a table to craft.
9. Stone pickaxe: Yes, the player has crafted a stone pickaxe in a previous step, meeting the requirement of a table to craft.
10. Furnace: Yes, the player placed a furnace in step 170, indicating the requirement of stone to craft was met.

Question: List top 3 sub-tasks the player should follow. Indicate their priority out of 5.
Answer:
1. Collect more iron (Priority: 5/5): The player currently has only one piece of iron in their inventory. Collecting more iron will enable them to craft better tools and weapons such as an iron sword, which will help them defeat stronger monsters.

2. Find and drink water (Priority: 4/5): The player's drink level is at 5/9, which is not critical but needs attention soon. Increasing their drink level will ensure the player avoids dehydration and maintains optimal health in the game.

3. Collect wood and craft an axe (Priority: 3/5): The player currently lacks a wood or stone axe to collect wood from trees. Since wood is a crucial resource for creating tools, weapons, and shelters, obtaining an axe and collecting wood should be one of the next tasks to accomplish.

Question: What are the requirements for the top sub-task? What should the player do first?
Answer:
Top Sub-task: Collect more iron.

Requirements:
1. Locate an iron deposit.
2. Have a stone pickaxe (or better) to mine the iron.

The player already has a stone_pickaxe in their inventory, which can be used to mine iron.

What should the player do first:
1. Explore the environment to find an iron deposit.
2. Once an iron deposit is found, use the stone_pickaxe to mine the iron.

Question: List top 5 actions the player should take and the requirement for each action. Choose ONLY from the list of all actions. Indicate their priority out of 5.
Answer:
1. Priority 1: Move (North, South, East, or West) - Requirement: Flat ground in the desired direction.
   - To explore the environment and find an iron deposit.

2. Priority 2: Move (North, South, East, or West) - Requirement: Flat ground in the desired direction.
   - To continue exploring the environment and get closer to the iron deposit.

3. Priority 3: Move (North, South, East, or West) - Requirement: Flat ground in the desired direction.
   - To reach the iron deposit.

4. Priority 4: Do (mine iron) - Requirement: Facing iron deposit; have necessary tool (stone_pickaxe or better).
   - Mine the iron deposit using the stone_pickaxe.

5. Priority 5: Move (North, South, East, or West) - Requirement: Flat ground in the desired direction.
   - To return to the player's base or a safe location after collecting the iron.

Question: For each action in the list, are the requirements met?
Answer:
1. Priority 1: Move (North, South, East, or West) - Requirement: Flat ground in the desired direction.
   - Yes, the requirement is met, as the player faces grass to their west, which is flat ground.

2. Priority 2: Move (North, South, East, or West) - Requirement: Flat ground in the desired direction.
   - The specific direction cannot be determined. However, there should be at least one flat ground location available for the player to move to by choosing a suitable direction.

3. Priority 3: Move (North, South, East, or West) - Requirement: Flat ground in the desired direction.
   - Similar to Priority 2, the specific direction cannot be determined, but there should be at least one flat ground location available for the player to move to by choosing a suitable direction.

4. Priority 4: Do (mine iron) - Requirement: Facing iron deposit; have necessary tool (stone_pickaxe or better).
   - No, the requirement is not met. The player is not currently facing an iron deposit and they face grass at their front (west).

5. Priority 5: Move (North, South, East, or West) - Requirement: Flat ground in the desired direction.
   - As with Priority 2 and 3, the specific direction cannot be determined, but there should be at least one flat ground location available for the player to move to by choosing a suitable direction.

Question: Did the last player action succeed? Why?
Answer:
Yes, the last player action (move_east) succeeded because there was flat ground to the east of the player, allowing them to move in that direction.

Question: Choose the best executable action from above.
Answer:
The best executable action from above is Priority 1: Move (North, South, East, or West) - Requirement: Flat ground in the desired direction. In this case, the player can move west as there is flat ground (grass) in that direction. This action will allow the player to explore the environment and find an iron deposit.



============Step: 87, Cumulative Reward: 0.0============
You took action move_west.

You see:
- stone 3 steps to your south-east
- path 2 steps to your north-east
- tree 4 steps to your north-west
- table 2 steps to your south-east
- cow 5 steps to your north-west

You face grass at your front (west).

Your status:
- health: 9/9
- food: 6/9
- drink: 5/9
- energy: 7/9

Your inventory:
- stone: 1
- coal: 2
- iron: 1
- wood_pickaxe: 1
- stone_pickaxe: 1
------------ QA ------------
Question: Action
Answer:
Move West

Question: List objects in the current observation. For each object, breifly answer what resource it provides and its requirement.
Answer:
- Stone: Provides stone resource; requires a wood_pickaxe or better to collect.
- Path: Provides easier movement and navigation; no requirements to use.
- Tree: Provides wood resource; requires a wood_pickaxe or better to collect.
- Table: Crafting station for making tools and weapons; requires wood to craft.
- Cow: Provides food (meat) when defeated; requires a weapon, such as a wood_sword or better, to defeat.
- Grass: A basic terrain feature; no resources or requirements.

Question: What was the last action taken by the player?
Answer:
The last action taken by the player was "move_west."

Question: For each object in the list, are the requirements met for interaction?
Answer:
- Stone: Yes, the requirements are met as you have a wood_pickaxe and a stone_pickaxe in your inventory, which can be used to collect stone.
- Path: Yes, the requirements (no requirements) are met, and you can walk on the path for easier movement.
- Tree: Yes, the requirements are met as you have a wood_pickaxe in your inventory, which can be used to collect wood.
- Table: Yes, the requirements (having wood) are met, as you already have a table placed 2 steps to your south-east that you can interact with for crafting.
- Cow: No, the requirements are not met yet, as you do not have a weapon like a wood_sword in your inventory to defeat the cow for food.
- Grass: Yes, the requirements (no requirements) are met, and you can walk on the grass.

Question: List top 3 sub-tasks the player should follow. Indicate their priority out of 5.
Answer:
1. Craft a wood_sword (Priority: 5) - Crafting a wood_sword will allow you to defeat the cow, providing food and helping to restore your food level.

2. Collect wood from the tree (Priority: 4) - Collecting wood is essential for crafting new items, such as a wood_sword, and building structures.

3. Collect more stone (Priority: 3) - Collecting additional stone would be beneficial for crafting more advanced tools, weapons, and building materials, such as a furnace or stone_sword.

Question: What are the requirements for the top sub-task? What should the player do first?
Answer:
Requirements for the top sub-task (Craft a wood_sword):
1. Have wood in your inventory - You will need wood to craft the wood_sword.
2. Have access to a table - You will need a table nearby to craft the wood_sword.

What the player should do first:
Since you already have a table at 2 steps to your south-east, you should collect wood from the tree that is 4 steps to your north-west. To do this, you can move towards the tree by moving north-west and then use your wood_pickaxe to collect wood.

Question: List top 5 actions the player should take and the requirement for each action. Choose ONLY from the list of all actions. Indicate their priority out of 5.
Answer:
1. Priority 1: Move West (Requirement: Flat ground left to the agent)
   Purpose: To move towards the tree to collect wood.

2. Priority 2: Move West (Requirement: Flat ground left to the agent)
   Purpose: To continue moving towards the tree to collect wood.

3. Priority 3: Move North (Requirement: Flat ground above the agent)
   Purpose: To reach the tree to collect wood.

4. Priority 4: Do (Facing creature or material; have necessary tool)
   Requirement: Face the tree and have wood_pickaxe in inventory.
   Purpose: Use wood_pickaxe to collect wood from the tree.

5. Priority 5: Move South (Requirement: Flat ground below the agent)
   Purpose: To move back towards the table for crafting after collecting wood.

Question: For each action in the list, are the requirements met?
Answer:
1. Priority 1: Move West (Requirement: Flat ground left to the agent)
   Yes, the requirement is met. The player faces grass, which is flat ground on the left.

2. Priority 2: Move West (Requirement: Flat ground left to the agent)
   The requirement is likely met due to the presence of grass in the front (west). However, the player needs to complete Priority 1 action and reevaluate the surroundings for flat ground after moving.

3. Priority 3: Move North (Requirement: Flat ground above the agent)
   Since the current observation only shows the immediate surroundings, the requirement might be met after moving west twice. The player needs to update their observation after moving west to confirm if there's flat ground above.

4. Priority 4: Do (Facing creature or material; have necessary tool)
   Requirement: Face the tree and have wood_pickaxe in inventory.
   For this action, the player has the required wood_pickaxe in the inventory. However, they need to reach the tree by completing Priority 1, 2, and 3 actions before they can face the tree and execute this action.

5. Priority 5: Move South (Requirement: Flat ground below the agent)
   Like Priority 3 action, this action's requirement will need to be confirmed after the player updates their observation after completing Priority 1, 2, 3, and 4 actions. Currently, the player's surroundings aren't fully known.

Question: Did the last player action succeed? Why?
Answer:
Yes, the last player action "move_west" succeeded because there were no obstacles or restrictions preventing the player from moving in that direction, and their energy level allowed them to perform the action.

Question: Choose the best executable action from above.
Answer:
The best executable action from the list above is Priority 1: Move West (Requirement: Flat ground left to the agent) as the requirement is currently met, and it will take the player closer to the tree to collect wood.



============Step: 88, Cumulative Reward: 0.0============
You took action move_west.

You see:
- stone 3 steps to your south-east
- path 2 steps to your north-east
- tree 4 steps to your north-west
- table 2 steps to your south-east
- cow 6 steps to your north-west

You face grass at your front (west).

Your status:
- health: 9/9
- food: 6/9
- drink: 5/9
- energy: 7/9

Your inventory:
- stone: 1
- coal: 2
- iron: 1
- wood_pickaxe: 1
- stone_pickaxe: 1
------------ QA ------------
Question: Action
Answer:
Move South

Question: List objects in the current observation. For each object, breifly answer what resource it provides and its requirement.
Answer:
1. Stone: Provides stone resource; requires a wood_pickaxe or better to collect.
2. Path: No resource provided; allows easier navigation with no requirements.
3. Tree: Provides wood resource; requires an empty hand or any tool to collect.
4. Table: Allows crafting tools and weapons; requires wood or stone to craft and place.
5. Cow: Provides food when defeated; requires a weapon or empty hand to defeat.
6. Grass: No resource provided; can be traversed with no requirements.

Question: What was the last action taken by the player?
Answer:
The last action taken by the player was move_west.

Question: For each object in the list, are the requirements met for interaction?
Answer:
1. Stone: Yes, the requirements are met as the player has a wood_pickaxe and a stone_pickaxe in the inventory.
2. Path: Yes, the requirements are met since there is no specific requirement to navigate the path.
3. Tree: Yes, the requirements are met as the player has an empty hand and can also use the wood_pickaxe or stone_pickaxe to collect wood.
4. Table: Yes, the player can interact with the table to craft tools and weapons as they have wood and stone resources in their inventory.
5. Cow: Yes, the player can interact with the cow to obtain food, as they can use an empty hand or the wood_pickaxe or stone_pickaxe to defeat it.
6. Grass: Yes, the requirements are met as the player can traverse the grass with no specific requirements.

Question: List top 3 sub-tasks the player should follow. Indicate their priority out of 5.
Answer:
1. Priority: 5/5 - Increase drink level: The player should find and drink water from a lake or another source to restore their thirst, as the current drink level is 5/9.

2. Priority: 4/5 - Increase food level: The player should consider defeating the cow to obtain food, as their food level is 6/9. Alternatively, they can collect saplings and grow fruits to increase their food supply.

3. Priority: 3/5 - Collect more resources: The player should collect more stone, wood, and possibly iron to craft better tools and weapons, build shelters, or create other useful items.

Question: What are the requirements for the top sub-task? What should the player do first?
Answer:
The requirements for the top sub-task (Increase drink level) are:

1. Finding a water source such as a lake, river, or other body of water.

To complete this top sub-task, the player should do the following first:

1. Explore the surroundings by moving either north, east, or south (since the stone is located to the south-east, the water source may be found in that direction) in search of a water source. Keep an eye out for visible water sources, and once found, approach and drink from it to increase the drink level.

Question: List top 5 actions the player should take and the requirement for each action. Choose ONLY from the list of all actions. Indicate their priority out of 5.
Answer:
1. Priority 1 - Move South (or East): Requirement: Flat ground below (or right) of the agent. This action is required to explore the surroundings and find a water source to increase the drink level.

2. Priority 2 - Drink from Water Source: Requirement: Finding a water source (lake or river) nearby. This action replenishes the player's drink level.

3. Priority 3 - Move towards the Cow: Requirement: Flat ground in the direction of the cow (north-west). This action gets the player closer to the cow to increase the food level by interacting with the cow.

4. Priority 4 - Interact with the Cow: Requirement: Facing the cow and having necessary tools (wood sword, stone sword or iron sword). This action allows the player to obtain food by interacting with the cow.

5. Priority 5 - Eat Food: Requirement: Food in inventory (obtained from the cow). This action increases the player's food level and restores health points.

Question: For each action in the list, are the requirements met?
Answer:
1. Priority 1 - Move South (or East): The requirement for moving south or east is having flat ground below or right of the agent. The requirements are met, as there is no indication of obstacles in the southern or eastern directions.

2. Priority 2 - Drink from Water Source: Drinking from water source requires finding a water source nearby. As there is no water source mentioned in the given observations, the requirement is not met.

3. Priority 3 - Move towards the Cow: The requirement for moving towards the cow is having flat ground in the north-west direction. The observation does not mention any obstacles in the north-west direction, so the requirement is met.

4. Priority 4 - Interact with the Cow: The requirement for interacting with the cow includes facing the cow and having necessary tools (wood sword, stone sword, or iron sword). The player is not currently facing the cow and does not possess any swords in the inventory. Therefore, the requirement is not met.

5. Priority 5 - Eat Food: The requirement for eating food is having food in the inventory. The player does not have any food in the inventory currently, so the requirement is not met.

Question: Did the last player action succeed? Why?
Answer:
Yes, the last player action move_west succeeded because there were no objects or obstacles blocking the player's path to the west, and the ground was flat, allowing them to move in that direction.

Question: Choose the best executable action from above.
Answer:
The best executable action from the list is Priority 1 - Move South (or East) because its requirement of flat ground below (or right) of the agent is met, allowing the player to explore their surroundings and possibly find a water source or other resources.



============Step: 89, Cumulative Reward: 0.0============
You took action move_south.

You see:
- stone 3 steps to your south-east
- path 2 steps to your north-east
- tree 4 steps to your north-west
- table 2 steps to your south-east
- cow 6 steps to your north-west

You face grass at your front (west).

Your status:
- health: 9/9
- food: 6/9
- drink: 5/9
- energy: 7/9

Your inventory:
- stone: 1
- coal: 2
- iron: 1
- wood_pickaxe: 1
- stone_pickaxe: 1
------------ QA ------------
Question: Action
Answer:
Do

Question: List objects in the current observation. For each object, breifly answer what resource it provides and its requirement.
Answer:
1. Stone: Provides stone for crafting tools, weapons, and furnaces. Requirement: Stone pickaxe to collect.
2. Path: Provides a route to travel between different locations. Requirement: None.
3. Tree: Provides wood for crafting tools and structures. Requirement: Wood pickaxe to collect.
4. Table: Provides a surface for crafting tools, weapons, and other items. Requirement: Wood to craft.
5. Cow: Provides food to restore the player's hunger. Requirement: Wood sword or stronger to defeat.
6. Grass: Provides no resource. Requirement: None.

Question: What was the last action taken by the player?
Answer:
The last action taken by the player was move_south.

Question: For each object in the list, are the requirements met for interaction?
Answer:
1. Stone: Yes, the requirements are met as you have a stone_pickaxe in your inventory.
2. Path: Yes, there are no requirements for interacting with the path.
3. Tree: Yes, the requirements are met as you have a wood_pickaxe in your inventory.
4. Table: Yes, the requirements are met as you have a table in your vicinity (2 steps to your south-east).
5. Cow: Yes, the requirements are met as you have a wood sword or stronger (stone_pickaxe) in your inventory.
6. Grass: Yes, there are no requirements for interacting with grass.

Question: List top 3 sub-tasks the player should follow. Indicate their priority out of 5.
Answer:
1. Priority 5/5: Restore drink - Your drink level is at 5/9. Find a water source to drink and replenish your hydration.

2. Priority 4/5: Restore food - Your food level is at 6/9. Defeat the cow you see nearby (6 steps to your north-west) to collect and eat meat.

3. Priority 3/5: Collect resources - Collect more stone (3 steps to your south-east), wood (from the tree 4 steps to your north-west), and iron to improve tools and craft more items.

Question: What are the requirements for the top sub-task? What should the player do first?
Answer:
Top sub-task: Restore drink

Requirements:
- Find a water source (lake, river, or well) to drink from.

What should the player do first:
- Move to a water source to drink and replenish your hydration. You may need to explore your surroundings to find one.

Question: List top 5 actions the player should take and the requirement for each action. Choose ONLY from the list of all actions. Indicate their priority out of 5.
Answer:
1. Priority 1: Move (North, South, East, or West) - Requirement: Flat ground in the direction of movement. This action is essential for the player to explore their surroundings, find resources and water sources.

2. Priority 2: Collect Drink - Requirement: Facing a water source (lake, river, or well). It is necessary to maintain hydration levels.

3. Priority 3: Make Wood Pickaxe - Requirement: Nearby table; wood in inventory. Crafting a wood pickaxe allows the player to gather essential resources like stone and coal.

4. Priority 4: Collect Wood - Requirement: No requirements. The player should collect wood as it is a basic resource needed for crafting tools, weapons, and creating shelters.

5. Priority 5: Make Wood Sword - Requirement: Nearby table; wood in inventory. Crafting a wood sword provides the player with a basic weapon to defend themselves against monsters.

Question: For each action in the list, are the requirements met?
Answer:
1. Move (North, South, East, or West): Yes, the requirement for flat ground in the direction of movement is met. The player can move in any direction as long as there is flat ground.

2. Collect Drink: No, the requirement of facing a water source is not met currently. The player will need to find a water source (lake, river, or well) to satisfy the requirement.

3. Make Wood Pickaxe: No, the requirement of having wood in the inventory is not met. The player already has a wood_pickaxe in the inventory, so this action is unnecessary.

4. Collect Wood: Yes, there is no specific requirement for collecting wood. However, the player needs to move closer to the tree mentioned in the observation (4 steps to the north-west) before attempting to collect wood.

5. Make Wood Sword: No, the requirement of having wood in inventory is not met. The player needs to collect wood before attempting to craft a wood sword.

Question: Did the last player action succeed? Why?
Answer:
Yes, the last player action succeeded. The player moved south because there were no obstacles or restrictions mentioned in the player's observation, indicating that the path was clear and the action could be executed without any issues.

Question: Choose the best executable action from above.
Answer:
The best executable action from the list above is to Move (North, South, East, or West) as the requirement for flat ground in the desired direction is met. Specifically, the player can move towards the tree by moving North or West to collect wood, which will help fulfill the requirements for other actions.



============Step: 90, Cumulative Reward: 0.0============
You took action do.

You see:
- stone 3 steps to your south-east
- path 2 steps to your north-east
- tree 4 steps to your north-west
- table 2 steps to your south-east
- cow 5 steps to your north-west

You face grass at your front (west).

Your status:
- health: 9/9
- food: 6/9
- drink: 5/9
- energy: 7/9

Your inventory:
- stone: 1
- coal: 2
- iron: 1
- wood_pickaxe: 1
- stone_pickaxe: 1
------------ QA ------------
Question: Action
Answer:
Move North

Question: List objects in the current observation. For each object, breifly answer what resource it provides and its requirement.
Answer:
1. Stone: 3 steps to your south-east
   Resource provided: Collect more stones for crafting tools and building.
   Requirement: Stone pickaxe

2. Path: 2 steps to your north-east
   Resource provided: None, a path helps you navigate through the terrain easily.
   Requirement: None

3. Tree: 4 steps to your north-west
   Resource provided: Wood for crafting tools, weapons, and building shelters.
   Requirement: Wood pickaxe

4. Table: 2 steps to your south-east
   Resource provided: Crafting station for making advanced tools and weapons.
   Requirement: Wood to build the table

5. Cow: 5 steps to your north-west
   Resource provided: Food to restore your health.
   Requirement: Wood or better sword for killing the cow

Question: What was the last action taken by the player?
Answer:
The last action taken by the player was "do."

Question: For each object in the list, are the requirements met for interaction?
Answer:
1. Stone: 3 steps to your south-east
   Requirement: Stone pickaxe
   Requirement met: Yes, you have a stone_pickaxe in your inventory.

2. Path: 2 steps to your north-east
   Requirement: None
   Requirement met: Yes, no special requirement needed for interaction.

3. Tree: 4 steps to your north-west
   Requirement: Wood pickaxe
   Requirement met: Yes, you have a wood_pickaxe in your inventory.

4. Table: 2 steps to your south-east
   Requirement: Wood to build the table
   Requirement met: Yes, the table is already built and can be used for crafting.

5. Cow: 5 steps to your north-west
   Requirement: Wood or better sword for killing the cow
   Requirement met: No, you currently do not have a wood or better sword in your inventory.

Question: List top 3 sub-tasks the player should follow. Indicate their priority out of 5.
Answer:
1. Craft a wood sword (Priority: 5)
   Sub-task: Use the table to craft a wood sword, which will allow you to kill the cow for food and defend yourself from monsters.

2. Collect more wood (Priority: 4)
   Sub-task: Chop down the tree located 4 steps to your north-west using your wood_pickaxe to gather more wood for crafting and building purposes.

3. Drink water (Priority: 3)
   Sub-task: Find a water source such as a lake, to drink and replenish your thirst level, which is currently at 5/9.

Question: What are the requirements for the top sub-task? What should the player do first?
Answer:
Top Sub-task: Craft a wood sword

Requirements:
- Wood in inventory (currently missing)
- Nearby crafting table (already available, 2 steps to the south-east)

What should the player do first:
1. Move towards the tree (4 steps to the north-west) to collect more wood.
2. Use your wood_pickaxe to chop down the tree to get the required wood.
3. Return to the crafting table (2 steps south-east) and craft a wood sword using the collected wood.

Question: List top 5 actions the player should take and the requirement for each action. Choose ONLY from the list of all actions. Indicate their priority out of 5.
Answer:
1. Priority 1: Move North (Move towards the tree)
   Requirement: Flat ground above the agent.

2. Priority 2: Move West (Move towards the tree)
   Requirement: Flat ground left to the agent.

3. Priority 3: Do (Chop down the tree with the wood_pickaxe)
   Requirement: Facing the tree (creature or material), and have the necessary tool (wood_pickaxe).

4. Priority 4: Move South (Move back to the crafting table)
   Requirement: Flat ground below the agent.

5. Priority 5: Make Wood Sword (Craft a wood sword at the crafting table)
   Requirement: Nearby table and wood in inventory.

Question: For each action in the list, are the requirements met?
Answer:
1. Priority 1: Move North (Move towards the tree)
   Requirement: Flat ground above the agent.
   Status: Yes, there is a path 2 steps to the agent's north-east, so moving north is possible.

2. Priority 2: Move West (Move towards the tree)
   Requirement: Flat ground left to the agent.
   Status: Yes, the agent is currently facing grass to the west, so moving west is possible.

3. Priority 3: Do (Chop down the tree with the wood_pickaxe)
   Requirement: Facing the tree (creature or material), and have the necessary tool (wood_pickaxe).
   Status: No, the agent is currently facing grass and not the tree. The agent needs to move closer to the tree and face it before executing this action.

4. Priority 4: Move South (Move back to the crafting table)
   Requirement: Flat ground below the agent.
   Status: Yes, the agent has just moved north (step 178), so moving south is possible, as there should be flat ground below the agent.

5. Priority 5: Make Wood Sword (Craft a wood sword at the crafting table)
   Requirement: Nearby table and wood in inventory.
   Status: No, the agent does not currently have wood in their inventory. The agent needs to chop down the tree to obtain wood before executing this action.

Question: Did the last player action succeed? Why?
Answer:
It's not clear if the last player action "do" succeeded, as there is no immediate change or result mentioned in the following observation. The player faced grass at their front (west) before and after the action, and there is no mention of an interaction with the grass or any item being used. There might not have been a valid target or action for the "do" command in this particular situation.

Question: Choose the best executable action from above.
Answer:
The best executable action from the list above is Priority 1: Move North (Move towards the tree), as it meets the requirement (flat ground above the agent) and allows the player to progress towards collecting wood, which is necessary for crafting a wood sword.



============Step: 91, Cumulative Reward: 0.0============
You took action move_north.

You see:
- stone 3 steps to your south-east
- path 2 steps to your north-east
- tree 4 steps to your north-west
- table 2 steps to your south-east

You face grass at your front (west).

Your status:
- health: 9/9
- food: 6/9
- drink: 5/9
- energy: 7/9

Your inventory:
- stone: 1
- coal: 2
- iron: 1
- wood_pickaxe: 1
- stone_pickaxe: 1
------------ QA ------------
Question: Action
Answer:
Move West

Question: List objects in the current observation. For each object, breifly answer what resource it provides and its requirement.
Answer:
1. Stone: Provides stone resource, requires a wood or stone pickaxe to collect.
2. Path: Provides a flat surface for walking, no requirement to interact.
3. Tree: Provides wood resource, requires an axe or any tool to collect.
4. Table: Provides a surface for crafting, no requirement to interact.
5. Grass: Provides no resources, no requirement to interact.
6. Cow (not visible in Step 182): Provides food, requires any melee weapon or tool to hunt.

Question: What was the last action taken by the player?
Answer:
The last action taken by the player was move_north.

Question: For each object in the list, are the requirements met for interaction?
Answer:
1. Stone: Yes, the requirements are met as you have both a wood_pickaxe and a stone_pickaxe in your inventory.
2. Path: Yes, there are no requirements to interact, and you can walk on the path.
3. Tree: Yes, the requirements are met as you have wood_pickaxe and stone_pickaxe that can be used to collect wood.
4. Table: Yes, there are no requirements to interact with the table, and you can use it for crafting.
5. Grass: Yes, there are no requirements to interact with the grass.
6. Cow (not visible in Step 182): Yes, the requirements are met as you have a wood_pickaxe and stone_pickaxe, which can be used as melee weapons for hunting.

Question: List top 3 sub-tasks the player should follow. Indicate their priority out of 5.
Answer:
1. Priority 5: Find and drink from a water source (lake or river) to restore the drink meter, which is currently at 5/9.
2. Priority 4: Collect more wood from the tree to create more tools, weapons, or shelter.
3. Priority 3: Find and interact with the cow to gather food and restore the food meter, which is currently at 6/9.

Question: What are the requirements for the top sub-task? What should the player do first?
Answer:
Requirements for the top sub-task (Find and drink from a water source):
1. Locate a water source such as a lake or river within the game.
2. Approach the water source.
3. Interact with the water source to drink and restore the drink meter.

The player should first locate a water source in the game by exploring the surroundings, then move towards the water source to approach it. Once near the water source, the player should interact with it to drink and restore their drink meter.

Question: List top 5 actions the player should take and the requirement for each action. Choose ONLY from the list of all actions. Indicate their priority out of 5.
Answer:
1. Priority 1: Move West (Flat ground left to the agent) - This action allows the player to explore the surroundings in the hope of finding a water source.

2. Priority 2: Move East (Flat ground right to the agent) - Similar to the previous action, this action allows the player to explore the surroundings and potentially find a water source.

3. Priority 3: Move North (Flat ground above of the agent) - Moving north can also help the player explore their surroundings and locate a water source.

4. Priority 4: Move South (Flat ground below the agent) - Like the other movement actions, moving south can help the player find a water source.

5. Priority 5: Drink (Facing a water source) - Once the player finds a water source and faces it, they should take this action to restore their drink meter.

Note: The player should try the movement actions in different sequences to increase the chances of finding a water source, rather than following the order 1-4 directly.

Question: For each action in the list, are the requirements met?
Answer:
For the current player observation (Step 182), the requirements for each listed action are as follows:

1. Priority 1: Move West (Flat ground left to the agent) - The requirement is met as there is flat ground (grass) to the west.

2. Priority 2: Move East (Flat ground right to the agent) - Requirement unknown, as the player observation does not provide information about the east direction.

3. Priority 3: Move North (Flat ground above the agent) - Requirement already met in the most recent step when the player moved north. Based on the current player observation, there might still be flat ground to move further north, but it's not directly mentioned.

4. Priority 4: Move South (Flat ground below the agent) - The requirement is met since the player has just moved from the south position and the ground is flat there.

5. Priority 5: Drink (Facing a water source) - The requirement is not met because the player observation does not yet show a water source.

Question: Did the last player action succeed? Why?
Answer:
Yes, the last player action (move_north) succeeded because there was flat ground available for the player to move north, as mentioned in the observation: "path 2 steps to your north-east" and no obstacles were blocking the player's way.

Question: Choose the best executable action from above.
Answer:
The best executable action from the list above is Priority 1: Move West (Flat ground left to the agent) because there is flat ground (grass) to the west as mentioned in the player's observation, and it allows the player to explore the surroundings in search of resources like a water source.



============Step: 92, Cumulative Reward: 0.0============
You took action move_west.

You see:
- stone 3 steps to your south-east
- path 2 steps to your north-east
- tree 4 steps to your north-west
- table 2 steps to your south-east
- cow 5 steps to your north-west

You face grass at your front (west).

Your status:
- health: 9/9
- food: 6/9
- drink: 5/9
- energy: 7/9

Your inventory:
- stone: 1
- coal: 2
- iron: 1
- wood_pickaxe: 1
- stone_pickaxe: 1
------------ QA ------------
Question: Action
Answer:
Move North

Question: List objects in the current observation. For each object, breifly answer what resource it provides and its requirement.
Answer:
1. Stone: Provides stone resource; requires a wood_pickaxe or better to collect it.
2. Path: Provides easier movement and navigation; no requirements to interact with.
3. Tree: Provides wood resource; requires a wood_pickaxe or better to collect it.
4. Table: Provides crafting possibilities; no requirements to interact with.
5. Cow: Provides food when defeated; requires a weapon, such as a wood or stone sword, to defeat it.
6. Grass: Can be used for landscaping or building purposes; no requirements to interact with.

Question: What was the last action taken by the player?
Answer:
The last action taken by the player was move_west.

Question: For each object in the list, are the requirements met for interaction?
Answer:
1. Stone: Yes, you have a wood_pickaxe and a stone_pickaxe in your inventory, either of which can be used to collect the stone.
2. Path: Yes, there are no specific requirements to interact with the path, as it simply makes movement and navigation easier.
3. Tree: Yes, you have a wood_pickaxe in your inventory, which can be used to collect wood from the tree.
4. Table: Yes, there are no specific requirements to interact with the table; it can be used for crafting when you are in its vicinity.
5. Cow: Yes, you can interact with the cow to defeat it and obtain food. While you don't currently have a sword in your inventory, you can craft one using resources and the table, or attempt to defeat the cow using any other available tools or methods.
6. Grass: Yes, there are no specific requirements to interact with the grass, and it can be used for various purposes, such as landscaping or building.

Question: List top 3 sub-tasks the player should follow. Indicate their priority out of 5.
Answer:
1. Priority 5/5: Craft a weapon, such as a wood or stone sword, using the table to improve your ability to defeat the cow and potentially other threats. This will provide you with food and increase your chances of survival.

2. Priority 4/5: Collect wood from the tree using your wood_pickaxe. Wood is an essential resource for crafting and building, expanding your options for items to use and shelter.

3. Priority 3/5: Collect stone from the stone block using your stone_pickaxe to gather more resources for crafting advanced tools, weapons, or building materials. This will allow you to progress further in the game and increase your overall capabilities.

Question: What are the requirements for the top sub-task? What should the player do first?
Answer:
For the top sub-task of crafting a weapon (wood or stone sword), the requirements are:

1. Have a nearby table, which is already 2 steps to the south-east of the player.
2. Have the necessary materials in the inventory:
   - For a wood sword: wood.
   - For a stone sword: wood and stone (you have 1 stone in your inventory already).

Since you don't have wood in your inventory currently, the first thing you should do is collect wood from the tree 4 steps to your north-west using your wood_pickaxe. Once you've collected enough wood, move to the table, and craft a weapon of your choice (either a wood or stone sword).

Question: List top 5 actions the player should take and the requirement for each action. Choose ONLY from the list of all actions. Indicate their priority out of 5.
Answer:
1. Priority 1: Action - Move North (3 times): Requirement - Flat ground above the agent.
   Purpose: Reach the tree in order to collect wood with the wood_pickaxe.

2. Priority 2: Action - Do: Requirement - Facing the tree; have wood_pickaxe in the inventory.
   Purpose: Collect wood from the tree using the wood_pickaxe.

3. Priority 3: Action - Move South (3 times): Requirement - Flat ground below the agent.
   Purpose: Move back to the table (2 steps to the south-east) to craft a weapon.

4. Priority 4: Action - Do: Requirement - Facing the table; have necessary materials (wood or wood & stone) and tools in the inventory.
   Purpose: Create a weapon (wood or stone sword) at the table.

5. Priority 5: Action - Move North, East, or West (based on situation): Requirement - Flat ground in the desired direction.
   Purpose: Explore the environment, look for resources or targets, and prepare for potential combat using the newly crafted weapon.

Question: For each action in the list, are the requirements met?
Answer:
1. Priority 1: Action - Move North (3 times): Requirement - Flat ground above the agent.
   Are requirements met? Yes. You can move north as there is flat ground above the agent.

2. Priority 2: Action - Do: Requirement - Facing the tree; have wood_pickaxe in the inventory.
   Are requirements met? No. You need to reach the tree first by moving north 3 times, as suggested in priority 1.

3. Priority 3: Action - Move South (3 times): Requirement - Flat ground below the agent.
   Are requirements met? Not yet. You need to complete priority 1 and 2 actions before moving south, as suggested.

4. Priority 4: Action - Do: Requirement - Facing the table; have necessary materials (wood or wood & stone) and tools in the inventory.
   Are requirements met? Not yet. You need to collect wood by performing priorities 1 and 2 actions first, and then move south as suggested in priority 3.

5. Priority 5: Action - Move North, East, or West (based on situation): Requirement - Flat ground in the desired direction.
   Are requirements met? Yes, but not recommended at this point, since priorities 1-4 actions should be completed first to ensure you have collected resources and crafted a weapon.

Question: Did the last player action succeed? Why?
Answer:
Yes, the last player action (move_west) succeeded. It succeeded because there was flat ground (grass) in front of the player's direction (west), which allowed them to move in that direction.

Question: Choose the best executable action from above.
Answer:
The best executable action from the list above is Priority 1: Action - Move North (3 times). The requirement of having flat ground above the agent is met, and this action will help the player reach the tree to collect wood.



============Step: 93, Cumulative Reward: 0.0============
You took action move_north.

You see:
- stone 3 steps to your south-east
- path 2 steps to your north-east
- tree 4 steps to your north-west
- table 2 steps to your south-east
- cow 5 steps to your north-west

You face grass at your front (west).

Your status:
- health: 9/9
- food: 6/9
- drink: 5/9
- energy: 7/9

Your inventory:
- stone: 1
- coal: 2
- iron: 1
- wood_pickaxe: 1
- stone_pickaxe: 1
------------ QA ------------
Question: Action
Answer:
Move West

Question: List objects in the current observation. For each object, breifly answer what resource it provides and its requirement.
Answer:
1. Stone: Provides stone resource, requires a wood or stone pickaxe to mine.
2. Path: Provides easier movement and traveling, no requirements.
3. Tree: Provides wood resource, requires an axe or hands to chop down.
4. Table: Provides a surface for crafting tools and items, requires wood to craft.
5. Cow: Provides food (meat) resource and leather, requires a weapon or hands to kill.
6. Grass: Provides a basic terrain, no direct resources, and no requirements.

Question: What was the last action taken by the player?
Answer:
The last action taken by the player was "move_north."

Question: For each object in the list, are the requirements met for interaction?
Answer:
1. Stone: Yes, the requirements are met for interaction as you have both a wood_pickaxe and a stone_pickaxe in your inventory, which can be used to mine stone.
2. Path: Yes, the requirements for interacting with the path are met, as there are no specific requirements to move on it. You can move freely.
3. Tree: No, the requirements for chopping down a tree are not met because you don't have an axe in your inventory. However, you can still try to collect wood using your hands, although it may take longer and be less efficient.
4. Table: Yes, the requirements to interact with the table are met because you have the necessary materials in your inventory to craft items on the table.
5. Cow: Yes, the requirements to interact with the cow are met as you can use your hands to fight even though a weapon would be more efficient. Since you have a stone_pickaxe, it can also be used as an improvised weapon to attack the cow.
6. Grass: Yes, the requirements for interacting with the grass are met; it's a basic terrain with no specific requirements.

Question: List top 3 sub-tasks the player should follow. Indicate their priority out of 5.
Answer:
1. Gather more wood (Priority: 4/5): Wood is essential for crafting tools and items in the game. You currently do not have wood in your inventory, making it a high priority to gather more wood for future use.

2. Drink water (Priority: 5/5): Your thirst level is at 5/9, which means it is important to find a water source and drink from it to restore your thirst and avoid negative effects due to dehydration. This is the highest priority task.

3. Find food (Priority: 3/5): Your food level is currently 6/9, and it is important to maintain a good food supply to avoid starvation. Although you can see a cow in your surroundings, prioritize finding and hunting animals or harvesting plants for food to keep your levels up.

Question: What are the requirements for the top sub-task? What should the player do first?
Answer:
The top sub-task is to gather more wood. The requirements for this task are:

1. Find a tree: Trees are the source of wood in the game. Currently, the player can see a tree 4 steps to their north-west.

What the player should do first:

1. Move towards the tree: The player needs to go 4 steps north-west to reach the tree. They can accomplish this by going west and north alternately or continuously going in one direction, followed by the other until they are in front of the tree.

Question: List top 5 actions the player should take and the requirement for each action. Choose ONLY from the list of all actions. Indicate their priority out of 5.
Answer:
1. Priority 1: Move West (Requirement: Flat ground left to the agent) - Move towards the tree that is 4 steps to the north-west.
2. Priority 2: Move North (Requirement: Flat ground above the agent) - Continue moving towards the tree, which is to the north-west.
3. Priority 3: Move West (Requirement: Flat ground left to the agent) - Keep moving towards the tree if it still is not directly in front of the player.
4. Priority 4: Move North (Requirement: Flat ground above the agent) - Continue moving towards the tree until it is directly in front of the player.
5. Priority 5: Do (Requirement: Facing creature or material; have necessary tool) - Use the wood_pickaxe to gather wood from the tree once the player is facing it.

Question: For each action in the list, are the requirements met?
Answer:
1. Priority 1: Move West (Requirement: Flat ground left to the agent) - Yes, the requirement is met as the player is currently facing grass to the west, which indicates flat ground.
2. Priority 2: Move North (Requirement: Flat ground above the agent) - Yes, the requirement is met as the player was able to move north in the previous step without any issues, indicating flat ground.
3. Priority 3: Move West (Requirement: Flat ground left to the agent) - Yes, the requirement is met since the player is facing grass to the west, indicating flat ground.
4. Priority 4: Move North (Requirement: Flat ground above the agent) -Yes, the requirement is met as the player previously moved north without any issues, indicating flat ground.
5. Priority 5: Do (Requirement: Facing creature or material; have necessary tool) - This requirement can only be determined once the player is in front of the tree. If the tree is directly in front of the player, and they have a wood_pickaxe in their inventory (which they do), then the requirement is met.

Question: Did the last player action succeed? Why?
Answer:
Yes, the last player action succeeded. The action "move_north" doesn't have any specific requirements other than having a flat ground above the agent, and nothing in the observation suggests any obstacle preventing the player from moving north. Therefore, the action was successful.

Question: Choose the best executable action from above.
Answer:
The best executable action from the list above is Priority 1: Move West. The requirement of flat ground left to the agent is met since the player is facing grass to the west, which indicates flat ground. This action brings the player closer to the tree, which is 4 steps to the north-west.



============Step: 94, Cumulative Reward: 0.0============
You took action move_west.

You see:
- stone 3 steps to your south-east
- path 2 steps to your north-east
- tree 4 steps to your north-west
- table 2 steps to your south-east
- cow 5 steps to your north-west

You face grass at your front (west).

Your status:
- health: 9/9
- food: 6/9
- drink: 5/9
- energy: 7/9

Your inventory:
- stone: 1
- coal: 2
- iron: 1
- wood_pickaxe: 1
- stone_pickaxe: 1
------------ QA ------------
Question: Action
Answer:
Move West

Question: List objects in the current observation. For each object, breifly answer what resource it provides and its requirement.
Answer:
1. Stone: Provides stone resource, required for crafting stone tools, weapons, and building structures. Requirement: wood_pickaxe or better to collect.

2. Path: Does not provide direct resources, but it can help navigate the terrain more easily.

3. Tree: Provides wood resource, required for crafting wooden tools, weapons, and other structures. Requirement: any pickaxe to collect.

4. Table: Provides a surface for crafting tools and weapons. Requirement: wood in inventory to place the table.

5. Cow: Provides food resource when defeated, which helps restore health. Requirement: any weapon/ tool to chase and defeat the cow.

Question: What was the last action taken by the player?
Answer:
The last action taken by the player was move_west.

Question: For each object in the list, are the requirements met for interaction?
Answer:
1. Stone: Yes, the requirements are met. You have a wood_pickaxe and a stone_pickaxe in your inventory, which can be used to collect stone.

2. Path: Yes, the requirements are met. You can walk on the path to navigate the terrain as there are no restrictions.

3. Tree: Yes, the requirements are met. You have a wood_pickaxe and a stone_pickaxe in your inventory, which can be used to collect wood from the tree.

4. Table: Yes, the requirements are met. The table is already placed and can be used to craft tools and weapons, as long as you have the necessary materials in your inventory.

5. Cow: Yes, the requirements are met. Although you don't have a sword in your inventory, you can still chase and defeat the cow using your pickaxes (wood_pickaxe or stone_pickaxe).

Question: List top 3 sub-tasks the player should follow. Indicate their priority out of 5.
Answer:
1. Priority 5/5: Drink water to restore thirst, as your current thirst level is at 5/9. The player should find a water source like a lake or river and interact with it to quench thirst.

2. Priority 4/5: Collect wood from the tree, as your inventory lacks wood. Wood is essential for crafting wooden tools, weapons, and structures. It can be collected by interacting with the tree using a wood_pickaxe or stone_pickaxe.

3. Priority 3/5: Chase and defeat the cow to obtain food and restore health. Although your health is at its maximum (9/9), food is needed to maintain it, and your current food level is 6/9. You can use your wood_pickaxe or stone_pickaxe to chase and defeat the cow.

Question: What are the requirements for the top sub-task? What should the player do first?
Answer:
Requirements for the top sub-task (Drink water to restore thirst):
1. Locate a water source such as a lake or river in the game environment.
2. Have enough energy to move towards the water source.

The player should first explore the surrounding area to locate a water source like a lake or river. When found, the player should move towards it and interact with it to drink water and restore thirst.

Question: List top 5 actions the player should take and the requirement for each action. Choose ONLY from the list of all actions. Indicate their priority out of 5.
Answer:
1. Priority 1 - Move West: Requirement: Flat ground left to the agent. (This is based on the most recent in-game observation where the player is facing grass (west). Going west would allow the player to explore more and potentially find a water source.)

2. Priority 2 - Move North: Requirement: Flat ground above of the agent. (Exploring north could increase the chances of finding a water source, fulfilling the drink objective.)

3. Priority 3 - Move East: Requirement: Flat ground right to the agent. (By exploring east, the player may come across a water source or other needed resources.)

4. Priority 4 - Move South: Requirement: Flat ground below the agent. (Going south would further the exploration and could yield a water source or other valuable assets.)

5. Priority 5 - Do (Drink Water): Requirement: Facing a water source like a lake or river at the agent's front. (This action would be taken once the player has found a water source and faced it. Drinking water would restore the player's thirst levels.)

Question: For each action in the list, are the requirements met?
Answer:
1. Move West: Yes, the requirement is met. The player is currently facing grass (west) which indicates flat ground left to the agent.

2. Move North: Based on the player observation, there is no information about what's directly north, so it's unclear if there is flat ground above the agent. However, there is a path 2 steps to the north-east and a tree 4 steps to the north-west, which may indicate that moving north could be possible.

3. Move East: Yes, the requirement is met. There is a path 2 steps to the agent's north-east and flat ground is right to the agent.

4. Move South: Based on the player observation, there is no information about what's directly south, so it's unclear if there is flat ground below the agent. However, there is a stone 3 steps to the agent's south-east and a table 2 steps to the agent's south-east, which may indicate that moving south could be possible.

5. Do (Drink Water): No, the requirement is not met as the player is currently facing grass (west) and there is no water source at the agent's front that the player can drink from. The player must first locate a water source and face it to fulfill this requirement.

Question: Did the last player action succeed? Why?
Answer:
Yes, the last player action (move_west) succeeded because there was flat ground to the west of the player, allowing them to move in that direction without any obstacles.

Question: Choose the best executable action from above.
Answer:
The best executable action from above is Move West because the requirement is met, as the player is currently facing grass (west) which indicates flat ground left to the agent. This action will allow the player to explore and potentially find valuable resources such as a water source.



============Step: 95, Cumulative Reward: 0.0============
You took action move_west.

You see:
- stone 3 steps to your south-east
- path 2 steps to your north-east
- tree 4 steps to your north-west
- table 2 steps to your south-east
- cow 5 steps to your north-west

You face grass at your front (west).

Your status:
- health: 9/9
- food: 6/9
- drink: 5/9
- energy: 8/9

Your inventory:
- stone: 1
- coal: 2
- iron: 1
- wood_pickaxe: 1
- stone_pickaxe: 1
------------ QA ------------
Question: Action
Answer:
Move East

Question: List objects in the current observation. For each object, breifly answer what resource it provides and its requirement.
Answer:
1. Stone - Provides resource: stone; Requirement: stone_pickaxe to collect.
2. Path - Provides resource: none; Requirement: none.
3. Tree - Provides resource: wood; Requirement: wood_pickaxe to collect.
4. Table - Provides resource: none; Requirement: used for crafting tools and weapons.
5. Cow - Provides resource: food and leather; Requirement: a weapon to defeat (e.g., wood, stone, or iron sword).
6. Grass - Provides resource: none; Requirement: none.

Question: What was the last action taken by the player?
Answer:
The last action taken by the player was move_west.

Question: For each object in the list, are the requirements met for interaction?
Answer:
1. Stone - Yes, the requirements are met as you have a stone_pickaxe in your inventory.
2. Path - Yes, there are no requirements for interacting with the path.
3. Tree - Yes, the requirements are met, as you have a wood_pickaxe in your inventory.
4. Table - Yes, the requirements are met, as you have crafted various tools already, and the table is available for further crafting.
5. Cow - Partially met, you don't currently have a weapon (wood, stone, or iron sword) in your inventory, but you can craft one using the table.
6. Grass - Yes, there are no requirements for interacting with the grass.

Question: List top 3 sub-tasks the player should follow. Indicate their priority out of 5.
Answer:
1. Craft a weapon (e.g., wood, stone, or iron sword) to defeat the cow and obtain food - Priority: 5/5, as your food level is 6/9, and you need to address this to maintain your health.

2. Collect wood from the tree using your wood_pickaxe to craft more items and tools - Priority: 4/5, considering the tree's proximity, it would be good to gather more wood resources for future crafting.

3. Drink water from a nearby water source to increase your drink level - Priority: 3/5, as your drink level is 5/9, addressing this soon will ensure your survival in the game.

Question: What are the requirements for the top sub-task? What should the player do first?
Answer:
The top sub-task is crafting a weapon (wood, stone, or iron sword) to defeat the cow and obtain food. 

Requirements for crafting a weapon are:
1. Wood Sword: Nearby table and wood in inventory.
2. Stone Sword: Nearby table, wood, and stone in inventory.
3. Iron Sword: Nearby table, furnace, wood, coal, and iron in inventory.

Based on your inventory and available resources, you currently have enough resources to craft a Stone Sword (you have a wood, stone, and a nearby table). 

What the player should do first: 
1. Move to the nearby table by going east two steps.
2. Craft the Stone Sword at the table (since you already have wood and stone in your inventory).

Question: List top 5 actions the player should take and the requirement for each action. Choose ONLY from the list of all actions. Indicate their priority out of 5.
Answer:
1. Priority 1: Move East - Requirement: Flat ground right to the agent. (Move towards the table)
2. Priority 2: Move East - Requirement: Flat ground right to the agent. (Move to the table position)
3. Priority 3: Make Stone Sword - Requirements: Nearby table; wood, stone in inventory. (Craft the Stone Sword)
4. Priority 4: Move West - Requirement: Flat ground left to the agent. (Move towards the cow)
5. Priority 5: Move North - Requirement: Flat ground above of the agent. (To reach the cow to defeat and get food)

Question: For each action in the list, are the requirements met?
Answer:
1. Priority 1: Move East - Requirement: Flat ground right to the agent. (Move towards the table)
   - Yes, the requirement is met.

2. Priority 2: Move East - Requirement: Flat ground right to the agent. (Move to the table position)
   - Yes, the requirement is met (assuming the ground remains flat after taking the first move east action).

3. Priority 3: Make Stone Sword - Requirements: Nearby table; wood, stone in inventory. (Craft the Stone Sword)
   - Yes, the requirements are met (after moving to the table).

4. Priority 4: Move West - Requirement: Flat ground left to the agent. (Move towards the cow)
   - Yes, the requirement is met.

5. Priority 5: Move North - Requirement: Flat ground above of the agent. (To reach the cow to defeat and get food)
   - The requirement will be met if the ground remains flat after moving towards the cow.

Question: Did the last player action succeed? Why?
Answer:
Yes, the last player action, move_west, succeeded. It succeeded because there was flat ground (grass) to the west of the player, allowing them to move in that direction.

Question: Choose the best executable action from above.
Answer:
The best executable action from above is Priority 1: Move East - Moving towards the table, as the requirement of flat ground right to the agent is met.



============Step: 96, Cumulative Reward: 0.0============
You took action move_east.

You see:
- stone 3 steps to your south-east
- path 2 steps to your north-east
- tree 4 steps to your north-west
- table 2 steps to your south-east
- cow 5 steps to your north-west

You face grass at your front (west).

Your status:
- health: 9/9
- food: 6/9
- drink: 5/9
- energy: 8/9

Your inventory:
- stone: 1
- coal: 2
- iron: 1
- wood_pickaxe: 1
- stone_pickaxe: 1
------------ QA ------------
Question: Action
Answer:
Move West

Question: List objects in the current observation. For each object, breifly answer what resource it provides and its requirement.
Answer:
1. Stone: Provides stone as a resource, which can be used for crafting tools, weapons, and building structures. Requirement: Stone pickaxe or higher to collect.

2. Path: Provides ease of movement and navigation. No direct resources provided. No requirements.

3. Tree: Provides wood as a resource for crafting tools, weapons, and building structures. Requirement: Wood or higher level of pickaxe to collect.

4. Table: Provides a surface for crafting various tools and weapons using the resources collected. Requirement: Wood to build and place the table.

5. Cow: Provides food as a resource when defeated, helping to replenish the hunger bar. Requirement: Defeat the cow using a weapon (e.g., wood sword, stone sword, iron sword).

Question: What was the last action taken by the player?
Answer:
The last action taken by the player was move_east.

Question: For each object in the list, are the requirements met for interaction?
Answer:
1. Stone: Yes, the requirements are met for interaction because you have a stone_pickaxe in your inventory, which is suitable for collecting stone.

2. Path: Yes, the requirements are met for interaction as there are no specific requirements for using the path. You can simply move along it for easier navigation.

3. Tree: Yes, the requirements are met for interaction because you have a wood_pickaxe in your inventory, which is suitable for collecting wood from the tree.

4. Table: Yes, the requirements are met for interaction because the table is already placed 2 steps to your south-east. You can interact with it to craft items given you have the necessary resources.

5. Cow: Yes, the requirements are met for interaction, although you don't have a specific weapon in your inventory. You can try to defeat the cow using your pickaxes or craft a weapon such as a wood sword, stone sword, or iron sword first for a more efficient interaction.

Question: List top 3 sub-tasks the player should follow. Indicate their priority out of 5.
Answer:
1. Priority 5: Replenish thirst by finding a source of water (lake or river) and interacting with it to collect a drink, as your current drink level is 5/9.

2. Priority 4: Increase your food supply by defeating the cow 5 steps to your north-west, which will provide you with more food to restore your hunger bar, currently at 6/9.

3. Priority 3: Craft a weapon (e.g., wood sword or stone sword) at the table 2 steps to your south-east to improve your ability to defend yourself and defeat creatures like the cow more efficiently.

Question: What are the requirements for the top sub-task? What should the player do first?
Answer:
Requirements for the top sub-task (Replenish thirst):
- Find a source of water (lake or river)
- Interact with the water source to collect a drink

What should the player do first:
The player should begin searching for a water source by observing their surroundings and then move toward it. If the player already knows the location of a nearby water source, such as a lake or river, they should start moving in that direction. Exploring the environment will help the player locate a suitable water source to replenish their thirst.

Question: List top 5 actions the player should take and the requirement for each action. Choose ONLY from the list of all actions. Indicate their priority out of 5.
Answer:
1. Priority 1: Move West (Requirement: Flat ground left to the agent) - The player currently faces grass to the west, and moving can help them explore the environment and find a water source.

2. Priority 2: Move North (Requirement: Flat ground above the agent) - If west is not the right direction towards a water source, moving north can help the player explore the environment and find a water source.

3. Priority 3: Move South (Requirement: Flat ground below the agent) - If west and north are not suitable directions, explore the environment by moving south to find water.

4. Priority 4: Move East (Requirement: Flat ground right to the agent) - Returning to the starting point and moving east will further help the player explore the environment to find a water source.

5. Priority 5: Do (Requirement: Facing creature or material; have necessary tool) - Once the player finds a water source, they should interact with it to collect a drink to replenish their thirst.

Question: For each action in the list, are the requirements met?
Answer:
1. Priority 1: Move West - Yes, the requirement is met. The player is currently facing grass to the west, which is flat ground.

2. Priority 2: Move North - Yes, the requirement is met. There is flat ground above the agent as no obstacle is mentioned in the north direction.

3. Priority 3: Move South - Yes, the requirement is met. There is flat ground below the agent since no obstacle is mentioned in the south direction.

4. Priority 4: Move East - Yes, the requirement is met. The player has just moved from east to west, so there must be flat ground to the east.

5. Priority 5: Do - No, the requirement is not met. The player has not discovered a water source to face and interact with yet.

Question: Did the last player action succeed? Why?
Answer:
Yes, the last player action (move_east) succeeded because there was flat ground to the east, which allowed the player to move in that direction.

Question: Choose the best executable action from above.
Answer:
The best executable action from above is Priority 1: Move West. The requirement is met, as there is flat ground to the agent's left (west), and it will help the player explore the environment in search of a water source.



============Step: 97, Cumulative Reward: 0.0============
You took action move_west.

You see:
- stone 3 steps to your south-east
- path 2 steps to your north-east
- tree 4 steps to your north-west
- table 2 steps to your south-east

You face grass at your front (west).

Your status:
- health: 9/9
- food: 6/9
- drink: 5/9
- energy: 8/9

Your inventory:
- stone: 1
- coal: 2
- iron: 1
- wood_pickaxe: 1
- stone_pickaxe: 1
------------ QA ------------
Question: Action
Answer:
Move East

Question: List objects in the current observation. For each object, breifly answer what resource it provides and its requirement.
Answer:
1. Stone: Provides stone material, which can be used to craft tools and weapons; requires a wood_pickaxe or a stone_pickaxe to collect.
2. Path: Provides a clear walkable area without any resources.
3. Tree: Provides wood material, which can be used for crafting and building; requires no particular tool to collect, although using an axe may speed up the process.
4. Table: Provides a surface to craft tools, weapons, and other items; requires wood to build and place.
5. Grass: Acts as a simple terrain element, no specific resources or requirements.

Question: What was the last action taken by the player?
Answer:
The last action taken by the player was "move_west."

Question: For each object in the list, are the requirements met for interaction?
Answer:
1. Stone: Yes, the requirements are met because you have a wood_pickaxe and a stone_pickaxe in your inventory.
2. Path: Yes, there are no specific requirements to interact with a path. You can walk on it freely.
3. Tree: Yes, there are no specific requirements to interact with a tree and collect wood from it. Although using an axe may speed up the process, it is not a requirement.
4. Table: Yes, the requirements are met because the table is already built and placed. You can use it for crafting without any additional requirements.
5. Grass: Yes, there are no specific requirements to interact with grass as it is just a terrain element. You can walk on it freely.

Question: List top 3 sub-tasks the player should follow. Indicate their priority out of 5.
Answer:
1. Priority 4/5: Find water and drink to restore thirst (drink: 5/9). Drinking water is essential to maintain your in-game survival.
2. Priority 3/5: Find food or hunt a cow (5 steps to your north-west) to restore food level (food: 6/9). Your food level is not high, and improving it will help ensure your survival.
3. Priority 2/5: Collect resources (wood, stone, iron) and craft better tools or weapons, like an iron pickaxe or sword. It will offer better efficiency and protection from monsters in the game.

Question: What are the requirements for the top sub-task? What should the player do first?
Answer:
Top Sub-task: Find water and drink to restore thirst.

Requirements:
- Locate a water source such as a lake, river, or stream.
- Move towards the water source.
- Interact with the water source to drink.

What should the player do first:
The player should start by locating a water source in their immediate surroundings by exploring their environment. Once a water source is found, they should move towards it and interact with it to drink and restore their thirst level.

Question: List top 5 actions the player should take and the requirement for each action. Choose ONLY from the list of all actions. Indicate their priority out of 5.
Answer:
1. Priority 1: Move East (Requirement: Flat ground right to the agent)
- Move in the east direction to get closer to the path and table.

2. Priority 2: Move North (Requirement: Flat ground above of the agent)
- Move in the north direction searching for a water source to restore thirst and also to get closer to the tree for collecting wood if needed.

3. Priority 3: Move West (Requirement: Flat ground left to the agent)
- Move in the west direction to explore the environment and potentially find a water source.

4. Priority 4: Move South (Requirement: Flat ground below the agent)
- Move in the south direction to explore the environment and potentially find a water source.

5. Priority 5: Do (Requirement: Facing creature or material; have a necessary tool)
- Interact with the water source if found while exploring, using the necessary tool (if any) to drink and restore thirst.

Question: For each action in the list, are the requirements met?
Answer:
1. Move East: Yes, the requirement for moving east (flat ground right to the agent) is met, as the player can see a path 2 steps to their north-east.

2. Move North: Yes, the requirement for moving north (flat ground above the agent) is met, as the player sees grass at their front (west), suggesting flat ground is available for movement in all cardinal directions.

3. Move West: Yes, the requirement for moving west (flat ground left to the agent) is met, as seen in Player Observation Step 194, where the agent took action move_west.

4. Move South: Yes, the requirement for moving south (flat ground below the agent) is met, as the player sees grass at their front (west), suggesting flat ground is available for movement in all cardinal directions.

5. Do: Currently, the requirement for "Do" action (facing creature or material; have necessary tool) is not met, as the player is only facing grass, and there is no mention of a nearby water source or the tool needed to interact with it. The player will need to move and explore to find the water source before this requirement is met.

Question: Did the last player action succeed? Why?
Answer:
Yes, the last player action "move_west" succeeded. The player was able to move in the west direction since there was flat ground (grass) in front of them, allowing them to change their position.

Question: Choose the best executable action from above.
Answer:
The best executable action from the list above is "Move East" as it will bring the player closer to the path and the table, and it meets the requirement of having flat ground to the right of the agent.



============Step: 98, Cumulative Reward: 0.0============
You took action move_east.

You see:
- stone 3 steps to your south-east
- path 2 steps to your north-east
- tree 4 steps to your north-west
- table 2 steps to your south-east

You face grass at your front (west).

Your status:
- health: 9/9
- food: 6/9
- drink: 5/9
- energy: 8/9

Your inventory:
- stone: 1
- coal: 2
- iron: 1
- wood_pickaxe: 1
- stone_pickaxe: 1
------------ QA ------------
Question: Action
Answer:
Move North

Question: List objects in the current observation. For each object, breifly answer what resource it provides and its requirement.
Answer:
1. Stone: 
   - Resource it provides: Stone material for crafting tools and building structures.
   - Requirement: Use a wood_pickaxe or better to collect the stone.

2. Path:
   - Resource it provides: Faster and easier movement across the terrain.
   - Requirement: No specific requirement, just walk on it.

3. Tree:
   - Resource it provides: Wood material for crafting tools and building structures.
   - Requirement: Use any type of pickaxe or axe to collect the wood.

4. Table:
   - Resource it provides: A place to craft items such as tools and weapons.
   - Requirement: Stand near the table to access the crafting interface.

5. Grass:
   - Resource it provides: Can be collected to create pathways or other landscape features.
   - Requirement: Use any type of pickaxe or shovel to collect the grass.

Question: What was the last action taken by the player?
Answer:
The last action taken by the player was move_east.

Question: For each object in the list, are the requirements met for interaction?
Answer:
1. Stone:
   - Requirements met: Yes, you have a wood_pickaxe and a stone_pickaxe in your inventory. You can use either to collect the stone.

2. Path:
   - Requirements met: Yes, you can walk on the path as there are no specific requirements for interaction.

3. Tree:
   - Requirements met: Yes, you have a wood_pickaxe and a stone_pickaxe in your inventory. While using an axe would be more efficient, you can still use either pickaxe to collect the wood from the tree.

4. Table:
   - Requirements met: Yes, you can interact with the table to craft items as long as you stand near it.

5. Grass:
   - Requirements met: Yes, you have a wood_pickaxe and a stone_pickaxe in your inventory. While using a shovel would be more efficient, you can still use either pickaxe to collect the grass.

Question: List top 3 sub-tasks the player should follow. Indicate their priority out of 5.
Answer:
1. Increase food and drink levels (Priority: 5)
   - Sub-task: Find and eat food (e.g. collect fruits, vegetables, or hunt animals) to increase your food level.
   - Sub-task: Find a water source (e.g. lake, river) to drink from and increase your drink level.

2. Collect resources for crafting and building (Priority: 4)
   - Sub-task: Collect stone and wood from nearby materials to improve and create tools, weapons, and structures.
   - Sub-task: Mine coal and iron if available to upgrade your tools and weapons.

3. Ensure safety and energy (Priority: 3)
   - Sub-task: Build or find a shelter to protect yourself from potential threats and rest.
   - Sub-task: Rest (e.g. sleep) when your energy level is low to recover and maintain high performance.

Question: What are the requirements for the top sub-task? What should the player do first?
Answer:
Requirements for the top sub-task (Increase food and drink levels):
1. Find sources of food such as fruits, vegetables, or animals (e.g., cows) for hunting.
2. Locate a water source such as a lake or a river for drinking.

The player should first try to find food since it has a slightly lower level (6/9) compared to the drink level (5/9). Look for plants with fruits or vegetables nearby, or hunt animals like cows to get food items, and then consume them to increase the food level.

Question: List top 5 actions the player should take and the requirement for each action. Choose ONLY from the list of all actions. Indicate their priority out of 5.
Answer:
Priority 1: Action - Move North
Requirement: Flat ground above the agent
Rationale: Moving towards the tree 4 steps to the north-west might lead to finding fruits, which can be a source of food.

Priority 2: Action - Do
Requirement: Facing creature or material; have necessary tool
Rationale: If a fruit source is located and the player is facing it, they can collect the fruits using the "Do" action.

Priority 3: Action - Eat Plant
Requirement: Place Plant
Rationale: After collecting fruits, the player can consume them to increase their food level.

Priority 4: Action - Move East or Move South
Requirement: Flat ground right to the agent (for Move East) and Flat ground below the agent (for Move South)
Rationale: The player should explore towards the east or south, as the path two steps to the north-east may lead to a water source.

Priority 5: Action - Collect Drink
Requirement: No requirements
Rationale: If a water source (like a lake) is found, the player should collect the drink to increase their drink level.

Question: For each action in the list, are the requirements met?
Answer:
Priority 1: Action - Move North
Requirements are met: The requirement is flat ground above the agent, and there is no indication that there is an obstruction.

Priority 2: Action - Do
Requirements are NOT met: The player is currently facing grass, not a creature or material.

Priority 3: Action - Eat Plant
Requirements are NOT met: The player has not yet placed a plant, so they cannot eat it.

Priority 4: Action - Move East or Move South
Requirements are met: Flat ground right to the agent and flat ground below the agent. The player took action move_east in the last observation, indicating the path is clear.

Priority 5: Action - Collect Drink
Requirements are NOT met: The player has not yet found a water source to collect a drink.

Question: Did the last player action succeed? Why?
Answer:
Yes, the last player action (move_east) succeeded. The action succeeded because there was flat ground (grass) to the east of the player, allowing them to move in that direction.

Question: Choose the best executable action from above.
Answer:
The best executable action from the list above is Priority 1: Action - Move North, as the requirements are met and it allows the player to move closer to the tree, potentially finding fruits as a food source.



============Step: 99, Cumulative Reward: 0.0============
You took action move_north.

You see:
- stone 3 steps to your south-east
- path 2 steps to your north-east
- tree 4 steps to your north-west
- table 2 steps to your south-east

You face grass at your front (west).

Your status:
- health: 9/9
- food: 6/9
- drink: 5/9
- energy: 8/9

Your inventory:
- stone: 1
- coal: 2
- iron: 1
- wood_pickaxe: 1
- stone_pickaxe: 1
------------ QA ------------
Question: Action
Answer:
Move East

Question: List objects in the current observation. For each object, breifly answer what resource it provides and its requirement.
Answer:
- Stone: Provides stone resource, requires wood_pickaxe or stone_pickaxe to collect.
- Path: Provides a clear passage to travel on, no requirements to interact with.
- Tree: Provides wood resource, requires a wood_pickaxe or stone_pickaxe to collect.
- Table: Allows crafting of tools and weapons; requires wood to create, and items such as wood, stone, coal, and iron to craft tools and weapons.
- Grass: A basic ground type; no specific resources provided or requirements to interact with.

Your inventory resources and their uses:
- Stone: Used to craft stone tools, weapons, and furnaces.
- Coal: Used as fuel to smelt iron in a furnace.
- Iron: Used to craft iron tools and weapons.
- Wood_pickaxe: Used to collect wood and stone resources.
- Stone_pickaxe: Used to collect wood, stone, and iron resources.

Question: What was the last action taken by the player?
Answer:
The last action taken by the player was "move_north."

Question: For each object in the list, are the requirements met for interaction?
Answer:
Based on your inventory and the available objects:

- Stone: Yes, the requirements are met since you have a wood_pickaxe and a stone_pickaxe to collect stone.
- Path: Yes, the requirements are met as there are no specific requirements to interact with the path, you can simply walk on it.
- Tree: Yes, the requirements are met since you have a wood_pickaxe and a stone_pickaxe to collect wood.
- Table: As the table is already placed, there are no additional requirements to interact with it. You can use it to craft tools and weapons if you have the necessary resources in your inventory.
- Grass: There are no specific requirements to interact with grass, so the requirements are met.

In summary, the requirements are met for interaction with all objects in the current observation.

Question: List top 3 sub-tasks the player should follow. Indicate their priority out of 5.
Answer:
1. Priority 4/5: Collect more resources. Since your inventory is relatively low on some resources, consider collecting more stones, wood, and iron. This will be helpful for crafting additional tools or weapons in the future.

2. Priority 5/5: Drink water to replenish your thirst. Your drink level is currently 5/9, so finding water in a nearby lake or a water source to drink should be a priority to prevent dehydration.

3. Priority 3/5: Craft better tools or weapons. If you collect enough resources, especially iron, you might want to craft an iron pickaxe and an iron sword for better efficiency and combat. For this, you will need more iron and a furnace to smelt it.

Question: What are the requirements for the top sub-task? What should the player do first?
Answer:
For the top sub-task (Collect more resources):

Requirements:
1. Locate nearby resources such as stones, wood, or iron deposits.
2. Ensure you have the correct tools to collect the resources (wood_pickaxe for stone, stone_pickaxe for iron).
3. Inventory space to hold the collected resources.

What should the player do first:
Find and approach nearby resource deposits, such as stones or wood from a tree. If you locate iron deposits, use your stone_pickaxe to collect it. If it's stone you're after, use your wood_pickaxe for efficient collection. Make sure to keep an eye on your inventory to ensure you have space for the collected resources.

Question: List top 5 actions the player should take and the requirement for each action. Choose ONLY from the list of all actions. Indicate their priority out of 5.
Answer:
1. Priority 1: Move East (Action 2) - Requirement: Flat ground right to the agent.
Move towards the table to access crafting capabilities.

2. Priority 2: Move South-East (Action not listed) or combine Move South (Action 4) and Move East (Action 2) - Requirement: Flat ground in the desired direction(s).
Approach the stone resource to collect more stones.

3. Priority 3: Do (Action 5) - Requirement: Facing stone resource and have a necessary tool (wood_pickaxe).
Collect more stone using the wood_pickaxe.

4. Priority 4: Move North-East (Action not listed) or combine Move North (Action 3) and Move East (Action 2) - Requirement: Flat ground in the desired direction(s).
Head towards the path to explore and find more resources.

5. Priority 5: Do (Action 5) - Requirement: Facing tree resource and have a necessary tool (wood_pickaxe).
Collect more wood from the tree using the wood_pickaxe.

Question: For each action in the list, are the requirements met?
Answer:
1. Move East (Priority 1): Yes, the requirement is met. The agent is on flat ground and can move to the right (east).

2. Move South-East (Priority 2) or combine Move South and Move East: Yes, the requirement is met. The agent can move south and east separately to reach the stone resource since the game does not have a direct south-east action.

3. Do (Priority 3): Yes, the requirement is met. When the agent gets to the stone resource and stands in front of it, the agent has the wood_pickaxe tool, which is needed to collect stone.

4. Move North-East (Priority 4) or combine Move North and Move East: Yes, the requirement is met. The agent can move north and east separately to reach the path since the game does not have a direct north-east action.

5. Do (Priority 5): Partially. The agent is not currently facing the tree, but upon moving and adjusting its position, the agent will be able to fulfill this requirement. The agent has a wood_pickaxe tool, which is needed to collect wood.

Question: Did the last player action succeed? Why?
Answer:
Yes, the last player action succeeded. The player was able to move north because there was flat ground above the agent, allowing them to move in that direction without any obstacles.

Question: Choose the best executable action from above.
Answer:
The best executable action from the list above is Priority 1: Move East (Action 2). The requirement for this action (flat ground right to the agent) is met, and it brings the player closer to the table, allowing them to access crafting capabilities.



============Step: 100, Cumulative Reward: 0.0============
You took action move_east.

You see:
- stone 3 steps to your south-east
- path 2 steps to your north-east
- tree 4 steps to your north-west
- table 2 steps to your south-east

You face grass at your front (west).

Your status:
- health: 9/9
- food: 6/9
- drink: 5/9
- energy: 8/9

Your inventory:
- stone: 1
- coal: 2
- iron: 1
- wood_pickaxe: 1
- stone_pickaxe: 1
------------ QA ------------
Question: Action
Answer:
Move South

Question: List objects in the current observation. For each object, breifly answer what resource it provides and its requirement.
Answer:
1. Stone: Provides stone resource, requires a wood_pickaxe or better to collect.
2. Path: Provides a flat area to walk on, no specific requirements to interact with.
3. Tree: Provides wood resource, requires no specific tool to collect, but using a tool like an axe is more efficient.
4. Table: Provides a crafting area to create tools and other items, requires wood to be placed and built.
5. Grass: No specific resource, serves as a flat surface for movement and building.

Question: What was the last action taken by the player?
Answer:
The last action taken by the player was "move_east."

Question: For each object in the list, are the requirements met for interaction?
Answer:
1. Stone: Yes, the requirements are met, as you have a wood_pickaxe and a stone_pickaxe in your inventory.
2. Path: Yes, the requirements are met, as no specific requirements are needed to interact with the path.
3. Tree: Yes, the requirements are met, although you don't have an axe, you can still collect wood without a specific tool.
4. Table: Yes, the requirements are met, as it is already placed nearby and can be used for crafting purposes.
5. Grass: Yes, the requirements are met, as no specific requirements are needed to interact with grass.

Question: List top 3 sub-tasks the player should follow. Indicate their priority out of 5.
Answer:
1. Collect more resources (Priority: 5) - Gather wood, stone, and iron to craft better tools and build better constructions. This is a high priority for progress in the game.

2. Increase food and drink levels (Priority: 4) - Find food sources like animals and plants to raise your food and drink meters. This is important to maintain good health and continue playing.

3. Craft additional tools or weapons if needed (Priority: 3) - Using the table, create any necessary tools or weapons to defend yourself and gather resources more efficiently. This has a medium priority and should be pursued after securing essential resources and food.

Question: What are the requirements for the top sub-task? What should the player do first?
Answer:
Top sub-task: Collect more resources

Requirements:
1. Use appropriate tools (wood_pickaxe, stone_pickaxe) to collect specific resources (stone, coal, iron) more efficiently.
2. Locate and approach the desired resources (wood, stone, iron).

What should the player do first:
1. Determine which resource is the highest priority based on the player's current needs and inventory.
2. Locate the nearest source of the desired resource.
3. Approach and use the appropriate tool (if available) to collect the resource efficiently. If the tool is not available, craft it using the materials in inventory, if possible.

Question: List top 5 actions the player should take and the requirement for each action. Choose ONLY from the list of all actions. Indicate their priority out of 5.
Answer:
1. Priority 1: Move South - Flat ground below the agent.
   Requirement: Orient yourself in the direction of the stone, which is 3 steps to your south-east.

2. Priority 2: Move East - Flat ground right to the agent. 
   Requirement: Continue moving toward the stone resource, which is currently 3 steps to your south-east.

3. Priority 3: Do - Facing creature or material; have the necessary tool. 
   Requirement: Once you are close enough and facing the stone, use the stone_pickaxe from your inventory to collect the stone efficiently.

4. Priority 4: Move North (toward the water source) - Flat ground above the agent.
   Requirement: Increase your drink value by finding a water source. The water source's location has to be determined.

5. Priority 5: Collect Drink - Requires finding a water source.
   Requirement: Once you have reached the water source, use the action Collect Drink to increase your drink value.

Question: For each action in the list, are the requirements met?
Answer:
1. Priority 1: Move South - The requirement is met as there is flat ground below the agent, and the stone is 3 steps to the south-east.

2. Priority 2: Move East - The requirement is met as there is flat ground to the right of the agent, and this move will bring the agent closer to the stone.

3. Priority 3: Do - The requirement is not yet met. The agent needs to move closer to the stone (south-east) and face it before using the stone_pickaxe to collect the stone.

4. Priority 4: Move North - The requirement is partially met. There is flat ground above the agent, but the exact location of the water source is still unknown. The agent should explore northward after collecting the stone.

5. Priority 5: Collect Drink - The requirement is not yet met. The agent has not located a water source to collect a drink from. The agent must find a water source before using the action Collect Drink.

Question: Did the last player action succeed? Why?
Answer:
Yes, the last player action of moving east succeeded. The move was successful because there was flat ground in the east direction that allowed the player to move.

Question: Choose the best executable action from above.
Answer:
The best executable action from the list above is Priority 1: Move South, as it has met the requirement of having flat ground below the agent, and it will bring the player closer to the stone resource.



============Step: 101, Cumulative Reward: 0.0============
You took action move_south.

You see:
- stone 3 steps to your south-east
- path 2 steps to your north-east
- tree 4 steps to your north-west
- table 2 steps to your south-east
- cow 6 steps to your north-west

You face grass at your front (west).

Your status:
- health: 9/9
- food: 6/9
- drink: 5/9
- energy: 8/9

Your inventory:
- stone: 1
- coal: 2
- iron: 1
- wood_pickaxe: 1
- stone_pickaxe: 1
------------ QA ------------
Question: Action
Answer:
Move South

Question: List objects in the current observation. For each object, breifly answer what resource it provides and its requirement.
Answer:
1. Stone: Provides stone as a resource, which can be used to craft tools and build structures. Requires a wood_pickaxe or stronger to collect.

2. Path: Provides no resources. Requires no requirements, it is just a terrain feature which may be easier to traverse.

3. Tree: Provides wood as a resource, which can be used to craft tools, weapons, and furniture. Requires a wood_pickaxe or stronger to collect.

4. Table: Provides a crafting surface. Requires wood to craft and place first.

5. Cow: Provides meat as a resource, which can be used as a food source to restore health. Requires no special tools to chase and defeat/eat.

6. Grass: Provides no resources. Requires no requirements, it is just a terrain feature.

7. Wood_pickaxe (in inventory): Provides the ability to collect resources like wood and stone. Requires wood to craft it.

8. Stone_pickaxe (in inventory): Provides the ability to collect resources like iron. Requires wood and stone to craft it.

9. Coal (in inventory): Provides fuel for the furnace to smelt iron. Requires a wood_pickaxe or stronger to collect.

10. Iron (in inventory): Provides the ability to craft iron tools and weapons. Can be collected with a stone_pickaxe or stronger.

Question: What was the last action taken by the player?
Answer:
The last action taken by the player was "move_south".

Question: For each object in the list, are the requirements met for interaction?
Answer:
1. Stone: Yes, the requirements are met as you have a wood_pickaxe and a stone_pickaxe in your inventory.

2. Path: Yes, there are no requirements to interact with the path, and it can be traversed freely.

3. Tree: Yes, the requirements are met as you have a wood_pickaxe in your inventory, which can be used to collect wood.

4. Table: Yes, the requirements are met as the table has already been placed, and you can use it for crafting.

5. Cow: Yes, there are no special tool requirements to interact with the cow, and it can be chased and defeated.

6. Grass: Yes, there are no requirements to interact with grass, and you can walk through it.

7. Wood_pickaxe (in inventory): Yes, the requirements were met when you crafted it, and it can now be used for collecting resources.

8. Stone_pickaxe (in inventory): Yes, the requirements were met when you crafted it, and it can now be used for collecting resources.

9. Coal (in inventory): Yes, the requirements were met when you collected it, and it can now be used as fuel in a furnace.

10. Iron (in inventory): Yes, the requirements were met when you collected it, and it can now be used to craft iron tools and weapons.

Question: List top 3 sub-tasks the player should follow. Indicate their priority out of 5.
Answer:
1. Priority 4/5: Collect more iron with the stone_pickaxe for crafting iron tools and weapons. Look for iron ore and continue collecting to strengthen your equipment.

2. Priority 5/5: Find a water source to drink, as your current drink level is 5/9. Drinking from a lake, river, or another water source will help restore your thirst level.

3. Priority 3/5: Chase and defeat/eat the cow 6 steps to your north-west to restore your health and food levels, as your current food level is 6/9. This will help maintain your character's overall health status.

Question: What are the requirements for the top sub-task? What should the player do first?
Answer:
Requirements for the top sub-task (Collect more iron):
1. Find an iron ore vein.
2. Use the stone_pickaxe in the inventory to collect the iron ore.

The player should first move towards the stone that is 3 steps to the south-east, as there is a chance it could be an iron ore vein. Once the player reaches the stone, they should use the stone_pickaxe to collect the iron ore if it is indeed an iron vein.

Question: List top 5 actions the player should take and the requirement for each action. Choose ONLY from the list of all actions. Indicate their priority out of 5.
Answer:
1. Priority 1 - Move South (Requirement: Flat ground below the agent)
2. Priority 2 - Move East (Requirement: Flat ground right to the agent)
3. Priority 3 - Move East (Requirement: Flat ground right to the agent)
4. Priority 4 - Do (Requirement: Facing iron or stone; have stone_pickaxe in inventory)
5. Priority 5 - Move North to find a water source for drinking (Requirement: Flat ground above the agent)

Question: For each action in the list, are the requirements met?
Answer:
1. Priority 1 - Move South: Yes, the requirement (flat ground below the agent) is met as the player just moved south and there is no indication of obstacles below them.
2. Priority 2 - Move East: Yes, the requirement (flat ground right to the agent) is met as there is no mention of obstacles to the player's right (east).
3. Priority 3 - Move East: Yes, the requirement (flat ground right to the agent) is met as there is no mention of obstacles to the east. However, after performing the second priority action (Move East), observe the surroundings to ensure that the path remains clear.
4. Priority 4 - Do: The requirement (facing iron or stone; have stone_pickaxe in inventory) is partially met. The player has a stone_pickaxe in their inventory, but they need to face iron or stone. After performing the previous actions (Move South and Move East), they will likely be facing the stone they saw at step 200.
5. Priority 5 - Move North to find a water source for drinking: The requirement (flat ground above the agent) is met at the current position, but the visibility of a water source is unknown. The player will need to move north and observe their surroundings to locate a water source.

Question: Did the last player action succeed? Why?
Answer:
Yes, the last player action "move_south" succeeded because there was flat ground to the south of the player, allowing them to move in that direction.

Question: Choose the best executable action from above.
Answer:
The best executable action from the list is Priority 1 - Move South, as its requirement (flat ground below the agent) is met, and it helps the player progress towards the stone they previously saw.



============Step: 102, Cumulative Reward: 0.0============
You took action move_south.

You see:
- stone 3 steps to your south-east
- path 2 steps to your north-east
- tree 4 steps to your north-west
- table 2 steps to your south-east

You face grass at your front (west).

Your status:
- health: 9/9
- food: 6/9
- drink: 5/9
- energy: 8/9

Your inventory:
- stone: 1
- coal: 2
- iron: 1
- wood_pickaxe: 1
- stone_pickaxe: 1
------------ QA ------------
Question: Action
Answer:
Move West

Question: List objects in the current observation. For each object, breifly answer what resource it provides and its requirement.
Answer:
1. Stone: Provides stone resource; requires a wood_pickaxe or better to mine it.
2. Path: Provides a flat ground for easy movement; no requirements.
3. Tree: Provides wood resource; requires a wood_pickaxe or better to chop it down.
4. Table: Provides a surface for crafting tools and weapons; requires wood to craft.
5. Grass: Provides a space for placing objects or walking; no requirements.

Note: The cow mentioned in step 202 is not present in step 204, so it is not included in the current observation list.

Question: What was the last action taken by the player?
Answer:
The last action taken by the player was move_south.

Question: For each object in the list, are the requirements met for interaction?
Answer:
1. Stone: Yes, the requirements are met for interaction, as you have both a wood_pickaxe and a stone_pickaxe in your inventory.
2. Path: Yes, the requirements are met for interaction, as there are no specific requirements for interacting with a path.
3. Tree: Yes, the requirements are met for interaction, as you have a wood_pickaxe in your inventory.
4. Table: Yes, the requirements are met for interaction, as it is already placed in the environment.
5. Grass: Yes, the requirements are met for interaction, as there are no specific requirements for interacting with grass.

Question: List top 3 sub-tasks the player should follow. Indicate their priority out of 5.
Answer:
1. Increase food level: Eat a cow or gather more food sources, like fruits (Priority: 4 out of 5).
2. Increase drink level: Find a lake or water source to drink from (Priority: 4 out of 5).
3. Collect more resources such as stone, iron, and wood: Collect more resources to create better tools, weapons, and structures (Priority: 3 out of 5).

Question: What are the requirements for the top sub-task? What should the player do first?
Answer:
Requirements for the top sub-task (Increase food level):
- Find a food source like a cow or gather fruits from trees.

The player should first:
- Locate a nearby cow or fruit-bearing tree and approach it.

Question: List top 5 actions the player should take and the requirement for each action. Choose ONLY from the list of all actions. Indicate their priority out of 5.
Answer:
1. Priority 1: Move West (Requirement: Flat ground to the west of the agent)
  - To approach the tree in the north-west direction to potentially collect fruits.

2. Priority 2: Move North (Requirement: Flat ground to the north of the agent)
   - After approaching the tree for fruit, head towards the cow in the north-west direction.

3. Priority 3: Do (Requirement: Facing cow)
   - To chase the cow and consume it for food restoration.

4. Priority 4: Move East (Requirement: Flat ground to the east of the agent)
  - To approach the table in the south-east direction.

5. Priority 5: Make Stone Sword (Requirement: Nearby table, wood and stone in inventory)
   - To craft a stone sword for better defense against monsters.

Question: For each action in the list, are the requirements met?
Answer:
1. Move West
- Requirement: Flat ground to the west of the agent.
- Is the requirement met? Yes, the player faces grass at their front (west), so the ground is flat.

2. Move North
- Requirement: Flat ground to the north of the agent.
- Is the requirement met? Yes, the player moved south in the last two steps, so flat ground is expected to the north.

3. Do (chase cow)
- Requirement: Facing cow
- Is the requirement met? No, currently, the player is not facing a cow and needs to approach it first.

4. Move East
- Requirement: Flat ground to the east of the agent.
- Is the requirement met? Yes, there's a path 2 steps to the player's north-east, so flat ground is expected in the east direction.

5. Make Stone Sword
- Requirement: Nearby table, wood and stone in inventory
- Is the requirement met? No, the player has only 1 stone in the inventory, and they need 2 stones to make a stone sword.

Question: Did the last player action succeed? Why?
Answer:
Yes, the last player action (move_south) succeeded. We can determine this because the player's observation in step 204 reflects the changes in the environment after moving south, and there are no indications of any issues or obstacles preventing the player from taking this action.

Question: Choose the best executable action from above.
Answer:
The best executable action from the list provided is to Move West. This action meets the requirement of having flat ground to the west of the agent, as the player faces grass at their front (west).



============Step: 103, Cumulative Reward: 0.0============
You took action move_west.

You see:
- stone 3 steps to your south-east
- path 2 steps to your north-east
- tree 4 steps to your north-west
- table 2 steps to your south-east

You face grass at your front (west).

Your status:
- health: 9/9
- food: 6/9
- drink: 5/9
- energy: 8/9

Your inventory:
- stone: 1
- coal: 2
- iron: 1
- wood_pickaxe: 1
- stone_pickaxe: 1
------------ QA ------------
Question: Action
Answer:
Move South

Question: List objects in the current observation. For each object, breifly answer what resource it provides and its requirement.
Answer:
1. Stone: Provides stone resource, requires a wood_pickaxe to collect.
2. Path: No resource provided, no requirement.
3. Tree: Provides wood resource, requires a wood_pickaxe to collect.
4. Table: Allows crafting of tools and weapons, requires wood in inventory for placement.
5. Grass: No resource provided, no requirement.

Question: What was the last action taken by the player?
Answer:
The last action taken by the player was move_west.

Question: For each object in the list, are the requirements met for interaction?
Answer:
1. Stone: Yes, the requirements of having a wood_pickaxe to collect stone are met, as the player has one in their inventory.
2. Path: There are no requirements for interacting with the path.
3. Tree: Yes, the requirements of having a wood_pickaxe to collect wood are met, as the player has one in their inventory.
4. Table: Yes, the player has a table nearby, so they can interact with it to craft tools and weapons if they have the necessary resources in their inventory.
5. Grass: There are no requirements for interacting with the grass.

Question: List top 3 sub-tasks the player should follow. Indicate their priority out of 5.
Answer:
1. Collect more resources (Priority: 4/5):
   Since the player's inventory is relatively low on stone and iron, it would be beneficial to collect more resources.

2. Restore food and drink levels (Priority: 5/5):
   The player's food and drink levels are 6/9 and 5/9, respectively. Restoring these levels should be a priority to maintain health and survival.

3. Craft better tools and weapons (Priority: 3/5):
   The player has a wood_pickaxe and a stone_pickaxe. Crafting better tools, such as an iron_pickaxe, and better weapons can improve their ability to collect resources and defend themselves against potential threats. However, this depends on the availability of required resources.

Question: What are the requirements for the top sub-task? What should the player do first?
Answer:
Top Sub-Task: Collect more resources

Requirements:
1. Locate and approach resource nodes (stone, iron, coal), while being mindful of any potential threats (monsters, falling hazards).
2. Use the appropriate tool (wood_pickaxe for stone and coal, stone_pickaxe for iron) for efficient and effective resource collection.
3. Monitor inventory space to avoid overfilling and being unable to carry additional resources.

First action:
The player should navigate to the stone they see 3 steps to their south-east and use their stone_pickaxe to collect it.

Question: List top 5 actions the player should take and the requirement for each action. Choose ONLY from the list of all actions. Indicate their priority out of 5.
Answer:
1. Priority 1: Move South-East (Action 3) - Requirements: Flat ground diagonal to the agent.
   Rationale: The stone resource is 3 steps to the south-east, and moving in that direction will get the player closer.

2. Priority 2: Move South (Action 4) - Requirements: Flat ground below the agent.
   Rationale: After moving south-east, the stone should be closer to the south. Moving south will directly bring the player to the stone resource.

3. Priority 3: Do (Action 5) - Requirements: Facing the stone resource (3 steps south-east); have the stone_pickaxe (already in the inventory).
   Rationale: To collect the stone resource, the player needs to face the stone and use the stone_pickaxe to mine it.

4. Priority 4: Move North-East (Action 2) - Requirements: Flat ground diagonal to the agent.
   Rationale: After collecting the stone, the player should navigate back to the table seen 2 steps south-east, and north-east will be the best way to return.

5. Priority 5: Place Furnace (Action 9) - Requirements: Stone in inventory (will be collected after mining the stone).
   Rationale: With the gathered stone, the player will be able to place a furnace near the table, allowing smelting and crafting of iron tools and weapons.

Question: For each action in the list, are the requirements met?
Answer:
1. Priority 1: Move South-East (Action 3) - Requirements: Flat ground diagonal to the agent.
   Yes, the requirements are met. The player can move south-east to get closer to the stone.

2. Priority 2: Move South (Action 4) - Requirements: Flat ground below the agent.
   No, the requirements are currently not met. After moving south-east, the player might have the flat ground below them, and then they can move south.

3. Priority 3: Do (Action 5) - Requirements: Facing the stone resource (3 steps south-east); have the stone_pickaxe (already in the inventory).
   No, the requirements are currently not met. The player needs to be near and facing the stone resource first, which will be possible after moving south-east and then south.

4. Priority 4: Move North-East (Action 2) - Requirements: Flat ground diagonal to the agent.
   No, the requirements are currently not met. The player first has to move south-east, south, and mine the stone before they can move northeast towards the table.

5. Priority 5: Place Furnace (Action 9) - Requirements: Stone in inventory (will be collected after mining the stone).
   No, the requirements are currently not met. The player needs to successfully mine the stone, and then they can place the furnace near the table.

Question: Did the last player action succeed? Why?
Answer:
Yes, the last player action (move_west) succeeded. The player was able to move west because there was flat ground (grass) in front of the player, allowing them to move in that direction.

Question: Choose the best executable action from above.
Answer:
The best executable action from above is Priority 1: Move South-East (Action 3) as it has its requirements met: there is flat ground diagonal to the agent. This action will bring the player closer to the stone resource.



============Step: 104, Cumulative Reward: 0.0============
You took action move_south.

You see:
- stone 3 steps to your south-east
- path 2 steps to your north-east
- tree 4 steps to your north-west
- table 2 steps to your south-east

You face grass at your front (west).

Your status:
- health: 9/9
- food: 6/9
- drink: 5/9
- energy: 8/9

Your inventory:
- stone: 1
- coal: 2
- iron: 1
- wood_pickaxe: 1
- stone_pickaxe: 1
------------ QA ------------
Question: Action
Answer:
Move East

Question: List objects in the current observation. For each object, breifly answer what resource it provides and its requirement.
Answer:
1. Stone: Provides stone resource for crafting tools, weapons, and building; requires a wood_pickaxe or better to collect it.
2. Path: Does not provide any resources, acts as an indication of direction or a route for navigation.
3. Tree: Provides wood resource for crafting tools, weapons, and building; can be collected by hand or using a wood_pickaxe or better.
4. Table: Does not provide resources but acts as necessary tool to craft tools and weapons; requires wood to be crafted.
5. Grass: Does not provide any notable resources or interact; acts as a visually distinguishable terrain.

In the current observation, the aforementioned objects can be utilized in the following manner:
- Stone: Collect using wood_pickaxe or stone_pickaxe as the user already has them in their inventory.
- Path: Use as guidance to navigate the environment.
- Tree: Collect wood with wood_pickaxe or stone_pickaxe.
- Table: Use it to craft tools and weapons.

Question: What was the last action taken by the player?
Answer:
The last action taken by the player was move_south.

Question: For each object in the list, are the requirements met for interaction?
Answer:
1. Stone: Yes, the requirements are met for interaction. The player has both wood_pickaxe and stone_pickaxe in their inventory, which are enough to collect stone.

2. Path: Yes, the requirements are met for interaction. The player can follow the path as there is no specific requirement to interact with a path.

3. Tree: Yes, the requirements are met for interaction. The player has a wood_pickaxe and stone_pickaxe, which can be used to collect wood from the tree.

4. Table: Yes, the requirements are met for interaction. The player can use the nearby table to craft tools and weapons as needed.

5. Grass: Yes, the requirements are met for interaction, but grass does not contribute any notable resources or interactions.

Question: List top 3 sub-tasks the player should follow. Indicate their priority out of 5.
Answer:
1. Priority 4/5: Increase drink level - The player's drink level is at 5/9, which should be increased to maintain a healthy status. The player should search for a water source (e.g., lake) and collect water to drink.

2. Priority 3.5/5: Increase food level - The player's food level is at 6/9, and it would be beneficial to find a food source like an animal (e.g., cow) or plant to increase their food level and maintain good health.

3. Priority 3/5: Collect more resources - The player should continue to gather more resources such as stone, wood, and iron to have a sufficient supply that would help in crafting more advanced tools and weapons or building shelters in the long run. This can be done by interacting with stones, trees, and potentially finding iron ores in the environment.

Question: What are the requirements for the top sub-task? What should the player do first?
Answer:
Requirements for the top sub-task (Increase drink level):

1. Locate a water source (e.g., lake, river, or pond) in the surroundings.
2. Approach the water source.
3. Collect drinkable water from it.

What should the player do first:

The player should first start exploring the environment to find a water source. They can move in any direction (north, east, south, or west) until they locate a water source nearby. Once located, they should approach the water source and collect drinkable water to increase their drink level.

Question: List top 5 actions the player should take and the requirement for each action. Choose ONLY from the list of all actions. Indicate their priority out of 5.
Answer:
1. Priority 1: Move East (Requirement: Flat ground right to the agent) - This will help the player explore and find a water source to increase their drink level.

2. Priority 2: Move North (Requirement: Flat ground above the agent) - This action will also aid in exploring the environment and finding a water source.

3. Priority 3: Move South (Requirement: Flat ground below the agent) - Similar to the previous actions, this will help the player continue searching for water.

4. Priority 4: Move West (Requirement: Flat ground left to the agent) - This action will help in exploration and locating water sources, as well.

5. Priority 5: Sleep (Requirement: Energy level is below maximum) - Once the energy level drops further, the player can use the sleep action to recover their energy, which will be necessary to maintain exploration and other required activities.

Question: For each action in the list, are the requirements met?
Answer:
1. Move East: Yes, the requirements are met as there is flat ground right to the agent.
2. Move North: Yes, the requirements are met as there is flat ground above the agent.
3. Move South: Yes, the requirements are met as there is flat ground below the agent (player just moved south in the last observation).
4. Move West: Yes, the requirements are met as there is flat ground left to the agent (player just moved west in the observation before the last one).
5. Sleep: No, the requirements are not met because the energy level is currently 8 out of 9, which is not below the maximum.

Question: Did the last player action succeed? Why?
Answer:
Yes, the last player action succeeded. The move_south action was successful because there was flat ground below the agent, allowing them to move in the south direction as per the requirements listed for that action.

Question: Choose the best executable action from above.
Answer:
The best executable action from above is Move East, as it meets the requirements and will help the player explore and potentially find a water source to increase their drink level.



============Step: 105, Cumulative Reward: 0.0============
You took action move_east.

You see:
- stone 3 steps to your south-east
- path 2 steps to your north-east
- tree 4 steps to your north-west
- table 2 steps to your south-east

You face grass at your front (west).

Your status:
- health: 9/9
- food: 6/9
- drink: 5/9
- energy: 8/9

Your inventory:
- stone: 1
- coal: 2
- iron: 1
- wood_pickaxe: 1
- stone_pickaxe: 1
------------ QA ------------
Question: Action
Answer:
Move West

Question: List objects in the current observation. For each object, breifly answer what resource it provides and its requirement.
Answer:
1. Stone: Provides stone resource when mined with a wood or stone pickaxe.
Requirement: Wood or stone pickaxe.

2. Path: Doesn't provide any resource, but helps with navigation and movement.
Requirement: None.

3. Tree: Provides wood resource when chopped with an ax or a pickaxe.
Requirement: Ax or pickaxe.

4. Table: Enables crafting of tools and weapons from collected resources.
Requirement: Must be placed on flat ground.

5. Grass: Doesn't provide any resource directly, but can be dug to collect dirt and clear the land for farming or building.
Requirement: Shovel (optional).

Question: What was the last action taken by the player?
Answer:
The last action taken by the player was move_east.

Question: For each object in the list, are the requirements met for interaction?
Answer:
1. Stone: Yes, the requirements are met as you have a wood_pickaxe and stone_pickaxe in your inventory, allowing you to mine the stone.

2. Path: Yes, the requirements are met, as no specific tools or items are needed to navigate and move on the path.

3. Tree: Yes, the requirements are met, as you have a wood_pickaxe in your inventory that can be used to chop the tree and collect wood.

4. Table: Yes, the requirements are met, as the table is already placed on flat ground, allowing you to use it for crafting tools and weapons from collected resources.

5. Grass: Yes, the requirements are met as you don't need any specific tools to interact with grass, although having a shovel would be more efficient for clearing land and collecting dirt.

Question: List top 3 sub-tasks the player should follow. Indicate their priority out of 5.
Answer:
1. Priority 4/5: Find and drink water to restore the drink level - Your drink level is currently 5/9, so finding water and drinking it should be a priority to avoid dehydration.

2. Priority 3/5: Collect more stone and iron - Having more stone and iron will enable you to craft better tools and weapons. You currently have only 1 stone and 1 iron in your inventory, so gathering more is essential.

3. Priority 2/5: Locate a food source to restore the food level - Your food level is currently 6/9. Although not an immediate concern, finding a food source soon will ensure that you're prepared to restore your food level later when needed.

Question: What are the requirements for the top sub-task? What should the player do first?
Answer:
Requirements for the top sub-task (Find and drink water to restore the drink level):
1. Locate a water source, such as a lake or a river.
2. Approach the water source.
3. Drink from the water source.

The player should first locate a water source in their surroundings by exploring the area. Once a water source is found, the player should move towards it and drink from the water source to increase their drink level.

Question: List top 5 actions the player should take and the requirement for each action. Choose ONLY from the list of all actions. Indicate their priority out of 5.
Answer:
Priority 1: Action: Move West
Requirement: Flat ground left to the agent.

Priority 2: Action: Move East
Requirement: Flat ground right to the agent.

Priority 3: Action: Move North
Requirement: Flat ground above of the agent.

Priority 4: Action: Move South
Requirement: Flat ground below the agent.

Priority 5: Action: Sleep
Requirement: Energy level is below the maximum.

The player should focus on moving in different directions (North, South, East, and West) to explore the surroundings and find a water source. Once their energy level is low, they should sleep to restore their energy.

Question: For each action in the list, are the requirements met?
Answer:
For the top 5 actions listed, let's check if their requirements are met:

Priority 1: Action: Move West
Requirement: Flat ground left to the agent.
Yes, the requirement is met since the player faces grass at their front (west).

Priority 2: Action: Move East
Requirement: Flat ground right to the agent.
Yes, the requirement is met since the player was able to move east in the most recent step, and no obstacles are mentioned.

Priority 3: Action: Move North
Requirement: Flat ground above of the agent.
Yes, the requirement is met since the player was able to move south in the previous step, and there are no obstacles mentioned in the current observation.

Priority 4: Action: Move South
Requirement: Flat ground below the agent.
Yes, the requirement is met since the player was able to move south two steps ago without any obstacles being mentioned.

Priority 5: Action: Sleep
Requirement: Energy level is below maximum.
Yes, the requirement is met since the player's energy level is 8/9, which is below the maximum.

Question: Did the last player action succeed? Why?
Answer:
It seems that the last player action, move_east, succeeded because the observation after the action indicates that the player's position changed, and there are no negative outcomes mentioned in the observation.

Question: Choose the best executable action from above.
Answer:
The best executable action from the above list is:

Priority 1: Action: Move West
Requirement: Flat ground left to the agent.

This action is chosen because the player faces grass at their front (west), and it allows the player to explore their surroundings and find additional resources, water sources, or potential shelter locations.



============Step: 106, Cumulative Reward: 0.0============
You took action move_west.

You see:
- stone 3 steps to your south-east
- path 2 steps to your north-east
- tree 4 steps to your north-west
- table 2 steps to your south-east

You face grass at your front (west).

Your status:
- health: 9/9
- food: 6/9
- drink: 5/9
- energy: 9/9

Your inventory:
- stone: 1
- coal: 2
- iron: 1
- wood_pickaxe: 1
- stone_pickaxe: 1
------------ QA ------------
Question: Action
Answer:
Move East

Question: List objects in the current observation. For each object, breifly answer what resource it provides and its requirement.
Answer:
1. Stone: Provides stone resource, can be mined with a wood or stone pickaxe.
2. Path: Provides no resource but allows for easy movement with no requirements.
3. Tree: Provides wood resource, can be chopped down with an ax or simply collected with your hands.
4. Table: Provides a crafting surface for creating tools and weapons, requires having a table in your inventory to place it.

In front of you (west):
5. Grass: Provides no resource but can be removed or flattened with a shovel or by other methods, no requirements for interacting with it.

Question: What was the last action taken by the player?
Answer:
The last action taken by the player was move_west.

Question: For each object in the list, are the requirements met for interaction?
Answer:
1. Stone: Yes, the requirements are met for interaction since you have both a wood_pickaxe and a stone_pickaxe in your inventory.
2. Path: Yes, the requirements are met for interaction as there are no specific requirements to move along the path.
3. Tree: Yes, the requirements are met for interaction, you can collect wood directly by interacting with the tree even without having an axe.
4. Table: Yes, the requirements are met since the table is already placed, and you can interact with it to craft tools and weapons.

In front of you (west):
5. Grass: Yes, the requirements are met for interaction, although you don't have a shovel, grass can be removed or flattened by other methods or simply walked over without any barriers.

Question: List top 3 sub-tasks the player should follow. Indicate their priority out of 5.
Answer:
1. Priority 5: Find a water source to drink from and replenish your thirst, as your current drink level is at 5/9.

2. Priority 4: Locate a food source such as a cow or fruits to increase your food level, which currently stands at 6/9.

3. Priority 3: Collect more resources, such as stone, wood, and iron, to enhance your tools and weapons or to create new ones, for better gameplay experience and survival.

Question: What are the requirements for the top sub-task? What should the player do first?
Answer:
Requirements for the top sub-task (Find a water source to drink from):

1. Locate a water source, such as a lake or river, while exploring the game world.
2. Approach the water source and face the water directly.

The player should first prioritize locating a water source. They can do this by exploring the game world, moving in various directions (north, east, south, or west) while keeping an eye on any visible lakes, rivers, or other bodies of water in the environment. Once they find a water source, they should approach it and take a drink to replenish their thirst.

Question: List top 5 actions the player should take and the requirement for each action. Choose ONLY from the list of all actions. Indicate their priority out of 5.
Answer:
1. Priority 1 - Move East: Requirements - Flat ground right to the agent. This action will bring the player closer to the table and stone observed in their current position.

2. Priority 2 - Move South: Requirements - Flat ground below the agent. Moving south will also bring the player closer to the stone and the table.

3. Priority 3 - Make Iron Pickaxe: Requirements - Nearby table, furnace; wood, coal, iron in inventory. Assuming the player has access to a furnace, this action will allow them to upgrade their pickaxe.

4. Priority 4 - Move North: Requirements - Flat ground above the agent. Exploring the area might help the player find a water source or other helpful resources.

5. Priority 5 - Move West: Requirements - Flat ground left to the agent. This action allows the player to explore and possibly find new resources or water sources. The player should also be prepared to change direction if an obstacle is encountered.

Question: For each action in the list, are the requirements met?
Answer:
1. Move East: Yes, the requirements are met. Flat ground is right to the player.

2. Move South: Yes, the requirements are met. Flat ground is below the player.

3. Make Iron Pickaxe: Partially met. The player has wood, coal, and iron in their inventory. However, it's unclear whether the player currently has access to a furnace. The player has to be nearby the table and the furnace to craft the iron pickaxe.

4. Move North: Yes, the requirements are met. Flat ground is above the player.

5. Move West: Yes, the requirements are met. Flat ground is left to the player.

Question: Did the last player action succeed? Why?
Answer:
Yes, the last player action (move_west) succeeded because the player was able to change their position without any obstacles or restrictions mentioned in the observation. The observation shows that the player moved from facing grass at their front (west) in step 210 to facing grass again at their front (west) in step 212.

Question: Choose the best executable action from above.
Answer:
The best executable action from above is to Move East. This action meets the requirements and brings the player closer to the table and stone, which are important for crafting and building.



============Step: 107, Cumulative Reward: 1.0============
You took action move_east.

You see:
- grass 1 steps to your west
- stone 2 steps to your south-east
- tree 3 steps to your south
- table 1 steps to your south

You face path at your front (east).

Your status:
- health: 9/9
- food: 6/9
- drink: 5/9
- energy: 9/9

Your inventory:
- stone: 1
- coal: 2
- iron: 1
- wood_pickaxe: 1
- stone_pickaxe: 1
------------ QA ------------
Question: Action
Answer:
Move South

Question: List objects in the current observation. For each object, breifly answer what resource it provides and its requirement.
Answer:
1. Grass: Provides no direct resources but can be used as a location marker or to be traversed; no requirements.
2. Stone: Provides stone resources for crafting tools and building; requires a wood pickaxe or better to be mined.
3. Path: Provides guidance and ease of travel in the game; no requirements.
4. Tree: Provides wood resources for crafting tools and building; requires any tool or bare hands to be chopped down.
5. Table: Provides a surface to craft tools and weapons; requires wood to be crafted initially.

Question: What was the last action taken by the player?
Answer:
The last action taken by the player was move_east.

Question: For each object in the list, are the requirements met for interaction?
Answer:
1. Grass: Yes, the requirements are met for interaction as there are no specific requirements to interact with grass.
2. Stone: Yes, the requirements are met for interaction as the player has a wood_pickaxe and a stone_pickaxe in their inventory.
3. Path: Yes, the requirements are met for interaction as there are no specific requirements to travel along the path.
4. Tree: Yes, the requirements are met for interaction as the player can use any tool or their bare hands to chop down the tree.
5. Table: Yes, the requirements are met for interaction as the player already has a table placed in their environment.

Question: List top 3 sub-tasks the player should follow. Indicate their priority out of 5.
Answer:
1. Collect more resources (Priority: 5/5)
   - Chop down the tree for more wood
   - Mine stone using the stone_pickaxe (faster than the wood_pickaxe)
   - Consider how to collect more iron, as the player currently has only one piece

2. Improve the player's health, food, and drink levels (Priority: 4/5)
   - Find food, such as fruits or hunt animals, to increase the food level
   - Drink from a water source, like a lake or river, to increase the drink level
   - Ensure the player avoids monsters and dangerous situations to maintain health

3. Craft better tools and weapons (Priority: 3/5)
   - Use the table to craft necessary or improved items
   - Consider crafting an iron sword or other iron tools when more iron is collected
   - Keep an eye on the durability of the current tools and prepare to replace them when needed

Question: What are the requirements for the top sub-task? What should the player do first?
Answer:
Requirements for the top sub-task (Collect more resources):

1. Tree nearby to chop for wood
2. Stone nearby to mine
3. Appropriate tools in the inventory (wood_pickaxe and stone_pickaxe)

Since the player already has the necessary tools and is near both a tree and stone, the player should first:

1. Move south towards the tree to chop it for more wood.

Question: List top 5 actions the player should take and the requirement for each action. Choose ONLY from the list of all actions. Indicate their priority out of 5.
Answer:
1. Priority 1: Move South (Requirement: Flat ground below the agent)
2. Priority 2: Do (Requirement: Facing tree; have wood_pickaxe)
3. Priority 3: Move South-East (Requirement: Flat ground to the south-east of the agent)
4. Priority 4: Do (Requirement: Facing stone; have stone_pickaxe)
5. Priority 5: Move to a water source and drink (Requirement: At a water source; thirst level below maximum)

Question: For each action in the list, are the requirements met?
Answer:
1. Priority 1: Move South (Requirement: Flat ground below the agent) - Yes, the requirement is met.
2. Priority 2: Do (Requirement: Facing tree; have wood_pickaxe) - No, the player is not facing a tree.
3. Priority 3: Move South-East (Requirement: Flat ground to the south-east of the agent) - Yes, the requirement is met.
4. Priority 4: Do (Requirement: Facing stone; have stone_pickaxe) - No, the player is not facing stone.
5. Priority 5: Move to a water source and drink (Requirement: At a water source; thirst level below maximum) - No, the player is not at a water source.

Question: Did the last player action succeed? Why?
Answer:
Yes, the last player action "move_east" succeeded. The player moved from one location (facing grass) to another location (facing a path), and the observation shows the change in scenery and objects around the player after the move. There were no obstacles or restrictions mentioned, so the move was executed successfully.

Question: Choose the best executable action from above.
Answer:
The best executable action from the list above is Priority 1: Move South, since its requirement (Flat ground below the agent) is met.



============Step: 108, Cumulative Reward: 0.0============
You took action move_south.

You see:
- grass 1 steps to your west
- stone 2 steps to your south-east
- path 1 steps to your north
- tree 3 steps to your south

You face table at your front (south).

Your status:
- health: 9/9
- food: 6/9
- drink: 5/9
- energy: 9/9

Your inventory:
- stone: 1
- coal: 2
- iron: 1
- wood_pickaxe: 1
- stone_pickaxe: 1
------------ QA ------------
Question: Action
Answer:
Move East

Question: List objects in the current observation. For each object, breifly answer what resource it provides and its requirement.
Answer:
1. Grass: 
   Resource provided: Can be used to create paths or collect seeds for farming.
   Requirement: Generally, no specific requirements to interact with grass.

2. Stone:
   Resource provided: Can be used to craft stone tools and build structures.
   Requirement: Requires a wood_pickaxe or better to collect stone.

3. Path:
   Resource provided: A guiding passage between different locations.
   Requirement: No specific requirement to interact, mainly used for navigation.

4. Tree:
   Resource provided: Provides wood for crafting tools, weapons, and structures.
   Requirement: Can be collected using any tool, but using an axe (wooden or better) is more efficient.

5. Table:
   Resource provided: A crafting station to create tools and weapons.
   Requirement: Must have wooden material to craft a table. To use, stand adjacent to the table and interact with required materials in inventory.

Question: What was the last action taken by the player?
Answer:
The last action taken by the player was move_south.

Question: For each object in the list, are the requirements met for interaction?
Answer:
1. Grass: 
   Requirements met: Yes, you can interact with grass without any specific tool or requirement.

2. Stone:
   Requirements met: Yes, since you have both a wood_pickaxe and a stone_pickaxe in your inventory, you can interact with the stone to collect it.

3. Path:
   Requirements met: Yes, you can freely move along the path for navigation without any specific requirement.

4. Tree:
   Requirements met: Although you do not have an axe, you can still interact with the tree using either the wood_pickaxe or stone_pickaxe from your inventory, but it will be less efficient than using an axe.

5. Table:
   Requirements met: Yes, you are adjacent to the table and can interact with it using the required materials in your inventory for crafting purposes.

Question: List top 3 sub-tasks the player should follow. Indicate their priority out of 5.
Answer:
1. Craft an iron_pickaxe and iron_sword (Priority: 5/5)

To achieve these goals, you would need more iron and to place a furnace. Once the furnace is placed, use the coal in the inventory for smelting additional iron.

2. Increase food and drink levels (Priority: 4/5)

Find sources of food (e.g., cows, fruits) and water (e.g., lakes, rivers). Interact with them to consume food and drink in order to restore your food and drink levels.

3. Collect more resources (Priority: 3/5)

Gather more wood, stone, and other essential resources to ensure you have enough materials for future crafting, building, and surviving potential threats.

Question: What are the requirements for the top sub-task? What should the player do first?
Answer:
To complete the top sub-task of crafting an iron_pickaxe and iron_sword, the player needs to meet the following requirements:

1. Collect more iron: The player currently has 1 iron in their inventory. Crafting an iron_pickaxe and iron_sword requires a total of 5 iron (3 for the pickaxe, 2 for the sword). To collect more iron, the player should use their stone_pickaxe to mine iron ores. 

2. Place a furnace: The player needs a furnace to smelt the iron ores. The player has enough stone in their inventory (1) to craft a furnace. The player should place their furnace in a nearby suitable location.

3. Smelt iron ores into iron ingots: Once the furnace is placed, the player can use the coal (2) in their inventory to smelt the iron ores into iron ingots.

4. Craft iron_pickaxe and iron_sword: With sufficient iron ingots and a placed furnace, the player can then use the nearby table to craft the iron_pickaxe and iron_sword.

The first thing the player should do is place the furnace in a suitable location, as they already have enough stone in their inventory to craft it. Then, they can proceed with collecting more iron using their stone_pickaxe.

Question: List top 5 actions the player should take and the requirement for each action. Choose ONLY from the list of all actions. Indicate their priority out of 5.
Answer:
1. Priority 1: Place Furnace (Requirement: Stone in inventory)
   The player has enough stone in their inventory to craft a furnace, which is needed for smelting iron ores later.

2. Priority 2: Move East (Requirement: Flat ground right to the agent)
   The player should start exploring their surroundings to locate iron ores which are necessary for crafting an iron_pickaxe and iron_sword.

3. Priority 3: Move South (Requirement: Flat ground below the agent)
   Continuing exploration in search of iron ores, the player should move in different directions.

4. Priority 4: Do (Requirement: Facing creature or material; have necessary tool)
   Once the player locates iron ores, they should use their stone_pickaxe to mine them for collecting more iron.

5. Priority 5: Move North (Requirement: Flat ground above the agent)
   As part of exploring the surroundings and possibly finding more resources, the player can move in this direction as well.

Question: For each action in the list, are the requirements met?
Answer:
1. Place Furnace: Yes, the requirement is met. The player has 1 stone in their inventory, which is enough to place a furnace.

2. Move East: Yes, the requirement is met. The player faces a path (flat ground) to their east.

3. Move South: No, the requirement is not met at this moment due to the player facing a table. The player needs to face flat ground to the south to move in that direction.

4. Do: No, the requirement is not met. The player has not found any iron ores and is not facing any creature or material to interact with.

5. Move North: Yes, the requirement is met. The player has a path (flat ground) 1 step to their north, which makes it possible to move in that direction.

Question: Did the last player action succeed? Why?
Answer:
Yes, the last player action (move_south) succeeded because there was a clear path with no obstacles in the south direction, allowing the player to move in that direction.

Question: Choose the best executable action from above.
Answer:
The best executable action from the provided list is action 2: Move East, as the player faces a path (flat ground) in the east direction and it allows the player to explore their surroundings, possibly finding more resources.



============Step: 109, Cumulative Reward: 0.0============
You took action move_east.

You see:
- grass 1 steps to your west
- path 1 steps to your north
- tree 4 steps to your south-west
- table 2 steps to your south-west

You face stone at your front (east).

Your status:
- health: 9/9
- food: 6/9
- drink: 5/9
- energy: 9/9

Your inventory:
- stone: 1
- coal: 2
- iron: 1
- wood_pickaxe: 1
- stone_pickaxe: 1
------------ QA ------------
Question: Action
Answer:
Move West

Question: List objects in the current observation. For each object, breifly answer what resource it provides and its requirement.
Answer:
1. Grass: Provides no specific resources. It can be used to build paths or walk on.
Requirements: Flat ground.

2. Stone: Provides stone resource when mined with a pickaxe (wooden or better).
Requirements: A pickaxe to mine and collect the stone.

3. Path: Allows easier movement and navigation. Provides no specific resources.
Requirements: Flat ground to walk on.

4. Tree: Provides wood resource when chopped with an axe or tool.
Requirements: An axe or tool to chop and collect the wood.

5. Table: Used for crafting items such as pickaxes, swords, and other tools.
Requirements: Wood to craft the table itself.

Question: What was the last action taken by the player?
Answer:
The last action taken by the player was move_east.

Question: For each object in the list, are the requirements met for interaction?
Answer:
1. Grass: Yes, the requirements are met for interaction as it is on flat ground and can be walked upon or used to build paths.

2. Stone: Yes, the requirements are met for interaction since you have both a wood_pickaxe and a stone_pickaxe in your inventory.

3. Path: Yes, the requirements are met for interaction as it is on flat ground, and you can walk on it.

4. Tree: The requirements for interaction are partially met. Although you have tools such as pickaxes in your inventory, you do not have an axe, which would be more effective for chopping wood. You can still attempt to collect wood using a pickaxe, but it would be less efficient than using an axe.

5. Table: Yes, the requirements are met for interaction because the table is already crafted using wood, and it can be used to craft items such as pickaxes, swords, and other tools.

Question: List top 3 sub-tasks the player should follow. Indicate their priority out of 5.
Answer:
1. Find a water source and drink to restore the thirst level (Priority: 5/5)
Since the player's drink level is 5/9, finding a water source to replenish their thirst should be a high priority.

2. Collect more wood from a tree (Priority: 4/5)
Wood is essential for crafting various tools and objects. Although the player currently has pickaxes, obtaining more wood will be useful for future crafting and building.

3. Craft a furnace and smelt iron (Priority: 3/5)
The player has enough coal and iron in their inventory to smelt more iron, but they need to craft a furnace first. Smelting iron will allow them to craft an iron pickaxe or sword, which is more effective than their current stone tools.

Question: What are the requirements for the top sub-task? What should the player do first?
Answer:
Top sub-task: Find a water source and drink to restore the thirst level.

Requirements:
1. Locate a water source such as a lake, pond, or river.
2. Approach the water source.
3. Drink the water to replenish thirst levels.

First, the player should explore the environment to find a water source. Once they have located a nearby water source, the player should move towards it and drink from it to restore their thirst level.

Question: List top 5 actions the player should take and the requirement for each action. Choose ONLY from the list of all actions. Indicate their priority out of 5.
Answer:
1. Priority 1: Move West (Requirement: Flat ground left to the agent) - To explore and search for a water source.

2. Priority 2: Move East (Requirement: Flat ground right to the agent) - To explore and search for a water source.

3. Priority 3: Move North (Requirement: Flat ground above of the agent) - To explore and search for a water source.

4. Priority 4: Move South (Requirement: Flat ground below the agent) - To explore and search for a water source.

5. Priority 5: Do (Facing creature or material; have necessary tool) - This action is to drink water once they have found and approached it. However, the player should only perform this action when facing a water source and have no tool requirements needed to drink water.

Question: For each action in the list, are the requirements met?
Answer:
1. Move West: Yes, the requirement is met as there is flat ground (grass) 1 step to the agent's west.
  
2. Move East: Yes, the requirement is met since there is flat ground at agent's east side.

3. Move North: Yes, the requirement is met as there is flat ground (path) 1 step to the agent's north.

4. Move South: Yes, the requirement is met since there is flat ground at agent's south side.

5. Do (drinking water): No, the requirements are not met since the player is currently not facing a water source. The player needs to find and approach a water source before attempting to drink water.

Question: Did the last player action succeed? Why?
Answer:
Yes, the last player action (move_east) succeeded because there was a flat ground (path) to the east of the player, allowing them to move in that direction.

Question: Choose the best executable action from above.
Answer:
The best executable action from the above list is Move West, as it meets the requirement of having flat ground to the agent's west (grass) and allows the player to explore and search for a water source.



============Step: 110, Cumulative Reward: 0.0============
You took action move_west.

You see:
- stone 2 steps to your south-east
- path 1 steps to your north
- tree 3 steps to your south
- table 1 steps to your south

You face grass at your front (west).

Your status:
- health: 9/9
- food: 6/9
- drink: 5/9
- energy: 9/9

Your inventory:
- stone: 1
- coal: 2
- iron: 1
- wood_pickaxe: 1
- stone_pickaxe: 1
------------ QA ------------
Question: Action
Answer:
Move East

Question: List objects in the current observation. For each object, breifly answer what resource it provides and its requirement.
Answer:
1. Stone: Provides stone resources for crafting tools, weapons, and placing blocks; requires a wood_pickaxe or better to collect.
2. Grass: No resources provided and no requirements.
3. Path: No resources provided and no requirements.
4. Tree: Provides wood resources for crafting tools, weapons, and other objects; requires no specific tool to collect, but an axe is more efficient.
5. Table: Allows crafting of tools and weapons; requires wood to place one.

Resources not in the immediate observation but in your inventory:
6. Coal: Used to smelt iron, and cooking food; requires a wood_pickaxe or better to collect.
7. Iron: Used to craft better tools and weapons; requires a stone_pickaxe or better to collect, and a furnace with coal to smelt it.

Question: What was the last action taken by the player?
Answer:
The last action taken by the player was "move_west."

Question: For each object in the list, are the requirements met for interaction?
Answer:
Based on the current observation and inventory, here is the list of objects and whether the requirements are met for interaction:

1. Stone: Yes, you have a wood_pickaxe and stone_pickaxe in your inventory, allowing you to interact with the stone.
2. Grass: Yes, no specific requirements are needed to interact with the grass.
3. Path: Yes, no specific requirements are needed to interact with the path.
4. Tree: Yes, although you don't have an axe in your inventory, you can still interact with the tree and collect wood.
5. Table: Yes, the table is in your surroundings, allowing you to craft tools and weapons.

Resources not in the immediate observation but in your inventory:
6. Coal: You have coal in your inventory, but you cannot directly interact with it in your current environment.
7. Iron: You have iron in your inventory, but you cannot directly interact with it in your current environment. However, you can smelt the iron using a furnace and coal once you place a furnace, which you can craft given the stone in your inventory.

Question: List top 3 sub-tasks the player should follow. Indicate their priority out of 5.
Answer:
1. Crafting and placing a Furnace (Priority: 5/5) - Crafting a furnace will allow you to smelt iron and create better tools and weapons. You have enough stone in your inventory to craft a furnace.

2. Smelting Iron (Priority: 4/5) - Once the furnace is placed, smelt the iron from your inventory using the coal you have. This will allow you to craft an iron pickaxe and iron sword, which will help you gather resources faster and be better prepared for combat.

3. Collecting Wood (Priority: 3/5) - You'll need more wood resources for crafting additional items, fueling the furnace, and building other structures. Find a nearby tree and gather wood. Although it's not necessary, crafting an axe will make wood collection faster.

While not in the top 3, consider increasing your food and drink levels in the near future to avoid negative effects on your health and energy.

Question: What are the requirements for the top sub-task? What should the player do first?
Answer:
For the top sub-task (Crafting and placing a Furnace):

Requirements:
- Have enough stone in the inventory (at least 8)
- Be near a placed table

Since you already have 1 stone in your inventory, the first thing the player should do is:

1. Collect 7 more stones using the stone_pickaxe. Find and move towards stone blocks in your surroundings to mine them with your pickaxe.

Question: List top 5 actions the player should take and the requirement for each action. Choose ONLY from the list of all actions. Indicate their priority out of 5.
Answer:
Priority 1: Move East - Requirement: Flat ground right to the agent.
Reason: Move back towards the stone block to mine more stone.

Priority 2: Do (Mine Stone) - Requirement: Facing stone and have a necessary tool (stone_pickaxe).
Reason: Mine the stone block to collect more stone and reach the required amount for crafting a furnace.

Priority 3: Move North (Optional) - Requirement: Flat ground above the agent.
Reason: This action helps explore new locations in case the current surroundings do not have enough stone blocks.

Priority 4: Move West (Optional) - Requirement: Flat ground left to the agent.
Reason: This action is also helpful for exploring new locations if not enough stone blocks are found nearby.

Priority 5: Make Furnace - Requirement: Nearby table; having at least 8 stone in the inventory.
Reason: Once you have collected enough stone, craft a furnace by standing near the table and utilizing the acquired resources.

Please note that the priority 3 and 4 actions are optional, as they depend on whether there are enough stone blocks in the immediate surroundings to collect the required materials for the furnace.

Question: For each action in the list, are the requirements met?
Answer:
Priority 1: Move East - Requirement: Flat ground right to the agent.
- Yes, the requirement is met as the player had just moved west from a flat ground.

Priority 2: Do (Mine Stone) - Requirement: Facing stone and have a necessary tool (stone_pickaxe).
- In the Player Observation Step 220, the player faces grass. So, the player must first move east (Priority 1) to face the stone block. The player has a stone_pickaxe in their inventory, so they have the necessary tool.

Priority 3: Move North (Optional) - Requirement: Flat ground above the agent.
- Yes, the requirement is met as there is a path 1 step to the north.

Priority 4: Move West (Optional) - Requirement: Flat ground left to the agent.
- Yes, the requirement is met since grass is mentioned in Player Observation Step 220 when the player moved west.

Priority 5: Make Furnace - Requirement: Nearby table; having at least 8 stone in the inventory.
- The player is currently near a table, but they only have 1 stone in their inventory. The requirement is not met yet. The player needs to perform Priority 2 action (Do - Mine Stone) until they have enough stone.

Question: Did the last player action succeed? Why?
Answer:
Yes, the last player action (move_west) succeeded. It succeeded because the player was able to move to a location with flat ground (grass) to their west without any obstacles or restrictions.

Question: Choose the best executable action from above.
Answer:
The best executable action from the list above is Priority 1: Move East, as the requirement is met (flat ground right to the agent), and it will position the player to face the stone block, which is essential for the following actions.



============Step: 111, Cumulative Reward: 0.0============
You took action move_east.

You see:
- grass 1 steps to your west
- path 1 steps to your north
- tree 4 steps to your south-west
- table 2 steps to your south-west

You face stone at your front (east).

Your status:
- health: 9/9
- food: 6/9
- drink: 5/9
- energy: 9/9

Your inventory:
- stone: 1
- coal: 2
- iron: 1
- wood_pickaxe: 1
- stone_pickaxe: 1
------------ QA ------------
Question: Action
Answer:
Do

Question: List objects in the current observation. For each object, breifly answer what resource it provides and its requirement.
Answer:
1. Grass (1 step to the west): Provides no resources. No requirements to interact with it.

2. Path (1 step to the north): Provides no resources. No requirements to interact with it.

3. Tree (4 steps to the south-west): Provides wood. Requires a wood pickaxe or better to chop it down.

4. Table (2 steps to the south-west): Allows crafting of tools and weapons. Requires wood to craft items.

5. Stone (facing east): Provides stone. Requires a wood pickaxe or better to mine it.

Question: What was the last action taken by the player?
Answer:
The last action taken by the player was move_east.

Question: For each object in the list, are the requirements met for interaction?
Answer:
1. Grass (1 step to the west): No requirements to interact with it. So, yes the requirements are met.

2. Path (1 step to the north): No requirements to interact with it. So, yes the requirements are met.

3. Tree (4 steps to the south-west): Requires a wood pickaxe or better to chop it down. The player has a wood_pickaxe and a stone_pickaxe in the inventory. So, yes the requirements are met.

4. Table (2 steps to the south-west): Requires wood to craft items. The player's inventory does not have wood for crafting. So, no the requirements are not met.

5. Stone (facing east): Requires a wood pickaxe or better to mine it. The player has a wood_pickaxe and a stone_pickaxe in the inventory. So, yes the requirements are met.

Question: List top 3 sub-tasks the player should follow. Indicate their priority out of 5.
Answer:
1. Mine Stone (Priority: 5/5): The player is facing a stone, and since they have a stone_pickaxe in their inventory, they should mine the stone to collect more stone resources.

2. Collect Wood (Priority: 4/5): The player needs wood to be able to craft items at the table. They should move towards the tree and chop it down using their wood_pickaxe or stone_pickaxe to collect wood.

3. Restore Food and Drink (Priority: 3/5): The player's food is at 6/9, and drink is at 5/9. They should hunt for food sources, such as cows or fruit, and find a water source, like a lake or river, to increase their food and drink levels.

Question: What are the requirements for the top sub-task? What should the player do first?
Answer:
Requirements for the top sub-task (Mine Stone):
1. Face the stone: The player is already facing the stone (east direction).
2. Have a pickaxe in inventory: The player has both wood_pickaxe and stone_pickaxe in their inventory.

What should the player do first:
The player should use their stone_pickaxe to mine the stone they are facing. Thus, the first step is to take action 'do' with their stone_pickaxe to mine and collect the stone.

Question: List top 5 actions the player should take and the requirement for each action. Choose ONLY from the list of all actions. Indicate their priority out of 5.
Answer:
1. Priority 1: Do (stone_pickaxe)
   Requirement: Facing stone (already facing stone in the east direction), have a pickaxe in inventory (stone_pickaxe is present in the player's inventory)

2. Priority 2: Move North
   Requirement: Flat ground to the north of the agent (there is a path 1 step to the north)

3. Priority 3: Place Furnace
   Requirement: Stone in inventory (after mining the stone in priority 1, the player will have the required stones)

4. Priority 4: Make Iron Sword
   Requirement: Nearby table, furnace (needs to be placed as per action in priority 3), wood, coal, and iron in inventory (coal and iron are already in the player's inventory)

5. Priority 5: Sleep
   Requirement: Energy level is below maximum (once the energy level becomes lower)

Question: For each action in the list, are the requirements met?
Answer:
1. Priority 1: Do (stone_pickaxe)
   Requirements Met: Yes, the player is facing stone and has a stone_pickaxe in their inventory.

2. Priority 2: Move North
   Requirements Met: Yes, there is a flat ground (path) to the north of the agent.

3. Priority 3: Place Furnace
   Requirements Met: No, the player currently has only 1 stone in their inventory. However, after mining the stone in Priority 1, the player will obtain enough stone to place a furnace.

4. Priority 4: Make Iron Sword
   Requirements Met: No, the player needs to place the furnace first (Priority 3) and then also have a table nearby. Once a furnace is placed and a table is nearby, this requirement will be met.

5. Priority 5: Sleep
   Requirements Met: No, the player's energy level is currently at maximum (9/9). The requirement will be met once the energy level decreases.

Question: Did the last player action succeed? Why?
Answer:
Yes, the last player action succeeded because there were no obstructions or limitations preventing the player from moving east. The player was able to move east and face a stone in front of them.

Question: Choose the best executable action from above.
Answer:
The best executable action from the above list is Priority 1: Do (stone_pickaxe), as the requirements are met. The player is facing the stone and has a stone_pickaxe in their inventory.



============Step: 112, Cumulative Reward: 0.0============
You took action do.

You see:
- grass 1 steps to your west
- stone 1 steps to your south
- tree 4 steps to your south-west
- table 2 steps to your south-west

You face path at your front (east).

Your status:
- health: 9/9
- food: 6/9
- drink: 5/9
- energy: 9/9

Your inventory:
- stone: 2
- coal: 2
- iron: 1
- wood_pickaxe: 1
- stone_pickaxe: 1
------------ QA ------------
Question: Action
Answer:
Move South

Question: List objects in the current observation. For each object, breifly answer what resource it provides and its requirement.
Answer:
1. Grass: Resource provided is none, Requirement: None
2. Stone: Resource provided is Stone, Requirement: Wood pickaxe or Stone pickaxe
3. Tree: Resource provided is wood, Requirement: None
4. Table: Resource provided is crafting area, Requirement: wood
5. Path: Resource provided is easy walkable terrain, Requirement: None

Current inventory resources:
1. Stone: 2
2. Coal: 2
3. Iron: 1
4. Wood_pickaxe: 1
5. Stone_pickaxe: 1

Question: What was the last action taken by the player?
Answer:
The last action taken by the player was "do".

Question: For each object in the list, are the requirements met for interaction?
Answer:
1. Grass: Requirements met for interaction? Yes
2. Stone: Requirements met for interaction? Yes (you have a wood_pickaxe and stone_pickaxe in your inventory)
3. Tree: Requirements met for interaction? Yes
4. Table: Requirements met for interaction? Yes (it's already placed in the environment)
5. Path: Requirements met for interaction? Yes (it's a walkable terrain)

Current inventory resources:
1. Stone: 2
2. Coal: 2
3. Iron: 1
4. Wood_pickaxe: 1
5. Stone_pickaxe: 1

You can interact with all objects mentioned in the list as their requirements have been met.

Question: List top 3 sub-tasks the player should follow. Indicate their priority out of 5.
Answer:
1. Collect more resources: Priority 4/5
   - Interact with nearby tree to gather wood, and with stone to gather more stone.

2. Address the Food and Drink needs: Priority 5/5
   - Find and consume food, such as hunting cows or growing plants, to restore food levels.
   - Find a water source, such as a lake, and drink to restore drink levels.

3. Craft advanced tools and weapons: Priority 3/5
   - Use the table to craft an Iron Sword and Iron Pickaxe for better combat and resource collection. (Requires more iron, usage of the furnace for smelting, and coal).

The highest priority should be addressing food and drink needs while collecting more resources and crafting advanced tools and weapons being secondary objectives.

Question: What are the requirements for the top sub-task? What should the player do first?
Answer:
Top Sub-Task: Address the Food and Drink needs: Priority 5/5

Requirements:
- Restore Food Level: Find and consume food, such as hunting cows or growing plants.
- Restore Drink Level: Find a water source, such as a lake, and drink.

First Step: The player should focus on restoring their food level. They can achieve this by hunting cows or growing and consuming plants. To do this, they can search the area for cows, saplings, or plantations to obtain food.

Question: List top 5 actions the player should take and the requirement for each action. Choose ONLY from the list of all actions. Indicate their priority out of 5.
Answer:
1. Action: Move West (Priority: 3/5)
   Requirement: Flat ground left to the agent.
   Reason: Move towards the tree to collect wood for crafting, farming, or building.

2. Action: Make Wood Sword (Priority: 2/5)
   Requirement: Nearby table; wood in inventory.
   Reason: Crafting a wood sword will help the player defend themselves against monsters.

3. Action: Place Plant (Priority: 4/5)
   Requirement: Sapling in inventory.
   Reason: Placing a plant will allow the player to grow and collect food to restore the food level.

4. Action: Move South (Priority: 1/5)
   Requirement: Flat ground below the agent.
   Reason: Approach the table to craft tools or weapons, and move closer to potential water sources.

5. Action: Move East (Priority: 5/5)
   Requirement: Flat ground right to the agent.
   Reason: Search for a water source to drink and restore the drink level.

Question: For each action in the list, are the requirements met?
Answer:
1. Action: Move West
   Requirement: Flat ground left to the agent.
   Met? Yes, there is grass 1 step to the west, so the requirement is met.

2. Action: Make Wood Sword
   Requirement: Nearby table; wood in inventory.
   Met? No, the player does not have wood in their inventory, so the requirement is not met.

3. Action: Place Plant
   Requirement: Sapling in inventory.
   Met? No, the player does not have a sapling in their inventory, so the requirement is not met.

4. Action: Move South
   Requirement: Flat ground below the agent.
   Met? Yes, there is a stone 1 step to the south, so the requirement is met.

5. Action: Move East
   Requirement: Flat ground right to the agent.
   Met? Yes, there is a path at the front (east) of the player, so the requirement is met.

Question: Did the last player action succeed? Why?
Answer:
Yes, the last player action succeeded. The player used the "do" action while facing stone and had the necessary tool (either a wood_pickaxe or a stone_pickaxe) in their inventory. As a result, they were able to collect the stone, which increased their stone inventory from 1 to 2.

Question: Choose the best executable action from above.
Answer:
The best executable action from the above list is:

Action: Move South
Requirement: Flat ground below the agent.

This action meets the requirement as there is a stone 1 step to the south. It will also bring the player closer to the table and potential water sources.



============Step: 113, Cumulative Reward: 0.0============
You took action move_south.

You see:
- grass 1 steps to your west
- path 1 steps to your north
- tree 4 steps to your south-west
- table 2 steps to your south-west

You face stone at your front (south).

Your status:
- health: 9/9
- food: 6/9
- drink: 5/9
- energy: 9/9

Your inventory:
- stone: 2
- coal: 2
- iron: 1
- wood_pickaxe: 1
- stone_pickaxe: 1
------------ QA ------------
Question: Action
Answer:
Do

Question: List objects in the current observation. For each object, breifly answer what resource it provides and its requirement.
Answer:
1. Grass: 
   - Resource provided: Saplings (by doing with empty hands).
   - Requirement: None.

2. Stone:
   - Resource provided: Stone (by doing with a pickaxe).
   - Requirement: A pickaxe (wood, stone, or iron).

3. Path:
   - Resource provided: None.
   - Requirement: None.

4. Tree:
   - Resource provided: Wood (by doing with an axe or empty hands).
   - Requirement: None or an axe to collect more efficiently.

5. Table:
   - Resource provided: Crafting capabilities (Allows crafting of tools and weapons by doing).
   - Requirement: Wood in inventory, previously placed.

In your inventory:
6. Wood_pickaxe:
   - Resource provided: Allows mining of stone resources.
   - Requirement: Crafting with wood at a table.

7. Stone_pickaxe:
   - Resource provided: Allows mining of iron resources, mines stone faster.
   - Requirement: Crafting with stone and wood at a table.

Question: What was the last action taken by the player?
Answer:
The last action taken by the player was "move_south".

Question: For each object in the list, are the requirements met for interaction?
Answer:
1. Grass: 
   - Requirements met: Yes, you can interact with grass using empty hands to collect saplings.

2. Stone:
   - Requirements met: Yes, you have a wood_pickaxe and a stone_pickaxe in your inventory, allowing you to mine stone resources.

3. Path:
   - Requirements met: N/A (No interaction with path).

4. Tree:
   - Requirements met: Yes, you can interact with the tree using empty hands or an axe (not in the inventory) to collect wood.

5. Table:
   - Requirements met: Yes, the table is placed, and you have resources in your inventory to craft tools and weapons by interacting with it.

In your inventory:
6. Wood_pickaxe:
   - Requirements met: Yes, the wood_pickaxe has already been crafted and can be used to mine stone resources.

7. Stone_pickaxe:
   - Requirements met: Yes, the stone_pickaxe has already been crafted and can be used to mine iron and stone resources more efficiently.

Question: List top 3 sub-tasks the player should follow. Indicate their priority out of 5.
Answer:
1. Priority 4/5: Collect more iron by mining the stone you're facing using the stone_pickaxe, as you have only one iron in the inventory. This will help you progress further by crafting better tools and weapons.

2. Priority 3/5: Collect wood by doing with the tree 4 steps to your south-west. You'll need the wood for crafting tools and weapons or building shelters.

3. Priority 4/5: Increase food and drink supplies to maintain health. You can achieve this by finding saplings from grass and planting them to grow fruits, and by finding a water source to drink.

Question: What are the requirements for the top sub-task? What should the player do first?
Answer:
Requirements for the top sub-task (Collect more iron by mining the stone you're facing using the stone_pickaxe):

1. Facing the stone: You are already facing the stone at your front (south).

2. Having a stone_pickaxe in the inventory: You have a stone_pickaxe in your inventory.

What should the player do first:
Since you are already facing the stone and have the stone_pickaxe in your inventory, you can start mining the stone in front of you. To do this, use the "do" action with the stone_pickaxe equipped.

Question: List top 5 actions the player should take and the requirement for each action. Choose ONLY from the list of all actions. Indicate their priority out of 5.
Answer:
1. Priority 1 (Highest): Action "Do" (Mine the stone you're facing using the stone_pickaxe)
   Requirement: Facing the stone and having a stone_pickaxe in the inventory (both requirements met from the last observation).

2. Priority 2: Action "Move West" (Get closer to the tree)
   Requirement: Flat ground left to the agent (grass 1 step to your west).

3. Priority 3: Action "Do" (Chop down the tree to collect more wood)
   Requirement: Facing the tree and having a wood_pickaxe in the inventory (wood_pickaxe already in inventory, move closer to the tree first).

4. Priority 4: Action "Move North" (Head back to the table)
   Requirement: Flat ground above the agent (path 1 step to your north).

5. Priority 5: Action "Sleep" (Restore energy)
   Requirement: Energy level is below maximum (currently at 9/9 energy, perform other actions to decrease energy before taking this action).

Question: For each action in the list, are the requirements met?
Answer:
1. Priority 1: Action "Do" (Mine the stone you're facing using the stone_pickaxe)
   - Are the requirements met? Yes, the agent is facing the stone and has a stone_pickaxe in their inventory.

2. Priority 2: Action "Move West" (Get closer to the tree)
   - Are the requirements met? Yes, there is flat ground (grass) one step to the agent's west.

3. Priority 3: Action "Do" (Chop down the tree to collect more wood)
   - Are the requirements met? No, although the agent has a wood_pickaxe in their inventory, they're not facing the tree yet. The agent must move closer to the tree first.

4. Priority 4: Action "Move North" (Head back to the table)
   - Are the requirements met? Yes, there is flat ground (path) one step to the agent's north.

5. Priority 5: Action "Sleep" (Restore energy)
   - Are the requirements met? No, the agent's energy level is currently at its maximum (9/9). The agent needs to perform other actions to decrease their energy before taking this action.

Question: Did the last player action succeed? Why?
Answer:
Yes, the last player action "move_south" succeeded because there was a flat ground (path) in the south direction that allowed the player to move.

Question: Choose the best executable action from above.
Answer:
The best executable action from the list above is Priority 1: Action "Do" (Mine the stone you're facing using the stone_pickaxe), as it has all the requirements met (facing the stone and having a stone_pickaxe in the inventory).



============Step: 114, Cumulative Reward: 0.0============
You took action do.

You see:
- grass 1 steps to your west
- stone 2 steps to your south
- tree 4 steps to your south-west
- table 2 steps to your south-west

You face path at your front (south).

Your status:
- health: 9/9
- food: 5/9
- drink: 5/9
- energy: 9/9

Your inventory:
- stone: 3
- coal: 2
- iron: 1
- wood_pickaxe: 1
- stone_pickaxe: 1
------------ QA ------------
Question: Action
Answer:
Move East

Question: List objects in the current observation. For each object, breifly answer what resource it provides and its requirement.
Answer:
1. Grass: Provides saplings when interacted with, no specific requirement.
2. Path: Helps with navigation and movement, no specific requirement.
3. Tree: Provides wood when interacted with a wood_pickaxe or better.
4. Table: Allows crafting of tools and weapons, requires wood or better materials in inventory.
5. Stone: Provides more stone when interacted with a wood_pickaxe, can be used for crafting tools, furnaces, and weapons.
6. Wood_pickaxe: Used to interact with resources such as stone, requires crafting at a table with wood.
7. Stone_pickaxe: Used to interact with resources like coal and iron, requires crafting at a table with wood and stone.

Question: What was the last action taken by the player?
Answer:
The last action taken by the player was "do."

Question: For each object in the list, are the requirements met for interaction?
Answer:
1. Grass: Yes, the requirements are met for interaction as there is no specific requirement to interact with grass.
2. Path: Yes, the requirements are met for interaction; there are no specific requirements to navigate or move along the path.
3. Tree: Yes, the requirements are met with a wood_pickaxe in the inventory.
4. Table: Yes, the requirements are met as there are wood and stone materials in the inventory to craft tools and weapons.
5. Stone: Yes, the requirements are met with a wood_pickaxe or better (e.g., a stone_pickaxe) in the inventory.
6. Wood_pickaxe: The wood_pickaxe is already in the inventory, so there is no need to craft a new one.
7. Stone_pickaxe: The stone_pickaxe is already in the inventory, so there is no need to craft a new one.

Question: List top 3 sub-tasks the player should follow. Indicate their priority out of 5.
Answer:
1. Improve Food and Drink levels (Priority: 5/5) - The player's food and drink levels are at 5/9 each, and it's essential to maintain these levels for survival. The player should interact with grass to get saplings, grow fruits, or find animals to eat and drink water from a lake or river.

2. Craft an Iron Pickaxe (Priority: 4/5) - The player needs to collect more iron and smelt it in a furnace to craft an Iron Pickaxe. This will allow collecting higher-level resources like Diamond.

3. Build a Shelter (Priority: 3/5) - To protect the player from potential threats and provide a safe place to rest, building a shelter using the resources at hand could improve survival chances.

Question: What are the requirements for the top sub-task? What should the player do first?
Answer:
Requirements for the top sub-task (Improve Food and Drink levels):
1. Find a source of food, either by locating fruit-bearing plants or animals like cows.
2. Find a source of drink, such as a lake or river where the player can replenish their thirst.

First, the player should focus on finding a source of food. They can interact with grass to get saplings and grow fruit plants or search for animals like cows to consume for food. Once a reliable food source is found, the player should then look for a water source like a lake or river to maintain their drink levels.

Question: List top 5 actions the player should take and the requirement for each action. Choose ONLY from the list of all actions. Indicate their priority out of 5.
Answer:
1. Priority 1: Move East (Requirement: Flat ground right to the agent) - As there is no food source or water visible, the player should explore their surroundings by moving East.

2. Priority 2: Move West (Requirement: Flat ground left to the agent) - If East does not lead to any food or water source, the player can explore by moving West.

3. Priority 3: Move North (Requirement: Flat ground above the agent) - If East and West do not bring the player closer to a food or water source, moving North can help find resources.

4. Priority 4: Move South (Requirement: Flat ground below the agent) - If the first three directions yield no results, the player should try moving South.

5. Priority 5: Do (Requirement: Facing creature or material; have the necessary tool) - When the player finds a food source like a fruit-bearing plant, saplings, or a cow, or a water source, they can use the "Do" action to interact with the resource to improve their food and drink levels. The player should have the appropriate tool or inventory space to perform the action.

Question: For each action in the list, are the requirements met?
Answer:
1. Priority 1: Move East (Requirement: Flat ground right to the agent) - Yes, the requirements are met as there is no mention of any obstacle to the East.

2. Priority 2: Move West (Requirement: Flat ground left to the agent) - Yes, the requirements are met as there is grass to the West, which is flat and walkable.

3. Priority 3: Move North (Requirement: Flat ground above the agent) - Yes, the requirements are met as there is a path to the North, which is flat and walkable.

4. Priority 4: Move South (Requirement: Flat ground below the agent) - Yes, the requirements are met as there is no mention of any obstacle to the South.

5. Priority 5: Do (Requirement: Facing creature or material; have the necessary tool) - No, the requirements are currently not met, as there is no mention of a food source, water source, or any other resources the player should interact with in their immediate surroundings.

Question: Did the last player action succeed? Why?
Answer:
Yes, the last player action succeeded. The player took the "do" action while facing stone using a stone_pickaxe (which is in the inventory), allowing them to mine the stone. As a result, the stone count in the inventory increased from 2 to 3.

Question: Choose the best executable action from above.
Answer:
The best executable action from the list above is Priority 1: Move East (Requirement: Flat ground right to the agent). It meets the requirements and will allow the player to explore their surroundings in search of food and drink sources.



============Step: 115, Cumulative Reward: 0.0============
You took action move_east.

You see:
- grass 2 steps to your west
- stone 1 steps to your south
- tree 5 steps to your south-west
- table 3 steps to your south-west

You face path at your front (east).

Your status:
- health: 9/9
- food: 5/9
- drink: 4/9
- energy: 9/9

Your inventory:
- stone: 3
- coal: 2
- iron: 1
- wood_pickaxe: 1
- stone_pickaxe: 1
------------ QA ------------
Question: Action
Answer:
Move South

Question: List objects in the current observation. For each object, breifly answer what resource it provides and its requirement.
Answer:
1. Grass: Provides no resources, but it can be a flat ground for movement. Requirement: No specific requirement.
2. Stone: Provides stone resource for crafting tools, weapons, and building structures. Requirement: Wood pickaxe or a better pickaxe to collect.
3. Tree: Provides wood resource for crafting tools and building structures. Requirement: No specific tool required to collect; using hands is sufficient.
4. Table: Provides a surface for crafting tools and weapons. Requirement: Wood resource to build.
5. Path: Provides a clear direction for walking and exploring. Requirement: Flat ground for movement.

Question: What was the last action taken by the player?
Answer:
The last action taken by the player was move_east.

Question: For each object in the list, are the requirements met for interaction?
Answer:
1. Grass: Yes, the requirements are met as there is flat ground for movement.
2. Stone: Yes, the requirements are met as you have both a wood pickaxe and a stone_pickaxe in your inventory.
3. Tree: Yes, the requirements are met as no specific tool is required to collect wood; using hands is sufficient.
4. Table: Yes, the requirements are met as you've already built the table using wood resources.
5. Path: Yes, the requirements are met as there is flat ground for movement.

Question: List top 3 sub-tasks the player should follow. Indicate their priority out of 5.
Answer:
1. Priority 4/5: Refill your drink status: Find a lake or water source to drink from, as your current drink status is 4/9.
2. Priority 3/5: Collect more wood: Locate trees and collect wood, as it is a basic resource required for building and crafting.
3. Priority 3/5: Craft a furnace: Use the stone resources in your inventory to craft a furnace for smelting iron and other ores.

Question: What are the requirements for the top sub-task? What should the player do first?
Answer:
Requirements for the top sub-task (Refill your drink status):
1. Locate a water source, such as a lake or river.
2. Position yourself near the water source.

The player should first locate a water source nearby, such as a lake or river, and then move towards it to drink and replenish their drink status.

Question: List top 5 actions the player should take and the requirement for each action. Choose ONLY from the list of all actions. Indicate their priority out of 5.
Answer:
1. Priority 1: Move South (Requirement: Flat ground below the agent)
- Move towards the stone, as it will be useful for crafting and building.

2. Priority 2: Do (Requirement: Facing creature or material; have necessary tool)
- Use the stone_pickaxe in the inventory to collect more stone once you are facing it.

3. Priority 3: Move West (Requirement: Flat ground left to the agent)
- Head back towards the table after collecting stone, as it will be required for crafting.

4. Priority 4: Move toward a water source (Requirement: Flat ground in the direction of the water source)
- Locate a water source (lake or river) and move closer to it to refill drink status.

5. Priority 5: Collect Drink (Requirement: At a water source)
- Once near the water source, drink to replenish drink status.

Question: For each action in the list, are the requirements met?
Answer:
1. Priority 1: Move South (Requirement: Flat ground below the agent)
- Yes, the requirement is met. There is a flat ground below the agent, and moving south will bring the agent closer to the stone.

2. Priority 2: Do (Requirement: Facing creature or material; have necessary tool)
- No, the requirement is not met yet. The agent needs to move south first to face the stone, and then the requirement will be met since the agent has a stone_pickaxe in the inventory.

3. Priority 3: Move West (Requirement: Flat ground left to the agent)
- Yes, the requirement is met. There is a flat ground to the west of the agent, and moving west will bring the agent closer to the table.

4. Priority 4: Move toward a water source (Requirement: Flat ground in the direction of the water source)
- We cannot determine if the requirement is met, as the current observation does not mention a nearby water source. Further exploration is needed to find a water source and check if there is flat ground leading towards it.

5. Priority 5: Collect Drink (Requirement: At a water source)
- No, the requirement is not met. The agent needs to find a water source and move to its location before being able to collect a drink.

Question: Did the last player action succeed? Why?
Answer:
Yes, the last player action succeeded because there was flat ground (path) in the direction the player was facing (east), which allowed the player to move east without any obstructions or requirements unmet.

Question: Choose the best executable action from above.
Answer:
The best executable action from the list above is Priority 1: Move South, as it has met the requirement of having flat ground below the agent, and it will bring the player closer to the stone resource.



============Step: 116, Cumulative Reward: 0.0============
You took action move_south.

You see:
- grass 2 steps to your west
- path 1 steps to your west
- tree 5 steps to your south-west
- table 3 steps to your south-west

You face stone at your front (south).

Your status:
- health: 9/9
- food: 5/9
- drink: 4/9
- energy: 9/9

Your inventory:
- stone: 3
- coal: 2
- iron: 1
- wood_pickaxe: 1
- stone_pickaxe: 1
------------ QA ------------
Question: Action
Answer:
Do

Question: List objects in the current observation. For each object, breifly answer what resource it provides and its requirement.
Answer:
- Grass: Provides no direct resources but can be used to create paths or clear land. No requirements.
- Stone: Provides stone resource when mined, which can be used to craft tools and build structures. Requires a wood or stone pickaxe to mine.
- Path: Provides no resources but allows easier movement. No requirements.
- Tree: Provides wood resource when chopped, used in crafting or building. Requires a wood or stone axe (not mentioned in previous instructions) to chop.
- Table: A crafting station that allows crafting of advanced tools and items, already placed. No requirements for interaction.

Question: What was the last action taken by the player?
Answer:
The last action taken by the player was move_south.

Question: For each object in the list, are the requirements met for interaction?
Answer:
- Grass: No requirements for interaction, so the requirements are met.
- Stone: Requires a wood or stone pickaxe to mine. You have both a wood_pickaxe and a stone_pickaxe in your inventory; thus, the requirements are met.
- Path: No requirements for interaction, so the requirements are met.
- Tree: Requires a wood or stone axe (not mentioned in previous instructions) to chop. Currently, you don't have an axe in your inventory; thus, the requirements are not met.
- Table: No requirements for interaction, so the requirements are met.

Question: List top 3 sub-tasks the player should follow. Indicate their priority out of 5.
Answer:
1. Priority 4/5: Collect more resources - Mine the stone to increase your stone resources. You already have a wood_pickaxe and a stone_pickaxe, which meet the requirement for mining stone.

2. Priority 5/5: Replenish food and drink - Your food and drink levels are 5/9 and 4/9, respectively. Find a source of food and drink, like a cow or a lake, to maintain a healthy state.

3. Priority 3/5: Craft additional tools and weapons - Use the table in proximity to craft more tools or weapons (e.g., an iron pickaxe or iron sword), which can help you acquire more valuable resources and defend against threats. You might need to collect more iron and wood for crafting.

Question: What are the requirements for the top sub-task? What should the player do first?
Answer:
For the top sub-task, "Collect more resources," the requirements are:

1. Having an appropriate tool to mine the stone: Since the player already has a wood_pickaxe and a stone_pickaxe in their inventory, this requirement is met.

The player should first use either their wood_pickaxe or stone_pickaxe to mine the stone by facing the stone and taking the action "Do." This action will help increase the player's stone resources, which can later be used for crafting items or building structures.

Question: List top 5 actions the player should take and the requirement for each action. Choose ONLY from the list of all actions. Indicate their priority out of 5.
Answer:
1. Priority 1: Do (Mine the stone) - Requirement: Facing the stone and having the necessary tool (wood_pickaxe or stone_pickaxe) in the inventory.

2. Priority 2: Move West (Towards the table) - Requirement: Flat ground to the west of the agent.

3. Priority 3: Make Iron Pickaxe - Requirement: Nearby table, furnace, and having wood, coal, and iron in the inventory.

4. Priority 4: Move North (Explore the area) - Requirement: Flat ground above the agent.

5. Priority 5: Move South (Explore the area) - Requirement: Flat ground below the agent.

Question: For each action in the list, are the requirements met?
Answer:
1. Do (Mine the stone) - Yes, requirements are met as the player is facing the stone and has a stone_pickaxe in their inventory.
 
2. Move West (Towards the table) - Yes, requirements are met as there is a flat ground (path) to the west of the agent.

3. Make Iron Pickaxe - No, requirements are not met since the player is not near a table or a furnace.

4. Move North (Explore the area) - No, requirements are not met as the player is facing the stone at their front (south).

5. Move South (Explore the area) - Yes, requirements are met as there is a flat ground (stone) below the agent.

Question: Did the last player action succeed? Why?
Answer:
Yes, the last player action to move_south succeeded because there was flat ground in the south direction, allowing the player to move there.

Question: Choose the best executable action from above.
Answer:
The best executable action from the list is to "Do (Mine the stone)" since the player is facing the stone and has the necessary tool (stone_pickaxe) in their inventory. This action meets all the requirements and would help the player collect more resources for crafting or building.



============Step: 117, Cumulative Reward: 0.0============
You took action do.

You see:
- grass 2 steps to your west
- stone 2 steps to your north
- tree 5 steps to your south-west
- table 3 steps to your south-west

You face path at your front (south).

Your status:
- health: 9/9
- food: 5/9
- drink: 4/9
- energy: 9/9

Your inventory:
- stone: 4
- coal: 2
- iron: 1
- wood_pickaxe: 1
- stone_pickaxe: 1
------------ QA ------------
Question: Action
Answer:
Move West

Question: List objects in the current observation. For each object, breifly answer what resource it provides and its requirement.
Answer:
1. Grass: Provides no resources directly but can be used as flat ground for movement. No requirement to interact with it.
2. Path: Provides no resources directly, it is flat ground for movement. No requirement to interact with it.
3. Tree: Provides wood resource which can be used to craft tools, weapons, and other items. Requires wood_pickaxe to be collected.
4. Table: Provides a surface for crafting tools and weapons. No additional resources can be obtained from it directly. No requirement to interact with it.
5. Stone (collected): Provides stone resource which can be used to craft tools, weapons, and other items. Requires wood_pickaxe or stone_pickaxe to be collected.

Additionally, since you have collected the stone during Player Observation Step 234, in your inventory, you have:
6. Coal: Provides fuel for smelting materials in a furnace. Requires wood_pickaxe or stone_pickaxe to be collected.
7. Iron: Provides a resource for crafting advanced tools and weapons. Requires stone_pickaxe to be collected.
8. Wood pickaxe: Provides a tool for collecting wood and stone resources. Requires wood and a table for crafting.
9. Stone_pickaxe: Provides a tool for collecting stone, coal, and iron resources. Requires wood, stone, and a table for crafting.

Question: What was the last action taken by the player?
Answer:
The last action taken by the player was "do."

Question: For each object in the list, are the requirements met for interaction?
Answer:
1. Grass: No requirements for interaction, so it's met.
2. Path: No requirements for interaction, so it's met.
3. Tree: Requires a wood_pickaxe to collect the wood resource. You have a wood_pickaxe in your inventory, so the requirement is met.
4. Table: No requirements for interaction, so it's met.
5. Stone (collected): Requires a wood_pickaxe or stone_pickaxe to collect. You have both a wood_pickaxe and stone_pickaxe in your inventory, so the requirement was met when you collected it.

For the items in your inventory:
6. Coal: Requires a wood_pickaxe or stone_pickaxe to collect. You have both a wood_pickaxe and stone_pickaxe in your inventory, so the requirement was met when you collected it.
7. Iron: Requires a stone_pickaxe to collect. You have a stone_pickaxe in your inventory, so the requirement was met when you collected it.
8. Wood pickaxe: You've already crafted this, so the requirement of wood and a table was met previously.
9. Stone_pickaxe: You've already crafted this, so the requirement of wood, stone, and a table was met previously.

Question: List top 3 sub-tasks the player should follow. Indicate their priority out of 5.
Answer:
1. Priority 5/5: Increase food and drink levels - Find a source of food (e.g., cows, plants, or fruit) and water (e.g., lakes or rivers) to restore your food (5/9) and drink (4/9) levels, so you can maintain health and energy.

2. Priority 4/5: Craft a furnace - You already have sufficient stone (4) in your inventory, and crafting a furnace will allow you to smelt iron and produce advanced tools and weapons, such as an iron pickaxe and iron sword.

3. Priority 3/5: Collect more iron and coal - To make the most out of crafting a furnace, gather more iron and coal using your stone_pickaxe. This will allow you to create advanced tools and weapons for improved efficiency and combat capabilities.

Question: What are the requirements for the top sub-task? What should the player do first?
Answer:
Requirements for the top sub-task (Increase food and drink levels):

1. Locate a source of food: Look for cows, plants, or fruit that can be collected and consumed to restore your food levels.

2. Locate a source of water: Find a lake or river where you can drink water to restore your drink levels.

What should the player do first:

Since both food and drink levels are essential for maintaining good health and energy, the player should start by locating a nearby source of food or water. If the player sees a cow, plant, or fruit in their vicinity, they should prioritize moving towards it to collect food. If there's a lake or river nearby, the player should move towards it to drink water. If multiple sources are available, the player can choose based on proximity and the ease of reaching them.

Question: List top 5 actions the player should take and the requirement for each action. Choose ONLY from the list of all actions. Indicate their priority out of 5.
Answer:
1. Priority 1: Move West (Requirement: Flat ground left to the agent) - Moving west will bring the player closer to the grass, which may provide food (plants or fruit) to restore their food levels.

2. Priority 2: Do (Requirement: Facing creature or material; have the necessary tool) - If the player encounters a cow or another source of food, they can perform 'Do' to collect food.

3. Priority 3: Move South (Requirement: Flat ground below the agent) - Moving south will bring the player closer to the tree, which may also provide food (fruit) or wood for crafting purposes.

4. Priority 4: Drink (Requirement: Facing a water source, such as a lake or river) - If the player comes across a water source, they should drink from it to restore their drink levels.

5. Priority 5: Sleep (Requirement: Energy level is below maximum) - While the current energy level is at maximum, it's essential to remember to sleep when the energy level decreases, ensuring good health and productivity.

Question: For each action in the list, are the requirements met?
Answer:
1. Move West: Yes, the requirement for flat ground to the left of the agent is met since there's a path 1 step to the west.

2. Do: Currently, the player is not facing any creature or material like a cow or plant required for this action, so the requirement is not met.

3. Move South: Yes, the requirement for flat ground below the agent is met since there's a path to the south of the player.

4. Drink: The player's observation does not mention any water sources in sight, so the requirement for the 'Drink' action is not met.

5. Sleep: No, the requirement is not met since the player's energy level is currently at the maximum (9/9).

Question: Did the last player action succeed? Why?
Answer:
Yes, the last player action succeeded. The player was facing a stone at their front (south) and used the "do" action, which resulted in successfully collecting the stone. This can be confirmed by the increase in the stone count in the inventory from 3 to 4.

Question: Choose the best executable action from above.
Answer:
The best executable action from the above list is "Move West" since its requirement of having flat ground to the left of the agent is met.



============Step: 118, Cumulative Reward: 0.0============
You took action move_west.

You see:
- stone 2 steps to your south
- path 1 steps to your north
- tree 4 steps to your south-west
- table 2 steps to your south-west

You face grass at your front (west).

Your status:
- health: 9/9
- food: 5/9
- drink: 4/9
- energy: 9/9

Your inventory:
- stone: 4
- coal: 2
- iron: 1
- wood_pickaxe: 1
- stone_pickaxe: 1
------------ QA ------------
Question: Action
Answer:
Move West

Question: List objects in the current observation. For each object, breifly answer what resource it provides and its requirement.
Answer:
1. Stone: Provides stone resource; requires a wood_pickaxe or stone_pickaxe for collection.
2. Path: No resource provided; requires no action as it's open ground for movement.
3. Tree: Provides wood resource; requires a wood_pickaxe or stone_pickaxe for collection.
4. Table: Allows crafting of tools and weapons; requires wood in inventory for crafting.
5. Grass: No immediate resource provided; can be cleared with a pickaxe or dug up, used for pathways or creating flat ground for construction.

Question: What was the last action taken by the player?
Answer:
The last action taken by the player was move_west.

Question: For each object in the list, are the requirements met for interaction?
Answer:
1. Stone: Yes, the requirements for interacting with the stone are met, as the player has both a wood_pickaxe and a stone_pickaxe in their inventory.
2. Path: Yes, the requirements for interacting with the path are met, as it is open ground and doesn't require any special tools for movement.
3. Tree: Yes, the requirements for interacting with the tree are met, as the player has both a wood_pickaxe and a stone_pickaxe in their inventory.
4. Table: Yes, the requirements for interacting with the table are met, as the player has wood and stone in their inventory, which can be used for crafting tools and weapons.
5. Grass: Yes, the requirements for interacting with the grass are met, as the player has both a wood_pickaxe and a stone_pickaxe in their inventory, which can be used to clear the grass or dig it up.

Question: List top 3 sub-tasks the player should follow. Indicate their priority out of 5.
Answer:
1. Improve food and drink levels (Priority: 5/5): The player's food level is at 5/9, and their drink level is at 4/9. To avoid depletion, find and consume food and drink resources to replenish these levels.

2. Collect more wood (Priority: 4/5): Wood is essential for crafting tools, weapons, and shelter. The player should interact with nearby trees using their wood_pickaxe or stone_pickaxe to collect more wood.

3. Craft an iron pickaxe or iron sword (Priority: 3/5): The player has some iron in their inventory. Crafting an iron pickaxe or iron sword requires more iron, coal, and a furnace. This task involves locating additional iron and coal, as well as creating a furnace if not nearby. The iron tools will help progress further in the game and provide better defense.

Question: What are the requirements for the top sub-task? What should the player do first?
Answer:
Top Sub-task: Improve food and drink levels

Requirements:
1. Find a source of food, such as animals or plants that can be consumed.
2. Find a source of drink, such as a lake, river or other water body.

What should the player do first?
1. The player should prioritize finding sources of food and drink nearby. They can start by looking for animals like cows or fruit-bearing plants in their surroundings and collecting them.
2. Then, the player should consume the food and drink they have collected to replenish their levels and maintain their survival in the game.

Question: List top 5 actions the player should take and the requirement for each action. Choose ONLY from the list of all actions. Indicate their priority out of 5.
Answer:
Top 5 actions the player should take, along with their requirements and priority:

1. Priority 1: Move West
   Requirement: Flat ground left to the agent
   Reason: There is grass to the west, and the player may find more food and drink resources in the vicinity.

2. Priority 2: Move South
   Requirement: Flat ground below the agent
   Reason: The player will likely find food resources like fruit-bearing plants and animals near the trees in the south-west.

3. Priority 3: Do (if facing a food source)
   Requirement: Facing creature or material; have necessary tool
   Reason: If the player finds a food source like a cow or fruit, they should interact with it to obtain the food.

4. Priority 4: Do (if facing a water body)
   Requirement: Facing a water body; no necessary tool
   Reason: If the player finds a water body like a lake or river, they should interact with it to obtain drink.

5. Priority 5: Noop
   Requirement: Always applicable
   Reason: This action will skip an agent's turn, applies when none of the other actions are possible or necessary.

Question: For each action in the list, are the requirements met?
Answer:
For the provided player observation, let's determine if the requirements for each action in the list are met:

1. Priority 1: Move West
   Requirement: Flat ground left to the agent
   Status: Met (there is grass 1 step to the west)

2. Priority 2: Move South
   Requirement: Flat ground below the agent
   Status: Met (there is a path 1 step to the south)

3. Priority 3: Do (if facing a food source)
   Requirement: Facing creature or material; have necessary tool
   Status: Not met (the player is facing grass, not a food source)

4. Priority 4: Do (if facing a water body)
   Requirement: Facing a water body; no necessary tool
   Status: Not met (the player is facing grass, not a water body)

5. Priority 5: Noop
   Requirement: Always applicable
   Status: Met (noop can always be used if other actions aren't necessary or possible)

Question: Did the last player action succeed? Why?
Answer:
Yes, the last player action (move_west) succeeded. The action was successful because the player was able to move from their initial position to a new position one step to the west, as indicated by the change in their surroundings and the updated observation.

Question: Choose the best executable action from above.
Answer:
The best executable action from the list above is Priority 1: Move West, since its requirement (flat ground left to the agent) is met and has the highest priority among the possible actions.



============Step: 119, Cumulative Reward: 0.0============
You took action move_west.

You see:
- stone 3 steps to your south-east
- path 1 steps to your north
- tree 3 steps to your south
- table 1 steps to your south

You face grass at your front (west).

Your status:
- health: 9/9
- food: 5/9
- drink: 4/9
- energy: 9/9

Your inventory:
- stone: 4
- coal: 2
- iron: 1
- wood_pickaxe: 1
- stone_pickaxe: 1
------------ QA ------------
Question: Action
Answer:
Move South

Question: List objects in the current observation. For each object, breifly answer what resource it provides and its requirement.
Answer:
1. Stone: Provides stone resource, requires a wood or stone pickaxe to collect.
2. Path: Provides a passable route, no requirements to traverse.
3. Tree: Provides wood resource, can be collected with an axe or a wood or stone pickaxe.
4. Table: Allows crafting of tools and weapons, requires wood to craft and place.
5. Grass: Provides no resource, can be walked on with no requirements.

Note: The observation also mentions the player's inventory and status.

Question: What was the last action taken by the player?
Answer:
The last action taken by the player was move_west.

Question: For each object in the list, are the requirements met for interaction?
Answer:
1. Stone: Yes, the requirements are met as the player has a wood_pickaxe and a stone_pickaxe in their inventory.
2. Path: Yes, the requirements are met as the player can simply move on the path without any additional tools or resources needed.
3. Tree: Partially, the player has a wood_pickaxe and a stone_pickaxe which can be used to collect wood, but ideally, an axe would be more efficient. However, the player does not have an axe currently in their inventory.
4. Table: Yes, the requirements are met as the player has already crafted and placed a table within proximity (1 step to the south).
5. Grass: Yes, the requirements are met as the player can walk on grass without any additional tools or resources needed.

Question: List top 3 sub-tasks the player should follow. Indicate their priority out of 5.
Answer:
1. Collect more wood (Priority: 4/5) - Wood is essential for crafting tools and building structures. The player currently doesn't have wood in their inventory, so it's an important task.

2. Drink water (Priority: 3/5) - The player's drink level is 4 out of 9, which means they should prioritize finding a water source like a lake or river to refill their drink level soon.

3. Find and collect more iron (Priority: 3/5) - The player only has one iron in their inventory, and they'll need more to craft better tools and weapons such as an iron pickaxe and an iron sword.

Question: What are the requirements for the top sub-task? What should the player do first?
Answer:
Top sub-task: Collect more wood

Requirements:
1. Locate a tree.
2. Use an appropriate tool (wood_pickaxe or stone_pickaxe) in the inventory.
3. Carry out the action to collect wood.

The player should first locate a tree. In the recent observation, there is a tree 3 steps to the player's south. The player should move towards it to collect more wood.

Question: List top 5 actions the player should take and the requirement for each action. Choose ONLY from the list of all actions. Indicate their priority out of 5.
Answer:
1. Priority 1: Move South (towards the tree)
Requirement: Flat ground below the agent.

2. Priority 2: Move South (continue moving towards the tree)
Requirement: Flat ground below the agent.

3. Priority 3: Do (collect wood from the tree)
Requirement: Facing the tree; having a wood_pickaxe or stone_pickaxe in the inventory.

4. Priority 4: Locate water source (increase drink)
Requirement: Flat ground in any direction and visibility to search for a water source.

5. Priority 5: Collect Drink (drink water to restore thirst)
Requirement: Facing a water source.

Question: For each action in the list, are the requirements met?
Answer:
1. Priority 1: Move South (towards the tree)
Requirement: Flat ground below the agent.
Yes, the requirement is met.

2. Priority 2: Move South (continue moving towards the tree)
Requirement: Flat ground below the agent.
Yes, the requirement is met after completing the first action.

3. Priority 3: Do (collect wood from the tree)
Requirement: Facing the tree; having a wood_pickaxe or stone_pickaxe in the inventory.
Yes, the requirements are met after completing the first two actions, as the player has a wood_pickaxe and a stone_pickaxe in the inventory.

4. Priority 4: Locate water source (increase drink)
Requirement: Flat ground in any direction and visibility to search for a water source.
Based on the current observation (steps 236 and 238), the requirement cannot be determined since a water source is not currently visible.

5. Priority 5: Collect Drink (drink water to restore thirst)
Requirement: Facing a water source.
Currently, the requirement is not met as there is no water source visible in the player's observations.

Question: Did the last player action succeed? Why?
Answer:
Yes, the last player action (move_west) succeeded because there was flat ground (grass) to the west of the player, allowing them to move in that direction.

Question: Choose the best executable action from above.
Answer:
The best executable action from the list is Priority 1: Move South (towards the tree), as its requirement (flat ground below the agent) is met.



============Step: 120, Cumulative Reward: 0.0============
You took action move_south.

You see:
- grass 1 steps to your west
- stone 3 steps to your south-east
- path 1 steps to your north
- tree 3 steps to your south

You face table at your front (south).

Your status:
- health: 9/9
- food: 5/9
- drink: 4/9
- energy: 9/9

Your inventory:
- stone: 4
- coal: 2
- iron: 1
- wood_pickaxe: 1
- stone_pickaxe: 1
------------ QA ------------
Question: Action
Answer:
Move West

Question: List objects in the current observation. For each object, breifly answer what resource it provides and its requirement.
Answer:
1. Stone: Provides stone resource, requires a wood pickaxe or better for collecting.
2. Path: Provides a clear and flat area to move through, no requirements.
3. Tree: Provides wood resource, requires an appropriate tool like an axe to collect, but can also be collected by hand albeit slower.
4. Grass: Provides no resources but can be cleared for building or farming, no requirements.
5. Table: Provides the ability to craft advanced tools, weapons, and items, requires wood for crafting.

Question: What was the last action taken by the player?
Answer:
The last action taken by the player was move_south.

Question: For each object in the list, are the requirements met for interaction?
Answer:
1. Stone: Yes, the requirements are met for interaction since you have both a wood_pickaxe and a stone_pickaxe in your inventory.
2. Path: Yes, the requirements are met for interaction as there are no specific requirements for moving through a path.
3. Tree: Yes, the requirements are met for interaction since there are no absolute requirements to collect wood. However, having an axe would make the process more efficient.
4. Grass: Yes, the requirements are met for interaction as there are no requirements for clearing grass.
5. Table: Yes, the requirements are met for interaction since you have wood, stone, coal, and iron in your inventory, which can be used to craft tools, weapons, and items at the table.

Question: List top 3 sub-tasks the player should follow. Indicate their priority out of 5.
Answer:
1. Priority 5/5: Replenish your drink and food levels by finding a water source to drink from and food resources like plants or animals to eat. Your current food level is 5/9 and drink level is 4/9.

2. Priority 4/5: Craft items at the nearby table using your available resources to improve your tools, weapons, or build helpful objects (e.g., an iron pickaxe or iron sword). 

3. Priority 3/5: Collect more resources such as wood, stone, and iron to expand your inventory, improve your tools, and gain the ability to craft more advanced items. Be cautious of your energy levels and prioritize rest if needed.

Question: What are the requirements for the top sub-task? What should the player do first?
Answer:
Top sub-task requirements: Replenish your drink and food levels.

1. Find a water source: Explore the surrounding area to locate a lake, river or another source of water. When you're near a water source, interact with it (use the action "Collect Drink") to restore your drink level.

2. Find food resources: Look for fruit-bearing plants, or hunt animals like cows which can be eaten when defeated. When you have food items in your inventory, use the actions "Place Plant" and "Eat Plant," or "Eat Cow" to restore your food level.

First steps: Start by looking for a water source in your vicinity (move in any of the four directions). As you explore, keep an eye out for food resources. Once you locate a water source, interact with it to collect drink and restore your thirst level. Many of the actions that involve movement and exploration can be carried out simultaneously, so exploring the surroundings should be the first priority.

Question: List top 5 actions the player should take and the requirement for each action. Choose ONLY from the list of all actions. Indicate their priority out of 5.
Answer:
1. Priority 1: Move (in any direction: West, East, North, South) - Requirement: Flat ground in the chosen direction. This will help you explore the surrounding area and locate a water source and food resources.

2. Priority 2: Collect Drink - Requirement: A water source like a lake or river nearby. Quench your thirst by interacting with the water source to increase your drink level.

3. Priority 3: Collect Sampling or Defeat Cow - Requirement: Finding fruit-bearing plants or encountering cows. Collect saplings from plants or defeat cows to obtain food resources.

4. Priority 4: Place Plant or Eat Cow - Requirement: Sapling or cow meat in the inventory. Use the obtained food resources from the previous action to replenish your food level.

5. Priority 5: Eat Plant - Requirement: Placed plant in your vicinity. Consume the growing fruits to further satisfy your food needs and ensure your survival in the game.

Question: For each action in the list, are the requirements met?
Answer:
Based on the player's most recent observation (Step 240), let's analyze if the requirements are met for the top 5 actions:

1. Move (in any direction: West, East, North, South): Yes, the requirement of flat ground in the chosen direction is met. The player can move West to grass, North to a path, or East from their current position.

2. Collect Drink: No, the current observation does not mention any water sources like lakes or rivers nearby. The player may need to explore more to find a water source.

3. Collect Sampling or Defeat Cow: No, the current observation does not mention fruit-bearing plants or cows nearby. The player may need to explore more to find food resources.

4. Place Plant or Eat Cow: No. The player does not have any saplings or cow meat in their inventory.

5. Eat Plant: No, the current observation does not mention any placed plants in the vicinity for the player to consume fruits.

Question: Did the last player action succeed? Why?
Answer:
Yes, the last player action (move_south) succeeded because there was a flat ground (table) available in front of the player at the south direction, and the player was able to move there without any restrictions or obstacles.

Question: Choose the best executable action from above.
Answer:
The best executable action from the list above is:

1. Move (in any direction: West, East, North, South)

As the requirements for this action are met, the player can choose to move West towards grass, North to a path, or East from their current position to explore the surroundings and potentially locate resources.



============Step: 121, Cumulative Reward: 0.0============
You took action move_west.

You see:
- stone 4 steps to your south-east
- path 2 steps to your north-east
- tree 4 steps to your north-west
- table 2 steps to your south-east

You face grass at your front (west).

Your status:
- health: 9/9
- food: 5/9
- drink: 4/9
- energy: 9/9

Your inventory:
- stone: 4
- coal: 2
- iron: 1
- wood_pickaxe: 1
- stone_pickaxe: 1
------------ QA ------------
Question: Action
Answer:
Move North

Question: List objects in the current observation. For each object, breifly answer what resource it provides and its requirement.
Answer:
1. Stone: Provides stone resources, requires a wood pickaxe or better to collect.
2. Path: Provides ease of movement, no requirements to interact.
3. Tree: Provides wood resources, requires an axe or tool capable of chopping wood.
4. Table: Provides a crafting surface, requires wood to craft and place.
5. Grass: Aesthetic element, no specific resource or requirement.
6. Inventory (items): 
   - Stone: Used to craft tools and structures, requires collecting with a pickaxe.
   - Coal: Used as fuel for furnaces, requires collecting with a pickaxe.
   - Iron: Used to craft advanced tools and items, requires collecting with a stone pickaxe and smelting.
   - Wood_pickaxe: Tool for collecting stone resources, requires wood to craft.
   - Stone_pickaxe: Tool for collecting iron resources, requires stone and wood to craft.

Question: What was the last action taken by the player?
Answer:
The last action taken by the player was move_west.

Question: For each object in the list, are the requirements met for interaction?
Answer:
1. Stone: Yes, you have a wood_pickaxe and a stone_pickaxe in your inventory, so you can collect stone.
2. Path: Yes, no specific requirements are needed to interact with the path.
3. Tree: No, you don't have an axe or a tool capable of chopping wood in your inventory.
4. Table: Yes, the table is already placed, and you can use it for crafting.
5. Grass: Yes, though there is no specific interaction or resource provided by grass.
6. Inventory (items): 
   - Stone: Yes, you have already collected stone.
   - Coal: Yes, you have already collected coal.
   - Iron: Yes, you have already collected iron.
   - Wood_pickaxe: Yes, you have a wood_pickaxe in your inventory.
   - Stone_pickaxe: Yes, you have a stone_pickaxe in your inventory.

Question: List top 3 sub-tasks the player should follow. Indicate their priority out of 5.
Answer:
1. Collect wood (Priority: 5/5) - Collecting wood is essential for crafting tools, weapons, and building materials. You are currently lacking an axe, so this should be a top priority.

2. Find water source (Priority: 4/5) - Your drink level is 4/9, so finding a water source to replenish your thirst will be important to avoid dehydration.

3. Craft and consume food (Priority: 4/5) - Your food level is 5/9, and securing a food source and raising your food level will help maintain your health and energy levels. Crafting or finding food should be a priority.

Question: What are the requirements for the top sub-task? What should the player do first?
Answer:
For the top sub-task (Collect wood):

Requirements:
- Locate a tree to collect wood.
- Ensure the wood_pickaxe or another compatible tool is in the inventory or hand.

What should the player do first:
- The player should first locate a tree nearby to collect wood from. In the observation, it was mentioned that there's a tree 4 steps to their north-west. The player can start by moving in that direction to reach the tree.

Question: List top 5 actions the player should take and the requirement for each action. Choose ONLY from the list of all actions. Indicate their priority out of 5.
Answer:
1. Priority: 1 - Action: Move North
   Requirement: Flat ground above of the agent.
   
2. Priority: 2 - Action: Move West
   Requirement: Flat ground left to the agent.

3. Priority: 3 - Action: Move West (Repeat once more to reach the tree)
   Requirement: Flat ground left to the agent.

4. Priority: 4 - Action: Do (Collect wood from the tree)
   Requirement: Facing the tree; Having a wood_pickaxe or another compatible tool.

5. Priority: 5 - Action: Move East / Move South (Returning to the table to craft items)
   Requirement: Flat ground right to the agent (for Move East) or Flat ground below the agent (for Move South).

Question: For each action in the list, are the requirements met?
Answer:
1. Priority: 1 - Action: Move North
   Requirement: Flat ground above of the agent.
   Status: Met (There is a path 2 steps to the north-east)

2. Priority: 2 - Action: Move West
   Requirement: Flat ground left to the agent.
   Status: Met (There is grass at the agent's front)

3. Priority: 3 - Action: Move West (Repeat once more to reach the tree)
   Requirement: Flat ground left to the agent.
   Status: Met (Assuming there is flat ground continuing westward toward the tree)

4. Priority: 4 - Action: Do (Collect wood from the tree)
   Requirement: Facing the tree; Having a wood_pickaxe or another compatible tool.
   Status: Met (The agent has a wood_pickaxe in their inventory)

5. Priority: 5 - Action: Move East / Move South (Returning to the table to craft items)
   Requirement: Flat ground right to the agent (for Move East) or Flat ground below the agent (for Move South).
   Status: Met (There is a path 2 steps to the north-east and a table 2 steps to the south-east)

Question: Did the last player action succeed? Why?
Answer:
Yes, the last player action (move_west) succeeded because there was flat ground (grass) available to the west of the player's position, allowing them to move in that direction.

Question: Choose the best executable action from above.
Answer:
The best executable action from the list above is:

Priority: 1 - Action: Move North
Requirement: Flat ground above of the agent.



============Step: 122, Cumulative Reward: 0.0============
You took action move_north.

You see:
- stone 4 steps to your north-east
- path 1 steps to your east
- tree 3 steps to your north-west
- table 3 steps to your south-east

You face grass at your front (north).

Your status:
- health: 9/9
- food: 5/9
- drink: 4/9
- energy: 9/9

Your inventory:
- stone: 4
- coal: 2
- iron: 1
- wood_pickaxe: 1
- stone_pickaxe: 1
------------ QA ------------
Question: Action
Answer:
Move East

Question: List objects in the current observation. For each object, breifly answer what resource it provides and its requirement.
Answer:
1. Stone: Provides stone resource, requires a wood_pickaxe or better to collect.
2. Path: Provides a flat ground to walk on, requires no interaction.
3. Tree: Provides wood resource, requires a wood_pickaxe or better to collect.
4. Table: Allows crafting tools and weapons, requires wood to craft.
5. Grass (in front): Can be walked on, no resources, requires no interaction.

Question: What was the last action taken by the player?
Answer:
The last action taken by the player was "move_north."

Question: For each object in the list, are the requirements met for interaction?
Answer:
1. Stone: Yes, the player has a wood_pickaxe and a stone_pickaxe for collecting stone.
2. Path: Yes, the player can walk on the path without any additional requirements.
3. Tree: Yes, the player has a wood_pickaxe for collecting wood.
4. Table: Yes, the player already has a table nearby and can use it for crafting.
5. Grass (in front): Yes, the player can walk on the grass without any additional requirements.

Question: List top 3 sub-tasks the player should follow. Indicate their priority out of 5.
Answer:
1. Collect more resources (Priority 4/5): Focus on gathering more stone, wood, coal, and iron to improve tools, weapons, and create new structures.

2. Increase food and drink levels (Priority 5/5): Player's food and drink levels are low. Search for sources of food, such as cows, and sources of water, such as lakes, to replenish food and drink levels.

3. Craft better tools and weapons (Priority 3/5): Use the collected resources to upgrade tools (e.g., crafting an iron pickaxe) and weapons (e.g., crafting an iron sword) at the nearby table once enough resources are gathered.

Question: What are the requirements for the top sub-task? What should the player do first?
Answer:
For the top sub-task (Collect more resources), the requirements are:

- Wood: Use a wood_pickaxe or stone_pickaxe to interact with trees.
- Stone: Use the stone_pickaxe to interact with stone blocks.
- Coal: Use the stone_pickaxe to interact with coal deposits.
- Iron: Use the stone_pickaxe to interact with iron deposits.

The first action the player should do is move towards the nearest resources to gather them. Since there's stone 4 steps to the north-east and a tree 3 steps to the north-west, the player should head to either resource and use the appropriate pickaxe to collect it.

Question: List top 5 actions the player should take and the requirement for each action. Choose ONLY from the list of all actions. Indicate their priority out of 5.
Answer:
1. Priority: Move East (5/5)
   Requirement: Flat ground to the east of the agent.
   
2. Priority: Move North-East (4/5)
   Requirement: Flat ground in the north-east direction from the agent.

3. Priority: Do (Use stone_pickaxe on stone) (3/5)
   Requirement: Facing the stone block and having a stone_pickaxe equipped.

4. Priority: Do (Use stone_pickaxe on tree) (2/5)
   Requirement: Facing the tree and having a stone_pickaxe equipped.

5. Priority: Move North (1/5)
   Requirement: Flat ground to the north of the agent.

Question: For each action in the list, are the requirements met?
Answer:
1. Move East (5/5)
   Requirement: Flat ground to the east of the agent.
   Status: Yes, the requirement is met. There is a path 1 step to the east.

2. Move North-East (4/5)
   Requirement: Flat ground in the north-east direction from the agent.
   Status: No, the requirement is not met. There is no mention of flat ground in the north-east direction.

3. Do (Use stone_pickaxe on stone) (3/5)
   Requirement: Facing the stone block and having a stone_pickaxe equipped.
   Status: No, the requirement is not met. The agent is not currently facing the stone block.

4. Do (Use stone_pickaxe on tree) (2/5)
   Requirement: Facing the tree and having a stone_pickaxe equipped.
   Status: No, the requirement is not met. The agent is not currently facing the tree.

5. Move North (1/5)
   Requirement: Flat ground to the north of the agent.
   Status: Yes, the requirement is met. The agent is currently facing grass at their front (north).

Question: Did the last player action succeed? Why?
Answer:
Yes, the last player action "move_north" succeeded because there was flat ground ("grass") to the north, allowing the player to move in that direction without any obstacles.

Question: Choose the best executable action from above.
Answer:
The best executable action from the given list is "Move East" (priority 5/5) since it has a flat ground (path) to the east of the agent, and the requirement for this action is met.



============Step: 123, Cumulative Reward: 0.0============
You took action move_east.

You see:
- grass 1 steps to your west
- stone 3 steps to your north-east
- tree 4 steps to your north-west
- table 2 steps to your south

You face path at your front (east).

Your status:
- health: 9/9
- food: 5/9
- drink: 4/9
- energy: 9/9

Your inventory:
- stone: 4
- coal: 2
- iron: 1
- wood_pickaxe: 1
- stone_pickaxe: 1
------------ QA ------------
Question: Action
Answer:
Move North

Question: List objects in the current observation. For each object, breifly answer what resource it provides and its requirement.
Answer:
1. Grass: 
- Resource it provides: Saplings (when interacting with grass, there's a chance to find saplings)
- Requirement: Interact (no specific tool required)

2. Stone:
- Resource it provides: Stone (basic crafting material)
- Requirement: Interact with a Wood or Stone Pickaxe

3. Path:
- Resource it provides: None (serves as a way to easily move around)
- Requirement: None

4. Tree:
- Resource it provides: Wood (basic crafting material)
- Requirement: Interact with a Wood or Stone Pickaxe

5. Table:
- Resource it provides: Crafting station (allows crafting of tools and weapons)
- Requirement: Interact (no specific tool required)

Question: What was the last action taken by the player?
Answer:
The last action taken by the player was "move_east."

Question: For each object in the list, are the requirements met for interaction?
Answer:
Based on your current inventory and player status, let's check if the requirements for interaction are met for each object:

1. Grass:
- Requirement: Interact (no specific tool required)
- Requirements met: Yes

2. Stone:
- Requirement: Interact with a Wood or Stone Pickaxe
- Requirements met: Yes, you have both a Wood and Stone Pickaxe in your inventory.

3. Path:
- Requirement: None
- Requirements met: Yes, there are no specific requirements needed to move along the path.

4. Tree:
- Requirement: Interact with a Wood or Stone Pickaxe
- Requirements met: Yes, you have both a Wood and Stone Pickaxe in your inventory.

5. Table:
- Requirement: Interact (no specific tool required)
- Requirements met: Yes

Question: List top 3 sub-tasks the player should follow. Indicate their priority out of 5.
Answer:
1. Replenish drink (priority: 5/5): Your current drink level is 4/9, and it's crucial to ensure your survival. Find a water source or collect drinks to increase your drink level.

2. Replenish food (priority: 4/5): Your current food level is 5/9. To maintain high energy and health, you should find food resources, such as collecting fruits, hunting animals, or growing crops to increase your food level.

3. Collect more materials (priority: 3/5): Continue gathering resources like wood and iron to prepare for crafting more advanced tools or constructing shelters. This would also help in the defense against monsters and make it easier to achieve game objectives.

Question: What are the requirements for the top sub-task? What should the player do first?
Answer:
Top Sub-task: Replenish drink (priority: 5/5)

Requirements: 
1. Find a water source (lake, river, etc.) or collect drink items (water bottles, fresh fruits, etc.)
2. Inventory space to store the collected drinks, if necessary

What should the player do first:
The player should search for a nearby water source (lake or river) to drink directly from it, or alternatively, look for drink items such as water bottles and fresh fruits to consume and replenish their drink level.

Question: List top 5 actions the player should take and the requirement for each action. Choose ONLY from the list of all actions. Indicate their priority out of 5.
Answer:
1. Move North (Priority: 5/5)
   Requirement: Flat ground above of the agent.

2. Move South (Priority: 4/5)
   Requirement: Flat ground below the agent.

3. Move East (Priority: 3/5)
   Requirement: Flat ground right to the agent.

4. Move West (Priority: 2/5)
   Requirement: Flat ground left to the agent.

5. Do (Priority: 1/5)
   Requirement: Facing creature or material; have necessary tool.

Question: For each action in the list, are the requirements met?
Answer:
1. Move North (Priority: 5/5)
   Requirement: Flat ground above of the agent.
   Yes, the requirement is met: You face grass at your front (north).

2. Move South (Priority: 4/5)
   Requirement: Flat ground below the agent.
   Yes, the requirement is met: There's a table 2 steps to your south.

3. Move East (Priority: 3/5)
   Requirement: Flat ground right to the agent.
   Yes, the requirement is met: You face path at your front (east).

4. Move West (Priority: 2/5)
   Requirement: Flat ground left to the agent.
   Yes, the requirement is met: There's grass 1 step to your west.

5. Do (Priority: 1/5)
   Requirement: Facing creature or material; have necessary tool.
   No, the requirement is not met: You face path at your front (east), which doesn't require any tool or interaction.

Question: Did the last player action succeed? Why?
Answer:
Yes, the last player action "move_east" succeeded. The action succeeded because there was a path in front of the player (east), allowing them to move in that direction without any obstacles blocking their way.

Question: Choose the best executable action from above.
Answer:
The best executable action from above is "Move North" with priority 5/5. The requirement for this action is met, as you face grass at your front (north), which means there is flat ground above the agent.



============Step: 124, Cumulative Reward: 0.0============
You took action move_north.

You see:
- grass 1 steps to your west
- stone 2 steps to your east
- tree 3 steps to your north-west
- table 3 steps to your south
- cow 5 steps to your north-west

You face path at your front (north).

Your status:
- health: 9/9
- food: 5/9
- drink: 4/9
- energy: 9/9

Your inventory:
- stone: 4
- coal: 2
- iron: 1
- wood_pickaxe: 1
- stone_pickaxe: 1
------------ QA ------------
Question: Action
Answer:
Move East

Question: List objects in the current observation. For each object, breifly answer what resource it provides and its requirement.
Answer:
1. Grass: 
   - Resource: None
   - Requirement: None

2. Stone:
   - Resource: Stone (for making tools and building)
   - Requirement: Wood or Stone Pickaxe to collect

3. Tree:
   - Resource: Wood (for making tools, crafting tables and building)
   - Requirement: Any tool (wooden or stone) can be used to collect

4. Table:
   - Resource: Crafting station for tools and weapons
   - Requirement: Must have wood to craft a table

5. Cow:
   - Resource: Food (restores health)
   - Requirement: Must be nearby and defeated to obtain food

6. Path:
   - Resource: Provides direction for movement
   - Requirement: Flat ground for smooth movement

Question: What was the last action taken by the player?
Answer:
The last action taken by the player was move_north.

Question: For each object in the list, are the requirements met for interaction?
Answer:
1. Grass: 
   - Requirements met: Yes, you can interact with grass anytime as there are no specific requirements.

2. Stone:
   - Requirements met: Yes, you have both a wood_pickaxe and a stone_pickaxe in your inventory, which allows you to collect stone.

3. Tree:
   - Requirements met: Yes, you have wood_pickaxe and stone_pickaxe in your inventory, which can be used to collect wood from the tree.

4. Table:
   - Requirements met: Yes, the table is already placed, so you can interact with it to craft tools and weapons.

5. Cow:
   - Requirements met: No, the cow is currently 5 steps away to your north-west; you need to be closer to interact with and defeat it for food.

6. Path:
   - Requirements met: Yes, the path has flat ground, so you can interact with it and move in any direction.

Question: List top 3 sub-tasks the player should follow. Indicate their priority out of 5.
Answer:
1. Priority 4/5: Collect more resources:
   - Approach and interact with the nearby stone to collect more stone with your stone_pickaxe, which can be used for crafting and building.

2. Priority 5/5: Find a water source:
   - As your drink level is 4/9, finding a water source and interacting with it to restore your drink level should be a priority.

3. Priority 3/5: Obtain food:
   - Since your food level is 5/9, either approach a tree to collect more wood or defeat the cow to obtain food, restoring your health or food level as needed.

Question: What are the requirements for the top sub-task? What should the player do first?
Answer:
Top Sub-task: Collect more resources

Requirements:
- Be close to the stone (2 steps to the east)
- Have a suitable pickaxe (you have a stone_pickaxe in the inventory)

What should the player do first:
- Move 2 steps east to reach the stone. Once in front of the stone, use your stone_pickaxe to start collecting more stone resources.

Question: List top 5 actions the player should take and the requirement for each action. Choose ONLY from the list of all actions. Indicate their priority out of 5.
Answer:
1. Move East (Priority: 5/5)
Requirement: Flat ground right to the agent.

2. Move East (Priority: 4/5)
Requirement: Flat ground right to the agent.

3. Do (Priority: 3/5)
Requirement: Facing stone (after moving east twice); have stone_pickaxe available in the inventory.

4. Move West (Priority: 2/5)
Requirement: Flat ground left to the agent.

5. Move South (Priority: 1/5)
Requirement: Flat ground below the agent.

Question: For each action in the list, are the requirements met?
Answer:
1. Move East (Priority: 5/5)
Requirement: Flat ground right to the agent.
- Yes, the requirements are met; there is a path to the east.

2. Move East (Priority: 4/5)
Requirement: Flat ground right to the agent.
- The requirements for this action can only be determined after taking the first action (Move East). We cannot currently determine if the requirements are met.

3. Do (Priority: 3/5)
Requirement: Facing stone (after moving east twice); have stone_pickaxe available in the inventory.
- The requirements for this action can only be determined after taking the first two actions (Move East twice). We cannot determine if the requirements are met at this moment.

4. Move West (Priority: 2/5)
Requirement: Flat ground left to the agent.
- Yes, the requirements are met; there is grass to the west.

5. Move South (Priority: 1/5)
Requirement: Flat ground below the agent.
- Yes, the requirements are met; there is a path to the south.

Question: Did the last player action succeed? Why?
Answer:
Yes, the last player action succeeded because there was a flat path (grass) in the north direction, allowing the player to move north without any obstructions.

Question: Choose the best executable action from above.
Answer:
The best executable action from the list is to Move East (Priority: 5/5) because its requirements are met, and it has the highest priority.



============Step: 125, Cumulative Reward: 0.0============
You took action move_east.

You see:
- grass 2 steps to your west
- path 1 steps to your west
- tree 4 steps to your north-west
- table 4 steps to your south-west
- cow 7 steps to your north-west

You face stone at your front (east).

Your status:
- health: 9/9
- food: 5/9
- drink: 4/9
- energy: 9/9

Your inventory:
- stone: 4
- coal: 2
- iron: 1
- wood_pickaxe: 1
- stone_pickaxe: 1
------------ QA ------------
Question: Action
Answer:
Move North

Question: List objects in the current observation. For each object, breifly answer what resource it provides and its requirement.
Answer:
1. Grass: Provides no direct resources, serves as a decoration and a walkable path. Requirements: None.

2. Stone: Provides stone resource when mined. Requirements: Needs a wood_pickaxe or better to mine.

3. Path: Provides a clear and straightforward path to travel. Requirements: None.

4. Tree: Provides wood resource when chopped. Requirements: Needs any type of axe to chop.

5. Table: Crafting station to create tools, weapons, or other structures. Requirements: Needs wood in inventory to place.

6. Cow: Provides food when defeated. Requirements: Needs a weapon or tool capable of defeating the animal.

7. Iron: Provides iron resource when mined, but not present in the current observation. Requirements: Needs a stone_pickaxe or better to mine.

Note: Coal is not in the current observation but is in the inventory. It serves to smelt iron when used in a furnace. Requirements: Needs a wood_pickaxe or better to mine.

Question: What was the last action taken by the player?
Answer:
The last action taken by the player was move_east.

Question: For each object in the list, are the requirements met for interaction?
Answer:
1. Grass: No requirements needed for interaction. You can move and walk on the grass.

2. Stone: Requirements are met. You have a wood_pickaxe and a stone_pickaxe in your inventory. You can mine the stone.

3. Path: No requirements needed for interaction. You can move and walk on the path.

4. Tree: Not in the immediate vicinity (4 steps to your north-west). Once in range, the requirement of an axe would be needed, which you currently don't have in your inventory.

5. Table: Not in the immediate vicinity (4 steps to your south-west). Once in range, no further requirements needed for interaction.

6. Cow: Not in the immediate vicinity (7 steps to your north-west). Once in range, you have a wood_pickaxe and a stone_pickaxe that can be used to defeat the cow for food, although a sword would be more efficient.

7. Iron: Not present in the current observation but previously mentioned. Once in range, you have a stone_pickaxe in your inventory that meets the requirement to mine iron.

Question: List top 3 sub-tasks the player should follow. Indicate their priority out of 5.
Answer:
1. Priority 5/5: Increase drink level: Your drink level is at 4/9, which needs improvement. You should find a water source like a lake or river and drink from it to increase your drink level.

2. Priority 4/5: Increase food level: Your food level is at 5/9. You should approach the cow (7 steps to your north-west) and use a pickaxe to defeat it for food, though using a sword would be more efficient. Alternatively, locate other food sources such as saplings for plants or fruits.

3. Priority 3/5: Collect more resources: Gather additional stone, coal, and iron to craft better tools, weapons, and structures. The stone is present in your immediate vicinity, you can mine it using the stone_pickaxe. For coal and iron, you should explore the surroundings to find their deposits.

Question: What are the requirements for the top sub-task? What should the player do first?
Answer:
Top Sub-task: Increase drink level

Requirements:
1. Locate a water source such as a lake or river.

What the player should do first:
1. The player should explore their surroundings to find a water source. They can try moving in different directions (move_north, move_east, move_south, or move_west) to look for a lake or river in the in-game observations. Once the water source is found, the player can approach it and drink from it to increase their drink level.

Question: List top 5 actions the player should take and the requirement for each action. Choose ONLY from the list of all actions. Indicate their priority out of 5.
Answer:
Priority 1:
Action: Move North/East/South/West
Requirement: Flat ground in the desired direction (choose the direction based on in-game observations to find a water source).

Priority 2:
Action: Collect Drink
Requirement: Being near a water source, such as a lake or river.

Priority 3:
Action: Move North/East/South/West
Requirement: Flat ground in the desired direction (choose the direction based on in-game observations to find food or resources).

Priority 4:
Action: Eat Cow or Eat Plant
Requirement: Cow, plant, or a collected food item in the inventory.

Priority 5:
Action: Sleep
Requirement: Energy level below maximum; being in a safe location away from monsters.

Question: For each action in the list, are the requirements met?
Answer:
Priority 1:
Action: Move North/East/South/West
Requirement: Flat ground in the desired direction
Status: Yes, the requirement is met, as there are flat grounds in several directions.

Priority 2:
Action: Collect Drink
Requirement: Being near a water source
Status: No, the requirement is not met since there's no water source mentioned in the player's observations.

Priority 3:
Action: Move North/East/South/West
Requirement: Flat ground in the desired direction
Status: Yes, the requirement is met, as there is flat ground in several directions.

Priority 4:
Action: Eat Cow or Eat Plant
Requirement: Cow, plant, or a collected food item in the inventory
Status: No, the requirement is not met as there is no cow, plant, or collected food item in the inventory.

Priority 5:
Action: Sleep
Requirement: Energy level below maximum; being in a safe location away from monsters
Status: Yes, the requirement is met as the energy level is 9/9, and no monsters are mentioned in the player's observations. However, the current energy level is at maximum, so this action is not necessary.

Question: Did the last player action succeed? Why?
Answer:
Yes, the last player action (move_east) succeeded because there was flat ground (path) to the east of the player, allowing the player to move in that direction without any obstruction.

Question: Choose the best executable action from above.
Answer:
The best executable action from the list is Priority 1:

Action: Move North/East/South/West
Requirement: Flat ground in the desired direction

The player can choose a direction based on in-game observations to explore and find a water source, food, or additional resources.



============Step: 126, Cumulative Reward: 0.0============
You took action move_north.

You see:
- grass 2 steps to your west
- path 1 steps to your west
- tree 3 steps to your west
- lava 6 steps to your north-east
- cow 7 steps to your north-west

You face stone at your front (north).

Your status:
- health: 9/9
- food: 5/9
- drink: 4/9
- energy: 9/9

Your inventory:
- stone: 4
- coal: 2
- iron: 1
- wood_pickaxe: 1
- stone_pickaxe: 1
------------ QA ------------
Question: Action
Answer:
Move West

Question: List objects in the current observation. For each object, breifly answer what resource it provides and its requirement.
Answer:
1. Grass: Provides no direct resources, but flat terrain for walking. No requirements.
2. Path: Provides no direct resources, but flat terrain for walking. No requirements.
3. Tree: Provides wood and saplings for crafting and planting. Requires a wood pickaxe or better to chop down.
4. Lava: Provides no direct resources, but can be dangerous and cause damage. Requires caution when approaching or crossing.
5. Cow: Provides food when killed. Requires a weapon to kill, ideally a sword.
6. Stone (facing): Provides stone for crafting tools and building materials. Requires a wood pickaxe or better to mine.
7. Stone (facing): Provides stone for crafting tools and building materials. Requires a wood pickaxe or better to mine.
8. Inventory Stone: Provides crafting materials for tools and structures. Obtained by mining stone.
9. Inventory Coal: Provides fuel for smelting iron. Obtained by mining coal.
10. Inventory Iron: Provides materials for crafting iron tools and weapons. Obtained by mining iron with a stone pickaxe or better.
11. Inventory Wood Pickaxe: Provides the ability to mine stone, coal, and iron. Crafted with wood.
12. Inventory Stone Pickaxe: Provides the ability to mine iron and diamond. Crafted with stone and wood.

Question: What was the last action taken by the player?
Answer:
The last action taken by the player was move_north.

Question: For each object in the list, are the requirements met for interaction?
Answer:
1. Grass: Yes, the requirements for walking on flat terrain are met.
2. Path: Yes, the requirements for walking on flat terrain are met.
3. Tree: Yes, the requirements for chopping down a tree are met, as you have a wood_pickaxe in your inventory.
4. Lava: Yes, the requirement for caution when approaching or crossing lava are met, but it is advised to stay away from it, as it can cause damage.
5. Cow: Yes, the requirements for killing a cow are met, though you don't have a sword, using a wood or stone pickaxe can still be used to kill the cow for food.
6. Stone (facing): Yes, the requirements for mining stone are met, as you have a wood_pickaxe in your inventory.
7. Stone (facing): Yes, the requirements for mining stone are met, as you have a wood_pickaxe in your inventory.
8. Inventory Stone: Yes, the requirements for using stone in crafting are met as you have collected stone.
9. Inventory Coal: Yes, the requirements for using coal as fuel for smelting are met, as you have collected coal.
10. Inventory Iron: Yes, the requirements for crafting iron tools are met. You have collected iron, and you have a stone_pickaxe and coal in your inventory, which are needed to smelt the iron into a usable form.
11. Inventory Wood Pickaxe: Yes, the requirements for using the wood_pickaxe to mine stone, coal, and iron are met.
12. Inventory Stone Pickaxe: Yes, the requirements for using the stone_pickaxe to mine iron and diamond are met.

Question: List top 3 sub-tasks the player should follow. Indicate their priority out of 5.
Answer:
1. Priority 5/5: Increase your food and drink levels - You have low food (5/9) and drink (4/9) levels. You can kill the cow (7 steps to your north-west) for food and find a source of water to restore your drink levels. 

2. Priority 4/5: Craft an iron sword and iron pickaxe - You have iron in your inventory, but it needs to be smelted using a furnace and coal. If you don't already have a furnace, craft one using stone in your inventory. After smelting the iron, craft an iron sword and iron pickaxe using the smelted iron, as they will be more effective for combat and mining.

3. Priority 3/5: Collect more resources - Mine the facing stone for additional building materials and find more coal and iron to create more advanced tools, weapons, and armor. Keep exploring your surroundings for valuable resources such as diamonds using your stone_pickaxe.

Question: What are the requirements for the top sub-task? What should the player do first?
Answer:
The requirements for the top sub-task (Increasing your food and drink levels) are:

1. Killing the cow for food: You need to have a weapon, such as a wood, stone, or iron sword, to kill the cow effectively. In your inventory, you already have a wood_pickaxe and a stone_pickaxe, which can be used as makeshift weapons to kill the cow.

2. Finding a source of water to drink: You need to find a drinkable water source, such as a lake or river, and drink from it to restore your drink level.

What the player should do first:

1. Approach the cow: Move towards the cow (7 steps to your north-west) to get in a closer range for attacking.

2. Kill the cow: Use your wood_pickaxe, stone_pickaxe, or (if available) a sword to kill the cow and collect the meat.

3. Find a water source: Explore the environment to locate a lake or river.

4. Drink water: Approach the water source and drink from it to restore your drink level.

Question: List top 5 actions the player should take and the requirement for each action. Choose ONLY from the list of all actions. Indicate their priority out of 5.
Answer:
1. Priority 1: Move West (Move towards the cow and water source)
   Requirement: Flat ground left to the agent.

2. Priority 2: Move North (Continue moving towards the cow and possibly the water source)
   Requirement: Flat ground above the agent.

3. Priority 3: Do (Kill the cow and collect meat)
   Requirement: Facing the cow and having a necessary tool in inventory (wood_pickaxe or stone_pickaxe).

4. Priority 4: Move West/North/East/South (Find a water source by exploring the environment)
   Requirement: Flat ground in the direction the agent wants to move.

5. Priority 5: Collect Drink (Drink water to restore the drink level)
   Requirement: Facing a drinkable water source like a lake or river.

Question: For each action in the list, are the requirements met?
Answer:
1. Priority 1: Move West
   Requirement: Flat ground left to the agent.
   Yes, the requirements are met. There is a path 1 step to the west.

2. Priority 2: Move North
   Requirement: Flat ground above the agent.
   Yes, the requirements are met. The agent moved north in the previous step, and the ground seems to be flat.

3. Priority 3: Do (Kill the cow and collect meat)
   Requirement: Facing the cow and having a necessary tool in inventory (wood_pickaxe or stone_pickaxe).
   Not yet. The agent is currently facing stone and 7 steps away from the cow. The wood_pickaxe or stone_pickaxe can be used for this action when the agent is closer and facing the cow.

4. Priority 4: Move West/North/East/South (Find a water source by exploring the environment)
   Requirement: Flat ground in the direction the agent wants to move.
   Yes, the requirements are met for moving west and north. Flat ground is located in both directions. Moving east and south are not ideal as they move the agent away from the cow and potential water source.

5. Priority 5: Collect Drink (Drink water to restore the drink level)
   Requirement: Facing a drinkable water source like a lake or river.
   No, the agent hasn't found a drinkable water source yet. The agent needs to explore the world to discover a water source first.

Question: Did the last player action succeed? Why?
Answer:
Yes, the last player action (move_north) succeeded because there was flat ground available to the north, allowing the player to move in that direction.

Question: Choose the best executable action from above.
Answer:
The best executable action from above is Priority 1: Move West. This action meets the requirements (flat ground to the west) and moves the player closer to the cow and the potential water source.



============Step: 127, Cumulative Reward: 0.0============
You took action move_west.

You see:
- stone 2 steps to your north-east
- path 1 steps to your south
- tree 2 steps to your west
- lava 7 steps to your north-east

You face grass at your front (west).

Your status:
- health: 9/9
- food: 5/9
- drink: 4/9
- energy: 9/9

Your inventory:
- stone: 4
- coal: 2
- iron: 1
- wood_pickaxe: 1
- stone_pickaxe: 1
------------ QA ------------
Question: Action
Answer:
Move West

Question: List objects in the current observation. For each object, breifly answer what resource it provides and its requirement.
Answer:
- Stone: Provides stone material, requires a wood or stone pickaxe to collect.
- Grass: No direct resources, but can be turned into a path or used to build a shelter.
- Path: No direct resources, provides accessible ground for movement.
- Tree: Provides wood, requires a wood or stone pickaxe to collect.
- Lava: Dangerous, no direct resources, but can be used for smelting if you have a bucket.
- Cow: Provides food, requires a wood, stone, or iron sword to defeat and collect meat.

Question: What was the last action taken by the player?
Answer:
The last action taken by the player was move_west.

Question: For each object in the list, are the requirements met for interaction?
Answer:
- Stone: Yes, the requirements are met since you have both a wood_pickaxe and a stone_pickaxe in your inventory.
- Grass: Yes, you can interact with grass without any specific tool requirements.
- Path: Yes, you can move on the path without any requirements.
- Tree: Yes, you have both a wood_pickaxe and a stone_pickaxe in your inventory to collect wood from the tree.
- Lava: No, you currently do not have a bucket to interact with the lava for smelting purposes.
- Cow: No, you do not have a wood, stone, or iron sword in your inventory to defeat the cow and collect its meat.

Question: List top 3 sub-tasks the player should follow. Indicate their priority out of 5.
Answer:
1. Priority 4/5: Craft a sword (wood, stone, or iron) to defeat the cow and collect meat to increase the food level.
2. Priority 4/5: Find a source of water or drinkable liquid to restore the drink level.
3. Priority 3/5: Collect more resources like wood and iron to build better tools and weapons or craft shelter.

Question: What are the requirements for the top sub-task? What should the player do first?
Answer:
Requirements for crafting a sword (wood, stone, or iron):

For a wood sword:
1. Collect wood.
2. Have access to a placed crafting table.

For a stone sword:
1. Collect wood and stone.
2. Have access to a placed crafting table.

For an iron sword:
1. Collect wood, stone, coal, and iron.
2. Have access to a placed crafting table and a placed furnace.

The player should first check their inventory for the resources required for each type of sword. If they have enough wood, they can immediately craft the wood sword. If not, they should focus on gathering wood first.

If the player has wood and stone in their inventory, they should craft the stone sword instead, as it is a better option.

Lastly, if the player has wood, stone, coal, and iron in their inventory and has access to a furnace, they should aim for the iron sword. If they don't have a furnace, they can craft one using the stone in their inventory and then smelt the iron using the coal.

Question: List top 5 actions the player should take and the requirement for each action. Choose ONLY from the list of all actions. Indicate their priority out of 5.
Answer:
1. Priority 1: Move West - Flat ground left to the agent; Grass is observed at the front.
Requirements: Ability to reach the tree observed 2 steps to the west.

2. Priority 2: Do - Facing creature or material; have necessary tool; Chop down the tree in order to get wood.
Requirements: Face the tree you have approached.

3. Priority 3: Place Table - Wood in inventory; Place the crafting table after collecting wood from the tree.
Requirements: Sufficient space to place the table on the flat ground.

4. Priority 4: Make Wood Sword - Nearby table; wood in inventory; Craft a wood sword using the placed crafting table.
Requirements: Enough wood to craft the sword; within reach of the placed crafting table.

5. Priority 5: Make Stone Sword - Nearby table; wood, stone in inventory; Craft a stone sword using the placed crafting table.
Requirements: Enough wood and stone to craft the sword; within reach of the placed crafting table.

Question: For each action in the list, are the requirements met?
Answer:
1. Priority 1: Move West - Yes, the requirements for this action are met. The agent has flat ground to the west (grass), so the agent can move in that direction.

2. Priority 2: Do - Currently, the requirements for this action are not met. The agent needs to first move west and be facing the tree before performing the "Do" action to chop down the tree.

3. Priority 3: Place Table - No, the requirements for this action are not met. The agent needs to collect wood from a tree first by performing action 2.

4. Priority 4: Make Wood Sword - No, the requirements for this action are not met. The agent needs to have enough wood in their inventory and needs to place a table (action 3) before being able to craft a wood sword.

5. Priority 5: Make Stone Sword - No, the requirements for this action are not met. The agent needs to have enough wood and stone in their inventory (from previous actions) and needs to place a table (action 3) before being able to craft a stone sword.

Question: Did the last player action succeed? Why?
Answer:
Yes, the last player action succeeded. The player successfully moved west because there was a flat ground (grass) in the direction they moved, allowing for unrestricted movement in that direction.

Question: Choose the best executable action from above.
Answer:
The best executable action from the list above is:

Priority 1: Move West - This action can be executed since the agent has flat ground (grass) to the west, allowing the agent to move in that direction.



============Step: 128, Cumulative Reward: 0.0============
You took action move_west.

You see:
- grass 1 steps to your north
- stone 3 steps to your north-east
- path 1 steps to your east

You face tree at your front (west).

Your status:
- health: 9/9
- food: 5/9
- drink: 4/9
- energy: 9/9

Your inventory:
- stone: 4
- coal: 2
- iron: 1
- wood_pickaxe: 1
- stone_pickaxe: 1
------------ QA ------------
Question: Action
Answer:
Do

Question: List objects in the current observation. For each object, breifly answer what resource it provides and its requirement.
Answer:
1. Grass: Provides saplings when collected; requires no tools to collect.
2. Stone: Provides stone material when collected; requires a wood_pickaxe or better to collect.
3. Path: Provides no resources but facilitates movement.
4. Tree: Provides wood when collected; requires a wood_pickaxe or better to collect.

In the current observation:
- Grass is 1 step to the north and can be collected for saplings without any tools required.
- Stone is 3 steps to the north-east and can be collected for stone material with a wood_pickaxe or better.
- Path is 1 step to the east and provides no resources but can be used for easier movement.
- Tree is directly west of the player and can be collected for wood with a wood_pickaxe or better.

Question: What was the last action taken by the player?
Answer:
The last action taken by the player was move_west.

Question: For each object in the list, are the requirements met for interaction?
Answer:
1. Grass: Yes, the requirements are met for interaction. You do not need any tools to collect saplings from the grass.

2. Stone: Yes, the requirements are met for interaction. You have both a wood_pickaxe and a stone_pickaxe in your inventory, which can be used to collect the stone material.

3. Path: Yes, the requirements are met for interaction. No tools or materials are needed to walk on the path.

4. Tree: Yes, the requirements are met for interaction. You have a wood_pickaxe in your inventory that can be used to collect wood from the tree.

Question: List top 3 sub-tasks the player should follow. Indicate their priority out of 5.
Answer:
1. Priority 4/5: Collect wood from the tree to the west. Wood is an essential resource for crafting various tools, weapons, and structures.

2. Priority 3/5: Collect saplings from the grass to the north. Saplings can be planted to grow more trees for a renewable source of wood.

3. Priority 2/5: Collect more stone material from the stone to the north-east. Stone can be used for crafting more advanced tools, weapons, and structures.

Note: In the current situation, the player should also consider finding food and drink sources as their food and drink levels are getting low. However, there are no immediate sources in the current observation.

Question: What are the requirements for the top sub-task? What should the player do first?
Answer:
Requirements for the top sub-task (collect wood from the tree to the west):
1. Face the tree at the west.
2. Have a necessary tool to chop the tree, such as a wood pickaxe or a stone pickaxe.

The player should first face the tree to the west to fulfill the first requirement.

Question: List top 5 actions the player should take and the requirement for each action. Choose ONLY from the list of all actions. Indicate their priority out of 5.
Answer:
1. Priority 1: Do (Chop down the tree)
   Requirement: Face the tree to the west and have a necessary tool, such as a wood pickaxe or a stone_pickaxe.

2. Priority 2: Make Wooden Sword
   Requirement: Nearby table and wood in inventory.

3. Priority 3: Drink
   Requirement: Find a water source, like a lake or river.

4. Priority 4: Eat
   Requirement: Find food sources like fruits, vegetables, or animals to hunt.

5. Priority 5: Sleep
   Requirement: Energy level is below maximum; find a safe place to rest.

Question: For each action in the list, are the requirements met?
Answer:
1. Priority 1: Do (Chop down the tree)
   Requirement: Face the tree to the west and have a necessary tool, such as a wood pickaxe or a stone_pickaxe.
   Are the requirements met? Yes, the player is currently facing the tree and has a wood_pickaxe and a stone_pickaxe in their inventory.

2. Priority 2: Make Wooden Sword
   Requirement: Nearby table and wood in the inventory.
   Are the requirements met? No, the player does not have a table nearby, and there is no wood in their inventory.

3. Priority 3: Drink
   Requirement: Find a water source, like a lake or river.
   Are the requirements met? No, the player does not currently see a water source in their surroundings.

4. Priority 4: Eat
   Requirement: Find food sources like fruits, vegetables, or animals to hunt.
   Are the requirements met? No, the player does not currently have any food items in their inventory or see any food sources in their surroundings.

5. Priority 5: Sleep
   Requirement: Energy level is below maximum; find a safe place to rest.
   Are the requirements met? No, the player's energy level is currently at the maximum (9/9).

Question: Did the last player action succeed? Why?
Answer:
Yes, the last player action succeeded. The player moved west because there were no obstacles in the way, and the action was applicable according to the game rules.

Question: Choose the best executable action from above.
Answer:
The best executable action from the list above is:

Priority 1: Do (Chop down the tree)
Requirement: Face the tree to the west and have a necessary tool, such as a wood pickaxe or a stone_pickaxe.



============Step: 129, Cumulative Reward: 0.0============
You took action do.

You see:
- stone 3 steps to your north-east
- path 1 steps to your east
- tree 4 steps to your west
- cow 5 steps to your north-west

You face grass at your front (west).

Your status:
- health: 9/9
- food: 5/9
- drink: 4/9
- energy: 9/9

Your inventory:
- wood: 1
- stone: 4
- coal: 2
- iron: 1
- wood_pickaxe: 1
- stone_pickaxe: 1
------------ QA ------------
Question: Action
Answer:
Move West

Question: List objects in the current observation. For each object, breifly answer what resource it provides and its requirement.
Answer:
1. Stone: 3 steps to your north-east
- Resource provided: stone, which can be used to craft tools and build structures.
- Requirement: Stone needs to be extracted using a pickaxe, preferably a wood_pickaxe or better.

2. Path: 1 step to your east
- Resource provided: No direct resources, but paths may lead to other resources or locations.
- Requirement: None.

3. Tree: 4 steps to your west
- Resource provided: wood, which can be used as basic material for crafting tools and structures.
- Requirement: Wood can be collected by chopping down a tree. No specific tool requirement, but using an axe speeds up the process.

4. Cow: 5 steps to your north-west
- Resource provided: meat, which can be cooked and consumed to restore food points.
- Requirement: To get meat, the cow needs to be killed, preferably using a weapon like a sword.

5. Grass: at your front (west)
- Resource provided: grass provides no direct resources, but can be cleared to make room for structures or crops.
- Requirement: To clear grass, simply walk over it or use any tool.

Additional resources from the current observation:
6. Wood: 1 (in your inventory)
- Resource provided: can be used as a crafting material for tools and structures.
- Requirement: Collected from trees.

7. Stone: 4 (in your inventory)
- Resource provided: can be used for crafting tools and building structures.
- Requirement: Collected from stone using a pickaxe.

8. Coal: 2 (in your inventory)
- Resource provided: used as fuel for smelting materials in a furnace, such as iron.
- Requirement: Collected from coal deposits using a pickaxe.

9. Iron: 1 (in your inventory)
- Resource provided: used to craft stronger tools and weapons.
- Requirement: Collected from iron deposits using a stone_pickaxe or better.

Question: What was the last action taken by the player?
Answer:
The last action taken by the player was "do."

Question: For each object in the list, are the requirements met for interaction?
Answer:
1. Stone: 3 steps to your north-east
- Yes, the requirement for interaction is met. You have a stone_pickaxe in your inventory, which can be used to extract the stone.

2. Path: 1 step to your east
- Yes, no specific requirements for interaction. You can simply walk on the path by moving east.

3. Tree: 4 steps to your west
- Yes, although it's prefered to use an axe to speed up the process, you can still interact with the tree with your hands or other tools to collect wood.

4. Cow: 5 steps to your north-west
- The requirement for interaction (a weapon like a sword) is not fully met. You have wood and stone materials in your inventory, so you can craft a wood_sword or stone_sword first to interact more efficiently.

5. Grass: at your front (west)
- Yes, the requirement for interaction is met. You can simply walk over it or use any tool to clear it.

Based on the resources in your inventory:
6. Wood: 1 (in your inventory)
- Yes, the requirement for interaction (collected from trees) is already met, so the wood is available for use.

7. Stone: 4 (in your inventory)
- Yes, the requirement for interaction (collected from stone using a pickaxe) is already met, so the stone is available for use.

8. Coal: 2 (in your inventory)
- Yes, the requirement for interaction (collected from coal deposits using a pickaxe) is already met, so the coal is available for use.

9. Iron: 1 (in your inventory)
- Yes, the requirement for interaction (collected from iron deposits using a stone_pickaxe or better) is already met, so the iron is available for use.

Question: List top 3 sub-tasks the player should follow. Indicate their priority out of 5.
Answer:
1. Priority 4/5: Craft a weapon (wood_sword or stone_sword)
Since you have the necessary materials (wood and stone) in your inventory, crafting a weapon will help you efficiently interact with the cow to collect meat for restoring food points.

2. Priority 3/5: Collect more wood from the tree
You have only 1 unit of wood in your inventory, and considering the importance of wood as a basic crafting material, collecting more wood from the nearby tree should be a priority.

3. Priority 5/5: Find and drink water to restore drink points
Your drink points are at 4/9, indicating that you need to find a water source (like a lake or river) and drink from it to maintain your hydration levels. Prioritize this task to avoid negative effects from dehydration.

Question: What are the requirements for the top sub-task? What should the player do first?
Answer:
For the top sub-task (crafting a weapon - wood_sword or stone_sword), the requirements are:

1. A nearby crafting table
2. Adequate resources in the inventory (wood and stone)

Since you currently have wood and stone in your inventory, you need to place a crafting table first. To do this, complete the following steps:

1. Craft a table using your wood: make sure you have at least 2 units of wood in your inventory (currently, you have only 1 unit).
2. Place a table: After crafting the table, find a suitable location and place the table.

Once you have placed the table, interact with it to craft a wooden sword or a stone sword to complete the top sub-task. Remember that you need more wood for crafting the table, so focus on collecting wood from the nearby tree before placing the table.

Question: List top 5 actions the player should take and the requirement for each action. Choose ONLY from the list of all actions. Indicate their priority out of 5.
Answer:
1. Priority 1 - Move West: Requirement - Flat ground left to the agent. (To get closer to the tree for collecting more wood)
2. Priority 2 - Do: Requirement - Facing creature or material; have necessary tool. (To collect wood from the tree with your wood_pickaxe)
3. Priority 3 - Place Table: Requirement - Wood in inventory. (To place a crafting table, after collecting the required wood)
4. Priority 4 - Make Wood Sword or Make Stone Sword: Requirement - Nearby table; wood and/or stone in inventory. (To craft a weapon to defend yourself)
5. Priority 5 - Move North or North-West: Requirement - Flat ground above or above-left of the agent. (To approach the cow for a possible food source)

Question: For each action in the list, are the requirements met?
Answer:
1. Move West: Yes, the requirement of flat ground is met, as you are facing grass in front of you (west).

2. Do: Yes, the requirement is met because you have a necessary tool (wood_pickaxe) in your inventory and are facing a tree (material).

3. Place Table: Yes, the requirement is met because you have wood in your inventory.

4. Make Wood Sword or Make Stone Sword: Partially met, as you have wood and stone in your inventory. The requirement for the nearby table is not met yet since you still need to place the crafting table (action 3).

5. Move North or North-West: Yes, the requirement is met as there is flat ground above you (north), and the cow is 5 steps to your north-west, which also implies flat ground.

Question: Did the last player action succeed? Why?
Answer:
Yes, the last player action "do" succeeded. This can be inferred because the player's inventory now has an additional wood, which indicates that the player successfully chopped down the tree that was in front of them (west).

Question: Choose the best executable action from above.
Answer:
The best executable action from the list above is "Move West." The requirement of flat ground is met, as you are facing grass in front of you (west), and it allows you to get closer to the tree for collecting more wood.



============Step: 130, Cumulative Reward: 0.0============
You took action move_west.

You see:
- stone 4 steps to your north-east
- path 2 steps to your east
- tree 3 steps to your west
- cow 4 steps to your north-west

You face grass at your front (west).

Your status:
- health: 9/9
- food: 5/9
- drink: 4/9
- energy: 9/9

Your inventory:
- wood: 1
- stone: 4
- coal: 2
- iron: 1
- wood_pickaxe: 1
- stone_pickaxe: 1
------------ QA ------------
Question: Action
Answer:
Move North

Question: List objects in the current observation. For each object, breifly answer what resource it provides and its requirement.
Answer:
1. Stone: Provides stone resource; requires a wood or stone pickaxe to collect.
2. Path: Does not provide resources; allows for easier movement.
3. Tree: Provides wood resource; requires a wood or stone pickaxe or an axe to collect.
4. Cow: Provides food (meat) when killed; requires a weapon, like a sword or bow, to kill.
5. Grass: Provides no direct resources; can be used for empty spaces to move or build on.

Question: What was the last action taken by the player?
Answer:
The last action taken by the player was "move_west."

Question: For each object in the list, are the requirements met for interaction?
Answer:
1. Stone: Yes, the requirements are met as the player has both wood_pickaxe and stone_pickaxe in their inventory.
2. Path: Yes, no specific requirements needed to interact with the path. The player can walk on it.
3. Tree: Yes, the requirements are met as the player has both wood_pickaxe and stone_pickaxe in their inventory. An axe would be more efficient, but it is not necessary.
4. Cow: The player has not crafted a sword yet, but they have the necessary materials (wood and stone) to craft one. So, currently, the requirements are not met for interaction, but the player can craft a weapon to meet the requirement.
5. Grass: Yes, no specific requirements needed to interact with grass. The player can walk on it or build on it.

Question: List top 3 sub-tasks the player should follow. Indicate their priority out of 5.
Answer:
1. Craft a weapon (wood or stone sword) to kill the cow for food. Priority: 5/5, as the player's food level is currently at 5/9.

2. Approach and collect wood from the tree to gather more building and crafting materials. Priority: 4/5, as wood is a crucial resource for various tasks, and the player currently has only 1 wood in their inventory.

3. Find a water source to replenish the player's drink level. Priority: 4/5, as the player's drink level is currently at 4/9, and it will need to be replenished soon.

Question: What are the requirements for the top sub-task? What should the player do first?
Answer:
The requirements for the top sub-task (Craft a weapon to kill the cow for food) are:

1. Have enough wood or stone in the inventory.
2. Place a table (if not already placed) to use for crafting.

The player currently has 1 wood and 4 stone in their inventory, which is enough to craft either a wood or stone sword. Since the player has enough materials, they should first ensure that a crafting table is placed. If there is no existing table in the vicinity, the player should place one using the wood in their inventory. After that, the player can proceed to craft either a wood or stone sword using the placed table.

Question: List top 5 actions the player should take and the requirement for each action. Choose ONLY from the list of all actions. Indicate their priority out of 5.
Answer:
1. Priority 1 - Place Table (Requirement: Wood in inventory; Player has 1 wood)
2. Priority 2 - Make Wood Sword (Requirement: Nearby table; wood in inventory; Player has 1 wood)
3. Priority 3 - Make Stone Sword (Requirement: Nearby table; wood and stone in inventory; Player has 1 wood, 4 stone)
4. Priority 4 - Move North (Requirement: Flat ground above the agent; path or grass above the agent)
5. Priority 5 - Move West (Towards the cow; Requirement: Flat ground to the agent's left; path or grass to the agent's west)

1) Place a table to be able to craft a weapon
2) Make a wood sword to kill the cow using available wood in inventory
3) Alternatively, make a stone sword since a stone sword is stronger and the player has enough stone
4) Move north towards the cow to get closer for interaction
5) Move west towards the cow to reach it before the player's food depletes

Question: For each action in the list, are the requirements met?
Answer:
1. Place Table - Yes, the requirement of having wood in the inventory is met (Player has 1 wood).
2. Make Wood Sword - No, the requirement of having a nearby table is not met since the player has not placed a table yet. The wood requirement is met (Player has 1 wood).
3. Make Stone Sword - No, the requirement of having a nearby table is not met since the player has not placed a table yet. The wood and stone requirements are met (Player has 1 wood and 4 stone).
4. Move North - Yes, the flat ground above the agent is met since there is either a path or grass above the player.
5. Move West - Yes, the flat ground to the agent's left is met since there is either a path or grass to the player's west.

Question: Did the last player action succeed? Why?
Answer:
Yes, the last player action "move_west" succeeded because there was flat ground (grass) to the west of the player, allowing them to move in that direction.

Question: Choose the best executable action from above.
Answer:
The best executable action from the list is "Move North" since its requirements are met, and it will bring the player closer to the cow.



============Step: 131, Cumulative Reward: 0.0============
You took action move_north.

You see:
- stone 3 steps to your east
- path 3 steps to your south-east
- tree 4 steps to your west
- lava 7 steps to your north-east
- cow 3 steps to your north-west

You face grass at your front (north).

Your status:
- health: 9/9
- food: 5/9
- drink: 4/9
- energy: 9/9

Your inventory:
- wood: 1
- stone: 4
- coal: 2
- iron: 1
- wood_pickaxe: 1
- stone_pickaxe: 1
------------ QA ------------
Question: Action
Answer:
Move West

Question: List objects in the current observation. For each object, breifly answer what resource it provides and its requirement.
Answer:
1. Stone: Provides stone resource, requires a wood or better pickaxe to mine.
2. Path: Provides no direct resources, but might lead to other locations or resources.
3. Tree: Provides wood resource, requires an axe or can be punched to collect.
4. Lava: No direct resource, but can be used as a fuel source if you have a bucket to collect it.
5. Cow: Provides meat and leather when killed, requires a weapon or tool to attack.
6. Grass: Provides wheat seeds when broken, requires no tool to collect.

Question: What was the last action taken by the player?
Answer:
The last action taken by the player was "move_north."

Question: For each object in the list, are the requirements met for interaction?
Answer:
1. Stone: Yes, you have a wood_pickaxe and a stone_pickaxe in your inventory, so you can mine the stone.
2. Path: Yes, there are no requirements to follow a path; you can simply walk along it.
3. Tree: Yes, while you do not have an axe, you can still gather wood by punching the tree with your hands (no tool required).
4. Lava: No, you do not have a bucket in your inventory to collect the lava as a fuel source.
5. Cow: Yes, you can attack the cow with a weapon (like the stone_pickaxe in your inventory) or even with your hands.
6. Grass: Yes, there are no requirements to collect wheat seeds from grass; you can break it with any tool or your hands.

Question: List top 3 sub-tasks the player should follow. Indicate their priority out of 5.
Answer:
1. Collect more wood from the tree (Priority: 4/5) - Gathering additional wood is important for crafting tools, shelters, and other items in the game.

2. Follow the path to find other resources or locations (Priority: 3/5) - Exploring the path may lead you to other valuable resources, and it might be worth your time to investigate.

3. Kill the cow for meat and leather (Priority: 2/5) - Killing the cow will provide you with food (meat) and possibly leather, useful for crafting armor and increasing your survival chances. However, make sure to preserve some cows for breeding purposes.

Question: What are the requirements for the top sub-task? What should the player do first?
Answer:
The requirements for the top sub-task (Collect more wood from the tree) are:

1. Be within reach of a tree.
2. Have a tool (preferably an axe) or use your hands to collect wood.

Given the current situation, the first thing the player should do is move towards the tree, which is 4 steps to the west. Then, use the wood_pickaxe or your hands to gather wood from the tree.

Question: List top 5 actions the player should take and the requirement for each action. Choose ONLY from the list of all actions. Indicate their priority out of 5.
Answer:
1. Priority 1: Move West: Requirement: You have flat ground to your west. (This will bring you closer to the tree, which is 4 steps west)

2. Priority 2: Move West (again): Requirement: Same as step 1. (Continuing to move towards the tree)

3. Priority 3: Move West (one more time): Requirement: Same as step 1. (You will be right in front of the tree)

4. Priority 4: Do (Gather wood from the tree): Requirement: You are facing a tree and are within reach; have a wood_pickaxe, an axe (if available), or use your hands.

5. Priority 5: Move South: Requirement: You have flat ground to your south. (Returning to the path to proceed further)

Question: For each action in the list, are the requirements met?
Answer:
1. Priority 1: Move West: Requirement: You have flat ground to your west. (This will bring you closer to the tree, which is 4 steps west)
- Yes, the requirement is met; you face grass at your front (west).

2. Priority 2: Move West (again): Requirement: Same as step 1. (Continuing to move towards the tree)
- Cannot be determined from the current observation. You need to observe your surroundings after completing Priority 1 action.

3. Priority 3: Move West (one more time): Requirement: Same as step 1. (You will be right in front of the tree)
- Same as Priority 2 action, we cannot determine if the requirement is met from the current observation. Observe the surroundings after completing Priority 2 action.

4. Priority 4: Do (Gather wood from the tree): Requirement: You are facing a tree and are within reach; have a wood_pickaxe, an axe (if available), or use your hands.
- Cannot be determined from the current observation. You need to observe your surroundings after completing Priority 3 action to confirm if the requirement is met.

5. Priority 5: Move South: Requirement: You have flat ground to your south. (Returning to the path to proceed further)
- Yes, the requirement is met; since you have just moved north in the previous step, the ground to your south should be flat and navigable.

Question: Did the last player action succeed? Why?
Answer:
Yes, the last player action "move_north" succeeded because there was flat ground (grass) in the north direction where the player was facing, allowing the player to move there.

Question: Choose the best executable action from above.
Answer:
The best executable action from the list is Priority 1: Move West because its requirement (having flat ground to your west) is met, and it will bring you closer to the tree for gathering wood.



============Step: 132, Cumulative Reward: 0.0============
You took action move_west.

You see:
- stone 4 steps to your east
- path 4 steps to your south-east
- tree 3 steps to your west
- cow 3 steps to your north-west

You face grass at your front (west).

Your status:
- health: 9/9
- food: 5/9
- drink: 4/9
- energy: 9/9

Your inventory:
- wood: 1
- stone: 4
- coal: 2
- iron: 1
- wood_pickaxe: 1
- stone_pickaxe: 1
------------ QA ------------
Question: Action
Answer:
Place Table

Question: List objects in the current observation. For each object, breifly answer what resource it provides and its requirement.
Answer:
1. Stone: Provides stone resource; requires a wood_pickaxe or better to collect.
2. Path: Provides easier movement; no requirement to use.
3. Tree: Provides wood resource; requires no specific tool to collect, but a wood_pickaxe or better is recommended.
4. Lava: Dangerous, avoid getting too close; no direct resource provided.
5. Cow: Provides food when defeated; no specific tool required, but using a sword or other weapon is recommended.
6. Grass: Provides a place to walk on; no requirement to use.
7. Wood_pickaxe (in inventory): Used for collecting wood and stone; requires wood to craft.
8. Stone_pickaxe (in inventory): Used for collecting iron and better materials; requires wood and stone to craft.

Question: What was the last action taken by the player?
Answer:
The last action taken by the player was move_west.

Question: For each object in the list, are the requirements met for interaction?
Answer:
1. Stone: Yes, the requirements are met. You have a wood_pickaxe and stone_pickaxe in your inventory, which allows you to collect stone.
2. Path: Yes, the requirements are met. The path provides easier movement, and you can walk on it without any specific tool or requirement.
3. Tree: Yes, the requirements are met. You have a wood_pickaxe in your inventory, which can be used to collect wood from the tree more efficiently.
4. Lava: N/A, interaction is not recommended as lava is dangerous and you should avoid getting too close.
5. Cow: Yes, the requirements are met. Although no specific tool is required, using a weapon to defeat the cow for food is recommended. You can create a sword with the materials in your inventory.
6. Grass: Yes, the requirements are met. You can walk on the grass without any specific tool or requirement.
7. Wood_pickaxe (in inventory): N/A, this object is already crafted and in your inventory.
8. Stone_pickaxe (in inventory): N/A, this object is already crafted and in your inventory.

Question: List top 3 sub-tasks the player should follow. Indicate their priority out of 5.
Answer:
1. Priority 4/5: Craft a weapon (such as a wood or stone sword) using materials in your inventory to protect yourself and defeat the cow for food, as your food level is 5/9.

2. Priority 5/5: Locate and drink from the closest water source (lake, river, etc.) as your drink level is 4/9.

3. Priority 3/5: Collect more wood and stone resources by interacting with the nearby tree and stone objects to build a shelter and prepare for future crafting needs.

Question: What are the requirements for the top sub-task? What should the player do first?
Answer:
Requirements for crafting a weapon (wood or stone sword):
- A table to craft the weapon
- Sufficient wood and stone in the inventory

Since you already have enough wood and stone in your inventory, the first thing you should do is to place a table using your available wood. Afterwards, you can use the table to craft a weapon, either a wood or stone sword, to protect yourself and defeat the cow for food.

Question: List top 5 actions the player should take and the requirement for each action. Choose ONLY from the list of all actions. Indicate their priority out of 5.
Answer:
1. Place Table (Priority 1): 
   - Requirement: Wood in inventory

2. Make Stone Sword (Priority 2): 
   - Requirement: Nearby table and wood, stone in inventory

3. Move North (Priority 3): 
   - Requirement: Flat ground above the agent

4. Perform Do with Stone Sword (Priority 4): 
   - Requirement: Facing the cow and having a stone sword

5. Drink from a nearby water source or lake (Priority 5): 
   - Requirement: Locate and face water source

These actions will allow the player to set up a crafting table, create a weapon to defend themselves, get closer to the cow, defeat the cow for food, and find and drink water to replenish their thirst.

Question: For each action in the list, are the requirements met?
Answer:
1. Place Table (Priority 1): 
   - Requirement: Wood in inventory
   - Met: Yes (player has 1 wood)

2. Make Stone Sword (Priority 2): 
   - Requirement: Nearby table and wood, stone in inventory
   - Met: No (player needs to place the table first)

3. Move North (Priority 3): 
   - Requirement: Flat ground above the agent
   - Met: Yes (player faces grass to the north)

4. Perform Do with Stone Sword (Priority 4): 
   - Requirement: Facing the cow and having a stone sword
   - Met: No (player needs to make a stone sword first and get closer to the cow)

5. Drink from a nearby water source or lake (Priority 5): 
   - Requirement: Locate and face water source
   - Met: No (player needs to find and face a water source)

Question: Did the last player action succeed? Why?
Answer:
Yes, the last player action succeeded. The player was able to move west because there was flat ground (grass) in front of the player in that direction.

Question: Choose the best executable action from above.
Answer:
The best executable action from the above list is:

1. Place Table (Priority 1):
   - Requirement: Wood in inventory
   - Met: Yes (player has 1 wood)



============Step: 133, Cumulative Reward: 0.0============
You took action place_table.

You see:
- stone 4 steps to your east
- path 4 steps to your south-east
- tree 3 steps to your west
- cow 2 steps to your north

You face grass at your front (west).

Your status:
- health: 9/9
- food: 5/9
- drink: 4/9
- energy: 9/9

Your inventory:
- wood: 1
- stone: 4
- coal: 2
- iron: 1
- wood_pickaxe: 1
- stone_pickaxe: 1
------------ QA ------------
Question: Action
Answer:
Move West

Question: List objects in the current observation. For each object, breifly answer what resource it provides and its requirement.
Answer:
1. Stone: Provides stone resource, requires a wood or better pickaxe to collect.
2. Path: Provides easier movement, no requirements to interact.
3. Tree: Provides wood resource, requires an axe or hands to collect.
4. Cow: Provides food resource, requires a sword or other weapon to defeat.
5. Grass: Provides no critical resources, no interaction requirements.

Question: What was the last action taken by the player?
Answer:
The last action taken by the player was "place_table."

Question: For each object in the list, are the requirements met for interaction?
Answer:
1. Stone: Yes, the requirements are met since you have a wood_pickaxe and a stone_pickaxe in your inventory.
2. Path: Yes, the requirements are met as there are no specific requirements to interact with the path.
3. Tree: Yes, the requirements are met even though you don't have an axe, you can still collect wood using your hands.
4. Cow: Yes, the requirements are met even though you don't have a sword, you can use other means to defeat the cow (such as the wood or stone pickaxe) to obtain food.
5. Grass: Yes, the requirements are met as there are no specific requirements to interact with the grass.

Question: List top 3 sub-tasks the player should follow. Indicate their priority out of 5.
Answer:
1. Collect more wood (Priority: 5/5) - Wood is essential for crafting more tools, and you only have one piece of wood in your inventory. Collecting more wood would allow you to build more advanced tools like a sword for defense or an axe for faster wood collection.

2. Restore drink level (Priority: 4/5) - Your drink level is low at 4/9. Find a clean water source like a lake or river and drink from it to restore your thirst level.

3. Defeat the cow for food (Priority: 3/5) - Your food level is at 5/9, which is not critical yet, but defeating the cow would provide you with a food resource to restore your health later when needed. Use one of your pickaxes as a weapon until you can craft a sword for more efficient combat.

Question: What are the requirements for the top sub-task? What should the player do first?
Answer:
Requirements for the top sub-task (Collect more wood):
1. Locate a nearby tree.
2. Have a tool for chopping wood, such as a wood/stone pickaxe or craft an axe for more efficient wood collection if you have enough materials.
 
What should the player do first:
- The player should first move towards the nearby tree, which is 3 steps to their west; move_west for 3 steps in a row to reach the tree.

Question: List top 5 actions the player should take and the requirement for each action. Choose ONLY from the list of all actions. Indicate their priority out of 5.
Answer:
1. Priority 1: Move West
   Requirement: Flat ground to the west of the agent
   Purpose: Reach the tree 3 steps to the west for collecting additional wood

2. Priority 2: Move West
   Requirement: Flat ground to the west of the agent
   Purpose: Continue moving towards the tree for collecting additional wood

3. Priority 3: Move West
   Requirement: Flat ground to the west of the agent
   Purpose: Reach the tree for collecting additional wood

4. Priority 4: Do
   Requirement: Facing a tree and having a necessary tool, such as a wood or stone pickaxe, which the player already has in their inventory
   Purpose: Chop down the tree and collect wood

5. Priority 5: Drink
   Requirement: Must be facing or standing next to a drinkable water source such as a lake or river
   Purpose: Restore the player's thirst level

Question: For each action in the list, are the requirements met?
Answer:
1. Priority 1: Move West
   Requirement: Flat ground to the west of the agent
   Status: Met - The player faces grass at their front (west), indicating the ground is flat.

2. Priority 2: Move West
   Requirement: Flat ground to the west of the agent
   Status: Can't be determined from the current observation - After the player moves west one step, we will need to analyze the new observation to check if the ground is still flat to the west.

3. Priority 3: Move West
   Requirement: Flat ground to the west of the agent
   Status: Can't be determined from the current observation - The player must move twice before checking if the ground is still flat to the west.

4. Priority 4: Do
   Requirement: Facing a tree and having a necessary tool, such as a wood or stone pickaxe, which the player already has in their inventory
   Status: Partially met - The player has the necessary tools (wood_pickaxe and stone_pickaxe) in their inventory. The player must move closer to the tree and face it before they can use 'Do' to chop down the tree.

5. Priority 5: Drink
   Requirement: Must be facing or standing next to a drinkable water source such as a lake or river
   Status: Not met - The current observation does not mention any drinkable water sources nearby.

Question: Did the last player action succeed? Why?
Answer:
Yes, the last player action succeeded because the player had enough wood in their inventory to place a table, and there were no obstacles or restrictions mentioned in the observations that would prevent them from doing so.

Question: Choose the best executable action from above.
Answer:
The best executable action from the list above is Priority 1: Move West, as its requirement of flat ground to the west of the agent is met.



============Step: 134, Cumulative Reward: 0.0============
You took action move_west.

You see:
- path 5 steps to your south-east
- tree 2 steps to your west
- cow 3 steps to your north-east

You face grass at your front (west).

Your status:
- health: 9/9
- food: 5/9
- drink: 4/9
- energy: 9/9

Your inventory:
- wood: 1
- stone: 4
- coal: 2
- iron: 1
- wood_pickaxe: 1
- stone_pickaxe: 1
------------ QA ------------
Question: Action
Answer:
Make Stone Sword

Question: List objects in the current observation. For each object, breifly answer what resource it provides and its requirement.
Answer:
1. Stone: The stone provides material for crafting tools, weapons, and furnaces. The requirement to collect stone is to have a wood pickaxe.

2. Path: The path is not a resource provider, but it serves as a safe passage to navigate the game world without facing resistance from the terrain.

3. Tree: The tree provides wood, which is needed for crafting tools, weapons, and furniture. No specific requirements are needed to collect wood, but having an axe can speed up the process.

4. Cow: The cow provides food (meat) to restore the player's food level. The requirement to collect food from the cow is to have a weapon, like a wood or stone sword, to defeat it.

5. Grass: Grass does not provide any resources directly but can be a flat surface to move on or place items. No requirements are needed to interact with grass.

Question: What was the last action taken by the player?
Answer:
The last action taken by the player was move_west.

Question: For each object in the list, are the requirements met for interaction?
Answer:
1. Stone: Yes, the requirements are met for interaction, as the player has a wood_pickaxe in their inventory.

2. Path: As there are no requirements for interacting with the path, this can be considered as met.

3. Tree: Yes, the requirements are met for interaction. While there's no specific requirement for collecting wood, having an axe can speed up the process. However, even without an axe, the player can still collect wood.

4. Cow: The player does not have a sword (wood or stone) in their inventory, so the requirements for interaction with the cow (to collect food) are not met.

5. Grass: As there are no requirements for interacting with grass, this can be considered as met.

Question: List top 3 sub-tasks the player should follow. Indicate their priority out of 5.
Answer:
1. Craft a weapon (Priority: 5/5): Crafting either a wood sword or a stone sword should be the player's top priority as they need a weapon to defend themselves and defeat cows to collect food.

2. Collect food (Priority: 4/5): The player has a low food level (5/9), so defeating a cow is crucial for obtaining food and replenishing their food meter. This task will become possible once the player has crafted a weapon.

3. Collect drink (Priority: 3/5): The player's drink level is at 4/9, so finding a water source, like a lake or river, will help replenish their thirst. This task can be performed while focusing on crafting a weapon and collecting food.

Question: What are the requirements for the top sub-task? What should the player do first?
Answer:
For the top sub-task (Crafting a weapon), the requirements are:

1. Crafting a Wood Sword:
   - Requires a nearby placed table
   - Wood in inventory (which the player already has: 1 wood)

2. Crafting a Stone Sword:
   - Requires a nearby placed table
   - Wood and stone in inventory (which the player already has: 1 wood and 4 stone)

The player already placed a table in Step 266, and they have the necessary materials in their inventory. Therefore, the player should craft a weapon (either a wood or stone sword) first using the placed table. Crafting a stone sword would be more effective, as it is a stronger weapon compared to the wood sword.

Question: List top 5 actions the player should take and the requirement for each action. Choose ONLY from the list of all actions. Indicate their priority out of 5.
Answer:
1. Priority 1: Make Stone Sword
   - Requirements: Nearby placed table, wood, and stone in inventory
     (player has: 1 wood, 4 stone, and placed table)

2. Priority 2: Move North-East
   - Requirements: Flat ground to the north-east of the agent
     (player faces grass in front, and there's a cow 3 steps to the north-east)

3. Priority 3: Do (attack cow) 
   - Requirements: Facing a creature (cow); have necessary tool (stone sword)
     (player will have stone sword after Priority 1, and cow nearby after Priority 2)

4. Priority 4: Eat Cow
   - Requirements: No specific requirements
     (player's current food level is 5/9, eating cow will restore health)

5. Priority 5: Move South-East
   - Requirements: Flat ground to the south-east of the agent
     (there's a path 5 steps to the player's south-east)

Question: For each action in the list, are the requirements met?
Answer:
1. Priority 1: Make Stone Sword
   - Yes, requirements are met: the player has a nearby placed table, wood (1), and stone (4) in the inventory.

2. Priority 2: Move North-East
   - Yes, requirements are met: the player faces grass, and there's a flat ground in the north-east direction.

3. Priority 3: Do (attack cow) 
   - The requirements will be met after completing Priority 1 and Priority 2 actions: the player will have a stone sword, and the cow will be nearby after moving north-east.

4. Priority 4: Eat Cow
   - The requirements will be met after completing Priority 3 action: the player will attack and defeat the cow to eat it and restore health.

5. Proiirty 5: Move South-East
   - Yes, the requirements are met: the player has a flat ground to the south-east direction, and there's a path 5 steps away.

Question: Did the last player action succeed? Why?
Answer:
Yes, the last player action, move_west, succeeded because there was flat ground (grass) to the west, allowing the player to move in that direction.

Question: Choose the best executable action from above.
Answer:
The best executable action from the list is Priority 1: Make Stone Sword, as the requirements are met and it is the first priority action. The player has a nearby placed table, wood (1), and stone (4) in the inventory, which are required for crafting the stone sword.



============Step: 135, Cumulative Reward: 0.0============
You took action make_stone_sword.

You see:
- path 5 steps to your south-east
- tree 2 steps to your west
- cow 3 steps to your north-east

You face grass at your front (west).

Your status:
- health: 9/9
- food: 5/9
- drink: 4/9
- energy: 9/9

Your inventory:
- wood: 1
- stone: 4
- coal: 2
- iron: 1
- wood_pickaxe: 1
- stone_pickaxe: 1
------------ QA ------------
Question: Action
Answer:
Move West

Question: List objects in the current observation. For each object, breifly answer what resource it provides and its requirement.
Answer:
1. Path: Provides a way to navigate the terrain with ease. No requirement.
2. Tree: Provides wood when chopped. Requires a wood or better pickaxe to chop.
3. Cow: Provides food when killed. Requires a weapon (e.g., wood/stone/iron sword) to kill.
4. Grass: Provides no resources. No requirement. This is what the player currently faces.

Note: The player's inventory and status are not objects in the observation, but provide information about the player's resources and health.

Question: What was the last action taken by the player?
Answer:
The last action taken by the player was "make_stone_sword."

Question: For each object in the list, are the requirements met for interaction?
Answer:
1. Path: No requirement to interact. The path is 5 steps to your south-east, so the player can move towards it.

2. Tree: Requires a wood or better pickaxe to chop. The player has both a wood_pickaxe and stone_pickaxe in their inventory, so the requirements are met.

3. Cow: Requires a weapon to kill. Although the player just made a stone sword in the previous step, the current observation does not show the stone sword in the player's inventory. This observation might be outdated or there might be an error in the game state. Based on this observation, the requirements are not met.

4. Grass: No resource provided and no requirement for interaction. The player can move forward (west) without any issue.

Question: List top 3 sub-tasks the player should follow. Indicate their priority out of 5.
Answer:
1. Priority 4/5: Find a source of water to drink. The player's thirst level is only 4/9, so finding water would be crucial to replenish their thirst.

2. Priority 3/5: Chopping the tree to collect more wood. The player only has 1 wood in their inventory, and having more wood would be helpful for future crafting and building activities.

3. Priority 5/5: Check the player's inventory for the recently crafted stone sword. If the sword is missing, try crafting another stone sword to be equipped for any combat situations that might arise, such as defending against monsters or killing the cow for food. The player's food level is 5/9, so hunting the cow could help to replenish this resource.

Question: What are the requirements for the top sub-task? What should the player do first?
Answer:
For the top sub-task of finding a source of water to drink:

Requirements:
1. Locate a body of water such as a lake, river, or pond.
2. Have an empty container to collect the water or be close enough to drink directly from the source.

The player should first explore their surroundings to locate a body of water nearby. They can move in any direction (north, south, east, or west) to find a water source. Once they find a water source, they should interact with it to drink and replenish their thirst level. If they have an empty container, they can use it to collect and store water for later consumption.

Question: List top 5 actions the player should take and the requirement for each action. Choose ONLY from the list of all actions. Indicate their priority out of 5.
Answer:
1. Priority 1: Move North, East, South, or West (depending on the direction of the water source) - Requirement: Flat ground in the chosen direction.
2. Priority 2: Drink water (Do) - Requirement: Facing a water source (lake, river, or pond); close enough to interact.
3. Priority 3: Move North, East, South, or West (to approach a tree for collecting wood) - Requirement: Flat ground in the chosen direction.
4. Priority 4: Collect Wood (Do) - Requirement: Facing a tree; have the necessary tool (wood_pickaxe) in the inventory.
5. Priority 5: Craft a Wood Sword (make_wood_sword) - Requirement: Nearby table; wood in the inventory.

Question: For each action in the list, are the requirements met?
Answer:
1. Priority 1: Move North, East, South, or West (to find a water source) - Yes, the requirement of flat ground in the chosen direction is met. However, the player needs to discover a water source first.
2. Priority 2: Drink water (Do) - No, the requirement of facing a water source is not met. The player needs to find a water source before they can drink.
3. Priority 3: Move North, East, South, or West (to approach a tree for collecting wood) - Yes, the requirement of flat ground in the chosen direction is met. The tree is 2 steps to the west.
4. Priority 4: Collect Wood (Do) - No, the player is not currently facing a tree. They must move closer to the tree and face it to meet the requirement.
5. Priority 5: Craft a Wood Sword (make_wood_sword) - No, the requirement of having a nearby table is not met. The player needs to place a table first to craft a wood sword.

Question: Did the last player action succeed? Why?
Answer:
It is unclear whether the last player action "make_stone_sword" succeeded, as there is no indication of success or failure in the provided observations. However, the player's inventory does not show a stone sword, which suggests that the action might not have succeeded.

Question: Choose the best executable action from above.
Answer:
The best executable action from the list above is Priority 3: Move West (to approach a tree for collecting wood), as it satisfies the requirement of flat ground in the chosen direction and helps the player get closer to a valuable resource.



============Step: 136, Cumulative Reward: 0.0============
You took action move_west.

You see:
- grass 1 steps to your north
- cow 4 steps to your north-east

You face tree at your front (west).

Your status:
- health: 9/9
- food: 5/9
- drink: 3/9
- energy: 9/9

Your inventory:
- wood: 1
- stone: 4
- coal: 2
- iron: 1
- wood_pickaxe: 1
- stone_pickaxe: 1
------------ QA ------------
Question: Action
Answer:
Make Wood Sword

Question: List objects in the current observation. For each object, breifly answer what resource it provides and its requirement.
Answer:
1. Path: The path provides easy navigation and requires no interaction.

2. Tree (2 steps to the west): The tree provides wood when chopped with a wood or stone pickaxe. The requirement is having a wood_pickaxe or stone_pickaxe in your inventory.

3. Cow (4 steps to the north-east): The cow provides food when killed. The requirement is having a wood, stone, or iron sword in your inventory to kill the cow.

4. Grass (1 step to the north): The grass provides no significant resource but can be cleared for navigation or construction.

Your inventory resources:

5. Wood: Wood is a basic crafting material and is required for crafting items like pickaxes, swords, and tables.

6. Stone: Stone is a crafting material and is required for crafting items like stone pickaxes, swords, and furnaces.

7. Coal: Coal is a fuel resource used for smelting iron in the furnace.

8. Iron: Iron is a crafting material and is required for crafting items like iron pickaxes and swords.

9. Wood_pickaxe: The wood pickaxe allows you to collect stone and coal. It requires wood for crafting.

10. Stone_pickaxe: The stone pickaxe allows you to collect iron and has a faster chopping speed. It requires wood and stone for crafting.

Question: What was the last action taken by the player?
Answer:
The last action taken by the player was "move_west".

Question: For each object in the list, are the requirements met for interaction?
Answer:
1. Path: Yes, the requirements are met for interaction as there are no additional requirements to use the path.

2. Tree (2 steps to the west): Yes, the requirements are met for interaction since you have both a wood_pickaxe and a stone_pickaxe in your inventory.

3. Cow (4 steps to the north-east): The requirements for interaction are not met since you do not have a wood, stone, or iron sword in your inventory to kill the cow.

4. Grass (1 step to the north): Yes, the requirements are met for interaction as there are no additional requirements to clear or walk over the grass.

Your inventory resources:

5. Wood: Yes, the requirements are met for using wood as a crafting material in recipes that call for it.

6. Stone: Yes, the requirements are met for using stone as a crafting material in recipes that call for it.

7. Coal: Yes, the requirements are met for using coal as a fuel resource in the furnace.

8. Iron: Yes, the requirements are met for using iron as a crafting material in recipes that call for it.

9. Wood_pickaxe: Yes, the requirements are met for using the wood pickaxe since you have one in your inventory.

10. Stone_pickaxe: Yes, the requirements are met for using the stone pickaxe since you have one in your inventory.

Question: List top 3 sub-tasks the player should follow. Indicate their priority out of 5.
Answer:
1. Increase food level (Priority: 5/5): Since the food level is 5/9, it is important to find and kill a cow or develop another food source to restore the food level.

2. Increase drink level (Priority: 4/5): The drink level is 3/9, which is low, and the player should look for a water source like a lake or river to drink from and restore the drink level.

3. Craft a sword (Priority: 3/5): Crafting a wood, stone, or iron sword is necessary to defend against monsters and kill the cow for food. The player has enough resources in their inventory to craft a sword at a table.

Question: What are the requirements for the top sub-task? What should the player do first?
Answer:
The top sub-task is to increase the food level. The requirements for this task are:

1. Locate a food source, such as a cow or a fruit, to kill or harvest.
2. Have a weapon (wood, stone, or iron sword) to kill the cow.

The player should first craft a sword since they already have the necessary resources in their inventory. They will need to find and place a table, and then craft the sword using the materials available. Once they have a sword, they can kill the cow, which is 4 steps to their north-east, to obtain food and increase their food level.

Question: List top 5 actions the player should take and the requirement for each action. Choose ONLY from the list of all actions. Indicate their priority out of 5.
Answer:
1. Make Wood Sword (Priority 5): This action requires the player to be near a table and have wood in their inventory. The player currently has 1 wood in their inventory, which is enough to make a wood sword. This action will provide the player with a weapon to defend themselves and kill the cow for food.

2. Move North (Priority 4): This action requires a flat ground above of the agent. The player wants to get closer to the cow, which is 4 steps to their north-east. Moving north will bring the player one step closer to the cow.

3. Move East (Priority 3): This action requires a flat ground to the right of the agent. After moving north, moving east will bring the player even closer to the cow, which is their primary target for obtaining food.

4. Do (Priority 2): This action requires the player to face a creature (in this case, the cow) and have the necessary tool (the wood sword they crafted). Once the player is close enough to the cow and facing it, they can perform the "Do" action to kill the cow and collect food.

5. Sleep (Priority 1): This action requires the player's energy level to be below maximum. After obtaining food from the cow, the player may choose to sleep (when necessary) to restore energy and allow time for other resources (like trees, fruits, etc.) to spawn. However, this should only be done when it is safe to do so (e.g., near a shelter or a protected area).

Question: For each action in the list, are the requirements met?
Answer:
1. Make Wood Sword: Yes, the requirements are met. The player has 1 wood in their inventory, and they have previously built a table (from previous observations).

2. Move North: Yes, the requirements are met. The flat ground above the agent (grass) is available for the player to move north.

3. Move East: Yes, the requirements are met. After moving north, the flat ground would be available to the agent's right for them to move east.

4. Do: No, the requirements are not met yet. The player first needs to get closer to the cow (by performing the "Move North" and "Move East" actions) and face the cow before they can perform the "Do" action using the wood sword.

5. Sleep: Yes, the requirements are met. The player's energy level is currently 9/9, which is below the maximum. They can sleep when it is deemed necessary and safe to do so. However, it is not currently an urgent action.

Question: Did the last player action succeed? Why?
Answer:
Yes, the last player action "move_west" succeeded. The player was able to move west because there were no obstacles in the way and the ground was flat.

Question: Choose the best executable action from above.
Answer:
The best executable action from the list above is:

1. Make Wood Sword: The player has 1 wood in their inventory, and they have previously built a table. Crafting a wood sword will provide the player with a weapon to defend themselves and kill the cow for food.



============Step: 137, Cumulative Reward: 0.0============
You took action make_wood_sword.

You see:
- grass 1 steps to your north
- cow 5 steps to your north-east

You face tree at your front (west).

Your status:
- health: 9/9
- food: 5/9
- drink: 3/9
- energy: 8/9

Your inventory:
- wood: 1
- stone: 4
- coal: 2
- iron: 1
- wood_pickaxe: 1
- stone_pickaxe: 1
------------ QA ------------
Question: Action
Answer:
Move North

Question: List objects in the current observation. For each object, breifly answer what resource it provides and its requirement.
Answer:
1. Grass: Provides no resource but can be used as a flat ground to move on; no requirement.
2. Cow: Provides food when the player decides to eat it; requires wood sword (crafted in previous step) to be able to eat the cow.
3. Tree: Provides wood when the player decides to collect it; requires wood_pickaxe in inventory (already available).
4. Wood (in inventory): Required for crafting wood swords, wood pickaxes, or placing tables; obtained from trees.
5. Stone (in inventory): Required for crafting stone swords, stone pickaxes, or placing furnaces; obtained using wood_pickaxe or stone_pickaxe on stone block.
6. Coal (in inventory): Required for smelting and crafting iron tools and weapons; obtained using wood_pickaxe or stone_pickaxe on coal block.
7. Iron (in inventory): Required for crafting iron tools and weapons; obtained using stone_pickaxe on iron block.
8. Wood_pickaxe (in inventory): Used for collecting resources like wood, stone, and coal; no requirement.
9. Stone_pickaxe (in inventory): Used for collecting resources like stone, coal, and iron; requires crafting at the table with wood and stone available in inventory.

Question: What was the last action taken by the player?
Answer:
The last action taken by the player was making a wood sword.

Question: For each object in the list, are the requirements met for interaction?
Answer:
1. Grass: No specific interaction required, the requirement for moving on grass (flat ground surrounding the player) is met.
2. Cow: Yes, the wood_sword is available in the inventory, so the player can interact with and eat the cow.
3. Tree: Yes, the wood_pickaxe is available in the inventory, so the player can interact with and collect wood from the tree.
4. Wood (in inventory): N/A (already collected)
5. Stone (in inventory): N/A (already collected)
6. Coal (in inventory): N/A (already collected)
7. Iron (in inventory): N/A (already collected)
8. Wood_pickaxe (in inventory): N/A (already crafted)
9. Stone_pickaxe (in inventory): N/A (already crafted)

Question: List top 3 sub-tasks the player should follow. Indicate their priority out of 5.
Answer:
1. Drink water to restore thirst (Priority: 5/5): Locate a nearby lake or river to drink water, as the player's thirst level is at 3/9.

2. Collect more wood (Priority: 4/5): Collect wood from the tree in front of the player (to the west) using the wood_pickaxe to ensure enough resources for crafting and building in the future.

3. Eat the cow to restore food (Priority: 3/5): Move towards the cow by going north-east and eat the cow using the wood_sword to restore the player's food level, which is currently at 5/9.

Question: What are the requirements for the top sub-task? What should the player do first?
Answer:
Top Sub-Task: Drink water to restore thirst (Priority: 5/5)

Requirements:
1. Locate a nearby lake or river as the source of water.
2. Move towards the water source.

The player should first locate a nearby lake or river as a source of water to drink from. Once the water source is found, the player should move towards it in order to drink and restore their thirst level.

Question: List top 5 actions the player should take and the requirement for each action. Choose ONLY from the list of all actions. Indicate their priority out of 5.
Answer:
1. Action: Move East (Priority: 5/5)
   Requirement: Flat ground right to the agent
   Reason: Move towards the water source for drinking.

2. Action: Move North (Priority: 5/5)
   Requirement: Flat ground above the agent
   Reason: Move towards the water source for drinking.

3. Action: Move West (Priority: 4/5)
   Requirement: Flat ground left to the agent
   Reason: Explore the surroundings and look for water sources, if not found to the east or north.

4. Action: Move South (Priority: 4/5)
   Requirement: Flat ground below the agent
   Reason: Explore the surroundings and look for water sources, if not found to the east or north.

5. Action: Do (Priority: 3/5)
   Requirement: Facing creature or material; have necessary tool
   Reason: Interact with any creatures or materials that can be helpful for survival, like collecting resources or defending against monsters.

Question: For each action in the list, are the requirements met?
Answer:
1. Action: Move East
   Requirement: Flat ground right to the agent
   Yes, the requirement is met as there is no information indicating that the ground to the agent's right is not flat.

2. Action: Move North
   Requirement: Flat ground above the agent
   Yes, the requirement is met as the agent can see grass one step to the north, indicating that the ground is flat.

3. Action: Move West
   Requirement: Flat ground left to the agent
   Yes, the requirement is met since the agent successfully took this action in the previous step (Step 272).

4. Action: Move South
   Requirement: Flat ground below the agent
   There is no information about the ground to the agent's south in the player observations. Thus, it is not possible to determine if the requirement is met.

5. Action: Do
   Requirement: Facing creature or material; have necessary tool
   The agent is currently facing a tree (material), but there is no information about whether the agent has the necessary tool (wood_axe) in the inventory. Thus, it is not possible to determine if the requirement is met.

Question: Did the last player action succeed? Why?
Answer:
It's not clear if the last player action succeeded, as there is no mention of a wood sword in the inventory in the Player Observation Step 274. Additionally, the observation numbers seem to skip from 272 to 274, so there might be missing information.

Question: Choose the best executable action from above.
Answer:
The best executable action from the list is to "Move North," as it meets the requirement (flat ground above the agent) and brings the player closer to the water source for drinking.



============Step: 138, Cumulative Reward: 0.0============
You took action move_north.

You see:
- grass 1 steps to your west
- cow 3 steps to your east

You face tree at your front (north).

Your status:
- health: 9/9
- food: 5/9
- drink: 3/9
- energy: 8/9

Your inventory:
- wood: 1
- stone: 4
- coal: 2
- iron: 1
- wood_pickaxe: 1
- stone_pickaxe: 1
------------ QA ------------
Question: Action
Answer:
Move East

Question: List objects in the current observation. For each object, breifly answer what resource it provides and its requirement.
Answer:
1. Grass:
   Resource it provides: A flat ground for moving and placing objects.
   Requirement: None.

2. Cow:
   Resource it provides: Food (meat) when defeated.
   Requirement: A weapon (wood sword, stone sword, or iron sword) to defeat the cow.

3. Tree:
   Resource it provides: Wood when chopped down.
   Requirement: A tool (wood pickaxe, stone pickaxe, or iron pickaxe) to chop down the tree.

Question: What was the last action taken by the player?
Answer:
The last action taken by the player was move_north.

Question: For each object in the list, are the requirements met for interaction?
Answer:
1. Grass:
   Requirements met: Yes, the grass can be moved on or used for placing objects without any requirements.

2. Cow:
   Requirements met: Yes, you have a wood sword in your inventory which can be used to defeat the cow and collect food.

3. Tree:
   Requirements met: Yes, you have a wood_pickaxe and a stone_pickaxe in your inventory, both can be used to chop down the tree and collect wood.

Question: List top 3 sub-tasks the player should follow. Indicate their priority out of 5.
Answer:
1. Drink water to restore thirst (Priority: 5/5)
   Your drink level is currently at 3/9, so finding a source of water like a lake should be a priority to avoid dehydration.

2. Defeat the cow to collect food (Priority: 4/5)
   Your food level is at 5/9, and defeating the cow will provide you with meat to increase your food supply.

3. Chop down the tree to collect more wood (Priority: 3/5)
   Although you currently have one wood in your inventory, collecting more wood will allow you to craft additional tools and build items as needed.

Question: What are the requirements for the top sub-task? What should the player do first?
Answer:
Top sub-task: Drink water to restore thirst (Priority: 5/5)

Requirements: Find a water source such as a lake or river and drink from it.

What should the player do first: Explore the area by moving in different directions (north, south, east, or west) to locate a water source. Once a water source is found, approach it and perform the "drink" action to restore the thirst level.

Question: List top 5 actions the player should take and the requirement for each action. Choose ONLY from the list of all actions. Indicate their priority out of 5.
Answer:
1. Action: Move East (Priority: 5/5)
Requirement: Flat ground to the right (east) of the agent.
Purpose: Move closer to the cow, which could be used as a food source.

2. Action: Move North (Priority: 4/5)
Requirement: Flat ground above (north) of the agent.
Purpose: Explore the area to find a water source to restore thirst.

3. Action: Move West (Priority: 3/5)
Requirement: Flat ground to the left (west) of the agent.
Purpose: Search for a water source and other resources in the opposite direction.

4. Action: Move South (Priority: 2/5)
Requirement: Flat ground below (south) of the agent.
Purpose: Explore the surroundings to find a water source, food source or other resources.

5. Action: Sleep (Priority: 1/5)
Requirement: Energy level is below maximum.
Purpose: Restore energy levels, although currently the energy level is not a priority as it's at 8/9.

Question: For each action in the list, are the requirements met?
Answer:
1. Action: Move East
Requirement: Flat ground to the right (east) of the agent.
Status: Met, since there was no mention of an obstacle to the east.

2. Action: Move North
Requirement: Flat ground above (north) of the agent.
Status: Met, as the agent successfully moved north in the previous observation step.

3. Action: Move West
Requirement: Flat ground to the left (west) of the agent.
Status: Met, since there was no mention of an obstacle to the west.

4. Action: Move South
Requirement: Flat ground below (south) of the agent.
Status: Met, since there was no mention of an obstacle to the south.

5. Action: Sleep
Requirement: Energy level is below maximum.
Status: Not Met, as the agent's energy level is currently 8/9, and sleeping would only be needed when energy levels are lower or closer to depletion.

Question: Did the last player action succeed? Why?
Answer:
Yes, the last player action (move_north) succeeded because there was flat ground (grass) to the north of the player, allowing them to move in that direction.

Question: Choose the best executable action from above.
Answer:
The best executable action from the list is to move east (Action 1) as it has the highest priority (5/5) and its requirement is met. This action helps the player move closer to the cow, which can be used as a food source.



============Step: 139, Cumulative Reward: 0.0============
You took action move_east.

You see:
- path 6 steps to your south-east
- tree 2 steps to your north-west
- cow 2 steps to your east

You face grass at your front (east).

Your status:
- health: 9/9
- food: 5/9
- drink: 3/9
- energy: 8/9

Your inventory:
- wood: 1
- stone: 4
- coal: 2
- iron: 1
- wood_pickaxe: 1
- stone_pickaxe: 1
------------ QA ------------
Question: Action
Answer:
Do

Question: List objects in the current observation. For each object, breifly answer what resource it provides and its requirement.
Answer:
1. Grass: Provides no resources, no requirement.
2. Path: Provides no resources, no requirement.
3. Tree: Provides wood, requires wood_pickaxe or stone_pickaxe.
4. Cow: Provides food, requires wood_sword or stone_sword.

Question: What was the last action taken by the player?
Answer:
The last action taken by the player was move_east.

Question: For each object in the list, are the requirements met for interaction?
Answer:
1. Grass: No requirements needed for interaction; requirements met.
2. Path: No requirements needed for interaction; requirements met.
3. Tree: Requires wood_pickaxe or stone_pickaxe; you have a wood_pickaxe and a stone_pickaxe, so requirements met.
4. Cow: Requires wood_sword or stone_sword; you don't have either of them, so requirements not met.

Question: List top 3 sub-tasks the player should follow. Indicate their priority out of 5.
Answer:
1. Craft a wood_sword or a stone_sword (Priority: 5) - This will allow you to interact with and obtain resources from the cow, providing food to restore your food meter.

2. Find a water source to drink from (Priority: 4) - Drinking from a water source will help restore your drink meter, ensuring your survival.

3. Collect more wood (Priority: 3) - Wood is a crucial resource for crafting various essential tools and items; you currently have only 1 unit of wood, so collecting more will be useful.

Question: What are the requirements for the top sub-task? What should the player do first?
Answer:
For the top sub-task, which is crafting a wood_sword or a stone_sword, the requirements are as follows:

1. Wood Sword:
- Place a crafting table (requires wood in inventory)
- Wood (2 units) in inventory

2. Stone Sword:
- Place a crafting table (requires wood in inventory)
- Wood (1 unit) and stone (2 units) in inventory

Since the player already has 1 unit of wood and 4 units of stone, they can craft a stone_sword. However, the player needs to have a crafting table.

What the player should do first:
1. Place a crafting table using the wood from their inventory.
2. Craft a stone_sword using the wood and stone in their inventory.

Question: List top 5 actions the player should take and the requirement for each action. Choose ONLY from the list of all actions. Indicate their priority out of 5.
Answer:
1. Priority 1: Place Table (Place_Table)
Requirement: Wood in inventory (Player has 1 unit of wood)

2. Priority 2: Make Stone Sword (Make_Stone_Sword)
Requirement: Nearby table; wood (1 unit) and stone (2 units) in inventory (Player has a table, 1 unit of wood, and 4 units of stone)

3. Priority 3: Move to Cow (Move_East)
Requirement: Flat ground right to the agent (The cow is 2 steps to the player's east)

4. Priority 4: Defeat Cow (Do)
Requirement: Facing the creature and having a necessary tool (Player needs to be facing the cow and have a sword)

5. Priority 5: Drink Water (Collect_Drink)
Requirement: Player near a water source (Player's thirst level is 3/9, so drinking water should be a priority)

Question: For each action in the list, are the requirements met?
Answer:
1. Place Table (Place_Table): Yes, the requirement is met as the player has 1 unit of wood in their inventory.

2. Make Stone Sword (Make_Stone_Sword): No, the requirement is not met because the player needs a nearby table. However, since the player has the necessary materials (1 unit of wood and 4 units of stone), they can first place the table and then make the stone sword.

3. Move to Cow (Move_East): Yes, the requirement is met as the cow is 2 steps to the player's east and there is flat ground.

4. Defeat Cow (Do): No, the requirement is not met because the player needs to have a sword and be facing the cow. Once the player has crafted the stone sword and is facing the cow, they can complete this action.

5. Drink Water (Collect_Drink): No, the requirement is not met because the player is not close to a water source. They need to find a water source first.

Question: Did the last player action succeed? Why?
Answer:
Yes, the last player action succeeded. The player moved east because there was a grass, which is a flat ground, in front of the player, allowing them to move in that direction.

Question: Choose the best executable action from above.
Answer:
The best executable action from the list above is "Move to Cow (Move_East)," as the requirement is met with flat ground and the cow is 2 steps to the player's east.



============Step: 140, Cumulative Reward: 1.0============
You took action do.

You see:
- path 6 steps to your south-east
- tree 2 steps to your north-west
- cow 3 steps to your south-east

You face grass at your front (east).

Your status:
- health: 9/9
- food: 4/9
- drink: 3/9
- energy: 8/9

Your inventory:
- sapling: 1
- wood: 1
- stone: 4
- coal: 2
- iron: 1
- wood_pickaxe: 1
- stone_pickaxe: 1
------------ QA ------------
Question: Action
Answer:
Place Table

Question: List objects in the current observation. For each object, breifly answer what resource it provides and its requirement.
Answer:
1. Path: The path provides easier navigation and no specific resources. Requirement: None.
2. Tree: The tree provides wood as a resource. Requirement: Wood pickaxe.
3. Grass: Grass may contain saplings, no specific resources. Requirement: None.
4. Cow: The cow provides food when defeated. Requirement: Weapon, such as wood, stone, or iron sword.
5. Sapling (in inventory): Sapling can be placed to grow trees and provide wood. Requirement: None.

Question: What was the last action taken by the player?
Answer:
The last action taken by the player was "do."

Question: For each object in the list, are the requirements met for interaction?
Answer:
1. Path: No requirements to interact with the path, so yes, the requirements are met.
2. Tree: You have a wood_pickaxe and a stone_pickaxe in your inventory, so yes, the requirements are met to interact with the tree.
3. Grass: No requirements to interact with the grass, so yes, the requirements are met.
4. Cow: You don't have a sword in your inventory, so the requirements to interact with the cow (defeat it) are not met.
5. Sapling (in inventory): There are no requirements to interact with the sapling in your inventory, so the requirements are met.

Question: List top 3 sub-tasks the player should follow. Indicate their priority out of 5.
Answer:
1. Craft a wood or stone sword to defeat the cow for food. Priority: 5/5, as the player's food level is low.
2. Find a source of water or a lake to collect drink and restore the player's thirst. Priority: 4/5, as the player's drink level is also low.
3. Place the sapling to grow more trees to collect wood later. Priority: 3/5, as this would provide a steady supply of wood but is not as urgent as food and drink.

Question: What are the requirements for the top sub-task? What should the player do first?
Answer:
Requirements for the top sub-task (Craft a wood or stone sword to defeat the cow for food):
- Wood or stone in inventory (the player has both)
- A nearby table, which has not been placed yet

The player should first place the table. To do this, the player should perform the action "Place Table" since there is wood in the inventory.

Question: List top 5 actions the player should take and the requirement for each action. Choose ONLY from the list of all actions. Indicate their priority out of 5.
Answer:
1. Priority 1: Place Table (Requirement: Wood in inventory)
2. Priority 2: Make Wood Sword (Requirement: Nearby table, wood in inventory) or Make Stone Sword (Requirement: Nearby table, wood and stone in inventory)
3. Priority 3: Move South-East until the player is near the cow. (Requirement: Flat ground in the south-east direction)
4. Priority 4: Do (Use the crafted sword to defeat the cow for food) (Requirement: Facing the cow, have the sword equipped)
5. Priority 5: Collect Drink by drinking from a lake (Requirement: Near a water source)

Question: For each action in the list, are the requirements met?
Answer:
1. Place Table: Yes, the requirement is met. The player has 1 wood in their inventory.
2. Make Wood Sword or Make Stone Sword: Yes, the requirements are met for both. The player has 1 wood and 4 stones in their inventory. However, after placing the table in action 1, the player will need to choose between making a Wood Sword or a Stone Sword.
3. Move South-East until the player is near the cow: Yes, the requirement is met. The player can move on flat ground in the south-east direction towards the cow.
4. Do (Use the crafted sword to defeat the cow for food): The requirement could be met if the player had a crafted sword. Currently, the player will need to craft a sword in action 2 before taking this action.
5. Collect Drink by drinking from a lake: The requirement is partially unclear. The player needs to be near a water source, which is not mentioned in the player's observation. However, if the player finds a water source, they can meet this requirement.

Question: Did the last player action succeed? Why?
Answer:
The last player action "do" did succeed, as it resulted in the player obtaining a sapling, as indicated by the inventory update in Player Observation Step 280. The player likely interacted with the grass in front of them to collect the sapling.

Question: Choose the best executable action from above.
Answer:
The best executable action from the list above is Priority 1: Place Table, as the requirement (wood in inventory) is met, and it is the first action to perform in the prescribed sequence.



============Step: 141, Cumulative Reward: 0.0============
You took action place_table.

You see:
- path 6 steps to your south-east
- tree 2 steps to your north-west
- cow 3 steps to your south-east

You face grass at your front (east).

Your status:
- health: 9/9
- food: 4/9
- drink: 3/9
- energy: 8/9

Your inventory:
- sapling: 1
- wood: 1
- stone: 4
- coal: 2
- iron: 1
- wood_pickaxe: 1
- stone_pickaxe: 1
------------ QA ------------
Question: Action
Answer:
Make Stone Sword

Question: List objects in the current observation. For each object, breifly answer what resource it provides and its requirement.
Answer:
1. Path: The path provides a clear and safe passage to move around. Requirement: Clear ground with no obstacles.
2. Tree: The tree provides wood, which is needed for crafting various tools, weapons, and structures. Requirement: Cutting down using a wood_pickaxe or better tool.
3. Cow: The cow provides food (meat) to restore the player's health. Requirement: Defeating using any type of sword or a ranged weapon.
4. Grass: Grass indicates an unoccupied space, which can be useful for building or planting saplings. Requirement: Clearing the grass using a tool or placing an object on it.
5. Table (as placed in step 282): The table allows the crafting of various tools and weapons. Requirement: Collect wood to craft and place the table.

Individual resources provided:
- Path: None
- Tree: Wood
- Cow: Food (meat)
- Grass: Space for building
- Table: Crafting platform

Question: What was the last action taken by the player?
Answer:
The last action taken by the player was placing a table (place_table).

Question: For each object in the list, are the requirements met for interaction?
Answer:
1. Path: Yes, the requirements for interaction are met as it provides a clear passage for movement.
2. Tree: Yes, the requirements for interaction are met as you have a wood_pickaxe in your inventory, which can be used to cut down the tree for wood.
3. Cow: Yes, the requirements for interaction are met because the cow can be defeated using any weapon, although you do not currently have a sword in your inventory, you can still craft one using wood or stone at the table.
4. Grass: Yes, the requirements for interaction are met as it indicates a space that can be used for building or planting saplings.
5. Table: Yes, the requirements for interaction are met since you have the necessary resources in your inventory (wood, stone, coal, iron, wood_pickaxe, and stone_pickaxe) to craft items at the table.

Question: List top 3 sub-tasks the player should follow. Indicate their priority out of 5.
Answer:
1. Craft a weapon to defeat the cow and replenish food (Priority: 5/5): Having a weapon, such as a wood or stone sword, increases your ability to protect yourself and gather food resources. Crafting a weapon should be a high priority.

2. Increase drink levels to avoid dehydration (Priority: 4/5): Your current drink level is 3/9, and seeking a water source or collecting a drink item will help avoid negative effects from dehydration. This is an important task, but not as urgent as having a weapon for defense and hunting.

3. Plant the sapling to grow a tree and ensure a steady supply of wood (Priority: 3/5): Planting the sapling will eventually provide more wood resources for crafting and building. While it is beneficial in the long run, it is not an immediate necessity like having a weapon and increasing your drink levels.

Question: What are the requirements for the top sub-task? What should the player do first?
Answer:
Top Sub-Task: Craft a weapon to defeat the cow and replenish food.

Requirements:
1. Be near a placed table (already placed in Step 282)
2. Have wood or stone in your inventory (you have 1 wood and 4 stones)

As the player already has a placed table and the necessary resources in their inventory, they can proceed to craft a weapon. Since they have both wood and stone available, it's recommended to craft a stone sword for better efficiency and durability.

First action: Craft a Stone Sword (Action: Make Stone Sword) by using the nearby table and stone from the inventory.

Question: List top 5 actions the player should take and the requirement for each action. Choose ONLY from the list of all actions. Indicate their priority out of 5.
Answer:
1. Priority 1: Make Stone Sword (Requires nearby table and stone in inventory)
Action: Craft a Stone Sword using the nearby table and the stone from your inventory for better efficiency and durability in defeating the cow.

2. Priority 2: Move South (Requires flat ground below the agent)
Action: Move closer to the cow for 2 steps until it is right next to the cow.

3. Priority 3: Do (Requires facing creature or material and having the necessary tool)
Action: Face the cow and use the Stone Sword to defeat it in order to replenish your food supply.

4. Priority 4: Collect Drink (No requirements)
Action: Find water (such as a lake or river) and drink it to restore your thirst levels.

5. Priority 5: Place Plant (Requires sapling in inventory)
Action: Plant the sapling collected in your inventory to secure a steady food supply of growing fruits in the future.

Question: For each action in the list, are the requirements met?
Answer:
1. Priority 1: Make Stone Sword (Requires nearby table and stone in inventory)
Requirements met: Yes. You have placed a table in your previous step, and you also have 4 stones in your inventory.

2. Priority 2: Move South (Requires flat ground below the agent)
Requirements met: Yes. There are no indications of obstacles below the agent, so flat ground can be assumed and the player can move south to approach the cow.

3. Priority 3: Do (Requires facing creature or material and having the necessary tool)
Requirements met: Not yet. You will have to move closer to the cow and craft the Stone Sword to be able to do this action.

4. Priority 4: Collect Drink (No requirements)
Requirements met: Yes. There are no specific requirements for this action. However, you will need to find a source of water like a lake or river before you can perform this action.

5. Priority 5: Place Plant (Requires sapling in inventory)
Requirements met: Yes. You have a sapling in your inventory, which you can plant on the ground.

Question: Did the last player action succeed? Why?
Answer:
Yes, the last player action succeeded. The player placed a table (place_table) and it was successful because the player had the required wood in their inventory.

Question: Choose the best executable action from above.
Answer:
The best executable action from the list above is:

Priority 1: Make Stone Sword (Requires nearby table and stone in inventory)



============Step: 142, Cumulative Reward: 0.0============
You took action make_stone_sword.

You see:
- path 6 steps to your south-east
- tree 2 steps to your north-west
- cow 3 steps to your south-east

You face grass at your front (east).

Your status:
- health: 9/9
- food: 4/9
- drink: 3/9
- energy: 8/9

Your inventory:
- sapling: 1
- wood: 1
- stone: 4
- coal: 2
- iron: 1
- wood_pickaxe: 1
- stone_pickaxe: 1
------------ QA ------------
Question: Action
Answer:
Place Furnace

Question: List objects in the current observation. For each object, breifly answer what resource it provides and its requirement.
Answer:
1. Path (6 steps to the south-east): Provides easier movement and navigation; no requirements.
2. Tree (2 steps to the north-west): Provides wood when chopped down; requires a wood_pickaxe or better tool to chop.
3. Cow (3 steps to the south-east): Provides food when killed; requires a weapon (e.g., wood_sword, stone_sword) to kill efficiently.
4. Grass (facing east): Can be cleared for building or collecting saplings; no requirements.

In the inventory:
5. Sapling: Can be planted to grow more trees; requires an open space on the ground.
6. Wood: Can be used for crafting and building; requires collecting from trees.
7. Stone: Can be used for crafting and building; requires collecting with a wood_pickaxe or better tool.
8. Coal: Can be used in a furnace to smelt materials; requires collecting with a wood_pickaxe or better tool.
9. Iron: Can be used to craft better tools and weapons; requires collecting with a stone_pickaxe and smelting in a furnace with coal.
10. Wood Pickaxe: Can be used to collect stone, coal and break blocks; requires crafting with wood.
11. Stone Pickaxe: Can be used to collect iron and break blocks; requires crafting with stone and wood.

Question: What was the last action taken by the player?
Answer:
The last action taken by the player was "make_stone_sword".

Question: For each object in the list, are the requirements met for interaction?
Answer:
1. Path (6 steps to the south-east): Yes, the path requires no specific tool or interaction. You can move along the path freely.
2. Tree (2 steps to the north-west): Yes, you have a wood_pickaxe and a stone_pickaxe available in your inventory for chopping down the tree.
3. Cow (3 steps to the south-east): Yes, after making the stone_sword in your previous action, you have a suitable weapon to kill the cow efficiently.
4. Grass (facing east): Yes, there are no specific requirements to clear the grass.

In the inventory:
5. Sapling: Yes, you can plant the sapling in an open space on the ground.
6. Wood: Already collected, no further interaction needed.
7. Stone: Already collected, no further interaction needed.
8. Coal: Already collected, no further interaction needed.
9. Iron: Yes, you have coal and iron to smelt in a furnace, but you will need to place the furnace first.
10. Wood Pickaxe: Already crafted, no further interaction needed.
11. Stone Pickaxe: Already crafted, no further interaction needed.

Question: List top 3 sub-tasks the player should follow. Indicate their priority out of 5.
Answer:
1. Place the furnace and smelt the iron (Priority: 5) - Placing the furnace and smelting the iron will enable you to craft better tools and weapons, essential for advancing in the game.

2. Drink water and eat food (Priority: 4) - Your food and drink levels are at 4/9 and 3/9 respectively. It is important to replenish them to maintain your health and energy.

3. Plant the sapling (Priority: 3) - Planting the sapling will establish a renewable source of wood, which will be useful for crafting, building, and creating more tools and weapons in the future.

Question: What are the requirements for the top sub-task? What should the player do first?
Answer:
For the top sub-task, the requirements are:

1. Collect stone: You already have 4 stones in your inventory, which is enough to craft a furnace.
2. Place the furnace: Since you have the necessary stones, you can place the furnace.
3. Collect coal: You have 2 coal in your inventory, which can be used as a fuel source for smelting the iron.
4. Collect iron: You already have 1 iron in your inventory.

Given the requirements, the player should first place the furnace as they already have the necessary resources in their inventory. After placing the furnace, they can then proceed to smelt the iron using the collected coal as fuel.

Question: List top 5 actions the player should take and the requirement for each action. Choose ONLY from the list of all actions. Indicate their priority out of 5.
Answer:
1. Priority 1: Place Furnace
   Requirement: Stone in inventory (you already have 4 stones)

2. Priority 2: Move West
   Requirement: Flat ground left to the agent (assuming flat ground is available; if not, choose alternative suitable direction to find water)

3. Priority 3: Move West (or another suitable direction)
   Requirement: Flat ground left to the agent (keep moving in search of water, adjust the direction as needed)

4. Priority 4: Collect Drink
   Requirement: Reaching a water source like a lake (after moving towards the water)

5. Priority 5: Eat Cow
   Requirement: No requirements (a cow is within 3 steps to your south-east)

Note: These priorities are based on your current status, which indicates that your drink is at 3/9 and food is at 4/9. Therefore, it's important to find a water source and consume food to avoid health issues. Placing the furnace is the first priority to progress in crafting advanced tools.

Question: For each action in the list, are the requirements met?
Answer:
1. Priority 1: Place Furnace
   Requirement: Stone in inventory (you already have 4 stones)
   Status: Requirements met

2. Priority 2: Move West
   Requirement: Flat ground left to the agent (assuming flat ground is available; if not, choose alternative suitable direction to find water)
   Status: Unable to determine if requirements are met based on the given information about the surroundings; make sure flat ground exists in the desired direction.

3. Priority 3: Move West (or another suitable direction)
   Requirement: Flat ground left to the agent (keep moving in search of water, adjust the direction as needed)
   Status: Same as Priority 2, unable to determine without additional information about the surroundings.

4. Priority 4: Collect Drink
   Requirement: Reaching a water source like a lake (after moving towards the water)
   Status: Requirements not currently met; you need to find a water source (lake or river) first.

5. Priority 5: Eat Cow
   Requirement: No requirements (a cow is within 3 steps to your south-east)
   Status: Requirements met

Question: Did the last player action succeed? Why?
Answer:
Yes, the last player action "make_stone_sword" succeeded, because the player had the necessary requirements for the action. The player had wood and stone in their inventory and was close to a table, which are the requirements for making a stone sword.

Question: Choose the best executable action from above.
Answer:
The best executable action from the list above is Priority 1: Place Furnace. The requirements for this action are met as you have enough stone (4) in your inventory.



============Step: 143, Cumulative Reward: 1.0============
You took action place_furnace.

You see:
- grass 1 steps to your west
- path 6 steps to your south-east
- tree 2 steps to your north-west
- cow 3 steps to your south-east

You face furnace at your front (east).

Your status:
- health: 9/9
- food: 4/9
- drink: 3/9
- energy: 8/9

Your inventory:
- sapling: 1
- wood: 1
- coal: 2
- iron: 1
- wood_pickaxe: 1
- stone_pickaxe: 1
------------ QA ------------
Question: Action
Answer:
Move South

Question: List objects in the current observation. For each object, breifly answer what resource it provides and its requirement.
Answer:
1. Path: Provides a way to move quickly; no requirements.
2. Tree: Provides wood when chopped; requires a wood or stone pickaxe.
3. Cow: Provides food (meat) when killed; requires a wood or stone sword.
4. Grass: Currently no resources provided; no requirements.
5. Furnace: Provides a way to smelt iron and cook food; requires stone and being placed.

Question: What was the last action taken by the player?
Answer:
The last action taken by the player was placing the furnace (action: place_furnace).

Question: For each object in the list, are the requirements met for interaction?
Answer:
1. Path: Yes, the requirements are met for interaction; you can move quickly on the path.
2. Tree: Yes, the requirements are met for interaction; you have a wood_pickaxe and stone_pickaxe in your inventory.
3. Cow: Yes, the requirements are met for interaction; you can use your newly crafted stone sword to kill the cow for food.
4. Grass: Yes, the requirements are met for interaction; no specific tool or action is needed.
5. Furnace: Yes, the requirements are met for interaction; the furnace has been placed and can be used for smelting iron or cooking food.

Question: List top 3 sub-tasks the player should follow. Indicate their priority out of 5.
Answer:
1. Increase Food and Drink levels: Priority 4.5/5
   - Kill the cow for food (meat)
   - Find a water source to drink from

2. Smelt Iron: Priority 4/5
   - Use the furnace to smelt iron with the available coal in your inventory

3. Explore and Collect more resources: Priority 3.5/5
   - Chop the tree for more wood
   - Keep searching for additional resources such as stone, coal, and iron

Question: What are the requirements for the top sub-task? What should the player do first?
Answer:
Top sub-task: Increase Food and Drink levels

Requirements for the top sub-task:
1. Kill the cow for food (meat):
   - Approach the cow 3 steps to your south-east.
   - Use your stone sword (or any available weapon) to kill the cow.
   - Collect the dropped meat.

2. Find a water source to drink from:
   - Explore the area to locate a lake, river, or any other water source.
   - Approach the water source and drink from it to restore your drink level.

The player should first approach the cow to kill it and collect meat to increase the food level.

Question: List top 5 actions the player should take and the requirement for each action. Choose ONLY from the list of all actions. Indicate their priority out of 5.
Answer:
1. Priority 1: Move South-East (3 steps) - Requirement: Flat ground in the south-east direction.
   - This action will bring the player closer to the cow for food.

2. Priority 2: Do (kill the cow with the stone sword) - Requirement: Facing the cow, having the stone sword in inventory.
   - This action will kill the cow and provide meat to increase the food level.

3. Priority 3: Move (explore) - Requirement: Flat ground in any direction, energy level is not depleted.
   - The player needs to explore the area to find a water source, which is essential to increase the drink level.

4. Priority 4: Move (approach the water source) - Requirement: Flat ground in the appropriate direction towards the water source.
   - This action will bring the player close to the water source for drinking.

5. Priority 5: Drink - Requirement: Standing next to a lake or river, or any other water source.
   - This action will increase the player's drink level and help in survival.

Question: For each action in the list, are the requirements met?
Answer:
Based on the current observations, let's check if the requirements for each action are met:

1. Priority 1: Move South-East (3 steps) - Requirement: Flat ground in the south-east direction.
   - Requirement met, as there is a path 6 steps to the south-east and there is no mention of obstacles.

2. Priority 2: Do (kill the cow with the stone sword) - Requirement: Facing the cow, having the stone sword in inventory.
   - Requirement partially met. The player has a stone sword in inventory, but they need to reach the cow's position and face it first.

3. Priority 3: Move (explore) - Requirement: Flat ground in any direction, energy level is not depleted.
   - Requirement met, as the player has 8/9 energy and there's flat ground in multiple directions.

4. Priority 4: Move (approach the water source) - Requirement: Flat ground in the appropriate direction towards the water source.
   - Requirement not met because the player has not discovered a water source yet.

5. Priority 5: Drink - Requirement: Standing next to a lake or river, or any other water source.
   - Requirement not met because the player has not found a water source yet.

Question: Did the last player action succeed? Why?
Answer:
Yes, the last player action (place_furnace) succeeded. The player had the required resources in their inventory (stone) and was facing a valid location to place the furnace.

Question: Choose the best executable action from above.
Answer:
The best executable action from the list above is Priority 1: Move South-East (3 steps), as the requirements are fully met. There is a flat ground in the south-east direction and no obstacles mentioned. This action will bring the player closer to the cow for food.



============Step: 144, Cumulative Reward: 0.0============
You took action move_south.

You see:
- path 5 steps to your south-east
- tree 2 steps to your west
- furnace 2 steps to your north-east
- cow 2 steps to your east

You face grass at your front (south).

Your status:
- health: 9/9
- food: 4/9
- drink: 3/9
- energy: 8/9

Your inventory:
- sapling: 1
- wood: 1
- coal: 2
- iron: 1
- wood_pickaxe: 1
- stone_pickaxe: 1
------------ QA ------------
Question: Action
Answer:
Move West

Question: List objects in the current observation. For each object, breifly answer what resource it provides and its requirement.
Answer:
1. Path: Provides a clear walkway to navigate through the environment. Requirement: None.
2. Tree: Provides wood resource when chopped down with a wood pickaxe or better. Requirement: Wood pickaxe or better.
3. Furnace: Allows smelting of iron using coal. Requirement: Stone, coal, and iron.
4. Cow: Provides food resource when killed. Requirement: Weapon or tool to kill the cow.
5. Grass: Doesn't provide any specific resources, but can be used to navigate through the environment or place saplings/items. Requirement: None.

Question: What was the last action taken by the player?
Answer:
The last action taken by the player was move_south.

Question: For each object in the list, are the requirements met for interaction?
Answer:
1. Path: Yes, the requirements are met as there are no specific requirements for interaction.
2. Tree: Yes, the requirements are met as you have a wood_pickaxe in your inventory.
3. Furnace: Yes, the requirements are met as you have stone, coal, and iron in your inventory.
4. Cow: Yes, the requirements are met as you have a wood_pickaxe that can be used as a tool to kill the cow, although it is not the most efficient weapon.
5. Grass: Yes, the requirements are met as there are no specific requirements for interaction.

Question: List top 3 sub-tasks the player should follow. Indicate their priority out of 5.
Answer:
1. Priority 5: Collect more iron using the stone_pickaxe, to make an iron sword for better combat efficiency.
2. Priority 4: Kill the cow with the wood_pickaxe to increase the food supply.
3. Priority 3: Find a water source and drink to increase the drink level.

Question: What are the requirements for the top sub-task? What should the player do first?
Answer:
Requirements for the top sub-task (collect more iron using the stone_pickaxe):
1. Locate a source of iron ore.
2. Equip the stone_pickaxe.

The player should first locate a source of iron ore to mine. To do this, they can explore the surroundings (move to different directions) and search for iron ore veins in caves, cliffs, or underground areas. Once they locate the iron ore, they should equip the stone_pickaxe and use it to mine the iron ore.

Question: List top 5 actions the player should take and the requirement for each action. Choose ONLY from the list of all actions. Indicate their priority out of 5.
Answer:
1. Move West (Priority: 1) - Requirement: Flat ground left to the agent. The player should move to a location where they are more likely to find iron ore.
2. Move North (Priority: 2) - Requirement: Flat ground above of the agent. Exploring surroundings to locate iron ore.
3. Move East (Priority: 3) - Requirement: Flat ground right to the agent. Further exploration to locate iron ore.
4. Move South (Priority: 4) - Requirement: Flat ground below the agent. Expanding exploration to increase the likelihood of finding iron ore.
5. Do (mine iron ore) (Priority: 5) - Requirement: Facing the iron ore vein; have the necessary tool (stone_pickaxe). Mining iron ore once it is located during exploration.

Question: For each action in the list, are the requirements met?
Answer:
1. Move West - Yes, the requirement is met since there is grass 1 step to the west.
2. Move North - No, the requirement is not met as there is a tree 2 steps to the north-west which might be blocking north movement.
3. Move East - Yes, the requirement is met since the cow is 2 steps to the east, indicating there is an open space for movement.
4. Move South - Yes, the requirement is met since the agent currently faces grass to the south.
5. Do (mine iron ore) - No, the requirement is not met as the player has not located any iron ore vein in their current surroundings.

Question: Did the last player action succeed? Why?
Answer:
Yes, the last player action succeeded. The action was to move south, and the observation after the action shows a change in the surrounding environment, indicating a successful move. The player now faces grass at their front (south), and the furnace is now 2 steps to the player's north-east.

Question: Choose the best executable action from above.
Answer:
The best executable action from above is Move West.



============Step: 145, Cumulative Reward: 0.0============
You took action move_west.

You see:
- grass 1 steps to your north
- furnace 3 steps to your north-east
- cow 3 steps to your east

You face tree at your front (west).

Your status:
- health: 9/9
- food: 4/9
- drink: 3/9
- energy: 8/9

Your inventory:
- sapling: 1
- wood: 1
- coal: 2
- iron: 1
- wood_pickaxe: 1
- stone_pickaxe: 1
------------ QA ------------
Question: Action
Answer:
Move East

Question: List objects in the current observation. For each object, breifly answer what resource it provides and its requirement.
Answer:
1. Path: Provides a clear path to travel or move around. Requirement: None.
2. Tree: Provides wood when chopped down. Requirement: Wood pickaxe or better.
3. Furnace: Provides the ability to smelt items such as iron ore. Requirement: Coal and items to smelt, like iron ore.
4. Cow: Provides food (meat) when killed. Requirement: Weapon (wood sword or better).
5. Grass: Does not provide any resources. Requirement: None.

Question: What was the last action taken by the player?
Answer:
The last action taken by the player was move_west.

Question: For each object in the list, are the requirements met for interaction?
Answer:
1. Path: Yes, the requirements for interacting with a path are met, as there are no specific tools needed to move along the path.

2. Tree: Yes, the requirements for interacting with a tree are met, as the player has a wood_pickaxe in their inventory, which is needed to chop down a tree to collect wood.

3. Furnace: Partially, the player has a furnace and coal in their inventory, which are required to smelt items. However, the player only has 1 iron which may not be enough to smelt a new item.

4. Cow: Yes, the requirements for interacting with a cow are met, as the player can use their wood_pickaxe or stone_pickaxe as a makeshift weapon to kill the cow and collect meat.

5. Grass: Yes, there are no requirements for interacting with grass, as it does not provide any resources. The player can simply move onto or over the grass.

Question: List top 3 sub-tasks the player should follow. Indicate their priority out of 5.
Answer:
1. Priority 5: Collect food - Kill the cow and collect meat to increase the food level. This is a top priority as the player's food level is currently 4/9. 

2. Priority 4: Collect wood - Chop down the tree using the wood_pickaxe to gather wood. Additional wood will allow the player to craft more tools or build structures.

3. Priority 3: Collect drink - Find a water source (e.g., lake, river) to drink for restoring the drink level. Currently, the drink level is 3/9, which will need to be replenished soon.

Question: What are the requirements for the top sub-task? What should the player do first?
Answer:
Top sub-task: Collect food - Kill the cow and collect meat to increase the food level.

Requirements: The player should have a weapon, preferably a sword (wood, stone, or iron) to kill the cow more efficiently. However, a weapon is not strictly necessary, as the player can still attempt to kill the cow with tools like the pickaxe or even without any equipment by using their hands.

What should the player do first: Since the cow is currently 3 steps to the east, the player should move eastward to approach the cow. Once close enough, the player can try to attack and kill the cow using a sword, a tool, or their hands to collect the meat.

Question: List top 5 actions the player should take and the requirement for each action. Choose ONLY from the list of all actions. Indicate their priority out of 5.
Answer:
1. Priority: 1 - Action: Move East
   Requirement: Flat ground to the east of the agent.
   
2. Priority: 2 - Action: Move East
   Requirement: Flat ground to the east of the agent.
   
3. Priority: 3 - Action: Move East
   Requirement: Flat ground to the east of the agent.
   
4. Priority: 4 - Action: Do (Attack the cow)
   Requirement: Facing the cow; having a weapon, like a sword or a tool, like a pickaxe (although not strictly necessary).
   
5. Priority: 5 - Action: Sleep
   Requirement: Energy level is below maximum.

Question: For each action in the list, are the requirements met?
Answer:
1. Priority: 1 - Action: Move East
   Requirement: Flat ground to the east of the agent.
   Status: Met (Player's observation indicates a flat ground to the east)

2. Priority: 2 - Action: Move East
   Requirement: Flat ground to the east of the agent.
   Status: Unknown (Not enough information from the current observation, player must move east first)

3. Priority: 3 - Action: Move East
   Requirement: Flat ground to the east of the agent.
   Status: Unknown (Not enough information from the current observation, player must move east twice)

4. Priority: 4 - Action: Do (Attack the cow)
   Requirement: Facing the cow; having a weapon, like a sword or a tool, like a pickaxe (although not strictly necessary).
   Status: Partially met (Player has a wood_pickaxe and stone_pickaxe in inventory, but not currently facing the cow, need to move east first)

5. Priority: 5 - Action: Sleep
   Requirement: Energy level is below maximum.
   Status: Unmet (Player's energy level is at 8/9, sufficient to continue without sleep for now)

Question: Did the last player action succeed? Why?
Answer:
Yes, the last player action succeeded. The player moved west because there were no obstacles or restrictions preventing them from doing so.

Question: Choose the best executable action from above.
Answer:
The best executable action from the above list is:

Priority: 1 - Action: Move East
Requirement: Flat ground to the east of the agent.
Status: Met (Player's observation indicates a flat ground to the east)



============Step: 146, Cumulative Reward: 0.0============
You took action move_east.

You see:
- path 5 steps to your south-east
- tree 2 steps to your west
- furnace 2 steps to your north-east
- cow 2 steps to your east

You face grass at your front (east).

Your status:
- health: 9/9
- food: 4/9
- drink: 3/9
- energy: 8/9

Your inventory:
- sapling: 1
- wood: 1
- coal: 2
- iron: 1
- wood_pickaxe: 1
- stone_pickaxe: 1
------------ QA ------------
Question: Action
Answer:
Do

Question: List objects in the current observation. For each object, breifly answer what resource it provides and its requirement.
Answer:
1. Grass - Resource: None, Requirement: No requirements to interact.
2. Furnace - Resource: Used for smelting iron and other materials, Requirement: Requires stone to craft and place.
3. Cow - Resource: Provides food (meat) when killed, Requirement: No requirements to interact; can use a tool or weapon to kill for food.
4. Tree - Resource: Provides wood when chopped down, Requirement: Requires a wood pickaxe, stone pickaxe, or higher level tool to chop down.
5. Path - Resource: None, Requirement: No requirements to interact; can move along the path for easier navigation.
6. Wood Pickaxe - Resource: Tool for chopping down trees and collecting stone, Requirement: Requires wood to craft.
7. Stone Pickaxe - Resource: Tool for collecting iron, Requirement: Requires stone and wood to craft.
8. Sapling - Resource: Can be used to grow trees when planted, Requirement: No requirements to interact; can be collected when found or obtained by chopping down trees.

Question: What was the last action taken by the player?
Answer:
The last action taken by the player was move_east.

Question: For each object in the list, are the requirements met for interaction?
Answer:
1. Grass - Yes, the requirements are met for interaction as there are no specific requirements to interact with grass.
2. Furnace - Yes, the requirements are met for interaction as you have collected stone and placed a furnace on the map.
3. Cow - Yes, the requirements are met for interaction as there are no specific requirements to interact with a cow. You can use a weapon or tool to kill it for food.
4. Tree - Yes, the requirements are met for interaction as you have a wood_pickaxe and a stone_pickaxe in your inventory that can be used to chop down the tree and collect wood.
5. Path - Yes, the requirements are met for interaction as there are no specific requirements to interact or move along the path.
6. Wood Pickaxe - Yes, the requirements are met, as you have wood in your inventory and have already crafted a wood_pickaxe.
7. Stone Pickaxe - Yes, the requirements are met, as you have wood and stone in your inventory and have already crafted a stone_pickaxe.
8. Sapling - Yes, the requirements are met for interaction as there are no specific requirements to interact with a sapling. You can plant it to grow a tree.

Question: List top 3 sub-tasks the player should follow. Indicate their priority out of 5.
Answer:
1. Priority 4/5: Drink water to restore thirst - With a thirst level of 3/9, the player should prioritize finding a water source and drinking from it to stay hydrated.

2. Priority 4/5: Collect food and eat - The player's food level is 4/9, which indicates that they should focus on finding a stable food source. The cow nearby can be killed to provide meat for immediate consumption, and the sapling can be planted to grow more trees, which could eventually provide fruits.

3. Priority 3/5: Collect more resources - As the player has only 1 wood, 2 coal, and 1 iron in their inventory, they should continue collecting resources such as iron, stone, and wood for crafting items and expanding the shelter. Chopping down trees and mining iron are potential sub-tasks to achieve this goal.

Question: What are the requirements for the top sub-task? What should the player do first?
Answer:
Top sub-task: Drink water to restore thirst.

Requirements:
1. Find a water source, such as a lake, river, or well.
2. Approach the water source (move towards it using move_north, move_south, move_east, or move_west based on its relative location).
3. Drink water when standing next to the water source.

What should the player do first:
The player should first scout for a water source nearby. If no water source is immediately visible, they should explore the surrounding areas by moving in different directions until they locate one. Once a water source is found, the player should approach and drink from it to restore their thirst level.

Question: List top 5 actions the player should take and the requirement for each action. Choose ONLY from the list of all actions. Indicate their priority out of 5.
Answer:
1. Priority 1: Move (move_north, move_south, move_east, or move_west) - Requirement: Flat ground in the respective direction. This action is a priority as it would help the player explore the environment and locate resources, such as water sources or trees.

2. Priority 2: Drink - Requirement: Standing next to a water source. Drinking is a priority as it will restore the player's thirst level, which is currently low (3/9).

3. Priority 3: Do (chop down a tree) - Requirement: Facing a tree; have the necessary tool (wood_pickaxe, which the player has in their inventory). This action will help the player gather wood, an essential resource for crafting and building shelter.

4. Priority 4: Sleep - Requirement: Energy level is below maximum. Sleeping will restore the player's energy, although the energy level is still high (8/9) so it could be done later.

5. Priority 5: Make Wood Sword - Requirement: Nearby table; wood in inventory. Crafting a wood sword is important for self-defense against potential threats like zombies and skeletons, but it's a slightly lower priority than finding water and collecting wood as the player's health level is currently full (9/9).

Question: For each action in the list, are the requirements met?
Answer:
1. Move (move_north, move_south, move_east, or move_west): Yes, the requirements are met. The player can choose a direction with flat ground (such as move_north towards the grass) to explore the environment.

2. Drink: No, the requirements are not met. The player needs to be standing next to a water source to drink, which is not the case in the current observation.

3. Do (chop down a tree): Yes, the requirements are met. The player is facing a tree in the west direction and has a wood_pickaxe in their inventory.

4. Sleep: No, the requirement is not met. Although the player's energy level is below maximum, it is still quite high (8/9), so sleeping would not provide much benefit at this time.

5. Make Wood Sword: Yes, the requirements are met. The player has wood in their inventory, but they will need to place a table first (using wood from the inventory) before making the wood sword.

Question: Did the last player action succeed? Why?
Answer:
Yes, the last player action (move_east) succeeded. It succeeded because there was flat ground (grass) to the east of the player, allowing them to move in that direction.

Question: Choose the best executable action from above.
Answer:
The best executable action from the list above is chopping down a tree. To do this, the player should take the "Do" action, which has met requirements: The player is facing a tree and has a wood_pickaxe in their inventory. This action will help gather more wood, an essential resource for crafting and building shelter.



============Step: 147, Cumulative Reward: 0.0============
You took action do.

You see:
- grass 1 steps to your west
- path 5 steps to your south-east
- tree 2 steps to your west
- furnace 2 steps to your north-east

You face cow at your front (east).

Your status:
- health: 9/9
- food: 4/9
- drink: 3/9
- energy: 8/9

Your inventory:
- sapling: 1
- wood: 1
- coal: 2
- iron: 1
- wood_pickaxe: 1
- stone_pickaxe: 1
------------ QA ------------
Question: Action
Answer:
Move South

Question: List objects in the current observation. For each object, breifly answer what resource it provides and its requirement.
Answer:
1. Grass: Provides no specific resources; requires no requirements to interact with.
2. Path: Provides easy and flat terrain for movement; requires no requirements to be accessed.
3. Tree: Provides wood resource for crafting; requires wood_pickaxe or better to collect wood.
4. Furnace: Provides a tool to smelt resources like iron; requires stone and crafting to place.
5. Cow: Provides food when hunted; requires any type of weapon (e.g., wood_sword, stone_sword) to effectively hunt it.

In the current observation, the player faces a cow, which can be hunted for food.

Question: What was the last action taken by the player?
Answer:
The last action taken by the player was "do."

Question: For each object in the list, are the requirements met for interaction?
Answer:
1. Grass: No specific requirements for interaction, so it's met.
2. Path: No specific requirements for accessing, so it's met.
3. Tree: Requires a wood_pickaxe or better to collect wood; the player has a wood_pickaxe and a stone_pickaxe in their inventory, so the requirement is met.
4. Furnace: Already placed on the map, so no requirements needed for interaction; it's met. To smelt resources with the furnace, the player needs coal, which they already have, so that requirement is also met.
5. Cow: Requires any type of weapon (e.g., wood_sword, stone_sword) to effectively hunt; the player doesn't have a sword in their inventory, so the requirement is not met.

Question: List top 3 sub-tasks the player should follow. Indicate their priority out of 5.
Answer:
1. Craft a sword for hunting and self-defense (Priority: 5/5)
- Use the existing wood or stone to craft a wood_sword or a stone_sword at a nearby table to effectively hunt the cow and protect yourself from monsters.

2. Hunt the cow and increase food supply (Priority: 4/5)
- Once a sword is crafted, hunt the cow to replenish your food and maintain your health.

3. Find a water source to drink and replenish thirst (Priority: 3/5)
- Your drink level is currently low, so finding a water source to drink from should be a priority to avoid dehydration.

Question: What are the requirements for the top sub-task? What should the player do first?
Answer:
The top sub-task is to craft a sword for hunting and self-defense. The requirements for it are:

1. Locate a nearby crafting table or place a table using your wood inventory.
2. If you have enough wood, create a wood_sword at the table.
3. If you have enough stone, create a stone_sword at the table, which is preferred due to its better durability and damage.

The player should first locate a nearby crafting table or place one using the available wood in their inventory. Then, depending on the available resources (wood or stone) in the inventory, the player should craft either a wood_sword or a stone_sword at the table.

Question: List top 5 actions the player should take and the requirement for each action. Choose ONLY from the list of all actions. Indicate their priority out of 5.
Answer:
1. Priority 1: Move South (5 steps): Flat ground below the agent. This action will help the player reach the path for better navigation.
2. Priority 2: Drink: Find a water source to drink from as the player's current drink level is low (3/9).
3. Priority 3: Do: Facing cow at front (east); no tool required for hunting. This action will provide food to the player.
4. Priority 4: Place Furnace: Stone in inventory. The player will need a furnace to smelt iron later.
5. Priority 5: Make Iron Sword: Nearby table, furnace; wood, coal, iron in inventory. Crafting an iron sword will improve the player's defense and hunting capabilities.

Question: For each action in the list, are the requirements met?
Answer:
1. Priority 1: Move South (5 steps): Yes, the requirement of flat ground below the agent is met.
2. Priority 2: Drink: The requirement of finding a water source is not met in the current observations.
3. Priority 3: Do: Yes, the requirement of facing cow at front (east) is met, and no tool is required for hunting.
4. Priority 4: Place Furnace: Yes, the requirement of having stone in the inventory is met.
5. Priority 5: Make Iron Sword: No, the requirement of having a nearby table is not met in the current observations. Also, the player has only 1 iron in their inventory which may not be enough to craft an iron sword.

Question: Did the last player action succeed? Why?
Answer:
The last player action appears to have succeeded, as the player is now facing a cow at the front (east). This indicates that the "do" action made the player interact with the cow, as intended.

Question: Choose the best executable action from above.
Answer:
The best executable action from above is Priority 1: Move South (5 steps) since the requirement of flat ground below the agent is met.



============Step: 148, Cumulative Reward: 0.0============
You took action move_south.

You see:
- path 4 steps to your east
- tree 1 steps to your west
- furnace 3 steps to your north-east
- cow 2 steps to your north-east

You face grass at your front (south).

Your status:
- health: 9/9
- food: 4/9
- drink: 3/9
- energy: 8/9

Your inventory:
- sapling: 1
- wood: 1
- coal: 2
- iron: 1
- wood_pickaxe: 1
- stone_pickaxe: 1
------------ QA ------------
Question: Action
Answer:
Move North

Question: List objects in the current observation. For each object, breifly answer what resource it provides and its requirement.
Answer:
1. Grass: Resource - It doesn't provide any resources. Requirement - A clear area to build or interact with.
2. Path: Resource - It helps with navigating through the game world. Requirement - None.
3. Tree: Resource - Provides wood when cut down, can be used for crafting. Requirement - Wood pickaxe.
4. Furnace: Resource - Smelts metals and cooks food, such as iron and coal. Requirement - Stone to craft and place.
5. Cow: Resource - Provides food when killed. Requirement - Any weapon to kill it or bare hands.
6. Sapling (inventory): Resource - Can be planted to grow trees for wood. Requirement - None, already in inventory.
7. Wood (inventory): Resource - Used for crafting and building. Requirement - None, already in inventory.
8. Coal (inventory): Resource - Fuel for smelting metals and cooking food in a furnace. Requirement - None, already in inventory.
9. Iron (inventory): Resource - Used for crafting metal tools and weapons. Requirement - None, already in inventory.
10. Wood Pickaxe (inventory): Resource - Tool to collect stone and wood. Requirement - None, already in inventory.
11. Stone Pickaxe (inventory): Resource - Tool to collect iron and mine faster. Requirement - None, already in inventory.

Question: What was the last action taken by the player?
Answer:
The last action taken by the player was "move_south."

Question: For each object in the list, are the requirements met for interaction?
Answer:
1. Grass: Yes, the requirements are met for interaction as there are no specific requirements to interact with grass.
2. Path: Yes, the requirements are met for interaction as there are no specific requirements to interact with or follow a path.
3. Tree: Yes, the requirements are met for interaction as you have the wood pickaxe in your inventory, which is needed to cut down the tree.
4. Furnace: Yes, the requirements are met for interaction as you have the necessary resources (coal and iron) in your inventory to use the furnace for smelting.
5. Cow: Yes, the requirements are met for interaction as you can use any weapon, such as a wood pickaxe, stone pickaxe, or even your bare hands to kill the cow and obtain food.
6. Sapling (inventory): Yes, the requirements are met for interaction as the sapling is already in your inventory and can be planted without any specific requirements.
7. Wood (inventory): Yes, the requirements are met for interaction as the wood is already in your inventory and can be used for crafting and building without any specific requirements.
8. Coal (inventory): Yes, the requirements are met for interaction as the coal is already in your inventory and can be used as fuel in the furnace without any specific requirements.
9. Iron (inventory): Yes, the requirements are met for interaction as the iron is already in your inventory and can be smelted in the furnace and used for crafting without any specific requirements.
10. Wood Pickaxe (inventory): Yes, the requirements are met for interaction as the wood pickaxe is already in your inventory and can be used to collect resources without any specific requirements.
11. Stone Pickaxe (inventory): Yes, the requirements are met for interaction as the stone pickaxe is already in your inventory and can be used to collect resources without any specific requirements.

Question: List top 3 sub-tasks the player should follow. Indicate their priority out of 5.
Answer:
1. Regain Food and Drink: Priority 5/5
   Sub-tasks:
   - Kill the cow to obtain meat.
   - Cook the meat using the furnace and coal as fuel.
   - Eat the cooked meat to regain food points.
   - Locate a water source nearby and drink to regain thirst points.

2. Plant Sapling and Grow Tree: Priority 4/5
   Sub-tasks:
   - Find a suitable location to plant the sapling.
   - Plant the sapling and watch it grow into a tree, which will provide more wood for crafting.

3. Craft Iron Tools: Priority 3/5
   Sub-tasks:
   - Smelt the iron in the furnace using coal as fuel.
   - Find a table to use for crafting.
   - Craft an iron pickaxe and iron sword to collect resources more efficiently and fight monsters more effectively.

Question: What are the requirements for the top sub-task? What should the player do first?
Answer:
Requirements for top sub-task (Regain Food and Drink):
- Find a cow to obtain meat.
- Have access to a furnace and coal for cooking meat.
- Locate a water source nearby to drink.

What should the player do first?

1. Approach the cow that is 2 steps to the north-east.
2. Kill the cow to obtain meat.
3. Go to the furnace 3 steps to the north-east.
4. Cook the meat using the furnace and coal as fuel.
5. Eat the cooked meat to regain food points.
6. Locate a water source nearby and drink to regain thirst points.

Question: List top 5 actions the player should take and the requirement for each action. Choose ONLY from the list of all actions. Indicate their priority out of 5.
Answer:
1. Priority 1: Move North-East: Requirement: Flat ground in the north-east direction.
2. Priority 2: Do (Kill Cow): Requirement: Facing cow and have a weapon or tool.
3. Priority 3: Move North-East: Requirement: Flat ground in the north-east direction.
4. Priority 4: Place Furnace: Requirement: Stone in inventory.
5. Priority 5: Make Iron Sword: Requirement: Nearby table, furnace; wood, coal, and iron in inventory.

Question: For each action in the list, are the requirements met?
Answer:
1. Priority 1: Move North-East: Requirement: Flat ground in the north-east direction. (Yes, the observation mentioned the furnace 3 steps to your north-east)
2. Priority 2: Do (Kill Cow): Requirement: Facing cow and have a weapon or tool. (Not yet, you face grass towards the south)
3. Priority 3: Move North-East: Requirement: Flat ground in the north-east direction. (Yes, the observation mentioned the cow 2 steps to your north-east)
4. Priority 4: Place Furnace: Requirement: Stone in inventory. (Not met, there is no stone in the inventory)
5. Priority 5: Make Iron Sword: Requirement: Nearby table, furnace; wood, coal, and iron in inventory. (Not met, you don't have a nearby table and haven't crafted the furnace yet)

Question: Did the last player action succeed? Why?
Answer:
Yes, the last player action succeeded. The player moved south because there was flat ground (grass) in front of them, and there were no obstructions blocking their path.

Question: Choose the best executable action from above.
Answer:
The best executable action from above is Priority 1: Move North-East, as its requirement of having flat ground in the north-east direction is met.



============Step: 149, Cumulative Reward: 0.0============
You took action move_north.

You see:
- path 5 steps to your south-east
- tree 2 steps to your west
- furnace 2 steps to your north-east
- cow 2 steps to your south-east

You face grass at your front (north).

Your status:
- health: 9/9
- food: 4/9
- drink: 3/9
- energy: 8/9

Your inventory:
- sapling: 1
- wood: 1
- coal: 2
- iron: 1
- wood_pickaxe: 1
- stone_pickaxe: 1
------------ QA ------------
Question: Action
Answer:
Move West

Question: List objects in the current observation. For each object, breifly answer what resource it provides and its requirement.
Answer:
1. Path: Provides faster movement across the terrain; no specific requirements.
2. Tree: Provides wood for crafting and building; requires a wood pickaxe or better to collect.
3. Furnace: Allows smelting of iron and other ores for better tools and weapons; requires stone to craft and place.
4. Cow: Provides food to restore hunger; requires defeating the cow to collect.
5. Grass: Decorative, no specific resources; no specific requirements.
6. Sapling (inventory): Can be planted to grow more trees; requires placing on suitable ground.
7. Wood (inventory): Used for crafting and building; requires collecting from trees.
8. Coal (inventory): Used as fuel for furnaces to smelt ores; requires mining with a wood pickaxe or better.
9. Iron (inventory): Used for crafting better tools and weapons; requires mining with a stone pickaxe or better and smelting with a furnace.
10. Wood pickaxe (inventory): Basic tool for mining stone and coal; requires crafting with wood.
11. Stone pickaxe (inventory): Better tool for mining iron and coal; requires crafting with wood and stone.

Question: What was the last action taken by the player?
Answer:
The last action taken by the player was move_north.

Question: For each object in the list, are the requirements met for interaction?
Answer:
1. Path: No interaction required; requirements met as it's part of the terrain.
2. Tree: Yes, requirements met; the wood_pickaxe in the inventory can be used to collect wood from the tree.
3. Furnace: Yes, requirements met; the furnace has already been placed and can be used for smelting.
4. Cow: Yes, requirements met; no specific tool required to defeat the cow, although a weapon would make it easier.
5. Grass: Yes, requirements met; no specific interaction needed with grass.
6. Sapling (inventory): Yes, requirements met; sapling can be placed on suitable ground.
7. Wood (inventory): Already collected, no further interaction required.
8. Coal (inventory): Already collected, no further interaction required.
9. Iron (inventory): Already collected, no further interaction required.
10. Wood pickaxe (inventory): Already crafted, can be used for mining stone and coal.
11. Stone pickaxe (inventory): Already crafted, can be used for mining iron and coal.

Question: List top 3 sub-tasks the player should follow. Indicate their priority out of 5.
Answer:
1. Priority 5/5: Find a source of water (lake, river) to drink and restore the thirst level.
2. Priority 4/5: Defeat the cow to collect food, and consume it to restore the hunger level.
3. Priority 3/5: Use the sapling from the inventory to plant a tree to ensure a future wood resource.

Question: What are the requirements for the top sub-task? What should the player do first?
Answer:
Requirements for the top sub-task (Find a source of water to drink):
1. Locate water source: Explore the surroundings to find a lake or river.

What should the player do first:
1. Move around the area (move_north, move_south, move_east, move_west) to explore and locate a water source.

Question: List top 5 actions the player should take and the requirement for each action. Choose ONLY from the list of all actions. Indicate their priority out of 5.
Answer:
1. Move West (Priority: 5): Requirement - Flat ground left to the agent. (Explore surroundings to find a water source)

2. Move East (Priority: 4): Requirement - Flat ground right to the agent. (Explore surroundings to find a water source)

3. Move North (Priority: 3): Requirement - Flat ground above of the agent. (Explore surroundings to find a water source)

4. Move South (Priority: 2): Requirement - Flat ground below the agent. (Explore surroundings to find a water source)

5. Do (Priority: 1): Facing creature or material; have necessary tool. (Interact with the found water source to drink and restore thirst)

Question: For each action in the list, are the requirements met?
Answer:
1. Move West: Yes, the requirement is met as there is flat ground left to the agent.

2. Move East: Yes, the requirement is met as there is flat ground right to the agent.

3. Move North: Yes, the requirement is met as there is flat ground above the agent.

4. Move South: Yes, the requirement is met as there is flat ground below the agent.

5. Do: No, the requirement is not met. The player currently does not face a creature or material to interact with, and the necessary tool (water source) to restore thirst is not found yet.

Question: Did the last player action succeed? Why?
Answer:
Yes, the last player action succeeded. The player was able to move north because there was a flat grassy ground in the north direction, allowing the player to move in that direction.

Question: Choose the best executable action from above.
Answer:
The best executable action from above is Move West. This action meets the requirement of having flat ground to the left of the agent, allowing the player to continue exploring the surroundings in search of a water source to restore thirst.



============Step: 150, Cumulative Reward: 0.0============
You took action move_west.

You see:
- grass 1 steps to your north
- furnace 3 steps to your north-east
- cow 2 steps to your east

You face tree at your front (west).

Your status:
- health: 9/9
- food: 4/9
- drink: 3/9
- energy: 8/9

Your inventory:
- sapling: 1
- wood: 1
- coal: 2
- iron: 1
- wood_pickaxe: 1
- stone_pickaxe: 1
------------ QA ------------
Question: Action
Answer:
Move East

Question: List objects in the current observation. For each object, breifly answer what resource it provides and its requirement.
Answer:
1. Tree: Provides wood when chopped down. Requirement: Wood pickaxe or a stronger tool.
2. Furnace: Smelts iron and other ores, cooks food. Requirement: Have or collect stone to craft, also requires coal or other fuel source to operate.
3. Cow: Provides food (meat) when killed. Requirement: Any weapon or tool to attack and defeat the cow.
4. Grass: Used to breed animals and create paths. Requirement: None or a tool such as a shovel for more efficient collection.
5. Path: No resource provided directly, helps in navigation. Requirement: None.

Question: What was the last action taken by the player?
Answer:
The last action taken by the player was move_west.

Question: For each object in the list, are the requirements met for interaction?
Answer:
1. Tree: Yes, the requirements are met for interaction as the player has a wood pickaxe and a stone pickaxe in their inventory.
2. Furnace: Yes, the requirements are met for interaction as the player has collected stone, coal, and iron in their inventory.
3. Cow: Yes, the requirements are met for interaction as the player can use their stone_pickaxe or wood_pickaxe as a weapon to attack and defeat the cow.
4. Grass: Yes, the requirements are met for interaction since no specific tool is required, although having a tool like a shovel would make collection more efficient.
5. Path: Yes, the requirements are met for interaction as there are no specific requirements to simply walk on or follow the path.

Question: List top 3 sub-tasks the player should follow. Indicate their priority out of 5.
Answer:
1. Priority 5/5: Get more food - The player's food level is currently 4/9, so they should prioritize killing the cow to obtain meat, which can be cooked in the furnace for additional food points.

2. Priority 4/5: Find a water source - The player's drink level is 3/9, which is low. Locating a water source to restore their thirst is essential for survival.

3. Priority 3/5: Chop down the tree - Collect more wood from the tree to expand the player's resource inventory, which can be used for crafting or building shelters.

Question: What are the requirements for the top sub-task? What should the player do first?
Answer:
For the top sub-task of getting more food, here are the requirements and the steps the player should follow:

Requirements:
1. Kill the cow.
2. Obtain raw meat.
3. Cook the raw meat in a furnace.

The player should first approach the cow, which is now 2 steps to their east. Then, they should use an appropriate weapon, such as a wood or stone sword, to kill the cow and obtain the raw meat. Once they have the raw meat, they can cook it in the furnace (now 3 steps to their north-east) using coal from their inventory as fuel. This will provide them with cooked meat, which will restore more food points when eaten.

Question: List top 5 actions the player should take and the requirement for each action. Choose ONLY from the list of all actions. Indicate their priority out of 5.
Answer:
1. Priority 1: Move East (2 steps) 
Requirement: Flat ground right to the agent
Reason: Approach the cow that is 2 steps to the east.

2. Priority 2: Make Wood Sword (or Stone Sword if available)
Requirement: Nearby table; wood (or wood and stone) in inventory
Reason: Equip a weapon to kill the cow.

3. Priority 3: Do (Kill the cow)
Requirement: Facing creature (cow); have necessary weapon (wood or stone sword)
Reason: Kill the cow to obtain raw meat for food.

4. Priority 4: Move North-East (3 steps)
Requirement: Flat ground in north-east direction
Reason: Approach the furnace (3 steps to the north-east) for cooking raw meat.

5. Priority 5: Do (Cook the raw meat)
Requirement: Facing furnace; raw meat and coal in inventory
Reason: Cook the raw meat using the furnace and coal to obtain cooked meat, to restore more food points when eaten.

Question: For each action in the list, are the requirements met?
Answer:
1. Priority 1: Move East (2 steps) 
Requirement: Flat ground right to the agent
Answer: Yes, the requirements are met. The agent can move east.

2. Priority 2: Make Wood Sword (or Stone Sword if available)
Requirement: Nearby table; wood (or wood and stone) in inventory
Answer: No, the requirements are not met. The agent does not have a table nearby.

3. Priority 3: Do (Kill the cow)
Requirement: Facing creature (cow); have necessary weapon (wood or stone sword)
Answer: No, the requirements are not met. The agent does not have a sword to kill the cow.

4. Priority 4: Move North-East (3 steps)
Requirement: Flat ground in north-east direction
Answer: Yes, the requirements are met. The agent can move north-east to approach the furnace.

5. Priority 5: Do (Cook the raw meat)
Requirement: Facing furnace; raw meat and coal in inventory
Answer: No, the requirements are not met. The agent does not have raw meat in their inventory (cow needs to be killed first).

Question: Did the last player action succeed? Why?
Answer:
Yes, the last player action succeeded. The player was able to move west because there was no obstacle blocking their path, and they had enough energy to perform the action.

Question: Choose the best executable action from above.
Answer:
The best executable action from the above list is Priority 1: Move East (2 steps) since it meets the requirement of having flat ground right to the agent. This action will allow the player to approach the cow, which is 2 steps to the east.



============Step: 151, Cumulative Reward: 0.0============
You took action move_east.

You see:
- path 5 steps to your south-east
- tree 2 steps to your west
- furnace 2 steps to your north-east
- cow 2 steps to your south-east

You face grass at your front (east).

Your status:
- health: 9/9
- food: 4/9
- drink: 3/9
- energy: 8/9

Your inventory:
- sapling: 1
- wood: 1
- coal: 2
- iron: 1
- wood_pickaxe: 1
- stone_pickaxe: 1
------------ QA ------------
Question: Action
Answer:
Move South

Question: List objects in the current observation. For each object, breifly answer what resource it provides and its requirement.
Answer:
1. Path: Provides a clear space to walk on, no requirements.
2. Tree: Provides wood when chopped, requires a wood_pickaxe or better tool.
3. Furnace: Allows smelting of iron and cooking of food (requires coal as a fuel source), requires stone to create.
4. Grass: Doesn't provide any specific resources, can be used to build on or for placing plants, no requirements.
5. Cow: Provides meat for food when killed, requires any weapon (wood sword or better) to defeat.
6. Sapling (inventory): Can be placed on grass to grow into a tree, no requirements.

Question: What was the last action taken by the player?
Answer:
The last action taken by the player was move_east.

Question: For each object in the list, are the requirements met for interaction?
Answer:
1. Path: Yes, the requirements are met as there are no specific requirements for interacting with a path.
2. Tree: Yes, the requirements are met as you have a wood_pickaxe in your inventory.
3. Furnace: Yes, the requirements are met as you have coal in your inventory.
4. Grass: Yes, the requirements are met as there are no specific requirements for interaction with grass.
5. Cow: Yes, the requirements are met as you have a wood_pickaxe in your inventory, which can be used as a weapon.
6. Sapling (inventory): Yes, the requirements are met since there is grass in your environment and there are no specific requirements for placing a sapling.

Question: List top 3 sub-tasks the player should follow. Indicate their priority out of 5.
Answer:
1. Drink water to restore thirst level (Priority: 5/5) - Your drink level is at 3/9, and increasing it is essential for survival.

2. Eat food or kill the cow for meat to restore the food level (Priority: 4/5) - Your food level is at 4/9, and eating will replenish it and help maintain your health.

3. Chop the tree to obtain more wood for crafting and building (Priority: 3/5) - Wood is a basic resource used for crafting various items and building structures, which will be useful in progressing the game.

Question: What are the requirements for the top sub-task? What should the player do first?
Answer:
Top Sub-task: Drink water to restore thirst level

Requirements: Find a water source (like a lake, river, or well) to drink from.

What should the player do first: The player should explore the surroundings to find a water source and then move towards it to drink water and restore their thirst level.

Question: List top 5 actions the player should take and the requirement for each action. Choose ONLY from the list of all actions. Indicate their priority out of 5.
Answer:
1. Priority 1 - Action: Move South (towards the cow)
   Requirement: Flat ground below the agent.

2. Priority 2 - Action: Move East (to find a water source)
   Requirement: Flat ground right to the agent.

3. Priority 3 - Action: Do (to drink water from the water source)
   Requirement: Facing the water source; have no immediate danger from monsters.

4. Priority 4 - Action: Move West (to approach the tree)
   Requirement: Flat ground left to the agent.

5. Priority 5 - Action: Do (to collect wood from the tree)
   Requirement: Facing the tree; have a wood_pickaxe in inventory.

Question: For each action in the list, are the requirements met?
Answer:
1. Priority 1 - Action: Move South (towards the cow)
   Requirement: Flat ground below the agent.
   Met: Yes, there is no obstruction reported below the agent in the observation.

2. Priority 2 - Action: Move East (to find a water source)
   Requirement: Flat ground right to the agent.
   Met: Yes, there is grass in the agent's path, so the ground is flat.

3. Priority 3 - Action: Do (to drink water from the water source)
   Requirement: Facing the water source; have no immediate danger from monsters.
   Met: No, a water source is not observed in the player's observations.

4. Priority 4 - Action: Move West (to approach the tree)
   Requirement: Flat ground left to the agent.
   Met: Yes, the agent has just moved from the west, so the ground is flat.

5. Priority 5 - Action: Do (to collect wood from the tree)
   Requirement: Facing the tree; have a wood_pickaxe in inventory.
   Met: Yes, the agent is facing the tree and has a wood_pickaxe in their inventory.

Question: Did the last player action succeed? Why?
Answer:
Yes, the last player action (move_east) succeeded. It succeeded because there was flat grassland (grass) in front of the player, allowing them to move in the east direction without any obstruction.

Question: Choose the best executable action from above.
Answer:
The best executable action from above is Priority 1 - Action: Move South (towards the cow), as it has the highest priority and its requirement is met.



============Step: 152, Cumulative Reward: 0.0============
You took action move_south.

You see:
- path 4 steps to your east
- tree 1 steps to your west
- furnace 3 steps to your north-east
- cow 2 steps to your south-east

You face grass at your front (south).

Your status:
- health: 9/9
- food: 4/9
- drink: 3/9
- energy: 8/9

Your inventory:
- sapling: 1
- wood: 1
- coal: 2
- iron: 1
- wood_pickaxe: 1
- stone_pickaxe: 1
------------ QA ------------
Question: Action
Answer:
Move East

Question: List objects in the current observation. For each object, breifly answer what resource it provides and its requirement.
Answer:
1. Path: Provides a clear route to travel through the game without obstacles; no requirements.
2. Tree: Provides wood for crafting tools, weapons, and shelter; requires a wood pickaxe or better to collect wood.
3. Furnace: Allows smelting of iron and other ores, as well as cooking food; requires collecting stone and crafting a furnace.
4. Cow: Provides food (meat) upon defeating it, which can be cooked and consumed to restore the food meter; requires no tools, but a weapon is recommended.
5. Grass: Can be collected and converted into dirt, which is useful for building and creating farmland for growing crops; requires no tools.

Question: What was the last action taken by the player?
Answer:
The last action taken by the player was "move_south".

Question: For each object in the list, are the requirements met for interaction?
Answer:
1. Path: Yes, the requirements are met for interaction, as there are no specific requirements to travel on the path.

2. Tree: Yes, the requirements are met for interaction since you have a wood_pickaxe in your inventory.

3. Furnace: Yes, the requirements are met for interaction, as you are within visible proximity to the furnace, which means you can access it.

4. Cow: Yes, the requirements are met for interaction. While no tools are specifically required, you have a wood_pickaxe and a stone_pickaxe that can be used as makeshift weapons to defeat the cow.

5. Grass: Yes, the requirements are met for interaction since no tools are specifically needed to interact with grass.

Question: List top 3 sub-tasks the player should follow. Indicate their priority out of 5.
Answer:
1. Priority 4/5: Restore the drink meter by finding a water source and drinking from it. This is essential for maintaining the player's overall well-being.

2. Priority 3/5: Approach and defeat the cow to obtain meat, which can be cooked and consumed to restore the food meter.

3. Priority 2/5: Collect more wood from the nearby tree using the wood_pickaxe. Wood is a crucial resource for crafting various items and building structures.

Question: What are the requirements for the top sub-task? What should the player do first?
Answer:
Requirements for the top sub-task (Restore the drink meter by finding a water source and drinking from it):

1. Locate a nearby water source such as a lake, river, or pond.
2. Approach the water source.
3. Drink from the water source to restore the drink meter.

The player should first locate a water source in their surroundings. If a water source is not immediately visible, they can explore the environment by moving in different directions until they find a suitable source.

Question: List top 5 actions the player should take and the requirement for each action. Choose ONLY from the list of all actions. Indicate their priority out of 5.
Answer:
Priority 1: Move East (Requirement: Flat ground right to the agent)
- The player should move towards the path since water sources are often found along paths.

Priority 2: Move South (Requirement: Flat ground below the agent)
- The player should continue moving to explore the area in search of a water source.

Priority 3: Move West (Requirement: Flat ground left to the agent)
- If the player has not yet found a water source, they can try moving in another direction to find one.

Priority 4: Move North (Requirement: Flat ground above the agent)
- As a final exploration option, the player can move in the opposite direction of their current path to search for a water source.

Priority 5: Drink (Requirement: Facing a water source)
- Once the player has found a water source, such as a lake or a river, they should drink from the water source to restore their drink meter.

Question: For each action in the list, are the requirements met?
Answer:
Priority 1: Move East (Requirement: Flat ground right to the agent)
- Yes, the requirement for this action is met. The agent is currently on flat grass, and there is no obstacle mentioned to the agent's right (east).

Priority 2: Move South (Requirement: Flat ground below the agent)
- Yes, the requirement for this action is met. The agent just moved to the south from the previous step, indicating flat ground in that direction, and there is no obstacle directly in front of (south) the agent.

Priority 3: Move West (Requirement: Flat ground left to the agent)
- Yes, the requirement for this action is met. The agent is currently on flat grass, and there is no obstacle mentioned to the agent's left (west).

Priority 4: Move North (Requirement: Flat ground above the agent)
- Yes, the requirement for this action is met. The agent is currently on flat grass, and there is no obstacle mentioned to the agent's north.

Priority 5: Drink (Requirement: Facing a water source)
- No, the requirement for this action is not met. Based on the most recent two steps of the player's in-game observation, the player is not currently facing a water source. The agent should continue to explore (moving east, south, west, or north) in search of water before attempting to drink.

Question: Did the last player action succeed? Why?
Answer:
Yes, the last player action succeeded. The player moved south because there was flat ground (grass) in the south direction, allowing the player to move to that location.

Question: Choose the best executable action from above.
Answer:
The best executable action from the above list is Priority 1: Move East, since the requirement (flat ground right to the agent) is met and it takes the player towards the path where water sources are often found.



============Step: 153, Cumulative Reward: 0.0============
You took action move_east.

You see:
- stone 5 steps to your north-east
- path 3 steps to your east
- tree 2 steps to your west
- furnace 2 steps to your north
- cow 2 steps to your south

You face grass at your front (east).

Your status:
- health: 9/9
- food: 4/9
- drink: 3/9
- energy: 8/9

Your inventory:
- sapling: 1
- wood: 1
- coal: 2
- iron: 1
- wood_pickaxe: 1
- stone_pickaxe: 1
------------ QA ------------
Question: Action
Answer:
Move South

Question: List objects in the current observation. For each object, breifly answer what resource it provides and its requirement.
Answer:
1. Stone: Provides stone resource; requires a wood pickaxe or better to mine.
2. Path: Provides easier movement across the terrain; no requirements.
3. Tree: Provides wood resource; requires punching or using an axe to chop down.
4. Furnace: Provides a location for smelting iron and cooking food; requires stone to craft and place.
5. Cow: Provides food when killed; requires a weapon or hands to kill.
6. Grass: Provides no resources; no requirements.

Question: What was the last action taken by the player?
Answer:
The last action taken by the player was move_east.

Question: For each object in the list, are the requirements met for interaction?
Answer:
1. Stone: Yes, the requirements are met as you have a wood_pickaxe and a stone_pickaxe in your inventory.
2. Path: Yes, the requirements are met since there are no specific requirements for interacting with a path.
3. Tree: Yes, the requirements are met as you can punch the tree or use an axe to chop it down.
4. Furnace: Yes, the requirements are met as the furnace is already placed and you have coal and iron in your inventory for smelting.
5. Cow: Yes, the requirements are met as you can use a weapon or your hands to kill the cow for food.
6. Grass: Yes, the requirements are met as there are no specific requirements to interact with grass.

Question: List top 3 sub-tasks the player should follow. Indicate their priority out of 5.
Answer:
1. Priority 5/5: Collect more food and drink to increase your food and drink levels. Killing the cow nearby for food and finding a water source to increase your drink level.

2. Priority 4/5: Use the furnace to smelt the iron in your inventory, using the coal as fuel. This will allow you to craft better tools and weapons.

3. Priority 3/5: Chop down the tree to your west to collect more wood, which can be used for crafting additional items or structures in the future.

Question: What are the requirements for the top sub-task? What should the player do first?
Answer:
Requirements for the top sub-task (Collect more food and drink):

1. Killing the cow nearby for food: No specific requirements, just approach the cow and use one of your weapons to kill it, which will allow you to collect meat.

2. Finding a water source to increase your drink level: Finding a lake or river will provide a source for you to collect water and increase your drink level. You do not need any specific items to interact with water sources, just walk up to it and drink.

The player should first approach the cow 2 steps to the south and attempt to kill it using a weapon like the wood or stone pickaxe to obtain meat. This will address the immediate concern of low food levels.

Question: List top 5 actions the player should take and the requirement for each action. Choose ONLY from the list of all actions. Indicate their priority out of 5.
Answer:
1. Priority 1: Move South (Requirement: Flat ground below the agent) - Move towards the cow to kill it for food.

2. Priority 2: Do (Requirement: Facing creature or material; have necessary tool) - Kill the cow using your stone_pickaxe or wood_pickaxe when you reach it.

3. Priority 3: Move to find a water source (Possible Actions: Move West, Move East, Move North, Move South; Requirement: Flat ground in the direction of movement) - Look around for a nearby water source, like a lake or a river, to increase your drink level.

4. Priority 4: Do (Requirement: Facing creature or material) - Drink water from the water source once you've found it.

5. Priority 5: Do (Requirement: Furnace nearby, coal and iron in inventory) - Smelt Iron using the furnace you saw 2 steps to your north (3 steps after moving towards the cow) to progress towards crafting an iron sword and pickaxe.

Question: For each action in the list, are the requirements met?
Answer:
1. Move South: Yes, the requirement (Flat ground below the agent) is met, as you face grass to your south.

2. Do (Kill the cow): No, the requirement is not met yet because you are not facing the cow. You need to move closer and face the cow before you can use this action.

3. Move to find a water source: Yes, the requirement (Flat ground in the direction of movement) is met for all possible actions (Move West, Move East, Move North, Move South). However, the actual water source location is unknown, so trial and possible exploration are needed.

4. Do (Drink water): No, the requirement (Facing a water source) is not met, as you haven't found a water source yet. You need to find a water source before you can use this action.

5. Do (Smelt Iron): No, the requirement (Furnace nearby, coal and iron in inventory) is partially met - you have coal and iron in your inventory, but you are not close enough to the furnace (2 steps to your north). You need to move closer to the furnace before you can use this action.

Question: Did the last player action succeed? Why?
Answer:
Yes, the last player action succeeded. The player moved east because there was flat ground (grass) to the east, allowing them to move in that direction.

Question: Choose the best executable action from above.
Answer:
The best executable action from the above list is to move south (Priority 1) because the requirement (Flat ground below the agent) is met, and it will bring you closer to the cow, which you can kill for food.



============Step: 154, Cumulative Reward: 0.0============
You took action move_south.

You see:
- grass 1 steps to your west
- stone 6 steps to your north-east
- path 3 steps to your east
- tree 3 steps to your north-west
- table 6 steps to your south-east
- furnace 3 steps to your north

You face cow at your front (south).

Your status:
- health: 9/9
- food: 4/9
- drink: 3/9
- energy: 8/9

Your inventory:
- sapling: 1
- wood: 1
- coal: 2
- iron: 1
- wood_pickaxe: 1
- stone_pickaxe: 1
------------ QA ------------
Question: Action
Answer:
Move East

Question: List objects in the current observation. For each object, breifly answer what resource it provides and its requirement.
Answer:
1. Stone: Provides stone resource; requires a wood_pickaxe or better tool to collect.
2. Path: No resource provided; serves as a route for easy movement.
3. Tree: Provides wood resource; can be collected with bare hands or any tool.
4. Furnace: Provides the ability to smelt iron and cook food; requires stone to build.
5. Cow: Provides food upon defeating; no requirements for defeating, but a weapon (wood_sword, stone_sword, or iron_sword) is recommended.
6. Grass: No significant resource provided; can be trampled or removed if needed.
7. Table: Provides a surface for crafting tools and weapons; requires wood to build.

Question: What was the last action taken by the player?
Answer:
The last action taken by the player was move_south.

Question: For each object in the list, are the requirements met for interaction?
Answer:
1. Stone: Yes, requirements met. You have a wood_pickaxe and a stone_pickaxe in your inventory, so you can collect the stone.
2. Path: Yes, requirements met. The path is accessible and can be walked upon with no specific requirements.
3. Tree: Yes, requirements met. You can collect wood either with your bare hands or using a tool.
4. Furnace: Yes, requirements met. You have already placed the furnace and can use it for smelting or cooking with coal as a fuel, which you have in your inventory.
5. Cow: Yes, requirements met. You can defeat the cow to get food without any specific requirements. A weapon is preferred but not necessary.
6. Grass: Yes, requirements met. You can trample or remove the grass without any specific requirements.
7. Table: Partially met. You have wood in your inventory, which is needed for building the table. To interact with it, you need to be adjacent to the table, which is currently six steps away.

Question: List top 3 sub-tasks the player should follow. Indicate their priority out of 5.
Answer:
1. Priority 5/5: Increase your food and drink levels. Defeat the cow in front of you to obtain food, and find water (lake or river) to drink and replenish your thirst.

2. Priority 4/5: Smelt the iron in the furnace (3 steps to your north) using the coal in your inventory. This will allow you to create better tools and weapons.

3. Priority 3/5: Build and place a table using the wood in your inventory, so you can craft tools and weapons without having to travel to the distant table.

Question: What are the requirements for the top sub-task? What should the player do first?
Answer:
The requirements for the top sub-task (defeating the cow to obtain food and finding water to drink) are:

1. Defeat the cow - Ensure you have a weapon equipped, such as the wood or stone sword. If you don't have a weapon yet, craft one first using a table. Next, face the cow and use the "Do" action to defeat it. Once defeated, collect food drops from the cow.

2. Find water - Locate a nearby water source, like a lake or a river. It may require exploring the area by moving in different directions until you find water. Once found, use the "Collect Drink" action to drink water and replenish your thirst level.

The player should first focus on defeating the cow in front of them to obtain food.

Question: List top 5 actions the player should take and the requirement for each action. Choose ONLY from the list of all actions. Indicate their priority out of 5.
Answer:
Based on the current situation, the top 5 actions the player should take are:

1. Priority 1: Make Wood Sword (Requirement: Nearby table; wood in inventory) - Craft a wood sword to defend yourself against the cow and any upcoming monsters.

2. Priority 2: Do (Requirement: Facing creature or material; have necessary tool) - Use the crafted wood sword to defeat the cow, obtaining food in the process.

3. Priority 3: Move East (Requirement: Flat ground right to the agent) - Move towards the path to the east to search for a water source.

4. Priority 4: Collect Drink (Requirement: Near water source) - Find and drink water from a water source like a lake or river to replenish your thirst.

5. Priority 5: Place Furnace (Requirement: Stone in inventory) - Place the furnace to smelt iron, allowing you to eventually craft better tools and weapons such as the iron pickaxe and iron sword.

Question: For each action in the list, are the requirements met?
Answer:
1. Make Wood Sword: The requirement "nearby table" is not met, as the table is 6 steps to the southeast. The player needs to be closer to the table. Other requirements are met (wood in inventory).

2. Do: The player facing the cow, but they don't have the necessary tool (wood sword) in their inventory. The requirement to have a wood sword is not met.

3. Move East: The requirement is met, as there is flat ground to the east of the player.

4. Collect Drink: Currently, the player hasn't discovered a water source nearby. The requirement of being near a water source is not met.

5. Place Furnace: The requirement of having stone in the inventory is not met as the player currently doesn't have any stones in their inventory.

Question: Did the last player action succeed? Why?
Answer:
Yes, the last player action (move_south) succeeded because there was a clear path to the south, allowing the player to move in that direction.

Question: Choose the best executable action from above.
Answer:
The best executable action from the list above is:

Move East (Requirement: Flat ground right to the agent) - Move towards the path to the east to search for a water source.



============Step: 155, Cumulative Reward: 0.0============
You took action move_east.

You see:
- stone 4 steps to your east
- path 2 steps to your east
- tree 4 steps to your north-west
- table 5 steps to your south-east
- furnace 4 steps to your north-west
- cow 2 steps to your south-west

You face grass at your front (east).

Your status:
- health: 9/9
- food: 4/9
- drink: 3/9
- energy: 8/9

Your inventory:
- sapling: 1
- wood: 1
- coal: 2
- iron: 1
- wood_pickaxe: 1
- stone_pickaxe: 1
------------ QA ------------
Question: Action
Answer:
Move South

Question: List objects in the current observation. For each object, breifly answer what resource it provides and its requirement.
Answer:
1. Stone:
Resource it provides: Stone is used to craft stone pickaxes, furnaces, and stone swords.
Requirement: You need a wood pickaxe to collect stone.

2. Path:
Resource it provides: Paths don't provide resources but allow you to move easily between locations.
Requirement: No requirement.

3. Tree:
Resource it provides: Trees provide wood, which is essential for crafting tools and building structures.
Requirement: No requirement to interact, but you need an axe or a similar tool to break the tree and collect wood.

4. Table:
Resource it provides: The table is used to craft various tools and weapons, e.g., pickaxes and swords.
Requirement: You need wood to craft a table.

5. Furnace:
Resource it provides: The furnace is used to smelt iron from iron ores (requires coal) and can be used to create iron tools and weapons.
Requirement: You need stone to craft a furnace.

6. Cow:
Resource it provides: Cows provide food (meat) to replenish your food levels.
Requirement: No requirement to interact, but you need a weapon, like a sword, to kill the cow and collect meat.

7. Grass:
Resource it provides: Grass does not provide any significant resources but can be used as an indication of a flat, walkable surface.
Requirement: No requirement.

Question: What was the last action taken by the player?
Answer:
The last action taken by the player was move_east.

Question: For each object in the list, are the requirements met for interaction?
Answer:
1. Stone:
Requirement for interaction: Wood pickaxe
Requirement met: Yes, you have a wood_pickaxe in your inventory.

2. Path:
Requirement for interaction: None
Requirement met: Yes, no requirements to interact with the path.

3. Tree:
Requirement for interaction: Axe or similar tool
Requirement met: No, you don't have an axe in your inventory but you have a wood_pickaxe that may work.

4. Table:
Requirement for interaction: None
Requirement met: Yes, no requirements to interact with the table.

5. Furnace:
Requirement for interaction: None
Requirement met: Yes, no requirements to interact with the furnace.

6. Cow:
Requirement for interaction: Weapon (e.g., sword)
Requirement met: No, you don't have a sword in your inventory.

7. Grass:
Requirement for interaction: None
Requirement met: Yes, no requirements to interact with the grass.

Question: List top 3 sub-tasks the player should follow. Indicate their priority out of 5.
Answer:
1. Craft a weapon (Priority: 5/5)
To defend yourself against monsters and attack the cow for food, you need to craft a weapon, such as a wood sword or a stone sword. You have wood and stone in your inventory to do so.

2. Increase food and drink levels (Priority: 4/5)
Your food level is 4/9 and drink level is 3/9. After crafting a weapon, you should kill the cow for meat to increase your food level. Additionally, you should find a water source like a lake or river to drink from to replenish your drink level.

3. Collect more resources (Priority: 3/5)
You need more resources like wood, stone, and iron to craft better tools and weapons. Collect wood from trees using your wood_pickaxe and stone from nearby stone using your stone_pickaxe. Look for iron ores once you have an ample supply of wood and stone.

Question: What are the requirements for the top sub-task? What should the player do first?
Answer:
Top Sub-task: Craft a weapon

Requirements: 
1. A nearby crafting table: You have a table 5 steps to your south-east.
2. Necessary materials in your inventory: You have 1 wood and 1 stone.

What should the player do first:
Move to the crafting table by going south-east. Once you're near the table, interact with it, and craft a weapon (either a wood sword with the available wood, or a stone sword using the available stone).

Question: List top 5 actions the player should take and the requirement for each action. Choose ONLY from the list of all actions. Indicate their priority out of 5.
Answer:
1. Priority 1: Move South-East
- Requirement: Flat ground towards the south-east direction
- Purpose: To approach the crafting table.

2. Priority 2: Make Stone Sword
- Requirement: Nearby table; wood, stone in inventory
- Purpose: To craft a stronger weapon for battling monsters.

3. Priority 3: Move West or North to find a water source
- Requirement: Flat ground to the north or west
- Purpose: To collect water for drinking and restoring thirst.

4. Priority 4: Collect Drink
- Requirement: Facing a water source
- Purpose: To restore thirst and maintain a healthy status.

5. Priority 5: Eat Cow
- Requirement: Be near a cow
- Purpose: To restore food levels and maintain a healthy status.

Question: For each action in the list, are the requirements met?
Answer:
1. Priority 1: Move South-East
- Requirements met: Yes, as there is no obstruction mentioned, the player can move in the south-east direction to approach the crafting table.

2. Priority 2: Make Stone Sword
- Requirements met: No, the player is not currently near a table, and there is no stone in the inventory. The player needs to move closer to the table and collect stone to make a stone sword.

3. Priority 3: Move West or North to find a water source
- Requirements met: Partially, the player can move west or north, as there are flat grounds in these directions. However, it is uncertain if there is a water source in these directions.

4. Priority 4: Collect Drink
- Requirements met: No, the player needs to find and face a water source before collecting a drink. It is not mentioned whether the player is currently near a water source.

5. Priority 5: Eat Cow
- Requirements met: No, the player needs to move closer to the cow mentioned two steps to the south-west from player's current position, and then the requirement will be met.

Question: Did the last player action succeed? Why?
Answer:
Yes, the last player action, move_east, succeeded because there was flat ground (grass) to the east of the player, allowing them to move in that direction without any obstacles.

Question: Choose the best executable action from above.
Answer:
The best executable action from the list above is Priority 1: Move South-East. The requirements are met, and it will help the player to approach the crafting table.



============Step: 156, Cumulative Reward: 0.0============
You took action move_south.

You see:
- stone 5 steps to your north-east
- path 2 steps to your east
- tree 5 steps to your north-west
- table 4 steps to your south-east
- cow 1 steps to your west

You face grass at your front (south).

Your status:
- health: 9/9
- food: 4/9
- drink: 3/9
- energy: 8/9

Your inventory:
- sapling: 1
- wood: 1
- coal: 2
- iron: 1
- wood_pickaxe: 1
- stone_pickaxe: 1
------------ QA ------------
Question: Action
Answer:
Place Plant

Question: List objects in the current observation. For each object, breifly answer what resource it provides and its requirement.
Answer:
1. Stone: Provides stone as a resource for crafting tools and building materials. Requires a wood pickaxe or stronger to collect.
2. Path: Provides no direct resources but helps with navigation and movement. No specific requirements.
3. Tree: Provides wood for crafting tools, building materials, and fuel. Requires a wood_pickaxe or stronger to collect.
4. Table: Provides a surface for crafting various tools and weapons. Requires wood to craft and place.
5. Furnace: Provides a place to smelt ores (such as iron) and process certain materials. Requires stone to craft and place.
6. Cow: Provides food when killed. No specific tool requirement, but a weapon like a wood sword or stronger is recommended to defeat it more efficiently.
7. Grass: No direct resource provided. No specific requirements, can be walked on or replaced with other blocks.

Question: What was the last action taken by the player?
Answer:
The last action taken by the player was move_south.

Question: For each object in the list, are the requirements met for interaction?
Answer:
1. Stone: Yes, the requirements are met for interaction. You have a wood_pickaxe and a stone_pickaxe in your inventory which can be used to collect stone.
2. Path: Yes, the requirements are met for interaction. No specific requirements are needed for paths, as they are just for navigation and movement.
3. Tree: Yes, the requirements are met for interaction. You have a wood_pickaxe in your inventory, which can be used to collect wood from the tree.
4. Table: Not applicable, as the table has already been placed, but you meet the requirements to craft one since you have wood in your inventory.
5. Furnace: Not applicable, as the furnace has already been placed, but you meet the requirements to craft one since you have stone in your inventory.
6. Cow: Yes, the requirements are met for interaction. Although you don't have a sword in your inventory, you can still interact with and defeat the cow without it; however, having a weapon would make it more efficient.
7. Grass: Yes, the requirements are met for interaction. Grass does not have any specific requirements, and you can walk on it or replace it with other blocks.

Question: List top 3 sub-tasks the player should follow. Indicate their priority out of 5.
Answer:
1. Priority 4/5: Collect drink to restore thirst. Your drink level is at 3/9, which is relatively low. You should find a water source like a lake or river to drink from, so you don't get dehydrated in the game.

2. Priority 3/5: Increase food level. Your food level is at 4/9, which is not critical, but it's important to maintain a steady food supply. Consider interacting with the cow or find another source of food like fruit to keep your energy up.

3. Priority 2/5: Collect more resources (wood and iron) for crafting and upgrading tools. You have a few resources in your inventory, but gathering more wood and iron will allow you to craft better tools and weapons, as well as build more complex structures for shelter and other purposes. It would be ideal to collect more wood from the nearby tree and use your stone_pickaxe to collect more iron.

Question: What are the requirements for the top sub-task? What should the player do first?
Answer:
Top sub-task: Collect drink to restore thirst.

Requirements: Find a water source like a lake or river to drink from.

What should the player do first: Start by exploring your surroundings to locate a water source. You can move in any direction (west, east, north, or south) to look for a lake or river. Keep an eye on your observation messages for hints of water nearby. Once you find the water source, approach it and drink from it to restore your thirst level.

Question: List top 5 actions the player should take and the requirement for each action. Choose ONLY from the list of all actions. Indicate their priority out of 5.
Answer:
1. Priority 1: Move East, South, North, or West (chosen based on proximity to water or other resources)
   Requirement: Flat ground in the direction the agent moves.

2. Priority 2: Drink (Do)
   Requirement: Facing the water source (lake or river) and be close enough to interact with it.

3. Priority 3: Make Wood Sword (for protection)
   Requirement: Nearby table and wood in inventory.

4. Priority 4: Place Plant (to grow food)
   Requirement: Sapling in inventory.

5. Priority 5: Eat Plant (to restore food)
   Requirement: Facing a fully grown plant (sapling turned into a fruit-bearing plant).

Question: For each action in the list, are the requirements met?
Answer:
1. Priority 1: Move East, South, North, or West
   Requirement: Flat ground in the direction the agent moves.
   Status: Met - the agent can move in any direction with flat ground observed.

2. Priority 2: Drink (Do)
   Requirement: Facing the water source (lake or river) and be close enough to interact with it.
   Status: Not met - no water source is currently observed.

3. Priority 3: Make Wood Sword (for protection)
   Requirement: Nearby table and wood in inventory.
   Status: Not met - there's a table observed, but it's far (5 steps south-east). However, the player already has 1 wood in inventory.

4. Priority 4: Place Plant (to grow food)
   Requirement: Sapling in inventory.
   Status: Met - the player has a sapling in their inventory.

5. Priority 5: Eat Plant (to restore food)
   Requirement: Facing a fully grown plant (sapling turned into a fruit-bearing plant).
   Status: Not met - no fully grown plant is currently observed.

Question: Did the last player action succeed? Why?
Answer:
Yes, the last player action succeeded. The move_south action was successful because there was flat ground (grass) in the direction the player wanted to move (south) and the player had enough energy to perform the action.

Question: Choose the best executable action from above.
Answer:
The best executable action from the list above is Priority 4: Place Plant (to grow food), as its requirements are met and the player has a sapling in their inventory.



============Step: 157, Cumulative Reward: 1.0============
You took action place_plant.

You see:
- grass 1 steps to your north
- stone 5 steps to your north-east
- path 2 steps to your east
- tree 5 steps to your north-west
- table 4 steps to your south-east
- cow 1 steps to your west

You face plant at your front (south).

Your status:
- health: 9/9
- food: 4/9
- drink: 2/9
- energy: 8/9

Your inventory:
- wood: 1
- coal: 2
- iron: 1
- wood_pickaxe: 1
- stone_pickaxe: 1
------------ QA ------------
Question: Action
Answer:
Do

Question: List objects in the current observation. For each object, breifly answer what resource it provides and its requirement.
Answer:
1. Stone: Provides stone resource; requires a wood pickaxe or better to collect.
2. Path: Provides ease of movement; no requirements.
3. Tree: Provides wood resource; requires no tool or a wood axe to collect.
4. Table: Allows crafting of tools and weapons; requires wood to make and place.
5. Cow: Provides food (meat) when defeated; no tool required, but a weapon makes it easier.
6. Grass: No resources; no requirements.
7. Plant: Will grow and provide food; requires sapling to place, time and protection to grow.

Question: What was the last action taken by the player?
Answer:
The last action taken by the player was "place_plant".

Question: For each object in the list, are the requirements met for interaction?
Answer:
1. Stone: Yes, the requirements for interaction are met as you have a wood_pickaxe and a stone_pickaxe in your inventory.
2. Path: Yes, there are no specific requirements to interact with the path.
3. Tree: Yes, the requirements for interaction are met as you don't need a tool to collect wood or you can use an axe (not in your inventory currently).
4. Table: Yes, the requirements for interaction are met since you have wood in your inventory and you can place a table if needed.
5. Cow: Yes, the requirements for interaction are met as you can defeat the cow without a tool or by using a weapon (not in your inventory currently).
6. Grass: Yes, there are no specific requirements to interact with grass.
7. Plant: Yes, the requirements for interaction with the plant are met since you have placed the sapling and it is now growing. Protection and time are required for the plant to grow fully.

Question: List top 3 sub-tasks the player should follow. Indicate their priority out of 5.
Answer:
1. Find a water source and drink: Priority 5/5 - With the drink level at 2/9, finding a water source and drinking from it is essential to prevent dehydration.

2. Eat food to replenish hunger: Priority 4/5 - The food level is low at 4/9, and the player needs to eat food, such as defeating the cow and eating its meat, to maintain good health.

3. Craft a weapon (wood/stone sword) for self-defense: Priority 3/5 - Although the immediate environment seems safe, the player should craft a weapon by interacting with the table to be prepared for any potential threats from hostile creatures.

Question: What are the requirements for the top sub-task? What should the player do first?
Answer:
Top Sub-task: Find a water source and drink.

Requirements: Locate a water source (such as a lake, river, or pond) and be close enough to interact with it.

What should the player do first: The player should start by exploring the surrounding area, moving in different directions to find a water source. Once located, stand at the water's edge and take the action "drink" to restore the drink level.

Question: List top 5 actions the player should take and the requirement for each action. Choose ONLY from the list of all actions. Indicate their priority out of 5.
Answer:
1. Priority 1: Move (Direction) - Requirement: Flat ground in the direction the player chooses (North, East, South, or West). This action is essential for exploring the environment and searching for resources like water sources, wood, and stones.

2. Priority 2: Drink - Requirement: Near a water source (such as a lake or river). This action is significant for restoring the player's drink level and maintaining their overall survival.

3. Priority 3: Make Wood Pickaxe - Requirement: Nearby table and wood in inventory. Crafting a wood pickaxe allows the player to collect stone and other essential materials for further crafting and progress in the game.

4. Priority 4: Collect Wood - Requirement: None. Collecting wood is fundamental for creating various crafting stations, tools, and shelter. It is also required for making a wood pickaxe, which is indispensable for gathering more advanced resources.

5. Priority 5: Place Table - Requirement: Wood in inventory. Placing a table enables the player to craft tools such as pickaxes and swords, essential for gathering resources and defending against monsters.

Question: For each action in the list, are the requirements met?
Answer:
1. Priority 1: Move (Direction) - Yes, the requirements are met for any direction with flat ground. The player can move North, East, South, or West based on their choice and explore their environment.

2. Priority 2: Drink - No, the requirement is not met as there is no mention of a water source (such as a lake or river) in the player's current observation.

3. Priority 3: Make Wood Pickaxe - No, the requirement is not met because the player already has a wood_pickaxe in their inventory.

4. Priority 4: Collect Wood - Yes, the requirement is met. There is a tree 5 steps to the player's north-west. The player can move towards the tree, and with the wood_pickaxe, collect more wood.

5. Priority 5: Place Table - No, the requirement is not met because there is already a table 4 steps to the player's south-east. Since the table is in the player's proximity, there is no need to place another table.

Question: Did the last player action succeed? Why?
Answer:
Yes, the last player action "place_plant" succeeded because the player had a sapling in their inventory, which allowed them to place the plant in the environment.

Question: Choose the best executable action from above.
Answer:
The best executable action from the list above is Priority 1: Move (Direction). The player can choose a direction with flat ground and explore their environment, such as moving towards the tree to collect more wood or searching for a water source to drink from.



============Step: 158, Cumulative Reward: 0.0============
You took action do.

You see:
- stone 5 steps to your north-east
- path 2 steps to your east
- tree 5 steps to your north-west
- table 4 steps to your south-east
- cow 2 steps to your south-west

You face grass at your front (south).

Your status:
- health: 9/9
- food: 4/9
- drink: 2/9
- energy: 8/9

Your inventory:
- wood: 1
- coal: 2
- iron: 1
- wood_pickaxe: 1
- stone_pickaxe: 1
------------ QA ------------
Question: Action
Answer:
Move West

Question: List objects in the current observation. For each object, breifly answer what resource it provides and its requirement.
Answer:
- Stone: Provides stone resource, requires wood or stone pickaxe to collect.
- Path: Provides a walking area, no requirements.
- Tree: Provides wood, requires any type of pickaxe to collect.
- Table: Craftable object (already created), provides a crafting surface to make tools and weapons, requires wood to create.
- Cow: Provides food, requires any weapon or tool to defeat.
- Grass: Provides no resource, can be walked on, no requirements.

Question: What was the last action taken by the player?
Answer:
The last action taken by the player was "do".

Question: For each object in the list, are the requirements met for interaction?
Answer:
- Stone: Yes, the requirements are met as you have a wood_pickaxe and a stone_pickaxe in your inventory.
- Path: Yes, the requirements are met as there is no restriction for walking on the path.
- Tree: Yes, the requirements are met as you have a wood_pickaxe and a stone_pickaxe in your inventory.
- Table: Yes, the requirements are met as the table is already created and can be used for crafting.
- Cow: Yes, the requirements are met as you can use any weapon or tool in your inventory to defeat the cow.
- Grass: Yes, the requirements are met as there are no restrictions for walking on or interacting with grass.

Question: List top 3 sub-tasks the player should follow. Indicate their priority out of 5.
Answer:
1. Drink water to restore thirst (Priority: 5/5) - Your drink level is currently 2/9, and it is essential to maintain hydration in the game.

2. Eat food to restore hunger (Priority: 4/5) - Your food level is currently 4/9. Defeating the cow nearby and eating its resources can help increase your food level.

3. Collect more resources, such as wood and stone, to craft better tools and weapons (Priority: 3/5) - This will help improve your in-game abilities and enable you to face challenges more effectively.

Question: What are the requirements for the top sub-task? What should the player do first?
Answer:
Top sub-task: Drink water to restore thirst (Priority: 5/5)

Requirements: Locate a source of water, such as a lake, river, or well, to drink from.

What the player should do first: The player should explore their surroundings, moving in different directions to find a nearby water source. Once a water source is found, the player should approach it and perform the action to drink the water and restore their thirst level.

Question: List top 5 actions the player should take and the requirement for each action. Choose ONLY from the list of all actions. Indicate their priority out of 5.
Answer:
1. Priority: 5/5
   Action: Move West, East, North, or South 
   Requirement: Flat ground in the chosen direction
   Purpose: Explore the surroundings to find a water source

2. Priority: 4/5
   Action: Do
   Requirement: Facing a water source
   Purpose: Drink water to restore thirst level

3. Priority: 4/5
   Action: Move West, East, North, or South
   Requirement: Flat ground in the chosen direction
   Purpose: Approach a cow to restore the food level

4. Priority: 3/5
   Action: Do
   Requirement: Facing a cow
   Purpose: Eat a cow to restore the food level

5. Priority: 2/5
   Action: Sleep
   Requirement: Energy level is below maximum
   Purpose: Restore energy level

Question: For each action in the list, are the requirements met?
Answer:
1. Action: Move West, East, North, or South
   Requirement: Flat ground in the chosen direction
   Status: Met - All directions have flat ground for movement

2. Action: Do
   Requirement: Facing a water source
   Status: Not met - The player is not currently facing a water source

3. Action: Move West, East, North, or South
   Requirement: Flat ground in the chosen direction
   Status: Met - All directions have flat ground for movement

4. Action: Do
   Requirement: Facing a cow
   Status: Not met - The player is not currently facing a cow

5. Action: Sleep
   Requirement: Energy level is below maximum
   Status: Not met - Player's energy level is 8/9, which is not below maximum

Question: Did the last player action succeed? Why?
Answer:
It is not clear whether the last player action "do" succeeded, as the provided observation does not specify the intended result of the action or any changes in the player's environment, inventory, or status after performing the action.

Question: Choose the best executable action from above.
Answer:
The best executable action from the list is the first action:

Action: Move West, East, North, or South
Requirement: Flat ground in the chosen direction

This action has its requirements met, and the player can explore the surroundings to find a water source.



============Step: 159, Cumulative Reward: 0.0============
You took action move_west.

You see:
- stone 7 steps to your north-east
- path 3 steps to your east
- tree 4 steps to your north-west
- table 5 steps to your south-east
- cow 1 steps to your south

You face grass at your front (west).

Your status:
- health: 9/9
- food: 4/9
- drink: 2/9
- energy: 8/9

Your inventory:
- wood: 1
- coal: 2
- iron: 1
- wood_pickaxe: 1
- stone_pickaxe: 1
------------ QA ------------
Question: Action
Answer:
Do

Question: List objects in the current observation. For each object, breifly answer what resource it provides and its requirement.
Answer:
1. Stone: Provides stone, a resource needed for crafting tools and furnaces; requires a wood pickaxe or higher to collect.
2. Path: Flat ground that makes it easier to walk; no requirements, provides ease of movement.
3. Tree: Provides wood, a resource for crafting, building, and fuel; requires an axe or chopping tool to collect.
4. Table: A crafting station for crafting tools and weapons; requires wood to build and place.
5. Cow: Provides food (meat) for restoring health; requires defeating the cow to collect meat.

Question: What was the last action taken by the player?
Answer:
The last action taken by the player was "move_west."

Question: For each object in the list, are the requirements met for interaction?
Answer:
1. Stone: Yes, the requirements are met for interaction as you have both a wood_pickaxe and a stone_pickaxe in your inventory.
2. Path: Yes, the requirements are met for interaction as there are no specific requirements for interacting with a path.
3. Tree: No, the requirements for interaction are not met, as you don't have an axe or chopping tool in your inventory.
4. Table: Yes, the requirements are met for interaction as you have wood in your inventory.
5. Cow: Yes, the requirements are met for interaction, but you don't have a specific tool or weapon to defeat the cow, which might make the process easier or faster.

Question: List top 3 sub-tasks the player should follow. Indicate their priority out of 5.
Answer:
1. Obtain drinking water (Priority: 5/5) - Your current drink level is low (2/9), so finding a water source like a lake, river, or well to quench your thirst is crucial for survival.

2. Collect food (Priority: 4/5) - Your current food level is at 4/9, meaning you need to find more food, such as by defeating the cow to collect meat, or growing and harvesting plants to maintain your health.

3. Craft necessary tools/weapons (Priority: 3/5) - Crafting an axe for chopping wood and a sword for defeating monsters like zombies, skeletons, and spiders faster will help you progress in the game and keep you safe. You already have wood and stone in your inventory, which can be used to craft an axe and/or a sword.

Question: What are the requirements for the top sub-task? What should the player do first?
Answer:
Requirements for the top sub-task (Obtain drinking water):

1. Find a water source like a lake, river, or well.

First, the player should explore the surroundings to locate a nearby water source. The player can move in different directions (west, east, north, or south) to search for a water source. Once they find a water source, they should approach it and interact with it to drink the water and replenish their thirst.

Question: List top 5 actions the player should take and the requirement for each action. Choose ONLY from the list of all actions. Indicate their priority out of 5.
Answer:
Priority 1:
Action: Move (North, South, East, or West)
Requirement: Flat ground in the direction the player is moving.

Priority 2:
Action: Collect Drink
Requirement: Player facing a water source (lake, river, or well).

Priority 3:
Action: Do (interact with cow)
Requirement: Player facing a cow; have necessary tool (like a sword).

Priority 4:
Action: Sleep 
Requirement: Energy level is below maximum.

Priority 5:
Action: Do (interact with tree)
Requirement: Player facing a tree; have necessary tool (wood_pickaxe or stronger).

Question: For each action in the list, are the requirements met?
Answer:
Priority 1:
Action: Move (North, South, East, or West)
Requirement: Flat ground in the direction the player is moving.
Requirement Met: Yes. The player can move in any direction, as there are no obstacles mentioned.

Priority 2:
Action: Collect Drink
Requirement: Player facing a water source (lake, river, or well).
Requirement Met: No. There is no mention of a water source in the player's observations.

Priority 3:
Action: Do (interact with cow)
Requirement: Player facing a cow; have necessary tool (like a sword).
Requirement Met: No. The player is not currently facing the cow.

Priority 4:
Action: Sleep 
Requirement: Energy level is below maximum.
Requirement Met: Yes. The player's energy is at 8/9.

Priority 5:
Action: Do (interact with tree)
Requirement: Player facing a tree; have necessary tool (wood_pickaxe or stronger).
Requirement Met: No. The player is not currently facing the tree and has no stronger tool than wood_pickaxe.

Question: Did the last player action succeed? Why?
Answer:
Yes, the last player action succeeded. The player successfully moved west because there was flat ground (grass) in that direction, allowing the player to move.

Question: Choose the best executable action from above.
Answer:
The best executable action from above is:

Priority 1:
Action: Move (North, South, East, or West)
Requirement: Flat ground in the direction the player is moving.
Requirement Met: Yes. The player can move in any direction, as there are no obstacles mentioned.



============Step: 160, Cumulative Reward: 0.0============
You took action do.

You see:
- stone 7 steps to your north-east
- path 3 steps to your east
- tree 4 steps to your north-west
- table 5 steps to your south-east
- cow 2 steps to your south

You face grass at your front (west).

Your status:
- health: 9/9
- food: 4/9
- drink: 2/9
- energy: 8/9

Your inventory:
- wood: 1
- coal: 2
- iron: 1
- wood_pickaxe: 1
- stone_pickaxe: 1
------------ QA ------------
Question: Action
Answer:
Move North

Question: List objects in the current observation. For each object, breifly answer what resource it provides and its requirement.
Answer:
1. Stone: Provides stone resource when mined; requires a wood or better pickaxe to mine.
2. Path: No specific resources; facilitates easier navigation between locations.
3. Tree: Provides wood resource when chopped; requires a wood or better axe to chop down.
4. Table: Aims for crafting tools and weapons; requires wood to craft.
5. Cow: Provides food when killed; no specific requirement to kill, but using a weapon like a wood or better sword is helpful.

Question: What was the last action taken by the player?
Answer:
The last action taken by the player was "do."

Question: For each object in the list, are the requirements met for interaction?
Answer:
1. Stone: Yes, the requirements are met for interaction as you have a wood_pickaxe and a stone_pickaxe in your inventory.
2. Path: Yes, the requirements are met for interaction as there are no specific requirements to navigate paths.
3. Tree: No, the requirements are not met for interaction as you don't have an axe in your inventory. You need a wood or better axe to chop down the tree.
4. Table: Yes, the requirements are met for interaction as you have wood in your inventory, which is required to craft most tools and weapons.
5. Cow: Yes, the requirements are met for interaction. Although you don't have a sword in your inventory, you can still try to kill the cow using other means, but having a wood or better sword would be more efficient.

Question: List top 3 sub-tasks the player should follow. Indicate their priority out of 5.
Answer:
1. Priority 4/5: Increase drink level - Find a water source such as a lake or river to restore your character's thirst level.
 
2. Priority 4/5: Increase food level - Kill the nearby cow to obtain food, increasing your character's food level.

3. Priority 3/5: Make a sword - Interact with the table and craft a sword, either wood, stone, or iron, depending on the resources available, to efficiently kill the cow and defend against potential monsters.

Question: What are the requirements for the top sub-task? What should the player do first?
Answer:
Top sub-task: Increase drink level - Find a water source such as a lake or river to restore your character's thirst level.

Requirements for the top sub-task:
- Locate a water source in the environment.

What the player should do first:
- Explore the environment by moving in different directions (north, east, south, or west) to search for a water source. Once found, interact with the water source to drink and restore thirst levels.

Question: List top 5 actions the player should take and the requirement for each action. Choose ONLY from the list of all actions. Indicate their priority out of 5.
Answer:
1. Priority 1: Move North/East/South/West (choose the direction that leads to a potential water source)
   Requirement: Flat ground in the chosen direction

2. Priority 2: Collect Drink (to restore thirst)
   Requirement: Facing a water source such as a lake or river

3. Priority 3: Move North/East/South/West (to find more wood and food sources)
   Requirement: Flat ground in the chosen direction

4. Priority 4: Do (to collect wood or gather food from trees, cows, or any other source)
   Requirement: Facing the resource (tree or cow) and having the necessary tool (if required)

5. Priority 5: Eat Cow or Eat Plant (to restore food level)
   Requirement: For Eat Cow, a dead cow as a food source; for Eat Plant, a grown plant as a food source

Question: For each action in the list, are the requirements met?
Answer:
1. Priority 1: Move North/East/South/West (choose the direction that leads to a potential water source)
   Requirement: Flat ground in the chosen direction
   - The requirement is met, as there is flat ground (grass) in the front (west) direction.

2. Priority 2: Collect Drink (to restore thirst)
   Requirement: Facing a water source such as a lake or river
   - The requirement is NOT met, as the player is not currently facing a water source.

3. Priority 3: Move North/East/South/West (to find more wood and food sources)
   Requirement: Flat ground in the chosen direction
   - The requirement is met, as there is flat ground (grass) in the front (west) direction.

4. Priority 4: Do (to collect wood or gather food from trees, cows, or any other source)
   Requirement: Facing the resource (tree or cow) and having the necessary tool (if required)
   - The requirement is NOT met, as the player is facing grass and not a resource to collect.

5. Priority 5: Eat Cow or Eat Plant (to restore food level)
   Requirement: For Eat Cow, a dead cow as a food source; for Eat Plant, a grown plant as a food source
   - The requirement is NOT met, as the player does not have a dead cow or a grown plant in their inventory.

Question: Did the last player action succeed? Why?
Answer:
It is unclear if the last player action "do" succeeded because there is no specified interaction or context provided. The player is facing grass, and there are no indications of any changes to their surroundings or inventory after taking the "do" action. It is possible that the "do" action did not perform anything relevant due to the lack of context or interaction.

Question: Choose the best executable action from above.
Answer:
The best executable action from the list above is:

Priority 1: Move North/East/South/West (choose the direction that leads to a potential water source)
Requirement: Flat ground in the chosen direction

Since the player is currently facing flat ground (grass) to the west, this action can be executed. The player should choose a direction that leads them towards a potential water source.



============Step: 161, Cumulative Reward: 0.0============
You took action move_north.

You see:
- stone 6 steps to your north-east
- path 3 steps to your east
- tree 3 steps to your north-west
- table 6 steps to your south-east
- furnace 3 steps to your north
- cow 3 steps to your south

You face grass at your front (north).

Your status:
- health: 9/9
- food: 4/9
- drink: 2/9
- energy: 8/9

Your inventory:
- wood: 1
- coal: 2
- iron: 1
- wood_pickaxe: 1
- stone_pickaxe: 1
------------ QA ------------
Question: Action
Answer:
Move South

Question: List objects in the current observation. For each object, breifly answer what resource it provides and its requirement.
Answer:
1. Stone: Provides stone resource; requires a wood_pickaxe or better to collect.
2. Path: No resource provided; aids in navigation and movement.
3. Tree: Provides wood resource; can be collected barehanded or with tools.
4. Table: Crafting station; requires wood to craft.
5. Furnace: Smelting station; requires stone to craft.
6. Cow: Provides food when defeated; no specific requirements for defeating.
7. Grass: No significant resource provided; no requirements.
8. Iron, coal, and wood in your inventory: Used to craft and smelt tools and weapons; collect them from the environment with the appropriate tools.
9. Wood_pickaxe: Basic tool for collecting stone; requires wood to craft.
10. Stone_pickaxe: Improved tool for collecting iron; requires stone to craft.

Question: What was the last action taken by the player?
Answer:
The last action taken by the player was move_north.

Question: For each object in the list, are the requirements met for interaction?
Answer:
1. Stone: Yes, the requirements are met as you have both a wood_pickaxe and a stone_pickaxe in your inventory.
2. Path: Yes, as there are no specific requirements to interact with the path; just walk on it.
3. Tree: Yes, you can interact with the tree to collect wood either barehanded or using a tool from your inventory.
4. Table: Yes, the table is present in the environment, and you can use it to make tools or weapons with available resources in your inventory.
5. Furnace: Yes, the furnace is present in the environment, and you can use it to smelt iron with coal using your available resources.
6. Cow: Yes, you can interact with the cow either by attacking it barehanded or using one of your pickaxes as an improvised weapon. However, having a sword would be more effective.
7. Grass: Yes, you can interact with the grass by walking over it, but it does not provide any significant resources.
8. Iron, coal, and wood in your inventory: Yes, you can use these resources to craft and smelt tools and weapons as long as there is a table and furnace available.
9. Wood_pickaxe: Yes, you can use the wood_pickaxe in your inventory to collect stone.
10. Stone_pickaxe: Yes, you can use the stone_pickaxe to collect iron.

Question: List top 3 sub-tasks the player should follow. Indicate their priority out of 5.
Answer:
1. Priority 4/5 - Find drinkable water: As drink: 2/9, you are low on drink points, and finding a source of drinkable water such as a lake or river is essential to prevent any negative effects.

2. Priority 3/5 - Collect more wood and craft a weapon: Since you have a limited amount of wood and no weapon, collecting more wood from the nearby tree and crafting a weapon (such as a wood sword) at the available table is important for self-defense against possible threats like monsters.

3. Priority 2/5 - Defeat the cow for food: As food: 4/9, you may want to replenish food points soon. Engaging the nearby cow and defeating it will provide you with food resources to help maintain your health and energy levels.

Question: What are the requirements for the top sub-task? What should the player do first?
Answer:
The top sub-task is to find drinkable water.

Requirements:
- Locate a source of drinkable water such as a lake or river on the map.
- Move towards the identified water source using move actions (move_north, move_south, move_east, or move_west).

What should the player do first:
- Scan the surroundings in the player observation step to locate a water source (lake, river, or any other form of drinkable water) that may not be visible in the current observation. If a water source is spotted, the player should start moving toward it using appropriate move actions. If no water source is visible, continue exploring the area in different directions while keeping an eye out for any water sources.

Question: List top 5 actions the player should take and the requirement for each action. Choose ONLY from the list of all actions. Indicate their priority out of 5.
Answer:
Based on the current situation, here are the top 5 actions the player should take, along with their requirements and priority:

1. Priority 1: Move towards a water source to restore the drink level (Action: Move East, Move West, Move North or Move South). Requirements: Flat ground in the direction of the water source.

2. Priority 2: Drink water to restore the drink level (Action: Do). Requirements: Facing the water source and be adjacent to it.

3. Priority 3: Move towards the cow to retrieve food (Action: Move East, Move West, Move North or Move South). Requirements: Flat ground in the direction of the cow.

4. Priority 4: Interact with the cow to retrieve more food (Action: Do). Requirement: Facing the cow and being adjacent to it.

5. Priority 5: Move towards the table and/or furnace to craft necessary tools and items (Action: Move East, Move West, Move North or Move South). Requirements: Flat ground in the direction of the table and/or furnace.

Question: For each action in the list, are the requirements met?
Answer:
Based on the player's current situation:

1. Priority 1: Move towards a water source to restore the drink level (Action: Move East, Move West, Move North or Move South). Requirements: Flat ground in the direction of the water source. Requirements met? No, there's no mention of a water source in the player's observation steps.

2. Priority 2: Drink water to restore the drink level (Action: Do). Requirements: Facing the water source and be adjacent to it. Requirements met? No, as there is no mention of a water source in the player's observation steps.

3. Priority 3: Move towards the cow to retrieve food (Action: Move East, Move West, Move North, or Move South). Requirements: Flat ground in the direction of the cow. Requirements met? Yes, the player can move south to approach the cow as there's flat ground in the south direction.

4. Priority 4: Interact with the cow to retrieve more food (Action: Do). Requirement: Facing the cow and being adjacent to it. Requirements met? Not yet, the player needs to move towards the cow first (following Priority 3 action).

5. Priority 5: Move towards the table and/or furnace to craft necessary tools and items (Action: Move East, Move West, Move North or Move South). Requirements: Flat ground in the direction of the table and/or furnace. Requirements met? Yes, the player can move southeast to approach the table or north to approach the furnace as there's flat ground in those directions.

Question: Did the last player action succeed? Why?
Answer:
Yes, the last player action (move_north) succeeded because there was flat ground (grass) in the north direction, allowing the player to move there.

Question: Choose the best executable action from above.
Answer:
The best executable action from above is Priority 3: Move towards the cow to retrieve food (Action: Move South). The requirements for this action are met as there is flat ground in the direction of the cow.



============Step: 162, Cumulative Reward: 0.0============
You took action move_south.

You see:
- stone 7 steps to your north-east
- path 3 steps to your east
- tree 4 steps to your north-west
- table 5 steps to your south-east
- cow 2 steps to your south

You face grass at your front (south).

Your status:
- health: 9/9
- food: 4/9
- drink: 2/9
- energy: 8/9

Your inventory:
- wood: 1
- coal: 2
- iron: 1
- wood_pickaxe: 1
- stone_pickaxe: 1
------------ QA ------------
Question: Action
Answer:
Move East

Question: List objects in the current observation. For each object, breifly answer what resource it provides and its requirement.
Answer:
1. Stone: Provides stone, which can be used to craft tools and build structures; requires a wood pickaxe or higher to mine.
2. Path: Provides easier navigation with no resource; no requirement for interaction.
3. Tree: Provides wood, which is essential for crafting tools, weapons, and building structures; requires an axe, or it can be collected by hand.
4. Table: Provides a surface for crafting tools and weapons; requires wood to craft.
5. Furnace: Provides a heat source for smelting ores and cooking food; requires stone to craft.
6. Cow: Provides meat/food for restoring health and hunger; requires a weapon or can be killed using hands.
7. Grass: Provides no direct resources but can be used for navigation or clearing space; no requirement for interaction.

Question: What was the last action taken by the player?
Answer:
The last action taken by the player was move_south.

Question: For each object in the list, are the requirements met for interaction?
Answer:
1. Stone: Yes, the requirements are met as the player has a wood_pickaxe and a stone_pickaxe in their inventory.
2. Path: Yes, the requirements are met as there are no specific requirements for interaction with a path.
3. Tree: Yes, the requirements are met since wood can be collected by hand, though having an axe would be more efficient.
4. Table: Yes, the requirements are met as the player has wood in their inventory to craft or place a table.
5. Furnace: Yes, the requirements are met as the player has stone in their inventory to craft or place a furnace.
6. Cow: Yes, the requirements are met since the player can use their hands or a weapon (wood_pickaxe or stone_pickaxe) to kill the cow for meat.
7. Grass: Yes, the requirements are met as there are no specific requirements for interaction with grass.

Question: List top 3 sub-tasks the player should follow. Indicate their priority out of 5.
Answer:
1. Find or gather water to restore the player's thirst (drink: 2/9) - Priority: 5/5
2. Obtain food by killing the cow for meat or finding other food sources to restore the player's hunger (food: 4/9) - Priority: 4/5
3. Craft an axe to efficiently gather wood from the tree, which can be used for crafting more tools, weapons, and building structures - Priority: 3/5

Question: What are the requirements for the top sub-task? What should the player do first?
Answer:
The top sub-task is to find or gather water to restore the player's thirst. The requirements for this task include:

1. Approach a water source, such as a lake or river.
2. Interact with the water source to drink water.
3. Ensure the player's safety while drinking, as enemies could approach during this action. 

The player should first locate and head towards a water source. Since there isn't one visible in the current observations, the player should explore the surroundings to find a water source (e.g., moving east along the path or examining other directions).

Question: List top 5 actions the player should take and the requirement for each action. Choose ONLY from the list of all actions. Indicate their priority out of 5.
Answer:
1. Priority 1: Move East (Requirement: Flat ground right to the agent) - To explore the surroundings and possibly locate a water source.
2. Priority 2: Move North (Requirement: Flat ground above of the agent) - To further explore the surroundings in search of a water source, if not found after moving east.
3. Priority 3: Move West (Requirement: Flat ground left to the agent) - To continue exploring and finding a water source if not found after moving north.
4. Priority 4: Move South (Requirement: Flat ground below the agent) - To search in a different direction if the water source isn't found after moving west.
5. Priority 5: Do (Facing creature or material; have necessary tool) - Interaction with a water source to drink water or other useful resources once they are found. This will depend on the player's location and surroundings.

Question: For each action in the list, are the requirements met?
Answer:
1. Move East: Yes, the requirement of flat ground right to the agent is met. The player can move east, as there is a path 3 steps to the east.

2. Move North: Yes, the requirement of flat ground above of the agent is met. The player already moved north in Step 322.

3. Move West: Yes, the requirement of flat ground left to the agent is met. The player can move west, as there is a tree 4 steps to the north-west.

4. Move South: Yes, the requirement of flat-ground below the agent is met. The player can move south and already performed this action in Step 324.

5. Do: This action is dependent on the player's location and the object the player is facing. Currently, the player is facing grass to the south, and there is no specific material or creature to interact with. Therefore, the requirement for the "Do" action is not met at this moment.

Question: Did the last player action succeed? Why?
Answer:
Yes, the last player action (move_south) succeeded. The action was successful because there was flat ground below the agent, allowing them to move in the south direction.

Question: Choose the best executable action from above.
Answer:
The best executable action from the list above is to move east (Priority 1), as the requirement of flat ground right to the agent is met and it may lead to finding a water source or other resources.



============Step: 163, Cumulative Reward: 0.0============
You took action move_east.

You see:
- stone 5 steps to your north-east
- path 2 steps to your east
- tree 5 steps to your north-west
- table 4 steps to your south-east
- cow 2 steps to your south

You face grass at your front (east).

Your status:
- health: 9/9
- food: 4/9
- drink: 2/9
- energy: 8/9

Your inventory:
- wood: 1
- coal: 2
- iron: 1
- wood_pickaxe: 1
- stone_pickaxe: 1
------------ QA ------------
Question: Action
Answer:
Move South

Question: List objects in the current observation. For each object, breifly answer what resource it provides and its requirement.
Answer:
1. Stone: Provides stone resource, requires a wood or better pickaxe to mine.
2. Path: No direct resource, allows easier movement across the terrain.
3. Tree: Provides wood resource, requires a wood or better pickaxe to chop down.
4. Table: No direct resource, allows crafting tools and weapons, requires wood to craft.
5. Cow: Provides food (meat) resource when defeated, no specific requirement to chase or defeat.
6. Grass: No direct resource, indicates a ground surface to walk on.

Question: What was the last action taken by the player?
Answer:
The last action taken by the player was move_east.

Question: For each object in the list, are the requirements met for interaction?
Answer:
1. Stone: Yes, the requirements are met for interaction as you possess a wood_pickaxe and a stone_pickaxe in your inventory.
2. Path: Yes, the requirements are met for walking on the path as there are no specific requirements.
3. Tree: Yes, the requirements are met for interaction as you possess a wood_pickaxe and a stone_pickaxe in your inventory.
4. Table: Yes, the requirements are met for crafting tools and weapons as you have wood and pickaxes in your inventory.
5. Cow: Yes, the requirements are met for interaction as there are no specific requirements to chase or defeat the cow.
6. Grass: Yes, the requirements are met for walking on grass as you are already able to move around the terrain.

Question: List top 3 sub-tasks the player should follow. Indicate their priority out of 5.
Answer:
1. Increase drink level: Priority 5/5 - As the player's drink level is 2/9, finding a water source and drinking water should be a top priority.

2. Increase food level: Priority 4/5 - As the player's food level is 4/9, either defeating the cow to get meat or finding other food sources to eat and improve the food level is essential.

3. Craft Iron Pickaxe or Sword: Priority 3/5 - With iron, coal, and wood in the inventory and a table nearby, the player can prioritize crafting an iron pickaxe or iron sword to gather more advanced resources or defend against monsters more effectively. The player must find a furnace or place one to smelt the iron.

Question: What are the requirements for the top sub-task? What should the player do first?
Answer:
Top sub-task: Increase drink level

Requirements:
1. Find a water source (e.g., lake, river) and be close to it.

What should the player do first:
1. The player should search for a nearby water source by exploring the surroundings. Once the water source is found, the player should move closer to it to be able to drink and increase their drink level.

Question: List top 5 actions the player should take and the requirement for each action. Choose ONLY from the list of all actions. Indicate their priority out of 5.
Answer:
1. Priority 1: Move towards the water source (Action: Move West, Move East, Move North, or Move South)
Requirement: Identify the direction the water source is located in, and make sure there is flat ground in that direction.

2. Priority 2: Drink water to increase the drink level (Action: Do)
Requirement: Facing the water source and be close enough to it.

3. Priority 3: Move closer to the cow (Action: Move West, Move East, Move North, or Move South)
Requirement: Identify the direction of the cow and ensure there is flat ground in that direction.

4. Priority 4: Hunt the cow to get meat (Action: Do)
Requirement: Face the cow, and have a wood, stone, or iron sword in your inventory to kill the cow.

5. Priority 5: Increase food level by eating meat (Action: Eat Cow)
Requirement: Have the cow's meat in your inventory after killing the cow.

Question: For each action in the list, are the requirements met?
Answer:
1. Move towards the water source (Action: Move West, Move East, Move North, or Move South)

The water source's direction is not mentioned in the observations. So the requirement for choosing the correct direction is not met currently.

2. Drink water to increase the drink level (Action: Do)

The water source is not mentioned in the observations; hence the requirement of facing and being close enough to the water source is not met.

3. Move closer to the cow (Action: Move West, Move East, Move North, or Move South)

Yes, the cow is 2 steps to the south of the player, and the direction is known. The requirement for this action is met.

4. Hunt the cow to get meat (Action: Do)

The player needs to face the cow and have a wood, stone, or iron sword in their inventory. The player does not currently have a sword in their inventory, so the requirement for this action is not met.

5. Increase food level by eating meat (Action: Eat Cow)

The player has not hunted a cow yet, so they do not have the cow's meat in their inventory. The requirement for this action is not met.

Question: Did the last player action succeed? Why?
Answer:
Yes, the last player action (move_east) succeeded because there was flat ground (grass) in front of the player, allowing them to move east without any obstacles or restrictions.

Question: Choose the best executable action from above.
Answer:
The best executable action from the list above is:

3. Move closer to the cow (Action: Move South)

The requirement for this action is met, as the cow is 2 steps to the south of the player and the direction is known.



============Step: 164, Cumulative Reward: 0.0============
You took action move_south.

You see:
- grass 1 steps to your west
- stone 5 steps to your south-east
- path 3 steps to your north-east
- tree 5 steps to your south-east
- table 3 steps to your south-east

You face cow at your front (south).

Your status:
- health: 9/9
- food: 4/9
- drink: 2/9
- energy: 8/9

Your inventory:
- wood: 1
- coal: 2
- iron: 1
- wood_pickaxe: 1
- stone_pickaxe: 1
------------ QA ------------
Question: Action
Answer:
Move North

Question: List objects in the current observation. For each object, breifly answer what resource it provides and its requirement.
Answer:
1. Stone: Provides stone resource, requires a wood or higher-tier pickaxe to collect.
2. Path: Provides nothing but faster movement, no requirement.
3. Tree: Provides wood resource, requires a wood or higher-tier axe to collect (not available in your inventory).
4. Table: Provides a crafting area for making tools and weapons, no requirement.
5. Grass: Provides nothing, no requirement.
6. Cow: Provides food (eat cow) once defeated, no requirement.

Question: What was the last action taken by the player?
Answer:
The last action taken by the player was move_south.

Question: For each object in the list, are the requirements met for interaction?
Answer:
1. Stone: Yes, the requirements are met for interaction as you have a wood_pickaxe and stone_pickaxe in your inventory.
2. Path: Yes, the requirements are met for interaction as there are no specific requirements for using the path.
3. Tree: No, the requirements are not met for interaction because an axe is required to collect wood from the tree and it's not available in your inventory.
4. Table: Yes, the requirements are met for interaction as you can craft items with the table and you have necessary resources in your inventory.
5. Grass: Yes, the requirements are met for interaction as there are no specific requirements for interacting with grass.
6. Cow: Yes, the requirements are met for interaction as you don't need any specific tools to defeat the cow and get food.

Question: List top 3 sub-tasks the player should follow. Indicate their priority out of 5.
Answer:
1. Increase drink level (Priority: 5/5): The player's drink level is at 2/9, so finding a water source and drinking from it should be a high priority to avoid dehydration.

2. Defeat and eat cow (Priority: 4/5): The player's food level is at 4/9, which indicates a need for food. Defeating the cow and consuming it will help replenish the food level.

3. Craft better tools and weapons (Priority: 3/5): The player has some basic resources in their inventory. They should be encouraged to collect more resources and utilize the table to craft better tools and weapons, which will help in survival and efficient resource-gathering.

Question: What are the requirements for the top sub-task? What should the player do first?
Answer:
Top Sub-Task: Increase drink level (Priority 5/5)

Requirements:
- Locate a water source, like a river or lake
- Approach the water source
- Interact with the water source to drink from it

The player should first find a water source. In the current observations, there is no water source visible, so the player should explore their surroundings by moving in various directions until they discover a water source, such as a river or lake. Once found, the player should approach the water source and interact with it to drink and replenish their drink level.

Question: List top 5 actions the player should take and the requirement for each action. Choose ONLY from the list of all actions. Indicate their priority out of 5.
Answer:
1. Action: Move North (Priority: 4/5)
   Requirement: Flat ground above the agent.
   Reason: To explore surroundings in search of a water source for drinking and increase the drink level.

2. Action: Move West (Priority: 4/5)
   Requirement: Flat ground left of the agent.
   Reason: To explore surroundings in search of a water source for drinking, and increase the drink level.

3. Action: Move East (Priority: 4/5)
   Requirement: Flat ground right of the agent.
   Reason: To explore surroundings in search of a water source for drinking and increase the drink level.

4. Action: Move South (Priority: 4/5)
   Requirement: Flat ground below the agent.
   Reason: To explore surroundings in search of a water source for drinking, and increase the drink level.

5. Action: Do (Priority: 3/5)
   Requirement: Facing creature or material, and have the necessary tool.
   Reason: To interact with creatures or materials based on the situation, such as attacking monsters, mining resources, or interacting with a water source once found.

Question: For each action in the list, are the requirements met?
Answer:
1. Action: Move North
   Requirement: Flat ground above the agent.
   Requirements Met: Yes (Player Observation Step 328: You took action move_south, which means the agent moved from the North)

2. Action: Move West
   Requirement: Flat ground left of the agent.
   Requirements Met: Yes (Player Observation Step 328: You see grass 1 step to your west)

3. Action: Move East
   Requirement: Flat ground right of the agent.
   Requirements Met: Yes (Player Observation Step 326: You took action move_east, which means the agent moved from the East)

4. Action: Move South
   Requirement: Flat ground below the agent.
   Requirements Met: Yes (Player Observation Step 328: You face cow at your front (south), which means there is a flat ground below the agent)

5. Action: Do
   Requirement: Facing creature or material, and have the necessary tool.
   Requirements Met: Partially (Player Observation Step 328: You face the cow at your front (south), but the necessary tool to interact with the cow, like a sword or an axe, is not yet in the inventory)

Question: Did the last player action succeed? Why?
Answer:
Yes, the last player action (move_south) succeeded because the player was able to change their position from step 326 to step 328, moving closer to the cow as described in the observation. There were no obstacles or restrictions preventing the player from moving south, which is why the action succeeded.

Question: Choose the best executable action from above.
Answer:
The best executable action from the list above is Move North, as the requirement of flat ground above the agent is met, and it allows the player to explore the surroundings in search of a water source for drinking.



============Step: 165, Cumulative Reward: 0.0============
You took action move_north.

You see:
- stone 5 steps to your north-east
- path 2 steps to your east
- tree 5 steps to your north-west
- table 4 steps to your south-east
- cow 2 steps to your south

You face grass at your front (north).

Your status:
- health: 9/9
- food: 4/9
- drink: 2/9
- energy: 8/9

Your inventory:
- wood: 1
- coal: 2
- iron: 1
- wood_pickaxe: 1
- stone_pickaxe: 1
------------ QA ------------
Question: Action
Answer:
Move East

Question: List objects in the current observation. For each object, breifly answer what resource it provides and its requirement.
Answer:
1. Stone: Provides stone resource, requires a wood pickaxe or better to collect.
2. Path: No resource. No requirement.
3. Tree: Provides wood resource, which can be collected by hand, but an axe is more efficient.
4. Table: No resources provided. No requirement except to be placed with wood in inventory.
5. Cow: Provides food (meat) when killed, may require any weapon to kill efficiently, but hand-to-hand combat is also possible.
6. Grass: No resources provided. No requirement.

Question: What was the last action taken by the player?
Answer:
The last action taken by the player was move_north.

Question: For each object in the list, are the requirements met for interaction?
Answer:
1. Stone: Yes, the requirements are met as you have a wood_pickaxe and a stone_pickaxe in your inventory.
2. Path: Yes, there are no requirements for interaction.
3. Tree: Yes, the requirements are met as you can collect wood by hand, although using an axe would be more efficient.
4. Table: Yes, the requirements are met as it has already been placed and can be interacted with.
5. Cow: Yes, the requirements are met since you can attack it by hand or use a crafted weapon (which is not available in your inventory currently).
6. Grass: Yes, there are no requirements for interaction.

Question: List top 3 sub-tasks the player should follow. Indicate their priority out of 5.
Answer:
1. Priority 4/5: Find and drink water - Your drink level is at 2/9, so finding a water source and drinking from it should be a priority to avoid thirst.

2. Priority 3/5: Hunt the cow for food - Your food level is at 4/9, and hunting the nearby cow provides meat, which can be used as food. This task is not as urgent as finding water, but it's still essential to maintain your food level.

3. Priority 2/5: Collect more wood from the tree - You have only 1 wood in your inventory, and collecting more wood could be helpful for crafting more tools, weapons, or building shelters. This task has a lower priority than finding water and hunting the cow since it doesn't directly affect your survival.

Question: What are the requirements for the top sub-task? What should the player do first?
Answer:
The top sub-task is finding and drinking water.

Requirements: 
1. Locate a water source, such as a lake or river.

What should the player do first:
1. Explore the area and search for a water source. The player can move in different directions (west, east, north, and south) to discover new terrains and find a water source. Once the water source is found, the player should approach it and drink from it to restore their thirst level.

Question: List top 5 actions the player should take and the requirement for each action. Choose ONLY from the list of all actions. Indicate their priority out of 5.
Answer:
1. Priority 1 - Move East: Flat ground right to the agent.
   Requirement: Look for a water source in the east direction.

2. Priority 2 - Move North: Flat ground above of the agent.
   Requirement: If moving east doesn't reveal a water source, change the direction to north and continue the search.

3. Priority 3 - Move West: Flat ground left to the agent.
   Requirement: If moving north doesn't reveal a water source, change the direction to west and continue the search.

4. Priority 4 - Move South: Flat ground below the agent.
   Requirement: If moving west doesn't reveal a water source, change the direction to south and continue the search.

5. Priority 5 - Drink: At the water source (lake or river).
   Requirement: Successfully find a water source, approach it, and drink from it to restore the player's thirst level.

Question: For each action in the list, are the requirements met?
Answer:
1. Move East: Yes, the requirement is met as there is flat ground (path) 2 steps to the east of the player.

2. Move North: Yes, the requirement is met as there is flat ground (grass) right in front of the player (north).

3. Move West: Yes, the requirement is met as there is flat ground (grass) 1 step to the west of the player.

4. Move South: Yes, the requirement is met as there is flat ground (grass) and a cow 2 steps to the south of the player.

5. Drink: No, the requirement is not met as there is currently no visible water source (lake or river) in the player's surroundings. The player needs to explore the area by moving in various directions to find a water source.

Question: Did the last player action succeed? Why?
Answer:
Yes, the last player action (move_north) succeeded because there was flat ground (grass) in front of the player (north) that allowed them to move without any obstruction.

Question: Choose the best executable action from above.
Answer:
The best executable action from above is to "Move East" because there is flat ground (path) 2 steps to the east of the player, and it allows the player to explore and search for a water source.



============Step: 166, Cumulative Reward: 0.0============
You took action move_east.

You see:
- grass 1 steps to your west
- stone 4 steps to your north-east
- tree 6 steps to your north-west
- table 3 steps to your south-east
- cow 3 steps to your south-west

You face path at your front (east).

Your status:
- health: 9/9
- food: 3/9
- drink: 2/9
- energy: 8/9

Your inventory:
- wood: 1
- coal: 2
- iron: 1
- wood_pickaxe: 1
- stone_pickaxe: 1
------------ QA ------------
Question: Action
Answer:
Move West

Question: List objects in the current observation. For each object, breifly answer what resource it provides and its requirement.
Answer:
1. Stone: Provides stone resource; requires a wood_pickaxe or better tool.
2. Path: Provides a path to follow for easy navigation; no requirements.
3. Tree: Provides wood resource; requires a wood_pickaxe or better tool.
4. Table: Allows crafting of tools and weapons; requires collected wood to craft.
5. Cow: Provides food (meat) when hunted; requires a weapon (wood_sword, stone_sword, or better) to kill efficiently.
6. Grass: A basic terrain, no resources provided; no requirements.
7. Iron: Acquired from the player's inventory; provides material to craft better tools and weapons; requires a stone_pickaxe or better to collect from an iron ore (not visible in the current observation).

Question: What was the last action taken by the player?
Answer:
The last action taken by the player was move_east.

Question: For each object in the list, are the requirements met for interaction?
Answer:
1. Stone: Yes, the requirements are met as the player has both a wood_pickaxe and a stone_pickaxe in their inventory.
2. Path: Yes, the requirements are met as there are no requirements for interacting with the path.
3. Tree: Yes, the requirements are met since the player has a wood_pickaxe in their inventory.
4. Table: The table is already placed, so interaction requirements depend on the crafting the player wants to perform. The player has wood and iron in their inventory, which is sufficient for some interactions (e.g., crafting a wood sword or an iron sword).
5. Cow: The player does not currently have a weapon in their inventory to efficiently hunt the cow. They would need a wood_sword, stone_sword, or better.
6. Grass: Yes, the requirements are met as there are no requirements for interacting with the grass.
7. Iron: The player already has iron in their inventory. If they encounter iron ore, they can mine it efficiently since they possess a stone_pickaxe.

Question: List top 3 sub-tasks the player should follow. Indicate their priority out of 5.
Answer:
1. Find water and drink to restore thirst. Priority: 5/5 (as drink level is 2/9)
2. Hunt the cow or find another food source to increase food level. Priority: 4/5 (as food level is 3/9)
3. Craft a weapon, such as a wood or iron sword, using the table to prepare for potential combat or hunting. Priority: 3/5 (as the player doesn't currently have a weapon in their inventory)

Question: What are the requirements for the top sub-task? What should the player do first?
Answer:
Requirements for the top sub-task (Find water and drink to restore thirst):
1. Locate a source of water such as a lake or river
2. Approach the water source
3. Interact with the water source to drink

What should the player do first:
The first thing the player should do is search for a nearby water source in their surroundings. If not immediately visible, they should explore the environment while keeping their eyes open for a water source.

Question: List top 5 actions the player should take and the requirement for each action. Choose ONLY from the list of all actions. Indicate their priority out of 5.
Answer:
1. Priority 1 - Move West, East, North, or South: Look for a water source while exploring (Requirement: Flat ground in the respective direction)
2. Priority 2 - Drink: Restore thirst by drinking from the water source (Requirement: Facing the water source when found)
3. Priority 3 - Move West, East, North, or South: Approach the cow spotted in the surroundings (Requirement: Flat ground in the respective direction)
4. Priority 4 - Do: Interact with the cow to obtain food to restore hunger (Requirement: Facing the cow; no specific tool needed)
5. Priority 5 - Sleep: Restore energy by sleeping if energy decreases further (Requirement: Energy level is below maximum)

Question: For each action in the list, are the requirements met?
Answer:
1. Priority 1 - Move West, East, North, or South: Yes, the requirements are met. The player can move in any direction with flat ground.
2. Priority 2 - Drink: No, the requirements are not met yet. The player has not found a water source to drink from.
3. Priority 3 - Move West, East, North, or South: Yes, the requirements are met. The player can move in any direction with flat ground to approach the cow.
4. Priority 4 - Do: No, the requirements are not met yet. The player needs to approach the cow first and face it before interacting to obtain food.
5. Priority 5 - Sleep: Yes, the requirements are met. The player's energy level is at 8/9, which means it's below maximum, and they can sleep to restore energy.

Question: Did the last player action succeed? Why?
Answer:
Yes, the last player action (move_east) succeeded because there was a path in front of the player to the east, allowing the player to move in that direction.

Question: Choose the best executable action from above.
Answer:
The best executable action from the list is Priority 1 - Move West, East, North, or South: Look for a water source while exploring since the player needs to find a water source to drink and restore thirst. The requirement of flat ground in the respective direction is met.



============Step: 167, Cumulative Reward: 0.0============
You took action move_west.

You see:
- stone 5 steps to your north-east
- path 2 steps to your east
- tree 5 steps to your north-west
- table 4 steps to your south-east
- cow 1 steps to your south

You face grass at your front (west).

Your status:
- health: 9/9
- food: 3/9
- drink: 2/9
- energy: 8/9

Your inventory:
- wood: 1
- coal: 2
- iron: 1
- wood_pickaxe: 1
- stone_pickaxe: 1
------------ QA ------------
Question: Action
Answer:
Move North

Question: List objects in the current observation. For each object, breifly answer what resource it provides and its requirement.
Answer:
1. Stone: Provides stone resource; requires a wood_pickaxe or better to collect.
2. Path: Provides no resource; can be walked on without any requirement.
3. Tree: Provides wood resource; requires a wood_pickaxe or better to collect.
4. Table: Provides crafting abilities; requires wood to build.
5. Cow: Provides food when defeated; requires no specific tool, but a sword is recommended for easier defeat.
6. Grass: Provides no resource; can be quickly converted to path by walking over it.

Question: What was the last action taken by the player?
Answer:
The last action taken by the player was "move_west."

Question: For each object in the list, are the requirements met for interaction?
Answer:
1. Stone: Yes, the requirements are met for interaction, as the player has a wood_pickaxe and a stone_pickaxe in their inventory.
2. Path: Yes, the requirements are met for interaction; the player can walk on the path without any need for specific tools or resources.
3. Tree: Yes, the requirements are met for interaction, as the player has a wood_pickaxe in their inventory.
4. Table: The table is not nearby (4 steps to the south-east), but once in proximity, the player can interact with it, as they have wood in their inventory to craft items.
5. Cow: Yes, the requirements are met for interaction; the player can defeat the cow without a specific tool or weapon, although using a sword would make it easier.
6. Grass: Yes, the requirements are met for interaction; the player can walk over the grass to convert it into a path without any need for specific tools or resources.

Question: List top 3 sub-tasks the player should follow. Indicate their priority out of 5.
Answer:
1. Find water and drink to replenish thirst. Priority: 5/5
2. Acquire more food by defeating the cow or finding other food sources. Priority: 4/5
3. Locate a safe shelter or build one to protect the player from any potential threats. Priority: 4/5

Question: What are the requirements for the top sub-task? What should the player do first?
Answer:
Requirements for the top sub-task (Find water and drink to replenish thirst): Locate a water source such as a lake or river and interact with it to drink water and replenish the thirst level.

What should the player do first: The player should explore their surroundings to locate a water source nearby. Moving in different directions, such as north, east, south, or west, may help them find water faster. Once the water source is found, the player should interact with it to drink water and replenish their thirst level.

Question: List top 5 actions the player should take and the requirement for each action. Choose ONLY from the list of all actions. Indicate their priority out of 5.
Answer:
1. Priority 1: Move North, East, South, or West (Requirements: Flat ground in the chosen direction) and The player should move in different directions to find a water source for drinking.

2. Priority 2: Drink (Requirements: Facing a water source like a lake or river) and The player should drink from the water source once found to replenish their thirst level.

3. Priority 3: Move towards the cow (Requirements: Flat ground in the chosen direction) - The player should move in the direction of the cow to hunt it for food.

4. Priority 4: Do (Requirements: Facing creature or material; have necessary tool) and The player should face the cow and perform the 'Do' action to hunt it for food.

5. Priority 5: Eat Cow (Requirements: No specific requirements) and The player should eat the hunted cow to restore their food level.

Question: For each action in the list, are the requirements met?
Answer:
1. Priority 1: Move North, East, South, or West (Requirements: Flat ground in the chosen direction) and Yes, the requirements are met for all directions since the player faces flat ground in each direction.

2. Priority 2: Drink (Requirements: Facing a water source like a lake or river) and No, the requirements are not met as the player has not found a water source for drinking yet.

3. Priority 3: Move towards the cow (Requirements: Flat ground in the chosen direction) - Yes, the requirements are met as there is a cow one step to the player's south, and flat ground is present.

4. Priority 4: Do (Requirements: Facing creature or material; have necessary tool) and No, the requirements are not met, as the player is not yet facing the cow.

5. Priority 5: Eat Cow (Requirements: No specific requirements) and No, the requirements are not complete as the player hasn't hunted the cow and obtained its meat yet.

Question: Did the last player action succeed? Why?
Answer:
Yes, the last player action succeeded. The player was able to move west because there was grass at their front (west) and no obstacles blocking their movement.

Question: Choose the best executable action from above.
Answer:
The best executable action from the list above is Priority 1: Move North, East, South, or West (Requirements: Flat ground in the chosen direction). In this case, the player should move south towards the cow since it's one step to the player's south and there is flat ground available.



============Step: 168, Cumulative Reward: 0.0============
You took action move_north.

You see:
- stone 4 steps to your east
- path 2 steps to your east
- tree 4 steps to your north-west
- table 5 steps to your south-east
- furnace 4 steps to your north-west
- cow 3 steps to your south-west

You face grass at your front (north).

Your status:
- health: 9/9
- food: 3/9
- drink: 2/9
- energy: 7/9

Your inventory:
- wood: 1
- coal: 2
- iron: 1
- wood_pickaxe: 1
- stone_pickaxe: 1
------------ QA ------------
Question: Action
Answer:
Move East

Question: List objects in the current observation. For each object, breifly answer what resource it provides and its requirement.
Answer:
1. Stone: Provides stone resource; requires a wood_pickaxe or better to collect.
2. Path: Provides easier movement; no requirement.
3. Tree: Provides wood resource; requires a wood_pickaxe or better to collect.
4. Table: Allows crafting of tools and weapons; requires wood to place.
5. Furnace: Allows smelting of iron; requires stone to place and coal as fuel.
6. Cow: Provides food when defeated; no specific tool requirement.
7. Grass: No resource provided; no requirement.

Question: What was the last action taken by the player?
Answer:
The last action taken by the player was move_north.

Question: For each object in the list, are the requirements met for interaction?
Answer:
1. Stone: Yes, the requirements are met for interaction as you have a wood_pickaxe and a stone_pickaxe in your inventory.
2. Path: Yes, the requirements are met for interaction as there are no specific requirements for moving on the path.
3. Tree: Yes, the requirements are met for interaction as you have a wood_pickaxe in your inventory.
4. Table: Yes, the requirements are met for interaction as you have wood in your inventory, and the table is placed in the game.
5. Furnace: Yes, the requirements are met for interaction as you have coal and iron in your inventory for smelting, and the furnace is placed in the game.
6. Cow: Yes, the requirements are met for interaction as there are no specific tool requirements for interacting with a cow.
7. Grass: Yes, the requirements are met for interaction as there are no specific requirements for interacting with grass.

Question: List top 3 sub-tasks the player should follow. Indicate their priority out of 5.
Answer:
1. Drink water (Priority: 5/5): With the player's drink level at 2/9, it's crucial to find a water source to restore the thirst level and avoid negative effects on health and energy.

2. Eat food (Priority: 4/5): The player's food level is at 3/9, which indicates a need to restore it soon. Defeating the cow nearby and consuming it would help increase the food level and maintain the player's health.

3. Collect more wood (Priority: 3/5): Wood is essential for crafting tools, weapons, and building materials. The player currently has only one piece of wood in their inventory. Chopping down the nearby tree would provide more wood for the player's needs.

Question: What are the requirements for the top sub-task? What should the player do first?
Answer:
Requirements for the top sub-task (drink water):
1. Locate a water source such as a lake or river.
2. Move to the water source.

The player should first locate a water source nearby. Since the observation does not mention any water source, the player should explore the surroundings by moving to the east, west, north, or south, searching for and moving towards a lake or river to drink and replenish their thirst level.

Question: List top 5 actions the player should take and the requirement for each action. Choose ONLY from the list of all actions. Indicate their priority out of 5.
Answer:
1. Priority: 5 - Move East (Flat ground right to the agent): This action will help you explore your surroundings and potentially locate a water source. Requirement: flat ground to the east.

2. Priority: 4 - Move West (Flat ground left to the agent): This action will also help you explore your surroundings and potentially locate a water source. Requirement: flat ground to the west.

3. Priority: 3 - Move North (Flat ground above of the agent): This action will help you explore your surroundings and find a water source. Requirement: flat ground to the north.

4. Priority: 2 - Move South (Flat ground below the agent): This action will help you explore your surroundings and locate a water source. Requirement: flat ground to the south.

5. Priority: 1 - Do (Facing creature or material; have necessary tool): This action will allow you to interact with the environment, such as collecting resources or combatting monsters. Depending on the situation and the objects present, requirements may vary. For example, if you're facing a cow, you'll need a tool, such as a sword or a pickaxe to defeat it to get food.

Question: For each action in the list, are the requirements met?
Answer:
1. Priority: 5 - Move East (Flat ground right to the agent): Yes, the requirements are met as there is flat ground to the east.

2. Priority: 4 - Move West (Flat ground left to the agent): Yes, the requirements are met as there is flat ground to the west.

3. Priority: 3 - Move North (Flat ground above of the agent): Yes, the requirements are met as there is flat ground to the north.

4. Priority: 2 - Move South (Flat ground below the agent): Yes, the requirements are met as there is flat ground to the south.

5. Priority: 1 - Do (Facing creature or material; have necessary tool): The current observation shows that you are facing grass, which does not require any particular action or tool. The requirements are not met for any meaningful action at the moment.

Question: Did the last player action succeed? Why?
Answer:
Yes, the last player action (move_north) succeeded because there was flat ground (grass) in the north direction, allowing the player to move in that direction.

Question: Choose the best executable action from above.
Answer:
The best executable action from above is Priority 5 - Move East (Flat ground right to the agent). This action will help you explore your surroundings and potentially locate a water source. The requirement of flat ground to the east is met.



============Step: 169, Cumulative Reward: 0.0============
You took action move_east.

You see:
- grass 1 steps to your west
- stone 3 steps to your east
- tree 5 steps to your north-west
- table 4 steps to your south-east
- furnace 5 steps to your north-west
- cow 5 steps to your south-west

You face path at your front (east).

Your status:
- health: 9/9
- food: 3/9
- drink: 2/9
- energy: 7/9

Your inventory:
- wood: 1
- coal: 2
- iron: 1
- wood_pickaxe: 1
- stone_pickaxe: 1
------------ QA ------------
Question: Action
Answer:
Do

Question: List objects in the current observation. For each object, breifly answer what resource it provides and its requirement.
Answer:
1. Grass: Provides no resources, no requirement.
2. Stone: Provides stone resource, requires a wood_pickaxe or better to collect.
3. Tree: Provides wood resource, requires an empty hand or an axe to collect.
4. Path: Provides no resources, no requirement.
5. Table: Craft and place using wood. Required for crafting higher-level tools and weapons.
6. Furnace: Craft and place using stone. Required for smelting iron.
7. Cow: Provides food (meat) when killed, no requirement to interact with it.
8. Player's Inventory: Holds the collected resources and crafted items, updated as the player interacts with objects and crafts.

Question: What was the last action taken by the player?
Answer:
The last action taken by the player was move_east.

Question: For each object in the list, are the requirements met for interaction?
Answer:
1. Grass: No requirements for interaction; requirements met.
2. Stone: Requires a wood_pickaxe or better to collect; requirements met (player has wood_pickaxe and stone_pickaxe).
3. Tree: Requires an empty hand or an axe to collect; requirements met (player can use an empty hand).
4. Path: No requirements for interaction; requirements met.
5. Table: Crafted and placed using wood, required for crafting higher-level tools and weapons; requirements met (player has wood in inventory).
6. Furnace: Crafted and placed using stone, required for smelting iron; requirements met (player has stone_pickaxe to collect more stone if needed).
7. Cow: No requirement to interact with it; requirements met.
8. Player's Inventory: Holds the collected resources and crafted items; requirements met as the player interacts with objects and crafts.

Question: List top 3 sub-tasks the player should follow. Indicate their priority out of 5.
Answer:
1. Priority 4/5: Drink water (from lake or collect Drink) to restore the player's thirst (current thirst level: 2/9).
2. Priority 5/5: Collect food (kill cow or grow fruits) to restore the player's food level (current food level: 3/9).
3. Priority 3/5: Collect more resources such as wood and stone to craft better tools, weapons, and build shelter.

Question: What are the requirements for the top sub-task? What should the player do first?
Answer:
The top sub-task is to drink water to restore the player's thirst.

Requirements: Find a water source like a lake, pond, or a river nearby.

What should the player do first: The player should explore the environment and locate a water source. Once found, the player can drink from it to restore their thirst.

Question: List top 5 actions the player should take and the requirement for each action. Choose ONLY from the list of all actions. Indicate their priority out of 5.
Answer:
1. Priority 1: Move (Direction: West, East, North, or South) - Requirement: Flat ground in the chosen direction. The player should explore the environment and locate a water source.

2. Priority 2: Drink - Requirement: Be adjacent to a water source like a lake, pond, or river. Once the player is next to a water source, they should drink to restore their thirst.

3. Priority 3: Move (Direction: West, East, North, or South) - Requirement: Flat ground in the chosen direction. The player should continue exploring the environment and locate a tree for collecting wood, a food source, or a shelter to recover energy, given the low food and energy levels.

4. Priority 4: Do (Collect wood, chase cow, or other action) - Requirement: Facing the desired resource/material; have the necessary tool (if applicable). Engage in an action to collect wood, food, or any other necessary resources, depending on what the player has located.

5. Priority 5: Sleep - Requirement: Energy level is below maximum. The player should find a safe place (like a shelter or cave) and sleep to recover energy when it is low.

Question: For each action in the list, are the requirements met?
Answer:
1. Priority 1: Move (Direction: West, East, North, or South) - Yes, the requirements are met. The player can move to explore and locate a water source. The player can move in any direction where there is flat ground.

2. Priority 2: Drink - No, the requirements are not met, as the player hasn't found a water source yet.

3. Priority 3: Move (Direction: West, East, North, or South) - Yes, the requirements are met. The player can continue exploring the environment and locate more resources such as wood or food, or find a shelter to recover energy.

4. Priority 4: Do (Collect wood, chase cow, or other action) - Partially met, the player has a wood_pickaxe and stone_pickaxe in the inventory, which can help collect wood or stone. However, the player is not yet close to a tree to collect more wood, and although they see a cow, it is 5 steps away.

5. Priority 5: Sleep - Yes, the requirements are met. The player's energy level is below maximum, so they can sleep. However, the player must find a safe place to sleep first, like a shelter or cave.

Question: Did the last player action succeed? Why?
Answer:
Yes, the last player action (move_east) succeeded because there was a flat path in the east direction, allowing the player to move there without any obstacles.

Question: Choose the best executable action from above.
Answer:
The best executable action from the list above is Priority 1: Move (Direction: West, East, North, or South). The player can move to explore the environment and locate a water source or other resources. This action can be executed since there is flat ground for movement in various directions.



============Step: 170, Cumulative Reward: 0.0============
You took action do.

You see:
- grass 1 steps to your west
- stone 3 steps to your east
- tree 5 steps to your north-west
- table 4 steps to your south-east
- furnace 5 steps to your north-west
- cow 6 steps to your south-west

You face path at your front (east).

Your status:
- health: 9/9
- food: 3/9
- drink: 2/9
- energy: 7/9

Your inventory:
- wood: 1
- coal: 2
- iron: 1
- wood_pickaxe: 1
- stone_pickaxe: 1
------------ QA ------------
Question: Action
Answer:
Move East

Question: List objects in the current observation. For each object, breifly answer what resource it provides and its requirement.
Answer:
1. Grass: It provides no specific resources and has no requirement.
2. Stone: It provides the stone resource and requires a pickaxe (wood or better) to mine.
3. Tree: It provides wood resource and requires an axe or hands (though it's slower) to chop it down.
4. Table: It's a crafting station that enables crafting tools and weapons. It requires wood to make and place.
5. Furnace: It's a smelting station that allows smelting iron and other ores. It requires stone to make and place.
6. Cow: It provides food (meat) when killed and has no specific requirement.

Question: What was the last action taken by the player?
Answer:
The last action taken by the player was "do".

Question: For each object in the list, are the requirements met for interaction?
Answer:
1. Grass: No requirements, so they are met. You can interact with it.
2. Stone: Requires a pickaxe (wood, stone, or iron) to mine. You have both wood and stone pickaxes in your inventory, so requirements are met.
3. Tree: Requires an axe or hands to chop it down. You don't have an axe in your inventory, but you can use your hands, so you can still interact with it.
4. Table: Requires wood to make and place. You have 1 wood in your inventory, so requirements are met if you want to make another table.
5. Furnace: Requires stone to make and place. You don't have stone in your inventory, so if you want to make another furnace, you need to collect stone first.
6. Cow: No specific requirement for interaction. You can interact with it.

Question: List top 3 sub-tasks the player should follow. Indicate their priority out of 5.
Answer:
1. Collect Drink (Priority 5/5): Your drink level is at 2/9, which is critical. You should find a water source like a lake or river to collect drink and replenish your thirst.

2. Collect Food (Eat Cow) (Priority 4/5): Your food level is 3/9, which is low. You should kill the nearby cow and collect meat to eat and increase your food level.

3. Collect Stone (Priority 3/5): As you currently have no stone in your inventory, you should collect stone using your pickaxe for future crafting or building purposes.

Question: What are the requirements for the top sub-task? What should the player do first?
Answer:
Top Sub-task: Collect Drink
Requirements: Find a water source like a lake or river to collect drink.

The player should first look for a water source in their surroundings. If not visible, the player can try moving in different directions (north, south, east, or west) to explore the environment and locate a water source. Once found, the player should move towards the water source and collect drink to replenish their thirst level.

Question: List top 5 actions the player should take and the requirement for each action. Choose ONLY from the list of all actions. Indicate their priority out of 5.
Answer:
1. Priority 1: Move East (Move closer to the water source)
Requirement: Flat ground right to the agent.

2. Priority 2: Move West (Move closer to the water source)
Requirement: Flat ground left to the agent.

3. Priority 3: Move North (Move closer to the water source)
Requirement: Flat ground above of the agent.

4. Priority 4: Move South (Move closer to the water source)
Requirement: Flat ground below the agent.

5. Priority 5: Collect Drink (Replenish thirst level)
Requirement: Facing water source (lake or river).

Question: For each action in the list, are the requirements met?
Answer:
1. Priority 1: Move East (Move closer to the water source)
Requirements met: Yes, flat ground is present to the right of the agent.

2. Priority 2: Move West (Move closer to the water source)
Requirements met: Yes, flat ground is present to the left of the agent.

3. Priority 3: Move North (Move closer to the water source)
Requirements met: Unknown, as the player's observation does not provide information about northward conditions.

4. Priority 4: Move South (Move closer to the water source)
Requirements met: Unknown, as the player's observation does not provide information about southward conditions.

5. Priority 5: Collect Drink (Replenish thirst level)
Requirements met: No, the player is not currently facing a water source (lake or river). They need to move closer to a water source first.

Question: Did the last player action succeed? Why?
Answer:
It is not clear if the last player action "do" succeeded because there is no information about the intended outcome or what the player was trying to do. The surrounding environment and the player's status and inventory remain unchanged between steps 338 and 340, which indicates that nothing significant occurred as a result of the action. It's possible that the "do" action was not valid for the current context or lacked a specific target.

Question: Choose the best executable action from above.
Answer:
The best executable action from the list above is Priority 1: Move East (Move closer to the water source), since its requirement is met with flat ground present to the right of the agent.



============Step: 171, Cumulative Reward: 0.0============
You took action move_east.

You see:
- grass 1 steps to your west
- stone 2 steps to your east
- table 3 steps to your south
- furnace 6 steps to your north-west
- cow 7 steps to your south-west

You face path at your front (east).

Your status:
- health: 9/9
- food: 3/9
- drink: 2/9
- energy: 7/9

Your inventory:
- wood: 1
- coal: 2
- iron: 1
- wood_pickaxe: 1
- stone_pickaxe: 1
------------ QA ------------
Question: Action
Answer:
Do

Question: List objects in the current observation. For each object, breifly answer what resource it provides and its requirement.
Answer:
1. Grass: Provides no resources but indicates flat ground for movement; no requirements.

2. Stone: Provides stone resource for crafting tools, weapons, and placing on the ground; requires a wood or stone pickaxe to collect.

3. Table: Allows crafting of tools and weapons; requires wood to craft.

4. Furnace: Allows smelting of iron for crafting iron tools and weapons; requires stone to craft.

5. Cow: Provides food when defeated, which helps restore food level; no requirements to interact.

6. Path: Flat ground for movement; no requirements.

7. Wood Pickaxe (inventory): Allows collecting stone and coal resources; requires wood and a table to craft.

8. Stone Pickaxe (inventory): Allows collecting iron resources; requires wood, stone, and a table to craft.

9. Wood (inventory): Basic resource for crafting tools, weapons, and placing on the ground; can be collected from trees.

10. Coal (inventory): Used for smelting iron in a furnace; can be collected using a wood or stone pickaxe.

11. Iron (inventory): Used to craft iron tools and weapons; obtained after smelting iron ore in a furnace using coal.

Question: What was the last action taken by the player?
Answer:
The last action taken by the player was move_east.

Question: For each object in the list, are the requirements met for interaction?
Answer:
1. Grass: No requirements for interaction, so the requirements are met.

2. Stone: Requires a wood or stone pickaxe to collect; you have both a wood_pickaxe and a stone_pickaxe in your inventory, so the requirements are met.

3. Table: No requirements for interaction, so the requirements are met. However, crafting tools and weapons on the table requires specific resources in the inventory.

4. Furnace: No requirements for interaction, so the requirements are met. However, smelting iron in the furnace requires coal in the inventory.

5. Cow: No requirements for interaction, so the requirements are met.

6. Path: No requirements for interaction, so the requirements are met.

7. Wood Pickaxe (inventory): No requirements for interaction, so the requirements are met.

8. Stone Pickaxe (inventory): No requirements for interaction, so the requirements are met.

9. Wood (inventory): Cannot interact with wood directly in the inventory, but it can be used to craft.

10. Coal (inventory): Cannot interact with coal directly in the inventory, but it can be used for smelting.

11. Iron (inventory): Cannot interact with iron directly in the inventory, but it can be used to craft iron tools and weapons (with access to a table and a furnace).

Question: List top 3 sub-tasks the player should follow. Indicate their priority out of 5.
Answer:
1. Restore drink level: Your drink level is currently at 2/9, which is low, putting this task at high priority. Find a water source, like a lake or a river, to drink from and restore your drink level. Priority: 5/5.

2. Restore food level: Your food level is at 3/9, which is also low. Defeat the cow you spotted to get food, then eat it to restore your food level. This task is of medium-high priority. Priority: 4/5.

3. Craft Iron Sword and Iron Pickaxe: You currently have some iron to craft stronger tools and weapons (iron_pickaxe and iron_sword), but you may need more iron. Locate more iron and coal if needed, use your furnace to smelt the iron, and craft the tools at a table. This task is of medium priority since you already have a stone_pickaxe and a wood_pickaxe. Priority: 3/5.

Question: What are the requirements for the top sub-task? What should the player do first?
Answer:
For the top sub-task, which is to restore the drink level, the requirements are:

1. Find a water source, such as a lake or a river.
2. Drink the water to restore your drink level.

To accomplish this, the player should first move towards the nearest water source. If a water source is not immediately visible, the player should explore the surroundings to locate one. Once found, the player can drink the water to restore their drink level.

Question: List top 5 actions the player should take and the requirement for each action. Choose ONLY from the list of all actions. Indicate their priority out of 5.
Answer:
Based on the player's current status and inventory, the top 5 actions they should take are:

Priority 1:
Action: Move (in any direction West, East, North, or South)
Requirement: Flat ground in the chosen direction.
Purpose: To explore the surroundings and locate a water source to drink from.

Priority 2:
Action: Do
Requirement: Facing a water source to drink from.
Purpose: To drink water and restore the drink level.

Priority 3:
Action: Do
Requirement: Facing a tree, and having the necessary tool (wood_pickaxe) in inventory.
Purpose: To collect more wood for crafting and building.

Priority 4:
Action: Make Wood Sword
Requirement: Nearby table; wood in inventory.
Purpose: To create a weapon for defending against any encountered monsters.

Priority 5:
Action: Move (in any direction West, East, North, or South)
Requirement: Flat ground in the chosen direction.
Purpose: To find and chase a cow to restore the food level by eating it.

Question: For each action in the list, are the requirements met?
Answer:
Priority 1: 
Action: Move (in any direction West, East, North, or South)
Requirement: Flat ground in the chosen direction.
- Requirements Met: Yes, there is flat ground to the west and east, which the player can move to.

Priority 2:
Action: Do
Requirement: Facing a water source to drink from.
- Requirements Met: No, the player has not located a water source in the current observations.

Priority 3:
Action: Do
Requirement: Facing a tree, and having the necessary tool (wood_pickaxe) in inventory.
- Requirements Met: Partially, the player has a wood_pickaxe in their inventory, but a tree is not directly in front of them.

Priority 4:
Action: Make Wood Sword
Requirement: Nearby table; wood in inventory.
- Requirements Met: Yes, the player has 1 wood in their inventory, and there is a table 3 steps to the south.

Priority 5:
Action: Move (in any direction West, East, North, or South)
Requirement: Flat ground in the chosen direction.
- Requirements Met: Yes, there is flat ground to the west and east, which the player can move to in order to find and chase a cow to restore the food level.

Question: Did the last player action succeed? Why?
Answer:
Yes, the last player action, move_east, succeeded because there was flat ground (a path) to the east of the player, allowing the player to move in that direction.

Question: Choose the best executable action from above.
Answer:
The best executable action from the list above is:

Priority 1:
Action: Move (in any direction West, East, North, or South)
Requirement: Flat ground in the chosen direction.

This action will allow the player to explore the surroundings and locate a water source, find more resources, and chase a cow to restore their food level.



============Step: 172, Cumulative Reward: 0.0============
You took action do.

You see:
- grass 1 steps to your west
- stone 2 steps to your east
- table 3 steps to your south
- furnace 6 steps to your north-west
- cow 7 steps to your south-west

You face path at your front (east).

Your status:
- health: 9/9
- food: 3/9
- drink: 2/9
- energy: 7/9

Your inventory:
- wood: 1
- coal: 2
- iron: 1
- wood_pickaxe: 1
- stone_pickaxe: 1
------------ QA ------------
Question: Action
Answer:
Move North

Question: List objects in the current observation. For each object, breifly answer what resource it provides and its requirement.
Answer:
1. Grass - Provides: Saplings when interacted with. Requirement: None.
2. Stone - Provides: Stone as a resource when mined with the wood_pickaxe or stone_pickaxe. Requirement: Wood_pickaxe or Stone_pickaxe.
3. Table - Provides: A way to craft tools and weapons. Requirement: Wood to create and place.
4. Furnace - Provides: Smelting of various resources such as iron. Requirement: Stone to create and place.
5. Cow - Provides: Food (meat) when hunted or defeated. Requirement: None.

Question: What was the last action taken by the player?
Answer:
The last action taken by the player was "do."

Question: For each object in the list, are the requirements met for interaction?
Answer:
1. Grass - Requirements met: Yes, no specific tool needed to collect saplings from the grass.
2. Stone - Requirements met: Yes, the player has a wood_pickaxe and a stone_pickaxe in their inventory, which can be used to mine stone.
3. Table - Requirements met: Yes, the table is already placed and available for crafting tools and weapons.
4. Furnace - Requirements met: Yes, the furnace is already placed and available for smelting resources such as iron.
5. Cow - Requirements met: Yes, the player can interact with the cow without any specific tool or weapon. However, having a weapon would make it easier to defeat the cow and obtain food.

Question: List top 3 sub-tasks the player should follow. Indicate their priority out of 5.
Answer:
1. Drink water (5/5 priority) - The player's drink status is low (2/9). The player should find a water source (lake or river) and drink to restore their thirst.

2. Obtain food (4/5 priority) - The player's food status is low (3/9). The player should consider hunting the nearby cow or finding other food sources like growing fruits or eating existing plants to restore their hunger.

3. Smelt iron (3/5 priority) - The player has iron, coal, and a furnace in their vicinity. The player can smelt the iron to obtain iron ingots, which can be used to craft more advanced tools and weapons like an iron sword or iron pickaxe.

Question: What are the requirements for the top sub-task? What should the player do first?
Answer:
Top sub-task: Drink water

Requirements:
1. Find a water source, such as a lake or river, near the player's location.
2. Walk up to the water source.

What the player should do first:
Since the player's current location doesn't show a water source nearby, the player should explore their surroundings (move north, south, east, or west) to find a lake or river where they can drink water to restore their thirst.

Question: List top 5 actions the player should take and the requirement for each action. Choose ONLY from the list of all actions. Indicate their priority out of 5.
Answer:
1. Priority 1: Move North (or any other direction) - Requirement: Flat ground in the direction of movement. The player should explore their surroundings to find a water source or resources they need.

2. Priority 2: Do (drink water) - Requirement: Facing a water source (lake or river). This action is necessary to restore the player's thirst, which is currently low (2/9).

3. Priority 3: Move towards a Cow - Requirement: Flat ground in the direction of movement. The player should get closer to the cow to be able to eat and restore their food.

4. Priority 4: Do (eat cow) - Requirement: Facing the cow; having the necessary tool (preferably a crafted sword). Eating the cow will restore the player's food.

5. Priority 5: Sleep - Requirement: Energy level is below maximum. Sleeping will help restore the player's energy, which is currently 7/9. However, the player should find a safe place to sleep, away from potential threats like monsters or other hazards.

Question: For each action in the list, are the requirements met?
Answer:
1. Priority 1: Move North (or any other direction) - Yes, the requirement for flat ground is met for moving in any direction since the player's observation mentions grass, stone, table, furnace, and cow in various directions.

2. Priority 2: Do (drink water) - No, the requirement of facing a water source is not met, as the player's observation does not mention any water nearby.

3. Priority 3: Move towards a Cow - Yes, the requirement is met. The player can move towards the cow by navigating the flat ground in the cow's direction (south-west).

4. Priority 4: Do (eat cow) - Not yet, as the player needs to get closer to the cow and face it before meeting the requirement.

5. Priority 5: Sleep - Partially, the energy requirement (7/9) is met. However, finding a safe place to sleep is not mentioned in the observation, so this requirement remains uncertain.

Question: Did the last player action succeed? Why?
Answer:
We cannot determine if the last player action ("do") succeeded because there is no information provided about what the player was trying to do in that step. The game observation remains the same between steps 342 and 344, so it is unclear what the intended outcome or interaction was.

Question: Choose the best executable action from above.
Answer:
The best executable action from the list above is Priority 1: Move North (or any other direction). The player can explore their surroundings and potentially find water sources or other necessary resources.



============Step: 173, Cumulative Reward: 0.0============
You took action move_north.

You see:
- stone 2 steps to your north-east
- path 1 steps to your south
- lava 7 steps to your north-east
- furnace 5 steps to your north-west
- cow 7 steps to your south-west

You face grass at your front (north).

Your status:
- health: 9/9
- food: 3/9
- drink: 2/9
- energy: 7/9

Your inventory:
- wood: 1
- coal: 2
- iron: 1
- wood_pickaxe: 1
- stone_pickaxe: 1
------------ QA ------------
Question: Action
Answer:
Move East

Question: List objects in the current observation. For each object, breifly answer what resource it provides and its requirement.
Answer:
1. Stone: Provides stone resource; requires stone pickaxe or better to collect.
2. Path: No resource provided; no requirement.
3. Lava: Provides light and can dispose of unwanted items; no requirement to interact with, but caution is advised as it can cause damage.
4. Furnace: Allows for smelting iron and cooking food; requires coal as fuel and items to be smelted or cooked.
5. Cow: Provides meat when defeated; no specific requirement to defeat, but having a weapon is recommended.
6. Grass: No direct resource, but can sometimes drop seeds to plant and grow crops; no requirement to interact with.
7. Wood (in inventory): Provides wood to craft tools and build structures; requires a wood pickaxe or better to collect from trees.
8. Coal (in inventory): Used as fuel in furnaces; requires wood or better pickaxe to collect from coal ore.
9. Iron (in inventory): Used for crafting iron tools and weapons; requires stone or better pickaxe to collect from iron ore, and a furnace to smelt.
10. Wood pickaxe (in inventory): Used to collect wood, stone, and coal resources; requires wood and a crafting table to create.
11. Stone_pickaxe (in inventory): Used to collect stone, coal, and iron resources; requires wood and stone, as well as a crafting table, to create.

Question: What was the last action taken by the player?
Answer:
The last action taken by the player was "move_north".

Question: For each object in the list, are the requirements met for interaction?
Answer:
1. Stone: Yes, requirements met with stone_pickaxe in inventory.
2. Path: No requirements for interaction; just move as needed.
3. Lava: Caution is advised; no specific interaction requirements, but avoid getting too close.
4. Furnace: Requirements met with coal in inventory, but currently no items to smelt or cook.
5. Cow: Yes, requirements met as no specific weapon is needed, but having a weapon would help.
6. Grass: No requirements for interaction; just interact if desired.
7. Wood (in inventory): N/A, already in inventory.
8. Coal (in inventory): N/A, already in inventory.
9. Iron (in inventory): N/A, already in inventory.
10. Wood pickaxe (in inventory): N/A, already in inventory.
11. Stone_pickaxe (in inventory): N/A, already in inventory.

Question: List top 3 sub-tasks the player should follow. Indicate their priority out of 5.
Answer:
1. Find a water source and drink from it to restore thirst level (Priority: 5/5)
2. Defeat a cow and cook its meat to restore food level (Priority: 4/5)
3. Use the furnace to smelt iron and craft iron tools and weapons to improve survival chances (Priority: 3/5)

Question: What are the requirements for the top sub-task? What should the player do first?
Answer:
The requirements for the top sub-task (finding a water source and drinking from it) are:

1. Locate a water source (like a lake, river or pond) within the game.
2. Approach the water source.
3. Drink from the water source to restore thirst.

The player should first try to locate a water source within the game, which may require moving around the environment and exploring different areas. They should look for indications like bodies of water or listen to the sounds of flowing water, and keep an eye on the map or visual cues to help them navigate.

Question: List top 5 actions the player should take and the requirement for each action. Choose ONLY from the list of all actions. Indicate their priority out of 5.
Answer:
1. Action: Move East (Priority: 1)
Requirement: Flat ground right to the agent.

2. Action: Move North (Priority: 2)
Requirement: Flat ground above the agent.

3. Action: Move West (Priority: 3)
Requirement: Flat ground left to the agent.

4. Action: Move South (Priority: 4)
Requirement: Flat ground below the agent.

5. Action: Sleep (Priority: 5)
Requirement: Energy level is below maximum.

These top 5 actions prioritize exploring the environment by moving in different directions to locate essential resources and manage the player's energy level by sleeping when necessary.

Question: For each action in the list, are the requirements met?
Answer:
1. Action: Move East (Priority: 1)
Requirement: Flat ground right to the agent.
Status: Met (Stone is two steps to the east)

2. Action: Move North (Priority: 2)
Requirement: Flat ground above the agent.
Status: Met (Grass is at the player's front in the north)

3. Action: Move West (Priority: 3)
Requirement: Flat ground left to the agent.
Status: Met (Grass is one step to the west)

4. Action: Move South (Priority: 4)
Requirement: Flat ground below the agent.
Status: Met (Path is one step to the south)

5. Action: Sleep (Priority: 5)
Requirement: Energy level is below maximum.
Status: Met (Energy level is 7/9, which is below the maximum level)

Question: Did the last player action succeed? Why?
Answer:
Yes, the last player action "move_north" succeeded. The player was able to move north because there was flat ground (grass) in that direction, allowing them to move without any obstacles blocking their way.

Question: Choose the best executable action from above.
Answer:
The best executable action from the above list is "Move East" (Priority: 1). The requirement for this action - flat ground right to the agent - is met as there is stone two steps to the east.



============Step: 174, Cumulative Reward: 0.0============
You took action move_east.

You see:
- grass 2 steps to your west
- stone 1 steps to your north
- lava 6 steps to your north-east
- furnace 6 steps to your north-west

You face path at your front (east).

Your status:
- health: 9/9
- food: 3/9
- drink: 2/9
- energy: 7/9

Your inventory:
- wood: 1
- coal: 2
- iron: 1
- wood_pickaxe: 1
- stone_pickaxe: 1
------------ QA ------------
Question: Action
Answer:
Move East

Question: List objects in the current observation. For each object, breifly answer what resource it provides and its requirement.
Answer:
1. Stone: Provides stone resource when mined; requires either a wood_pickaxe or stone_pickaxe to collect.
2. Path: Provides no resources; no requirement to interact.
3. Lava: Dangerous, provides no resources; avoid getting too close.
4. Furnace: Used for smelting resources like iron; requires stone to craft and place.
5. Grass: Provides no resources; can be walked on with no requirements.
6. Wood: Provides wood resource when chopped; requires an axe to collect, but not present in the current observation.
7. Coal: Provides coal resource when mined; requires a pickaxe to collect, not present in the current observation.
8. Iron: Provides iron resource when mined; requires a stone_pickaxe or iron_pickaxe to collect.
9. Cow: Provides food when killed; no specific requirement besides attacking with any weapon or tool.
10. Energy: Increases when resting, sleeping, or consuming food and drink; agent needs to manage energy to avoid exhaustion.

Note: Some objects mentioned in the list are not directly present in the observation but added based on earlier observations and inventory.

Question: What was the last action taken by the player?
Answer:
The last action taken by the player was "move_east".

Question: For each object in the list, are the requirements met for interaction?
Answer:
1. Stone: Yes, the requirements are met as the player has both wood_pickaxe and stone_pickaxe in their inventory.
2. Path: Yes, there are no requirements to interact with the path.
3. Lava: Not applicable, as the player should avoid interacting with the lava.
4. Furnace: Yes, the player can interact with the furnace if they are close enough, as they have coal and iron in their inventory for smelting.
5. Grass: Yes, there are no requirements to interact with grass.
6. Wood: Not present in the current observation. If it were present, the player would need an axe to collect wood, which they currently don't have in their inventory.
7. Coal: Not present in the current observation. If it were present, the player would have the required wood_pickaxe or stone_pickaxe in their inventory.
8. Iron: Yes, the player meets the requirements with stone_pickaxe and iron_pickaxe in their inventory.
9. Cow: Yes, the player can interact (attack) with the cow using any weapon or even a tool.
10. Energy: Yes, the player can manage their energy by consuming food, drink, and resting or sleeping when needed.

Question: List top 3 sub-tasks the player should follow. Indicate their priority out of 5.
Answer:
1. Priority 4/5: Collect Drink - The player has low drink (2/9), so it is crucial to find a water source and drink from it to restore thirst.

2. Priority 3/5: Restore Food - The player has low food (3/9). They should either kill the cow they see 7 steps to the southwest or find other food sources to eat and increase their food level.

3. Priority 5/5: Find a Safe Place to Rest - The player's energy level is 7/9. They should find a safe place to rest or sleep to ensure their energy level stays high, enabling them to perform other tasks effectively.

Question: What are the requirements for the top sub-task? What should the player do first?
Answer:
The top sub-task is "Collect Drink" with a priority of 4/5.

Requirements:
1. Find a water source, such as a lake or river.

What should the player do first:
1. The player should explore the surroundings to locate a water source. They can move in different directions (north, south, east, or west) to search for water. Once they find a water source, they should drink from it to restore their thirst.

Question: List top 5 actions the player should take and the requirement for each action. Choose ONLY from the list of all actions. Indicate their priority out of 5.
Answer:
1. Action: Move East (Priority: 5/5)
   Requirement: Flat ground right to the agent.
   Reason: To explore the surroundings and search for a water source.

2. Action: Move West (Priority: 5/5)
   Requirement: Flat ground left to the agent.
   Reason: To explore the surroundings and search for a water source.

3. Action: Move North (Priority: 5/5)
   Requirement: Flat ground above of the agent.
   Reason: To explore the surroundings and search for a water source.

4. Action: Move South (Priority: 5/5)
   Requirement: Flat ground below the agent.
   Reason: To explore the surroundings and search for a water source.

5. Action: Do (Priority: 3/5)
   Requirement: Facing water source, such as a lake or river.
   Reason: To drink from the water source and restore the agent's thirst.

Question: For each action in the list, are the requirements met?
Answer:
1. Action: Move East
   Requirement: Flat ground right to the agent.
   Status: Met (Currently facing the path to the east)

2. Action: Move West
   Requirement: Flat ground left to the agent.
   Status: Met (There's grass 2 steps to the west)

3. Action: Move North
   Requirement: Flat ground above of the agent.
   Status: Met (Currently facing grass to the north)

4. Action: Move South
   Requirement: Flat ground below the agent.
   Status: Met (There's a path 1 step to the south)

5. Action: Do
   Requirement: Facing water source, such as a lake or river.
   Status: Not Met (No water source is observed in the immediate surroundings)

Question: Did the last player action succeed? Why?
Answer:
Yes, the last player action of "move_east" succeeded because there was a flat ground (path) in front of the player, allowing them to move in that direction.

Question: Choose the best executable action from above.
Answer:
The best executable action from above is to "Move East" because it meets the requirement of having a flat ground (path) to the agent's right and allows them to explore the surroundings in search of a water source to restore their thirst level.



============Step: 175, Cumulative Reward: 0.0============
You took action move_east.

You see:
- grass 3 steps to your west
- stone 1 steps to your north
- lava 5 steps to your north-east
- iron 5 steps to your north-east

You face path at your front (east).

Your status:
- health: 9/9
- food: 3/9
- drink: 2/9
- energy: 7/9

Your inventory:
- wood: 1
- coal: 2
- iron: 1
- wood_pickaxe: 1
- stone_pickaxe: 1
------------ QA ------------
Question: Action
Answer:
Move West

Question: List objects in the current observation. For each object, breifly answer what resource it provides and its requirement.
Answer:
- Grass: Provides no specific resource, but is a flat surface; no requirements.
- Stone: Provides stone resource, which can be used for crafting tools and building; requires a wood_pickaxe or better to collect.
- Lava: Dangerous to touch, does not provide any resources directly; avoid coming into contact with lava.
- Iron: Provides iron resource, which can be used for crafting stronger tools and weapons; requires a stone_pickaxe or better to collect.
- Path: Provides a flat surface to walk on; no requirements.
- Wood (in inventory): Provides wood resource, which is used for crafting tools and building; collected from trees using any type of tool or bare hands.
- Coal (in inventory): Provides fuel for smelting in furnaces and a light source for torches; requires a wood_pickaxe or better to collect.
- Wood_pickaxe (in inventory): Used to collect basic resources like stone and coal; requires wood and a crafting table to create.
- Stone_pickaxe (in inventory): Used to collect intermediate resources like iron; requires stone, wood, and a crafting table to create.

Question: What was the last action taken by the player?
Answer:
The last action taken by the player was move_east.

Question: For each object in the list, are the requirements met for interaction?
Answer:
- Grass: Yes, the requirements are met as it is just a flat surface on which you can walk.
- Stone: Yes, the requirements are met since you have a wood_pickaxe in your inventory, which is required to collect stone.
- Lava: The requirement to avoid coming into contact with lava is met as long as you don't move towards it.
- Iron: Yes, the requirements are met since you have a stone_pickaxe in your inventory, which is required to collect iron.
- Path: Yes, the requirements are met as it is just a flat surface on which you can walk.
- Wood (in inventory): The wood is already in your inventory, and there are no requirements for interacting with it.
- Coal (in inventory): The coal is already in your inventory, and there are no requirements for interacting with it.
- Wood_pickaxe (in inventory): No requirements are needed for interaction as it is already in your inventory.
- Stone_pickaxe (in inventory): No requirements are needed for interaction as it is already in your inventory.

Question: List top 3 sub-tasks the player should follow. Indicate their priority out of 5.
Answer:
1. Find and drink water to restore the player's thirst level (Priority: 5/5)
2. Find food or collect resources to craft food items to restore the player's hunger level (Priority: 4/5)
3. Collect more iron to craft an iron sword for better defense and to progress further in the game (Priority: 3/5)

Question: What are the requirements for the top sub-task? What should the player do first?
Answer:
Requirements for the top sub-task (Find and drink water to restore the player's thirst level):
1. Locate a water source such as a lake or river.
2. Move towards the water source.
3. Interact with the water source to drink and restore thirst.

What the player should do first:
1. Look for a nearby water source in the player's observations or move around the area to find one.

Question: List top 5 actions the player should take and the requirement for each action. Choose ONLY from the list of all actions. Indicate their priority out of 5.
Answer:
1. Priority 1: Move towards the water source (Move West, Move East, Move North, or Move South)
   Requirements: Flat ground in the direction of the water source.

2. Priority 2: Drink from the water source (Do)
   Requirements: Facing the water source.

3. Priority 3: Collect more wood (Do)
   Requirements: Facing a tree or wood; have a wood pickaxe or other necessary tool.

4. Priority 4: Eat food to restore player's food level (Eat Cow or Eat Plant)
   Requirements: Cow or plant available in the inventory.

5. Priority 5: Craft an Iron Pickaxe (Make Iron Pickaxe)
   Requirements: Nearby table, furnace; wood, coal, and iron in inventory.

Question: For each action in the list, are the requirements met?
Answer:
1. Move towards the water source: Yes, the requirement of flat ground in the direction of the water source is met. The player can move east or west to approach the water source.

2. Drink from the water source: No, the player is not currently facing the water source. They need to move towards it first and then face it.

3. Collect more wood: No, the player is neither facing a tree nor wood. The player needs to locate a tree and face it before they can collect wood with their wood_pickaxe.

4. Eat food to restore player's food level: No, the player does not currently have a cow or plant in their inventory to eat. They need to find a cow or plant and obtain it before they can eat.

5. Craft an Iron Pickaxe: No, the player needs a nearby table and furnace, as well as wood, coal, and iron in their inventory. They currently have wood, coal, and iron, but are not near a table and furnace. They need to locate and approach a table and furnace to fulfill this requirement.

Question: Did the last player action succeed? Why?
Answer:
Yes, the last player action (move_east) succeeded because there was a flat path in front of the player to the east, allowing them to move in that direction.

Question: Choose the best executable action from above.
Answer:
The best executable action from the list above is:

1. Move towards the water source (Move West, Move East, Move North, or Move South)
   Requirements: Flat ground in the direction of the water source.

This action can be executed since the ground is flat in the direction of the water source, and the player needs to restore their drink level. The player can move east or west to approach the water source.



============Step: 176, Cumulative Reward: 0.0============
You took action move_west.

You see:
- grass 2 steps to your west
- stone 1 steps to your north
- lava 6 steps to your north-east
- furnace 6 steps to your north-west

You face path at your front (west).

Your status:
- health: 9/9
- food: 3/9
- drink: 2/9
- energy: 7/9

Your inventory:
- wood: 1
- coal: 2
- iron: 1
- wood_pickaxe: 1
- stone_pickaxe: 1
------------ QA ------------
Question: Action
Answer:
Move West

Question: List objects in the current observation. For each object, breifly answer what resource it provides and its requirement.
Answer:
1. Grass: 
   Resource it provides: Saplings or seeds when interacted with.
   Requirement: None.

2. Stone: 
   Resource it provides: Stone material when mined with a pickaxe.
   Requirement: Wood or stone pickaxe to mine.

3. Lava: 
   Resource it provides: Light and possibly a source of energy when handled carefully.
   Requirement: Avoid direct contact, as it can cause damage or death.

4. Iron: 
   Resource it provides: Iron material when mined with a pickaxe.
   Requirement: Stone or iron pickaxe to mine.

5. Furnace: 
   Resource it provides: Allows smelting of iron and other ores.
   Requirement: Requires coal as fuel and ores (e.g., iron) to smelt.

6. Path: 
   Resource it provides: Allows the player to move and navigate the game world.
   Requirement: Flat ground to move on.

Question: What was the last action taken by the player?
Answer:
The last action taken by the player was "move_west".

Question: For each object in the list, are the requirements met for interaction?
Answer:
1. Grass:
   Requirements met: Yes, you can interact with grass without any specific tool or requirement.

2. Stone:
   Requirements met: Yes, you have a wood_pickaxe and a stone_pickaxe in your inventory, which can be used to mine stone.

3. Lava:
   Requirements met: No, direct interaction with lava can be harmful. To safely interact with it, you need a container (e.g., a bucket) to move or use lava without causing damage.

4. Iron:
   Requirements met: Yes, you have a stone_pickaxe in your inventory, which can be used to mine iron.

5. Furnace:
   Requirements met: Yes, you have coal as fuel and iron ore in your inventory for smelting.

6. Path:
   Requirements met: Yes, you can move on the path as the ground is flat.

Question: List top 3 sub-tasks the player should follow. Indicate their priority out of 5.
Answer:
1. Collect water (Priority: 5/5): Given that your drink level is 2/9, finding a water source and replenishing your thirst is the top priority.

2. Collect food (Priority: 4/5): Your food level is at 3/9, so you need to find sources like fruits, vegetables, or animals to increase your food supply and avoid starvation.

3. Smelt iron (Priority: 3/5): With coal and iron in your inventory and a furnace nearby, smelting iron ore into iron ingots will help you create better tools and weapons, improving your ability to survive in the game.

Question: What are the requirements for the top sub-task? What should the player do first?
Answer:
Top sub-task: Collect water

Requirements:
1. Find a water source (such as a lake or river).
2. Have an empty container (such as a bottle or bucket) to collect the water. If you don't have a container, you'll need to craft one.

What the player should do first:
Since the player's current drink level is low, finding a water source should be the immediate priority. It is important to explore the environment, moving either east or west to locate a water source such as a lake, river, or small pond. Once the water source is found, the player can drink directly from it until they find or craft a suitable container for portability.

Question: List top 5 actions the player should take and the requirement for each action. Choose ONLY from the list of all actions. Indicate their priority out of 5.
Answer:
1. Priority 1 - Move East or Move West: Requirement: Flat ground in the chosen direction. The player should explore their surroundings to find a water source (lake or river) to quench their thirst.

2. Priority 2 - Drink: Requirement: Be adjacent to a water source. Once the player finds the water source, they should drink from it to replenish their drink level.

3. Priority 3 - Move North or Move South: Requirement: Flat ground in the chosen direction. The player should explore further to find more resources and possibly locate a place to set up shelter.

4. Priority 4 - Place Furnace: Requirement: Stone in inventory. The player should place the furnace to be able to smelt iron, which will allow crafting of better tools and weapons.

5. Priority 5 - Make Iron Pickaxe or Make Iron Sword: Requirement: Nearby table, furnace, wood, coal, and iron in inventory. Craft an iron pickaxe or an iron sword to improve mining and combat capabilities. Make Iron Sword might be prioritized since the player has only 1 iron in the inventory, and crafting an iron sword will help in defeating more dangerous creatures.

Question: For each action in the list, are the requirements met?
Answer:
1. Priority 1 - Move East or Move West: Yes, the requirements are met. The player faces a path in both directions (east and west), and needs to explore their surroundings to find a water source.

2. Priority 2 - Drink: Currently, the requirements are not met. The player needs to find a water source (lake or river) first by exploring the environment.

3. Priority 3 - Move North or Move South: Yes, the requirements are met. There is a stone to the north, and moving south has no obstacles mentioned in the recent observation, indicating the player can choose either direction to explore further.

4. Priority 4 - Place Furnace: Yes, the requirements are met. The player has stone in their inventory, so they can place the furnace to smelt iron.

5. Priority 5 - Make Iron Pickaxe or Make Iron Sword: No, the requirements are not met yet. The player should first place the furnace to be able to smelt iron, which requires action 4 to be executed. Moreover, the player currently has only 1 iron, which is not enough for crafting both an iron pickaxe and iron sword. They should prioritize crafting an iron sword first.

Question: Did the last player action succeed? Why?
Answer:
Yes, the last player action succeeded. The player was able to move west because there was a flat path in front of them in that direction.

Question: Choose the best executable action from above.
Answer:
The best executable action from the list above is Priority 1 - Move East or Move West. The requirements are met, and the player should explore their surroundings to find a water source (lake or river) to quench their thirst.



============Step: 177, Cumulative Reward: 0.0============
You took action move_west.

You see:
- stone 2 steps to your north-east
- path 1 steps to your south
- lava 7 steps to your north-east
- furnace 5 steps to your north-west

You face grass at your front (west).

Your status:
- health: 9/9
- food: 3/9
- drink: 2/9
- energy: 7/9

Your inventory:
- wood: 1
- coal: 2
- iron: 1
- wood_pickaxe: 1
- stone_pickaxe: 1
------------ QA ------------
Question: Action
Answer:
Place Furnace

Question: List objects in the current observation. For each object, breifly answer what resource it provides and its requirement.
Answer:
1. Stone: The stone provides a resource for crafting tools like pickaxes and swords or constructing objects like furnaces. The requirement to collect stone is to have a wood_pickaxe or better in the inventory.

2. Path: The path provides easier movement and indicates previously visited locations. It doesn't give a direct resource. No requirements to interact with the path.

3. Lava: Lava is dangerous and can harm or kill the player if they come into contact with it. It can be utilized as a heat source for furnaces if collected with an iron bucket, but it's not a direct resource itself. To collect lava, you need an iron_bucket in your inventory.

4. Furnace: The furnace is an object required for smelting materials like iron and converting it into resources (iron_ingot) for crafting weapons and tools. To place a furnace, you need stone in your inventory.

5. Grass: Grass does not provide any direct resource but can be used to find saplings when broken with a tool or hand. No requirements to interact with grass, but having a tool might increase the chances of obtaining a sapling.

Question: What was the last action taken by the player?
Answer:
The last action taken by the player was move_west.

Question: For each object in the list, are the requirements met for interaction?
Answer:
1. Stone: Yes, the requirements are met for interaction because the player has a wood_pickaxe and a stone_pickaxe in their inventory. These are the necessary tools to collect stone.

2. Path: Yes, the requirements are met for interaction. There are no particular requirements to interact with the path, as it is used for easier movement and just indicates previously visited locations.

3. Lava: No, the requirements are not met for interaction, because the player does not have an iron_bucket in their inventory, which is required to collect lava.

4. Furnace: Yes, the requirements are met for interaction because the player has a nearby furnace, and they have coal and iron in their inventory. These are necessary resources for using the furnace to smelt iron into iron_ingots.

5. Grass: Yes, the requirements are met for interaction because there are no specific requirements to interact with grass. The player can break grass with a tool or their hand to find saplings.

Question: List top 3 sub-tasks the player should follow. Indicate their priority out of 5.
Answer:
1. Find and drink water (Priority: 5/5): The player's drink level is at 2/9, which is quite low. They need to find a water source like a lake or river, and take action 'Collect Drink' to replenish their thirst.

2. Find and eat food (Priority: 4/5): The player's food level is at 3/9, which can be improved. They should search for a cow, fruit, or growable plants to increase their food supply. After acquiring food, they should take action 'Eat Cow', 'Eat Plant', or 'Eat Fruit', depending on the type of food found.

3. Collect resources for crafting an iron pickaxe and iron sword (Priority: 3/5): Although the player's energy level is not too low, they can benefit from having better tools and weapons. The player has the necessary coal and iron, as well as access to a furnace. They should take action 'Make Iron Pickaxe' and 'Make Iron Sword' after placing a table using the wood in their inventory.

Question: What are the requirements for the top sub-task? What should the player do first?
Answer:
For the top sub-task, "Find and drink water", the requirement is to locate a water source like a lake or river.

The player should do the following steps:

1. Explore the surroundings: The player needs to move around (Move West, Move East, Move North, Move South) to find a water source. Observations from the environment will indicate if a water source is nearby.

2. Collect Drink: Once the player finds a water source, they should take action 'Collect Drink' to replenish their thirst. This action will increase their drink level, ensuring they stay hydrated in the game.

Question: List top 5 actions the player should take and the requirement for each action. Choose ONLY from the list of all actions. Indicate their priority out of 5.
Answer:
1. Priority 1: Collect Drink: Requirement: Locate a water source like a lake or river. The player's drink level is low, so finding and drinking water should be the top priority.

2. Priority 2: Place Furnace: Requirement: Stone in inventory. The player already has stone, and placing a furnace will allow them to smelt iron and create better tools and weapons.

3. Priority 3: Make Iron Pickaxe: Requirement: Nearby table, furnace; wood, coal, iron in inventory. The player has the required items in the inventory. Crafting an iron pickaxe will help them collect more valuable resources like diamonds.

4. Priority 4: Eat: Requirement: Have food in inventory or nearby food source. Increasing the player's food level would keep them healthier and active in the game.

5. Priority 5: Explore and collect resources: Requirement: None. The player should continue exploring the surroundings and collect essential resources like wood, stone, coal, and iron available in the environment to progress further in the game.

Question: For each action in the list, are the requirements met?
Answer:
1. Collect Drink: Requirements not met - The player needs to locate a water source like a lake or river, which is not mentioned in the latest observations.

2. Place Furnace: Requirements met - The player already has stone in their inventory.

3. Make Iron Pickaxe: Requirements met - The player has a nearby table, furnace, and the required wood, coal, and iron in their inventory.

4. Eat: Requirements not met - The player does not have any food in the inventory or a nearby food source mentioned in the observations.

5. Explore and collect resources: Requirements met - The player is free to explore the surroundings and collect resources.

Question: Did the last player action succeed? Why?
Answer:
Yes, the last player action (move_west) succeeded because there was a flat path (grass) in front of the player (west), allowing them to move in that direction.

Question: Choose the best executable action from above.
Answer:
The best executable action from the list above is to Place Furnace because the requirements for this action are met, and it will allow the player to smelt iron and create better tools and weapons.



============Step: 178, Cumulative Reward: 0.0============
You took action place_furnace.

You see:
- stone 2 steps to your north-east
- path 1 steps to your south
- lava 7 steps to your north-east
- furnace 5 steps to your north-west

You face grass at your front (west).

Your status:
- health: 9/9
- food: 3/9
- drink: 1/9
- energy: 7/9

Your inventory:
- wood: 1
- coal: 2
- iron: 1
- wood_pickaxe: 1
- stone_pickaxe: 1
------------ QA ------------
Question: Action
Answer:
Do

Question: List objects in the current observation. For each object, breifly answer what resource it provides and its requirement.
Answer:
1. Stone: Provides stone resources, requires a wood pickaxe or better to mine.
2. Path: Provides easier movement, no specific requirements to interact.
3. Lava: Can be used as a fuel source, requires a bucket to collect, but avoid direct contact as it causes damage.
4. Furnace: Allows smelting of iron and other ores, requires placing with stone in the inventory.
5. Grass: Covers the ground, can be dug up to create space but does not provide any specific resources.
6. Wood: Present in the inventory; used for crafting tools, weapons, and building materials.
7. Coal: Present in the inventory; used as fuel for the furnace for smelting.
8. Iron: Present in the inventory; used to craft iron tools and weapons.
9. Wood_pickaxe: Present in the inventory; used to mine stone and coal.
10. Stone_pickaxe: Present in the inventory; used to mine iron and other ores.

Question: What was the last action taken by the player?
Answer:
The last action taken by the player was placing the furnace (action place_furnace).

Question: For each object in the list, are the requirements met for interaction?
Answer:
1. Stone: Yes, the requirements are met as you have both a wood_pickaxe and a stone_pickaxe in your inventory.
2. Path: Yes, the requirements are met as there are no specific requirements to interact with the path.
3. Lava: No, the requirements are not met because you currently don't have a bucket to collect the lava safely.
4. Furnace: Yes, the requirements were met since you successfully placed the furnace as per Player Observation Step 356.
5. Grass: Yes, the requirements are met, and you can dig up the grass if needed. However, it does not provide any specific resources.
6. Wood: Yes, the requirements are met, and you can use the wood for crafting or building.
7. Coal: Yes, the requirements are met since you have it in your inventory and can use it as fuel in the furnace.
8. Iron: Yes, the requirements are met because you have iron in your inventory and a furnace available to smelt it into usable ingots for crafting.
9. Wood_pickaxe: Yes, the requirements are met. You have the wood_pickaxe in your inventory and can use it for mining stone and coal.
10. Stone_pickaxe: Yes, the requirements are met. You have the stone_pickaxe in your inventory and can use it for mining iron and other ores.

Question: List top 3 sub-tasks the player should follow. Indicate their priority out of 5.
Answer:
1. Find water and drink (Priority 5/5): Your drink level is currently at 1/9, which is very low. Finding a water source to drink from is crucial to avoid dehydration.

2. Find food (Priority 4/5): Your food level is 3/9, which is low as well. You should find food, such as hunting animals or collecting fruit, to replenish your hunger and maintain your health.

3. Smelt iron (Priority 3/5): Since you have a furnace, coal, and iron in your inventory, smelting the iron will allow you to create iron tools and weapons, improving your mining and combat capabilities.

Question: What are the requirements for the top sub-task? What should the player do first?
Answer:
For the top sub-task of finding water and drinking:

Requirements: 
1. Locate a water source, such as a lake, river, or pond.

What the player should do first:
1. Explore the environment and search for a water source. Move in the direction of any visible water body (e.g., move east, west, north, or south). If none is visible, choose a direction and continue exploring until a water source is found.

Question: List top 5 actions the player should take and the requirement for each action. Choose ONLY from the list of all actions. Indicate their priority out of 5.
Answer:
1. Priority 1: Move (East, West, North, or South) - Requirement: Flat ground in the chosen direction. This action will help the player explore the environment and locate a water source to fulfill the immediate need for drink.

2. Priority 2: Do - Requirement: Facing a water source and having an empty container or standing near a water source. This action allows the player to collect water.

3. Priority 3: Drink - Requirement: Having water in the inventory or having collected water. This action will help the player restore their thirst level.

4. Priority 4: Move (East, West, North, or South) - Requirement: Flat ground in the chosen direction. This action enables the player to explore their surroundings and locate food sources such as fruit trees, plants, or animals.

5. Priority 5: Do - Requirement: Facing an edible creature, plant, or food source; also having the necessary tool or weapon if required. This action allows the player to collect food, which they can eat to restore their food level.

Question: For each action in the list, are the requirements met?
Answer:
1. Priority 1: Move (East, West, North, or South) - Yes, the requirements are met. The player can choose any direction with flat ground to move.

2. Priority 2: Do - No, the requirements are not met. The player has not yet located any water source to collect water from. They need to find a water source first.

3. Priority 3: Drink - No, the requirements are not met. The player has not collected any water yet. They need to find and collect water before they can drink.

4. Priority 4: Move (East, West, North, or South) - Yes, the requirements are met. The player can choose any direction with flat ground to move and search for food sources.

5. Priority 5: Do - No, the requirements are not met. The player is not currently facing an edible creature, plant, or food source, and may not have the necessary tool or weapon. They need to locate a food source first.

Question: Did the last player action succeed? Why?
Answer:
Based on the observations before and after the action (steps 354 and 356), it appears that the last player action to place the furnace did not succeed. The reason for the failure is unclear, as the player seems to have enough stone in their inventory to place the furnace; however, perhaps the player was not standing on a valid location or did not face a suitable direction for placing the furnace.

Question: Choose the best executable action from above.
Answer:
The best executable action from the list above is Priority 1: Move (East, West, North, or South), which has the requirement of flat ground in the chosen direction. This action will help the player explore the environment and locate a water source and food sources to fulfill their immediate needs for drink and food.



============Step: 179, Cumulative Reward: 0.0============
You took action do.

You see:
- stone 2 steps to your north-east
- path 1 steps to your south
- lava 7 steps to your north-east
- furnace 5 steps to your north-west

You face grass at your front (west).

Your status:
- health: 9/9
- food: 3/9
- drink: 1/9
- energy: 7/9

Your inventory:
- wood: 1
- coal: 2
- iron: 1
- wood_pickaxe: 1
- stone_pickaxe: 1
------------ QA ------------
Question: Action
Answer:
Move East

Question: List objects in the current observation. For each object, breifly answer what resource it provides and its requirement.
Answer:
1. Stone: Provides stone resource; requires a wood or stone pickaxe to collect.
2. Path: Provides ease of movement; no requirement.
3. Lava: Can be a hazard; bucket needed to collect.
4. Furnace: Allows smelting of iron; no requirement.
5. Grass: Can be used to create farmland for planting crops; requires a hoe to till.
6. Wood (inventory): Provides wood resource; requires a nearby table to craft or build items.
7. Coal (inventory): Provides fuel for furnace; no requirement.
8. Iron (inventory): Provides iron resource; requires nearby furnace to smelt and table to craft items.
9. Wood_pickaxe (inventory): Tool to collect stone; no requirement.
10. Stone_pickaxe (inventory): Tool to collect iron, coal, and stone; no requirement.

Question: What was the last action taken by the player?
Answer:
The last action taken by the player was "do".

Question: For each object in the list, are the requirements met for interaction?
Answer:
1. Stone: Yes, the requirements are met, as the player has a wood_pickaxe and stone_pickaxe in their inventory.
2. Path: Yes, the requirements are met, as there are no additional requirements for interacting with a path.
3. Lava: No, the requirements are not met, as the player does not have a bucket in their inventory to collect lava.
4. Furnace: Yes, the requirements are met, as there are no additional requirements for interacting with a furnace.
5. Grass: No, the requirements are not met, as the player does not have a hoe in their inventory to till grass.
6. Wood (inventory): Yes, the requirements are met, as there are no additional requirements for using wood from the inventory.
7. Coal (inventory): Yes, the requirements are met, as there are no additional requirements for using coal from the inventory.
8. Iron (inventory): Yes, the requirements are met, as the player is nearby a furnace and can use the table to craft items with iron.
9. Wood_pickaxe (inventory): Yes, the requirements are met, as there are no additional requirements for using a wood_pickaxe from the inventory.
10. Stone_pickaxe (inventory): Yes, the requirements are met, as there are no additional requirements for using a stone_pickaxe from the inventory.

Question: List top 3 sub-tasks the player should follow. Indicate their priority out of 5.
Answer:
1. Drink water to restore thirst (Priority: 5/5): The player's thirst level is low, and finding a water source to drink from is essential for survival.
2. Find food or grow crops to increase food level (Priority: 4/5): The player's food level is also low, and replenishing it is necessary to maintain health and energy.
3. Collect more resources like wood and iron (Priority: 3/5): Having more resources in the inventory will allow the player to craft necessary tools and items to ensure survival and progress in the game.

Question: What are the requirements for the top sub-task? What should the player do first?
Answer:
For the top sub-task, "Drink water to restore thirst":

Requirements:
1. Find a water source like a lake, river, or pond.
2. Get close enough to the water source to drink.
3. If needed, create a container (such as a bucket) to carry water.

What should the player do first:
The player should first explore their surroundings to find a water source like a lake, river, or pond. Once they have found a water source, they should approach the water to drink and restore their thirst level.

Question: List top 5 actions the player should take and the requirement for each action. Choose ONLY from the list of all actions. Indicate their priority out of 5.
Answer:
Here are the top 5 prioritized actions the player should take based on their current situation:

1. Priority 1 - Move East: Requirement: Flat ground right to the agent.
   - The player should move east to navigate towards the water source they need to find to restore their thirst level.

2. Priority 2 - Move North/South: Requirement: Flat ground above/below the agent.
   - The player may need to move north or south to better position themselves for finding a water source or to avoid obstacles during their search.

3. Priority 3 - Place Table: Requirement: Wood in inventory.
   - After finding a water source and securing their thirst level, the player should focus on placing a table to begin crafting better tools and weapons.

4. Priority 4 - Make Wood Sword: Requirement: Nearby table; wood in inventory.
   - Once the table is placed, the player should craft a wood sword to be prepared to defend themselves against any monsters they may encounter.

5. Priority 5 - Sleep: Requirement: Energy level is below maximum.
   - If the player notices their energy level dropping, they should find a safe place to sleep so they can rest and regain energy.

Question: For each action in the list, are the requirements met?
Answer:
1. Move East: Requirements: Flat ground right to the agent.
   - This requirement is met, as there appears to be no indication of an obstacle or uneven terrain to the agent's right.

2. Move North/South: Requirements: Flat ground above/below the agent.
   - This requirement is met, as the agent's observation mentions a path 1 step to the south, implying flat ground. The agent's surroundings also don't mention any obstacles to the north, so we can assume flat ground in that direction as well.

3. Place Table: Requirements: Wood in inventory.
   - This requirement is met, as the agent's inventory shows 1 wood.

4. Make Wood Sword: Requirements: Nearby table; wood in inventory.
   - This requirement is partly met since the agent has wood in their inventory. However, they must place a table first before crafting the wood sword.

5. Sleep: Requirements: Energy level is below maximum.
   - This requirement is met since the agent's energy level is currently 7/9, which is below maximum.

Question: Did the last player action succeed? Why?
Answer:
It's not clear if the last player action "do" succeeded or not, as there is no observable change or result provided in the observations for step 358. The "do" action typically requires the player to face a creature or material and have the necessary tool, but it's unclear if these requirements were met.

Question: Choose the best executable action from above.
Answer:
The best executable action from above is "Move East," as the requirement of flat ground right to the agent is met, and it will help the player navigate towards a water source to restore their thirst level.



============Step: 180, Cumulative Reward: 0.0============
You took action move_east.

You see:
- grass 2 steps to your west
- stone 1 steps to your north
- lava 6 steps to your north-east
- furnace 6 steps to your north-west

You face path at your front (east).

Your status:
- health: 9/9
- food: 3/9
- drink: 1/9
- energy: 7/9

Your inventory:
- wood: 1
- coal: 2
- iron: 1
- wood_pickaxe: 1
- stone_pickaxe: 1
------------ QA ------------
Question: Action
Answer:
Do

Question: List objects in the current observation. For each object, breifly answer what resource it provides and its requirement.
Answer:
1. Grass: Provides no resources, no requirements.
2. Stone: Provides stone, requires using a wood_pickaxe or stone_pickaxe to collect.
3. Path: Provides easier movement, no requirements.
4. Lava: Provides no resources and can be dangerous, avoid getting too close.
5. Furnace: Used for smelting ores like iron, requires placing with stone in inventory.
6. Wood_pickaxe: Used for collecting stone, requires crafting with wood at a table.
7. Stone_pickaxe: Used for collecting iron, requires crafting with wood and stone at a table.

Question: What was the last action taken by the player?
Answer:
The last action taken by the player was move_east.

Question: For each object in the list, are the requirements met for interaction?
Answer:
1. Grass: No requirements for interaction.
2. Stone: Yes, you have a wood_pickaxe and a stone_pickaxe in your inventory.
3. Path: No requirements for interaction.
4. Lava: Be cautious and avoid interaction since it's dangerous.
5. Furnace: Not applicable since it's already placed, but you can use it for smelting ores.
6. Wood_pickaxe: Not applicable as it's already in your inventory.
7. Stone_pickaxe: Not applicable as it's already in your inventory.

Question: List top 3 sub-tasks the player should follow. Indicate their priority out of 5.
Answer:
1. Priority 5: Collect more food and drink to restore your food and drink levels. You currently have food: 3/9 and drink: 1/9.
2. Priority 4: Use the stone_pickaxe to collect the stone 1 step to your north.
3. Priority 3: Smelt the iron in your inventory using the furnace 6 steps to your north-west and the coal in your inventory.

Question: What are the requirements for the top sub-task? What should the player do first?
Answer:
Requirements for the top sub-task (Collect more food and drink to restore your food and drink levels):
- Find a water source (lake or river) to collect drink to restore your thirst.
- Find and chase cows or grow and harvest plants to collect food to restore your hunger.

The player should first find a water source to collect drink since their current drink level is 1/9, which is lower than the food level. To do this, they can move around the environment and look for a lake or river. Once they find a water source, they can collect drink from it to restore their thirst.

Question: List top 5 actions the player should take and the requirement for each action. Choose ONLY from the list of all actions. Indicate their priority out of 5.
Answer:
1. Priority 1: Move (North, South, East, or West): Flat ground in the desired direction.
Requirement: Explore the environment to find water and food sources.

2. Priority 2: Collect Drink: Find a water source (lake or river).
Requirement: Being near a water source and having an empty inventory space. 

3. Priority 3: Chase Cows or Harvest Plant: Find cows or edible plants in the environment.
Requirement: Being near cows or plants and having an empty inventory space.

4. Priority 4: Eat Cow or Eat Plant: Consume food to restore hunger.
Requirement: Having cooked cow meat or harvested plants in the inventory.

5. Priority 5: Sleep: Restore energy.
Requirement: Energy level below the maximum (below 9/9).

Question: For each action in the list, are the requirements met?
Answer:
1. Priority 1: Move (North, South, East, or West): Flat ground in the desired direction.
Requirement: Explore the environment to find water and food sources.
- Yes, the requirements are met. The player can move in any direction with flat ground.

2. Priority 2: Collect Drink: Find a water source (lake or river).
Requirement: Being near a water source and having an empty inventory space.
- No, the requirements are not met. The player has not found a water source yet.

3. Priority 3: Chase Cows or Harvest Plant: Find cows or edible plants in the environment.
Requirement: Being near cows or plants and having an empty inventory space.
- No, the requirements are not met. The player has not encountered cows or plants in the current observation.

4. Priority 4: Eat Cow or Eat Plant: Consume food to restore hunger.
Requirement: Having cooked cow meat or harvested plants in the inventory.
- No, the requirements are not met. The player does not have cooked cow meat or harvested plants in their inventory.

5. Priority 5: Sleep: Restore energy.
Requirement: Energy level below the maximum (below 9/9).
- Yes, the requirements are met. The player's energy level is currently 7/9.

Question: Did the last player action succeed? Why?
Answer:
Yes, the last player action, move_east, succeeded because there was flat ground (a path) to the east of the agent, allowing them to move in that direction.

Question: Choose the best executable action from above.
Answer:
The best executable action from the above list is Priority 1: Move (North, South, East, or West). The player can explore the environment by moving in any direction with flat ground to find water and food sources.



============Step: 181, Cumulative Reward: 0.0============
You took action do.

You see:
- grass 2 steps to your west
- stone 1 steps to your north
- lava 6 steps to your north-east
- furnace 6 steps to your north-west

You face path at your front (east).

Your status:
- health: 9/9
- food: 3/9
- drink: 1/9
- energy: 7/9

Your inventory:
- wood: 1
- coal: 2
- iron: 1
- wood_pickaxe: 1
- stone_pickaxe: 1
------------ QA ------------
Question: Action
Answer:
Move North

Question: List objects in the current observation. For each object, breifly answer what resource it provides and its requirement.
Answer:
1. Grass (2 steps to your west): Resource - can provide saplings if interacted with using the appropriate tool (wood or stone pickaxe); Requirement - wood or stone pickaxe.

2. Stone (1 step to your north): Resource - provides stone material when mined using a pickaxe, which can be used for crafting tools, weapons, and shelters; Requirement - wood or stone pickaxe.

3. Lava (6 steps to your north-east): Resource - could potentially provide light and heat, but dangerous to interact with; Requirement - caution and awareness, as lava is hazardous and may damage the player's health if directly interacted.

4. Furnace (6 steps to your north-west): Resource - essential for smelting iron, which can be used for crafting stronger tools and weapons; Requirement - coal and iron ore to smelt iron.

Question: What was the last action taken by the player?
Answer:
The last action taken by the player was "do".

Question: For each object in the list, are the requirements met for interaction?
Answer:
1. Grass: Yes, the requirements are met for interaction. You have a wood_pickaxe and a stone_pickaxe in your inventory, so you can use either of them to interact with the grass.

2. Stone: Yes, the requirements are met for interaction. You have a wood_pickaxe and a stone_pickaxe in your inventory, which can be used to mine the stone.

3. Lava: The requirement of caution and awareness is met, but direct interaction is not advisable due to the potential risk to your health. It's best to avoid close contact with the lava.

4. Furnace: Yes, the requirements are met for interaction. You have coal and iron in your inventory, which can be used to smelt iron using the furnace.

Question: List top 3 sub-tasks the player should follow. Indicate their priority out of 5.
Answer:
1. Drink water to restore thirst (Priority: 5/5) - Your drink level is currently 1/9, which is very low. Finding a water source and drinking from it should be your highest priority to avoid dehydration and maintain your health.

2. Find food and eat to increase food level (Priority: 4/5) - Your food level is at 3/9, which is low. To maintain your health and energy levels, you should find food sources like plants, fruits, or animals and consume them.

3. Smelt iron using the furnace (Priority: 3/5) - You have the required resources (coal and iron) and can locate the furnace (6 steps to the north-west). Smelting iron will allow you to craft improved tools and weapons, facilitating your survival in the game.

Question: What are the requirements for the top sub-task? What should the player do first?
Answer:
Requirements for the top sub-task (Drink water to restore thirst):
1. Find a water source, such as a lake or river.
2. Approach the water source.
3. Interact with the water source to drink from it.

What should the player do first:
The player should first find a water source. They can do this by exploring their surroundings, looking for a lake or river, or listening for the sound of flowing water. Once they have found a water source, they can proceed to approach and interact with it to drink water and restore their thirst level.

Question: List top 5 actions the player should take and the requirement for each action. Choose ONLY from the list of all actions. Indicate their priority out of 5.
Answer:
1. Priority 1: Move North (requirement: flat ground above the agent) - The player should move north because there's stone nearby, which they may need later for crafting or building, and to possibly find a water source.

2. Priority 2: Collect Stone (requirement: Make Wood Pickaxe, facing the stone) - Collect stone to use for crafting tools, weapons, and building structures.

3. Priority 3: Find Water Source (requirement: explore surroundings) - The player's drink level is low, so they need to find a water source to replenish their thirst.

4. Priority 4: Move to Water Source (requirement: flat ground in the direction of the water source) - Once the water source is found, move toward it to get closer.

5. Priority 5: Drink from Water Source (requirement: be adjacent to the water source) - Interact with the water source to restore the player's thirst level.

Question: For each action in the list, are the requirements met?
Answer:
1. Priority 1: Move North - Yes, the requirement "flat ground above the agent" is met in the observation since there is a stone 1 step to the north.

2. Priority 2: Collect Stone - Yes, the requirements "Make Wood Pickaxe" and "facing the stone" are met. The player has a wood_pickaxe and a stone_pickaxe in their inventory, and there is a stone 1 step to the north.

3. Priority 3: Find Water Source - This action does not have a specific requirement mentioned in the actions list. However, it is a necessary step for the player since their drink level is low. The player needs to explore further to find a water source.

4. Priority 4: Move to Water Source - This action's requirement, "flat ground in the direction of the water source," cannot be confirmed based on the player's recent in-game observation. The water source location needs to be determined first (through Priority 3).

5. Priority 5: Drink from Water Source - The requirement "be adjacent to the water source" cannot be determined from the recent in-game observation since we do not know the location of the water source. This action is dependent on the completion of Priority 3 and 4.

Question: Did the last player action succeed? Why?
Answer:
It is not clear from the given information whether the last action "do" succeeded or not, because there is no information on what the player was attempting to do or any indication of success or failure in the observations.

Question: Choose the best executable action from above.
Answer:
The best executable action from the list is Priority 1: Move North, as the requirement "flat ground above the agent" is met, and it brings the player closer to the stone, which can be useful for crafting and building.



============Step: 182, Cumulative Reward: 0.0============
You took action move_north.

You see:
- grass 2 steps to your west
- path 1 steps to your west
- lava 6 steps to your north-east
- furnace 6 steps to your north-west
- zombie 6 steps to your north-west

You face stone at your front (north).

Your status:
- health: 9/9
- food: 3/9
- drink: 1/9
- energy: 7/9

Your inventory:
- wood: 1
- coal: 2
- iron: 1
- wood_pickaxe: 1
- stone_pickaxe: 1
------------ QA ------------
Question: Action
Answer:
Move West

Question: List objects in the current observation. For each object, breifly answer what resource it provides and its requirement.
Answer:
1. Grass: Provides saplings when collected, requires no specific tool for collection.
2. Stone: Provides stone when collected, requires a wood_pickaxe or a better pickaxe for collection.
3. Path: Provides better movement, no resources collected from it, and requires no specific tool for interaction.
4. Lava: Dangerous, causes damage when in contact, provides no resources, and does not require any tools.
5. Furnace: Provides a means to smelt iron and other resources, requires stone to craft.
6. Zombie: Dangerous, attacks player, defeating it provides experience points, requires a weapon (wood sword, stone sword, or other) for better chance of survival.
7. Wood: Resource used for crafting, requires no specific tool to collect but an axe can speed up the process.
8. Coal: Resource used for powering furnaces, requires a wood_pickaxe or a better pickaxe for collection.
9. Iron: Resource used for crafting high-quality tools, requires a stone_pickaxe or a better pickaxe for collection.
10. Wood_pickaxe: Tool for collecting stone, coal, and iron. Requires wood to craft.
11. Stone_pickaxe: Tool for collecting iron, requires stone and wood to craft.

Question: What was the last action taken by the player?
Answer:
The last action taken by the player was "move_north."

Question: For each object in the list, are the requirements met for interaction?
Answer:
1. Grass: Yes, the requirements are met for interaction as no specific tool is needed for collection.
2. Stone: Yes, the requirements are met since the player has a wood_pickaxe and a stone_pickaxe in their inventory.
3. Path: Yes, there are no specific requirements for interaction with the path.
4. Lava: Interaction with lava is not recommended due to its dangerous nature, but there are no specific requirements.
5. Furnace: Yes, the player can interact with the furnace, as it is already placed.
6. Zombie: Yes, the player can interact with the zombie; however, they don't have a weapon in their inventory. A weapon (such as a wood or stone sword) would increase their chances of survival.
7. Wood: Yes, the player can collect wood, as no specific tool is required, although having an axe would speed up the process.
8. Coal: Yes, the requirements are met for interaction with coal, as the player has both a wood_pickaxe and a stone_pickaxe in their inventory.
9. Iron: Yes, the requirements are met for interaction with iron, because the player has a stone_pickaxe in their inventory.
10. Wood_pickaxe: The player already has a wood_pickaxe, and there's no need for additional interaction.
11. Stone_pickaxe: The player already has a stone_pickaxe, and there's no need for additional interaction.

Question: List top 3 sub-tasks the player should follow. Indicate their priority out of 5.
Answer:
1. Priority 5: Find a water source to drink and replenish the drink status, as it is currently at 1/9.
2. Priority 4: Collect more food, such as by hunting animals or finding plant sources to increase food status, currently at 3/9.
3. Priority 3: Craft a weapon, such as a wood or stone sword, to defend against the zombie and other potential hostile creatures.

Question: What are the requirements for the top sub-task? What should the player do first?
Answer:
Requirements for the top sub-task (Find a water source):
1. The player needs to locate a water source nearby, such as a lake or river.

What should the player do first:
The player should explore their surroundings by walking around and looking for a water source, such as moving east, west, north, or south. Keep an eye on the Player Observation for visual signs of water. When a water source is found, the player should approach it and take a drink to replenish their thirst status.

Question: List top 5 actions the player should take and the requirement for each action. Choose ONLY from the list of all actions. Indicate their priority out of 5.
Answer:
1. Priority 1: Move West (to approach the water source)
Requirement: Flat ground left to the agent.

2. Priority 2: Move East (to explore surroundings and search for a water source)
Requirement: Flat ground right to the agent.

3. Priority 3: Move North (to explore surroundings and search for a water source)
Requirement: Flat ground above of the agent.

4. Priority 4: Move South (to explore surroundings and search for a water source)
Requirement: Flat ground below the agent.

5. Priority 5: Place Furnace (to smelt iron and progress in the game)
Requirement: Stone in inventory.

Question: For each action in the list, are the requirements met?
Answer:
1. Priority 1: Move West (to approach the water source)
Requirement: Flat ground left to the agent.
Yes, the requirement is met since there is grass 2 steps to the west.

2. Priority 2: Move East (to explore surroundings and search for a water source)
Requirement: Flat ground right to the agent.
Yes, the requirement is met since there is a path at the front (east).

3. Priority 3: Move North (to explore surroundings and search for a water source)
Requirement: Flat ground above of the agent.
Yes, the requirement is met since there is stone at the front (north).

4. Priority 4: Move South (to explore surroundings and search for a water source)
Requirement: Flat ground below the agent.
Information not provided in the observations, cannot determine if the requirement is met.

5. Priority 5: Place Furnace (to smelt iron and progress in the game)
Requirement: Stone in inventory.
No, the requirement is not met since there is no stone in the inventory.

Question: Did the last player action succeed? Why?
Answer:
Yes, the last player action "move_north" succeeded because there was a flat ground and no obstacle blocking the player's path to the north, allowing the player to move in that direction.

Question: Choose the best executable action from above.
Answer:
The best executable action from above is Priority 1: Move West (to approach the water source). The requirement of flat ground to the west is met, as there is grass 2 steps to the west.



============Step: 183, Cumulative Reward: 0.0============
You took action move_west.

You see:
- stone 2 steps to your north-east
- path 1 steps to your south
- lava 7 steps to your north-east
- furnace 5 steps to your north-west
- zombie 4 steps to your north-west

You face grass at your front (west).

Your status:
- health: 9/9
- food: 3/9
- drink: 1/9
- energy: 7/9

Your inventory:
- wood: 1
- coal: 2
- iron: 1
- wood_pickaxe: 1
- stone_pickaxe: 1
------------ QA ------------
Question: Action
Answer:
Move East

Question: List objects in the current observation. For each object, breifly answer what resource it provides and its requirement.
Answer:
1. Stone: Provides stone resource, requires a wooden pickaxe or better to collect.
2. Path: Does not provide a resource, no requirement.
3. Lava: Can be used as a fuel or create obsidian, requires a bucket to collect.
4. Furnace: Provides the ability to smelt ores (e.g., iron) and cook food, no requirement to use but requires stone and a crafting table to craft.
5. Zombie: Can drop rotten flesh when defeated, which can be eaten but may cause food poisoning, requires a weapon to defeat (sword or equivalent).
6. Grass: Provides seeds when broken, which can be planted to grow crops, requires breaking (no specific tool needed).

For the player's inventory:
1. Wood: Required for crafting tools and objects, such as pickaxes and tables; can be collected by breaking wooden blocks.
2. Coal: Used as fuel in furnaces, requires a wooden pickaxe or better to collect.
3. Iron: Required for crafting iron tools and weapons and can be used to craft armor, requires a stone pickaxe or better to collect.
4. Wood_pickaxe: A basic tool to mine stone and coal, crafted from wood at a crafting table.
5. Stone_pickaxe: A tool to mine iron and other materials, crafted from stone at a crafting table.

Question: What was the last action taken by the player?
Answer:
The last action taken by the player was "move_west."

Question: For each object in the list, are the requirements met for interaction?
Answer:
Based on the player's inventory and observation:

1. Stone: Yes, the requirements are met for interaction. The player has a wood_pickaxe and a stone_pickaxe that can be used to collect stone.
2. Path: Yes, the requirements for interaction are met. The path is simply walkable terrain and doesn't require any specific action or tool.
3. Lava: No, the requirements are not met for interaction. The player needs a bucket to collect lava.
4. Furnace: Yes, the requirements are met for interaction. The player can use coal and iron in their inventory to smelt iron in the furnace.
5. Zombie: Yes, the requirements are met for interaction. Although the player doesn't have a sword, they can still attack the zombie using their available tools (wood_pickaxe or stone_pickaxe).
6. Grass: Yes, the requirements are met for interaction. The player can break grass without any specific tool needed.

For the player's inventory:
1. Wood: N/A, as it's an inventory item, not an object to interact with in the environment.
2. Coal: N/A, as it's an inventory item, not an object to interact with in the environment.
3. Iron: N/A, as it's an inventory item, not an object to interact with in the environment.
4. Wood_pickaxe: N/A, as it's an inventory item, not an object to interact with in the environment.
5. Stone_pickaxe: N/A, as it's an inventory item, not an object to interact with in the environment.

Question: List top 3 sub-tasks the player should follow. Indicate their priority out of 5.
Answer:
1. Priority 4/5: Find a water source (drink) and consume water to increase the player's drink level, which is currently at 1/9. This is important to avoid dehydration and maintain survival.

2. Priority 3/5: Craft a weapon, ideally a sword (wood or stone), to increase the effectiveness of combat against the nearby zombie and any future threats. This will improve the player's chances of survival against hostile mobs.

3. Priority 2/5: Use the furnace to smelt the iron in the player's inventory with the available coal. This will allow the player to craft iron tools or weapons, which are more efficient and powerful than their wooden or stone counterparts. This will increase the player's overall productivity and capability to defend themselves.

Question: What are the requirements for the top sub-task? What should the player do first?
Answer:
Requirements for the top sub-task (finding and consuming water):
1. Locate a water source such as a lake, river, or pond in the game environment.
2. Ensure there are no immediate hazards around the water source to avoid any trouble.
3. Move towards the water source safely, navigating through any obstacles or terrain.

The first thing the player should do is look around the environment for any signs of a water source. This can be achieved by exploring their surroundings and keeping an eye on the landscape for any bodies of water nearby. Once located, make sure the area is safe and then approach the water source to consume water and increase the drink level.

Question: List top 5 actions the player should take and the requirement for each action. Choose ONLY from the list of all actions. Indicate their priority out of 5.
Answer:
1. Priority 1 - Move East: Requirement: Flat ground right to the agent.
   Reason: To approach the stone seen two steps to the north-east.

2. Priority 2 - Make Wood Sword: Requirement: Nearby table; wood in inventory.
   Reason: To have a weapon to defend against the zombie seen four steps to the north-west.

3. Priority 3 - Place Furnace: Requirement: Stone in inventory.
   Reason: To establish a furnace to smelt iron for making an iron sword or pickaxe.

4. Priority 4 - Collect Drink: Requirement: No requirements.
   Reason: To replenish the player's drink level, which is currently low (1/9).

5. Priority 5 - Eat Cow: Requirement: No requirements.
   Reason: To replenish the player's food level, which is currently low (3/9).

Question: For each action in the list, are the requirements met?
Answer:
1. Move East: Requirement: Flat ground right to the agent.
   Requirement met? Yes, because the agent is currently facing grass in the west.

2. Make Wood Sword: Requirement: Nearby table; wood in inventory.
   Requirement met? No, because there is no table nearby in the player's observation.

3. Place Furnace: Requirement: Stone in inventory.
   Requirement met? No, because the player does not have any stone in their inventory.

4. Collect Drink: Requirement: No requirements.
   Requirement met? Yes, because there are no requirements for this action.

5. Eat Cow: Requirement: No requirements.
   Requirement met? Yes, because there are no requirements for this action.

Question: Did the last player action succeed? Why?
Answer:
Yes, the last player action "move_west" succeeded because the player was able to move in the west direction and observed their surroundings in the new position, including the grass at their front. There must have been flat ground to the west, allowing the player to move in that direction.

Question: Choose the best executable action from above.
Answer:
The best executable action from above is "Move East", as it fulfills the requirement of having flat ground right to the agent.



============Step: 184, Cumulative Reward: 0.0============
You took action move_east.

You see:
- grass 2 steps to your west
- stone 1 steps to your north
- lava 6 steps to your north-east
- furnace 6 steps to your north-west
- zombie 6 steps to your north-west

You face path at your front (east).

Your status:
- health: 9/9
- food: 3/9
- drink: 1/9
- energy: 7/9

Your inventory:
- wood: 1
- coal: 2
- iron: 1
- wood_pickaxe: 1
- stone_pickaxe: 1
------------ QA ------------
Question: Action
Answer:
Move South

Question: List objects in the current observation. For each object, breifly answer what resource it provides and its requirement.
Answer:
1. Grass: Provides no specific resource, but can be used to identify biomes and areas. No requirements.
2. Stone: Provides stone resource, which can be used to craft tools and build structures. Requires a wood or stone pickaxe to collect.
3. Lava: Provides light and can be used as an energy source for smelting, but is dangerous to interact with. Requires a bucket to collect safely.
4. Furnace: Allows smelting of iron and cooking food. Requires stone to craft and build.
5. Zombie: Can drop experience points, rotten flesh, and sometimes equipment when defeated. Requires a weapon (wooden sword or better) to defeat efficiently.
6. Path: Provides no specific resource but indicates a path between locations. No requirements.
7. Wood: Not mentioned in the current observation, but it's useful for crafting and building structures. Requires a wood or stone pickaxe to collect.

Question: What was the last action taken by the player?
Answer:
The last action taken by the player was move_east.

Question: For each object in the list, are the requirements met for interaction?
Answer:
1. Grass: No requirements for interaction, so the requirements are met.
2. Stone: Requires a wood or stone pickaxe to collect. The player has both a wood_pickaxe and a stone_pickaxe in their inventory, so the requirements are met.
3. Lava: Requires a bucket to collect safely. The player does not have a bucket in their inventory, so the requirements are not met.
4. Furnace: Already built and placed, so no requirements for interaction. If the player wants to smelt iron, they have coal and iron in their inventory, so in this case, the requirements are met.
5. Zombie: Requires a weapon (wooden sword or better) to defeat efficiently. The player does not have a sword in their inventory, so the requirements are not met. However, the player can still attempt to attack the zombie with their tools, but it will be less efficient.
6. Path: No requirements for interaction, so the requirements are met.
7. Wood: Requires a wood or stone pickaxe to collect. The player has both a wood_pickaxe and a stone_pickaxe in their inventory, so the requirements are met.

Question: List top 3 sub-tasks the player should follow. Indicate their priority out of 5.
Answer:
1. Priority 4/5: Craft a weapon (wood, stone, or iron sword) to defend against zombies or other monsters more efficiently. The player needs wood for a wood sword, and they already have wood and stone for a stone sword.

2. Priority 5/5: Find a water source and drink to improve the player's thirst level, which is currently at 1/9. 

3. Priority 3/5: Collect more wood and stone for crafting and building purposes. The player already has a wooden and stone pickaxe, so they can collect these resources efficiently.

Question: What are the requirements for the top sub-task? What should the player do first?
Answer:
For the top sub-task, which is to craft a weapon (wood or stone sword), the requirements are as follows:

Wood Sword:
- Nearby crafting table
- Wood in the inventory (already available)

Stone Sword:
- Nearby crafting table
- Wood (already available) and Stone in the inventory (already available)

The player should first locate a crafting table or craft one using the available wood. Once the crafting table is found or placed, they can use it to craft either a wood sword or a stone sword, depending on their preference. Since the player has both wood and stone, it is recommended to craft the more effective stone sword for better defense against monsters.

Question: List top 5 actions the player should take and the requirement for each action. Choose ONLY from the list of all actions. Indicate their priority out of 5.
Answer:
1. Priority 1: Place Table (Requirement: Wood in inventory) - Place a crafting table with the available wood, so that the player can craft weapons and tools.

2. Priority 2: Make Stone Sword (Requirement: Nearby table; wood, stone in inventory) - Craft a stone sword using the placed crafting table and available resources, so that the player has a weapon to defend against monsters.

3. Priority 3: Move South (Requirement: Flat ground below the agent) - Move towards the path, which could potentially lead to more resources or water.

4. Priority 4: Collect Drink (Requirement: No requirements) - Since the player's drink status is low, they should focus on finding a water source to replenish their thirst.

5. Priority 5: Sleep (Requirement: Energy level is below maximum) - Sleep when the energy becomes lower to ensure the player recovers some health and avoids exhaustion. Ensure to sleep in a safe place, away from monsters.

Question: For each action in the list, are the requirements met?
Answer:
1. Place Table: Yes, the requirements are met, as there is wood (1 unit) in the inventory.

2. Make Stone Sword: No, the requirements are not met as there is no stone in the inventory.

3. Move South: Yes, the requirements are met because there is a flat ground towards the south, as mentioned "path 1 step to your south" in the observation.

4. Collect Drink: No, the requirements are not yet met as there is no visible water source in the player's observation. They need to find a water source before drinking it.

5. Sleep: Yes, the requirements are met, as the energy level (7/9) is currently below the maximum. However, it is recommended to find a safe place before sleeping.

Question: Did the last player action succeed? Why?
Answer:
Yes, the last player action (move_east) succeeded. The action was successful because there was a flat ground (path) in the east direction, allowing the player to move in that direction.

Question: Choose the best executable action from above.
Answer:
The best executable action from the given options is "Move South," as it meets the requirement of having flat ground below the agent, and moving towards the path could potentially lead to more resources or a water source.



============Step: 185, Cumulative Reward: 0.0============
You took action move_south.

You see:
- grass 2 steps to your west
- stone 1 steps to your east
- table 4 steps to your south-west
- furnace 7 steps to your north-west
- zombie 6 steps to your north-west

You face path at your front (south).

Your status:
- health: 9/9
- food: 3/9
- drink: 1/9
- energy: 7/9

Your inventory:
- wood: 1
- coal: 2
- iron: 1
- wood_pickaxe: 1
- stone_pickaxe: 1
------------ QA ------------
Question: Action
Answer:
Move South

Question: List objects in the current observation. For each object, breifly answer what resource it provides and its requirement.
Answer:
1. Grass: Provides no resources but can be used to plant saplings. No requirements.
2. Stone: Provides stone resource when mined with a pickaxe. Requires stone_pickaxe or wood_pickaxe to mine.
3. Table: Allows you to craft tools and weapons. Requires wood to craft and place.
4. Furnace: Smelts iron and other ores. Requires stone to craft and place.
5. Zombie: Fighting zombies can provide experience points and potentially drop rare items. Requires a sword or other weapon to fight.

In your current observation, you do not have the necessary resources to interact with lava or the zombie directly.

Question: What was the last action taken by the player?
Answer:
The last action taken by the player was move_south.

Question: For each object in the list, are the requirements met for interaction?
Answer:
1. Grass: Yes, the requirements are met for interaction as there are no specific requirements or tools needed to interact with grass.
2. Stone: Yes, the requirements are met for interaction as you have a stone_pickaxe and a wood_pickaxe in your inventory, which can be used to mine stone.
3. Table: Yes, the requirements are met for interaction as you have wood in your inventory, which can be used to craft and place a table.
4. Furnace: Yes, the requirements are met for interaction as you have stone in your inventory, which is needed to craft and place a furnace.
5. Zombie: No, the requirements are not met for interaction as you do not have a sword or any other weapon in your inventory to fight the zombie.

Question: List top 3 sub-tasks the player should follow. Indicate their priority out of 5.
Answer:
1. Priority 5/5: Craft a weapon (wood sword or stone sword) to defend against the zombie and other potential threats. The table is found four steps to your south-west. Requires a table and wood, or table, wood, and stone in inventory.

2. Priority 4/5: Drink water to replenish the water level. Requires finding a water source like a lake or river.

3. Priority 3/5: Find and consume food to restore the food level. This could be done by hunting animals or finding edible plants. Requires finding food sources nearby.

Question: What are the requirements for the top sub-task? What should the player do first?
Answer:
Requirements for the top sub-task (Craft a weapon):

1. Have wood, and/or stone in the inventory. (Player already has 1 wood and 1 stone_pickaxe)
2. Be near the crafting table. (Table found 4 steps to your south-west)

What should the player do first:

1. The player should move towards the crafting table so that they can craft a weapon. In this case, the table is positioned 4 steps to the south-west.

Question: List top 5 actions the player should take and the requirement for each action. Choose ONLY from the list of all actions. Indicate their priority out of 5.
Answer:
1. Priority 1: Move West
   Requirement: Flat ground to the west of the agent. (Grass observed 2 steps to the west)

2. Priority 2: Move South
   Requirement: Flat ground to the south of the agent. (Player is facing a path to the south)

3. Priority 3: Move West
   Requirement: Flat ground to the west of the agent. (Ensure alignment with the crafting table)

4. Priority 4: Move South
   Requirement: Flat ground to the south of the agent. (Table observed 4 steps to your south-west)

5. Priority 5: Make Wood Sword
   Requirement: Nearby table; wood in inventory. (Player has 1 wood in inventory and will be at the table)

Question: For each action in the list, are the requirements met?
Answer:
1. Priority 1: Move West
   Requirement: Flat ground to the west of the agent.
   Requirements Met: Yes (Grass observed 2 steps to the west)

2. Priority 2: Move South
   Requirement: Flat ground to the south of the agent.
   Requirements Met: Yes (Player is facing a path to the south)

3. Priority 3: Move West
   Requirement: Flat ground to the west of the agent.
   Requirements Met: Cannot be confirmed at this step, as the player has not moved to the new location yet.

4. Priority 4: Move South
   Requirement: Flat ground to the south of the agent.
   Requirements Met: Cannot be confirmed at this step, as the player has not moved to the new location yet.

5. Priority 5: Make Wood Sword
   Requirement: Nearby table; wood in inventory.
   Requirements Met: Not yet, as the player needs to reach the table first. Once the player is near the table, the requirement will be met since the player has 1 wood in their inventory.

Question: Did the last player action succeed? Why?
Answer:
Yes, the last player action (move_south) succeeded because the player faced a path at their front (south) and there were no obstacles blocking their movement in that direction.

Question: Choose the best executable action from above.
Answer:
The best executable action from the list above is Priority 2: Move South, as its requirement (flat ground to the south of the agent) is met. The player is facing a path to the south, which indicates that the action can be executed successfully.



============Step: 186, Cumulative Reward: 0.0============
You took action move_south.

You see:
- grass 2 steps to your west
- stone 2 steps to your north-east
- table 3 steps to your south-west
- zombie 6 steps to your north-west

You face path at your front (south).

Your status:
- health: 9/9
- food: 3/9
- drink: 1/9
- energy: 7/9

Your inventory:
- wood: 1
- coal: 2
- iron: 1
- wood_pickaxe: 1
- stone_pickaxe: 1
------------ QA ------------
Question: Action
Answer:
Move West

Question: List objects in the current observation. For each object, breifly answer what resource it provides and its requirement.
Answer:
1. Grass: Resource it provides is saplings when harvested, and its requirement is to use your hand or any tool to collect it.
2. Stone: Resource it provides is stone pieces when mined with a pickaxe, and its requirement is to use a pickaxe (wood, stone or iron) to collect it.
3. Table: It doesn't provide a resource, but it is required as a crafting station for making tools and items.
4. Zombie: It provides Zombie drops (such as rotten flesh) when defeated which can be used as a low-quality food source, and its requirement is to use any weapon (wooden, stone, or iron sword) to defeat it.
5. Path: It doesn't directly provide a resource, but it indicates where you can walk without obstacles.

Your inventory contains:
1. Wood: Resource it provides is used in crafting tools, weapons, and structures, no requirements to use it; collect it from trees.
2. Coal: Resource it provides is used as fuel for a furnace to smelt ores, no requirements to use it; collect it from coal deposits.
3. Iron: Resource it provides is used in crafting higher-tier tools, weapons, and armor, requires a furnace to smelt it into iron bars.
4. Wood Pickaxe: This tool doesn't provide a direct resource but is required to collect stone and coal.
5. Stone Pickaxe: This tool doesn't provide a direct resource but is required to collect iron and more efficiently gather stone and coal.

Question: What was the last action taken by the player?
Answer:
The last action taken by the player was "move_south."

Question: For each object in the list, are the requirements met for interaction?
Answer:
1. Grass: Yes, the requirements to interact with the grass are met, as you can use your hand or any tool to collect it.
2. Stone: Yes, the requirements to interact with the stone are met, as you have both a wood pickaxe and a stone pickaxe in your inventory.
3. Table: Yes, the requirements to interact with the table are met, as it is within a few steps from your current location, and you have enough resources in your inventory for crafting.
4. Zombie: Yes, the requirements to interact with the zombie are met, as you have a weapon in your inventory (wood or stone sword) to defeat it.
5. Path: Yes, the requirements to interact with the path are met, as you can easily move along the path without any obstacles.

Regarding the items in your inventory:
1. Wood: Yes, the requirements to interact with wood are met, as you have it in your inventory and can use it for crafting.
2. Coal: Yes, the requirements to interact with coal are met, as you have it in your inventory and can use it as fuel for a furnace.
3. Iron: Yes, the requirements to interact with iron are met, as you have it in your inventory and a furnace nearby to smelt it into iron bars.
4. Wood Pickaxe: Yes, the requirements to interact with the wood pickaxe are met, as you have it in your inventory and can use it to mine stone and coal.
5. Stone Pickaxe: Yes, the requirements to interact with the stone pickaxe are met, as you have it in your inventory and can use it to mine iron, stone, and coal more efficiently.

Question: List top 3 sub-tasks the player should follow. Indicate their priority out of 5.
Answer:
1. Priority 4/5: Drink water to restore the thirst level: Find a water source such as a lake or river and drink water, as your current drink level is 1/9, which is very low.

2. Priority 3/5: Eat food to restore the food level: Consider consuming any available food or hunting for animals like cows or pigs, as your current food level is 3/9.

3. Priority 5/5: Smelt iron to create better tools and weapons: Utilize the furnace and the coal in your inventory to smelt the iron, which can then be used to create an iron pickaxe or an iron sword.

Question: What are the requirements for the top sub-task? What should the player do first?
Answer:
Top Sub-task: Drink water to restore the thirst level.

Requirements:
1. Locate a water source such as a lake, river, or pond.
2. Move towards the water source.
3. Interact with the water source to drink water and restore your thirst level.

First, the player should try to locate a water source in the game, such as a lake or a river. Keep an eye on nearby surroundings and move towards the direction where a water source might be present. Once the player finds a water source, move close to it and drink the water to restore the thirst level.

Question: List top 5 actions the player should take and the requirement for each action. Choose ONLY from the list of all actions. Indicate their priority out of 5.
Answer:
1. Priority 1: Move West/Move East/Move North/Move South
Requirement: Flat ground in the desired direction
Action: The player should move towards a water source to restore their thirst level. Depending on the direction of a nearby water source, choose the appropriate movement action.

2. Priority 2: Drink Water
Requirement: Near a water source like a lake or river
Action: Once the player reaches the water source, interact with it to drink water and restore the thirst level.

3. Priority 3: Move West/Move East/Move North/Move South
Requirement: Flat ground in the desired direction
Action: After restoring thirst, move towards the table for crafting or finding resources like wood, stone, or iron.

4. Priority 4: Make Iron Sword or Make Iron Pickaxe
Requirement: Nearby table, furnace; wood, coal, iron in inventory.
Action: If the player has enough resources, craft an Iron Sword to defend against monsters or an Iron Pickaxe to collect diamonds.

5. Priority 5: Sleep
Requirement: Energy level is below maximum.
Action: Find a safe place to sleep and restore energy levels. This will help the player perform actions more efficiently and ensure better chances of survival.

Question: For each action in the list, are the requirements met?
Answer:
1. Priority 1: Move West/Move East/Move North/Move South
Requirement: Flat ground in the desired direction
Requirements met? Yes. The player's observation shows there is a flat path open to the south, and the other directions can be attempted if needed.

2. Priority 2: Drink Water
Requirement: Near a water source like a lake or river
Requirements met? No. Current observations don't mention any water source or lake nearby.

3. Priority 3: Move West/Move East/Move North/Move South
Requirement: Flat ground in the desired direction
Requirements met? Yes, the player can move in any direction to reach the table or resources. The table is mentioned 3 steps to the south-west.

4. Priority 4: Make an Iron Sword or Make an Iron Pickaxe
Requirement: Nearby table, furnace; wood, coal, iron in inventory.
Requirements met? Partially. The player has wood, coal, and iron in their inventory, but they need to reach the table and furnace first. The table is mentioned 3 steps to the south-west, and the furnace is mentioned 7 steps to the north-west.

5. Priority 5: Sleep
Requirement: Energy level is below maximum.
Requirements met? Yes. The player's energy level is 7/9, which is below the maximum. The player can find a safe place and sleep to restore energy levels.

Question: Did the last player action succeed? Why?
Answer:
Yes, the last player action "move_south" succeeded because there was a flat path in front of the player to the south, allowing them to move in that direction.

Question: Choose the best executable action from above.
Answer:
The best executable action from the list above is:

Priority 1: Move West/Move East/Move North/Move South
Requirement: Flat ground in the desired direction

Action: The player should move towards a water source to restore their thirst level. In this case, since the water source location is not clear from the observations, the player should move towards the table (south-west) while exploring and searching for the water source.



============Step: 187, Cumulative Reward: 0.0============
You took action move_west.

You see:
- stone 3 steps to your north-east
- path 1 steps to your north
- table 2 steps to your south
- zombie 4 steps to your north-west

You face grass at your front (west).

Your status:
- health: 9/9
- food: 3/9
- drink: 1/9
- energy: 7/9

Your inventory:
- wood: 1
- coal: 2
- iron: 1
- wood_pickaxe: 1
- stone_pickaxe: 1
------------ QA ------------
Question: Action
Answer:
Move West

Question: List objects in the current observation. For each object, breifly answer what resource it provides and its requirement.
Answer:
1. Stone: Provides stone as a resource, which can be used to craft tools and shelter; requires a wood pickaxe or a stronger one to collect.
2. Path: Provides a clear route to walk on without obstacles; no requirements, can simply walk on it.
3. Table: Provides a surface for crafting items such as pickaxes and swords; requires wood to be crafted and placed.
4. Zombie: Represents a potential threat and needs to be defeated if it gets too close; requires a weapon like a wood or better sword for easier defeat.
5. Grass: Can be used as a decorative block, or used to move across without obstructions; no requirements, can simply walk on it.

Your status and inventory are not considered objects within the environment but are essential for describing your current progress and available resources.

Question: What was the last action taken by the player?
Answer:
The last action taken by the player was move_west.

Question: For each object in the list, are the requirements met for interaction?
Answer:
1. Stone: Yes, the requirements are met for interacting with the stone because you have a wood_pickaxe and a stone_pickaxe in your inventory, both of which can be used to collect stone.

2. Path: Yes, the requirements are met for interacting with the path because there are no obstacles, and you can simply walk on it.

3. Table: Yes, the requirements are met for interacting with the table because it is already placed, and you can use it for crafting as long as you have the necessary resources.

4. Zombie: Yes, the requirements are met for interacting with (defeating) the zombie because you have a wood_pickaxe and a stone_pickaxe in your inventory. Although they are not as effective as a sword, they can still be used as weapons to fight against the zombie.

5. Grass: Yes, the requirements are met for interacting with the grass because there are no obstacles, and you can simply walk on it.

Question: List top 3 sub-tasks the player should follow. Indicate their priority out of 5.
Answer:
1. Increase drink level (Priority: 5/5): Your drink level is only at 1/9, which is a critical concern. You should quickly search for a water source like a lake or river to drink from and avoid dehydration.

2. Increase food level (Priority: 4/5): Your food level is currently at 3/9, which is quite low. You should seek out food sources like hunting animals such as cows, or collecting and growing fruits to restore your hunger.

3. Defend against the nearby zombie (Priority: 3/5): The zombie is only 4 steps to your northwest. Keep an eye on its position and be prepared to fight with your wood_pickaxe or stone_pickaxe if it approaches closer. It would also be beneficial to craft a sword for more effective combat if resources allow.

Question: What are the requirements for the top sub-task? What should the player do first?
Answer:
For the top sub-task, which is to increase drink level, the requirements are:

1. Find a water source: Locate a nearby lake, river or other body of water so you can drink from it.

2. Close enough to the water: Be near the water source to interact with it.

The first thing the player should do is to search for a water source in the game environment. They can do this by moving around (using move_west, move_east, move_north, move_south actions) and observing their surroundings to find a nearby lake or river to drink from. Once they have found a water source, the player should approach it and drink from it to restore their thirst.

Question: List top 5 actions the player should take and the requirement for each action. Choose ONLY from the list of all actions. Indicate their priority out of 5.
Answer:
Here are the top 5 actions the player should take, along with their requirements and priority:

1. Priority 1: Move to find a water source (Move West, Move East, Move North, Move South)
   Requirement: Flat ground in the direction of movement.

2. Priority 2: Collect Drink (when facing a water source)
   Requirement: Close enough to the water source.

3. Priority 3: Make Iron Sword (Make Iron Sword)
   Requirement: Nearby table, furnace; wood, coal, and iron in inventory.

4. Priority 4: Eat Plant (Eat Plant)
   Requirement: Be near a plant or have placed a plant earlier.

5. Priority 5: Defeat Zombie (Defeat Zombie)
   Requirement: Facing a zombie; have a weapon equipped (such as a wood sword, stone sword, or iron sword).

The player should first focus on finding a water source to restore their thirst level, and then craft an Iron Sword for better combat efficiency. Next, they can eat a plant to restore some food level, and finally, defeat a nearby zombie using the crafted weapon. Remember that these actions should be carried out in the order of their priority to optimize the player's survival in the game.

Question: For each action in the list, are the requirements met?
Answer:
1. Move to find a water source (Move West, Move East, Move North, Move South)
   Requirement: Flat ground in the direction of movement.
   - Met: Yes (The player can move in any direction where there is flat ground.)

2. Collect Drink (when facing a water source)
   Requirement: Close enough to the water source.
   - Met: No (There is no water source mentioned in the player's recent observation.)

3. Make Iron Sword (Make Iron Sword)
   Requirement: Nearby table, furnace; wood, coal, and iron in inventory.
   - Met: No (The player does not currently have a furnace.)

4. Eat Plant (Eat Plant)
   Requirement: Be near a plant or have placed a plant earlier.
   - Met: No (There is no plant mentioned in the player's recent observation that they can eat.)

5. Defeat Zombie (Defeat Zombie)
   Requirement: Facing a zombie; have a weapon equipped (such as a wood sword, stone sword, or iron sword).
   - Met: Partially (The player has a wood sword and stone sword in their inventory, but they need to face the zombie first before they can defeat it with a weapon.)

Question: Did the last player action succeed? Why?
Answer:
Yes, the last player action succeeded because the player faced a grass block at their front (west), which is a flat and traversable terrain. This allowed the player to move west without any obstacles blocking their path.

Question: Choose the best executable action from above.
Answer:
The best executable action from the list above is:

1. Move to find a water source (Move West, Move East, Move North, Move South)
   Requirement: Flat ground in the direction of movement.

Since the other action requirements are not met, the player should focus on finding a water source to restore their thirst level. They can move in any direction where there is flat ground.



============Step: 188, Cumulative Reward: 0.0============
You took action move_west.

You see:
- stone 4 steps to your north-east
- path 1 steps to your east
- tree 6 steps to your north-west
- table 3 steps to your south-east
- zombie 2 steps to your north

You face grass at your front (west).

Your status:
- health: 9/9
- food: 3/9
- drink: 1/9
- energy: 7/9

Your inventory:
- wood: 1
- coal: 2
- iron: 1
- wood_pickaxe: 1
- stone_pickaxe: 1
------------ QA ------------
Question: Action
Answer:
Move West

Question: List objects in the current observation. For each object, breifly answer what resource it provides and its requirement.
Answer:
- Stone: Provides stone resource; requires a wood_pickaxe or stronger to mine.
- Path: No resource provided, but allows easy movement; no requirement.
- Tree: Provides wood resource; requires a wood_pickaxe or stronger to chop.
- Table: Allows crafting of tools and weapons; no requirement.
- Zombie: Provides a challenge, and potential loot when defeated; requires a weapon for better chances of winning.

Note that grass, which is faced by the player, does not provide any resources and has no requirement to interact with.

Question: What was the last action taken by the player?
Answer:
The last action taken by the player was "move_west" (moving west).

Question: For each object in the list, are the requirements met for interaction?
Answer:
- Stone: Yes, the requirements are met for interaction as the player has both a wood_pickaxe and a stone_pickaxe in their inventory.
- Path: Yes, the requirements are met for interaction as there are no special requirements for moving on paths.
- Tree: Yes, the requirements are met for interaction as the player has both a wood_pickaxe and a stone_pickaxe in their inventory.
- Table: Yes, the requirements are met for interaction as there are no special requirements for using a table.
- Zombie: Yes, the requirements are met for interaction as the player has both a wood_pickaxe and a stone_pickaxe, which can be used as weapons to battle the zombie, although crafting a sword (wood or stone) would be a better option.

Question: List top 3 sub-tasks the player should follow. Indicate their priority out of 5.
Answer:
1. Priority 4/5: Find a source of water or drink to restore the player's low drink level (1/9).
2. Priority 3/5: Find food to restore the player's low food level (3/9).
3. Priority 2/5: Craft a wood or stone sword at the table to better defend against the nearby zombie or other hostile creatures.

Question: What are the requirements for the top sub-task? What should the player do first?
Answer:
Requirements for the top sub-task (Find a source of water or drink to restore the player's low drink level):
1. Locate a water source, such as a lake, river, or water container.

What the player should do first:
1. Move around and explore the environment to find a water source. Keep an eye on the surrounding area for any bodies of water or items that can contain water. Once a water source is found, approach it and interact with it to restore the drink level.

Question: List top 5 actions the player should take and the requirement for each action. Choose ONLY from the list of all actions. Indicate their priority out of 5.
Answer:
1. Priority 1: Move West/Move East/Move North/Move South - Requirement: Flat ground in the desired direction. (Explore the environment to find a water source and keep an eye on surroundings).

2. Priority 2: Collect Drink - Requirement: Find a water source such as a lake or river (Restore drink level).

3. Priority 3: Collect Wood - Requirement: No requirements. (Additional wood can be used to craft necessary tools or create a shelter).

4. Priority 4: Make Wood Sword - Requirement: Nearby table; wood in inventory. (Equip a better weapon to defend against the nearby zombie).

5. Priority 5: Defeat Zombie - Requirement: No requirements. (Eliminate the threat of the nearby zombie to ensure player's survival).

Question: For each action in the list, are the requirements met?
Answer:
1. Priority 1: Move West/Move East/Move North/Move South - Requirement: Flat ground in the desired direction. (Requirements met for all directions, as there are no mentioned obstacles in the environment preventing movement).

2. Priority 2: Collect Drink - Requirement: Find a water source such as a lake or river. (Requirements not met, as no water source has been observed yet).

3. Priority 3: Collect Wood - Requirement: No requirements. (Requirements met, as there is a tree 6 steps to the north-west).

4. Priority 4: Make Wood Sword - Requirement: Nearby table; wood in inventory. (Requirements met, as the player has a table 3 steps to the south-east and 1 wood in their inventory).

5. Priority 5: Defeat Zombie - Requirement: No requirements. (Requirements met, as the player can face and engage the zombie without needing any specific items or conditions).

Question: Did the last player action succeed? Why?
Answer:
Yes, the last player action succeeded. The player was able to move west because there was flat ground (grass) in the west direction, allowing the player to move to that position without any obstacles.

Question: Choose the best executable action from above.
Answer:
The best executable action from the list above is Priority 1: Move West/Move East/Move North/Move South. The requirements are met for all directions, as there are no mentioned obstacles in the environment preventing movement. It allows the player to explore the environment to find a water source and keep an eye on surroundings.



============Step: 189, Cumulative Reward: 0.0============
You took action move_west.

You see:
- stone 5 steps to your north-east
- path 2 steps to your east
- tree 5 steps to your north-west
- table 4 steps to your south-east
- cow 7 steps to your south-west
- zombie 2 steps to your north

You face grass at your front (west).

Your status:
- health: 9/9
- food: 3/9
- drink: 1/9
- energy: 7/9

Your inventory:
- wood: 1
- coal: 2
- iron: 1
- wood_pickaxe: 1
- stone_pickaxe: 1
------------ QA ------------
Question: Action
Answer:
Move East

Question: List objects in the current observation. For each object, breifly answer what resource it provides and its requirement.
Answer:
1. Stone: Resource provided is stone; Requirement is a wood_pickaxe or better to mine it.
2. Path: Resource provided is easier movement; No requirements.
3. Tree: Resource provided is wood; Requirement is any type of pickaxe or an axe to cut it down.
4. Table: Resource provided is crafting station; No requirements for interaction.
5. Cow: Resource provided is food (meat); Requirement is a weapon like a sword or tool to kill it.
6. Zombie: Resource provided is a potential drop of random items upon defeat; Requirement is a weapon like a sword to fight it.
7. Grass: Resource provided is a sapling when "do" action is used on it; No requirements for interaction.

Question: What was the last action taken by the player?
Answer:
The last action taken by the player was "move_west".

Question: For each object in the list, are the requirements met for interaction?
Answer:
1. Stone: Yes, the requirements are met as you have both a wood_pickaxe and a stone_pickaxe in your inventory.
2. Path: Yes, the requirements are met as there are no specific requirements for interacting with a path.
3. Tree: Yes, the requirements are met as you have both a wood_pickaxe and a stone_pickaxe in your inventory, which can be used to cut down the tree.
4. Table: Yes, the requirements are met as there are no specific requirements for interacting with a table.
5. Cow: Yes, the requirements are met as you have a stone_pickaxe in your inventory, which can be used as a weapon to kill the cow.
6. Zombie: Yes, the requirements are met as you have a stone_pickaxe in your inventory, which can be used as a weapon to fight the zombie.
7. Grass: Yes, the requirements are met as there are no specific requirements for interacting with grass.

Question: List top 3 sub-tasks the player should follow. Indicate their priority out of 5.
Answer:
1. Find and drink water (Priority: 5/5) - Your current drink level is at 1/9, placing you at high risk of dehydration. Locate a water source such as a lake, river, or well to replenish your drink level.

2. Obtain food (Priority: 4/5) - Your current food level is at 3/9, indicating that you need food to sustain your health and energy. You can kill the nearby cow for meat or search for other food sources like fruits, crops, or saplings for plant-based options.

3. Defeat or avoid the nearby zombie (Priority: 3/5) - The zombie is 2 steps to your north and poses a threat to your safety. Depending on your preference or strategy, you can either use your stone_pickaxe to defeat the zombie or alter your path to avoid confrontation.

Question: What are the requirements for the top sub-task? What should the player do first?
Answer:
Top sub-task: Find and drink water

Requirements: Locate a water source such as a lake, river, or well and be close enough to interact with the water.

What should the player do first:
The player should move in a direction that may lead to a water source. Exploring the area and keeping an eye out for signs of water, such as a riverbank or the sound of flowing water, will be helpful. Once a water source is located, the player should stand close to it and take the action to drink water to replenish the drink level.

Question: List top 5 actions the player should take and the requirement for each action. Choose ONLY from the list of all actions. Indicate their priority out of 5.
Answer:
1. Priority 1: Move East (or other directions) - Requirement: Flat ground in the desired direction (East in this case)
   Explanation: The player should move around to find a water source and avoid the zombie.

2. Priority 2: Noop - Requirement: Always applicable
   Explanation: If moving to another direction is not possible, the player can choose to stay in place momentarily until a better option presents itself.

3. Priority 3: Drink water - Requirement: Close to a water source (lake or river)
   Explanation: Once the player finds a water source, drinking water will be essential to replenish the drink level.

4. Priority 4: Move towards the cow - Requirement: Flat ground in the direction of cow
   Explanation: The player needs to replenish the food level, and chasing a cow can help achieve this objective.

5. Priority 5: Eat cow - Requirement: Close to a cow
   Explanation: After getting close to a cow, the player should choose to eat the cow to replenish the food level.

Question: For each action in the list, are the requirements met?
Answer:
1. Move East: Yes, the requirement is met - the player can move back towards the path because flat ground (path) is observed 2 steps to the east.

2. Noop: Yes, the requirement is met - since 'Noop' is always applicable, the player can choose to stay in place momentarily.

3. Drink water: No, the requirement is not met - the player has not found a water source (lake or river) at this moment.

4. Move towards the cow: Yes, the requirement is met - moving towards the cow requires flat ground in the direction of the cow, and the cow is observed 7 steps to the southwest. The player can move south and/or west to approach the cow.

5. Eat cow: No, the requirement is not met - the player is not close enough to a cow, as it is located 7 steps to the south-west.

Question: Did the last player action succeed? Why?
Answer:
Yes, the last player action succeeded. The player was able to move west because there was flat ground (grass) in that direction, allowing the player to move without any obstacles.

Question: Choose the best executable action from above.
Answer:
The best executable action for the player is to "Move East" since it fulfills the requirement of having flat ground (path) in the desired direction, and it helps the player to move back towards the path and potentially avoid the zombie.



============Step: 190, Cumulative Reward: 0.0============
You took action move_east.

You see:
- grass 1 steps to your west
- stone 4 steps to your north-east
- tree 6 steps to your north-west
- table 3 steps to your south-east
- zombie 2 steps to your north

You face path at your front (east).

Your status:
- health: 9/9
- food: 3/9
- drink: 1/9
- energy: 7/9

Your inventory:
- wood: 1
- coal: 2
- iron: 1
- wood_pickaxe: 1
- stone_pickaxe: 1
------------ QA ------------
Question: Action
Answer:
Move South

Question: List objects in the current observation. For each object, breifly answer what resource it provides and its requirement.
Answer:
1. Grass: Provides no resources. No requirements.
2. Stone: Provides stone resources required for crafting tools and building structures. Requires a wood pickaxe or better to collect.
3. Path: Provides no resources, but makes travel easier. No requirements.
4. Tree: Provides wood resources for crafting tools, weapons, and building structures. Requires a wood pickaxe or better to collect.
5. Table: Used for crafting tools and weapons. Requires wood to craft.
6. Zombie: Provides experience points and possibly rare drops. Requires weapons, such as a wood or stone sword, to defeat.

Question: What was the last action taken by the player?
Answer:
The last action taken by the player was "move_east."

Question: For each object in the list, are the requirements met for interaction?
Answer:
1. Grass: No requirements for interaction, so yes, the requirements are met.
2. Stone: You have a wood_pickaxe and a stone_pickaxe in your inventory, so yes, the requirements for collecting stone are met.
3. Path: No requirements for interaction, so yes, the requirements are met.
4. Tree: You have a wood_pickaxe in your inventory, so yes, the requirements for collecting wood are met.
5. Table: You already have a table in your vicinity, so yes, the requirements for crafting with the table are met.
6. Zombie: You don't have a sword in your inventory, so no, the requirements for defeating the zombie are not met. You need to craft either a wood, stone, or iron sword to interact with the zombie effectively.

Question: List top 3 sub-tasks the player should follow. Indicate their priority out of 5.
Answer:
1. Craft a sword to defend against the zombie (Priority: 5/5). Since there's a zombie nearby, crafting a sword should be the highest priority for self-defense.

2. Collect wood from the tree to have enough resources for crafting and building (Priority: 4/5). Wood is essential for various crafting recipes and building structures, so it's important to gather more.

3. Increase drink level by finding a water source (Priority: 4/5). Your drink level is 1/9, which is quite low. Finding a water source to restore your thirst level is crucial for survival.

4. Find food and increase your food level (Priority: 3/5). Your food level is also quite low (3/9), although not as critical as your drink level. Finding and consuming food should be done after addressing the more pressing priorities.

Question: What are the requirements for the top sub-task? What should the player do first?
Answer:
The top sub-task is to craft a sword to defend against the zombie. The requirements for crafting a sword (either wood, stone, or iron) are as follows:

1. Be near the crafting table (there's a table 3 steps to your south-east).
2. Have the necessary resources in your inventory.

The wood sword requires wood in your inventory. The stone sword requires wood and stone in your inventory. The iron sword requires wood, coal, and iron in your inventory.

Since you already have wood (1), coal (2), and iron (1) in your inventory, you can make an iron sword for better defense.

First, move 3 steps south-east to reach the crafting table. Then, craft the iron sword using the wood, coal, and iron that you already have in your inventory.

Question: List top 5 actions the player should take and the requirement for each action. Choose ONLY from the list of all actions. Indicate their priority out of 5.
Answer:
1. Priority 1: Move South-East (3 steps) - Requirement: Flat ground in the desired direction (south-east).
   Reason: Moving towards the crafting table to create a sword.

2. Priority 2: Make Iron Sword - Requirements: Nearby table, furnace; wood, coal, and iron in inventory.
   Reason: To be able to defend against the zombie.

3. Priority 3: Move North (2 steps) - Requirement: Flat ground in the desired direction (north).
   Reason: Engage with the zombie once the iron sword is crafted.

4. Priority 4: Defeat Zombie - Requirements: No specific requirement.
   Reason: To eliminate the zombie threat and survive the encounter.

5. Priority 5: Drink water - Requirement: Access to a water source (lake, river, or any other water source).
   Reason: To replenish the player's thirst, which is currently at 1/9.

Question: For each action in the list, are the requirements met?
Answer:
1. Move South-East (3 steps): Yes, the requirements are met. The ground in the southeast direction is not explicitly mentioned, but considering the path and the table, it should be flat ground.

2. Make Iron Sword: No, the requirements are not met. Although the player has iron and a nearby table, they also need to be near a furnace. The furnace's location is not given in the recent observations.

3. Move North (2 steps): Yes, the requirements are met. Moving north twice brings the player closer to the zombie and will be a valid action.

4. Defeat Zombie: No, the requirements are not met yet. While there are no specific inventory requirements for this action, it is recommended to have a sword first (ideally the iron sword) to successfully defeat the zombie. Since the player hasn't crafted an iron sword yet, it would be better to wait until they do.

5. Drink water: No, the requirements are not met. Currently, there is no mention of a water source in the recent observations. The player needs to find a water source (lake, river, etc.) to drink from it.

Question: Did the last player action succeed? Why?
Answer:
Yes, the last player action "move_east" succeeded. The player was able to move to the east because there was a path (flat ground) in that direction, allowing the player to move without any obstacles.

Question: Choose the best executable action from above.
Answer:
The best executable action from above is to "Move South-East (3 steps)" considering that it meets the requirements of flat ground in the desired direction, and it brings the player closer to the crafting table.



============Step: 191, Cumulative Reward: 0.0============
You took action move_south.

You see:
- stone 4 steps to your south-east
- path 2 steps to your north-east
- tree 4 steps to your south-east
- table 2 steps to your south-east
- zombie 3 steps to your north

You face grass at your front (south).

Your status:
- health: 9/9
- food: 3/9
- drink: 1/9
- energy: 7/9

Your inventory:
- wood: 1
- coal: 2
- iron: 1
- wood_pickaxe: 1
- stone_pickaxe: 1
------------ QA ------------
Question: Action
Answer:
Move West

Question: List objects in the current observation. For each object, breifly answer what resource it provides and its requirement.
Answer:
1. Stone: Provides stone as a resource, requires wood pickaxe or better to collect.
2. Path: No resource provided, requires no requirement.
3. Tree: Provides wood as a resource, requires wood pickaxe or better to collect.
4. Table: Provides a crafting surface, requires wood to craft and place.
5. Zombie: Provides experience or loot (if the game supports it), requires a weapon (wood, stone, or iron sword) to defeat.
6. Grass: No resource provided, requires no requirement.

Question: What was the last action taken by the player?
Answer:
The last action taken by the player was "move_south."

Question: For each object in the list, are the requirements met for interaction?
Answer:
1. Stone: Yes, the requirements for interaction are met since the player has a wood_pickaxe and a stone_pickaxe in their inventory.
2. Path: Yes, the requirements for interaction are met as there are no specific requirements needed to interact with a path.
3. Tree: Yes, the requirements for interaction are met since the player has a wood_pickaxe and a stone_pickaxe in their inventory.
4. Table: Yes, the requirements for interaction are met since the table is already placed and available for crafting. The player also has wood in their inventory which can be used for crafting.
5. Zombie: Yes, the requirements for interaction are met, as the player can use the wood_pickaxe or stone_pickaxe currently in their inventory to defend themselves or craft a weapon to defeat the zombie.
6. Grass: Yes, the requirements for interaction are met as there are no specific requirements needed to interact with grass.

Question: List top 3 sub-tasks the player should follow. Indicate their priority out of 5.
Answer:
1. Increase Drink level: Priority 5. The player's drink level is currently at 1/9, which is critically low. They should locate a water source, such as a lake, and interact with it to replenish their drink level.

2. Increase Food level: Priority 4. The player's food level is at 3/9, which is low. They should focus on finding a source of food, like animals to hunt or plants to gather, to increase their food level.

3. Defend against the Zombie: Priority 3. There is a zombie 3 steps to the north, posing a potential threat. The player should prepare to defend themselves by crafting a weapon or using their current tools, like the wood_pickaxe or stone_pickaxe, to attack the zombie if necessary.

Question: What are the requirements for the top sub-task? What should the player do first?
Answer:
Requirements for the top sub-task (Increase Drink level):

1. Find a water source such as a lake, river, or well.
2. Interact with the water source to drink and replenish the drink level.

What should the player do first:

The player should first try to locate a water source nearby, like a lake or river, by exploring the area around them. They should move in a direction where they might find a water source or remember seeing one in previous steps of the game. Once they have found a water source, they should interact with it to drink and increase their drink level.

Question: List top 5 actions the player should take and the requirement for each action. Choose ONLY from the list of all actions. Indicate their priority out of 5.
Answer:
1. Priority 1: Move West (Requirement: Flat ground left to the agent) - As the player faces grass at their front, moving west brings them closer to the grass, which can help them find a nearby water source.

2. Priority 2: Move North (Requirement: Flat ground above the agent) - Moving north will help the player avoid the zombie and possibly come across water or other resources.

3. Priority 3: Move East (Requirement: Flat ground right to the agent) - If moving west or north doesn't provide any water source or if the player wants to explore further in search of water, moving east is the next option.

4. Priority 4: Move South (Requirement: Flat ground below the agent) - Moving south might provide the player with more exploration options; however, it will bring them closer to the zombie, which increases the risk.

5. Priority 5: Do (Facing creature or material; have necessary tool) - While searching for a water source, the player can interact with materials or creatures they come across, such as mining stone, cutting wood, or fighting monsters. But only if they have the necessary tool and health/energy to perform the action.

Question: For each action in the list, are the requirements met?
Answer:
Based on the player's current observations and status:

1. Priority 1: Move West: Yes, the requirement is met as there is flat ground (path) to the player's left (west).

2. Priority 2: Move North: Yes, the requirement is met as there is flat ground (path) above the player (north).

3. Priority 3: Move East: Yes, the requirement is met as there is flat ground (path) to the player's right (east).

4. Priority 4: Move South: Yes, the requirement is met as there is flat ground (grass) below the player (south).

5. Priority 5: Do: This action requires more context, as it depends on the material or creature the player is facing and the necessary tool they possess. Currently, the player has a wood_pickaxe and a stone_pickaxe in their inventory. If the action involves mining stone or cutting wood, the player meets the requirements. However, if the action involves combat with the zombie, the player does not have a sword and may need to craft one first.

Question: Did the last player action succeed? Why?
Answer:
Yes, the last player action "move_south" succeeded because the player was able to face grass at their front (south) without any obstacles or restrictions in their way.

Question: Choose the best executable action from above.
Answer:
The best executable action from the list, considering the player's current situation, would be Priority 1: Move West. This action meets the requirement of flat ground left to the agent, and it brings the player closer to the grass, which can potentially help them find a nearby water source.



============Step: 192, Cumulative Reward: 0.0============
You took action move_west.

You see:
- stone 5 steps to your south-east
- path 3 steps to your north-east
- tree 5 steps to your south-east
- table 3 steps to your south-east
- cow 5 steps to your south-west
- zombie 3 steps to your north-east

You face grass at your front (west).

Your status:
- health: 9/9
- food: 2/9
- drink: 1/9
- energy: 7/9

Your inventory:
- wood: 1
- coal: 2
- iron: 1
- wood_pickaxe: 1
- stone_pickaxe: 1
------------ QA ------------
Question: Action
Answer:
Move East

Question: List objects in the current observation. For each object, breifly answer what resource it provides and its requirement.
Answer:
1. Stone - Provides: Stone resource; Requirement: Stone pickaxe to collect.
2. Path - Provides: Easier navigation; Requirement: None.
3. Tree - Provides: Wood resource; Requirement: Wood pickaxe to collect.
4. Table - Provides: Crafting station; Requirement: None, place with wood in inventory.
5. Cow - Provides: Food (meat) when defeated; Requirement: Attack with a weapon or tool (e.g., sword, pickaxe).
6. Zombie - Provides: Defense against monsters (use weapons or tools to defeat); Requirement: A weapon or tool (e.g., sword, pickaxe) to defeat the zombie.

Question: What was the last action taken by the player?
Answer:
The last action taken by the player was move_west.

Question: For each object in the list, are the requirements met for interaction?
Answer:
1. Stone - Yes, the requirements are met, as you have a stone_pickaxe in your inventory.
2. Path - Yes, there are no requirements to interact with the path.
3. Tree - Yes, the requirements are met, as you have a wood_pickaxe in your inventory.
4. Table - Yes, there are no requirements to interact with the placed table.
5. Cow - Yes, the requirements are met, as you can use the wood_pickaxe or stone_pickaxe in your inventory to attack the cow.
6. Zombie - Yes, the requirements are met, as you can use the wood_pickaxe or stone_pickaxe in your inventory to defeat the zombie.

Question: List top 3 sub-tasks the player should follow. Indicate their priority out of 5.
Answer:
1. Priority 5/5: Collect Drink - Your current drink level is 1/9, which is low. So, find a lake or water source and drink from it to restore your thirst.

2. Priority 4/5: Eat Cow - Your food level is 2/9, which is quite low, and there's a cow nearby. Attack and eat the cow to restore your food supply.

3. Priority 3/5: Defeat Zombie - There's a zombie nearby that poses a threat. Equip a weapon or tool (wood_pickaxe or stone_pickaxe) to defeat the zombie and ensure your safety.

Question: What are the requirements for the top sub-task? What should the player do first?
Answer:
Requirements for the top sub-task (Collect Drink):
- Locate a water source (like a lake or river)

What the player should do first:
Move towards a water source, which can be done by exploring the surroundings, and take note of the directions of lakes or rivers. Once at the water source, the player should interact with it to collect the drink and restore their thirst level.

Question: List top 5 actions the player should take and the requirement for each action. Choose ONLY from the list of all actions. Indicate their priority out of 5.
Answer:
1. Priority 1: Move East (Move towards the water source to collect drink)
   Requirement: Flat ground right (east) of the agent.

2. Priority 2: Move South (Also move towards the water source to collect drink)
   Requirement: Flat ground below (south) of the agent. 

3. Priority 3: Do (interact with the water source for collecting drink)
   Requirement: Facing the water source (lake or river).

4. Priority 4: Move towards the cow (Move East or South based on the cow's location)
   Requirement: Flat ground in the direction of the cow (east or south) for moving.
   
5. Priority 5: Do (Interact with the cow to acquire food)
   Requirement: Facing the cow; possibly have a weapon (like a stone or iron sword) to defeat the cow and acquire food.

Question: For each action in the list, are the requirements met?
Answer:
1. Move East:
   Requirement: Flat ground right (east) of the agent.
   Status: Requirement met (player can move east)

2. Move South:
   Requirement: Flat ground below (south) of the agent.
   Status: Requirement met (player can move south)

3. Do (interact with the water source for collecting drink):
   Requirement: Facing the water source (lake or river).
   Status: Requirement not met (player currently does not see a water source)

4. Move towards the cow (Move East or South based on the cow's location):
   Requirement: Flat ground in the direction of the cow (east or south) for moving.
   Status: Requirement met (player can move east or south in the direction of the cow)

5. Do (Interact with the cow to acquire food):
   Requirement: Facing the cow; possibly have a weapon (like a stone or iron sword) to defeat the cow and acquire food.
   Status: Requirement not met (player is not facing the cow and lacks a proper weapon)

Question: Did the last player action succeed? Why?
Answer:
Yes, the last player action (move_west) succeeded because there was flat ground (grass) in the direction the player was moving (west), allowing the player to move without any obstruction.

Question: Choose the best executable action from above.
Answer:
The best executable action from the list above is to Move East, as the requirement for flat ground right (east) of the agent is met. This action will also help the player move towards the water source and cow.



============Step: 193, Cumulative Reward: -0.2============
You took action move_east.

You see:
- stone 4 steps to your south-east
- path 2 steps to your north-east
- tree 4 steps to your south-east
- table 2 steps to your south-east
- zombie 1 steps to your north

You face grass at your front (east).

Your status:
- health: 7/9
- food: 2/9
- drink: 1/9
- energy: 7/9

Your inventory:
- wood: 1
- coal: 2
- iron: 1
- wood_pickaxe: 1
- stone_pickaxe: 1
------------ QA ------------
Question: Action
Answer:
Move West

Question: List objects in the current observation. For each object, breifly answer what resource it provides and its requirement.
Answer:
1. Stone: Provides stone resource; requires a wood pickaxe or better to collect.
2. Path: Provides ease of movement; no requirements to interact.
3. Tree: Provides wood resource; no specific requirement to collect.
4. Table: Allows crafting of tools and weapons; requires wood to make.
5. Zombie: A potential threat; combat may be required; requires a weapon to defeat.
6. Grass: A basic terrain; no specific resources or requirements.

Question: What was the last action taken by the player?
Answer:
The last action taken by the player was move_east.

Question: For each object in the list, are the requirements met for interaction?
Answer:
1. Stone: Yes, the requirements are met as you have a wood_pickaxe and a stone_pickaxe in your inventory.
2. Path: Yes, there are no requirements to interact with a path.
3. Tree: Yes, there are no specific requirements to collect wood from a tree.
4. Table: Yes, the requirements are met; you have wood in your inventory to make a table.
5. Zombie: Yes, you can interact with the zombie using a weapon, although you only have a wood_pickaxe and a stone_pickaxe as potential weapons. Crafting a suitable weapon, like a wood sword or stone sword, would be more effective.
6. Grass: Yes, there are no specific requirements to interact with grass.

Question: List top 3 sub-tasks the player should follow. Indicate their priority out of 5.
Answer:
1. Increase Food and Drink levels (Priority: 5/5) - To ensure survival, prioritize finding a source of food (e.g., hunting the nearby cow) and drink (e.g., locating a lake or river to drink from).

2. Craft a suitable weapon (Priority: 4/5) - Crafting a wood sword or a stone sword will help you defend yourself more effectively against the zombie and other potential threats.

3. Collect more resources (Priority: 3/5) - Continue collecting wood, stone, and iron to craft better tools and weapons, and progress further in the game.

Question: What are the requirements for the top sub-task? What should the player do first?
Answer:
Requirements for increasing Food and Drink levels:

1. Find sources of food and drink - Locate a lake, river, or other water source to drink from and locate animals, plants, or other food sources.

2. Craft required tools or weapons (if necessary) - For hunting animals, such as the nearby cow, you might need to craft a weapon like a wood or stone sword. For gathering fruits or plants, you might need to craft tools like a sickle or a hoe.

The player should first focus on finding a water source to increase their drink level, as it is at 1/9. They should also be prepared to hunt the nearby cow to increase their food level, which is at 2/9.

Question: List top 5 actions the player should take and the requirement for each action. Choose ONLY from the list of all actions. Indicate their priority out of 5.
Answer:
1. Priority 1: Move to a water source - Requirement: Locate a lake, river, or other water source nearby to increase the drink level. (Move North, South, East, or West based on observations)

2. Priority 2: Drink from the water source - Requirement: Be adjacent to a water source (Collect Drink)

3. Priority 3: Craft a weapon to hunt animals for food - Requirement: Have sufficient materials (wood or stone) and be near a crafting table. (Make Wood Sword, Make Stone Sword according to existing inventory)

4. Priority 4: Move closer to the cow - Requirement: Locate the cow and move in the appropriate direction to get closer to it. (Move North, South, East, or West)

5. Priority 5: Hunt the cow - Requirement: Be adjacent to the cow and have a weapon ready for use. (Do or Eat Cow, depending on the context provided by the game)

Question: For each action in the list, are the requirements met?
Answer:
1. Move to a water source: The player has not currently observed a water source nearby, so this requirement is not met.

2. Drink from the water source: As there is no water source observed, this requirement is not met.

3. Craft a weapon to hunt animals for food: The player has already crafted a wood_pickaxe and a stone_pickaxe. Although not ideal weapons, they can potentially be used as weapons. The player can also use nearby wood from their inventory to craft a wood sword if they go to a crafting table. So, this requirement is partially met.

4. Move closer to the cow: The player observed a cow 5 steps to the south-west. The player can move west or south to get closer to the cow. This requirement is met.

5. Hunt the cow: Since the player is not adjacent to the cow and needs to move closer to it, the requirement for this action is not currently met.

Question: Did the last player action succeed? Why?
Answer:
Yes, the last player action succeeded. The player was able to move east because there was flat ground (grass) in front of the player in the east direction, allowing them to move in that direction.

Question: Choose the best executable action from above.
Answer:
The best executable action from the list above is to move closer to the cow. The player can move west or south to get closer to the cow since it is observed 5 steps to the south-west.



============Step: 194, Cumulative Reward: 0.0============
You took action move_west.

You see:
- stone 5 steps to your south-east
- path 3 steps to your north-east
- tree 5 steps to your south-east
- table 3 steps to your south-east
- cow 5 steps to your south-west
- zombie 1 steps to your north

You face grass at your front (west).

Your status:
- health: 7/9
- food: 2/9
- drink: 1/9
- energy: 7/9

Your inventory:
- wood: 1
- coal: 2
- iron: 1
- wood_pickaxe: 1
- stone_pickaxe: 1
------------ QA ------------
Question: Action
Answer:
Move West

Question: List objects in the current observation. For each object, breifly answer what resource it provides and its requirement.
Answer:
1. Stone: Provides stone resources; requires a wood_pickaxe or better to collect.
2. Path: Provides easier movement; no requirements for interaction.
3. Tree: Provides wood resources; requires a wood_pickaxe or better to collect.
4. Table: Provides the ability to craft items; requires wood to craft and place.
5. Cow: Provides food when defeated; no requirements for interaction.
6. Zombie: Must be defeated to avoid losing health; requires a weapon, such as a wood_sword or better, for effective combat.
7. Grass: Basic terrain element; no specific resources provided or interaction requirements.

Question: What was the last action taken by the player?
Answer:
The last action taken by the player was "move_west".

Question: For each object in the list, are the requirements met for interaction?
Answer:
1. Stone: Yes, the requirements are met for interaction, as you have a wood_pickaxe and a stone_pickaxe in your inventory.
2. Path: Yes, the requirements are met for interaction, as there are no specific requirements for interacting with a path.
3. Tree: Yes, the requirements are met for interaction, as you have a wood_pickaxe in your inventory.
4. Table: Yes, the requirements are met for interaction, as you have wood in your inventory.
5. Cow: Yes, the requirements are met for interaction, as there are no specific requirements for interacting with a cow.
6. Zombie: Technically, the requirements for interaction can be met without a weapon, but it is not advisable to engage in combat without one. Unfortunately, you do not currently have a weapon in your inventory.
7. Grass: Yes, the requirements are met for interaction, as there are no specific requirements for interacting with grass.

Question: List top 3 sub-tasks the player should follow. Indicate their priority out of 5.
Answer:
1. Find a water source to drink and restore thirst. Priority: 5/5
2. Craft a weapon (wood_sword or better) to defend against the zombie and other potential threats. Priority: 4/5
3. Collect more resources, such as wood and stone, to prepare for crafting and building better tools and items. Priority: 3/5

Question: What are the requirements for the top sub-task? What should the player do first?
Answer:
Requirements for the top sub-task (Find a water source to drink and restore thirst):
1. Locate a water source such as a lake or a river.
2. Approach the water source.

The player should first attempt to locate water in their area or nearby by exploring the environment around them. Once the player finds a water source, they should move towards it to drink and restore their thirst levels.

Question: List top 5 actions the player should take and the requirement for each action. Choose ONLY from the list of all actions. Indicate their priority out of 5.
Answer:
1. Priority 1: Move West (Move closer to the cow visible 5 steps to the south-west) - Requirement: Flat ground left to the agent.
2. Priority 2: Move South (Move towards the cow and eventually kill it for food) - Requirement: Flat ground below the agent.
3. Priority 3: Move East (Locate a water source to drink and restore thirst) - Requirement: Flat ground right to the agent.
4. Priority 4: Make Iron Sword (Craft an Iron Sword to defend against monsters) - Requirements: Nearby table, furnace; wood, coal, iron in inventory.
5. Priority 5: Sleep (Restore energy levels once safe from potential threats) - Requirement: Energy level is below maximum.

Question: For each action in the list, are the requirements met?
Answer:
1. Priority 1: Move West - Yes, the requirements are met. The player has flat ground to their left (west) to move.

2. Priority 2: Move South - Not currently met, since the player needs to move west first to approach the cow. After moving west, the requirement of flat ground below the agent (south) will likely be met.

3. Priority 3: Move East - Yes, the requirements are met. The player has flat ground to their right (east) to move.

4. Priority 4: Make Iron Sword - No, the requirements are partially met. The player is not currently near a table and furnace, and they only have 1 iron instead of the required 2. The player needs to find another iron ore and place a furnace near the table before they can craft an iron sword.

5. Priority 5: Sleep - Yes, the requirements are met. The player's energy level is below maximum (7/9). However, the player should find a safe location before attempting to sleep.

Question: Did the last player action succeed? Why?
Answer:
Yes, the last player action "move_west" succeeded because there was flat ground (grass) in the direction the player wanted to move (west), and there were no obstacles or barriers preventing the movement.

Question: Choose the best executable action from above.
Answer:
The best executable action from the list above is Priority 1: Move West. The requirements are met, as the player has flat ground to their left (west) to move, and it brings them closer to the cow, which will provide food upon killing.



============Step: 195, Cumulative Reward: 0.0============
You took action move_west.

You see:
- stone 6 steps to your south-east
- path 4 steps to your north-east
- tree 5 steps to your north-west
- table 4 steps to your south-east
- cow 4 steps to your south-west
- zombie 1 steps to your east

You face grass at your front (west).

Your status:
- health: 7/9
- food: 2/9
- drink: 1/9
- energy: 7/9

Your inventory:
- wood: 1
- coal: 2
- iron: 1
- wood_pickaxe: 1
- stone_pickaxe: 1
------------ QA ------------
Question: Action
Answer:
Move East

Question: List objects in the current observation. For each object, breifly answer what resource it provides and its requirement.
Answer:
1. Stone: Provides stone for crafting tools and building. Requirement: Use a wood or stone pickaxe to collect.
2. Path: Provides easy movement and navigation. Requirement: None.
3. Tree: Provides wood for crafting and building. Requirement: Use an ax or pickaxe to collect.
4. Table: Provides a surface for crafting tools and weapons. Requirement: Place using wood from inventory.
5. Cow: Provides food to restore health. Requirement: Chase and catch to consume.
6. Zombie: Poses a threat to health and safety; defeat it to clear the area. Requirement: Use a sword or other weapons to defeat.
7. Grass: Can be collected to create paths, animal feed, or other crafting materials. Requirement: None.

Question: What was the last action taken by the player?
Answer:
The last action taken by the player was move_west.

Question: For each object in the list, are the requirements met for interaction?
Answer:
1. Stone: Yes, you have a wood_pickaxe and a stone_pickaxe in your inventory for collecting stone.
2. Path: Yes, no specific requirements needed to interact with the path.
3. Tree: No, you don't have an ax in your inventory to collect wood. However, you can still use the wood_pickaxe or stone_pickaxe as an alternative.
4. Table: Yes, you already have wood in your inventory, so you can place a table if needed.
5. Cow: Yes, you can interact with the cow by chasing it and catching it to consume it as food.
6. Zombie: No, you don't have a sword in your inventory to defeat the zombie. You can still attempt to defeat it using other methods, like running away or using a pickaxe, but it may be less effective.
7. Grass: Yes, there are no specific requirements to interact with grass.

Question: List top 3 sub-tasks the player should follow. Indicate their priority out of 5.
Answer:
1. Restore drink meter (Priority: 5/5): Since the drink meter is 1/9, it is essential to find water to drink from a lake or river to avoid dehydration.

2. Fight or avoid the zombie (Priority: 4/5): The zombie is only one step away, so either defeat the zombie using a pickaxe (as a sword is not available) or try to avoid it by moving in a different direction.

3. Restore food meter (Priority: 4/5): The food meter is at 2/9, so catching and consuming the cow, which is four steps to the south-west, will help to restore the food meter and maintain health.

Question: What are the requirements for the top sub-task? What should the player do first?
Answer:
Top sub-task: Restore drink meter (Priority: 5/5)

Requirements:
- Find a source of water (lake, river, or pond)

What should the player do first:
1. Look for nearby water sources shown in the observations (e.g., a lake, river, or pond).
2. Move towards the water source by navigating accordingly (north, south, east, or west).
3. Once adjacent to the water source, take a drink action to replenish the drink meter.

Question: List top 5 actions the player should take and the requirement for each action. Choose ONLY from the list of all actions. Indicate their priority out of 5.
Answer:
1. Action: Move East (Priority: 5/5)
   Requirement: Flat ground right to the agent
   Reason: To move away from the zombie and head towards the table

2. Action: Drink (Priority: 5/5)
   Requirement: Be adjacent to a water source (lake, river, or pond)
   Reason: To replenish the drink meter

3. Action: Move towards the cow (Priority: 4/5)
   Requirement: Flat ground in the direction of the cow
   Reason: To eat cow and restore food meter

4. Action: Make Iron Sword (Priority: 3/5)
   Requirement: Nearby table, furnace; wood, coal, iron in the inventory
   Reason: To create a stronger weapon for defeating zombies and skeletons in the future

5. Action: Make Iron Pickaxe (Priority: 2/5)
   Requirement: Nearby table, furnace; wood, coal, iron in the inventory
   Reason: To upgrade the pickaxe and collect diamond in the future

Question: For each action in the list, are the requirements met?
Answer:
1. Action: Move East
   Requirement: Flat ground right to the agent
   Are the requirements met? Yes, there is flat ground to the east of the agent.

2. Action: Drink
   Requirement: Be adjacent to a water source (lake, river, or pond)
   Are the requirements met? No, there is no water source mentioned in the current observation.

3. Action: Move towards the cow
   Requirement: Flat ground in the direction of the cow
   Are the requirements met? Yes, there is flat ground between the agent and the cow.

4. Action: Make Iron Sword
   Requirement: Nearby table, furnace; wood, coal, iron in the inventory
   Are the requirements met? No, there is no furnace mentioned in the current observation.

5. Action: Make Iron Pickaxe
   Requirement: Nearby table, furnace; wood, coal, iron in the inventory
   Are the requirements met? No, there is no furnace mentioned in the current observation.

Question: Did the last player action succeed? Why?
Answer:
Yes, the last player action (move_west) succeeded. The player was able to move one step to the west because there was flat ground (grass) in that direction, allowing the player to move without any obstacles.

Question: Choose the best executable action from above.
Answer:
The best executable action from the above list is:

Action: Move East
Requirement: Flat ground right to the agent

This action allows the player to move away from the zombie and head towards the table while having the requirement met with flat ground to the east of the agent.



============Step: 196, Cumulative Reward: 0.0============
You took action move_east.

You see:
- grass 1 steps to your west
- stone 6 steps to your south-east
- path 4 steps to your north-east
- tree 5 steps to your north-west
- table 4 steps to your south-east
- cow 3 steps to your south-west

You face zombie at your front (east).

Your status:
- health: 7/9
- food: 2/9
- drink: 1/9
- energy: 7/9

Your inventory:
- wood: 1
- coal: 2
- iron: 1
- wood_pickaxe: 1
- stone_pickaxe: 1
------------ QA ------------
Question: Action
Answer:
Do

Question: List objects in the current observation. For each object, breifly answer what resource it provides and its requirement.
Answer:
1. Grass (west): Provides no specific resource, can be traversed.
2. Stone (south-east): Provides stone resource when mined with a wood or stone pickaxe.
3. Path (north-east): Provides easier movement, no specific resource.
4. Tree (north-west): Provides wood resource when chopped with an axe or a pickaxe.
5. Table (south-east): Allows crafting of various tools and weapons when nearby; requires wood to craft.
6. Cow (south-west): Can be interacted with to obtain food (Eat Cow) to restore the player's health; requires no specific tool to interact.
7. Zombie (east): A hostile creature that should be defeated to survive; can be attacked using a wood, stone, or iron sword.

Question: What was the last action taken by the player?
Answer:
The last action taken by the player was move_east.

Question: For each object in the list, are the requirements met for interaction?
Answer:
1. Grass (west): Yes, the requirements are met for interaction as you can simply move onto the grass or traverse it without any specific tool or resource.

2. Stone (south-east): Yes, the requirements are met for interaction because you have a wood_pickaxe and stone_pickaxe in your inventory, which can be used to mine the stone.

3. Path (north-east): Yes, the requirements are met for interaction as you can move onto the path without any specific resources or tools.

4. Tree (north-west): Yes, the requirements are met for interaction as you have a wood_pickaxe and stone_pickaxe in your inventory, which can be used to chop the tree and collect wood.

5. Table (south-east): Yes, the requirements are met for interaction as it only requires being nearby and you have wood in your inventory to craft tools and weapons.

6. Cow (south-west): Yes, the requirements are met for interaction as you can interact with the cow to obtain food (Eat Cow) without any specific tools.

7. Zombie (east): Yes, the requirements are met for interaction but since your health is low (7/9), it would be a bit risky. You may use the wood_pickaxe, stone_pickaxe, or craft a wood or stone sword to engage in combat with the zombie. However, it is recommended to restore your health first before engaging the zombie.

Question: List top 3 sub-tasks the player should follow. Indicate their priority out of 5.
Answer:
1. Prioritize finding a source of drink to restore thirst (Priority: 5/5): Your current thirst level is at 1/9, which is critically low. Locate a lake or river to drink from and rehydrate.

2. Obtain and consume food to restore health (Priority: 4/5): Your current health level is at 7/9, and food level is at 2/9. You can interact with the nearby cow (south-west) to obtain food (Eat Cow) and improve your health and food levels.

3. Defeat the zombie to ensure safety (Priority: 3/5): With a zombie close by (east), your safety is at risk. Craft a wood or stone sword using the table (south-east), or use your pickaxes to engage in combat with the zombie. However, replenishing your health and thirst should be prioritized first to increase the chances of survival during the fight.

Question: What are the requirements for the top sub-task? What should the player do first?
Answer:
For the top sub-task of finding a source of drink to restore thirst:

Requirements:
1. Locate a water source like a lake or river.

What the player should do first:
1. Explore your surroundings by moving in different directions (north, south, east, or west) to find a water source. Keep an eye out for any lakes or rivers in your vicinity. Once you locate a water source, move towards it and replenish your thirst (Collect Drink).

Question: List top 5 actions the player should take and the requirement for each action. Choose ONLY from the list of all actions. Indicate their priority out of 5.
Answer:
Priority 1: Move West, East, North or South - Requirement: Flat ground in the direction of movement.
Purpose: Move towards a water source to replenish thirst.

Priority 2: Collect Drink - Requirement: Near a water source (like a lake or river).
Purpose: Restore thirst level.

Priority 3: Move West, East, North or South - Requirement: Flat ground in the direction of movement.
Purpose: Move towards a cow to refuel food supply.

Priority 4: Defeat Zombie - Requirement: Facing the zombie.
Purpose: To protect yourself and avoid taking damage.

Priority 5: Place Table - Requirement: Wood in inventory.
Purpose: To craft additional tools and weapons.

Question: For each action in the list, are the requirements met?
Answer:
For the provided list of top 5 actions:

Priority 1: Move West, East, North or South - Requirement: Flat ground in the direction of movement.
Yes, the requirements are met. You can move West, East, North, or South as long as there's flat ground in the direction you intend to move. In the current observation, flat ground is available in the West direction where you faced grass in step 390, and other directions might also have flat ground.

Priority 2: Collect Drink - Requirement: Near a water source (like a lake or river).
No, the requirements are not met. In the current observation, no water source is mentioned in your surroundings.

Priority 3: Move West, East, North or South - Requirement: Flat ground in the direction of movement.
Yes, the requirements are met. Similar to Priority 1, you can move in any direction as long as there's flat ground. To approach the cow, you can move South or West, as the cow is 3 steps to your South-West.

Priority 4: Defeat Zombie - Requirement: Facing the zombie.
Yes, the requirements are met. You are currently facing a zombie in the East direction, as per step 392.

Priority 5: Place Table - Requirement: Wood in inventory.
Yes, the requirements are met. You have 1 wood in your inventory, which is enough to place a table.

Question: Did the last player action succeed? Why?
Answer:
Yes, the last player action, move_east, succeeded. The action succeeded because there were no obstacles or restrictions preventing the player from moving east. After moving east, the player now faces a zombie at the front (east).

Question: Choose the best executable action from above.
Answer:
The best executable action from the list above is Priority 4: Defeat Zombie, as it has a direct impact on your survival, and its requirement of facing the zombie is currently met.



============Step: 197, Cumulative Reward: 0.0============
You took action do.

You see:
- grass 1 steps to your west
- stone 6 steps to your south-east
- path 4 steps to your north-east
- tree 5 steps to your north-west
- table 4 steps to your south-east
- cow 2 steps to your south-west

You face zombie at your front (east).

Your status:
- health: 7/9
- food: 2/9
- drink: 1/9
- energy: 7/9

Your inventory:
- wood: 1
- coal: 2
- iron: 1
- wood_pickaxe: 1
- stone_pickaxe: 1
------------ QA ------------
Question: Action
Answer:
Do

Question: List objects in the current observation. For each object, breifly answer what resource it provides and its requirement.
Answer:
1. Grass: Provides a flat surface to move on. No special requirements.
2. Stone: Can be mined to gather resources. Requires a pickaxe for successful mining.
3. Path: Provides a flat surface to move on. No special requirements.
4. Tree: Can be chopped down to gather wood. Requires an axe for successful chopping.
5. Table: Enables crafting recipes. Requires wood to be placed.
6. Cow: Provides food when defeated. No special requirements.
7. Zombie: Must be defeated to avoid losing health. Requires weapons for efficient combat.

Question: What was the last action taken by the player?
Answer:
The last action taken by the player was "do."

Question: For each object in the list, are the requirements met for interaction?
Answer:
1. Grass: Yes, the requirements for interaction are met, as you can move on the flat surface.
2. Stone: Yes, the requirements for interaction are met, as you have a wood_pickaxe or stone_pickaxe in your inventory.
3. Path: Yes, the requirements for interaction are met, as you can move on the flat surface.
4. Tree: No, the requirements for interaction are not met, as you need an axe to chop down the tree, but you do not have one in your inventory.
5. Table: Yes, the requirements for interaction are met, as you have wood in your inventory.
6. Cow: Yes, the requirements for interaction are met, as you can interact with the cow without any special requirements.
7. Zombie: Yes, the requirements for interaction are met, as you have a wood_pickaxe in your inventory that can be used as a makeshift weapon. However, having a sword would be more efficient.

Question: List top 3 sub-tasks the player should follow. Indicate their priority out of 5.
Answer:
1. Priority 5: Defeat the zombie to protect your health. Use your wood_pickaxe as a makeshift weapon or craft a sword if possible (wood_sword requires a nearby table and wood in the inventory).

2. Priority 4: Collect food by defeating the cow to replenish the food meter, as it's currently at 2/9.

3. Priority 3: Find a source of water or drink to restore your thirst level, which is currently at 1/9. Drink from a lake or collect water in a container, if available.

Question: What are the requirements for the top sub-task? What should the player do first?
Answer:
Requirements for the top sub-task (Defeat the zombie):
1. A weapon or makeshift weapon, such as a wood_pickaxe, wood_sword, stone_sword or iron_sword. In the player's current inventory, they already have a wood_pickaxe, which can be used as a makeshift weapon.

What should the player do first:
1. Equip the wood_pickaxe if not equipped already (since it's in the inventory).
2. Face the zombie directly.
3. Attack the zombie with the equipped wood_pickaxe to defeat it, and protect your health.

Question: List top 5 actions the player should take and the requirement for each action. Choose ONLY from the list of all actions. Indicate their priority out of 5.
Answer:
Top 5 actions the player should take with their priority:

Priority 1 (Highest):
Action: Do (Attack the Zombie)
Requirements: Face the zombie directly, have wood_pickaxe or other weapon equipped (the player has wood_pickaxe in the inventory)

Priority 2:
Action: Move South
Requirements: Flat ground below the agent. (to get closer to the cow for food)

Priority 3:
Action: Do (Chase the cow)
Requirements: Facing the cow directly, once closer to the cow, chase it to increase food.

Priority 4:
Action: Collect Drink
Requirements: Find a lake or water source to restore thirst.

Priority 5:
Action: Sleep
Requirements: Energy level is below maximum (Player's energy is currently 7/9, so they will need to rest at some point)

Note: The priorities above should be addressed based on the player's current needs, focusing on health, food, drink, and energy management.

Question: For each action in the list, are the requirements met?
Answer:
For each action in the list, let's check if the requirements are met based on the latest player observation:

Priority 1 (Highest):
Action: Do (Attack the Zombie)
Requirements: Face the zombie directly, have wood_pickaxe or other weapon equipped.
Requirements Met: Yes (Player faces the zombie directly and has a wood_pickaxe)

Priority 2:
Action: Move South
Requirements: Flat ground below the agent.
Requirements Met: Cannot confirm based on the observations provided. The southern terrain is not mentioned, so it is unclear if there is flat ground below the agent.

Priority 3:
Action: Do (Chase the cow)
Requirements: Facing the cow directly, once closer to the cow, chase it to increase food.
Requirements Met: Not yet. The player should move closer to the cow by moving south first (if flat ground is available), then face the cow directly and chase it.

Priority 4:
Action: Collect Drink
Requirements: Find a lake or water source to restore thirst.
Requirements Met: No. There is no mention of a water source in the player's observations. The player needs to find a lake or water source to drink from.

Priority 5:
Action: Sleep
Requirements: Energy level is below maximum.
Requirements Met: Yes, the player's energy is 7/9, so they would be able to sleep when necessary.

Question: Did the last player action succeed? Why?
Answer:
It is unclear whether the last player action "do" succeeded or not, as there is no noticeable change in the player's status or environment between the two observations. It could have been an unsuccessful attempt or an action that had no significant impact.

Question: Choose the best executable action from above.
Answer:
The best executable action from the list above is:

Priority 1 (Highest):
Action: Do (Attack the Zombie)
Requirements: Face the zombie directly, have wood_pickaxe or other weapon equipped.
Requirements Met: Yes (Player faces the zombie directly and has a wood_pickaxe)



============Step: 198, Cumulative Reward: 0.0============
You took action do.

You see:
- stone 6 steps to your south-east
- path 4 steps to your north-east
- tree 5 steps to your north-west
- table 4 steps to your south-east
- cow 3 steps to your south-west
- zombie 2 steps to your south-east

You face grass at your front (east).

Your status:
- health: 7/9
- food: 2/9
- drink: 1/9
- energy: 7/9

Your inventory:
- wood: 1
- coal: 2
- iron: 1
- wood_pickaxe: 1
- stone_pickaxe: 1
------------ QA ------------
Question: Action
Answer:
Move West

Question: List objects in the current observation. For each object, breifly answer what resource it provides and its requirement.
Answer:
- Stone: Provides stone as a resource, required for crafting stone tools and weapons, as well as a furnace. Requires a wood pickaxe to collect.
- Path: This is a navigational feature and does not directly provide a resource.
- Tree: Provides wood as a resource, required for crafting wooden items and building structures. Requires no specific tool to collect but using an axe would make the process faster.
- Table: Provides a crafting station for creating tools and weapons. Requires wood to create and place.
- Cow: Provides food when defeated. Requires no specific tool to defeat, but using a weapon, like a sword, will make the process faster.
- Zombie: This enemy could potentially drop items and resources when defeated, such as rotten flesh which can be used as a low-quality food source. Can be defeated with any weapon but using a sword would be more effective.
- Grass: Generally doesn't provide a resource directly but sometimes can drop seeds for planting crops when broken. Requires no specific tool to collect.

Question: What was the last action taken by the player?
Answer:
The last action taken by the player was "do."

Question: For each object in the list, are the requirements met for interaction?
Answer:
- Stone: Yes, the requirements are met for interaction, as the player has a wood_pickaxe in their inventory.
- Path: There are no requirements for interacting with the path since it's a navigational feature.
- Tree: Yes, the requirements are met for interaction. Although the player doesn't have an axe, they can still gather wood without a specific tool.
- Table: The player can interact with the table to craft items since the table is already placed.
- Cow: Yes, the requirements are met for interaction. The player can attempt to defeat the cow even without a weapon, although a weapon would make the process quicker.
- Zombie: Yes, the requirements are met for interaction. The player can attempt to defeat the zombie even without a weapon, although a weapon like a sword would be more effective.
- Grass: Yes, the requirements are met for interaction, as there is no specific tool needed to collect grass.

Question: List top 3 sub-tasks the player should follow. Indicate their priority out of 5.
Answer:
1. Quench thirst (Priority: 5): The player has a drink level of 1/9, so finding water to drink is essential. Look for a water source, like a lake, and drink from it.

2. Increase food level (Priority: 4.5): The player has a food level of 2/9. There are cows nearby which can be defeated to obtain food, or they can look for other sources like plants.

3. Avoid or Defeat Zombie (Priority: 4): A zombie is two steps to the south-east, posing a potential threat. The player can either avoid the zombie by moving away or engage and try to defeat it using the available tools.

Question: What are the requirements for the top sub-task? What should the player do first?
Answer:
Top sub-task: Quench thirst (Priority: 5)

Requirements:
1. Locate a water source (e.g. lake, river) within the game.
2. Move towards the water source.
3. Interact with the water source to drink water.

First, the player should look around the environment to spot a water source. Once found, move towards it and then interact with it to drink water and increase the drink level.

Question: List top 5 actions the player should take and the requirement for each action. Choose ONLY from the list of all actions. Indicate their priority out of 5.
Answer:
1. Action: Move West (Priority: 4)
   Requirement: Flat ground left to the agent.
   Reason: There is grass 1 step to the west, which might lead the player to find water.

2. Action: Move East (Priority: 1)
   Requirement: Flat ground right to the agent.
   Reason: The player is currently facing grass to the east, but it didn't mention any water sources nearby.

3. Action: Move South (Priority: 3)
   Requirement: Flat ground below the agent.
   Reason: Moving south puts the player closer to both the cow and more visible game elements, potentially increasing chances of finding water.

4. Action: Do (Priority: 2)
   Requirement: Facing creature or material; have necessary tool.
   Reason: This action is useful for interacting with the environment, either to collect resources, craft items or fight creatures. However, as the player is currently facing grass without any tool, its priority is lower at the moment.

5. Action: Sleep (Priority: 5)
   Requirement: Energy level is below maximum.
   Reason: The player's energy level is not full, but still at 7/9. Though not urgent, sleeping would restore energy and make it easier to perform other actions in the future.

Question: For each action in the list, are the requirements met?
Answer:
1. Action: Move West
   Requirement: Flat ground left to the agent.
   Are requirements met? Yes, there is flat grass 1 step to the west.

2. Action: Move East
   Requirement: Flat ground right to the agent.
   Are requirements met? Yes, there is flat grass to the east.

3. Action: Move South
   Requirement: Flat ground below the agent.
   Are requirements met? Not specified, as there is no information about the terrain to the south in the provided player observations.

4. Action: Do
   Requirement: Facing creature or material; have necessary tool.
   Are requirements met? No, the player is currently facing grass and doesn't have a tool specifically for grass in their inventory.

5. Action: Sleep
   Requirement: Energy level is below maximum.
   Are requirements met? Yes, the player's energy level is at 7/9, which is below the maximum.

Question: Did the last player action succeed? Why?
Answer:
I cannot determine if the last player action succeeded or not, because there is no information about the outcome of the action "do." Additionally, the observations at steps 394 and 396 do not provide enough context to judge the success of the action.

Question: Choose the best executable action from above.
Answer:
The best executable action from the list is to Move West, as it fulfills the requirement of having flat ground to the west, and it might lead the player to find water sources.



============Step: 199, Cumulative Reward: 0.0============
You took action move_west.

You see:
- path 5 steps to your north-east
- tree 4 steps to your north-west
- table 5 steps to your south-east
- cow 2 steps to your south-west
- zombie 2 steps to your south-east

You face grass at your front (west).

Your status:
- health: 7/9
- food: 2/9
- drink: 0/9
- energy: 6/9

Your inventory:
- wood: 1
- coal: 2
- iron: 1
- wood_pickaxe: 1
- stone_pickaxe: 1
------------ QA ------------
Question: Action
Answer:
Move East

Question: List objects in the current observation. For each object, breifly answer what resource it provides and its requirement.
Answer:
1. Stone: Provides stone resource; requires a wood pickaxe or better to collect.
2. Path: No direct resource; provides easier movement and navigation.
3. Tree: Provides wood resource; requires no specific tool to collect, though an axe speeds up the process.
4. Table: Allows crafting of tools and weapons; requires wood to create.
5. Cow: Provides food (meat) when killed; no specific tool requirement, though a weapon like a sword is helpful.
6. Zombie: A dangerous creature to avoid or defeat; no specific tool requirement but a weapon like a sword is helpful.

Your status indicators:
- Health: Represents how much damage you can take before dying.
- Food: Indicates your hunger level; eat food to replenish.
- Drink: Indicates your thirst level; drink water to replenish.
- Energy: Represents your overall stamina; sleep or rest to replenish.

Your inventory items:
- Wood: A resource used for crafting and building.
- Coal: A resource used for smelting items in a furnace.
- Iron: A resource used for crafting higher-tier tools and weapons.
- Wood Pickaxe: A tool for collecting stone, coal, and iron.
- Stone Pickaxe: A stronger tool for collecting stone, coal, and iron faster.

Question: What was the last action taken by the player?
Answer:
The last action taken by the player was move_west.

Question: For each object in the list, are the requirements met for interaction?
Answer:
1. Stone: Yes, the requirements are met. You have a wood_pickaxe and a stone_pickaxe in your inventory, which allows you to collect stone.
2. Path: Yes, there are no specific requirements to interact with the path, and it is simply used for easier movement and navigation.
3. Tree: Yes, there are no specific tool requirements to collect wood from the tree, but having an axe would make the process faster.
4. Table: Yes, the requirements are met. You have already placed a table, and you can interact with it to craft tools and weapons as needed.
5. Cow: Yes, you can interact with the cow without any specific requirements. However, having a weapon such as a sword would make it easier to kill the cow for food.
6. Zombie: Yes, you can interact with the zombie without any specific requirements. However, having a weapon such as a sword would make it easier to defend yourself and defeat the zombie.

Since your food and drink levels are low, it is crucial to interact with the cow for food and find a water source to drink soon. Also, keep an eye on your energy level and find a safe place to rest when needed.

Question: List top 3 sub-tasks the player should follow. Indicate their priority out of 5.
Answer:
1. Find a water source and drink to replenish thirst level (Priority: 5/5) - Your thirst level is at 0/9, making it an urgent task to find a water source for drinking and prevent negative effects or death from dehydration.

2. Kill the cow for food and eat to replenish hunger level (Priority: 4/5) - Your food level is at 2/9, which is quite low. Killing the cow and eating the meat would help in restoring your hunger level and maintaining good health.

3. Find a safe place to rest and sleep to replenish energy level (Priority: 3/5) - Your energy level is at 6/9, which is still reasonable but will need replenishing eventually. Finding a safe place to rest will help ensure that you regain your energy without interruptions from potential dangers like monsters.

Remember to monitor your health, food, drink, and energy levels throughout the game and to adjust your priorities as needed.

Question: What are the requirements for the top sub-task? What should the player do first?
Answer:
For the top sub-task of finding a water source and drinking to replenish thirst level, the requirements are:

1. Locate a water source such as a lake, river, or pond in the game world.
2. Approach the water source safely, ensuring the path is clear of any monsters or hazards.
3. Interact with the water source to drink and replenish your thirst level.

The player should first focus on locating a water source in their vicinity. Since the player's current observations do not show any nearby water, they should explore the surrounding area by moving in different directions (north, south, east, or west) and observing their environment for any signs of water sources.

Question: List top 5 actions the player should take and the requirement for each action. Choose ONLY from the list of all actions. Indicate their priority out of 5.
Answer:
1. Priority 1: Move (Move East/West/North/South): Flat ground in the chosen direction. The player should explore the surroundings to find a water source, as their drink level is very low.

2. Priority 2: Do: Facing water source. The player must drink water to replenish thirst once the water source is located.

3. Priority 3: Do: Facing cow with wood/stone/iron sword (as the player doesn't possess a weapon currently). The player needs to defeat the cow for food to replenish their food level.

4. Priority 4: Make Wood Sword: Nearby table and wood in inventory. Crafting a wood sword with the wood in the inventory and the table 5 steps south-east will help the player defeat the cow for food.

5. Priority 5: Move closer to the table (Move East/South). Flat ground in the chosen direction. The player needs to approach the table to craft the wood sword.

Question: For each action in the list, are the requirements met?
Answer:
1. Priority 1: Move (Move East/West/North/South): Yes, the requirements are met. There is flat ground in each direction for the player to move.

2. Priority 2: Do: Facing water source: No, the requirements are not met. The player has not found a water source yet.

3. Priority 3: Do: Facing cow with wood/stone/iron sword: No, the requirements are not met. The player does not have a sword to defeat the cow.

4. Priority 4: Make Wood Sword: Yes, the requirements are met. The player has wood in their inventory, and there is a table nearby (5 steps to the southeast).

5. Priority 5: Move closer to the table (Move East/South): Yes, the requirements are met. There is flat ground in each direction for the player to approach the table.

Question: Did the last player action succeed? Why?
Answer:
Yes, the last player action (move_west) succeeded because there was flat ground (grass) to the west, allowing the player to move in that direction.

Question: Choose the best executable action from above.
Answer:
The best executable action from above is Priority 1: Move (Move East/West/North/South). The player should explore the surroundings to find a water source, as their drink level is very low, and the requirements for this action are met.
\end{lstlisting}

\end{document}